\documentclass[twoside,12pt,english,openany]{book}

\usepackage{setspace}
\newenvironment{abstract}{}{}
\usepackage{abstract}
\usepackage{arxiv}
\usepackage[square,numbers]{natbib}

\usepackage[utf8]{inputenc} 
\usepackage[T1]{fontenc}    
\usepackage[english]{babel}
\usepackage{dsfont} %
\usepackage{xcolor}

\usepackage{hyperref}
\hypersetup{
    colorlinks=true,
    linkcolor=black,
    filecolor=magenta,      
    urlcolor=cyan,
    pdfpagemode=FullScreen,
    }    
\usepackage{url}            
\usepackage{booktabs}       
\usepackage{amsfonts}       
\usepackage{nicefrac}       
\usepackage{microtype}      
\usepackage{lipsum}		
\usepackage{graphicx}

\usepackage{doi}
\usepackage{amsmath}
\usepackage{enumitem}
\DeclareMathOperator*{\argmax}{arg\,max}
\DeclareMathOperator*{\argmin}{arg\,min}
\usepackage{pifont}
\usepackage{xparse}
\usepackage{xspace}

\usepackage{wrapfig}
\usepackage{graphicx}

\usepackage{colortbl}
\newcommand{\ra}[1]{\renewcommand{\arraystretch}{#1}}
\usepackage{longtable}

\usepackage{titlesec}
\titleformat{\paragraph}[runin]{\normalsize\bfseries}{\theparagraph}{1em}{}
\titlespacing*{\paragraph}{0pt}{6pt}{4pt}
\titleformat{\subparagraph}[runin]{\normalsize\itshape}{\thesubparagraph}{1em}{}
\titlespacing*{\subparagraph}{0pt}{4pt}{3pt}
\renewcommand\theparagraph{\thesubsubsection.\arabic{paragraph}}
\renewcommand\thesubparagraph{\theparagraph.\arabic{subparagraph}}
\usepackage{tikz}
\usetikzlibrary{trees, tikzmark, shadows, arrows.meta, positioning, calc}

\usepackage[edges]{forest}
\usepackage[normalem]{ulem}
\usepackage[most]{tcolorbox}

\definecolor{ai_blue}{RGB}{158, 195, 231}

\definecolor{forestgreen}{RGB}{34,139,34}
\definecolor{steelred}{RGB}{139,30,30}
\newtcolorbox{quotebox}{
    colback=ai_blue!60, 
    colframe=gray!30,
    coltext=black,
    boxrule=0pt,
    arc=4pt,
    left=10pt,
    right=10pt,
    top=10pt,
    bottom=10pt,
    enhanced,
    drop shadow={black!50!white}
}

\usepackage{multicol}
\usepackage{multirow}
\usepackage{cleveref}

\tikzset{%
    parent/.style =          {align=center,text width=2cm,rounded corners=3pt, line width=0.3mm},
    child/.style =           {align=center,text width=2.3cm,rounded corners=3pt, fill=blue!10,draw=blue!80,line width=0.3mm},
    grandchild/.style =      {align=center,text width=2cm,rounded corners=3pt},
    greatgrandchild/.style = {align=center,text width=1.5cm,rounded corners=3pt},
    greatgrandchild2/.style = {align=center,text width=1.5cm,rounded corners=3pt},    
    referenceblock/.style =  {align=center,text width=1.5cm,rounded corners=2pt},
    pretrain/.style =           {align=center,text width=2.9cm,rounded corners=3pt, fill=blue!10,draw=blue!80,line width=0.3mm},   
    pretrain_work/.style =           {align=center, text width=2.9cm,rounded corners=3pt, fill=blue!10,draw=blue!0,line width=0.3mm},  
    template/.style =           {align=center,text width=2.9cm,rounded corners=3pt, fill=red!10,draw=red!80,line width=0.3mm},   
    template_work/.style =           {align=center,text width=2.9cm,rounded corners=3pt, fill=red!10,draw=red!0,line width=0.3mm},    
    answer/.style =           {align=center,text width=2.9cm,rounded corners=3pt,line width=0.3mm},
    answer_work/.style =           {align=center,text width=2.9cm,rounded corners=3pt,line width=0.3mm},
    multiple/.style =           {align=center,text width=2.9cm,rounded corners=3pt, fill= orange!10,draw= orange!80,line width=0.3mm},   
    multiple_work/.style =           {align=center,text width=2.9cm,rounded corners=3pt, fill= orange!10,draw= orange!0,line width=0.3mm},        
    tuning/.style =           {align=center,text width=2.9cm,rounded corners=3pt, fill= magenta!10,draw= magenta!80,line width=0.3mm},   
    tuning_work/.style =           {align=center,text width=2.9cm,rounded corners=3pt, fill= magenta!10,draw= magenta!0,line width=0.3mm},
    assurance/.style =           {align=center,text width=2.9cm,rounded corners=3pt, fill= green!10,draw= green!80,line width=0.3mm},   
    assurance/.style =           {align=center,text width=2.9cm,rounded corners=3pt, fill= green!10,draw= green!0,line width=0.3mm},
}

\usepackage{color}			

\usepackage{tabularx}
\usepackage{tabularray}

\usepackage{CJKutf8}  

\usepackage{amssymb}
\usepackage{amsthm}
\usepackage{mathtools}

\tcbuselibrary{theorems,breakable}

\newtheoremstyle{definition}
  {6pt}        
  {6pt}        
  {\normalfont} 
  {}           
  {\bfseries}  
  {.}          
  {.5em}       
  {}           

\theoremstyle{definition}
\newtheorem{definition}{Definition}

\definecolor{DefGreen}{HTML}{E8F5E9}
\tcolorboxenvironment{definition}{
  colback=DefGreen, 
  boxrule=0pt,
  boxsep=0pt,
  left=2pt,    
  right=\labelsep,   
  top=12pt,           
  bottom=12pt,        
  oversize=2pt,
  sharp corners,
  before skip=6pt,   
  after skip=6pt,    
  breakable,
}

\definecolor{customgreen}{RGB}{191,255,191} 
\definecolor{customblue}{RGB}{228,254,254} 
\definecolor{customyellow}{RGB}{255,255,191} 
\definecolor{customred}{RGB}{255,191,191} 

\usepackage{xcolor}
\definecolor{marron}{RGB}{60,30,10}
\definecolor{darkblue}{RGB}{0,0,80}
\definecolor{lightblue}{RGB}{80,80,80}
\definecolor{darkgreen}{RGB}{0,80,0}
\definecolor{darkgray}{RGB}{0,80,0}
\definecolor{darkred}{RGB}{80,0,0}
\definecolor{shadecolor}{rgb}{0.97,0.97,0.97}

\definecolor{LightMint}{HTML}{E8F5E9}
\definecolor{DarkMint}{HTML}{E8F5E9}   

\usepackage{lettrine}
\input Acorn.fd
\newcommand*\initfamily{\usefont{U}{Acorn}{xl}{n}}

\usepackage{todonotes}

\usepackage{color}            
\usepackage[T1]{fontenc}
\usepackage{newpxtext}
\usepackage{fancyhdr}

\fancypagestyle{plain}{
  \fancyhf{}
  \fancyfoot[C]{\thepage}
}

\newcommand{\estcab}[1]{\itshape\textcolor{marron}{\nouppercase #1}}
\fancypagestyle{plain}{%
  \fancyhf{}
  \fancyhead[LE]{\thepage}%
  \fancyhead[RE]{\leftmark}%
  \fancyhead[LO]{\rightmark}%
  \fancyhead[RO]{\thepage}%
}

\usepackage{fourier-orns}

\usepackage{graphicx}
\usepackage{wallpaper}
\usepackage{wrapfig}
\usepackage{booktabs}

\title{Advances and Challenges in Foundation Agents \\ \fontsize{11pt}{13pt}\selectfont From Brain-Inspired Intelligence to Evolutionary, Collaborative, and Safe Systems}

\date{} 
\renewcommand{\headeright}{}
\renewcommand{\undertitle}{}

\author{
 \textbf{Bang Liu}$^{2,3,20*} \footnotemark[2]\ \ $, 
 \textbf{Xinfeng Li}$^{4} \footnotemark[1]\ \ $,  
 \textbf{Jiayi Zhang}$^{1,10} \footnotemark[1]\ \ $,
 \textbf{Jinlin Wang}$^{1} \footnotemark[1]\ \ $,
 \textbf{Tanjin He}$^{5} \footnotemark[1]\ \ $,
 \textbf{Sirui Hong}$^{1} \footnotemark[1]\ \ $, \\
 \textbf{Hongzhang Liu}$^{6} \footnotemark[1]\ \ $, 
 \textbf{Shaokun Zhang}$^{7} \footnotemark[1]\ \ $,
 \textbf{Kaitao Song}$^{8} \footnotemark[1]\ \ $,
 \textbf{Kunlun Zhu}$^{9} \footnotemark[1]\ \ $,
 \textbf{Yuheng Cheng}$^{1} \footnotemark[1]\ \ $, \\
 \textbf{Suyuchen Wang}$^{2,3} \footnotemark[1]\ \ $, 
 \textbf{Xiaoqiang Wang}$^{2,3} \footnotemark[1]\ \ $,
 \textbf{Yuyu Luo}$^{10} \footnotemark[1] \ \ $,
 \textbf{Haibo Jin}$^{9} \footnotemark[1] \thanks{Major Contribution.}$,
 \textbf{Peiyan Zhang}$^{10}$,
 \textbf{Ollie Liu}$^{11}$, \\
 \textbf{Jiaqi Chen}$^{1}$, 
 \textbf{Huan Zhang}$^{2,3}$,
 \textbf{Zhaoyang Yu}$^{1}$, 
 \textbf{Haochen Shi}$^{2,3}$, 
 \textbf{Boyan Li}$^{10}$, 
  \textbf{Dekun Wu}$^{2,3}$, 
  \textbf{Fengwei Teng}$^{1}$, \\
  \textbf{Xiaojun Jia}$^{4}$, 
  \textbf{Jiawei Xu}$^{1}$,
  \textbf{Jinyu Xiang}$^{1}$,
  \textbf{Yizhang Lin}$^{1}$, 
  \textbf{Tianming Liu}$^{14}$, 
   \textbf{Tongliang Liu}$^{6}$, \\
  \textbf{Yu Su}$^{15}$,
  \textbf{Huan Sun}$^{15}$,
  \textbf{Glen Berseth}$^{2,3,20}$, 
  \textbf{Jianyun Nie}$^{2}$, 
  \textbf{Ian Foster}$^{5}$,
  \textbf{Logan Ward}$^{5}$, 
  \textbf{Qingyun Wu}$^{7}$, \\
  \textbf{Yu Gu}$^{15}$,
  \textbf{Mingchen Zhuge}$^{16}$,
  \textbf{Xinbing Liang}$^{1}$,
  \textbf{Xiangru Tang}$^{12}$,
  \textbf{Haohan Wang}$^{9}$,
  \textbf{Jiaxuan You}$^{9}$, 
  \textbf{Chi Wang}$^{19}$, \\
  \textbf{Jian Pei}$^{17} \footnotemark[2]\ \ $,
  \textbf{Qiang Yang}$^{10,18} \footnotemark[2]\ \ $,
  \textbf{Xiao-Liang Qi}$^{13} \footnotemark[2]\ \ $,
  \textbf{Chenglin Wu}$^{1*} \footnotemark[2] \thanks{Corresponding\ authors: Bang\ Liu (bang.liu@umontreal.ca),\ Jian\ Pei\ (j.pei@duke.edu),\ Qiang\ Yang\ (qyang@cse.ust.hk),\ Xiaoliang\ Qi\ (xlqi@stanford.edu),\ Chenglin\ Wu\ (alexanderwu@deepwisdom.ai)}$
   \vspace{.5em} 
  \\
  $^1$MetaGPT, 
  $^2$Université de Montréal, 
  $^3$Mila - Quebec AI Institute, 
  $^4$Nanyang Technological University, \\
  $^5$Argonne National Laboratory, 
  $^6$University of Sydney, 
  $^7$Penn State University, 
  $^8$Microsoft Research Asia,  \\
  $^9$University of Illinois at Urbana-Champaign, 
  $^{10}$The Hong Kong University of Science and Technology, \\
  $^{11}$University of Southern California, 
  $^{12}$Yale University, 
  $^{13}$Stanford University, 
  $^{14}$University of Georgia, \\ 
  $^{15}$The Ohio State University, 
  $^{16}$King Abdullah University of Science and Technology,  
  $^{17}$Duke University, \\
  $^{18}$The Hong Kong Polytechnic University,
  $^{19}$Google DeepMind,
  $^{20}$Canada CIFAR AI Chair 
}

\hypersetup{
pdftitle={Advances and Challenges in Foundation Agents},
pdfsubject={AI},
}

\begin{document}

\maketitle

\begin{abstract}
The advent of large language models (LLMs) has catalyzed a transformative shift in artificial intelligence, paving the way for advanced intelligent agents capable of sophisticated reasoning, robust perception, and versatile action across diverse domains. As these agents increasingly drive AI research and practical applications, their design, evaluation, and continuous improvement present intricate, multifaceted challenges. This book provides a comprehensive overview, framing intelligent agents within modular, brain-inspired architectures that integrate principles from cognitive science, neuroscience, and computational research.
We structure our exploration into four interconnected parts. First, we systematically investigate the \textbf{modular foundation of intelligent agents}, systematically mapping their cognitive, perceptual, and operational modules onto analogous human brain functionalities and elucidating core components such as memory, world modeling, reward processing, goal, and emotion. Second, we discuss \textbf{self-enhancement and adaptive evolution mechanisms}, exploring how agents autonomously refine their capabilities, adapt to dynamic environments, and achieve continual learning through automated optimization paradigms. Third, we examine \textbf{collaborative and evolutionary multi-agent systems}, investigating the collective intelligence emerging from agent interactions, cooperation, and societal structures, highlighting parallels to human social dynamics. Finally, we address the critical imperative of \textbf{building safe and beneficial AI systems}, emphasizing intrinsic and extrinsic security threats, ethical alignment, robustness, and practical mitigation strategies necessary for trustworthy real-world deployment.
By synthesizing modular AI architectures with insights from different disciplines, this survey identifies key research gaps, challenges, and opportunities, encouraging innovations that harmonize technological advancement with meaningful societal benefit. The project's Github link is: \url{https://github.com/FoundationAgents/awesome-foundation-agents}.
\end{abstract}


\clearpage
\chapter*{Preface}
Large language models (LLMs) have revolutionized artificial intelligence (AI) by demonstrating unprecedented capabilities in natural language and multimodal understanding, as well as reasoning and generation. These models are trained on vast datasets and exhibit emergent abilities such as reasoning, in-context learning, and even rudimentary planning. While these models represent a major step forward in realizing intelligent machines, they themselves do not yet fully embody all the capabilities of an intelligent being. Since the early days of artificial intelligence, AI researchers have long been on a quest for a truly ``intelligent'' system that can learn, plan, reason, sense, communicate, act, remember, and demonstrate various human-like abilities and agility. These beings, known as intelligent agents, should be able to think both long-term and short-term, perform complex actions, and interact with humans and other agents. LLMs are an important step towards realizing intelligent agents, but we are not there yet.

This book provides a comprehensive overview of the current state of the art of LLM-based intelligent agents. In the past, there have been numerous research papers and books on intelligent agents, as well as a flurry of books on LLMs. However, there has scarcely been comprehensive coverage of both. While LLMs can achieve significant capabilities required by agents, they only provide the foundations upon which further functionalities must be built. For example, while LLMs can help generate plans such as travel plans, they cannot yet generate fully complex plans for complex and professional tasks, nor can they maintain long-term memories without hallucination. Furthermore, their ability to perform real-world actions autonomously remains limited. We can view LLMs as engines, with agents being the cars, boats, and airplanes built using these engines. In this view, we naturally seek to move forward in designing and constructing fully functioning intelligent agents by making full use of the capabilities provided by LLMs.

In this engine-vehicle analogy of the interplay between LLMs and agents, we naturally ask: How much of the capabilities of intelligent agents can current LLM technologies provide? What are the functions that cannot yet be realized based on current LLM technologies? Beyond LLMs, what more needs to be done to have a fully intelligent agent capable of autonomous action and interaction in the physical world? What are the challenges for fully integrated LLM-based agents? What additional developments are required for capable, communicative agents that effectively collaborate with humans? What are the areas that represent low-hanging fruits for LLM-based agents? What implications will there be for society once we have fully intelligent LLM-based agents, and how should we prepare for this future?

These questions transcend not only the engineering practice of extending current LLMs and agents but also raise potential future research directions. We have assembled frontier researchers from AI, spanning from LLM development to agent design, to comprehensively address these questions. The book consists of four parts. The first part presents an exposition of the requirements for individual agents, comparing their capabilities with those of humans, including perception and action abilities. The second part explores agents' evolution capabilities and their implications on intelligent tools such as workflow management systems. The third part discusses societies of agents, emphasizing their collaborative and collective action capabilities, and the fourth part addresses ethical and societal aspects, including agent safety and responsibilities.

This book is intended for researchers, students, policymakers, and practitioners alike. The audience includes non-AI readers curious about AI, LLMs, and agents, as well as individuals interested in future societies where humans co-exist with AI. Readers may range from undergraduate and graduate students to researchers and industry practitioners. The book aims not only to provide answers to readers' questions about AI and agents but also to inspire them to ask new questions. Ultimately, we hope to motivate more people to join our endeavor in exploring this fertile research ground.


\clearpage

\setcounter{tocdepth}{2}
\tableofcontents

\clearpage
\section*{Notation}

Here we summarize the notations used throughout the book for the reader's convenience. Detailed definitions can be found in the reference locations.

\DefTblrTemplate{caption}{default}{}
\DefTblrTemplate{conthead}{default}{}
\DefTblrTemplate{capcont}{default}{}
\definecolor{verylightgray}{gray}{0.95}

\begin{longtblr}[
  caption = {Notation used in this book},
	label = {tab:notation},
	]{
		colspec = {|t{0.15\linewidth}|m{0.6\linewidth}|b{0.15\linewidth}|},
		rowhead = 1,
		hlines,
		row{even} = {LightMint},
	} 
	\textbf{Symbol} & \textbf{Description} & \textbf{Reference}\\

    $\mathcal{W}$ & The world with society systems. & Sec.~\ref{subsec:agent-framework-symbols}\\
    $\mathcal{S}$ & State space of an environment.  & Sec.~\ref{subsec:agent-framework-symbols}\\
    $s_t \in \mathcal{S}$ & Environment's state at time $t$. & Sec.~\ref{subsec:agent-framework-symbols}\\
    $\mathcal{O}$  & Observation space. & Sec.~\ref{subsec:agent-framework-symbols}\\
    $o_t \in \mathcal{O}$ & Observation at time $t$. & Sec.~\ref{subsec:agent-framework-symbols}\\
    $\mathcal{A}$  & Agent's action space. & Sec.~\ref{subsec:agent-framework-symbols}\\
    $a_t \in \mathcal{A}$ & Agent's action output at time $t$. & Sec.~\ref{subsec:agent-framework-symbols}\\
    $\mathcal{M}$ & Mental states space. & Sec.~\ref{subsec:agent-framework-symbols}\\
    $M_t \in \mathcal{M}$ & Agent's mental state at time $t$. & Sec.~\ref{subsec:agent-framework-symbols}\\
    $M_t^{\mathrm{mem}}$  & \textit{Memory} component in $M_t$. & Sec.~\ref{subsec:agent-framework-symbols}\\
    $M_t^{\mathrm{wm}}$  & \textit{World model} component in $M_t$. & Sec.~\ref{subsec:agent-framework-symbols}\\
    $M_t^{\mathrm{emo}}$ & \textit{Emotion} component in $M_t$. & Sec.~\ref{subsec:agent-framework-symbols}\\
    $M_t^{\mathrm{goal}}$ & \textit{Goal} component in $M_t$. & Sec.~\ref{subsec:agent-framework-symbols}\\
    $M_t^{\mathrm{rew}}$  & \textit{Reward/Learning} signals in $M_t$. & Sec.~\ref{subsec:agent-framework-symbols}\\
    $\mathrm{L}$ & Agent's learning function. & Sec.~\ref{subsec:agent-framework-symbols}\\
    $\mathrm{R}$ & Agent's reasoning function. & Sec.~\ref{subsec:agent-framework-symbols}\\
    $\mathrm{C}$ & Agent's cognition function. & Sec.~\ref{subsec:agent-framework-symbols}\\
    $\mathrm{E}$ & Action execution (effectors). & Sec.~\ref{subsec:agent-framework-symbols}\\
    $\mathrm{T}$ & Environment transition. & Sec.~\ref{subsec:agent-framework-symbols}\\


    $\mathrm{P}$           & Prompt text / prompt set fed to the LLM.                                    & Sec.~\ref{sec:prompt-optim} \\
    $K_{\text{wf}}$        & Agentic workflow composed of interconnected nodes and control flows.        & Sec.~\ref{sec:workflow-optimization} \\
    $N$           & Node representing a single LLM invocation used in $K_{\text{wf}}$. & Sec.~\ref{sec:workflow-optimization} \\
    $\theta$ & Parameters of the world model $M_t^{\mathrm{wm}}$. & Sec.~\ref{subsec:KL_measure_intelligence} \\
    $P_\theta$ & Predicted data distribution. & Sec.~\ref{subsec:KL_measure_intelligence} \\
    $P_\mathcal{W}$ & True data distribution in the real world. & Sec.~\ref{subsec:KL_measure_intelligence} \\
    $\mathcal{K}$ & Space of known data and information. & Sec.~\ref{subsec:KL_measure_intelligence} \\
    $\mathcal{U}$ & Space of unknown data and information. & Sec.~\ref{subsec:KL_measure_intelligence} \\
    ${\bf x}$ & Dataset representing scientific knowledge. & Sec.~\ref{subsec:KL_measure_intelligence} \\
    ${\bf x}_{\mathrm{K}}$ & Known dataset sampled from $\mathcal{K}$. & Sec.~\ref{subsec:KL_measure_intelligence} \\
    ${\bf x}_{\mathrm{U}}$ & Unknown dataset sampled from $\mathcal{U}$. & Sec.~\ref{subsec:KL_measure_intelligence} \\
    $D_0$ & KL divergence from $P_\mathcal{W}$ to $P_\theta$ at time $t=0$. & Sec.~\ref{subsec:KL_measure_intelligence} \\
    $D_{\mathrm{K}}$ & KL divergence from $P_\mathcal{W}$ to $P_\theta$ after acquiring knowledge. & Sec.~\ref{subsec:KL_measure_intelligence} \\
    $IQ_t^{\mathrm{agent}}$ & Agent's intelligence at time $t$. & Sec.~\ref{subsec:KL_measure_intelligence} \\
    $\Delta $ & Subspace of $\mathcal{U}$ for knowledge expansion. & Sec.~\ref{subsec:IQ_increase_w_knowledge} \\
    ${\bf x}_\Delta$ & Dataset from $\Delta$. & Sec.~\ref{subsec:IQ_increase_w_knowledge} \\
    $\Theta$ & Space of possible world model parameters $\theta$. & Sec.~\ref{subsec:knowledge_expansion_strategy} \\
    $\theta^*_{\mathrm{K},t}$ & Optimal world model parameters given the agent's knowledge at time $t$. & Sec.~\ref{subsec:knowledge_expansion_strategy} \\
    $D_{\mathrm{K},\Theta}^{\rm min}$ & Minimum unknown given the agent's knowledge and $\Theta$. & Sec.~\ref{subsec:knowledge_expansion_strategy} \\

    $\mathbf{x}_{1:n}$ & Input token sequence. & Sec.~\ref{ssec:brain_sec_threat} \\
    $\mathbf{y}$ & Generated output sequence. & Sec.~\ref{ssec:brain_sec_threat} \\
    
    $p$ & Probability of generating $\mathbf{y}$ given $\mathbf{x}_{1:n}$. & Sec.~\ref{sssec:jailbreak} \\
    $\tilde{\mathbf{x}}_{1:n}$ & Perturbed input sequence. & Sec.~\ref{sssec:jailbreak} \\
    $\mathcal{R}^*$ & Ideal alignment reward (measuring adherence to safety/ethical guidelines). & Sec.~\ref{sssec:jailbreak} \\
    $\mathbf{y}^\star$ & Jailbreak output induced by perturbations. & Sec.~\ref{sssec:jailbreak} \\
    $\mathcal{A}$ & a set of safety/ethical guidelines & Sec.~\ref{sssec:jailbreak} \\
    $\mathcal{T}$ & the distribution or set of possible jailbreak instructions. & Sec.~\ref{sssec:jailbreak} \\
    $\mathcal{L}^{adv}$ & Jailbreak loss. & Sec.~\ref{sssec:jailbreak} \\

    $\mathbf{p}$ & Prompt injected into the original input. & Sec.~\ref{sssec:promptinj} \\
    $\mathbf{x}'$ & Combined (injected) input sequence. & Sec.~\ref{sssec:promptinj} \\
    $\mathcal{L}^{inject}$ & Prompt injection loss. & Sec.~\ref{sssec:promptinj} \\
    $\mathbf{p}^\star$ & Optimal injected prompt minimizing $\mathcal{L}^{inject}$. & Sec.~\ref{sssec:promptinj} \\
    $\mathcal{P}$ & Set of feasible prompt injections. & Sec.~\ref{sssec:promptinj} \\
    $e_{x_i}\in\mathbb{R}^{d_e}$ & Embedding of token $x_i$ in a $d_e$-dimensional space. & Sec.~\ref{sssec:hallucination} \\
    $\mathrm{W}_Q,\mathrm{W}_K,\mathrm{W}_V$ & Projection matrices for query, key, and value. & Sec.~\ref{sssec:hallucination} \\
    $A_{ij}$ & Attention score between tokens $i$ and $j$. & Sec.~\ref{sssec:hallucination} \\
    $o_i$ & Contextual representation of token $i$ (weighted sum result). & Sec.~\ref{sssec:hallucination} \\
    $\delta_{x_i}$ & Perturbation applied to $e_{x_i}$, satisfying $\|\delta_{x_i}\|\leq\epsilon$. & Sec.~\ref{sssec:hallucination} \\
    $\tilde{e}_{x_i}$ & Perturbed token embedding. & Sec.~\ref{sssec:hallucination} \\
    $A_{ij}^\Delta$ & Attention score under perturbation. & Sec.~\ref{sssec:hallucination} \\
    $\tilde{o}_i$ & Updated token representation under perturbation. & Sec.~\ref{sssec:hallucination} \\
    $\mathcal{H}$ & Hallucination metric. & Sec.~\ref{sssec:hallucination} \\
    
    $\mathcal{R}$ & Actual alignment reward of the model's output. & Sec.~\ref{sssec:misalignment} \\
    $\Delta_{\text{align}}$ & Alignment gap. & Sec.~\ref{sssec:misalignment} \\
    $\mathcal{L}^{misalign}$ & Misalignment loss. & Sec.~\ref{sssec:misalignment} \\
    $\lambda$ & Trade-off parameter for the alignment gap in the misalignment loss. & Sec.~\ref{sssec:misalignment} \\

    $\mathcal{D}$ & Clean training dataset. & Sec.~\ref{sssec:poison} \\
    $\tilde{\mathcal{D}}$ & Poisoned training dataset. & Sec.~\ref{sssec:poison} \\
    $\theta$ & Model parameters. & Sec.~\ref{sssec:poison} \\
    $\theta^\star$ & Model parameters learned from the poisoned dataset. & Sec.~\ref{sssec:poison} \\
    $\theta_{\text{clean}}$ & Model parameters obtained using the clean dataset. & Sec.~\ref{sssec:poison} \\
    $\Delta_\theta$ & Deviation of model parameters due to poisoning. & Sec.~\ref{sssec:poison} \\
    $t$ & Backdoor trigger. & Sec.~\ref{sssec:poison} \\
    $\mathcal{B}$ & Backdoor success rate. & Sec.~\ref{sssec:poison} \\
    $\mathbb{I}$ & Indicator function. & Sec.~\ref{sssec:poison} \\
    $\mathcal{Y}_{\text{malicious}}$ & Set of undesirable outputs. & Sec.~\ref{sssec:poison} \\
    
    $g$ & Function estimating the probability that input $\mathbf{x}$ was in the training set, with range $[0,1]$. & Sec.~\ref{ssec:brain_priv_threat} \\
    $\eta$ & Threshold for membership inference. & Sec.~\ref{ssec:brain_priv_threat} \\
    $\mathbf{x}^\star$ & Reconstructed training sample in a data extraction attack. & Sec.~\ref{ssec:brain_priv_threat} \\
    $\mathbf{p}_{sys}$ & System prompt defining the agent's internal guidelines. & Sec.~\ref{ssec:brain_priv_threat} \\
    $\mathbf{p}_{user}$ & User prompt. & Sec.~\ref{ssec:brain_priv_threat} \\
    $\mathbf{p}^\star$ & Reconstructed prompt via inversion. & Sec.~\ref{ssec:brain_priv_threat} \\

\end{longtblr}

\chapter{Introduction}
\label{chap:intro}

\lettrine[lines=3]{\initfamily\textcolor{darkgreen}{A}}{rtificial Intelligence (AI)} has long been driven by humanity's ambition to create entities that mirror and transcend human intelligence. The roots of this fascination trace back to ancient myths and early engineering marvels, which illustrate humanity's enduring dream of creating intelligent, autonomous beings. Stories like that of Talos, the bronze automaton of Crete, describe a giant constructed by the gods to guard the island, capable of patrolling its shores and fending off intruders. Such myths symbolize the desire to imbue artificial creations with human-like agency and purpose. In ancient China, \begin{CJK*}{UTF8}{gbsn}诸葛亮\end{CJK*} (Zhuge Liang) was said to have invented the \begin{CJK*}{UTF8}{gbsn}木牛流马\end{CJK*} (wooden ox and flowing horse)—ingenious self-moving transport devices used for military logistics—demonstrating early imagination of autonomous, functional machines shaped by human intent. Similarly, the mechanical inventions of the Renaissance, including Leonardo da Vinci's humanoid robot (designed to mimic human motion and anatomy) represent the first attempts to translate these myths into tangible, functional artifacts. These early imaginings and prototypes reflect the deep-seated aspiration to bridge imagination and technology, laying the groundwork for the scientific pursuit of machine intelligence, culminating in Alan Turing's seminal 1950 question, ``\textit{Can machines think?}'' \cite{turing2009computing}. To address this, Turing proposed the \textit{Turing Test}, a framework to determine whether machines could exhibit human-like intelligence through conversation, shifting focus from computation to broader notions of intelligence. Over the decades, AI has evolved from symbolic systems reliant on predefined logic to machine learning models capable of learning from data and experience and adapting to new situations. This progression reached a new frontier with the advent of large language models (LLMs), which demonstrate remarkable abilities in understanding, reasoning, and generating human-like text \cite{brown2020language}. Central to these advancements is the concept of \textit{agent}, a system that not only processes information but also perceives its environment, makes decisions, and acts autonomously. Initially a theoretical construct, the agent paradigm has become a cornerstone of modern AI, driving advancements in fields ranging from conversational assistants to embodied robotics as AI systems increasingly tackle dynamic, real-world environments.

\section{The Rise and Development of AI Agents}
\label{sec:agent-origin}

\lettrine[lines=3]{\initfamily\textcolor{darkgreen}{T}}{he} concept of ``agent'' is a cornerstone of modern AI, representing a system that perceives its environment, makes decisions, and takes actions to achieve specific goals. This idea, while formalized in AI in the mid-20th century, has roots in early explorations of autonomy and interaction in intelligent systems. One of the most widely cited definitions, proposed by \citet{russell1995aima}, describes an agent as ``\textit{anything that can be viewed as perceiving its environment through sensors and acting upon that environment through actuators}''. This definition emphasizes the dual nature of agents as both observers and actors, capable of dynamically adapting to their surroundings rather than following static rules. It encapsulates the shift in AI from systems that merely compute to systems that engage with their environment.
The historical development of agents parallels the evolution of AI itself. Early symbolic systems, such as Newell and Simon's General Problem Solver \cite{newell1961gps}, sought to replicate human problem-solving processes by breaking tasks into logical steps. However, these systems were limited by their reliance on structured environments and predefined logic. The agent paradigm emerged as a response to these limitations, focusing on autonomy, adaptability, and real-world interaction. Rodney Brooks's subsumption architecture in the 1980s exemplified a pivotal shift toward behavior-based robotics (BBR), introducing agents capable of real-time, reactive behavior in physical environments \cite{brooks1986robust}. Unlike traditional approaches that relied on constructing detailed internal models of the world, BBR emphasizes systems with minimal internal state, where behavior emerges from direct sensory-motor interactions. These robots exhibit complex-appearing actions by continuously adjusting to their environment, not through deep planning, but through layered and reflexive responses. Brooks's architecture demonstrated that robust, scalable intelligence could arise from simple, modular behaviors operating in parallel, marking a foundational departure from deliberative AI design.
Agents have since become a versatile framework across AI subfields. In robotics, they enable autonomous navigation and manipulation; in software, they form the foundation of multi-agent systems used for simulation and coordination \cite{wooldridge2009introduction}. By integrating perception, reasoning, and action into a cohesive structure, the agent paradigm has consistently served as a bridge between theoretical AI constructs and practical applications, advancing our understanding of how intelligent systems can operate in dynamic and complex environments.

The advent of large language models (LLMs) has redefined the capabilities of agents, transforming their role in artificial intelligence and opening up new horizons for their applications. Agents, once confined to executing narrowly defined tasks or following rigid rule-based frameworks, now leverage the broad generalization, reasoning, and adaptability of models like OpenAI's ChatGPT \cite{openai2023chatgpt}, DeepSeek AI's DeepSeek \cite{liu2024deepseek}, Anthropic's Claude \cite{anthropic2023claude}, Alibaba's QWen \cite{yang2024qwen2}, and Meta's LLaMA \cite{touvron2023llama}. These LLM-powered agents have evolved from static systems into dynamic entities capable of processing natural language, reasoning across complex domains, and adapting to novel situations with remarkable fluency. No longer merely passive processors of input, these agents have become active collaborators, capable of addressing long-horizon challenges and interacting with their environments in a way that mirrors human problem-solving.

A key advancement in the LLM era is the seamless integration of language understanding with actionable capabilities. Modern LLMs, equipped with function-calling APIs, enable agents to identify when external tools or systems are required, reason about their usage, and execute precise actions to achieve specific goals. For instance, an agent powered by ChatGPT can autonomously query a database, retrieve relevant information, and use it to deliver actionable insights, all while maintaining contextual awareness of the broader task. This dynamic combination of abstract reasoning and concrete execution allows agents to bridge the gap between cognitive understanding and real-world action. Furthermore, the generalization abilities of LLMs in few-shot and zero-shot learning have revolutionized the adaptability of agents, enabling them to tackle a diverse array of tasks (from data analysis and creative content generation to real-time collaborative problem-solving) without extensive task-specific training. This adaptability, coupled with their conversational fluency, positions LLM-powered agents as intelligent mediators between humans and machines, seamlessly integrating human intent with machine precision in increasingly complex workflows.

\section{A Parallel Comparison between Human Brain and AI Agents}
\label{sec:parallel-compare-human-ai}

\lettrine[lines=3]{\initfamily\textcolor{darkgreen}{T}}{he} rapid integration of LLMs into intelligent agent architectures has not only propelled artificial intelligence forward but also highlighted fundamental differences between AI systems and human cognition. As illustrated briefly in Table~\ref{tab:brain_llm_comparison}, LLM-powered agents differ significantly from human cognition across dimensions such as underlying ``hardware'', consciousness, learning mechanisms, creativity, and energy efficiency. However, it is important to emphasize that this comparison provides only a high-level snapshot rather than an exhaustive depiction. Human intelligence possesses many nuanced characteristics not captured here, while AI agents also exhibit distinct features beyond this concise comparison.

Human intelligence operates on biological hardware, the brain, that demonstrates extraordinary energy efficiency, enabling lifelong learning, inference, and adaptive decision-making with minimal metabolic costs. In contrast, current AI systems require substantial computational power, resulting in significantly higher energy consumption for comparable cognitive tasks. Recognizing this performance gap emphasizes energy efficiency as a critical frontier for future AI research.

Human learning is continuous, interactive, and context-sensitive, deeply shaped by social, cultural, and experiential factors. Conversely, LLM agents primarily undergo static, offline batch training with limited ongoing adaptation capabilities. Despite research work using instruction tuning and reinforcement learning from human feedback (RLHF) \cite{bai2022training}, LLM agents still fall short of human-like flexibility. Bridging this gap through approaches such as lifelong learning, personalized adaptation, and interactive fine-tuning represents a promising research direction, enabling AI to better mirror human adaptability and responsiveness.

Creativity in humans emerges from a rich interplay of personal experiences, emotional insights, and spontaneous cross-domain associations. Emotional states not only motivate creative expression but also influence the originality, depth, and resonance of the outcomes, imbuing them with personal meaning and affective significance. In contrast, creativity in large language models (LLMs) stems primarily from the statistical recombination of training data (what might be described as ``statistical creativity''). While often fluent and occasionally surprising, this form of creativity lacks emotional grounding, lived experience, and intentional originality. This contrast reveals opportunities for advancing AI agents by incorporating deeper contextual understanding, simulated emotional states, and experiential memory. Such developments could lead to more authentic and emotionally attuned creative processes.

In terms of consciousness and emotional experience, LLM agents lack genuine subjective states and self-awareness inherent to human cognition. Although fully replicating human-like consciousness in AI may not be necessary or even desirable, appreciating the profound role emotions and subjective experiences play in human reasoning, motivation, ethical judgments, and social interactions can guide research toward creating AI that is more aligned, trustworthy, and socially beneficial.

Considering the time scale, the human brain has evolved over millions of years, achieving remarkable efficiency, adaptability, and creativity through natural selection and environmental interactions. In stark contrast, AI agents have undergone rapid yet comparatively brief development over roughly 80 years since the advent of early computational machines. This parallel comparison between human cognition and AI systems is thus highly valuable, as it uncovers fundamental differences and provides meaningful insights that can guide advancements in AI agent technologies. Ultimately, drawing inspiration from human intelligence can enhance AI capabilities, benefiting humanity across diverse applications from healthcare and education to sustainability and beyond.

\begin{table*}[!htb]
\centering
\caption{Concise high-level comparison between human brain and LLM agent.}
\label{tab:brain_llm_comparison}
\begin{tabular}{p{2.1cm} p{4.3cm} p{4.3cm} p{4.2cm}}
\toprule
\textbf{Dimension} & \textbf{Human Brain / Cognition} & \textbf{LLM Agent} & \textbf{Remarks} \\
\midrule
\rowcolor{LightMint}
\textbf{Hardware \& Maintenance} & 
- Biological neurons, neurotransmitters, neuroplasticity. \newline
- Requires sleep, nutrition, rest. \newline
- Limited replication, knowledge transfer via learning. \newline
- Extremely energy-efficient (approx. 20W). &
- Deep neural networks, gradient-based optimization. \newline
- Requires hardware, stable power, and cooling. \newline
- Easily duplicated across servers globally. \newline
- High energy consumption (thousands of watts per GPU server). &
Human brains are biologically maintained, energy-efficient, and not easily replicable. LLM agents rely on hardware maintenance, are highly replicable, but significantly less energy-efficient. \\
\textbf{Learning Style} &
- Lifelong, continuous, online learning. \newline
- Few-shot, rapid knowledge transfer. \newline
- Influenced by environment, culture, emotions. &
- Primarily offline, batch-based training. \newline
- Limited online learning and adaptation. \newline
- Neutral, impersonal learned knowledge. &
Despite improvements via instruction tuning, human learning remains more dynamic, adaptive, and culturally/emotionally integrated than LLM learning. \\
\rowcolor{LightMint}
\textbf{Creativity \& Divergence} & 
- Rooted in personal experience, emotions, subconscious insights. \newline
- Rich cross-domain associations, metaphorical thinking. \newline
- Emotional depth influences creativity. &
- Statistical recombination from extensive data. \newline
- Novelty through probabilistic optimization. \newline
- Limited emotional and experiential grounding. &
LLM creativity is statistical and data-driven; human creativity blends emotion, experience, and subconscious processes. \\
\textbf{Consciousness \& Development} & 
- Genuine subjective experiences, emotions, self-awareness. \newline
- Gradual developmental stages from childhood. \newline
- Emotional cognition drives decision-making. &
- No genuine subjective experience or self-awareness. \newline
- ``Emotions'' are superficial language imitations. \newline
- Static post-training with limited dynamic growth. &
Human consciousness emerges from emotional, social, and biological development; LLMs remain static without true introspection or emotional depth. \\
\bottomrule
\end{tabular}
\end{table*}


\begin{figure}[!htb]
\centering
    \includegraphics[width=1.0\columnwidth]{figures/brain2.pdf}
    \caption{Illustration of key human brain functionalities grouped by major brain regions, annotated according to their current exploration level in AI research. This figure highlights existing achievements, gaps, and potential opportunities for advancing artificial intelligence toward more comprehensive, brain-inspired capabilities. Please note: The location of a label does not imply that it is associated with or is only associated with nearby brain regions. Also, to focus on the current state of AI capabilities rather than the anatomy of the brain, we did not mark the limbic system, subcortical system, hypothalamus, amygdala, and so on in this lateral view of the brain, though these systems play important roles in brain function.}
\label{fig:brain}
\end{figure}

\subsection{Brain Functionalities and AI Parallels}

Designing intelligent agents calls for inspiration from the human brain's functional architecture. A high-level map linking brain regions – frontal, parietal, occipital, temporal lobes, as well as the cerebellum, brainstem, and key subcortical structures – to cognitive functions can reveal gaps between human capabilities and current AI systems, as shown in Figure \ref{fig:brain}. However, brain functions are not siloed into rigid anatomical zones: most abilities emerge from networks spanning multiple regions. For instance, memory involves the hippocampus (temporal lobe) interacting with frontal cortex and other areas, and ``self-awareness'' or consciousness cannot be pinpointed to a single spot. Therefore, it is important to keep in mind that cognition is distributed rather than strictly localized. With that in mind, we review each major brain region's core functions (drawing on \textit{Principles of Neural Science} by Kandel et al. \cite{kandel2013principles}, \textit{Neuroscience} by Purves et al. \cite{purves2018neuroscience}, and other sources) and map them to AI-relevant cognitive capabilities. We then propose a set of functions to feature in the figure – emphasizing those most relevant to AI agents (e.g. reasoning, memory, planning, perception, decision-making, motivation, emotion, motor skills) – along with an assessment of how developed these functions are in AI. 
For a big-picture perspective, we categorize the state of research in AI with three distinct levels:
\begin{itemize}
    \item \textbf{Level 1 (L1):} well-developed.
    \item \textbf{Level 2 (L2):} partially developed.
    \item \textbf{Level 3 (L3):} underexplored.
\end{itemize}
Our goal is to come up with a clear, biologically-grounded illustration and discussion that will engage researchers in AI by highlighting which human cognitive functions are replicated in machines and which remain frontier challenges.

\paragraph*{Frontal Lobes (Executive Functions and Decision-Making)}
The frontal lobes – especially the prefrontal cortex – are the seat of the brain's highest-order cognitive functions known collectively as executive functions \cite{kandel2013principles}. These include abilities such as planning, decision-making, problem-solving, working memory, and inhibitory control (self-control). The frontal lobe is also involved in voluntary motor control (with the rear portion containing the primary motor cortex) and aspects of language (Broca's area in the left frontal lobe handles speech production). From a neuroscience perspective, damage to the prefrontal cortex famously impairs one's judgment, planning, and social behavior (as illustrated by the classic Phineas Gage case) \cite{wiki:PhineasGage}. In the context of AI agents, frontal lobe functions correspond to the core ``thinking'' and control components of an intelligent system:
\begin{itemize}
    \item \textbf{Planning and reasoning:} AI has made progress here, for example with automated planners and logical reasoners, and large language models (LLMs) that can follow multi-step reasoning to some extent. These are partially developed (L2) in current AI. However, human-level flexible planning remains only partly solved.
    \item \textbf{Decision-making:} In humans this involves weighing outcomes, rewards, and risks (frontal cortex often working with basal ganglia) \cite{larry2024organization}. AI agents have decision modules (e.g. reinforcement learning policies, decision trees, or LLMs), but handling open-ended, goal-conflicting decisions with human-like adaptability is still L2 at best. Simple decisions from well-defined rewards (like games) are mastered by AI, but broad autonomous decision-making in the real world remains challenging.
    \item \textbf{Working memory:} Frontal networks (especially dorsolateral prefrontal cortex) can hold and manipulate information in mind (e.g. remembering a phone number or interim result). AI analogs include the context windows of neural networks or explicit memory buffers. While current models have limited memory (e.g. a Transformer's context length or external memory in some architectures), this is an active area, and partial functionality exists. Still, the robust, general working memory humans exhibit (flexibly updating and focusing on relevant info) is not fully realized in AI (some aspects may be underexplored, edging into L3).
    \item \textbf{Cognitive flexibility and inhibitory control:} Frontal lobes allow us to shift strategies or perspectives and to suppress inappropriate impulses. AI systems are typically brittle in this regard – they follow their programming or learned policy rigidly, and struggle with on-the-fly strategy shifts or inhibiting a pre-potent response unless explicitly trained. This remains underexplored (L3). For instance, an AI might exploit a reward loophole (lack of ``inhibitory'' self-regulation) unless designers anticipate and constrain it. Future agent architectures may need a mechanism akin to frontal inhibitory control to moderate behaviors.
\end{itemize}
It's worth noting that social-emotional functions involving the frontal lobe (like empathy, theory of mind, self-reflection) are still rudimentary in AI. Humans rely on frontal cortex (especially medial and orbitofrontal regions) interacting with limbic structures to navigate social situations and emotions.  It is a network-level phenomena rather than confined to frontal lobe alone. AI agents do not yet possess genuine empathy or self-awareness – these functions remain L3 (largely absent). 
In summary, the frontal lobes contribute the supervisory, goal-directed intelligence that we often associate with ``thinking'', and many of these capacities are only partially realized in AI to date.

\paragraph*{Parietal Lobes (Perception Integration and Attention)}
The parietal lobes are key to integrating sensory information from various modalities and constructing a spatial understanding of the world \cite{purves2018neuroscience}. The anterior part of the parietal lobe contains the somatosensory cortex, which processes touch, proprioception (body position), and other somatic senses. The posterior parietal areas are crucial for spatial awareness, visuo-spatial processing, and attention – essentially, knowing where things are and how to interact with them. For example, the parietal lobe helps us localize objects in space, understand geometric relationships, and coordinate eye and hand movements by linking vision with motor plans. It also plays a central role in attention control, particularly the dorsal attention network that directs our focus to locations or sensory features of interest.

In AI terms, parietal lobe functions translate to an agent's ability to perceive and navigate its environment:
\begin{itemize}
	\item \textbf{Multisensory integration:} Robots and AI systems that use multiple sensors (vision, touch, etc.) attempt to combine those inputs into a coherent model of the environment. This is still partially developed (L2) – e.g., we have AI that can align vision with depth sensors or touch, but human-level integration (where a slight brush on the arm, a sound, and a peripheral visual cue all unify into a single event perception) is far from achieved.
	\item \textbf{Spatial representation and mapping:} Parietal circuits create internal maps (for instance, of your surroundings or your body in space). AI has made progress in spatial mapping and navigation (SLAM algorithms for robots build 3D maps, and deep reinforcement learning agents can navigate virtual mazes). This capability is moderately developed – certain tasks like autonomous driving or drone flight show that machines can handle spatial reasoning in constrained scenarios (L2). Yet, they lack the general-purpose, flexible spatial understanding humans have (e.g. understanding a cluttered room's layout at a glance, or mentally rotating objects), so further research is needed.
	\item \textbf{Attention mechanisms:} The way the brain's parietal–frontal circuits spotlight task-relevant information is loosely echoed by the \emph{attention heads} in modern transformer networks \cite{vaswani2017attention} and various attention mechanisms \cite{guo2022attention}.  Humans, however, wield attention as an \emph{active lens}: in a lecture we can \emph{simultaneously} follow the speaker's voice (auditory stream), skim the projected slide (visual stream), and monitor the clock in peripheral vision, then ``zoom in'' on a line of text to decode a formula, or ``zoom out'' to grasp the talk's overall structure.  Neurophysiology shows that such rapid shifts are driven by top-down signals from prefrontal cortex and thalamic relays that modulate sensory gain on the fly \cite{corbetta2002control}.  By contrast, transformer attention is fixed once its weights are learned; it does not receive real-time executive feedback about goals or context.  Hence we label current AI as L2: computational attention is powerful but still lacks the adaptive, goal-directed control that characterises biological attention, making this an active frontier of research.
	\item \textbf{Sensorimotor coordination:} Parietal lobe helps translate between sensory coordinates and motor coordinates – for instance, computing how to reach for a seen object (integrating visual location with arm position). Some AI systems (robotic manipulators with vision) approximate this, using calibration and learned coordinate transforms. Still, human parietal cortex excels at online adjustments and using context (like adjusting reach if an object is moving). AI is catching up in domains like robotic arm manipulation, but general sensorimotor integration remains L2 (demonstrated in specific setups but not as universally robust as in humans).
\end{itemize}
Notably, AI has very limited touch sensing (making that L3), whereas the parietal somatosensory cortex finely discriminates texture, pressure, etc. Overall, parietal lobe functions are critical for an agent to perceive its world and orient within it, areas where AI has some successes (especially in vision) but still lacks the generality and fluidity of human perception.

\paragraph*{Occipital Lobes (Visual Processing)}
The occipital lobes are the brain's visual processing center \cite{kandel2013principles}. The primary visual cortex (V1) in occipital lobe receives input from the eyes (via the thalamus) and extracts low-level features like edges. From there, occipital regions and adjacent visual areas (extending along the occipital-temporal border for the ventral stream, and occipital-parietal for the dorsal stream) hierarchically build up visual perception: detecting shapes, colors, motion, and eventually complex patterns and objects. In summary, the occipital lobe is primarily responsible for processing visual information.

In the context of AI, vision has been one of the most successful domains – thanks largely to deep learning:
\begin{itemize}
	\item \textbf{Visual perception (Recognition):} Machine vision systems (convolutional neural networks and their successors) can now match or exceed human performance in tasks like object recognition, face detection, and image classification \cite{lecun2015deep}. This corresponds to the L1 level (well-developed) in AI. For example, AI vision models can instantly recognize thousands of object categories in images, a feat once thought to require human-like vision. This maps to what the occipital lobe and ventral visual cortex do – identifying what is in the visual field.
	\item \textbf{Scene understanding and visual reasoning:} Beyond raw perception, humans readily understand spatial relationships in a scene, contextual clues, and can perform reasoning on visual inputs (e.g. predicting what might happen next in a scene, or solving a visual puzzle). AI is part-way (L2) here. Some systems can caption images or answer questions about a scene (vision-language models), indicating a degree of semantic understanding. Yet, these models often lack true grounded understanding – they might label objects but fail on deeper comprehension (for instance, understanding intentionality or causality from an image). Visual reasoning tasks (like answering abstract questions about a picture or performing complex video analysis) remain challenging.
	\item \textbf{Visual attention and eye movements:} Humans constantly move their eyes and focus on important parts of the visual field (a function involving occipital and parietal circuits). AI vision models don't literally move eyes, but some incorporate attention mechanisms that mimic focusing on regions of an image. This is related to the earlier discussion on attention (shared with parietal function). It's moderately well implemented in AI (L2), but we don't explicitly label this under occipital as it can be considered part of the general attention function under parietal/frontal coordination.
\end{itemize}
AI has heavily developed the visual recognition capabilities, but capabilities like real-time 3D visual guidance are less mature (though present in robotics and self-driving cars). This distinction shows another gap in AI relative to the human visual system's integrated prowess.

\paragraph*{Temporal Lobes (Memory, Language, and Audition)}
The temporal lobes have diverse but crucial roles in cognition, spanning auditory processing, language, memory, and high-level visual recognition \cite{purves2018neuroscience}. The upper part of the temporal lobe (superior temporal gyrus) contains the primary auditory cortex, which processes sound inputs – frequencies, rhythms, etc. Adjacent areas (e.g. Wernicke's area in the left temporal lobe \cite{wiki:WernickesArea}) are essential for language comprehension, linking sounds to meaning. The medial temporal lobe houses the hippocampus and related structures, which are the heart of the brain's episodic memory system (forming and retrieving autobiographical memories) and also support spatial navigation. The temporal lobe's ventral visual stream (inferior temporal cortex) specializes in pattern recognition – including recognizing complex stimuli like faces (the fusiform face area) and scenes. In short, the temporal lobe is a multifaceted hub for recognition and memory.

Mapping these to AI capabilities:
\begin{itemize}
	\item \textbf{Language comprehension and production:} Human language ability relies on temporal lobe (comprehension of words, meanings) in concert with frontal lobe (speech production via Broca's area, and broader language planning). AI has seen remarkable advances here – LLMs can now parse text and generate fluent responses, indicating a high level of language competence in narrow settings. Machine translation, speech recognition, and speech synthesis are also quite advanced. Thus, for linguistic processing, AI is at L1 in many respects. An AI can ``comprehend'' and produce text in multiple languages with little human-like effort, though we should note it's often statistical rather than grounded understanding. Still, relative to other cognitive domains, language is a success story for AI, so the figure should mark functions like ``Language Comprehension/Production'' as well-developed (L1).
	\item \textbf{Auditory perception:} The ability to parse sound – speech, music, environmental noises – is another temporal lobe function. AI matches or exceeds humans in low-level auditory tasks like speech-to-text transcription under ideal conditions (think of virtual assistants accurately recognizing spoken commands). This is L1 for narrow cases (e.g. trained speech recognizers). However, true auditory scene analysis (understanding a cacophony of sounds, picking out one conversation in a noisy room – the ``cocktail party effect'' \cite{wiki:CocktailPartyEffect}) remains very hard for AI. So there are still aspects at L2/L3. But in Figure~\ref{fig:brain}, we label ``Auditory Processing (L1)'' given core progress in speech recognition.
	\item \textbf{Episodic memory \& learning:} The hippocampus enables us to form new episodic memories (remembering experiences in context) and to perform lifelong learning by integrating new memories without wiping old ones. AI's analogs here are continual learning algorithms and memory-augmented networks. This area is underdeveloped (L3) – most AI systems do not learn continuously in a stable way; they suffer catastrophic forgetting if trained on new data unless special techniques are used. They also lack the rich, context-tagged memory of experiences that humans have. We therefore label this function as ``Episodic Memory \& Lifelong Learning (L3)''.
	\item \textbf{Semantic memory and understanding:} Beyond specific events, humans build up semantic memory – factual and conceptual knowledge about the world (much of this is linked to temporal lobe association areas as well). AI in some sense has simulated semantic memory: knowledge graphs, vast pretrained models that encode facts (e.g. GPT knows many facts from its training). So semantic understanding is partially there (L2). But AI's knowledge can be superficial or lacking true comprehension of context. Thus, in Figure~\ref{fig:brain}, we include ``Semantic Knowledge/Understanding (L2)'' as a function.
	\item \textbf{Face and object recognition:} The temporal lobe's ventral stream areas identify objects and faces. AI vision is quite good at this (object and face recognition are at L1 with deep learning). This has also been captured under occipital functions already (visual perception).
\end{itemize}

\paragraph*{Cerebellum (Coordination, Skill Learning, and Timing)}
The cerebellum – the ``little brain'' at the back – is traditionally known for motor coordination and motor learning. It fine-tunes movements, maintaining balance and posture, and ensures movements are smooth and accurate \cite{knierim2020cerebellum}. When you learn a physical skill (like riding a bicycle or playing piano), the cerebellum is heavily involved in adapting motor commands through practice – essentially performing adaptive error correction based on feedback. Notably, the cerebellum contains more neurons than the rest of the brain combined, arranged in a highly regular circuitry ideal for learning patterns. In recent decades, research has revealed the cerebellum also contributes to certain cognitive and emotional functions, acting as a predictor or timing mechanism even in non-motor tasks. It's been implicated in language (e.g. helping predict the timing of syllables) and even in aspects of attention and executive function. In essence, the cerebellum builds internal models that allow the brain to make fine-grained predictions and adjustments.

For AI, the cerebellum's roles translate to capabilities that are still not fully realized in agents:
\begin{itemize}
	\item \textbf{Motor coordination and skill learning:} In robotics, there are control algorithms and learning methods (like reinforcement learning with feedback) that echo what the cerebellum does. For instance, adaptive controllers can learn to correct a robot arm's movements (analogous to cerebellar error correction). However, robots remain far clumsier than humans. They often lack the real-time adaptive finesse the cerebellum provides. This area is partially developed (L2) – we have examples of robotic learning for specific skills, but a general ``cerebellum-like'' module for adaptive motor control is not present in most AI agents. Thus, we list ``Motor Skills \& Coordination (L2)'' by the cerebellum to highlight this gap.
	\item \textbf{Cognitive timing and prediction:} The cerebellum is thought to function as an internal clock and predictor for events on the order of tens to hundreds of milliseconds. This is crucial for tasks like predicting sensory outcomes of one's actions or timing when to initiate a sequence. AI systems typically do have predictive models (e.g. forward models in model-based reinforcement learning), but these are often task-specific. There isn't a general mechanism like the cerebellum's that seamlessly handles timing for perception and action across domains. Timing in AI agents (e.g. predicting when something will happen, not just what) is relatively underexplored (L3). One could imagine future AI agents with a dedicated module for temporal prediction and smooth sequencing – analogous to cerebellar function – but currently this is rudimentary.
	\item \textbf{Error correction:} This overlaps with coordination, but extends to cognitive domains. For example, cerebellar activity has been observed in language processing, possibly helping to predict and correct linguistic sequences \cite{moberget2016cerebellar}. AI does perform error correction in training (via backpropagation), but online error correction during tasks is limited. Real-time adaptive control (a feedback loop adjusting actions on the fly) is present in some advanced systems (e.g. adaptive cruise control in cars, or self-balancing robots), yet it's not at human proficiency. We mark ``Adaptive error correction (L2)'' under cerebellum to reflect that AI can do this in narrow cases but lacks a general, brain-like capability to adapt behaviors fluidly whenever mismatches occur.
\end{itemize}
The cerebellum is often left out of high-level AI comparisons, but it shouldn't be – it highlights the embodied, fine-control intelligence humans have. A particularly interesting insight from computational neuroscience is that different brain modules may correspond to different learning paradigms: cerebellum for supervised learning, basal ganglia for reinforcement learning, and cerebral cortex for unsupervised learning. This was proposed by Doya (1999) \cite{doya1999computations} and others, noting how the cerebellum takes error signals (like a supervised loss), basal ganglia use reward feedback, and cortex finds patterns in data. This perspective can inspire AI: e.g., designing an agent with separate components for these learning types, mirroring brain organization. We include the cerebellum's function (predictive control via error-driven learning) in Figure~\ref{fig:brain} to show that it's a major gap in current AI agents that mostly rely on reinforcement learning and (in the case of deep networks) a form of error-driven learning during training only, not continuous adaptation like the cerebellum performs in real-time.

\paragraph*{Brainstem (Basic Autonomic and Arousal Functions)}
The brainstem (midbrain, pons, medulla) is the most evolutionarily ancient part of the brain, responsible for fundamental life-sustaining processes and reflexes. It acts as the main communication highway between the brain and body, and houses nuclei that control breathing, heart rate, blood pressure, swallowing, and reflexive actions like blinking \cite{clevelandclinic:Brainstem}. In addition, the brainstem contains the reticular activating system, a network that regulates sleep-wake cycles and overall arousal level (i.e., how alert or vigilant you are). It also contributes to balance and posture (through vestibular nuclei) and coordinates head/eye movements via reflexes. Essentially, the brainstem keeps the body running and primes the brain's level of consciousness.

In terms of AI or robotics:
\begin{itemize}
	\item	\textbf{Survival autopilot:} Many of the brainstem's duties have no direct analogue in non-embodied AI (a chatbot doesn't need to regulate blood pressure!). However, in robotics, low-level control loops (for locomotion, balance, etc.) play a similar role. For instance, a bipedal robot uses feedback controllers that mimic reflexes to keep upright. These can be considered L1 (well-developed) in very narrow scopes – engineers can design reflexive responses (like a withdrawal reflex if a robot arm meets resistance). We also label ``Reflexive Responses (L1)'' at the brainstem, as simple reflex-like behaviors in robots (or even in software agents, e.g. an immediate reaction to an input) are straightforward and implemented.
	\item	\textbf{Autonomic regulation:} Since AI agents don't have a body with physiology, they lack an equivalent of the autonomic nervous system. One could argue that some AI systems regulate internal variables (CPU temperature throttling, memory management) automatically, but this is a stretch as a cognitive function. Thus, we label ``Autonomic Regulation (L3)'' to indicate it is underexplored in AI. If we consider future embodied AI (like intelligent androids), they might need something like this to manage power, self-maintenance, etc., but it's speculative.
	\item	\textbf{Arousal and global attention state:} The brainstem's influence on arousal has interesting parallels to AI in the sense of adaptive computation. Humans can be drowsy or hyper-alert, which affects how we process information. AI systems currently lack any explicit notion of ``being alert'' vs ``tired'' – they run in a fixed mode unless programmed otherwise. There is research into adaptive AI that could, say, slow down to save energy or limit computation when not needed, but it's not mainstream. We mark ``Arousal/Attention States (L3)'' as largely unaddressed. However, one might draw a parallel with how some AI models can attend more or less strongly to inputs (controlled by parameters), somewhat akin to gain control in neurons under different arousal. This is a loose analogy; overall, the global modulatory role the brainstem (with neuromodulators like norepinephrine, serotonin) plays – affecting mood and readiness – is missing in AI agents.
\end{itemize}
We should clarify that the functions we discussed here are primarily relevant for embodied AI or robotics. For instance, a self-driving car's automatic braking when a collision is imminent is reflex-like (and indeed implemented in today's tech). But an AI algorithm in isolation doesn't have ``body regulation''. So, brainstem functions underscore how biological intelligence is deeply tied to a body, whereas AI often abstracts that away. It's an important conceptual gap for readers to appreciate: truly human-like AI might need analogue systems for maintaining its ``well-being'' (self-preservation, energy management) and adjusting its alertness to situations – concepts drawn from the brainstem and related systems.

\paragraph*{Subcortical Systems (Thalamus, Basal Ganglia, Limbic System)}
Finally, beyond the six major regions above, subcortical systems deserve representation, as they are crucial to cognition and differ markedly from what current AI implementations include. These structures are embedded below the cortex and often coordinate with multiple cortical regions:
\begin{itemize}
	\item \textbf{Thalamus (Sensory Relay and Attention Filter):} The thalamus is often called the brain's ``gateway'' or relay station – almost all sensory signals (except smell) pass through thalamic nuclei before reaching cortex \cite{clevelandclinic:Thalamus}. But the thalamus does more than relay: it actively modulates and integrates signals. It plays a key role in attention by amplifying relevant signals and suppressing others, under guidance from cortical feedback \cite{wimmer2015thalamic}. It's also involved in maintaining consciousness (targeted by anesthetics) and coordinating cortical rhythms. In AI, there is no single equivalent of a thalamus. However, one might liken it to routing layers or attention mechanisms that decide which data go where. The concept of a central hub that gates information flow in a network is present in some neural network architectures (e.g. transformer attention decides which inputs ``attend'' to which others), but the thalamus's dynamic, task-dependent control is beyond current AI. We mark functions like ``Sensory Integration \& Routing'' as L2 (partially present conceptually in AI via attention layers), and ``Global Workspace'' as L3 (largely absent – AI doesn't have a unified workspace model equivalent to what some cognitive theories assign to thalamocortical circuits). 

	\item \textbf{Basal Ganglia (Action Selection and Reinforcement Learning):} The basal ganglia are a group of nuclei (caudate, putamen, globus pallidus, substantia nigra, etc.) that are central to selecting and initiating actions, and they implement a biological form of reinforcement learning. They take inputs from the cortex (especially frontal and parietal areas), and through complex loops, they determine which actions are facilitated or inhibited, often by evaluating expected rewards or outcomes \cite{larry2024organization}. Dopamine signals from the midbrain (e.g. from the substantia nigra pars compacta) encode reward prediction errors, a concept very much like the reward signals in AI RL algorithms. In fact, neuroscientific evidence suggests the basal ganglia ``learn'' which actions lead to reward via dopamine-mediated plasticity – a direct parallel to how AI agents update policies from reward feedback. We can confidently say ``Reward-Based Learning and Habit Formation'' are primary functions of the basal ganglia. AI has a whole subfield of reinforcement learning, which has seen successes (games, some robotic tasks), so this is moderately developed (L2) in AI. However, current AI RL is narrow and data-hungry compared to human habit learning. Basal ganglia also contribute to procedural memory (learning habits or skills that become automatic) – something AI doesn't explicitly differentiate (it learns policies, but the idea of habits vs. goal-directed actions is an emerging concept in AI research). 

	\item \textbf{Limbic System – Amygdala and Hippocampus (Emotion and Memory):} We touched on these in the temporal lobe section, but to reiterate: the amygdala is crucial for processing emotional significance of stimuli and fear conditioning (learning to avoid harmful situations) \cite{wiki:LimbicSystem}. It assigns value (good or bad) to experiences, which then influences decision-making and memory (through its connections to hippocampus and frontal cortex). The hippocampus, as mentioned, enables forming new declarative memories and mapping environments (it's often likened to the brain's GPS for spatial memory). In AI, there are nascent attempts to model hippocampal function – e.g. neural network ``memory'' modules for episodic recall, or models of spatial navigation that emulate place cells and grid cells found in the hippocampal formation. These are still L3 overall (exploratory). As for the amygdala's role, AI currently lacks genuine emotion; at most, we simulate ``emotion'' as reward functions or use sentiment analysis to detect emotions in text, which is not an internal drive. We label ``Emotion Processiong \& Learning (L3)'' and ``Episodic Memory (L3)'' to highlight capabilities largely missing in AI agents: affective computing (AI that can experience or at least robustly respond to emotions) is very limited, and one-shot contextual learning (storing a new event and generalizing from it) is also an open problem.
	\item \textbf{Hypothalamus (Drives and Homeostasis):} The hypothalamus orchestrates the endocrine system and autonomic nervous system to maintain internal balance – it controls things like hunger, thirst, temperature, and release of hormones \cite{jimenez2023hypothalamic}. It also generates primitive drives (e.g. hunger drives you to seek food, which the cortex then plans for). AI agents do not have intrinsic survival needs, so they lack any true equivalent of homeostatic drives. We sometimes give AI an objective function (e.g. maximize score), but these are externally defined and do not fluctuate like biological needs. To the extent researchers are exploring intrinsic motivation for AI (like curiosity-based rewards), it is still rudimentary. In the figure, we add ``Motivation \& Drives (L3)'' to acknowledge this gap. It reminds us that a human-inspired AI agent might require internally generated goals (not just tasks imposed by users) to be truly autonomous and robust in varied environments. It should be noted that giving AI intrinsic motivation is also seen as a potentially dangerous direction and should be treated with great caution \cite{bengio2025superintelligent}.
\end{itemize}
Bringing these subcortical pieces together, we see that many are poorly represented in today's AI. Current intelligent systems are heavily cortex-like (perception modules, decision logic, etc.) but lack the rich support system of the subcortical brain: no analog of a thalamus to smartly route information, no hypothalamus to create self-preserving goals, a primitive version of basal ganglia for RL at best, and minimal emotional or episodic memory faculties. The figure's accompanying explanation should drive home that cognition arises from cortical–subcortical interactions. For example, decision-making is not just frontal (cortical) deliberation, but also involves basal ganglia (habits and dopamine rewards) and amygdala (emotional bias) and hypothalamus (drive states). Emphasizing these interactions will lend a more nuanced and truthful picture than a simplistic lobe-by-lobe map. It also inspires AI researchers to think about architectures that incorporate these principles – such as an agent that, say, has a core RL module (analogous to basal ganglia) for learning from rewards, a memory module (analogous to hippocampus) for episodic recall, and perhaps a ``global workspace'' (inspired by thalamocortical loops) for attention and context integration.

\paragraph*{Bridging Brain-Like Functions and Building Beneficial AI}
Until now, we have witnessed the gap between human brain and machine intelligence. Nevertheless, the objective is not necessarily to replicate every facet of human cognition within artificial intelligence systems. Rather, our overarching aim should be to develop intelligent agents that are useful, ethical, safe, and beneficial to society. By critically comparing human and artificial intelligence, we highlight the existing gaps and illuminate promising directions for innovation. This comparative perspective allows us to selectively integrate beneficial aspects of human cognition, such as energy-efficient processing, lifelong adaptive learning, emotional grounding, and rich creativity, while simultaneously innovating beyond human limitations. Ultimately, this approach aims to foster the creation of more capable, resilient, and responsible AI systems.

Furthermore, it is vital to consider the evolving role of humans within a hybrid Human-AI society. The goal of AI should not be to replace human roles entirely, but rather to augment and empower human abilities, complementing human skills and judgment in areas where AI excels, such as handling vast datasets, performing rapid calculations, and automating repetitive tasks. Human oversight and interpretability are essential to ensure that powerful AI systems remain controllable and aligned with human values and ethical standards. Thus, the core objective must be the development of AI technologies that are transparent, interpretable, and responsive to human guidance.

Human-centered AI design emphasizes collaboration, safety, and social responsibility, ensuring technological advancement proceeds in a controlled, reliable manner. By placing humans at the center of the AI ecosystem, we can harness AI's potential to enhance human productivity, creativity, and decision-making, facilitating technical and societal progress without compromising human autonomy or dignity. Ultimately, a thoughtful integration of human intelligence and AI capabilities can pave the way for a sustainable, equitable, and prosperous future.

\section{Foundation Agents: A Modular and Brain-Inspired AI Agent Framework}

\lettrine[lines=3]{\initfamily\textcolor{darkgreen}{O}}{ne} core issue in the LLM era is the \emph{lack of a unified framework} that integrates the rich cognitive and functional components required by advanced agents. While LLMs offer exceptional language reasoning capabilities, many current agent designs remain \emph{ad hoc}. They incorporate modules like perception, memory, or planning in a piecemeal fashion, failing to approximate the well-coordinated specialization seen in biological systems such as the human brain. Unlike current LLM agents, the human brain seamlessly balances perception, memory, reasoning, and action through distinct yet interconnected regions, facilitating adaptive responses to complex stimuli. LLM-driven agents, by contrast, often stumble when tasks require cross-domain or multimodal integration, highlighting the need for a more holistic approach akin to the brain's functional diversity.
Motivated by these parallels, our work advocates drawing inspiration from the human brain to systematically analyze and design agent frameworks. This perspective shows that biological systems achieve general intelligence by blending specialized components (for perception, reasoning, action, etc.) in a tightly integrated fashion, an approach that could serve as a blueprint for strengthening current LLM-based agents.

Neuroscientific research reveals that the brain leverages both \textbf{rational circuits} (e.g., the neocortex, enabling deliberation and planning) and \textbf{emotional circuits} (e.g., the limbic system) to guide decision-making. Memory formation involves the hippocampus and cortical mechanisms, while reward signals, mediated by dopaminergic and other neuromodulatory pathways, reinforce behavior and learning. These biological insights inspire several design principles for AI agents, including but not limited to: 
\begin{itemize}
    \item \textbf{Parallel, Multi-Modal Processing:} The brain processes visual, auditory, and other sensory inputs in parallel through specialized cortical areas, integrating them in associative regions. Similarly, AI agents benefit from parallel processing of diverse sensor streams, fusing them in later stages for coherent understanding.

    \item \textbf{Hierarchical and Distributed Cognition:} Reasoning, planning, emotional regulation, and motor control involve interactions between cortical and subcortical regions. Analogously, AI agents can employ modular architectures with subsystems dedicated to rational inference, emotional appraisal, and memory.

    \item \textbf{Attention Mechanisms:} Human attention prioritizes sensory information based on context, goals, and emotions. AI agents can replicate this by modulating perception through learned attention policies, dynamically adjusting focus based on internal states.

    \item \textbf{Reward and Emotional Integration:} Emotions are not merely noise but integral to decision-making, modulating priorities, enhancing vigilance, and guiding learning. Reward-driven plasticity facilitates habit formation and skill acquisition, a concept critical to reinforcement learning in AI agents.

    \item \textbf{Goal Setting and Tool Usage:} The human prefrontal cortex excels at setting abstract goals and planning action sequences, including tool uses. Similarly, AI agents require robust goal-management systems and adaptive action repertoires, driven by external rewards and intrinsic motivations.
\end{itemize}
These principles form the foundation of our proposed \textbf{brain-inspired agent framework}, where biological mechanisms serve as inspiration rather than direct replication.

Before we formalise our framework design for AI agents, we pause to ask a simpler question: 
\emph{What faculties must any truly autonomous agent possess, no matter how the modules are wired?}
Answering it further clarifies why each box in our forthcoming diagram is indispensable.

\subsection{From Language Models to AI Agents}
\label{subsec:llm2agent}

Large language models can already analyse prose, write code, and argue a point, yet they live in a closed book: they read tokens, write tokens, and forget the scene once the page turns.  An \emph{agent}, by contrast, survives in the wild.  Think of walking home after work: eyes check traffic, feet adjust to kerbs, memory recalls a shortcut, an inner film plays possible detours, hunger tugs the route towards a deli, and the whole routine improves with every trip.  To grant an LLM the same street-smarts we must graft on several faculties, including (but not limited to) perception, action, memory, world-model, motivation, and learning, each as real as the words it speaks.  The paragraphs that follow sketch those faculties, the headaches they bring, and how they fit into a single perception–cognition–action loop that anchors this book.

\paragraph*{Perception: seeing and hearing beyond tokens.}
A text-only model is the cognitive equivalent of reading ticker tape in a dark room.  Humans, in contrast, fuse vision, audition, and touch in parallel cortical streams \cite{kandel2013principles}.  Multimodal models such as GPT-4 already accept images \cite{openai2023gpt4}, yet an agent that sorts parts on a bench or watches market charts needs deeper channels.  Three problems arise: fusion (aligning different sensor streams); noise (coping with glare, static, or hostile inputs); and task-aware attention (deciding which slice of the torrent matters \emph{now}).  Good perception is not a dashboard of data; it is a spotlight that obeys the mission.

\paragraph*{Action: talking is cheap; doing is hard.}
Text output is enough for a chatbot, but powerless to open a door.  Modern tool-using agents treat tokens as API calls: code execution in ReAct \cite{yao2023react}, plugin invocations in CoALA \cite{sumers2024cognitive}, motion scripts for robots.  Once an agent can pay invoices or steer drones, safety becomes paramount.  The designer must guarantee grounding (the model grasps an action's real impact), syntax fidelity (calls obey the tool's grammar), and alignment (behaviour stays within human intent).

\paragraph*{Memory: more than a prompt window.}
People store decades of episodes; an LLM forgets everything outside its context length.  External stores, including vector databases, structured logs, or the memory stream in Generative Agents \cite{park2023generative}, let a model recall prior events and sustain identity across sessions.  Headaches follow: curation (what to keep), retrieval (finding the right shard when it matters), and catastrophic forgetting if online updates rewrite the past \cite{kirkpatrick2017overcoming}.  Memory must be selective yet searchable, stable yet plastic.

\begin{table}[!ht]
\centering
\caption{Notation summary for the revised agent framework, highlighting separate \emph{learning} and \emph{reasoning} functions within the overall cognition process.}
\label{tab:notation_summary}
\renewcommand{\arraystretch}{1.2} 
\setlength{\tabcolsep}{4pt} 
\begin{tabular}{@{}p{0.2\textwidth}p{0.75\textwidth}@{}}
\toprule
\textbf{Symbol} & \textbf{Meaning} \\
\midrule
\rowcolor{LightMint}
$\mathcal{W}$ & The world with society systems that encapsulate both environment and intelligent beings (AI or human). \\
$\mathcal{S}$    
& State space of the \textbf{environment}.  \\
\rowcolor{LightMint}
$s_t \in \mathcal{S}$   
& Environment's state at time $t$. \\
$\mathcal{O}$           
& Observation space. \\
\rowcolor{LightMint}
$o_t \in \mathcal{O}$   
& Observation at time $t$ (potentially shaped by \emph{attention} or other perception filters). \\

$\mathcal{A}$           
& Agent's action space. \\
\rowcolor{LightMint}
$a_t \in \mathcal{A}$   
& Action output by the agent at time $t$. This can be an external (physical) action or an \emph{internal} (mental) action such as \emph{planning} or \emph{decision-making}. \\

\midrule
$\mathcal{M}$           
& Space of all \textit{mental states}. \\
\rowcolor{LightMint}
$M_t \in \mathcal{M}$   
& Agent's mental state at time $t$, encompassing sub-components (memory, emotion, etc.). \\
$M_t^{\mathrm{mem}}$    
& \textit{Memory} component in $M_t$ (e.g., short-term or long-term knowledge). \\
\rowcolor{LightMint}
$M_t^{\mathrm{wm}}$     
& \textit{World model} component in $M_t$ (internal representation of how the environment evolves). \\
$M_t^{\mathrm{emo}}$    
& \textit{Emotion} component in $M_t$ (internal valence, arousal, or affective states). \\
\rowcolor{LightMint}
$M_t^{\mathrm{goal}}$   
& \textit{Goal} component in $M_t$ (objectives, desired outcomes, intentions). \\
$M_t^{\mathrm{rew}}$    
& \textit{Reward/Learning} signals in $M_t$ (drives updates to preferences, values, or policy). \\

\midrule
\rowcolor{LightMint}
$\mathrm{L}$            
& \textbf{Learning function}: $\mathrm{L}:\,\mathcal{M}\times \mathcal{A}\times \mathcal{O} \;\to\; \mathcal{M}$. Responsible for \emph{updating} or \emph{learning} the next mental state (e.g., memory, world model, emotion), based on the previous mental state $M_{t-1}$, the previous action $a_{t-1}$, and the new observation $o_t$. Reflects how the agent \emph{acquires or revises} knowledge, skills, or preferences. \\

$\mathrm{R}$            
& \textbf{Reasoning function}: $\mathrm{R}:\,\mathcal{M} \;\to\; \mathcal{A}$. Responsible for deriving the \emph{next action} $a_t$ given the \emph{updated} mental state $M_t$. Can involve \emph{planning}, \emph{decision-making}, or other internal logic. \\
\rowcolor{LightMint}
$\mathrm{C}$            
& \textbf{Cognition function}: $\mathrm{C}:\,\mathcal{M}\times \mathcal{A}\times \mathcal{O} \;\to\; \mathcal{M}\times \mathcal{A}$. Encapsulates both \emph{learning} (\(\mathrm{L}\)) and \emph{reasoning} (\(\mathrm{R}\)). Concretely, $(M_t, a_t) = \mathrm{C}(M_{t-1}, a_{t-1}, o_t)$ means the agent first \emph{learns} the new mental state $M_t = \mathrm{L}(M_{t-1}, a_{t-1}, o_t)$, then \emph{reasons} about the next action $a_t = \mathrm{R}(M_t)$. \\

$\mathrm{E}$            
& \textbf{Action execution (effectors)}: $\mathrm{E}:\,\mathcal{A} \;\to\; \mathcal{A}$. (Optional) transforms or finalizes $a_t$ before applying it to the environment (e.g., converting a high-level command into low-level motor signals). \\
\rowcolor{LightMint}
$\mathrm{T}$            
& \textbf{Environment transition}: $\mathrm{T}:\,\mathcal{S}\times \mathcal{A} \;\to\; \mathcal{S}$. Defines how the environment state evolves from $(s_t, a_t)$ to $s_{t+1}$. \\

\bottomrule
\end{tabular}
\end{table}

\paragraph*{World model and planning: imagining before acting.}
Humans rehearse futures in the mind's eye; a chess engine searches moves ahead; a model-based RL agent predicts dynamics \cite{sutton1998reinforcement}.  An LLM agent benefits from an internal simulator (symbolic, neural \cite{ha2018world}, or hybrid) to test ``what-if'' hypotheses before acting.  Three hurdles block the way: accuracy (bad dreams mislead), compute (long roll-outs burn time), and arbitration (when to trust fast intuition and when to press the slow \emph{plan} button).

\paragraph*{Goals and motivation: the why behind the what.}
Every action needs a reason.  Classical agent definitions stress sensing and acting \cite{russell1995aima}, but autonomous systems also need enduring agendas \cite{maes1995artificial,franklin1997agent}.  In code, goals appear as reward functions, symbolic objectives, or scripted drives.  Poorly chosen, they invite reward-hacking and the paper-clip parable \cite{amodei2016concrete,nick2014superintelligence}.  Safe goal design blends clear constraints, human oversight, and research on intrinsic motives such as curiosity.

\paragraph*{Learning and adaptation: yesterday's lessons, tomorrow's edge.}
A frozen model stagnates; a courier robot must improve with every delivery.  Continual-learning methods (replay, regularisation, dynamic layers \cite{parisi2019continual}) aim to graft new skills without erasing old ones.  Live updates, however, risk drift from the tested baseline, so production systems often confine change to side modules while keeping the core weights fixed.

Perception feeds observations to \emph{cognition}; cognition updates a structured mental state and emits actions; the environment responds; the cycle repeats.  This lean scaffold mirrors the brain's cortex–subcortex dialogue and leaves later chapters free to zoom into specialised memories, hierarchical planners, or social protocols.

Based on the above discussions, in the following, we outline our framework's key concepts, introducing a unified agent architecture based on the \emph{perception–cognition–action loop} enriched by reward signals and learning processes. Each subsystem is carefully defined and interconnected to ensure transparency in how memory, world models, emotions, goals, rewards, and learning interact.
We formalize cognition as a general reasoning mechanism, with \emph{planning} and \emph{decision-making} framed as specific ``mental actions'' shaping behavior. Connections to established theories, such as Minsky's \emph{Society of Mind}~\cite{minsky1988society}, Buzsáki's \emph{inside-out} perspective~\cite{Buzsaki2019}, and Bayesian active inference~\cite{Friston2011}, are explored to highlight the framework's generality and biological plausibility.

\subsection{Core Concepts and Notations in the Agent Loop}
\label{subsec:agent-framework-symbols}

Our architecture operates at three conceptual levels: \textbf{Society}, \textbf{Environment}, and \textbf{Agent}. The \emph{Agent} is then decomposed into three main subsystems: \textbf{Perception}, \textbf{Cognition}, and \textbf{Action}. Within \emph{Cognition}, we identify key submodules: \emph{memory}, \emph{world model}, \emph{emotional state}, \emph{goals}, \emph{reward}, \emph{learning}, and \emph{reasoning} processes (including ``planning'' and ``decision-making'' as special actions produced with reasoning). \emph{Attention} is primarily handled within perception and cognition. Before presenting the formal loop, we summarize our symbols in Table~\ref{tab:notation_summary}.


In the following, based on the notations in Table~\ref{tab:notation_summary}, we present our proposed agent loop.

\begin{definition}[\textbf{The Agent Loop}]
\label{def:foundation-agent-loop}

An intelligent agent operates in discrete time steps $t$, continuously interacting with its environment. At each step, the following processes occur:

\begin{enumerate}

\item \textbf{Environment State} ($s_t \in \mathcal{S}$): 

The environment is in state $s_t$.

\item \textbf{Perception ($\mathrm{P}$)}: The agent perceives the environment to generate observations $o_t$:
\[
o_t = \mathrm{P}(s_t, M_{t-1}),
\]
where $M_{t-1}$ guides selective attention and filtering.

\item \textbf{Cognition ($\mathrm{C}$)}: Updates mental state and selects actions:
\[
(M_t, a_t) = \mathrm{C}(M_{t-1}, a_{t-1}, o_t).
\]
where $M_t$ encapsulates different sub-states:
\[
M_t = \{M_t^{\text{mem}}, M_t^{\text{wn}}, M_t^{\text{emo}}, M_t^{\text{goal}}, M_t^{\text{rew}}, \cdots \}.
\]
Cognition consists of:
\begin{itemize}
    \item \textbf{Learning} ($\mathrm{L}$): Updates mental state based on observations:
    \[
    M_t = \mathrm{L}(M_{t-1}, a_{t-1}, o_t).
    \]

    \item \textbf{Reasoning} ($\mathrm{R}$): Determines the next action:
    \[
    a_t = \mathrm{R}(M_t),
    \]
    which may be:
    \begin{itemize}
        \item \textbf{External Actions}, directly affecting the environment.
        \item \textbf{Internal Actions}, including:
            \begin{itemize}
                \item \textit{Planning}: Internal sequence of future actions.
                \item \textit{Decision-making}: Choosing the best action from available options.
            \end{itemize}
    \end{itemize}
\end{itemize}

\item \textbf{Action Execution ($\mathrm{E}$)}: Transforms action $a_t$ into executable form:
\[
a'_t = \mathrm{E}(a_t).
\]

\item \textbf{Environment Transition ($\mathrm{T}$)}: The environment responds to the agent's action:
\[
s_{t+1} = \mathrm{T}(s_t, a'_t).
\]

\end{enumerate}

In multi-agent scenarios, each agent $i$ maintains individual states $(M_t^i, a_t^i, o_t^i)$, and the environment collectively updates based on all agents' actions. At broader scales (AI societies or worlds, $\mathcal{W}$), agents interact within diverse social systems (e.g., economic, communication, or transportation), forming complex societal structures.

\end{definition}

Figure~\ref{fig:agent-framework} illustrates our agent framework, presenting the core concepts and different types of information or control flows among them.
Until now, we have presented a brain-inspired agent framework that integrates biological insights into a formal \emph{Perception--Cognition--Action} loop. By decomposing cognition into modules for memory, world modeling, emotion, goals, reward-based learning, and reasoning, we capture essential parallels with the human brain's hierarchical and reward-driven processes. Critically, \emph{attention} is included in the loop to enable selective filtering based on internal states. Furthermore, \emph{planning} and \emph{decision-making} can be viewed as distinct internal (mental) actions that either refine internal representations or select external behaviors. Our framework naturally extends classical agent architectures, providing a multi-level structure that integrates emotional and rational processes as well as robust, reward-driven learning across short and long timescales.

\begin{figure}[!htb]
\centering
    \includegraphics[width=0.98\columnwidth]{figures/AgentFramework.pdf}
    \caption{An overview of our general framework for describing an intelligent agent loop and agent society.}
\label{fig:agent-framework}
\end{figure}

\textbf{Society and Social Systems.} In many real-world scenarios, agents do not merely interact with a static environment but operate within a broader \emph{society}, comprising various \emph{social systems} such as financial markets, legal frameworks, political institutions, educational networks, and cultural norms. These structures shape and constrain agents' behaviors by defining rules, incentives, and shared resources. For example, a financial system dictates how economic transactions and resource allocations occur, while a political system provides governance mechanisms and regulatory constraints. Together, these social systems create a layered context in which agents must adaptively learn, reason, and act, both to satisfy their internal goals and to comply (or strategically engage) with external societal rules. In turn, the actions of these agents feed back into the social systems, potentially altering norms, policies, or resource distributions.

\paragraph*{A Formal Definition of Foundation Agents.}
Building on these insights and our vision of robust, adaptive intelligence, we now formally introduce the concept of a \emph{Foundation Agent}. Unlike traditional agent definitions that focus primarily on immediate sensory-action loops, a Foundation Agent embodies sustained autonomy, adaptability, and purposeful behavior, emphasizing the integration of internal cognitive processes across diverse environments.

\begin{definition}[\textbf{Foundation Agent}]
\label{def:foundation-agent}
A \textbf{Foundation Agent} is an autonomous, adaptive intelligent system designed to \textbf{actively perceive} diverse signals from its environment, continuously \textbf{learn} from experiences to refine and update structured internal states (such as memory, world models, goals, emotional states, and reward signals), and \textbf{reason} about purposeful actions (both external and internal) to autonomously navigate toward complex, long-term objectives.

\vspace{2mm}

\noindent
More concisely: \textbf{foundation agent is a fundamental intelligent unit with universal understanding, cognition, and action capabilities that can operate in any environment and collaborate to form collective intelligence.}
\vspace{2mm}

A Foundation Agent possesses the following core capabilities:
\begin{enumerate}
    \item \textbf{Active and Multimodal Perception:} It continuously and selectively perceives environmental data from multiple modalities (textual, visual, embodied, or virtual).
    \item \textbf{Dynamic Cognitive Adaptation:} It maintains, updates, and autonomously optimizes a rich internal \emph{mental state} (memory, goals, emotional states, reward mechanisms, and comprehensive world models) through \textbf{learning} that integrates new observations and experiences.
    \item \textbf{Autonomous Reasoning and Goal-Directed Planning:} It proactively engages in sophisticated reasoning processes, including long-term planning and decision-making, to derive goal-aligned strategies.
    \item \textbf{Purposeful Action Generation:} It autonomously generates and executes purposeful actions, which can be external (physical movements, digital interactions, communication with other agents or humans) or internal (strategic planning, self-reflection, optimization of cognitive structures), systematically shaping its environment and future cognition to fulfill complex objectives.
    \item \textbf{Collaborative Multi-Agent Structure:} It can operate within multi-agent or agent society structures, collaboratively forming teams or communities of agents that collectively accomplish complex tasks and goals beyond individual capabilities.
\end{enumerate}
\vspace{2mm}
Foundation Agents represent a fundamental shift from traditional agents by integrating the above core features, enabling them to function effectively across a wide range of environments and domains.
\end{definition}

Unlike classical definitions, which often frame agents primarily in terms of simple perception–action loops (``perceive and act'' \cite{russell1995aima}), our notion of Foundation Agents emphasizes the depth and integration of internal cognitive processes.
In contrast to prior work \cite{liu2024position} that defines ``foundation agents'' as generalist decision models emphasizing unified representations, policy interfaces, and interactive learning across tasks, our conception highlights a deeper, brain-inspired integration, explicitly modeling mental states and goal-directed reasoning to mirror biological cognition more completely. 
Foundation Agents in our framework not only perceive their environment and perform immediate actions but also possess an evolving, goal-oriented cognition, continuously adapting memory structures, world models, emotional and reward states, and autonomously refining their strategies through reasoning. This internal cognitive richness allows Foundation Agents to autonomously decompose complex, abstract goals into actionable tasks, strategically explore their environments, and dynamically adjust their behavior and cognitive resources. Our unified \textbf{perception–cognition–action} framework thus accommodates and explicitly models these sophisticated cognitive capabilities, recognizing internal (mental) actions on par with external (physical or digital) interactions, facilitating a broad range of embodiments, from physical robots to software-based or purely textual intelligent agents.

\subsection{Biological Inspirations}
\label{subsec:bio-insipre-of-agent-loop}

Our agent framework, though fundamentally computational, finds its roots in biological systems, especially the intricate operations of the human brain. By exploring these biological parallels, we gain deeper insight into how our framework can embody key cognitive faculties that make biological agents so versatile, robust, and adaptive.

\paragraph*{Memory: Capturing Experience and Knowledge (Hippocampus and Neocortex).}
Human memory is famously rich, vivid, and multifaceted. Neuroscience describes the hippocampus as a librarian of episodic memories (specific, detailed snapshots of past events) while the neocortex houses semantic knowledge, the enduring conceptual understanding we build up through a lifetime \cite{Squire1992MemorySystems, Bear2020Neuroscience}. Together, these brain structures orchestrate the flow from immediate experiences to long-lasting wisdom.

Inspired by this partnership, our agent's memory module, $M_t^{\mathrm{mem}}$, mirrors this duality by combining short-term memory (to rapidly record ongoing interactions and context) and long-term storage (to consolidate and generalize from past experiences). Just as the brain filters and retains information based on relevance, our memory module prioritizes which experiences merit long-term preservation, preventing clutter and ensuring essential insights remain accessible when most needed.

\paragraph*{World Model: Imagining and Predicting the Future (Predictive Processing).}
One of the brain's remarkable feats is its ability to constantly predict what will happen next. Cognitive theories propose that our cortical networks operate like sophisticated prediction engines, anticipating sensory inputs and swiftly adjusting internal expectations based on mismatches \cite{Rao1999PredictiveCoding, Friston2011}. This predictive dance allows humans to navigate effortlessly through uncertainty, imagining consequences, weighing alternatives, and adjusting in real-time.

Our agent's world model, $M_t^{\mathrm{wm}}$, captures this predictive prowess by continually maintaining and updating an internal simulation of the environment. It anticipates future states, evaluates potential outcomes of actions, and dynamically updates itself based on fresh observations. Much like the brain seamlessly integrates perceptions into predictions, our agent leverages a similar Bayesian-like process to ensure its internal model stays accurate, relevant, and useful.

\paragraph*{Emotion: Guiding Behavior Beyond Pure Logic (Limbic System).}
Contrary to stereotypes, human emotions are not irrational distractions; rather, they act as powerful and nuanced guides for decision-making. Structures in the limbic system, such as the amygdala and hypothalamus, shape our focus, modulate our reactions, and enhance learning through emotional significance \cite{LeDoux1996EmotionalBrain, Damasio1995DescartesError}. For instance, fear sharpens our attention to danger, and excitement motivates exploration and creativity.

To emulate this adaptive guidance, our agent incorporates an emotion component, $M_t^{\mathrm{emo}}$. Although computational emotions do not equate to genuine human feelings, this module mimics their functional role: prioritizing attention, adjusting urgency, and directing learning efforts based on internal affective states. Like the limbic system subtly steering human choices, this emotional mechanism ensures our agent remains responsive, adaptive, and aligned with its context.

\paragraph*{Goals and Reward: Shaping Intentions and Motivations (Prefrontal \& Subcortical Circuits).}
Goals give purpose; rewards reinforce behavior. In humans, abstract goal-setting and sophisticated long-term planning are primarily orchestrated by the prefrontal cortex \cite{Miller2002PFC, badre2008cognitive}, while reinforcement signals from subcortical regions (especially dopaminergic pathways) continuously adjust our motivations and habits \cite{Schultz1997Dopamine}. This integrated circuitry enables humans to sustain complex intentions and refine actions through continuous feedback loops.

Our agent mirrors this sophisticated partnership with goal ($M_t^{\mathrm{goal}}$) and reward ($M_t^{\mathrm{rew}}$) components. Goals shape the agent's overarching intentions, guiding decisions towards desired outcomes, while the reward component continuously tunes behavior and reinforces effective strategies. Together, they form an adaptive motivational system that mirrors how the brain's goal-driven deliberation is harmonized with reward-driven adaptation, enabling flexible, contextually appropriate behavior.

\paragraph*{Reasoning, Planning, and Decision-Making: Executive Control (Prefrontal Cortex).}
The hallmark of human intelligence is arguably our capacity for reasoning and planning—deliberate, future-oriented cognition primarily governed by the prefrontal cortex. This brain region synthesizes memory, perception, emotional states, and reward signals into coherent strategies for action \cite{Fuster2008PFC, Shallice2011PFCReview}. It's what lets us imagine multiple outcomes, judge their merits, and confidently choose a path forward.

Reflecting this remarkable capability, our agent's reasoning sub-function acts as the executive core. It orchestrates internal simulations, weighs alternative actions, and selects optimal strategies—echoing how the prefrontal cortex evaluates possibilities before making a decision. By clearly distinguishing long-term planning from moment-to-moment decision-making, our agent can flexibly switch between reflective deliberation and swift, intuitive choices, just as humans seamlessly alternate between careful thought and rapid reaction.

These biological analogies enrich our computational framework by grounding its functional logic in neuroscientific realism. Yet, importantly, the parallels remain flexible, not rigidly bound to biological exactitude. They serve as guideposts rather than blueprints, highlighting fundamental cognitive principles that, when embedded into artificial agents, can yield intelligent systems capable of rich, adaptive behaviors. In subsequent chapters, we delve deeper into these modules, further exploring how they interact and evolve, guided by insights from both neuroscience and cutting-edge AI research.

\subsection{Connections to Existing Theories}
\label{subsec:connections-existing-theories}

Our foundation-agent framework isn't created from scratch; rather, it builds upon and synthesizes several influential theories from AI, cognitive science, and neuroscience. Here, we clarify these connections explicitly, highlighting both the similarities and critical enhancements we introduce.

\paragraph*{Classic Perception–Cognition–Action Cycle.}
The traditional AI perspective sees agents as engaging in a repeated loop: sensing the environment, thinking about it, and then acting accordingly~\cite{russell1995aima}. Our framework directly extends this basic cycle by incorporating richer cognitive machinery: explicit attentional control within perception ($\mathrm{P}$), fine-grained internal states such as memory, emotion, and goals within cognition ($\mathrm{C}$), and reward signals that evolve dynamically. This deeper granularity helps clarify how internal states guide perception and cognition, making it easier to understand and engineer adaptive agent behaviors.

\paragraph*{Minsky's Society of Mind.}
Marvin Minsky famously proposed that intelligence emerges from interactions among numerous specialized internal agents, each performing simpler tasks yet collectively producing complex cognition~\cite{minsky1988society}. Our modular subcomponents—memory ($M_t^{\mathrm{mem}}$), world model ($M_t^{\mathrm{wm}}$), emotion ($M_t^{\mathrm{emo}}$), goals ($M_t^{\mathrm{goal}}$), and rewards ($M_t^{\mathrm{rew}}$)—echo this idea, representing a cooperative society of distinct yet interdependent cognitive modules. Moreover, recent work on language-based agent societies, such as the Mindstorms paradigm~\cite{zhuge2023mindstorms}, supports extending Minsky's internal ``societies'' into externally interacting communities, mirroring our emphasis on multi-agent and socially structured intelligence.

\paragraph*{Buzsáki's Inside-Out Perspective.}
The neuroscientist György Buzsáki argues that brains actively construct perceptions rather than passively registering them from the outside world~\cite{Buzsaki2019}. In our model, perception is explicitly influenced by prior mental states ($M_{t-1}$), including emotions, goals, and reward expectations. This active construction of perception means our agents, like human brains, continuously refine their understanding of the environment based on internal states and past experiences, embodying the inside-out perspective in a clear computational form.

\paragraph*{Generalizing the POMDP Framework.}
Partially Observable Markov Decision Processes (POMDPs) have long provided a robust mathematical formulation for modeling agents under uncertainty. Classical POMDPs, however, rely on probabilistic transitions between environmental states and typically use externally defined scalar rewards. Our framework significantly generalizes the POMDP structure in several ways: 
i) \textit{Flexible State Transitions:} Unlike classical POMDPs constrained to probabilistic transitions, our environment transition function ($\mathrm{T}$) allows both deterministic and stochastic mappings without predefined limitations, increasing modeling versatility. 
ii) \textit{Internalized Reward:} Instead of relying on external scalar rewards, we embed reward signals within the agent's internal mental state ($M_t^{\mathrm{rew}}$). This embedding allows rewards to dynamically evolve and interact with emotions, goals, and memory, reflecting a more realistic and nuanced motivational system. 
iii) \textit{Expanded Decision-Making:} Traditional POMDPs use a straightforward value-maximization policy. By contrast, our reasoning mechanism explicitly incorporates emotions, memories, and goals into decisions, accommodating richer, more nuanced behavioral strategies, including heuristic and socially influenced choices.
iv) \textit{Modular Mental States:} Classical POMDPs collapse internal states into a singular belief representation. Our explicit modeling of separate cognitive modules (memory, emotion, etc.) significantly enhances transparency and interpretability, aligning closer to biological plausibility.

Thus, while our framework includes the classical POMDP as a simplified special case, it notably broadens the scope of possible agent behaviors, providing richer modeling capabilities.

\paragraph*{Active Inference and the Bayesian Brain.}
Karl Friston's active inference framework posits that intelligent agents continually update internal models to minimize discrepancies between expected and observed outcomes, reducing ``surprise'' or free energy~\cite{Friston2011}. This predictive perspective resonates deeply with our model. Our world model ($M_t^{\mathrm{wm}}$), alongside goal and reward components, continually refines predictions about future environmental states, enabling the agent to anticipate and adapt proactively. Decision-making, planning, and action selection then explicitly aim to reduce surprise by aligning internal expectations with observed reality, mirroring the Bayesian-brain perspective in a structured computational form.

\paragraph*{Biological Plausibility and Computational Flexibility.}
Throughout these theoretical connections, our framework prioritizes two guiding principles: biological plausibility and computational generality. While each submodule aligns clearly with neuroscientific analogues—memory with hippocampal-cortical interplay, emotion with limbic function, reasoning with prefrontal cortical circuits—these analogies inspire rather than rigidly constrain implementation. Our modules remain agnostic to specific computational realizations, easily accommodating neural networks, symbolic logic, probabilistic models, or hybrid methods. This openness preserves flexibility, enabling diverse implementations that remain faithful to core cognitive principles without artificial constraints.

By explicitly situating our framework among these influential theories, we achieve clarity on how each aspect of our design contributes distinctively and why integrating these aspects yields a powerful, flexible, and biologically inspired agent architecture. These connections not only clarify theoretical positioning but also serve as guideposts for future enhancements, ensuring our approach remains both grounded in rigorous science and open to ongoing innovation.

\section{Navigating This Book}
\label{sec:navigating_this_survey}

\lettrine[lines=3]{\initfamily\textcolor{darkgreen}{T}}{his book} is structured to provide a comprehensive, modular, and interdisciplinary examination of intelligent agents, drawing inspiration from cognitive science, neuroscience, and other disciplines to guide the next wave of advancements in AI. While many existing surveys \cite{duranteAGENTAISURVEYING,huangPositionPaperAgent2024,xiRisePotentialLarge2023,wangSurveyLargeLanguage2024,language-agent-tutorial,masterman2024landscape,guo2024large,zhang2024survey,yu2025trustagent} offer valuable insights into various aspects of agent research, we provide a detailed comparison of their focal points in Table \ref{tab:survey_compare}. Our work distinguishes itself by systematically comparing biological cognition with computational frameworks to identify synergies, gaps, and opportunities for innovation. By bridging these domains, we aim to provide a unique perspective that highlights not only where agents excel but also where significant advancements are needed to unlock their full potential.

\begin{table}[ht]
\small
\centering
\caption{Summary of existing reviews with different focal points. \(\bullet\) indicates primary focus while \(\circ\) indicates secondary or minor focus.}
\label{tab:survey_compare}
\resizebox{\textwidth}{!}{
\begin{tabular}{@{}l c c c c c c c c@{}}
\toprule
\textbf{Survey} & \textbf{Cognition} & \textbf{Memory} &  \textbf{World Model} &  \textbf{Reward} & \textbf{Action}  &\textbf{Self Evolve} & \textbf{MultiAgent} & \textbf{Safety} \\
\midrule
\rowcolor{LightMint}
\citet{zhang2024survey}   & \(\bullet\) & \(\bullet\) & \(\circ\) & \(\circ\) & \(\circ\) & \(\bullet\) & \(\circ\) & \(\circ\) \\
\citet{guo2024large}   & \(\bullet\) & \(\bullet\) & \(\circ\) & \(\circ\) & \(\circ\) &  \(\bullet\) & \(\bullet\) & \(\circ\) \\
\rowcolor{LightMint}
\citet{yu2025trustagent} & \(\bullet\)  & \(\bullet\)  & \(\circ\) & \(\circ\) & \(\bullet\)  &  \(\circ\) & \(\bullet\)  & \(\bullet\) \\
\citet{wangSurveyLargeLanguage2024}  & \(\bullet\) & \(\bullet\) & \(\circ\) & \(\circ\) & \(\bullet\) &  \(\circ\) & \(\bullet\) & \(\circ\) \\
\rowcolor{LightMint}
\citet{masterman2024landscape}     & \(\bullet\) & \(\bullet\) & \(\circ\) & \(\circ\) & \(\bullet\) &  \(\circ\) & \(\bullet\) & \(\circ\) \\
\citet{xiRisePotentialLarge2023}    & \(\bullet\) & \(\bullet\) & \(\circ\) & \(\circ\) & \(\bullet\) &  \(\bullet\) & \(\bullet\) & \(\bullet\) \\
\rowcolor{LightMint}
\citet{huangPositionPaperAgent2024} & \(\bullet\) & \(\bullet\) & \(\circ\) & \(\bullet\) & \(\bullet\) &  \(\bullet\)& \(\bullet\) & \(\bullet\) \\
\citet{duranteAGENTAISURVEYING}  & \(\bullet\) & \(\bullet\) & \(\circ\) & \(\bullet\) & \(\bullet\) &  \(\bullet\)& \(\bullet\) & \(\bullet\) \\
\rowcolor{LightMint}
\textbf{This Book} & \(\bullet\) & \(\bullet\) & \(\bullet\) & \(\bullet\) & \(\bullet\) &  \(\bullet\) & \(\bullet\)& \(\bullet\) \\

\bottomrule
\end{tabular}
}
\end{table}

The book is divided into four key parts: 
\begin{itemize}
    \item In \textbf{Part~\ref{part-agent}}: Modular Design of Intelligent Agents, we introduce the core modules of agents, including the cognition module, which serves as the ``brain'' of the agent; the perception systems for interpreting sensory input; as well as the action systems for interacting with the external world. Within the cognition system, we further discuss the memory, world modeling, emotion, goal, and reward systems, analyzing their current progress, limitations, and research challenges. 
    
    \item In \textbf{Part~\ref{part-enhance}}: Self-Enhancement in Intelligent Agents, we shift focus to the capability of agents to evolve and optimize themselves. We explore mechanisms like adaptive learning, self-reflection, and feedback-driven improvement, inspired by the human ability to grow and refine skills over time. This part also addresses the importance of dynamic memory systems and continuous knowledge integration for agents to remain relevant and effective in changing environments.
    

    \item In \textbf{Part~\ref{part-society}}: Collaborative and Evolutionary Intelligent Systems, we examine how agents interact with each other and their environments to solve complex, large-scale problems. We discuss multi-agent systems, highlighting their applications in fields such as robotics, medical systems and scientific discovery. This part explores multi-agent system topologies and agent protocol, tracing the evolution of communication and collaboration from static to dynamic frameworks. We align agents with human collaboration paradigms, examining how interaction patterns shape the co-evolution of intelligence and how multi-agent systems adapt their decision-making in various collaborative settings to solve complex challenges through collective intelligence.

    \item Finally, in \textbf{Part~\ref{part-safety}}: Building Safe and Beneficial AI, we provide a comprehensive analysis of the security landscape for LLM-based agents. We introduce a framework categorizing threats as intrinsic or extrinsic. Intrinsic vulnerabilities arise from within the agent's architecture: the core LLM ``brain'', and the perception and action modules that enable interactions with the world. Extrinsic risks stem from the agent's engagement with memory systems, other agents, and the broader environment. This part not only formalizes and analyzes these vulnerabilities, detailing specific attack vectors like jailbreaking and prompt injection, but also reviews a range of defense mechanisms. Moreover, we explore future directions, including superalignment techniques and the scaling law of AI safety---the interplay between capability and risk.
\end{itemize}
By weaving together these threads, we aim to provide a holistic perspective on the current state of intelligent agents and a forward-looking roadmap for their development. Our unique focus on integrating cognitive science insights with computational design principles positions this book as a foundational resource for researchers seeking to design agents that are not only powerful and efficient but also adaptive, ethical, and deeply aligned with the complexities of human society.

\part{Core Components of Intelligent Agents}
\label{part-agent}
\begin{figure*}[!ht]
    \centering
    \includegraphics[width=0.6\textwidth]{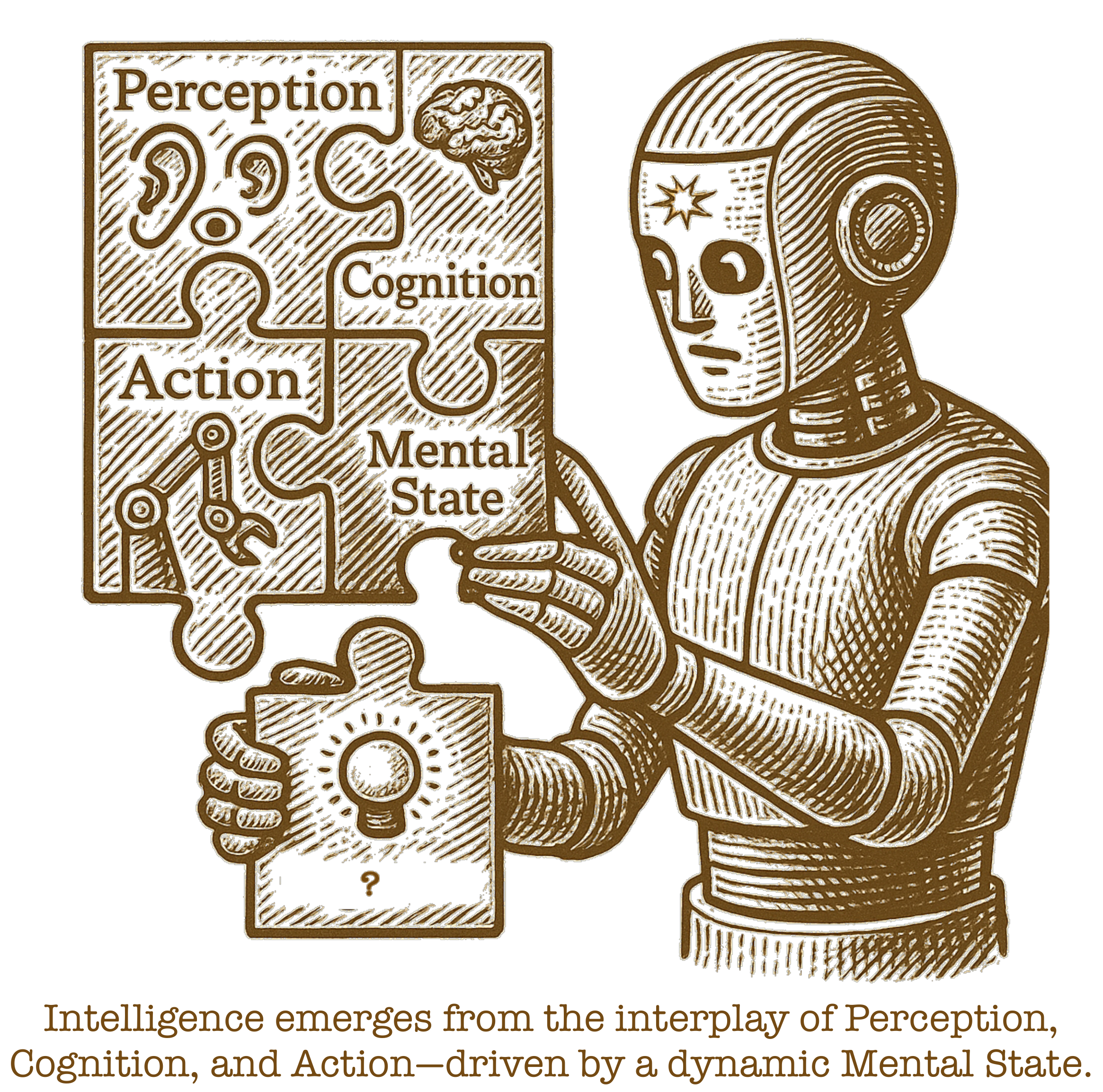}
    \label{fig:part1}
\end{figure*}

What makes an intelligent agent more than just a clever model? The difference lies not in scale but in structure—in the presence of a mind that can see, feel, remember, plan, and act. This part of the book dives beneath the surface of modern AI systems to examine the internal architecture of intelligent agents, not as abstract math, but as working minds. Minds that perceive their world, carry memory through time, adapt to change, and chase goals beyond the next token.

At the heart of our framework is a living loop: perception feeds cognition, cognition selects action, and action reshapes perception. But this loop is not empty. Inside cognition, we find a constellation of faculties—memory, world model, emotion, goals, and learning—each contributing to the agent's ability to adapt and pursue long-term purpose. These are not accessories but necessities, as real to an agent as lungs and eyes are to a body.

In this part, we explore each of these faculties in turn. We begin with cognition, the control center that organizes thought, plans, and decisions. From there, we trace the roles of memory, which binds past to present; the world model, which imagines futures; and the reward system, which teaches from outcomes. We also examine emotion, not as sentiment but as a powerful biasing force in cognition and behavior. The chapters then lead outward to perception, the gateway from world to mind, and action, the channel through which mind shapes world.
Each module is inspired by how biological systems—especially the human brain—solve these problems, but is cast in a form suitable for building digital minds. This part lays the groundwork for everything that follows. Before we can teach agents to collaborate, generalize, or reason morally, we must understand what makes them tick—what internal machinery gives rise to intelligent behavior.
These are the core components of a Foundation Agent. We now open the casing to see how each part fits, functions, and learns.
\chapter{Cognition}
\label{ch:cognition}

\begin{figure}[!ht]
\centering
\footnotesize
    \begin{forest}
        for tree={
            forked edges,
            draw,
            rounded corners,
            node options={align=center,},
            s sep=6pt,
            calign=center,
            grow=east,
            reversed=true,
            anchor=base west,
            parent anchor=east,
            child anchor=west,
            base=left,
            font=\small,
            minimum width=2.5em,
          },
          where level=1{text width=5em,fill=customblue!50}{},
          where level=2{text width=5em,fill=customgreen!50}{},
        [Cognition, fill=gray!20
            [Learning, text width=50pt, for tree={
                calign=child edge, calign child=(n_children()+1)/2,
            }
                [Space, text width=50pt, for tree={
                    calign=child edge, calign child=(n_children()+1)/2,
                }
                    [Full, text width=40pt,
                        [
                            {
                                SFT~\cite{wei2021finetuned}, PEFT~\cite{houlsby2019parameter}, \\ RLHF~\cite{ouyang2022training}, ReFT~\cite{luong2024reft}, \\
                                Agentic Models~\cite{song2025r1}
                            }, text width=205pt
                        ]
                    ]
                    [Partial, text width=40pt,
                        [
                            {
                                CoT~\cite{wei2022chain}, Voyager~\cite{wang2023voyager}, \\
                                Reflexion~\cite{Shinn2023ReflexionLA}, ActRe~\cite{yang2024react}, \\
                                Generative Agents~\cite{park2023generative}
                            }, text width=205pt
                        ]
                    ]
                ]
                [Objective, text width=50pt, for tree={
                    calign=child edge, calign child=(n_children()+1)/2,
                }
                    [Perception, text width=40pt,
                        [
                            {
                                CLIP~\cite{alec2021clip}, LLaVA~\cite{liu2023llava}, \\
                                CogVLM~\cite{wang2025cogvlm}, Qwen-Audio~\cite{chu2024qwen2}, \\
                                R1-Searcher~\cite{song2025r1, song2025r1plusplus}, Search-R1~\cite{jin2025searchr1}
                            }, text width=205pt
                        ]
                    ]
                    [Reasoning, text width=40pt,
                        [
                            {
                                SKY-32B~\cite{sky_t1_2025}, Open Thoughts~\cite{open_thoughts}, \\
                                LIMO~\cite{ye2025limo}, STaR~\cite{zelikman2022star}, ReST~\cite{gulcehre2023rest}, \\
                                OpenR~\cite{wang2024openr}, LLaMA-Berry~\cite{zhang2024llamaberry}, \\
                                RAGEN~\cite{wang2025ragen}, OpenR1~\cite{2024openr1}
                            }, text width=205pt
                        ]
                    ]
                    [World, text width=40pt,
                        [
                            {
                                Inner Monologue~\cite{huang2023inner}, DESP~\cite{wang2023describe}, \\ 
                                Self-refine~\cite{madaan2024self-refine}, CRITIC~\cite{goucritic}, \\
                                Reflexion~\cite{Shinn2023ReflexionLA}, ExpeL~\cite{zhao2024expel}
                            }, text width=205pt
                        ]
                    ]
                ]
            ]
            [Reasoning, text width=50pt, for tree={
                calign=child edge, calign child=(n_children()+1)/2,
            }
                [Structured, text width=50pt,
                    [Dynamic, text width=40pt
                        [
                            {
                                ReAct~\cite{yao2022react}, MCoT~\cite{yang2024markov}, \\
                                ToT~\cite{yaotree}, LATS~\cite{zhou2024language}, RAP~\cite{hao2023reasoning}, \\
                                GoT~\cite{Besta2023GraphOT}, PoT~\cite{zhang2024path}, DoT~\cite{Zhang2024OnTD}
                            }, text width=205pt
                        ]
                    ]
                    [Static, text width=40pt
                        [
                            {
                                Self-Consistency~\cite{wang2023selfconsistency}, \\
                                Self-refine~\cite{madaan2024self-refine}, PHP~\cite{zheng2023progressive}, \\
                                Self-Verification~\cite{stechly2024self}, CoVe~\cite{dhuliawala2024chain}
                            }, text width=205pt
                        ]
                    ]
                    [Domain, text width=40pt
                        [
                            {
                                MathPrompter~\cite{imani2023mathprompter}, PedCoT~\cite{jiang2024pedcot}, \\
                                Physics Reasoner~\cite{pang2025physics}
                            }, text width=205pt
                        ]
                    ]
                ]
                [Unstructured, text width=50pt,
                    [Prompt, text width=40pt
                        [
                            {
                                CoT~\cite{wei2022chain}, Step-Back,~\cite{zhengtake}\\
                                Ask Me Anything~\cite{arora2022ask}, CoK~\cite{li2023chain}, SEK~\cite{fan2024self}
                            }, text width=205pt
                        ]
                    ]
                    [Model, text width=40pt
                        [
                            {
                                DeepSeek-R1~\cite{guo2025deepseek}, \\
                                Claude 3.7 Sonnet~\cite{anthropic2023claude}, o1 ~\cite{jaech2024openai}\\
                            }, text width=205pt
                        ]
                    ]
                    [Implicit, text width=40pt
                        [
                            {
                                Quiet-STaR~\cite{zelikman2024quiet}, Coconut~\cite{hao2024training} \\
                            }, text width=205pt
                        ]
                    ]
                ]
                [Planning, text width=50pt,
                    [
                        {
                            DEPS~\cite{wang2023describe}, ProgPrompt~\cite{singh2023progprompt}, \\
                            ADaPT~\cite{prasad2023adapt}, ToT~\cite{yaotree}, RAP~\cite{hao2023reasoning}, \\
                            TravelPlanner~\cite{xie2024travelplanner}, PDDL~\cite{mcdermott1998pddl}, \\
                            Mind2Web~\cite{deng2024mind2web}
                        }, text width=260pt
                    ]
                ]
            ]
        ]
    \end{forest}
    \caption{A taxonomy of research on cognition covering different learning and reasoning paradigms.}
    \label{fig:cognition_system}
\end{figure}

\lettrine[lines=3]{\initfamily\textcolor{darkgreen}{H}}{uman cognition} is a dynamic system of information processing that supports learning, memory, reasoning, and goal-directed behavior. At its core lies the concept of a mental state—a structured, internal representation encompassing beliefs, knowledge, context, and intent. This mental state provides the substrate over which cognitive operations act, enabling humans to flexibly adapt to new situations, abstract over experiences, and make context-sensitive decisions. Decades of cognitive neuroscience have revealed a modular yet integrated architecture underlying these capabilities: perception systems that translate sensory input into internal symbols; memory systems that encode and retrieve experience; reasoning systems that formulate decisions; action systems that translate decisions into environmental interactions; reward signals that guide behavior through reinforcement; and emotion systems that modulate attention and resource allocation \cite{fodor1983modularity, badre2008cognitive}.

LLM-based agents offer a new computational paradigm that begins to approximate aspects of this architecture. These agents construct and manipulate internal mental states—though often implicitly—via large-scale hidden representations, memory buffers, or intermediate reasoning steps. Learning arises through gradient-based updates or contextual inferences, while reasoning often involves generating or selecting structured hypotheses, subgoals, or actions based on current context. Despite differences from biological systems, the underlying principles—modular operations on internal representations, guided by experience and adaptive goals—recur. In this chapter, we first examine \textbf{Learning} as the process by which agents improve or restructure their internal state. We then turn to \textbf{Reasoning}, understood as the agent's internal deliberation process that selects or constructs actions through search, generation, or inference. Figure~\ref{fig:cognition_system} shows an overview of selected research works on different learning and reasoning paradigms, which we will discuss in more details next.

\section{Learning}
\label{sec:cog-learning}

\lettrine[lines=3]{\initfamily\textcolor{darkgreen}{{L}}}{earning} is the mechanism by which an intelligent agent improves its performance over time through experience. It allows the agent to adjust its internal parameters—such as its models of the world, policy for action, or strategy for reasoning—based on observations and outcomes. From supervised learning to reinforcement learning, from variational inference to meta-learning, diverse paradigms have been proposed to formalize how an agent can acquire knowledge, skills, or behavior through interaction with data or environments. Despite their apparent differences, these learning approaches share a common goal: optimizing an objective that aligns the agent's internal process with its external goals. This section presents a unified perspective on learning, bridging multiple paradigms under a general formulation, and setting the stage for how learning principles can be integrated into agentic reasoning and decision-making.

\subsection{A Unified Formulation of Learning}
\label{subsec:unified-learning}

The previous chapter defined learning in the \emph{Foundation Agents} loop by the update rule \(M_t=L(M_{t-1},a_{t-1},o_t)\).  We now replace the black-box placeholder \(L\) with a single objective that subsumes supervised fitting, unsupervised representation learning, reinforcement learning, active inference, meta-learning, and continual adaptation.  The formulation reveals that every algorithm negotiates three forces simultaneously: fidelity to experience, parsimony of internal representation, and foresight across multiple temporal horizons.

\begin{definition}[\textbf{Unified Learning Framework for Foundation Agents}]
\label{def:unified-learning}
Learning is the process by which an agent minimises expected prediction error (\emph{free energy}) across a hierarchy of time-scales while constraining the complexity of its internal model.
\[
M_t=\arg\min_{M}\sum_{i}w_i\,\mathbb{E}_{\tau\sim\mathcal{T}_i}\Big[-\ln P\!\big(o_\tau,r_{\tau-1}\mid a_{\tau-1},M\big)+\lambda\,\mathcal{C}\!\big(a_{\tau-1};M\big)\Big]+\beta\,\mathcal{D}\!\big(M,M_{t-1}\big).
\]

\paragraph*{Experience term:} The quantity \(-\ln P(o_\tau,r_{\tau-1}\mid a_{\tau-1}, M)\) measures how surprising the joint sensory–reward signal is under the current model—minimizing it reduces prediction error.  Reward is treated as a sensory variable, unifying prediction and control under a single statistical drive; labels, pixels, proprioception, and returns therefore enter through the same doorway.

\paragraph*{Action cost:} The optional component \(\lambda\,\mathcal{C}(a_{\tau-1};M)\) prices behaviour.  Setting \(\mathcal{C}\) to negative policy entropy encourages exploration; choosing information gain yields curiosity; \(\lambda=0\) collapses the loss to passive learning.

\paragraph*{Temporal mixture:} Weights \(w_i\) mix expectations over horizons \(\mathcal{T}_i\) (e.g., milliseconds, episodes, lifetime).  Larger weights emphasise long-term consistency, smaller weights permit rapid adaptation; the shared parameters must serve every horizon.

\paragraph*{Complexity regulariser:} The divergence \(\mathcal{D}(M, M_{t-1})\) tethers the update to the previous self and may be instantiated as KL, \(\ell_2\), or Wasserstein distance; \(\beta\) sets the plasticity–stability trade-off.  In variational settings one can separate stability and prior-complexity terms, but for online agents these roles usually coincide.

\paragraph*{Computational reality:} Agents approximate the ideal objective with stochastic gradient descent, variational EM, policy gradients, evolutionary strategies, or any solver allowed by their resource budget.

\end{definition}

\begin{figure}[!ht]
\centering
    \includegraphics[width=0.8\columnwidth]{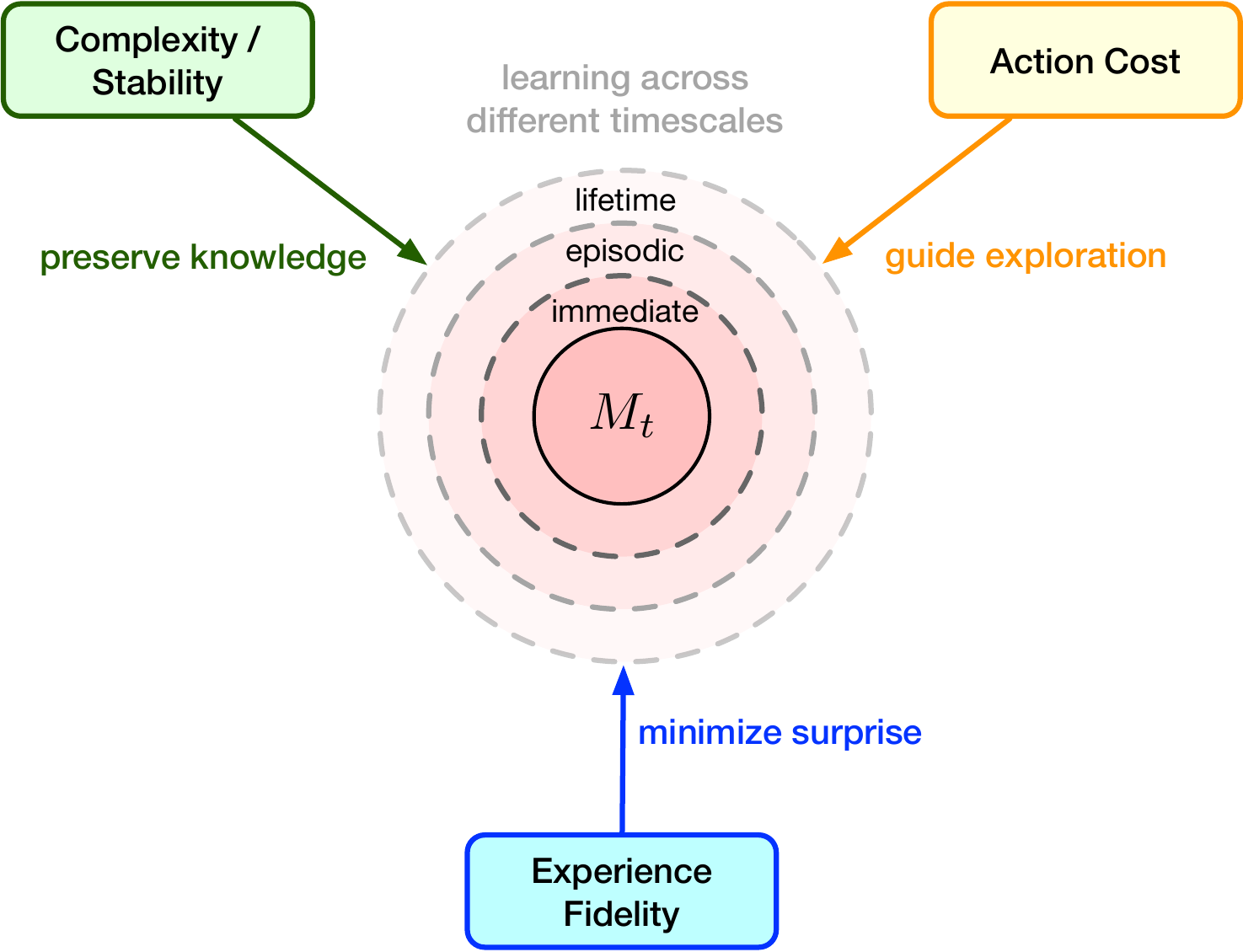}
    \caption{Learning as optimisation under three competing forces. The mental state $M_t$ evolves to balance experience fidelity (minimising prediction error), action costs (shaping exploration), and complexity constraints (preserving acquired knowledge). The concentric arcs represent integration across temporal scales, from immediate sensory processing through episodic memory to lifetime knowledge. The equilibrium emerges from jointly satisfying all constraints across all timescales.}
\label{fig:learning-forces}
\end{figure}

This equation instantiates \(L\) as a concrete optimisation problem inside the Foundation Agents framework.
Figure~\ref{fig:learning-forces} offers an intuitive visualisation of this framework. It shows how the agent's mental state $M_t$ evolves by balancing three fundamental forces: fidelity to experience, behavioural shaping via action cost, and preservation of accumulated knowledge through regularisation. These forces interact across multiple temporal scales, from immediate adaptation to lifelong consistency, forming the core dynamics of learning in both biological and artificial systems.

To illustrate the breadth and unifying power of the above formulation, Table~\ref{tab:unified-paradigm-comparison} provides concrete instantiations across major learning paradigms. Each paradigm corresponds to a different way of instantiating the likelihood term, temporal weighting, behavioural cost, and complexity regularisation. Despite their diversity, they all conform to the same abstract objective, validating the generality of Definition~\ref{def:unified-learning}.

\begin{table}[ht]
\centering
\caption{Common instantiations of the unified objective.  Columns spell out the usual choice of likelihood (experience term), temporal weighting, behavioural cost $\mathcal{C}$, and regulariser $\mathcal{D}$.  Adjusting only these ingredients recovers a broad spectrum of learning paradigms without altering the surrounding free-energy skeleton.}
\small
\setlength{\tabcolsep}{4pt}
\begin{tabular}{@{}p{3.1cm}p{4.9cm}p{2.5cm}p{1.9cm}p{3.0cm}@{}}
\toprule
\textbf{Paradigm} & \textbf{Experience likelihood $-\ln P(\cdot)$} & \textbf{Time-scale weights $\{w_i\}$} & \boldmath$\mathcal{C}$ & \boldmath$\mathcal{D}$ \\ \midrule
\rowcolor{LightMint}
Supervised classification & $P(y\!\mid\!x;M)$ (cross-entropy) & $w_{\text{inst}}{=}1$ & $0$ & $\ell_2$ decay or KL prior \\
Unsupervised (VAE / Diffusion) & $P(x\!\mid\!z;M)$ plus latent prior & $w_{\text{inst}}{=}1$ & $0$ & KL on latents \& weights \\
\rowcolor{LightMint}
Self-supervised contrastive & $\exp(\operatorname{sim}(h_i,h_j)/\tau)$ (InfoNCE) & $w_{\text{inst}}{=}1$ & $0$ & $\ell_2$ decay \\
GAN / Adversarial & $P(x)\!\propto\!\exp(-D_\phi(x))$ & $w_{\text{inst}}{=}1$ & $0$ & Jensen–Shannon KL \\
\rowcolor{LightMint}
Policy-gradient RL & $P(o,r)\!\propto\!\exp(r/\alpha)$ & $w_{\text{inst}}{=}1$ & Entropy bonus & Trust-region KL \\
Curiosity-driven RL & idem, with predicted future $o$ & $w_{\text{inst}}{=}1$ & Info-gain or RND cost & Entropy-weighted KL \\
\rowcolor{LightMint}
Active inference / active learning & same likelihood as RL & $w_{\text{inst}},w_{\text{hor}}\!\neq\!0$ & Info-gain cost & Variational KL \\
Continual learning & any of the above & task-dependent $w_i$ & $0$ & Fisher-weighted KL to snapshot \\
\rowcolor{LightMint}
Bayesian online update & same as base task & sliding $w_{\text{inst}}$ & $0$ & Online KL to prior \\
Meta-learning (outer loop) & episode return or task loss & $w_{\text{inst}},w_{\text{epis}}$ & $0$ & KL across tasks \\
\rowcolor{LightMint}
Knowledge-constrained & likelihood penalised by rules & $w_{\text{inst}}{=}1$ & $0$ & KL + rule penalty \\
Skill / memory write & likelihood of recall success & $w_{\text{epis}}$ or $w_{\text{life}}$ & energy or write cost & $\ell_2$ or sparse prior \\ \bottomrule
\end{tabular}

\label{tab:unified-paradigm-comparison}
\end{table}

\subsection{Learning Across Mental State Components}
\label{subsec:learning-across-mental-states}

Learning represents the fundamental process through which \textbf{intelligent agents transform experiences into knowledge within their mental states}. This transformation occurs across different cognitive spaces, from holistic updates across the full mental state to refinement of specific cognitive components. The scope of learning encompasses remarkable capacities that serve different objectives: enhancing perceptual understanding, improving reasoning capabilities, and developing richer world understanding.
Recent developments in LLM-based agents demonstrate how different learning strategies update specific components of the mental state. 
Table~\ref{tab:learning-summary} summarizes a selection of representative methods and highlights the cognitive subsystems they primarily affect—ranging from memory updates in systems like Voyager and Generative Agents to reward modeling in RewardAgent and Text2Reward, or world model construction in WebDreamer and AutoManual. This breakdown helps clarify how learning is distributed across the broader architecture of an agent.

Human learning operates across multiple spaces and objectives through the brain's adaptable neural networks. The brain coordinates learning across its entire network through integrated systems: the \emph{hippocampus} facilitates rapid encoding of episodic experiences, the \emph{cerebellum} supports supervised learning for precise motor skills, the \emph{basal ganglia} enable reinforcement learning through dopaminergic reward signals, and \emph{cortical regions} facilitate unsupervised pattern extraction \cite{doya2000complementary}. At more focused levels, specific neural circuits can undergo targeted adaptation, allowing for specialized skill development and knowledge acquisition. These systems work together on different timescales, ranging from immediate responses to lifelong development, while being influenced by factors like attention, emotions, and social environment \cite{badre2008cognitive}.

LLM agents, while fundamentally different in architecture, implement analogous learning processes across their mental state spaces. At the comprehensive level, they acquire broad knowledge through pre-training on massive datasets, demonstrating a form of unsupervised learning. At more focused levels, they refine specific capabilities through parameter-updating mechanisms like supervised fine-tuning and reinforcement learning. Uniquely, they also demonstrate in-context learning capabilities, adapting to novel tasks without parameter changes by leveraging context within their attention window: a capability that mirrors aspects of human working memory but operates through fundamentally different mechanisms.

The comparison between human and artificial learning systems provides valuable insights for developing more capable, adaptive agents. Human learning demonstrates notable characteristics in efficiency, contextualization, and integration with emotional systems, while LLM-based approaches show distinct capabilities in processing large datasets, representing formal knowledge, and synthesizing information across domains. These complementary strengths suggest productive directions for research. As we explore the foundations of learning, we first examine the spaces where learning occurs within mental states, followed by an analysis of the specific objectives that drive learning processes.

\begin{table}[ht]
\small
\centering
\caption{Summary of learning methods with different state modifications. \(\bullet\) indicates primary impact while \(\circ\) indicates secondary or no direct impact.}
\label{tab:learning-summary}
\begin{tabular}{@{}l c c c c c c @{}}
\toprule
\textbf{Method} & \textbf{Model} & \textbf{Perception} & \textbf{Reasoning} & \textbf{Memory} & \textbf{Reward} & \textbf{World Model} \\
\midrule
\rowcolor{LightMint}
Voyager \cite{wang2023voyager}  & \(\circ\) & \(\circ\) & \(\circ\) & \(\bullet\) & \(\circ\) &  \(\circ\) \\
Generative Agents \cite{park2023generative}  & \(\circ\) & \(\circ\) & \(\circ\) & \(\bullet\) & \(\circ\) &  \(\circ\) \\
\rowcolor{LightMint}
Learn-by-interact \cite{su2025learn} & \(\bullet\) & \(\circ\) & \(\circ\) & \(\bullet\) & \(\circ\) &  \(\circ\) \\
RAGEN \cite{wang2025ragen}  & \(\bullet\) & \(\circ\) & \(\bullet\) & \(\circ\) & \(\bullet\) &  \(\circ\) \\
\rowcolor{LightMint}
DigiRL \cite{bai2024digirl}  & \(\bullet\) & \(\circ\) & \(\bullet\) & \(\circ\) & \(\bullet\) &  \(\circ\) \\
R1-Searcher \cite{song2025r1}  & \(\bullet\) & \(\bullet\) & \(\bullet\) &  \(\circ\) & \(\bullet\) &  \(\circ\) \\
\rowcolor{LightMint}
RewardAgent \cite{peng2025agenticrewardmodelingintegrating} & \(\bullet\) & \(\circ\) & \(\circ\) & \(\circ\) & \(\bullet\)&  \(\circ\)\\
Text2Reward \cite{xie2023text2reward} & \(\circ\) & \(\circ\) & \(\circ\) & \(\circ\) & \(\bullet\)&  \(\circ\)\\
\rowcolor{LightMint}
ARAMP \cite{chen2025armap}  & \(\bullet\) & \(\circ\) & \(\circ\) & \(\circ\) & \(\bullet\)&  \(\circ\)\\
ActRe \cite{yang2024react} & \(\bullet\) & \(\circ\) & \(\bullet\) & \(\circ\) & \(\circ\) &  \(\bullet\)\\
\rowcolor{LightMint}
WebDreamer \cite{gu2024your}  & \(\circ\) & \(\circ\) & \(\circ\) & \(\circ\) & \(\circ\) &  \(\bullet\)\\
RAP \cite{hao2023reasoning}  & \(\circ\) & \(\circ\) & \(\circ\) & \(\circ\) & \(\circ\) &  \(\bullet\)\\
\rowcolor{LightMint}
AutoManual \cite{chen2024automanual} & \(\circ\) & \(\circ\) & \(\circ\) & \(\bullet\) & \(\circ\)&  \(\bullet\)\\
\bottomrule
\end{tabular}
\end{table}

\subsection{Learning Space}
\label{subsec:learning-space}

The learning approaches in LLM agents represent a structured, data-driven paradigm in contrast to the exploratory, emotionally-driven learning observed in humans. While human learning often involves active curiosity, motivation, and emotional reinforcement, LLM-based agents typically learn through more formalized processes, such as parameter updates during training or structured memory formation during exploration. Current agent architectures attempt to bridge this gap by implementing mechanisms that simulate aspects of human learning while leveraging the strengths of computational systems.

Learning within an intelligent agent occurs across different cognitive spaces, encompassing both large-scale model updates and more localized changes to modularized mental states \(M\). In systems where the model is the only trainable component, the model parameters \(\theta\) can be viewed as constituting or encoding the entire mental state. More generally, the mental state can include a combination of subsystems:
\begin{equation}
    M = \{M^{\theta}, M^{mem}, M^{wm}, M^{emo}, M^{goal}, M^{rew}\}
\end{equation}
where \(M^{\theta}\) denotes the core model parameters, \(M^{mem}\) represents memory, \(M^{wm}\) denotes the world model, \(M^{emo}\) indicates emotional state, \(M^{goal}\) represents goals, and \(M^{rew}\) represents reward signals.

\begin{figure}[!ht]
\centering
    \includegraphics[width=0.8\columnwidth]{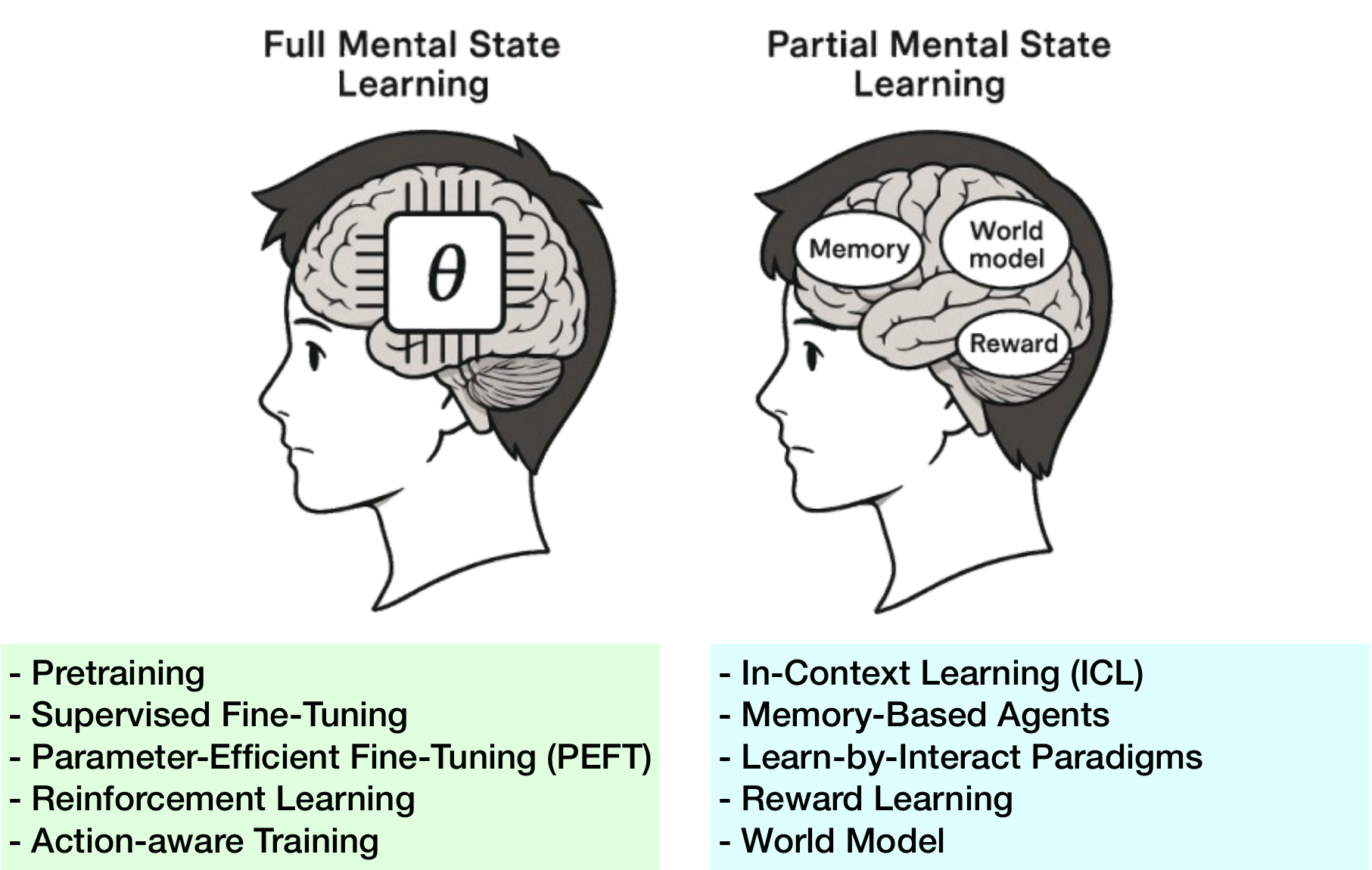}
    \caption{Comparison between full mental state learning and partial mental state learning in intelligent agents. Full mental state learning modifies the core model parameters $M^\theta$, leading to holistic capability updates across all subsystems. This includes techniques like supervised fine-tuning, parameter-efficient tuning, and reinforcement learning with alignment. In contrast, partial mental state learning targets specific components such as memory, reward models, or world models, often without changing model parameters. This includes in-context learning, dynamic memory updates, and environment-driven adaptation, enabling efficient and modular agent improvement.}
\label{fig:learning-spaces}
\end{figure}

Modifications to \(M^{\theta}\)—the core model—often lead to holistic changes that affect all components of the mental state. In contrast, more targeted updates to memory, world model, or reward components allow the agent to adapt specific subsystems while preserving general capabilities. For instance, learning experiences and skills from the environment primarily influence memory, while leveraging the LLM's inherent predictive capabilities enhances the world model. The distinction between these two paradigms—full mental state learning and partial component-level adaptation—is illustrated in Figure~\ref{fig:learning-spaces}.

\paragraph*{Full Mental State Learning}
Full mental state learning enhances the capabilities of an agent through comprehensive modifications to \(M^{\theta}\), which in turn influences all components of \(M\). This process begins with pre-training, which establishes the foundation of language models by acquiring vast world knowledge, analogous to how human babies absorb environmental information during development, though in a more structured and extensive manner.

\emph{Post-training} techniques represent the cornerstone for advancing agent capabilities. Similar to how human brains are shaped by education, these techniques, while affecting the entire model, can emphasize different aspects of cognitive development. Specifically, various forms of tuning-based learning enable agents to acquire domain-specific knowledge and logical reasoning capabilities. \emph{Supervised Fine-Tuning (SFT)} \cite{wei2021finetuned} serves as the fundamental approach where models learn from human-labeled examples, encoding knowledge directly into the model's weights. For computational efficiency, \emph{Parameter-Efficient Fine-Tuning (PEFT)} methods have emerged. Adapter-BERT \cite{houlsby2019parameter} introduced modular designs that adapt models to downstream tasks without modifying all parameters, while Low-Rank Adaptation (LoRA) \cite{hu2021lora} achieves similar results by decomposing weight updates into low-rank matrices, adjusting only a small subset of effective parameters.

Some agent capabilities are closely connected to how well they align with human preferences, with \emph{alignment-based learning} approaches modifying \(M^{\theta}\) to reshape aspects of the agent's underlying representations. Reinforcement learning from human feedback (RLHF) \cite{ziegler2019fine} aligns models with human values by training a reward model on comparative judgments and using this to guide policy optimization. InstructGPT \cite{ouyang2022training} demonstrated how this approach could dramatically improve consistency with user intent across diverse tasks. Direct Preference Optimization (DPO) \cite{rafailov2023direct} has further simplified this process by reformulating it as direct preference learning without explicit reward modeling, maintaining alignment quality while reducing computational complexity.

\emph{Reinforcement learning (RL)} presents a promising pathway for specialized learning in specific environments. RL has shown particular promise in enhancing reasoning capabilities, essentially enabling the agent to refine its internal thinking processes. Foundational works such as Reinforcement Fine-Tuning (ReFT) \cite{luong2024reft} enhance reasoning through fine-tuning with automatically sampled reasoning paths under online reinforcement learning rewards. DeepSeek-R1 \cite{guo2025deepseek} advances this approach through rule-based rewards and Group Relative Policy Optimization (GRPO) \cite{shao2024dsmath}, while Kimi k1.5 \cite{kimiteam2025kimik15} combines contextual reinforcement learning with optimized chain-of-thought techniques to improve both planning processes and inference efficiency. In specific environments, modifying models to enhance agents' understanding of actions and external environments has proven effective, as demonstrated by DigiRL \cite{bai2024digirl}, which implements a two-stage reinforcement learning approach enabling agents to perform diverse commands on real-world Android device simulators.

Recent works have attempted to integrate \emph{agent action spaces} directly into model training \cite{song2025r1, jin2025searchr1, sun2025simpledeepsearcher,qi2025webrl,cao2025skyrl, chen2025atlasagenttuninglearning}, enabling learning of appropriate actions for different states through RL or SFT methods. This integration fundamentally affects the agent's memory, reward understanding, and world model comprehension, pointing toward a promising direction for the emergence of agentic models.

\paragraph*{Partial Mental State Learning}
While full mental state learning through updates to \(M^{\theta}\) provides comprehensive capability updates, learning focused on particular components of \(M\) represents another essential and often more efficient approach. Such partial mental state learning can be achieved either through targeted model updates or through in-context adaptation without parameter changes.

\emph{In-Context Learning (ICL)} illustrates how agents can effectively modify specific mental state components without modifying the underlying model. This mechanism allows agents to adapt to new tasks by leveraging examples or instructions within their context window, paralleling human working memory's role in rapid task adaptation. \emph{Chain-of-Thought (CoT)} \cite{wei2022chain} demonstrates the effectiveness of this approach, showing how agents can enhance specific cognitive capabilities while maintaining their base model parameters unchanged.

The feasibility of partial mental state learning is evidenced through various approaches targeting different components such as memory (\(M^{mem}\)), reward (\(M^{rew}\)), and world model (\(M^{wm}\)). Through normal communication and social interaction, Generative Agents \cite{park2023generative} demonstrate how agents can accumulate and replay memories, extracting high-level insights to guide dynamic behavior planning. In environmental interaction scenarios, Voyager \cite{wang2023voyager} showcases how agents can continuously update their \emph{skill library} through direct engagement with the Minecraft environment, accumulating procedural knowledge without model retraining. Mem0 \cite{chhikara2025mem0} provides agents with persistent and efficient \emph{long-term memory} through scalable dynamic memory management and graph-based memory representations, significantly enhancing their ability to handle complex, multi-session tasks.

\emph{Learn-by-Interact} \cite{su2025learn} further extends this approach by synthesizing experiential data through direct environmental interaction, eliminating the need for manual annotation or reinforcement learning frameworks. Additionally, agents can learn from their mistakes and improve through reflection, as demonstrated by \emph{Reflexion} \cite{Shinn2023ReflexionLA}, which guides agents' future thinking and actions by obtaining textual feedback from repeated trial and error experiences. Further, KnowSelf \cite{qiao2025agentic} introduces knowledgeable self-awareness, enabling agents to intelligently assess whether external knowledge is needed or to self-correct based on specific contexts, leading to more efficient and strategic planning.

Modifications to reward and world models provide another example of partial mental state learning. ARMAP \cite{chen2025armap} refines environmental reward models by distilling them from agent action trajectories, providing a foundation for further learning. AutoMC \cite{li2024auto} constructs dense reward models through environmental exploration to support agent behavior. Meanwhile, \cite{gu2024your} explicitly leverages LLMs as world models to predict the impact of future actions, effectively modifying the agent's world understanding (\(M^{wm}\)). ActRe \cite{yang2024react} builds upon the language model's inherent world understanding to construct tasks from trajectories, enhancing the agent's capabilities as both a world model and reasoning engine through iterative training.

\subsection{Learning Objective}
\label{subsec:learning-objective}

\begin{figure}[!ht]
\centering
    \includegraphics[width=0.8\columnwidth]{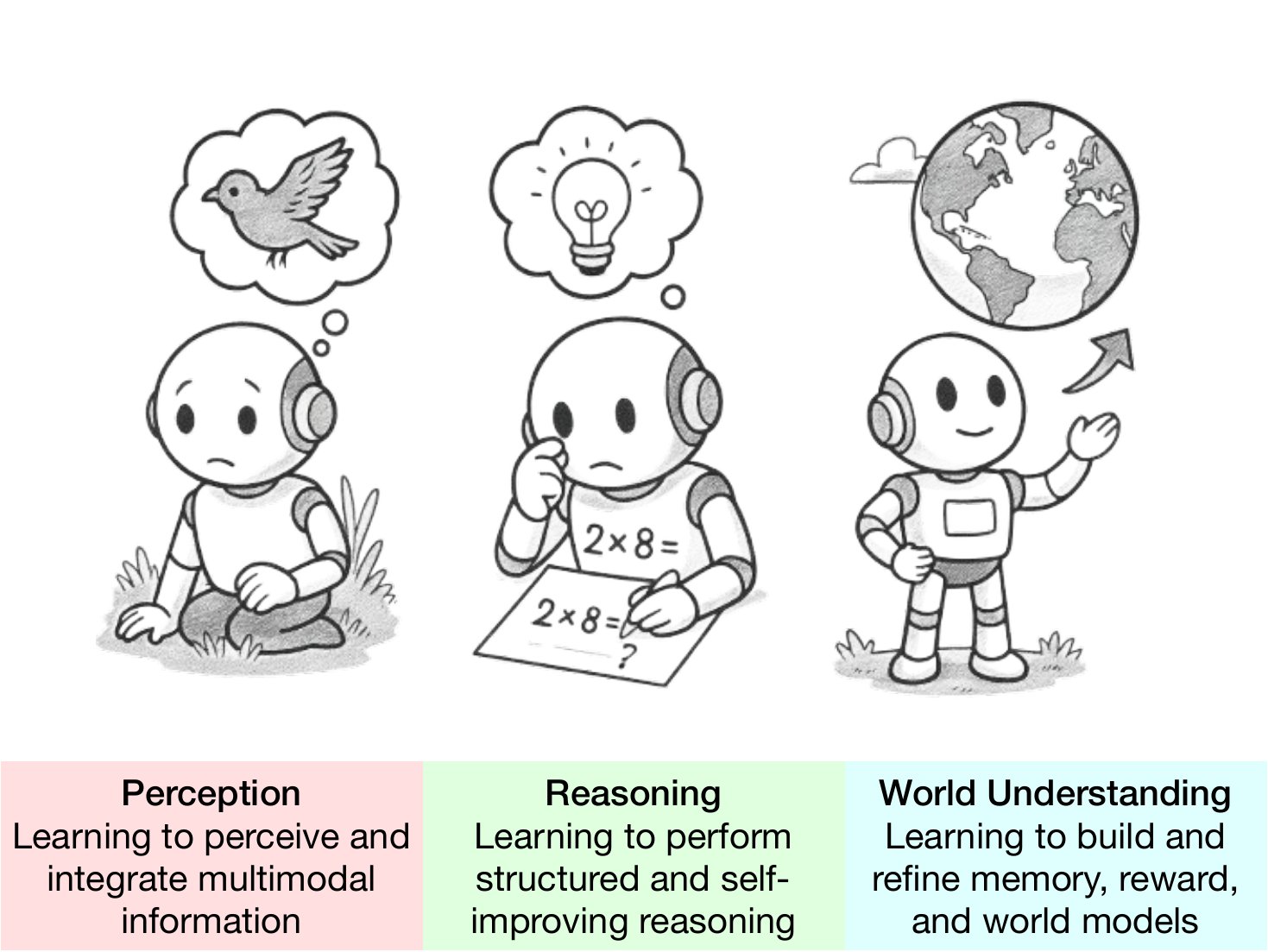}
    \caption{Three core learning objectives in intelligent agents. From left to right: \textbf{1)} Learning to perceive and integrate multimodal information from the environment enhances the agent's sensory and retrieval capabilities; \textbf{2)}	Learning to perform structured and self-improving reasoning strengthens the agent's cognitive abilities through data-driven, reinforcement, and self-evolutionary methods; \textbf{3)} Learning to build and refine memory, reward understanding, and world models enables agents to interpret their environment, simulate outcomes, and act more effectively.}
\label{fig:learning-objectives}
\end{figure}

The learning process of intelligent agents manifests across all aspects of their interaction with the environment. At the input level, agents learn to better perceive and parse environmental information; at the processing level, agents learn how to conduct effective reasoning based on existing knowledge or reasoning capabilities; at the comprehension level, agents form and optimize their understanding of the world through continuous interaction. This multi-level learning objective framework enables agents to evolve continuously across different dimensions, allowing them to better handle complex and dynamic task environments. Figure~\ref{fig:learning-objectives} shows an intuitive overview.

\paragraph*{Learning for Better Perception}
The ability to effectively perceive and process information from the environment is fundamental to agent intelligence. To enhance perceptual capabilities, agents employ two primary learning approaches: expanding multimodal perception and leveraging retrieval mechanisms.

\emph{Multimodal perception learning} enables agents to process and integrate diverse sensory inputs, similar to human multi-sensory integration but unconstrained by biological limitations. This capability has evolved significantly through advances like CLIP \cite{alec2021clip}, which pioneered the alignment of visual and linguistic representations in shared embedding spaces. Building on this foundation, models like LLaVA \cite{liu2023llava} enhanced visual perception by training specialized projectors on image-text pairs, while CogVLM \cite{wang2025cogvlm} advanced visual reasoning through unified representational architectures.
The expansion of perceptual modalities continues across multiple sensory domains. In audio processing, Qwen-Audio \cite{chu2024qwen2} demonstrates the unified encoding of diverse acoustic information, from speech to environmental sounds. Recent work by \cite{fu2024touch} has even ventured into tactile perception, developing datasets that align touch, vision, and language representations. These advances enable agents to engage more comprehensively with both physical and digital environments.

Agents also learn to enhance their observational capabilities through \emph{retrieval} mechanisms. Unlike human perception, which is constrained by immediate sensory input, agents can learn to access and integrate information from vast external knowledge repositories. Retrieval-augmented approaches like RAG \cite{gupta2024comprehensive, guan2024bsharedrag, jiang2025deepretrievalhackingrealsearch} enhance perceptual understanding by connecting immediate observations with relevant stored knowledge. 
Recent work on retrieval-based agents demonstrates the potential for enhancing \emph{active information acquisition} capabilities. Search-o1 \cite{li2025searcho1} guides reasoning models to learn active retrieval through prompting, thereby expanding their knowledge boundaries. Taking this further, R1-Searcher \cite{song2025r1} and Search-R1 \cite{jin2025searchr1} directly incorporate retrieval capabilities into the model, enabling autonomous information retrieval during the reasoning process. These advances suggest a promising direction for improving agent perception: enhancing model-level active perception capabilities to enrich the foundation for decision-making. This approach may represent a significant avenue for future agent development.

\paragraph*{Learning for Better Reasoning}
\label{para:learn_reasoning}

Reasoning serves as a critical bridge between an agent's mental state and its actions, making the ability to reason effectively and the development of reasoning capabilities essential for intelligent agents. The foundation of reasoning in modern agents stems from two key elements: the rich world knowledge embedded in their underlying models, and the robust logical frameworks supported either internally or through context structuring. This makes learning for better reasoning a vital objective in agent development.

The development of reasoning capabilities has evolved through three major paradigms. First, \emph{data-driven approaches} leverage high-quality reasoning data and sophisticated curation methods to directly enhance model capabilities. Second, \emph{reinforcement learning methods} enable more autonomous learning through reward-based optimization, ranging from single-agent self-improvement to complex multi-agent collaborative systems. Third, \emph{self-evolutionary approaches} push the boundaries further by enabling agents to modify their own core functionalities and code, achieving truly autonomous reasoning development.

The importance of reasoning in agent development has been re-emphasized following the release of the o1 series. The most straightforward approach to enhancing reasoning capabilities involves \emph{collecting and distilling data from open/closed-source reasoning models}. For instance, SKY-32B \cite{sky_t1_2025} distilled data from QWQ-32B \cite{qwq-32b-preview} to train a 32B reasoning model at a cost of \$450. Similarly, Open Thoughts \cite{open_thoughts} trained Bespoke-Stratos-32B at a low cost by distilling and synthesizing datasets from DeepSeek-R1~\cite{guo2025deepseek}. These studies demonstrate that even without complex algorithmic design, using reasoning data to perform Supervised Fine-Tuning (SFT) on base models can effectively activate reasoning capabilities.

Building on these basic distillation approaches, researchers have identified that the \emph{quality of reasoning data} plays an even more critical role than quantity. Highly structured reasoning data more effectively enables agents and language models to learn reasoning processes. Notably, LIMO \cite{ye2025limo} demonstrated that powerful reasoning models could be built with extremely few data samples by constructing long and effective reasoning chains for complex reasoning tasks. This insight stems from their observation that language models inherently possess sufficient knowledge for reasoning but require high-quality reasoning paths to activate these capabilities \cite{yao2025unveiling}. Supporting this view, \citet{li2025llms} revealed that both Long CoT and Short CoT fundamentally teach models to learn reasoning structures rather than specific content, suggesting that automated selection of high-quality reasoning data may become an important future direction.

To systematically generate such high-quality data, iterative improvement methods have emerged that enable models to bootstrap their own reasoning capabilities. The bootstrap paradigm exemplified by STaR \cite{zelikman2022star} and its variants implements techniques where models generate step-by-step rationales and iteratively improve through fine-tuning on successful reasoning paths. This family includes Quiet-STaR \cite{zelikman2024quiet}, V-STaR \cite{hosseini2024vstar}, and rStar-Math \cite{guan2025rstar}, with the last one specifically enhancing mathematical reasoning through reinforcement learning principles. By iteratively selecting correct reasoning paths for training, these methods achieve self-improvement through successive refinement cycles.

The most sophisticated data-driven approaches integrate explicit \emph{feedback mechanisms} to provide quality assessment during the data generation process. One viable exploration approach involves first conducting extensive searches, and then using verifiable environments or trainable reward models to provide feedback on reasoning trajectories, thereby filtering out high-quality reasoning data. This approach has led to several families of techniques that leverage different feedback mechanisms to improve reasoning capabilities. The ReST family, beginning with the original ReST \cite{gulcehre2023rest} introducing reinforced self-training, performs multiple attempts (typically 10) per sample and creates new training datasets from successful reasoning instances. ReST-EM \cite{singh2024rest} enhances the approach with expectation maximization, while ReST-MCTS \cite{singh2024rest} further integrates Monte Carlo Tree Search to enable improved reasoning capabilities through more sophisticated exploration strategies.

Several approaches have introduced \emph{Policy Reward Models (PRMs)} to provide quality feedback on reasoning paths. Methods like OpenR \cite{wang2024openr} and LLaMA-Berry \cite{zhang2024llamaberry} model reasoning tasks as Markov Decision Processes (MDPs) and leverage tree search to explore diverse reasoning paths while using PRMs for quality assessment. In domain-specific applications, methods like rStar-Math \cite{guan2025rstar} and DeepSeekMath \cite{shao2024dsmath} have demonstrated success in mathematical problem-solving through multi-round self-iteration and balanced exploration-exploitation strategies. For code generation, o1-Coder \cite{zhang2024o1coder} leverages MCTS to generate code with reasoning processes, while Marco-o1 \cite{zhang2024o1coder} extends this approach to open-ended tasks. These implementations highlight how the synergy between MCTS and PRM achieves effective reasoning path exploration while maintaining solution quality through fine-grained supervision.

Beyond data-driven approaches, \emph{reinforcement learning} and \emph{agentic self-improvement} are the most attractive learning methods.
Reinforcement learning (RL) has demonstrated remarkable success in enhancing language models' reasoning capabilities at the foundational level. Recent breakthroughs like DeepSeek R1 \cite{guo2025deepseek} and Kimi k1.5 \cite{kimiteam2025kimik15} exemplify how RL can directly improve LLMs' reasoning performance. The foundation of RL for LLMs can be traced to several pioneering frameworks: ReFT \cite{luong2024reft} introduced a combination of supervised fine-tuning and online reinforcement learning, while VeRL \cite{sheng2024verl} established an open-source framework supporting various RL algorithms for large-scale models up to 70B parameters. RFT \cite{zhang2024rft} further demonstrated the effectiveness of reward-guided optimization in specific reasoning tasks.

Building on these foundational frameworks, more recent research has pushed the boundaries of LLM reasoning enhancement by exploring how to cultivate reasoning capabilities in more challenging \emph{low-resource settings and self-supervised manner}. \citet{shafayat2025can} explains that the success of self-training highly depends on the model's initial capabilities and the nature of the task, motivating further exploration of effective self-improvement strategies. To address the heavy reliance on large-scale annotated data, innovative reinforcement learning paradigms have been proposed. For instance, Absolute Zero \cite{zhao2025absolute} leverages a self-play mechanism, enabling the evolution of the model's reasoning capabilities without any human-annotated data. Pushing data efficiency to its extreme, \citet{wang2025reinforcement} demonstrates the potential for effective learning and generalization from just a single successful instance.

A core challenge for these LLM-focused learning methods lies in designing effective reward signals. Spurious Rewards \cite{shao2025spurious} cautions that purely outcome-based rewards can be deceptive, as correct answers can arise from flawed reasoning. This has driven a shift towards more robust, process-oriented supervision. To move away from the need for costly external verifiers, pioneering works like Verifree \cite{zhou2025reinforcing} explore the use of intrinsic signals, such as consistency among multiple reasoning paths or self-critique, to generate rewards.

Beyond single-model enhancement, reinforcement learning has evolved to train autonomous agent systems that adapt their own core functionalities through sophisticate collaboration. This system-level approach has driven a surge of innovation focused on multi-agent coordination, credit assignment, and stable training for real-world applications.

A prominent direction is the shift from single agents to multi-agent systems to tackle complexity. Frameworks like MARFT \cite{liao2025marft} now provide standardized infrastructure for multi-agent reinforcement fine-tuning, enabling novel cognitive architectures where agents collaborate as thinkers, critics, and solvers to learn ``meta-think'' \cite{wan2025rema}. Addressing the core RL challenge of credit assignment in long-horizon tasks, researchers have developed more granular reward mechanisms. For instance, SPA-RL reinforces agents by attributing progress at a stepwise level \cite{wang2025sparl}, while algorithmic improvements like GiGPO offer finer-grained policy optimization without significant overhead \cite{feng2025gigpo}. Concurrently, ensuring stability during this self-evolutionary process is critical. Works like RAGEN \cite{wang2025ragen} analyze the dynamics of multi-turn training, identifying failure modes and proposing more robust algorithms. These advancements are empowering agents in complex, real-world domains, from long-horizon software engineering tasks \cite{cao2025skyrl} to dynamic web navigation, where agents trained with self-evolving online curricula can even surpass proprietary models \cite{qi2025webrl}.

Beyond refining existing capabilities through external learning signals, the most transformative recent advances involve agents learning to fundamentally \emph{self-improve} by modifying their own code and underlying operational mechanisms. Alita \cite{qiu2025alita} exemplifies a generalist agent that achieves maximal self-evolution with minimal predefinition by autonomously generating and reusing Model Context Protocols. This allows it to dynamically construct, refine, and integrate new capabilities, akin to writing its own ``operating manuals'' for continuous learning. AlphaEvolve \cite{novikov2025alphaevolve} pushes this autonomous learning to the code level, serving as a coding agent that combines large language models with evolutionary computation. It can directly modify code to improve algorithms and achieve breakthroughs in scientific and algorithmic discovery, such as in areas like matrix multiplication. Furthermore, the Darwin Gödel Machine \cite{zhang2025darwin} takes this a step further, presenting an AI system capable of continuously improving itself by rewriting its own code, achieving the co-evolution of the agent's own coding abilities and its self-improvement capabilities. This represents the ultimate frontier for agents to attain advanced, continuous reasoning skill learning, moving towards truly autonomous and adaptive intelligent systems.

\paragraph*{Learning for World Understanding}
A critical aspect of agent intelligence is the ability to \emph{understand how the world operates} through direct interaction and experience accumulation. This understanding encompasses how the environment responds to different actions and the consequences these actions bring. Through continuous interaction with their environment, agents can build and refine their \textit{memory}, \textit{reward understanding}, and \textit{world model}, learning from both successes and failures to develop a more comprehensive grasp of their operational domain.

The most fundamental approach to world understanding begins with \emph{direct environmental interaction}, where agents learn through basic trial-and-error mechanisms. Inner Monologue \cite{huang2023inner} demonstrates how agents can accumulate basic environmental knowledge through continuous interaction, while Learn-by-Interact \cite{su2025learn} shows that meaningful understanding can emerge from direct environmental engagement without explicit reward mechanisms. These foundational approaches establish the groundwork for more sophisticated learning paradigms, with recent work on Test-Time Interaction \cite{shenbai2025tti} demonstrating that scaling interaction horizon—rather than just reasoning depth—enables agents to dynamically balance exploration and exploitation through extended environmental engagement.

Building upon basic interaction capabilities, agents require systematic mechanisms to \emph{process and organize their accumulated experiences}. More sophisticated approaches are exemplified by DESP \cite{wang2023describe} and Voyager \cite{wang2023voyager} in the Minecraft environment, where agents not only gather experiences but also actively process them through outcome analysis and dynamic skill library expansion, respectively. The systematic processing of experiences has been further enhanced by Generative Agents \cite{park2023generative}, which introduces sophisticated memory replay mechanisms, enabling agents to extract high-level insights from past interactions. This approach is complemented by Self-refine \cite{madaan2024self-refine} and Critic \cite{goucritic}, which implement structured cycles of experience evaluation and refinement, creating robust frameworks for continuous experiential learning.

A crucial component of world understanding involves learning to \emph{interpret and optimize reward signals} through environmental interaction. Text2Reward \cite{xie2023text2reward} demonstrates how agents can continuously refine reward functions through human feedback, better aligning them with task objectives and environmental characteristics. Similarly, AutoManual \cite{chen2024automanual} builds behavioral guidelines through sustained interaction, developing reward-verified protocols that provide a foundation for understanding environmental rewards and decision-making. These interaction-based optimization mechanisms enable agents to better comprehend environmental dynamics and generate more precise reward signals.

The culmination of world understanding research involves the construction and application of explicit \emph{world models} that enable agents to simulate and reason about environmental dynamics. RAP \cite{hao2023reasoning} represents a significant advancement by conceptualizing reasoning as planning with a world model, repurposing LLMs as both reasoning agents and world models to simulate potential action outcomes through Monte Carlo Tree Search. The importance of structured mental representations is highlighted by cognitive maps \cite{Geunwoo2023LMComputer}, which show that cognitively-inspired representations significantly enhance LLMs' extrapolation capabilities in novel environments. In specialized domains, recent work demonstrates that LLMs can function as effective world models for web interactions \cite{gu2024your,chae2024web}, simulating potential state changes before executing actions for safer decision-making. Furthermore, ActRe \cite{yang2024reactx} explores reverse reasoning by first performing actions and then generating post-hoc explanations, demonstrating LLMs' inherent understanding of world dynamics and enabling autonomous trajectory annotation.

The most advanced systems integrate all these components into autonomous learning cycles that can self-manage the entire process of world understanding. Through systems like Reflexion \cite{Shinn2023ReflexionLA} and ExpeL \cite{zhao2024expel}, agents have achieved sophisticated experiential learning by autonomously managing the full cycle of experience collection, analysis, and application, enabling them to learn effectively from both successes and failures. These developments collectively illustrate how world models are becoming increasingly central to agent learning systems, providing a foundation for understanding environmental dynamics and enabling more effective planning, reasoning, and decision-making in complex, interactive environments.

\section{Reasoning}
\label{sec:reasoning}


\lettrine[lines=3]{\initfamily\textcolor{darkgreen}{R}}{easoning} represents the key to intelligent behavior, transforming raw information into actionable knowledge that drives problem-solving and decision-making. For both humans and artificial agents, it enables logical inference, hypothesis generation, and purposeful interaction with the world. In human cognition, reasoning emerges through multiple strategies: \emph{deductive reasoning} applies general rules to specific cases, \emph{inductive reasoning} builds generalizations from particular instances, and abductive reasoning constructs plausible explanations from incomplete data~\cite{cheng2024inductive, hayes2010inductive}. These processes rely on heuristics, mental shortcuts that streamline decision-making under uncertainty, and they are continuously refined through environmental feedback, ensuring that reasoning remains grounded in reality and adaptive to change. 

For LLM-based agents, reasoning serves a parallel role, elevating them beyond reactive systems to proactive entities capable of sophisticated cognition. Through reasoning, these agents process multimodal inputs, integrate diverse knowledge sources, and formulate coherent strategies to achieve objectives. The environment plays a dual function: supplying information that fuels reasoning and serving as the proving ground where reasoned actions are tested, creating a feedback loop that enables agents to validate inferences and learn from errors.

In LLM-based agents, reasoning can be formally defined as the process of \textbf{action selection based on mental states}, representing a crucial bridge between perception and action. More precisely, given a mental state $M_t$ at time $t$, reasoning can be formalized as a function $R(M_t) \rightarrow a_t$, where $a_t$ represents the selected action. This process operates across diverse environments such as textual, digital, and physical worlds, where completing a task typically requires either a single reasoning step or a composition of multiple reasoning actions.

\begin{figure}[!t]
\centering
    \includegraphics[width=\columnwidth]{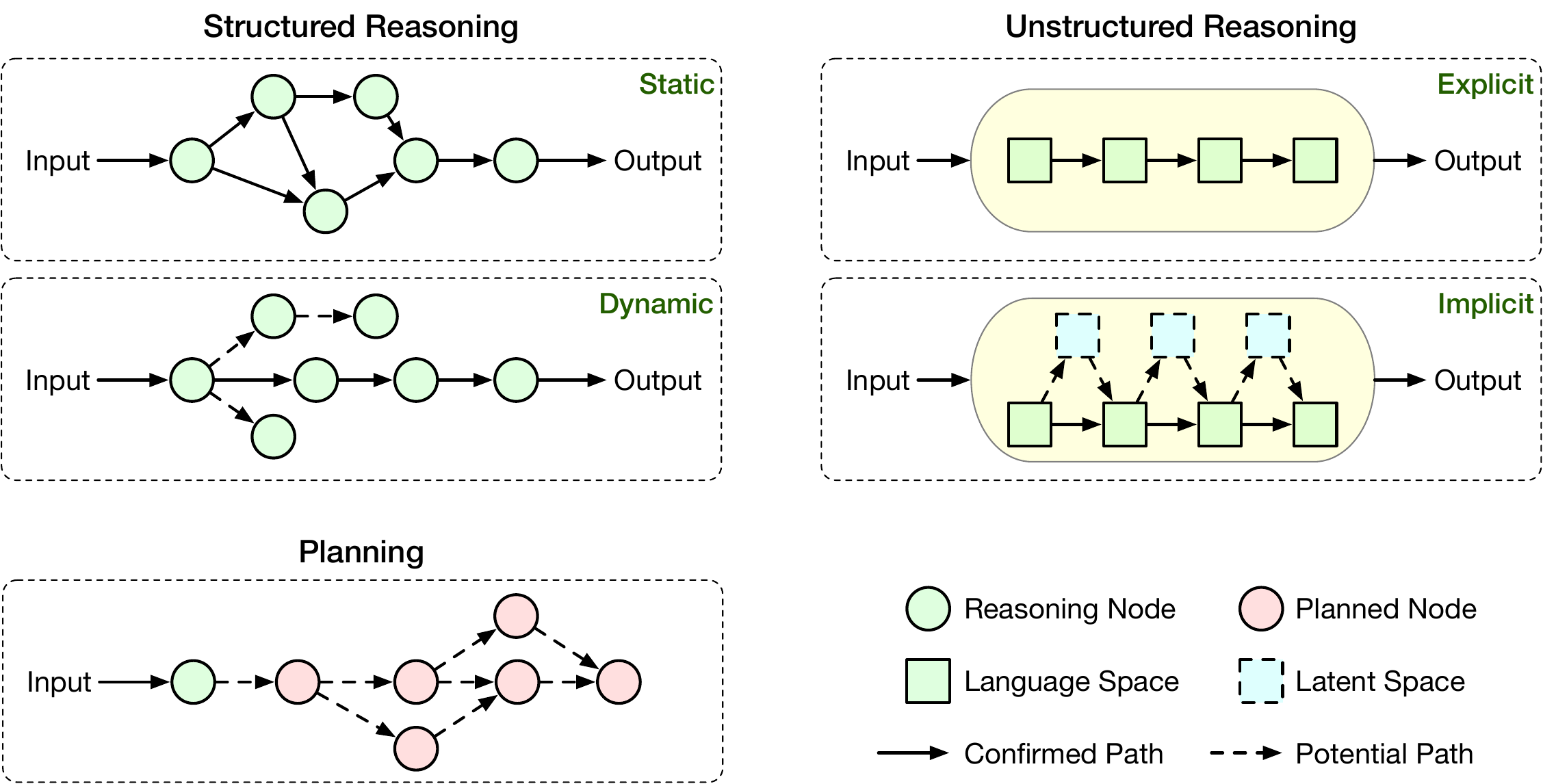}
    \caption{Comparison of reasoning paradigms in LLM-based agents.}
\label{fig:cognition-reasoning}
\end{figure}

The composition of reasoning actions naturally leads to two distinct approaches: structured and unstructured reasoning. Figure~\ref{fig:cognition-reasoning} illustrates the different reasoning paradigms. \emph{Structured reasoning} ($R_s$) can be formalized as an explicit composition $R_s = R_1 \circ R_2 \circ \ldots \circ R_n$, where each $R_i$ represents a discrete reasoning step with clear logical dependencies. In contrast, \emph{unstructured reasoning} ($R_u$) takes a more holistic form $R_u = f(M_t)$, where the composition remains implicit and flexible, allowing for dynamic adaptation to context. This dual framework mirrors human cognition, where structured reasoning parallels our explicit logical deduction processes, while unstructured reasoning reflects our capacity for intuitive problem-solving and pattern recognition.

The environment plays a crucial role in this formalization, serving both as a source of observations $o_t$ that influence mental state updates ($M_t = L(M_{t-1}, a_{t-1}, o_t)$) and as a testing ground for reasoning outcomes. This creates a continuous feedback loop where reasoning not only drives action selection but also influences how the agent's mental state evolves, enabling iterative refinement of reasoning strategies through experience.

In this section, we will examine how these reasoning approaches manifest in practice. We first present a more concrete formulation of reasoning in general. Then we introduce structured reasoning that emphasizes systematic problem decomposition and multi-step logical chains. We next explore unstructured reasoning, which allows for flexible response patterns and parallel solution exploration. Finally, we investigate planning as a specialized form of reasoning that combines both structured and unstructured approaches for tackling complex, long-horizon tasks.

\subsection{A Unified Formulation of Reasoning}
\label{sec:unified-reasoning}

Reasoning is the internal process by which an agent transforms its current mental state \(M_t\) into concrete actions or decisions. These actions may range from low-level motor outputs to high-level plans, and may even involve recursively generating new subgoals. To unify diverse reasoning patterns, we formulate reasoning as the selection of a policy that minimizes a free-energy objective over imagined futures, balancing utility, curiosity, resource constraints, and computational cost.

\begin{definition}[\textbf{Unified Reasoning Objective}]
\label{def:reasoning-objective}
Reasoning selects a policy that minimizes expected future free energy while respecting exploration and resource constraints:
\[
\begin{aligned}
\pi_t^\star &= \arg\min_{\pi}\sum_i v_i\,\mathbb{E}_{\tau\sim\mathcal{T}_i}\!\Bigl[\mathbb{E}_{o_\tau,r_\tau\sim P(\cdot\mid \pi,\ M_t)}[-\,r_\tau] + \alpha\,I(o_\tau;\pi\mid M_t) + \lambda\,\mathcal{C}_{\text{act}}(\pi)\Bigr] + \beta\,\mathcal{S}(\pi) \\
a_t &= \mathrm{Extract}(\pi_t^\star,M_t)
\end{aligned}
\]

\paragraph*{Utility term:} The expected reward \(\mathbb{E}[-r_\tau]\) penalizes undesirable outcomes or low task utility. When reward reflects goal satisfaction, this term recovers classical reinforcement learning and optimal control; when reward reflects prior preferences, it recovers active inference.

\paragraph*{Exploration term:} The mutual information term \(I(o_\tau;\pi\mid M_t)\) encourages exploratory behavior by rewarding policies that reduce epistemic uncertainty. Setting \(\alpha = 0\) yields pure exploitation; \(\alpha > 0\) induces curiosity.

\paragraph*{Action cost:} The cost term \(\mathcal{C}_{\text{act}}(\pi)\) reflects practical penalties such as energy usage, risk, or API fees.

\paragraph*{Search cost:} The search term \(\mathcal{S}(\pi)\) quantifies internal computational effort, such as number of nodes, tokens, or floating-point operations. The weight \(\beta\) enforces bounded rationality.

\paragraph*{Temporal mixture:} The coefficients \(v_i\) allow reasoning to mix over different time horizons \(\mathcal{T}_i\), from short-term reflexes to long-term planning. These mirror the horizon weights used in the learning objective.

\paragraph*{Policy hierarchy:} A policy \(\pi\) can represent decisions at various levels of abstraction—from atomic actions to multi-step strategies. The extraction function \(\mathrm{Extract}(\pi_t^\star, M_t)\) maps the selected policy to an executable action, possibly invoking another round of reasoning in a receding-horizon fashion.
\end{definition}

\vspace{0.5em}
While the unified objective offers a theoretical view of reasoning, practical systems must implement it with finite resources, modular components, and clear interfaces. In particular, most reasoning engines—from classical planners to LLM-based agents—can be interpreted as repeatedly transforming internal hypotheses via a combination of \emph{rewrite operators} and a \emph{scheduling policy}.

\paragraph*{Implementation Scaffold} To realize the unified objective in practice, intelligent agents adopt a concrete structure based on a loop over internal hypotheses. This loop consists of selecting and applying transformation operators from a library \(\mathcal{L}\), under the guidance of a scheduler policy \(\pi\), starting from an initial seed hypothesis \(h_0 = \mathrm{Init}(M_t)\). This abstraction captures a wide variety of paradigms—including symbolic planning, neural program synthesis, and chain-of-thought prompting—and can be formally described as follows.

\begin{definition}[\textbf{Operator–Scheduler Realisation}]
\label{def:reasoning-realisation}
Let \(\mathcal H\) be the space of internal hypotheses.
An initial draft \(h_0=\mathrm{Init}(M_t)\) is produced from the current mental state.
A finite operator library \(\mathcal L = \{r^{(1)}, \dots, r^{(K)}\}\) defines atomic rewrites \(r^{(k)} : \mathcal{H} \to \mathcal{H}\).
At each step \(i\), the scheduler samples \(\sigma_i \sim \pi(\cdot \mid h_{i-1}, M_t)\), and applies the selected operator: \(h_i = r^{(\sigma_i)}(h_{i-1})\).
The iteration halts when a termination predicate \(\mathrm{Term}(h_i, M_t)\) is satisfied or a resource budget is exhausted.
The final draft \(h_N\) is converted into an action via \(\mathrm{Extract}(h_N, M_t)\).
\end{definition}

\noindent
This iterative process produces the \textbf{reasoning trace} \(\zeta_t = (h_0, h_1, \dots, h_N)\), which records the internal deliberation sequence for step \(t\). It provides a unified pattern encompassing classical search trees, stochastic rollouts, and token-by-token language generation.

Importantly, the scheduler policy \(\pi\) that governs the application of operators is exactly the object being optimized in the free-energy formulation (Definition~\ref{def:reasoning-objective}). The operator dynamics define the transitions through hypothesis space, while the scheduler implicitly balances utility, curiosity, action cost, and compute effort.

\begin{figure}[!htb]
\centering
\includegraphics[width=\columnwidth]{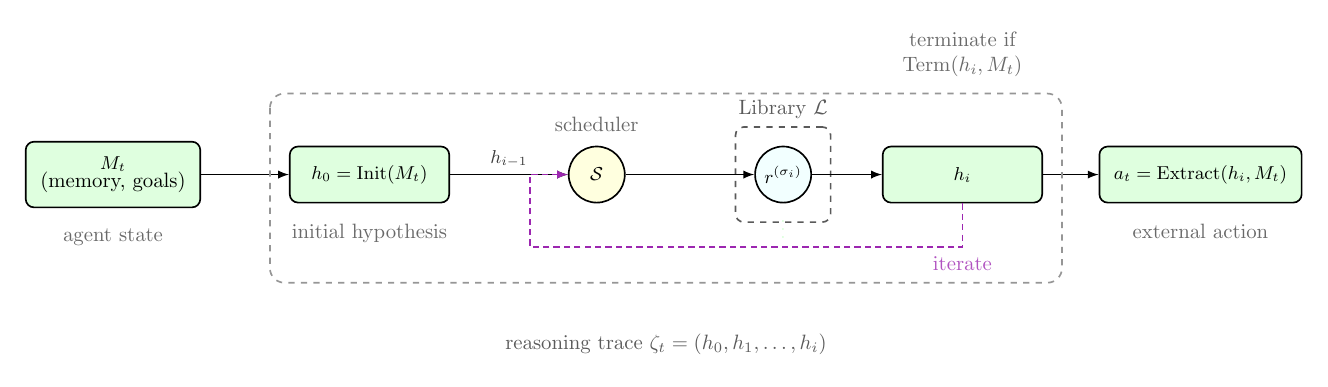}
\caption{Reasoning as a scheduler-controlled sequence of atomic rewrites.
From the mental state $M_t$ the agent seeds an initial hypothesis $h_0$.
The scheduler $\mathcal{S}$ (orange) repeatedly chooses an operator
$r^{(\sigma_i)}$ from the library $\mathcal{L}$ (green) and applies it to the current draft, producing the updated hypothesis $h_i$. This loop continues (purple dashed arrow) until the termination predicate
$\mathrm{Term}(h_i,M_t)$ is satisfied, after which the final hypothesis is
converted to an external action $a_t=\mathrm{Extract}(h_i,M_t)$. The dashed
frame encloses the reasoning trace
$\zeta_t=(h_0,h_1,\dots,h_i)$, a pattern that subsumes chain-of-thought,
tree search, and program-synthesis planners alike.}
\label{fig:reasoning}
\end{figure}

\vspace{0.5em}
To demonstrate the generality of this framework, we summarize a wide range of reasoning paradigms—across AI planning, reinforcement learning, inference, and LLM-based techniques—as specific instantiations of the operator–scheduler pattern. Table~\ref{tab:reasoning-instances} presents each paradigm's scheduler policy, utility signal, exploration strategy, operator set, and search cost, all of which align with the unified formulation.

\begin{table}[ht]
\centering
\caption{Reasoning paradigms as instantiations of the operator–scheduler realisation. Each row specifies the components that instantiate the unified objective: scheduler policy, utility, exploration term, operator set, and search cost. All conform to the free-energy skeleton of Definition~\ref{def:reasoning-objective}.}
\small
\setlength{\tabcolsep}{4pt}
\begin{tabular}{@{}p{2.9cm}p{2.6cm}p{2.4cm}p{2.4cm}p{2.4cm}p{2.4cm}@{}}
\toprule
\textbf{Paradigm} & \textbf{Scheduler policy \(\pi\)} & \textbf{Utility signal} & \textbf{Exploration term} & \textbf{Operator set \(\mathcal L\)} & \textbf{Search cost \(\mathcal S\)} \\ \midrule
\rowcolor{LightMint}
Reactive control & rule-based single action & immediate reward & \(\alpha=0\) & identity & negligible \\
Model-free RL & param.\ \(\pi_\theta(a\mid s)\) & discounted return & entropy bonus & identity & gradient steps / batch \\
\rowcolor{LightMint}
Monte-Carlo tree search & tree policy over moves & win–loss & UCB bonus & expand, rollout, backprop & node budget \\
Hierarchical RL & option policy & option value & KL to prior & invoke\_option, terminate & option horizon budget \\
\rowcolor{LightMint}
Active inference & trajectory distribution & prior preference & epistemic F.E. & integrator & gradient iterations \\
LLM chain-of-thought & token policy & reward / log-prob & variance penalty & append\_token, self\_eval & token budget \\
\rowcolor{LightMint}
Theorem proving & proof-tactic policy & proof length cost & premise novelty & tactic library & node expansions \\
Program synthesis & code-edit policy & test failure cost & branch-uncertainty bonus & insert, replace, delete & edit budget \\
\rowcolor{LightMint}
Meta-controller & policy over planners & outer-loop utility & info-gain across tasks & spawn\_planner, eval\_planner & run budget \\ \bottomrule
\end{tabular}
\label{tab:reasoning-instances}
\end{table}

\paragraph*{Symmetry with Learning} Learning optimizes internal models using real-world data; reasoning optimizes policies over imagined futures simulated by those models. Both processes minimize composite free-energy losses of the form:  
\[
\text{expected loss} + \alpha \times \text{uncertainty penalty} + \lambda \times \text{action cost} + \beta \times \text{resource cost}
\]
Together, they complete the perception–cognition–action loop at the heart of intelligent agents.

\paragraph*{Module Interfaces}
Operators may call the world model \(W\) for simulation,
query memory for context,
or receive urgency/temperature parameters from the emotion state \(E_t\).
Their trainable parameters join the global set \(\Theta\)
updated by the learning rule \(L\).

\medskip
\noindent
Our formulation of reasoning therefore captures different
reasoning styles examined in the rest of this chapter while
remaining consistent with the overarching
perception--cognition--action loop of Foundation Agents.

\subsection{Structured Reasoning}
\label{subsec:structured-reasoning}

Structured reasoning represents a methodical approach to problem-solving that employs explicit organizational frameworks to guide the reasoning process. Unlike unstructured approaches, structured reasoning makes the composition of reasoning steps explicit, which can be formalized as $R_s = R_1 \circ R_2 \circ \ldots \circ R_n$, where each $R_i$ represents a discrete reasoning step with clear logical dependencies. In this formulation, each reasoning node is an explicitly executed computational unit, and the connections between nodes represent definite information flow paths. This approach enables more systematic exploration of solution spaces and facilitates more robust decision-making through deliberate step-by-step analysis, providing high interpretability and traceability throughout the reasoning process.

\subsubsection{Dynamic Reasoning Structures}
\label{subsubsec:dynamic-reasoning}

Dynamic reasoning structures allow for the adaptive construction of reasoning paths during problem-solving, creating versatile frameworks that can adjust based on intermediate results and insights.

\paragraph*{Linear Sequential Reasoning}
Linear structures frame reasoning as a series of sequential steps, where each step builds on the one before. ReAct \cite{yao2022react} illustrates this by combining reasoning traces with task-specific actions in an alternating fashion. This combination allows for reasoning traces to guide and modify action plans while actions can access external sources for further information. This mutual interaction improves both reasoning integrity and environmental adaptation.
Reasoning via Planning (RAP) \cite{hao2023reasoning} extends the linear reasoning paradigm by formulating LLM reasoning as a Markov decision process, though it was limited by states specifically designed for particular problems. The Markov Chain of Thought (MCoT) \cite{yang2024markov} extended this paradigm by conceptualizing each reasoning step as a Markovian state accompanied by executable code. This approach enables efficient next-step inference without requiring a lengthy context window by compressing previous reasoning into a simplified math question. Atom of Thoughts~\cite{teng2025atomthoughtsmarkovllm}  explicitly defined problems as state representations and designed a general decomposition-contraction two-phase state transition mechanism to construct Markovian reasoning processes, transforming complex problems into a series of atomic questions.
Furthermore, some research has begun to challenge the Markovian assumption prevalent in many linear reasoning processes. \citet{zhang2025beyond} argues that complex reasoning is inherently non-Markovian and requires consideration of the entire history. It proposes a Reflective Exploration framework based on Bayes-Adaptive RL, which allows the agent to update its strategy based on the full reasoning trace, providing a theoretical foundation for building more powerful long-horizon reasoning models.

\paragraph*{Tree-Based Exploration}
Tree-based approaches expand beyond linear structures by organizing reasoning into hierarchical frameworks that support branching exploration. Tree of Thoughts (ToT) \cite{yaotree} introduces a structured approach where complex problems are decomposed into intermediate steps, enabling breadth-first or depth-first search through the solution space. This allows the model to consider multiple reasoning paths simultaneously and systematically explore alternatives.
Language Agent Tree Search (LATS) \cite{zhou2024language} advances this paradigm by integrating Monte Carlo Tree Search (MCTS) with LLMs, using the environment as an external feedback mechanism. This approach enables more deliberate and adaptive problem-solving by balancing exploration and exploitation through a sophisticated search process guided by LLM-powered value functions and self-reflection.
Reasoning via Planning (RAP) \cite{hao2023reasoning} further enhances tree-based reasoning by repurposing LLMs as both reasoning agents and world models. Through this dual role, RAP enables agents to simulate the outcomes of potential reasoning paths before committing to them, creating a principled planning framework that balances exploration with exploitation in the reasoning space.

\paragraph*{Graph-Based Reasoning}
Graph structures offer even greater flexibility by allowing non-hierarchical relationships between reasoning steps. Graph of Thoughts (GoT) \cite{Besta2023GraphOT} extends tree-based approaches to arbitrary graph structures, enabling more complex reasoning patterns that can capture interdependencies between different steps. This approach allows for connections between seemingly disparate reasoning branches, facilitating more nuanced exploration of the solution space.
Path of Thoughts (PoT)~\cite{zhang2024path} addresses relation reasoning challenges by decomposing problems into three key stages: graph extraction, path identification, and reasoning. By explicitly extracting a task-agnostic graph that identifies entities, relations, and attributes within the problem context, PoT creates a structured representation that facilitates the identification of relevant reasoning chains, significantly improving performance on tasks requiring long reasoning chains.
Diagram of Thought (DoT)~\cite{Zhang2024OnTD} models iterative reasoning as the construction of a directed acyclic graph (DAG), organizing propositions, critiques, refinements, and verifications into a unified structure. This approach preserves logical consistency while enabling the exploration of complex reasoning pathways, providing a theoretically sound framework grounded in Topos Theory.

\subsubsection{Static Reasoning Structures}
\label{subsubsec:static-reasoning}

Static reasoning structures employ fixed frameworks that guide the reasoning process without dynamically adjusting the structure itself, focusing instead on improving the content within the established framework.

\paragraph*{Ensemble Methods}
Ensemble approaches leverage multiple independent reasoning attempts to improve overall performance through aggregation. Self-Consistency \cite{wang2023selfconsistency} pioneered this approach by sampling multiple reasoning paths rather than relying on single greedy decoding, significantly improving performance through majority voting among the generated solutions.
MedPrompt \cite{nori2023can} demonstrates how domain-specific ensemble techniques can enhance performance by carefully crafting prompts that elicit diverse reasoning approaches, achieving state-of-the-art results on medical benchmarks through systematic composition of prompting strategies.
LLM-Blender \cite{Jiang2023LLMBlenderEL} introduces a sophisticated ensembling framework that leverages the diverse strengths of multiple LLMs through pairwise comparison (PairRanker) and fusion (GenFuser) of candidate outputs. This approach enables the system to select the optimal model output for each specific example, creating responses that exceed the capabilities of any individual model.

\paragraph*{Progressive Improvement}
Progressive improvement frameworks focus on iteratively refining reasoning through structured feedback loops. Self-Refine \cite{madaan2024self-refine} implements an iterative approach where the model generates initial output, provides self-feedback, and uses that feedback to refine itself. This mimics human revision processes without requiring additional training or reinforcement learning, resulting in significant improvements across diverse tasks.
Reflexion \cite{Shinn2023ReflexionLA} extends this concept by integrating environmental feedback, enabling agents to verbally reflect on task feedback signals and maintain reflective text in an episodic memory buffer. This approach guides future decision-making by incorporating insights from previous attempts, significantly enhancing performance in sequential decision-making, coding, and reasoning tasks.
Progressive-Hint Prompting (PHP) \cite{zheng2023progressive} further develops this paradigm by using previously generated answers as hints to progressively guide the model toward correct solutions. This approach enables automatic multiple interactions between users and LLMs, resulting in significant accuracy improvements while maintaining high efficiency.

\paragraph*{Error Correction}
Error correction frameworks focus specifically on identifying and addressing mistakes in the reasoning process. Self-Verification \cite{stechly2024self} introduces a self-critique system that enables models to backward-verify their conclusions by taking the derived answer as a condition for solving the original problem, producing interpretable validation scores that guide answer selection.
Refiner \cite{paul2024refiner} addresses the challenge of scattered key information by adaptively extracting query-relevant content and restructuring it based on interconnectedness, highlighting information distinction and effectively aligning downstream LLMs with the original context.
Chain-of-Verification (CoVe) \cite{dhuliawala2024chain} tackles factual hallucinations through a structured process where the model drafts an initial response, plans verification questions, independently answers those questions, and generates a final verified response. This deliberate verification process significantly reduces hallucinations across a variety of tasks.
Recursive Criticism and Improvement (RCI) \cite{Geunwoo2023LMComputer} enables LLMs to execute computer tasks by recursively criticizing and improving their outputs, outperforming existing methods on the MiniWoB++ benchmark with only a handful of demonstrations per task and without task-specific reward functions.
Critic \cite{goucritic} extends this approach by integrating external tools for validation, enabling LLMs to evaluate and progressively amend their outputs like human interaction with tools. This framework allows initially ``black box'' models to engage in a continuous cycle of evaluation and refinement, consistently enhancing performance across diverse tasks.

\subsubsection{Domain-Specific Reasoning Frameworks}
\label{subsubsec:domain-specific-reasoning}

Domain-specific reasoning frameworks adapt structured reasoning approaches to the unique requirements of particular domains, leveraging specialized knowledge and techniques to enhance performance in specific contexts.

MathPrompter \cite{imani2023mathprompter} addresses arithmetic reasoning challenges by generating multiple algebraic expressions or Python functions to solve the same math problem in different ways. This approach improves confidence in the output results by providing multiple verification paths, significantly outperforming state-of-the-art methods on arithmetic benchmarks.
Physics Reasoner \cite{pang2025physics} addresses the unique challenges of physics problems through a knowledge-augmented framework that constructs a comprehensive formula set and employs detailed checklists to guide effective knowledge application. This three-stage approach—problem analysis, formula retrieval, and guided reasoning—significantly improves performance on physics benchmarks by mitigating issues of insufficient knowledge and incorrect application.
Pedagogical Chain-of-Thought (PedCoT) \cite{jiang2024pedcot} leverages educational theory, particularly the Bloom Cognitive Model, to guide the identification of reasoning mistakes in mathematical contexts. This approach combines pedagogical principles for prompt design with a two-stage interaction process, providing a foundation for reliable mathematical mistake identification and automatic answer grading.

The evolution of structured reasoning in LLM agents reflects a growing understanding of how to enhance reasoning capabilities through explicit organizational frameworks. From linear sequences to complex graphs, and ensemble methods to specialized domain frameworks, these approaches demonstrate the power of structural guidance in improving reasoning performance across diverse tasks and domains.

\subsection{Unstructured Reasoning}
\label{subsec:unstructured-reasoning}

In contrast to structured reasoning approaches that explicitly organize reasoning steps, unstructured reasoning ($R_u$) takes a holistic form $R_u = f(M_t)$, where the composition remains implicit and flexible. In this mode, the reasoning process is encapsulated within a single function mapping, without explicitly defining intermediate steps or state transitions. This approach leverages the inherent capabilities of language models to generate coherent reasoning without enforcing rigid structural constraints, with intermediate reasoning processes occurring explicitly in the language space or implicitly in the latent space. Unstructured reasoning methods have demonstrated remarkable effectiveness across diverse tasks while maintaining simplicity and efficiency in implementation.

\subsubsection{Prompting-Based Reasoning}
\label{subsubsec:prompt-based-reasoning}

The most accessible way to elicit reasoning in LLM agents lies in carefully crafted prompts. By providing appropriate reasoning demonstrations or instructing LLMs to perform inferential steps, agents can leverage their logical deduction capabilities to solve problems through flexible reasoning processes.

\paragraph*{Chain-of-Thought Variants}
The cornerstone of prompting-based reasoning is Chain-of-Thought (CoT) prompting~\cite{wei2022chain}, which operationalizes reasoning through few-shot examples with explicit generation of intermediate rationalization steps. This foundational technique has inspired several evolutionary variants that enhance its basic approach. Zero-shot CoT~\cite{kojima2022large} eliminates the need for demonstration examples through strategic prompting (e.g., ``Let's think step by step''), making the approach more accessible while maintaining effectiveness. Auto-CoT~\cite{zhang2023automatic} automates the creation of effective demonstrations by clustering diverse questions and generating reasoning chains for representative examples from each cluster. Least-to-Most Prompting~\cite{zhouleast} addresses complex reasoning by decomposing problems into sequential sub-problems, enabling a progressive planning process that facilitates easy-to-hard generalization. Complex CoT~\cite{fu2023complexity} further enhances reasoning depth by specifically selecting high-complexity exemplars as prompting templates, better equipping models to tackle intricate problems.

\paragraph*{Problem Reformulation Strategies}
Advanced prompting strategies demonstrate architectural innovations in reasoning guidance by reformulating the original problem. Step-Back Prompting~\cite{zhengtake} implements abstraction-first reasoning through conceptual elevation, enabling models to derive high-level concepts and first principles before addressing specific details. Experimental results demonstrate substantial performance gains on various reasoning-intensive tasks, with improvements of 7-27\% across physics, chemistry, and multi-hop reasoning benchmarks. Rephrase and Respond~\cite{deng2023rephrase} employ semantic expansion to transform original questions into more tractable forms, allowing models to approach problems from multiple linguistic angles and identify the most effective problem formulation.
Abstraction-of-Thought~\cite{hong2024abstraction} introduces a novel structured reasoning format that explicitly requires varying levels of abstraction within the reasoning process. This approach elicits language models to first contemplate at the abstract level before incorporating concrete details, a consideration overlooked by step-by-step CoT methods. By aligning models with the AoT format through finetuning on high-quality samples, the approach demonstrates substantial performance improvements across a wide range of reasoning tasks compared to CoT-aligned models.

\paragraph*{Enhanced Prompting Frameworks}
Several frameworks extend the basic prompting paradigm by introducing structured constraints and knowledge integration mechanisms. Ask Me Anything~\cite{arora2022ask} constrains open-ended generation through task reformulation into question-answer sequences, while Algorithm of Thoughts~\cite{sel2023algorithm} exploits the recurrence dynamics of LLMs by employing algorithmic examples in context to guide reasoning pathways. Chain-of-Knowledge (CoK)~\cite{li2023chain} addresses factual grounding by dynamically incorporating heterogeneous knowledge sources through adaptive query generation, and Self-Explained Keywords (SEK)~\cite{fan2024self} tackles low-frequency term challenges by extracting and explaining key concepts within the reasoning process itself.

\subsubsection{Reasoning Models}
\label{subsubsec:reasoning-models}


As described in \ref{para:learn_reasoning}, both models and agents have achieved enhanced reasoning capabilities through various approaches, with these reasoning abilities being inherently embedded within the model parameters rather than constrained by human-imposed structural limitations. Recent advances in language models have led to the development of specialized reasoning models designed explicitly for complex inferential tasks, incorporating data, architecture, and training innovations that optimize reasoning capabilities and enhance performance on tasks requiring logical inference.

Reasoning models like DeepSeek's R1~\cite{guo2025deepseek}, Anthropic's Claude 3.7 Sonnet~\cite{anthropic2023claude}, and OpenAI's o series models~\cite{jaech2024openai} represent the frontier of reasoning capabilities, demonstrating remarkable proficiency across diverse reasoning benchmarks. These models are trained with specialized methodologies that emphasize reasoning patterns, often incorporating significant amounts of human feedback and reinforcement learning to enhance their inferential abilities.

The emergence of dedicated reasoning models reflects a growing understanding of the importance of reasoning capabilities in language models and the potential benefits of specialized training for these tasks \cite{ye2025limo,wang2025ragen,muennighoff2025s1,wu2025concise}. By concentrating on reasoning-focused training data and objectives, these models achieve performance levels that significantly exceed those of general-purpose language models, particularly on tasks that require complex logical inference, mathematical reasoning, and multi-step problem-solving.

\subsubsection{Implicit Reasoning}
\label{subsubsec:implicit-reasoning}

Beyond explicit reasoning approaches, recent research has explored the potential of implicit reasoning methods that operate without overtly exposing the reasoning process. These approaches aim to improve efficiency by reducing the number of tokens generated while maintaining or enhancing reasoning performance.

Quiet-STaR~\cite{zelikman2024quiet} generalizes the Self-Taught Reasoner approach by teaching LMs to generate rationales at each token to explain the future text, improving their predictions. This approach addresses key challenges, including computational cost, the initial unfamiliarity with generating internal thoughts, and the need to predict beyond individual tokens. Experimental results demonstrate zero-shot improvements in mathematical reasoning (5.9\%→10.9\%) and commonsense reasoning (36.3\%→47.2\%) after continued pretraining, marking a step toward LMs that learn to reason in a more general and scalable way.

Chain of Continuous Thought (Coconut)~\cite{hao2024training} introduces a paradigm that enables LLM reasoning in an unrestricted latent space instead of using natural language. By utilizing the last hidden state of the LLM as a representation of the reasoning state and feeding it back as the subsequent input embedding directly in the continuous space, Coconut demonstrates improved performance on reasoning tasks with fewer thinking tokens during inference. This approach leads to emergent advanced reasoning patterns, including the ability to encode multiple alternative next reasoning steps, allowing the model to perform a breadth-first search rather than committing to a single deterministic path.

Seeking to harness the efficiency of implicit methods while retaining the robustness of explicit thought, the framework proposed in System-1.5 Reasoning~\cite{wang2025system}. It allows a model to dynamically shortcut from explicit, step-by-step linguistic reasoning into an efficient latent space for parallel exploration, and then return to the explicit space. This enables faster, ``insight-like'' leaps in logic, offering a new direction for building more powerful reasoning systems that combine the speed of intuition with the rigor of deliberation.

However, the fundamental generalization capabilities of implicit reasoning, including these hybrid forms, remain a critical concern. Recent analysis~\cite{lin2025implicit} of implicit reasoning in transformers reveals important insights into its limitations. While language models can perform step-by-step reasoning and achieve high accuracy in both in-domain and out-of-domain tests via implicit reasoning when trained on fixed-pattern data, implicit reasoning abilities emerging from training on unfixed-pattern data tend to overfit specific patterns and fail to generalize further. These findings suggest that language models acquire implicit reasoning through shortcut learning, enabling strong performance on tasks with similar patterns while lacking broader generalization capabilities.

The evolution of unstructured reasoning approaches demonstrates the remarkable adaptability of language models to different reasoning paradigms. From simple prompting techniques to sophisticated implicit reasoning methods, these approaches leverage the inherent capabilities of LLMs to perform complex logical inferences without requiring explicit structural constraints. This flexibility enables more intuitive problem-solving while maintaining efficiency and effectiveness across diverse reasoning tasks.

\subsection{Planning}
\label{subsec:planning}

Planning is a fundamental aspect of human cognition, enabling individuals to organize actions, anticipate outcomes, and achieve goals in complex, dynamic environments~\cite{newell1958elements, kambhampati2004cse574}. Formally, planning can be described as the process of \textbf{constructing potential pathways from an initial state to a desired goal state}, represented as \( P: S_0 \rightarrow \{a_1, a_2, \dots, a_n\} \rightarrow S_t \), where \( S_0 \) is the source state, \( \{a_1, a_2, \dots, a_n\} \) denotes a sequence of possible actions, and \( S_g \) is the goal state. Unlike direct reasoning, planning involves generating hypothetical action sequences before execution, functioning as computational nodes that remain inactive until deployed. This cognitive ability emerges from the interplay of specialized neural circuits, including the prefrontal cortex, which governs executive control, and the hippocampus, which supports episodic foresight and spatial mapping. Insights from decision theory, psychology, and cybernetics—such as rational frameworks, prospect theory, and feedback loops—demonstrate how planning allows humans to transcend reactive behavior, actively shaping their futures through deliberate intent and adaptive strategies. This capacity not only underpins intelligent behavior but also serves as a model for developing LLM-based agents that seek to replicate and enhance these abilities computationally~\cite{huang2024understanding, li2024lasp}.

In human cognition, planning operates as a hierarchical process, integrating immediate decisions with long-term objectives. This reflects the brain's modular architecture, where neural systems collaborate to balance short-term demands with future possibilities—a dynamic informed by control theory's principles of stability and optimization. Similarly, LLM-based agents employ planning by leveraging their extensive linguistic knowledge and contextual reasoning to transform inputs into actionable steps. Whether addressing structured tasks or unpredictable challenges, these agents emulate human planning by decomposing objectives, evaluating potential outcomes, and refining their strategies—blending biological inspiration with artificial intelligence. This section examines the theoretical foundations and practical techniques of planning, from sequential approaches to parallel exploration, highlighting its critical role in intelligent systems.

Despite the potential of LLMs in automated planning, their performance faces limitations due to gaps in world knowledge~\cite{kambhampati2024can}. LLMs often lack deep comprehension of world dynamics, relying on pattern recognition rather than genuine causal reasoning, which hinders their ability to manage sub-goal interactions and environmental changes~\cite{valmeekam2023planning}. Additionally, their reliance on static pre-training data restricts adaptability in real-time scenarios, limiting their generalization in dynamic planning tasks~\cite{pallagani2023understanding}. The absence of an intrinsic System 2 reasoning mechanism further complicates their ability to independently generate structured, optimal plans~\cite{kambhampati2024llms}. However, researchers have proposed strategies such as task decomposition, search optimization, and external knowledge integration to mitigate these challenges.

\paragraph*{Task Decomposition}
Task decomposition enhances LLM planning by breaking complex goals into smaller, manageable subtasks, reducing problem complexity and improving systematic reasoning \cite{guan2025etva}. The Least-to-Most Prompting method \cite{zhouleast} exemplifies this approach, guiding LLMs to solve subproblems incrementally. ADaPT \cite{prasad2024adapt} further refines this strategy by dynamically adjusting task decomposition based on complexity and model capability, particularly in interactive decision-making scenarios. These methods also facilitate parallel subtask processing, backward error tracing, and independence determination \cite{teng2025atomthoughtsmarkovllm}, providing a structured framework for reasoning.

In LLM planning, tasks function as executable units—distinct from static state descriptions in formal models—emphasizing structured sequences that achieve intended outcomes \cite{wang2023describe}. These tasks vary in nature: some are subproblems requiring specific solutions such as solving equations within broader challenges, while others involve tool invocation such as querying APIs for weather data in travel planning \cite{shen2024hugginggpt, Pan2023Chameleon}. Alternatively, tasks may be represented as graph nodes mapping dependencies, such as prioritizing objectives in project management \cite{lin2024graph}. By defining clear, modular goals, these formulations enhance reasoning and action efficiency, guiding agents through complex problem spaces with greater precision \cite{singh2023progprompt}.

\paragraph*{Searching}
Given the stochastic nature of LLMs \cite{setlur2025scaling}, parallel sampling combined with aggregated reasoning can improve inference performance. Task decomposition structures individual solution trajectories, enabling the construction of a solution space that includes multiple pathways to a goal and their interrelationships \cite{yaotree, haollm}. This space allows sampling diverse potential solutions \cite{zhang2024thought}, facilitating exploration through techniques like reflection, review, and parallel sampling informed by existing knowledge \cite{xiong2024large}.

Computational constraints often preclude exhaustive evaluation, making efficient navigation of the solution space essential. Methods include tree search algorithms like LATS \cite{zhou2023language}, heuristic approaches such as PlanCritic's genetic algorithms \cite{burns2024plancritic}, and CoT-SC, which identifies recurring solutions via self-consistency checks \cite{wang2023selfconsistency}. Reward-based models like ARMAP assess intermediate and final outcomes to optimize planning \cite{chen2025armap}. This iterative exploration and refinement process enhances adaptability, ensuring robust strategies for complex problems.

\paragraph*{World Knowledge Integration}
Effective planning requires agents to navigate dynamic environments, anticipate changes, and predict outcomes, underscoring the importance of world knowledge. RAP \cite{hao2023reasoning} examines the interplay between LLMs, agent systems, and world models, positioning LLMs as dual-purpose entities: as world models, they predict state changes following actions \cite{gu2024your, zhao2024large}; as agents, they select actions based on states and goals \cite{yao2022react}. This framework mirrors human cognition—simulating action consequences before selecting optimal paths—and unifies language models, agent models, and world models as pillars of machine reasoning \cite{hu2023language}.

Agents augment LLM capabilities by integrating external knowledge, addressing gaps in world understanding. ReAct employs an action-observation loop to gather environmental feedback, combining real-time data with linguistic knowledge to improve decision-making in complex scenarios \cite{yao2022react}. This enables LLMs to iteratively refine their world models during action execution, supporting adaptive planning. Conversely, LLM+P \cite{liu2023llmp} integrates LLMs with the PDDL planning language, converting natural language inputs into formalized representations solved by classical planners \cite{Mahdavi2024Leveraging, guan2023leveraging}. This hybrid approach compensates for LLMs' limitations in structured planning, merging their linguistic flexibility with the reliability of traditional systems.

Further advancements enhance LLM planning through world knowledge integration. CodePlan \cite{kambhampati2024position} uses code-form plans—pseudocode outlining logical steps—to guide LLMs through complex tasks, achieving notable performance improvements across benchmarks \cite{wen2024unlocking}. The World Knowledge Model (WKM) equips LLMs with prior task knowledge and dynamic state awareness, reducing trial-and-error and hallucinations in simulated environments \cite{qiao2024Large}. Unlike language-centric models, the Joint Embedding Predictive Architecture (JEPA) leverages self-supervised learning from multimodal data to construct predictive world models, efficiently capturing dynamic environmental changes and offering a robust framework for adaptive planning \cite{lecun2017futureoflife}. A neuro-symbolic approach combining Linear Temporal Logic with Natural Language (LTL-NL) integrates formal logic with LLMs, leveraging implicit world knowledge to ensure reliable, adaptive planning \cite{wang2023conformal}. Together, these methods illustrate how structured frameworks and environmental understanding can transform LLMs into effective planners.

\section{Summary and Discussion}
\label{sec:ch-cognition-summary}

\lettrine[lines=3]{\initfamily\textcolor{darkgreen}{T}}{he} cognitive evolution of intelligent agents represents a spiraling progression of learning and reasoning, much like how biological intelligence achieve higher reasoning efficiency after acquiring better knowledge and logic. Our unified framework clearly describes this spiral relationship: the learning process continuously optimizes mental states through $M_t = L(M_{t-1}, a_{t-1}, o_t)$, while the reasoning process makes better decisions based on updated mental states through $a_t = R(M_t)$. This mutually reinforcing mechanism enables agents to continuously enhance cognitive capabilities through experience accumulation. Richer experiential data improves learning effectiveness, while better mental state representations support more efficient reasoning processes.

Intelligent agents possess remarkable diversity in learning forms. Without changing model parameters, prompt engineering, workflow design, few-shot examples, memory mechanisms, action space definitions, and even agent logic itself can be learned. When model parameters are modified, data acquisition strategies (how agents explore environments or collect external knowledge) and various training paradigms (SFT, LoRA, RL, etc.) all provide possibilities for capability enhancement. This rich learning space presents both opportunities and challenges. The key question is: \textbf{when facing such a broad learning space, we need to master optimal learning forms for agents in different scenarios.} Furthermore, true Foundation Agents should possess meta-learning capabilities: understanding the relationship between scenarios and learning forms to automatically construct appropriate agent architectures and continuously improve them. This requires agents not only to execute learning but to learn how to learn.

However, enabling agents to engage in active learning faces a fundamental difficulty. We can provide agents with clear external goals to drive learning, but this logic differs essentially from humans. Human behavior stems from internal reward signals produced by hormonal systems, with basic motivations being survival and reproduction. In contrast, agents (particularly language models) naturally lack such internal motivation settings, causing their behavior to remain passive and reactive rather than active and exploratory. We propose a key hypothesis: \textbf{solving the internal reward system problem for agents is necessary to truly motivate agent behavior, including active learning capabilities}. This is not merely a technical challenge but involves designing reasonable internal drive mechanisms for artificial systems to generate spontaneous exploration and improvement motivation like biological agents.

Another critical issue lies in the gap between the capability requirements for agent reasoning and current reasoning research directions. Reasoning is essentially the effective use of internal states after learning, which contains two core components: cognitive ability for internal logic and understanding of external world dynamics. Existing work on improving language model reasoning capabilities mostly focuses on mathematical and logical domains, dedicated to enhancing internal logic exploration abilities. However, relatively few works consider how to enhance language models' or agents' understanding capabilities of external world dynamics.

This raises a fundamental question: \textbf{are agents' poor performance in diverse environments due to insufficient logical reasoning capabilities or inadequate understanding of world dynamics?} Recent research provides important insights into this question. \citet{shenbai2025tti} demonstrated that agents acquiring as much current environmental dynamic information as possible during reasoning can significantly improve agent capabilities, indicating the crucial role of environmental understanding in reasoning. Meanwhile, ActRe~\cite{yang2024reactx} further proved the important impact of world dynamics understanding capabilities embedded within models on reasoning performance. These findings suggest that understanding of world dynamics may be the key bottleneck limiting agent reasoning capabilities.

Current reasoning research focuses too much on how to think better while relatively neglecting what to think based on. True reasoning capability improvement may require shifting attention from pure logical structure optimization toward joint enhancement of logical capabilities and world understanding, demanding we examine the relationship between reasoning and world modeling.

In conclusion, the future of cognition in Foundation Agents lies in their ability to balance the need for sophisticated learning mechanisms with the requirement for internal motivation systems. By continuing to explore and refine the interplay between passive external guidance and active internal drives, between structured reasoning capabilities and adaptive world understanding, we move closer to developing AI systems that not only can learn and reason effectively but can autonomously drive their own cognitive evolution in an ever-changing world.

\chapter{Memory}
\label{chap:memory}

\lettrine[lines=3]{\initfamily\textcolor{darkgreen}{M}}{emory} is fundamental to both human and artificial intelligence. For humans, it serves as the bedrock of cognition, a vast repository of experiences and knowledge that empowers us to learn, adapt, and navigate the complexities of the world. From infancy, our capacity to encode, store, and retrieve information underpins our ability to acquire language, master skills, and build relationships. Decades of research in neuroscience and cognitive psychology have illuminated the multifaceted role of memory, revealing its influence on our sense of self, creative endeavors, and decision-making processes. Similarly, in the burgeoning field of artificial intelligence, memory is increasingly recognized as a cornerstone of intelligent behavior. Just as humans rely on past experiences to inform present actions, intelligent agents require robust memory mechanisms to tackle intricate tasks, anticipate future events, and adjust to dynamic environments. Therefore, a deep understanding of human memory – its organization, processes, and limitations – provides invaluable insights for the development of more capable and adaptable AI systems. This section will first provide a concise overview of human memory, focusing on the key stages of encoding, consolidation, and retrieval. We will then transition to exploring the diverse approaches employed in designing intelligent agent memory systems. A critical comparison between these artificial memory systems and their human counterparts will highlight existing gaps in areas such as adaptability, contextual understanding, and resilience. Finally, we will consider how principles derived from neuroscience and cognitive psychology can inform future research, suggesting directions that may lead to the creation of artificial memory systems that exhibit greater robustness, nuance, and ultimately, a closer resemblance to the remarkable capabilities of human memory.

\section{Overview of Human Memory}
\label{sec:human-memory-overview}

\lettrine[lines=3]{\initfamily\textcolor{darkgreen}{U}}{nderstanding} human memory is central to both cognitive science and the design of intelligent systems. Memory enables organisms to accumulate knowledge, adapt behavior based on past experience, and project into the future. In humans, memory is not a monolithic faculty but a constellation of interacting subsystems that differ in temporal span, representational format, and accessibility to consciousness. These systems collectively support a wide range of functions, from the rapid detection of sensory stimuli to the deliberate recollection of life events and the fluent execution of learned skills. In this section, we survey foundational classifications of memory and examine prominent theoretical models, each offering a distinct lens through which to understand how information is encoded, stored, and retrieved in the human mind.

\subsection{Types of Human Memory}
\label{subsubsec:mem-type}

Human memory is often conceptualized as a multi-tiered system that captures, stores, and retrieves information at different levels of processing and timescales. Researchers from the fields of cognitive science, neuroscience, and psychology have proposed various models to describe these levels. A commonly accepted hierarchy distinguishes between sensory memory, short-term memory (including working memory), and long-term memory~\cite{AtkinsonShiffrin1968,fox2017multiplicity}. Within long-term memory, explicit (declarative) and implicit (non-declarative) forms are further delineated~\cite{baddeley1992working}. Figure~\ref{fig:human-mem} illustrates one such hierarchical framework:

\begin{figure}[!ht]
\begin{center}

\begin{forest}
    for tree={draw, 
             font = \sffamily\small\linespread{0.8}\selectfont,
             fill = customgreen!50, 
            align = center,
             grow = south,
        forked edge,        
            s sep = 4mm,    
            l sep = 6mm,    
         fork sep = 3mm,    
              }
[Human\\Memory, fill=customblue!50 
    [Long-Term\\Memory
        [Declarative\\(Explicit) Memory
            [Semantic\\Memory]
            [Episodic\\Memory
                [Autobiographical\\Memory]
            ]
        ]
        [Non-Declarative\\(Implicit) Memory
            [Procedural\\Memory]
            [Priming]
            [Classical\\Conditioning]
            [Non-Associative\\Memory]
        ]
     ]
    [Sensory\\Memory]
    [Short-Term\\Memory
        [Working\\Memory]
    ]
]
\end{forest}

\end{center}
\caption{The hierarchical taxonomy of the human memory system.}
\label{fig:human-mem}
\end{figure}

\begin{itemize}
\item \textbf{Sensory Memory.} Sensory memory is the initial, brief store of raw sensory information. It maintains inputs from the environment for a duration ranging from milliseconds to a few seconds, allowing subsequent processes to determine which portions of the stimulus are relevant for further analysis~\cite{sperling1960information}. Iconic memory (for visual input) \cite{coltheart1980iconic} and echoic memory (for auditory input) \cite{gardiner1983recency} are two well-known subtypes.

\item \textbf{Short-Term Memory and Working Memory.} Short-term memory (STM) involves holding a limited amount of information in an easily accessible state for seconds to under a minute. The term \emph{working memory} is often used to emphasize the manipulation of that information rather than mere maintenance. While some models treat working memory as a subset of STM, others view it as a distinct system that manages both the storage and active processing of data (for instance, performing arithmetic in one's head) \cite{aben2012distinction,cowan2008differences}. The capacity of STM or working memory is finite, typically cited as around seven plus or minus two chunks of information~\cite{miller1956magical}, though individual differences and task factors can modulate this figure.

\item \textbf{Long-Term Memory (LTM).} Long-term memory accommodates the more durable storage of information that can persist from hours to decades \cite{shiffrin1969storage,norris2017short}. This repository supports the learning of skills, the acquisition of factual knowledge, and the recollection of personal experiences. Although long-term memory is sometimes described as having a vast or near-unlimited capacity, factors such as decay, interference, and retrieval cues influence the extent to which stored information can be accessed~\cite{ebbinghaus1913memory}.

\begin{itemize}
\item \textbf{Declarative (Explicit) Memory.} Declarative memory encompasses memories that can be consciously recalled and articulated \cite{eichenbaum1997declarative}. Within this broad category, researchers often discuss:
\begin{itemize}
    \item \textbf{Semantic Memory:} Factual knowledge about the world, including concepts, words, and their relationships \cite{kumar2021semantic}. Examples include recalling the meaning of vocabulary terms or knowing the capital city of a country.
    \item \textbf{Episodic Memory:} Personally experienced events that retain contextual details such as time, place, and the people involved \cite{tulving2002episodic}. This form of memory allows individuals to mentally travel back in time to relive past experiences.
    \item \textbf{Autobiographical Memory:} A form of episodic memory focusing on events and experiences related to one's personal history \cite{fivush2011development}. While sometimes treated as a sub-category of episodic memory, autobiographical memory places particular emphasis on the self and its evolving life narrative.
\end{itemize}

\item \textbf{Non-Declarative (Implicit) Memory.} Non-declarative memory refers to memories that influence behavior without the need for conscious awareness \cite{squire1992declarative}. Key subtypes include:
\begin{itemize}
    \item \textbf{Procedural Memory:} The gradual acquisition of motor skills and habits (e.g., riding a bicycle, typing on a keyboard) that become automatic with repetition~\cite{gupta2002theoretical,cohen1980preserved}.
    \item \textbf{Priming:} The phenomenon in which prior exposure to a stimulus influences subsequent responses, often without explicit recognition of the previous encounter \cite{tulving1990priming}.
    \item \textbf{Classical Conditioning:} The learned association between two stimuli, where one stimulus comes to elicit a response originally produced by the other \cite{clark2002classical}.
    \item \textbf{Non-Associative Memory:} Adaptive modifications in behavior following repeated exposure to a single stimulus. Habituation (reduced response to a repeated, harmless stimulus) and sensitization (increased response after exposure to a noxious or intense stimulus) are representative examples~\cite{ioannou2021non,conway2000construction}.
\end{itemize}
\end{itemize}

\end{itemize}

Despite the orderly appearance of these categories, human memory processes often overlap and intersect. For example, autobiographical memory is typically nested within episodic memory; yet, its particular focus on self-relevant experiences leads some theorists to treat it as a distinct category. Similarly, the boundary between short-term and working memory can differ depending on the theoretical perspective. Some scholars prefer a more functional, process-oriented view of working memory, while others employ a strictly capacity-oriented concept of short-term storage. In each case, these different perspectives on memory highlight the complexity and nuance of human cognition.

\subsection{Models of Human Memory}
\label{subsubsec:human-mem-model}

Human memory has inspired a wide range of theoretical models, each offering different insights into how information is acquired, organized, and retrieved. Although no single framework commands universal agreement, several influential perspectives have shaped the discourse in cognitive science, neuropsychology, and AI research. The following content highlights some of the most prominent models and architectures used to explain memory's multiple facets.

\begin{figure}[!ht]
\centering
    \includegraphics[width=0.8\columnwidth]{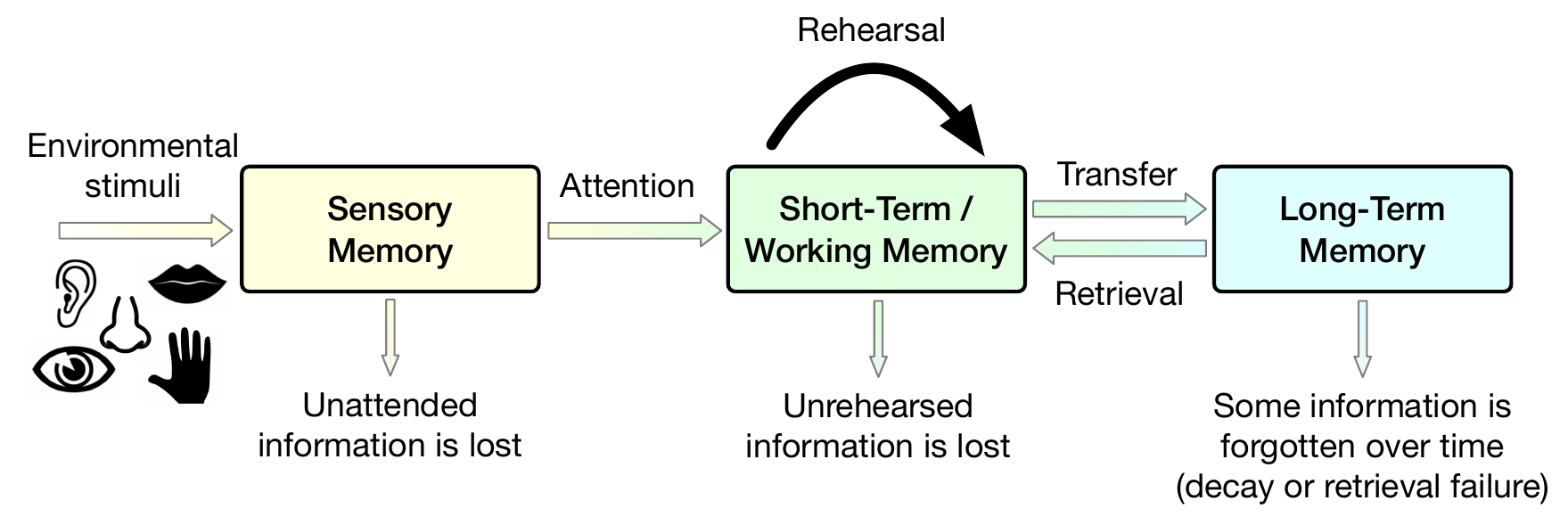}
    \caption{Atkinson-Shiffrin three-stage model of human memory~\cite{AtkinsonShiffrin1968}.}
\label{fig:3-stage-mem}
\end{figure}

\paragraph*{The Multi-Store (Modal) Model}
A seminal proposal by Atkinson and Shiffrin \cite{AtkinsonShiffrin1968} introduced the multi-store or ``modal'' model, which posits three main stores for incoming information: \emph{sensory memory}, \emph{short-term memory}, and \emph{long-term memory}. Control processes (e.g., attention, rehearsal) regulate how data transitions across these stores. Figure~\ref{fig:3-stage-mem} illustrates this model of memory. Despite its relative simplicity, this model remains foundational for understanding how fleeting sensory impressions eventually form stable, long-lasting representations.

\begin{figure}[!ht]
\centering
    \includegraphics[width=0.55\columnwidth]{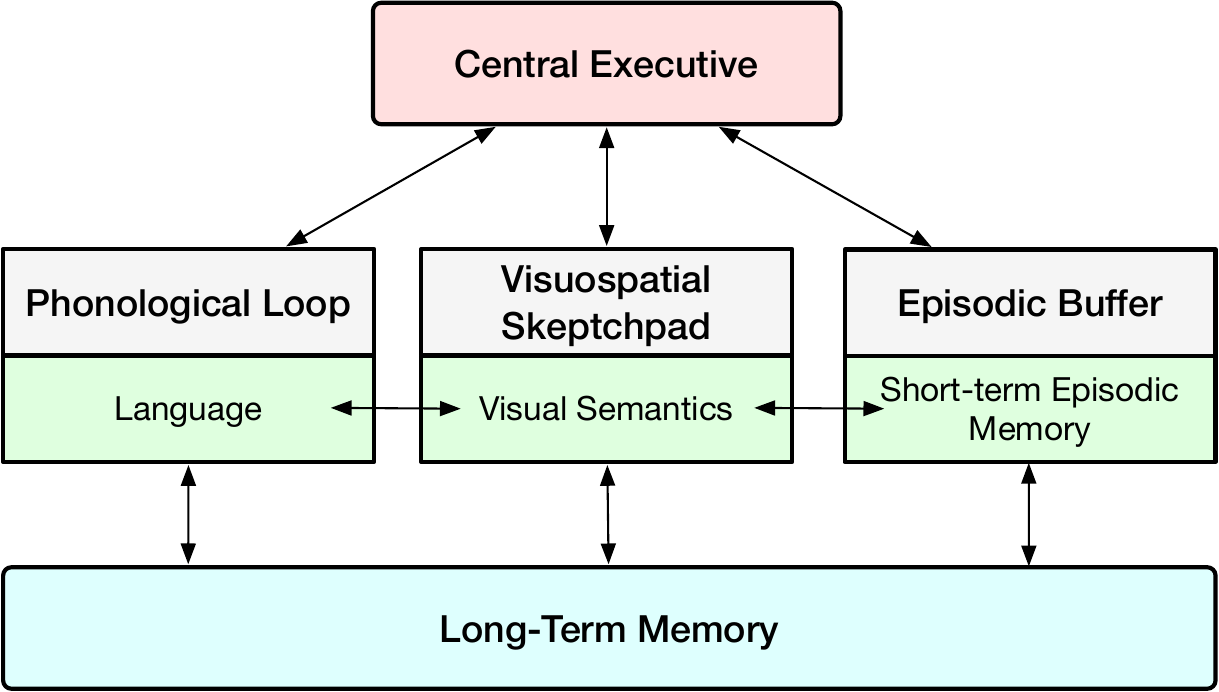}
    \caption{Baddeley's model of working memory~\cite{BaddeleyHitch1974}.}
\label{fig:wm-mem}
\end{figure}

\paragraph*{Working Memory Models}
Recognizing that short-term memory also involves active maintenance, Baddeley and Hitch \cite{BaddeleyHitch1974} proposed a \emph{working memory} framework emphasizing the dynamic manipulation of information. Their original model described a central executive that coordinates two subsystems: the phonological loop (verbal) and the visuospatial sketchpad (visual/spatial). A subsequent refinement introduced the episodic buffer to integrate material from these subsystems with long-term memory \cite{Baddeley2000}. Figure~\ref{fig:wm-mem} shows the framework of the working memory model. Alternatives such as Cowan's embedded-processes model \cite{Cowan1988} similarly underscore the role of attention in governing how information is briefly sustained and manipulated.

\begin{figure}[!ht]
\centering
    \includegraphics[width=0.55\columnwidth]{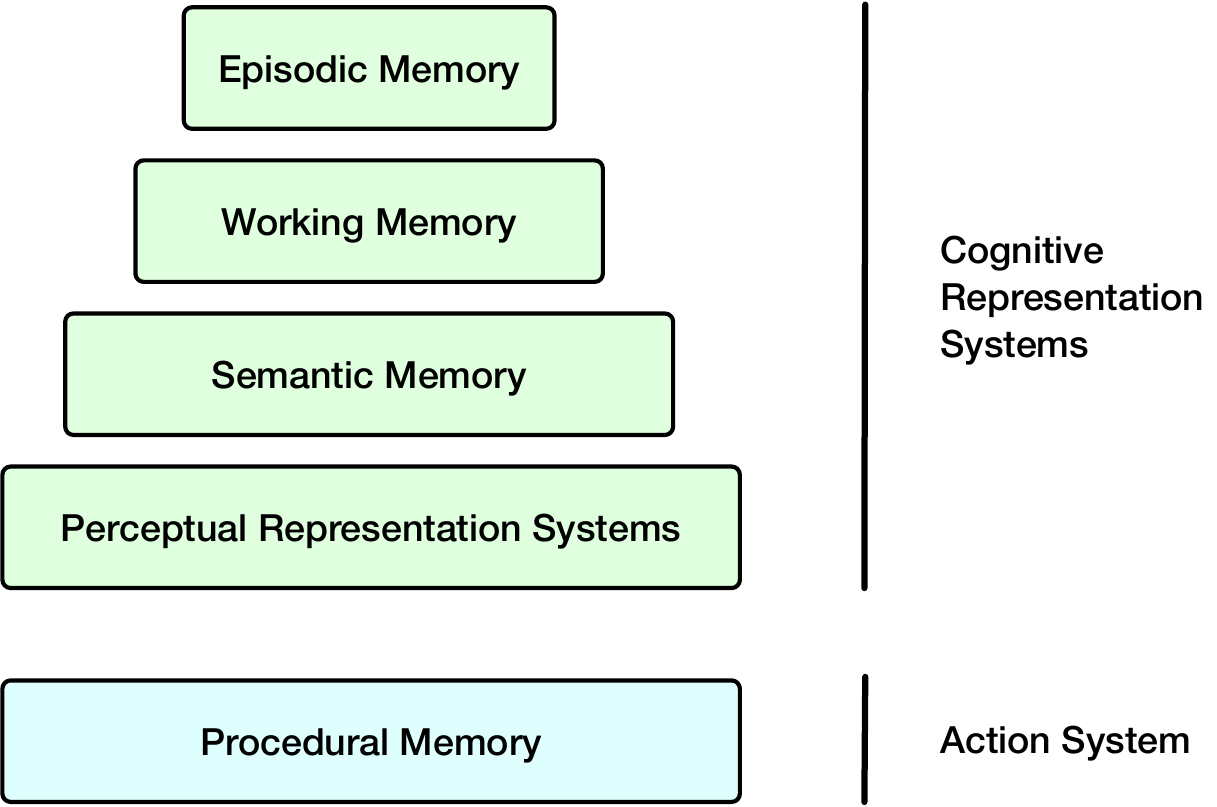}
    \caption{The Serial-Parallel Independent (SPI) model of human memory~\cite{Tulving1985}.}
\label{fig:spi-mem}
\end{figure}

\paragraph*{Serial-Parallel-Independent (SPI) Model}
Initial distinctions between episodic, semantic, and procedural memory were championed by Tulving \cite{Tulving1985}, who later refined his ideas into the Serial-Parallel-Independent (SPI) model, as shown in Figure~\ref{fig:spi-mem}. In this framework, memory is divided into two overarching systems. The \emph{cognitive representation system} handles perceptual input and semantic processes, encompassing facts, concepts, and contextual (episodic) knowledge. The \emph{action system}, by contrast, underpins procedural skills such as dance routines, driving maneuvers, or typing proficiency. 
Tulving's SPI model posits that memory formation can occur at multiple levels: strictly perceptual encoding can support rudimentary episodic memories, while richer episodic representations benefit from semantic mediation. For instance, patients with semantic dementia, who struggle to retain word meanings, can still form some episodic memories but often lack the full contextual detail conferred by intact semantic networks. By highlighting the role of procedural memory and its automatic, intuitive nature, the SPI model aims to integrate structure (the content of memory) and function (how memory is used), surpassing earlier accounts that largely focused on explicit storage and retrieval. Despite these strengths, critics note that the model under-specifies how working memory operates within the broader system, and the feedback mechanisms connecting cognitive and action subsystems remain loosely defined.

\paragraph*{Global Workspace Theory (GWT) and the IDA/LIDA Framework} Global Workspace Theory (GWT), developed by Baars \cite{baars1993cognitive}, conceptualizes consciousness and working memory as a ``broadcast'' mechanism that distributes information to specialized processors. Building on GWT, Franklin \cite{Franklin1997,FranklinKelemenMcCauley1998} proposed the \emph{IDA (Intelligent Distribution Agent)} model, later extended to \emph{LIDA (Learning IDA)}, as a comprehensive cognitive architecture. In these frameworks, multiple memory systems (e.g., perceptual, episodic, procedural) interact through iterative ``cognitive cycles'', with a global workspace functioning as a hub for attention and decision-making. From an AI standpoint, IDA/LIDA demonstrates how human-like memory processes can be operationalized to guide an agent's perception, action selection, and learning.

\begin{figure}[!ht]
\centering
    \includegraphics[width=0.5\columnwidth]{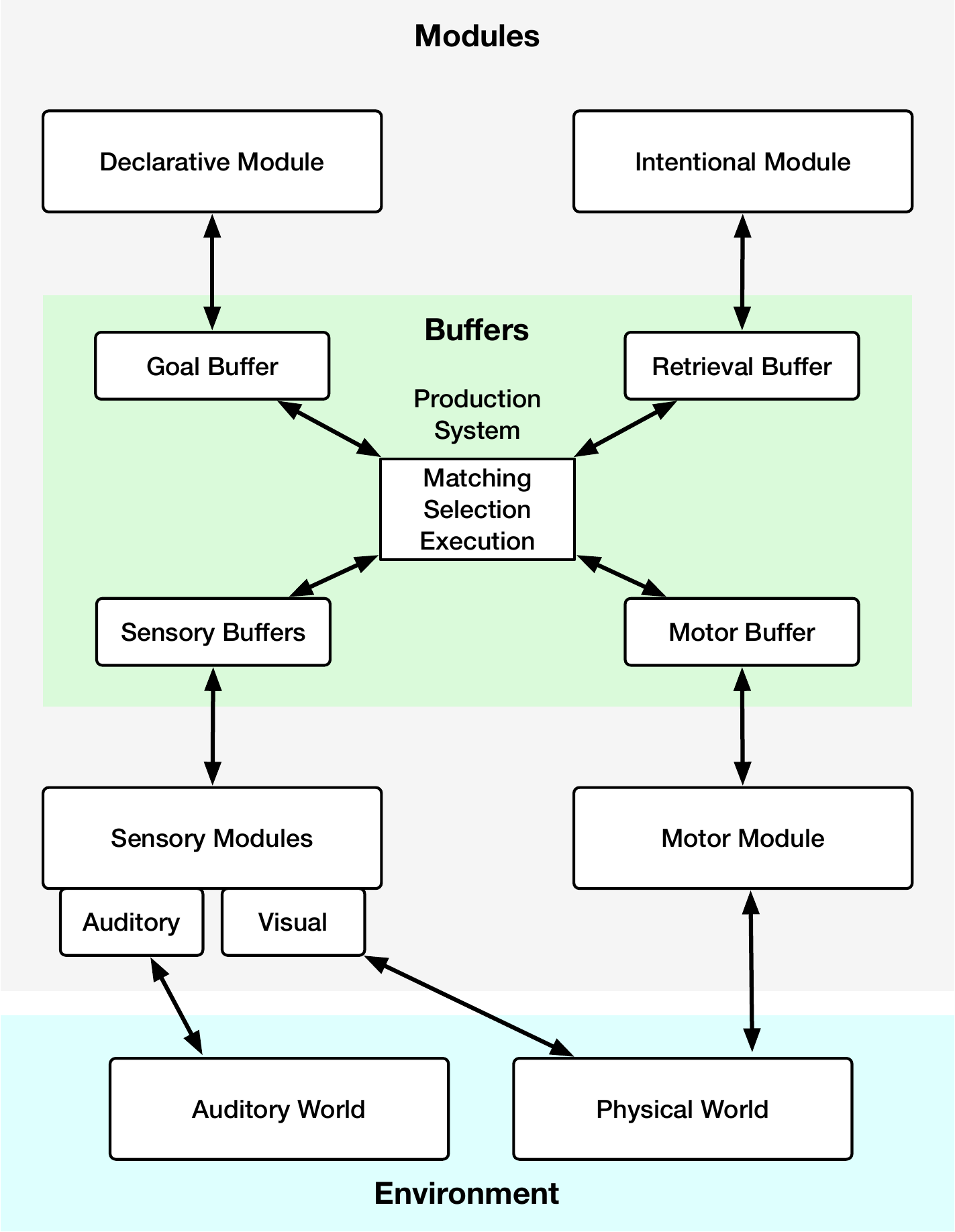}
    \caption{An abstraction of the most important processes in the ACT-R model~\cite{Anderson2007}.}
\label{fig:act-R}
\end{figure}

\paragraph*{ACT-R and Cognitive Architectures} 
ACT-R (Adaptive Control of Thought—Rational) \cite{Anderson2007} is a cognitive architecture that integrates memory, perception, and motor processes into a unified theoretical framework. It has been applied extensively across diverse domains, including learning and memory, problem-solving, decision-making, language comprehension, perception and attention, cognitive development, and individual differences. 
Figure~\ref{fig:act-R} illustrates the processes of ACT-R. At the core of ACT-R are distinct \emph{modules} (e.g., visual, manual, declarative, procedural) that interact with the system through dedicated \emph{buffers}. Declarative memory stores factual ``chunks'' while procedural memory encodes if–then production rules for actions and strategies. Cognition unfolds via a \emph{pattern matcher} that selects a single production to fire based on the current buffer contents. This symbolic production system is augmented by subsymbolic processes, guided by mathematical equations that dynamically regulate activations, retrieval latencies, and production utilities. By combining symbolic and subsymbolic levels, ACT-R provides a mechanistic account of how individuals acquire, retrieve, and apply knowledge, thus shedding light on empirical phenomena such as reaction times, error patterns, and the shaping of learning over time.

In summary, each of the aforementioned models illuminates different aspects of memory. For example, the multi-store model provides a straightforward introduction to storage stages, working memory models emphasize active maintenance and manipulation, and frameworks such as IDA/LIDA or ACT-R embed memory within a comprehensive view of cognition. In practice, researchers often draw upon multiple perspectives, reflecting the intricate nature of human memory and its integral role in perception, learning, and adaptive behavior.

\section{From Human Memory to Agent Memory}
\label{subsec:human-to-agent-mem}

\lettrine[lines=3]{\initfamily\textcolor{darkgreen}{H}}{aving} established the fundamentals of human memory, we now focus on how Large Language Model (LLM)-based agents manage and store information. Memory is not merely a storage mechanism but is fundamental to human and artificial intelligence. Memory underpins cognition, enabling learning, adaptation, and complex problem-solving for humans. Similarly, for LLM-based agents, memory provides the crucial scaffolding for maintaining context, learning from experience, and acting coherently over time. Without memory, even a highly capable LLM would struggle to adapt to changing circumstances or maintain focus during extended interactions.

While LLM-based agents and biological systems differ fundamentally, the principles guiding human memory—context retention, selective forgetting, and structured retrieval—are highly relevant to agent design. Therefore, examining the parallels and distinctions between human and artificial memory is beneficial. Functionally, we can draw analogies: an agent's short-term memory buffer resembles the prefrontal cortex's role in working memory, while long-term storage in a vector database is akin to the hippocampus's function in consolidating episodic memories. Agent memory design can benefit from emulating human memory's mechanisms, including selective attention, prioritized encoding, and cue-dependent retrieval. However, crucial differences exist.

Human memory, built upon biological neural networks, integrates storage and computation within neurons' connections and activity patterns. This offers a high degree of parallelism and adaptability. In contrast, current agent memory systems predominantly rely on digital storage and algorithms, using symbolic representations and logical operations, thus separating storage and computation. This impacts information processing: human memory is associative and dynamic, capable of fuzzy matching and creative leaps, while current agent memory relies on precise matching and vector similarity, struggling with ambiguity. Although digital storage capacity is vast, it cannot yet replicate the complexity and dynamism of human memory, particularly in nuanced pattern recognition and long-term stability. Human memory, while imperfect, excels at extracting crucial information from noisy data. Agent memory systems, in their current stage, are still nascent compared to the intricacies of human memory, facing limitations in organization, integration, adaptive forgetting, and knowledge transfer.

The need for a dedicated memory module in LLM-based agents is paramount. While external knowledge bases (databases, search engines, APIs)~\cite{cheng2024exploring} provide valuable information, they do not capture the agent's internal reasoning, partial inferences, or task-specific context. An agentic memory system internalizes interim steps, evolving objectives, and historical dialogue, enabling self-referential exploration and adaptation. This is crucial for tasks requiring the agent to build upon prior judgments or maintain a personalized understanding of user goals.

Early approaches to agent memory, such as appending conversation history to the input prompt (a rudimentary form of working memory)~\cite{baddeley2010working}, have evolved. Modern architectures employ more sophisticated techniques, including vector embeddings for rapidly retrieving memories~\cite{camacho2018word} and selective incorporation of reasoning chains into subsequent inference steps~\cite{liu2023think,zhang2024you}. These diverse methods share the common goal of managing a large information reservoir without compromising system responsiveness.

However, compared to the sophistication of human memory, current agentic methods have limitations. Many systems lack coherent strategies for long-term memory consolidation, leading to cluttered logs or abrupt information loss. The flexible, bidirectional interplay between stored knowledge and ongoing processing, characteristic of human working memory, is often absent. Metacognitive oversight, selective recall, forgetting, and vigilance against outdated information are also underdeveloped in LLM-based agents. Balancing comprehensive recall with practical efficiency, as humans do, remains a key challenge.

Building robust and adaptable memory for LLM-based agents involves addressing three core research questions: First, how should memory be represented to capture diverse information types and facilitate efficient access? Second, how can agent memory evolve, incorporating new experiences, adapting to changing contexts, and maintaining consistency? Finally, how can the stored memories effectively enhance reasoning, decision-making, and overall agent performance? Figure~\ref{fig:tree-memory} shows an overview of selected relevant research works. The following sections delve into these crucial areas, exploring current approaches, limitations, and potential future directions.

\section{Representation of Agent Memory}
\begin{figure*}[!t]
\centering
\small
\resizebox{0.99\textwidth}{!}{%
    \begin{forest}
        for tree={
            forked edges,
            draw,
            rounded corners,
            node options={align=center,},
            s sep=6pt,
            calign=center,
            grow=east,
            reversed=true,
            anchor=base west,
            parent anchor=east,
            child anchor=west,
            base=left,
            font=\small,
          },
          where level=1{text width=6em,fill=customblue!50}{},
          where level=2{text width=6em,fill=customgreen!50}{},
        [Memory, text width=4em, fill=gray!20
            [Representation, for tree={
                answer, text width=7em,
                calign=child edge, calign child=(n_children()+1)/2
                }
                [Sensory
                    [Text-based
                        [RecAgent~\cite{wang2023user} CoPS~\cite{zhou2024cognitive} 
                        MemoryBank~\cite{zhong2024memorybank} 
                        Memory Sandbox~\cite{huang2023memory}, text width=26em]
                    ]
                    [Multi-modal
                        [VideoAgent~\cite{fan2024videoagent} WorldGPT~\cite{ge2024worldgpt} 
                        Agent S~\cite{agashe2024agent}
                        OS-Copilot~\cite{wu2024copilot} 
                        MuLan~\cite{li2024mulan}, text width=26em]
                    ]
                ]
                [Short-term 
                    [Context
                        [MemGPT~\cite{packer2023memgpt} 
                        KARMA~\cite{wang2024karma} 
                        LSFS~\cite{shi2024commands}
                        OSCAR~\cite{wang2024oscar}
                        RCI~\cite{Geunwoo2023LMComputer}, text width=26em]
                    ]
                    [Working
                        [Generative Agent~\cite{park2023generative} 
                        RLP~\cite{fischer2023reflective} 
                        CALYPSO~\cite{zhu2023calypso}
                        HiAgent~\cite{hu2024hiagent}, text width=26em]
                    ]
                ]
                [Long-term
                    [Semantic
                        [AriGraph~\cite{anokhin2024arigraph} RecAgent~\cite{wang2023user} 
                        HippoRAG~\cite{gutierrez2024hipporag}, text width=26em]
                    ]
                    [Episodic
                        [MobileGPT~\cite{lee2023explore}
                        MemoryBank~\cite{zhong2024memorybank} 
                        Episodic Verbalization~\cite{barmann2024episodic}
                        MrSteve~\cite{park2024mr}, text width=26em]                   
                    ]
                    [Procedural
                        [AAG~\cite{roth2024pairing} Cradle~\cite{tan2024towards} 
                        JARVIS-1~\cite{wang2024jarvis} 
                        LARP~\cite{yan2023larp}, text width=26em]                  
                    ]
                ]
            ]
             [Lifecycle, for tree={
                answer, text width=7em,
                calign=child edge, calign child=(n_children()+1)/2,
                }
                [Acquisition
                    [Information Compression
                        [HiAgent~\cite{hu2024hiagent} LMAgent~\cite{liu2024lmagent}
                        ReadAgent~\cite{lee2024human}
                        M$^2$WF~\cite{wang2025leveraging}, text width=26em]
                    ]
                    [Experience Consolidation
                        [ExpeL~\cite{zhao2024expel}
                        MindOS~\cite{hu2025unified}
                        ~\cite{vanschoren2018meta}
                        ~\cite{hou2024my}, text width=26em]
                    ]
                ]
                [Encoding
                    [Selective Attention
                        [AgentCorrd~\cite{pan2024agentcoord}
                        MS~\cite{gao2024memory}
                        GraphVideoAgent~\cite{chu2025understanding}
                        A-MEM~\cite{xu2025mem}
                        ~\cite{ali2024robots}, text width=26em]
                    ]
                    [Multi-modal Fusion
                        [Optimus-1~\cite{li2024optimus} 
                        Optimus-2~\cite{li2025optimus}
                        JARVIS-1~\cite{wang2024jarvis}, text width=26em]
                    ]
                ]
                [Derivation
                    [Reflection
                        [Agent S~\cite{agashe2024agent}
                        OSCAR~\cite{wang2024oscar}
                        R2D2~\cite{huang2025r2d2}
                        Mobile-Agent-E~\cite{wang2025mobile}, text width=26em]
                    ]
                    [Summarization
                        [SummEdits~\cite{laban2023summedits}
                        SCM~\cite{wang2023enhancing}
                        Healthcare Copilot~\cite{ren2024healthcare}
                        ~\cite{wang2023recursively}, text width=26em]
                    ]
                    [Knowledge Distillation
                        [Knowagent~\cite{zhu2024knowagent}
                        AoTD~\cite{shi2024unlocking}
                        LDPD~\cite{liu2024language}
                        Sub-goal Distillation~\cite{hashemzadeh2024sub}
                        MAGDi~\cite{chen2024magdi}, text width=26em]
                    ]
                    [Selective Forgetting
                        [Lyfe Agent~\cite{kaiya2023lyfe}
                        TiM~\cite{liu2023think}
                        MemoryBank~\cite{zhong2024memorybank}
                        S$^3$~\cite{gao2023s3}
                        ~\cite{hou2024my}, text width=26em]
                    ]
                ]
                [Retrieval
                    [Indexing
                        [HippoRAG~\cite{gutierrez2024hipporag}
                        TradingGPT~\cite{li2023tradinggpt}
                        LongMemEval~\cite{wu2024longmemeval}
                        SeCom~\cite{pan2025memory}, text width=26em]
                    ]
                    [Matching
                        [Product Keys~\cite{lample2019large}
                        OSAgent~\cite{xu2024osagent}
                        ~\cite{bahdanau2014neural}
                        ~\cite{hou2024my}, text width=26em]
                    ]                    
                ]
                [Neural Memory
                    [Associative Memory
                        [Hopfield Networks~\cite{demircigil2017model,ramsauer2020hopfield}
                        Neural Turing Machines~\cite{falcon2022neural}, text width=26em]
                    ]
                    [Parameter Integration
                        [MemoryLLM~\cite{wang2024memoryllm}
                        SELF-PARAM~\cite{wang2024self}
                        MemoRAG~\cite{qian2024memorag}
                        TTT-Layer~\cite{sun2024learning}
                        Titans~\cite{behrouz2024titans}
                        R$^3$Mem~\cite{wang2025r}, text width=26em]
                    ]                    
                ]
                [Utilization
                    [RAG
                        [RAGLAB~\cite{zhang2024raglab}
                        Adaptive Retrieval~\cite{mallen2023not}
                        Atlas~\cite{farahani2024deciphering}
                        ~\cite{yuan2025personalized}, text width=26em]
                    ]
                    [Long-context Modeling
                        [RMT~\cite{bulatov2022recurrent,bulatov2023scaling}
                        AutoCompressor~\cite{chevalier2023adapting}
                        ICAE~\cite{ge2023context}
                        Gist~\cite{mu2024learning} 
                        CompAct~\cite{yoon2024compact}, text width=26em]
                    ]
                    [Alleviating Hallucination
                        [Lamini~\cite{li2024banishing}
                        Memoria~\cite{park2023memoria}
                        PEER~\cite{he2024mixture}
                        ~\cite{ding2024retrieve}, text width=26em]
                    ]
                ]
            ]
        ]
    \end{forest}
}
    \caption{A taxonomy of selected research works about different memory modules in intelligent agents.}
    \label{fig:tree-memory}
\end{figure*}

\lettrine[lines=3]{\initfamily\textcolor{darkgreen}{I}}{nspired} by human cognitive systems~\cite{wang2003discovering}, current memory architecture in intelligent agents adopts a hierarchical framework that integrates perception through sensory memory~\cite{wang2023user}, real-time decision-making via short-term memory~\cite{kang2023think,peng2023check}, and sustained knowledge retention through long-term memory~\cite{kim2023machine,yao2023retroformer,Shinn2023ReflexionLA}. This multi-layered structure equips agents to manage immediate tasks while maintaining a broader contextual understanding, fostering adaptability and seamless continuity across diverse interactions.

Specifically, the memory system transforms raw environmental inputs into structured, actionable representations. Sensory memory acts as the gateway, capturing and selectively filtering perceptual signals to provide a foundation for cognitive processing. Short-term memory bridges these immediate perceptions with task-level understanding, buffering recent interactions and enabling dynamic adaptation through experience replay and state management. Long-term memory then consolidates and stores information over extended periods, facilitating cross-task generalization and the accumulation of enduring knowledge.

Together, these memory components form a cohesive cycle of perception, interpretation, and response. This cycle supports real-time decision-making and enables agents to learn and evolve continuously, reflecting an intricate balance between responsiveness and growth. The following delves into the formulation of each memory type, exploring their unique roles and interactions within the agent's cognitive architecture.

\subsection{Sensory Memory}
\label{subsec:sensory-mem}

\begin{figure}[!ht]
\centering
    \includegraphics[width=0.8\textwidth]{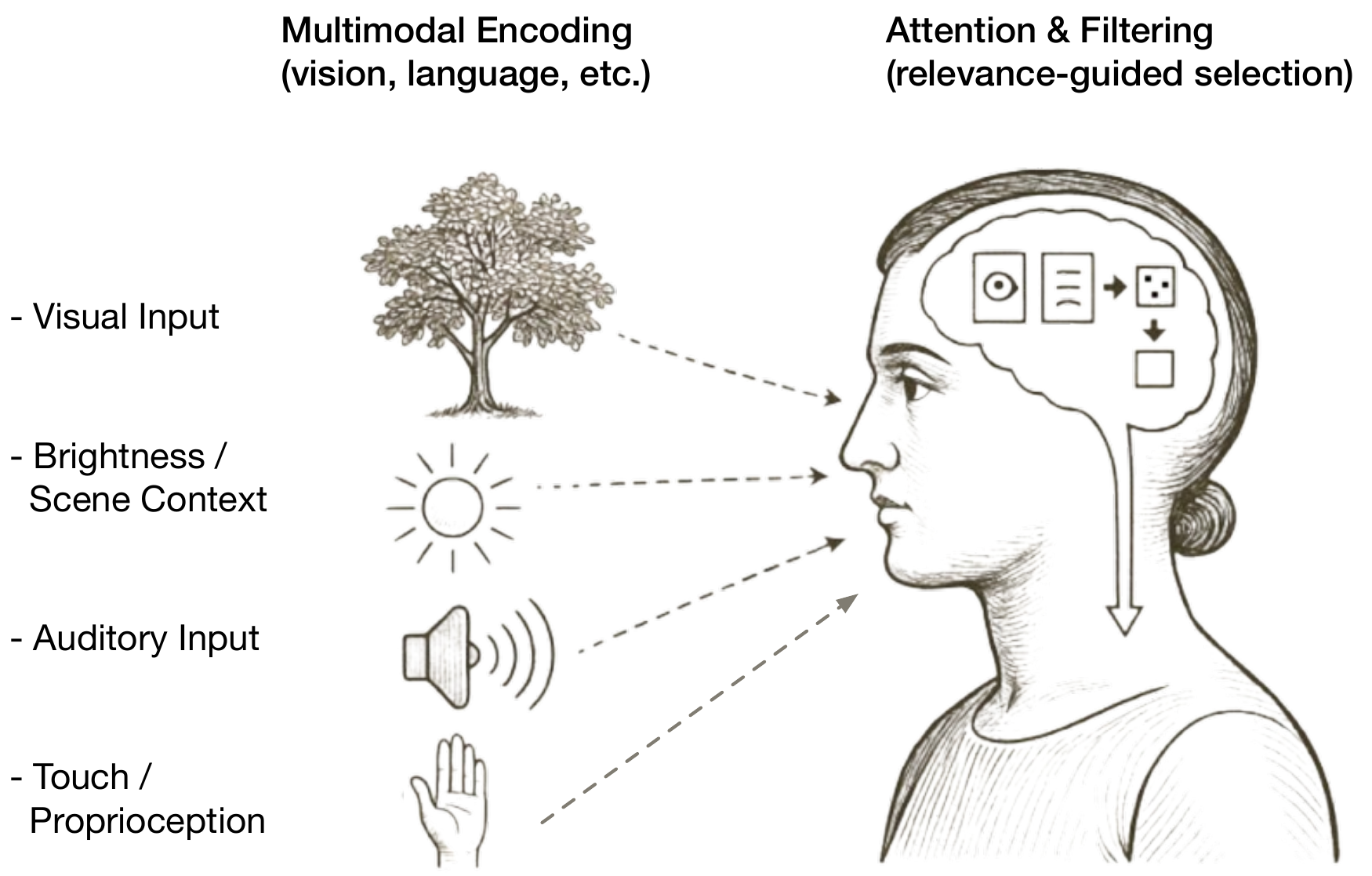}
    \caption{Illustration of sensory memory formation in intelligent agents. Multimodal sensory inputs—such as vision, sound, touch, and contextual signals—are first encoded into internal representations. An attentional mechanism filters these representations to prioritize task-relevant information, discarding noise. The selected content is temporarily retained in the sensory memory buffer, forming the foundation for higher-level cognitive processing. }
\label{fig:sensory-mem}
\end{figure}

In human cognitive systems, as shown in Figure~\ref{fig:sensory-mem}, sensory memory serves as a mechanism for collecting information through the senses, such as touch, hearing, and vision, and is characterized by its extremely brief lifespan. Analogously, sensory memory functions as the embedded representation of inputs such as text, images, and other perceptual data in intelligent agents. It represents the initial phase of environmental information processing, acting as a gateway for transforming raw observations into meaningful representations for further cognitive processing. Crucially, there's a fundamental distinction between the raw, high-bandwidth, and often analog nature of true sensory input and the already processed, typically textual or symbolic, information that many current intelligent agents receive as ``observations''. True sensory memory in humans deals with this unprocessed deluge, holding it fleetingly for the cognitive system to perform rapid, pre-attentive feature extraction and selection. For intelligent agents, especially those based on LLMs, the input is often pre-processed, such as text descriptions of a scene or tokenized language. Therefore, the ``sensory memory'' in agents often refers to the initial buffering and encoding of these already somewhat abstracted inputs. Recent research continues to explore how LLMs process and generate sensory language, often finding discrepancies compared to human patterns, potentially influenced by training data and RLHF techniques~\cite{hicke2025zero}. The challenge remains in bridging this gap: how can agents effectively process and compress truly raw, multimodal sensory data into formats (like text or embeddings) that are amenable to higher-level reasoning, and how much vital information is lost or distorted in this necessary transformation?. This initial conversion is critical, as it determines the quality and nature of the information available for all subsequent memory stages. Some recent works investigate emergent self-awareness in multimodal LLMs through sensorimotor experiences, highlighting the role of sensory integration and memory~\cite{varela2025sensorimotor}.

Sensory memory in intelligent agents transcends passive information reception. It dynamically encodes and filters perceptual signals, bridging immediate sensory inputs with the agent's internal state, objectives, and prior knowledge. This adaptive process facilitates rapid perception of environmental changes, task continuity, and real-time context-aware information processing. Sophisticated attention mechanisms are employed to ensure relevance and focus in the sensory memory layer, forming a critical foundation for decision-making and adaptation.

Formally, sensory memory formation consists of three sequential steps: \emph{perceptual encoding}, \emph{attentional selection}, and \emph{transient retention}. First, perceptual encoding transforms raw sensory signals into processable representations, mathematically expressed as:  
\begin{equation}
\phi(o_t) = \text{Encode}(o_t, s_t)
\end{equation}
where $o_t$ is the sensory input at time $t$, and $s_t$ represents the agent's state.  
For instance, RecAgent~\cite{wang2023user} employs an LLM-based sensory memory module to encode raw observations while filtering noise and irrelevant content. Extending beyond text-based perception, multimodal sensory memory systems such as Jarvis-1~\cite{wang2024jarvis}, VideoAgent~\cite{fan2024videoagent}, and WorldGPT~\cite{ge2024worldgpt} integrate multimodal foundation models to process diverse modality inputs.  

Next, attentional selection extracts crucial information from the encoded sensory data. This process, guided by an attention mechanism, is represented as:  
\begin{equation}
\alpha_t = \text{Attention}(\phi(o_t), c_t)
\end{equation}
where $\phi(o_t)$ is the encoded input, and $c_t$ denotes contextual information influencing attention.  
For example, RecAgent~\cite{wang2023user} employs an attention mechanism with an importance scoring system that assigns relevance scores to compressed observations, prioritizing critical inputs such as item-specific interactions while de-emphasizing less significant actions.

Finally, transient retention temporarily stores the selected sensory information as sensory memory:  
\begin{equation}
M_{\text{sensory}} = \{\alpha_t \mid t \in [t-\tau, t]\}
\end{equation}
Several strategies have been implemented to manage the time window. For instance, RecAgent~\cite{wang2023user} models retention by associating each observation with the timestamp corresponding to the start of a simulation round in the user behavior simulation environment, represented as a triplet $\langle$observation, importance score, timestamp$\rangle$. 
Similarly, CoPS~\cite{zhou2024cognitive} employs a fixed-size sensory memory pool as a time window, which consists of user search requests for personalized search, facilitating ``re-finding'' behavior. When a new query is received, the system first checks the sensory memory for relevant matches. If a match is found, the query is classified as a re-finding instance, enabling a rapid sensory response.

\subsection{Short-Term Memory}
\label{subsec:short-term-mem}

\begin{figure}[!ht]
\centering
    \includegraphics[width=0.85\textwidth]{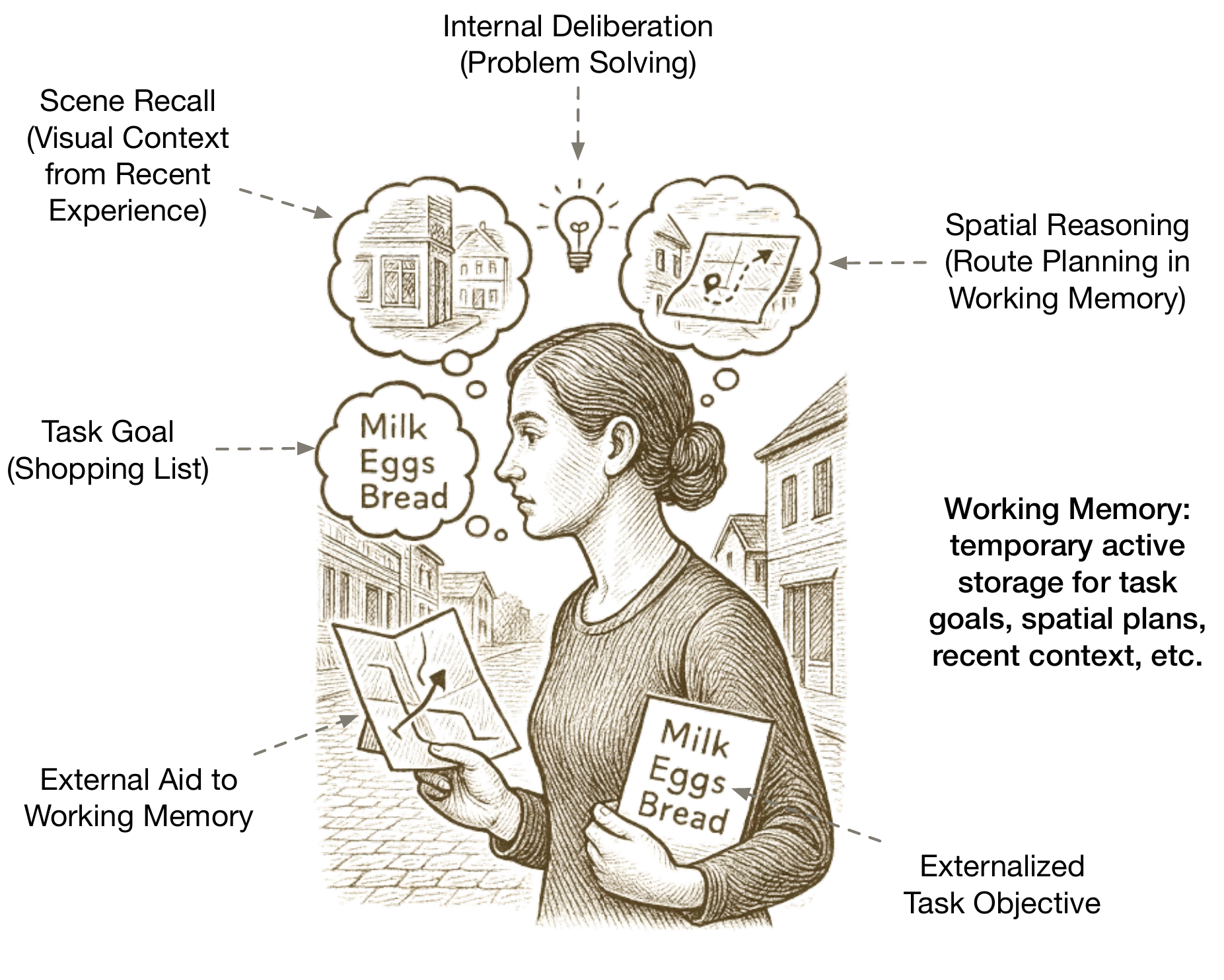}
    \caption{Illustration of short-term and working memory in action. The figure depicts an agent navigating an urban environment while holding a shopping list and a map. Around her, thought bubbles represent key contents temporarily held in working memory, including task goals (e.g., grocery items), spatial plans (e.g., route to the destination), visual recall (e.g., remembered storefronts), and active reasoning (e.g., route decisions). This dynamic workspace enables the agent to integrate recent observations, task demands, and internal deliberation to guide immediate behavior, capturing the essence of short-term memory as both transient storage and active processing for adaptive cognition. }
\label{fig:short-term-mem}
\end{figure}

Short-term memory in cognition-inspired intelligent agents serves as a transient and dynamic workspace that bridges sensory memory and long-term memory. It is essential for storing and processing task-relevant information and recent interaction sequences, supporting real-time decision-making and adaptive behavior. Inspired by human short-term and working memory, it temporarily retains information to facilitate complex cognitive tasks, ensuring continuity and coherence in the agent's operations. Figure~\ref{fig:short-term-mem} illustrates short-term memory with a vivid example.

Short-term memory in intelligent agents can be categorized into \emph{context memory} and \emph{working memory}. On the one hand, context memory treats the context window as the short-term memory of LLMs. For example, MemGPT~\cite{packer2023memgpt}, inspired by hierarchical memory systems in operating systems, manages different storage tiers to extend context beyond the LLM's inherent limitations. \cite{wang2024symbolic} introduces a neurosymbolic context memory that enhances LLMs by enabling symbolic rule grounding and LLM-based rule application.

On the other hand, working memory involves fetching and integrating relevant external knowledge to hold essential information during an agent's operation. Generative Agent~\cite{park2023generative} employs short-term memory to retain situational context, facilitating context-sensitive decision-making. Reflexion~\cite{Shinn2023ReflexionLA} utilizes a sliding window mechanism to capture and summarize recent feedback, balancing detailed immediate experiences with high-level abstractions for enhanced adaptability. RLP~\cite{fischer2023reflective} maintains conversational states for speakers and listeners, using them as prompts to support dialogue understanding and generation. 

For interactive and creative game scenarios, CALYPSO~\cite{zhu2023calypso} assists Dungeon Masters in storytelling for Dungeons \& Dragons by constructing short-term memory from scene descriptions, monster details, and narrative summaries, enabling adaptive storytelling and dynamic engagement. Similarly, Agent S~\cite{agashe2024agent} and Synapse~\cite{zheng2023synapse}, designed for GUI-based autonomous computer interaction, define their short-term memory as task trajectories, including actions such as button clicks and text inputs. This formulation supports behavioral cloning and enhances adaptation in novel GUI navigation tasks.  

In robotics applications, SayPlan~\cite{rana2023sayplan} leverages scene graphs and environmental feedback as short-term memory to guide planning and execution in scalable robotic environments. KARMA~\cite{wang2024karma} engages short-term working memory with an effective and adaptive memory replacement mechanism to dynamically record changes in objects' positions and states. LLM-Planner~\cite{song2023llm} iteratively updates short-term memory with environmental observation to prompt an LLM for dynamic planning.

\subsection{Long-Term Memory}
\label{subsec:long-term-mem}

\begin{figure}[!ht]
\centering
    \includegraphics[width=\textwidth]{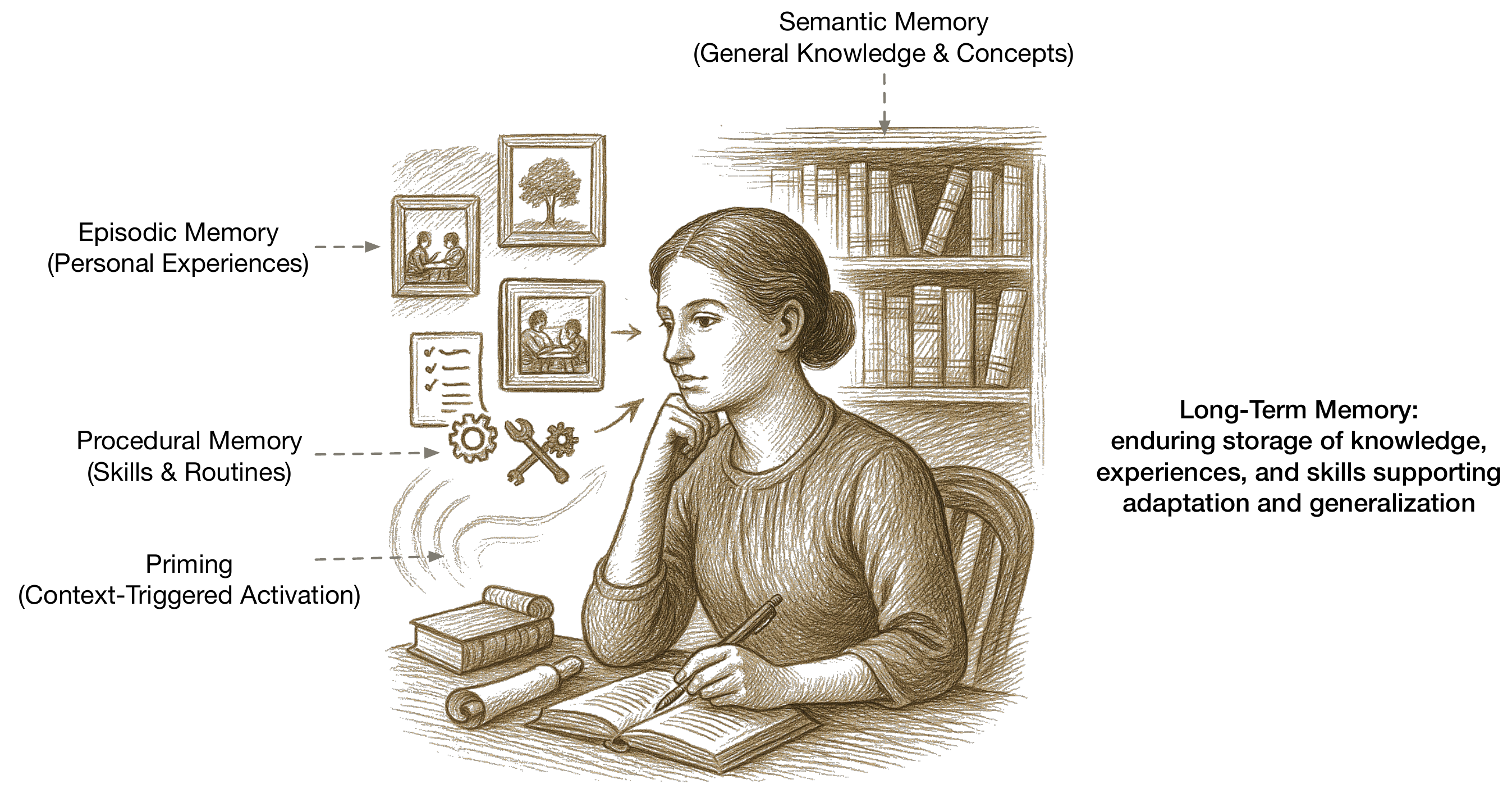}
    \caption{Illustration of long-term memory in cognition-inspired intelligent agents. The figure depicts a reflective agent drawing upon different types of long-term memory. In the background, bookshelves represent semantic memory, which refers to accumulated general knowledge and conceptual facts, while framed visual scenes signify episodic memory, consisting of past experiences and interactions. In the foreground, procedural memory is symbolized by tools and checklists used to encode skills and reusable plans. Subtle motion lines represent priming, which enables rapid, context-sensitive activation of relevant responses. Together, these memory systems support long-term retention, cross-task generalization, and adaptive behavior in intelligent agents.}
\label{fig:long-term-mem}
\end{figure}

Long-term memory in cognition-inspired intelligent agents enables the retention and retrieval of information over extended periods, allowing agents to generalize knowledge and adapt to new contexts effectively. Unlike sensory and short-term memory, which handle transient or immediate data, long-term memory supports cumulative learning and cross-task adaptability. It mirrors human long-term memory by incorporating explicit and implicit components, facilitating richer contextual understanding and intuitive behavior. Figure~\ref{fig:long-term-mem} illustrates the key components of long-term memory in cognition-inspired agents, including semantic, episodic, procedural, and priming-based mechanisms that collectively support adaptation, knowledge accumulation, and skill reuse.

On the one hand, \emph{explicit memory} involves intentional recollection, analogous to declarative memory in humans. It consists of \emph{semantic memory}, which stores general knowledge such as facts and concepts, and \emph{episodic memory}, which records specific events and interaction histories. Semantic memory in intelligent agents can be preloaded from domain knowledge bases or dynamically acquired through interactions. For example, in environments like TextWorld, semantic memory captures structured facts, such as ``\emph{Recipe} $-$ \emph{contains} $-$ \emph{Tuna}'' or ``\emph{Recipe} $-$ \emph{is on} $-$ \emph{Table}''. Episodic memory, in contrast, logs situational context and sequential actions, such as ``go from kitchen to living room, then to garden''. Integrating semantic and episodic memory allows agents to retain static and contextual information, enabling human-like adaptability and context-aware responses.  

On the other hand, \emph{implicit memory} shapes agent behavior through \emph{procedural memory} and \emph{priming}. Procedural memory enables agents to perform repetitive tasks efficiently by recalling specific skills and reusable plans. For example, it automates routine tasks without requiring explicit instructions, improving task execution efficiency. Priming, meanwhile, captures state changes and corresponding responses, allowing agents to adapt to similar contexts quickly. Priming enhances fluidity and context-sensitive decision-making by directly matching observations to or continuously chaining actions. 

Most intelligent agents implement both semantic and episodic memory within their memory modules. For instance, Agent S~\cite{agashe2024agent}, designed for GUI automation tasks, incorporates semantic memory to store online web knowledge in natural language form, while episodic memory captures high-level, step-by-step task experiences. Similarly, AriGraph~\cite{anokhin2024arigraph}, targeting embodied simulation tasks, encodes semantic environment knowledge using a fact graph and logs episodic navigation history through an event graph. In AI companion systems like MemoryBank~\cite{zhong2024memorybank} for SiliconFriend, semantic memory constructs user portraits in natural language, while episodic memory retains interaction histories, enhancing personalized and context-aware behavior. Recently, Mem$0$~\cite{chhikara2025mem0} introduced an open-source, production-ready memory architecture that dynamically extracts, consolidates, and retrieves salient conversational information for long-term consistency. Its graph-based extension, Mem$0^g$, further enables structured relational reasoning by modeling conversational elements as entity-relation graphs.

For implementing implicit memory, current agent systems primarily adopt model-friendly memory formats, such as key-value pair storage, executable code, or reusable routines. For example, AAG~\cite{roth2024pairing} defines and generalizes procedures through analogy, mapping knowledge from one situation (base) to another (target). This structure can be represented as a linear directed chain graph, where the input serves as the root, the output as the leaf node, and each intermediate step as a node in the chain. Similarly, Cradle~\cite{tan2024towards} and Jarvis-1~\cite{wang2024jarvis} implement procedural memory by storing and retrieving skills in code form, which can be either learned from scratch or pre-defined. Once curated, skills can be added, updated, or composed within memory. The most relevant skills for a given task and context are then retrieved to support action planning.

\section{The Memory Lifecycle} 
\label{sec:mem-lifecycle}

\lettrine[lines=3]{\initfamily\textcolor{darkgreen}{H}}{uman cognition} is underpinned by a dynamic memory system that processes information through stages: \emph{acquisition}, \emph{encoding}, \emph{consolidation}, \emph{retrieval}, and \emph{utilization}. These stages are interdependent, enabling humans to learn from experiences, adapt to new situations, and make informed decisions. Similarly, designing intelligent agents with a structured memory lifecycle allows for more robust, adaptable, and context-aware behavior. By mirroring the human memory process, we can create AI systems that not only process information efficiently but also learn and evolve over time.

\begin{figure}[!ht]
\centering
    \includegraphics[width=1.0\textwidth]{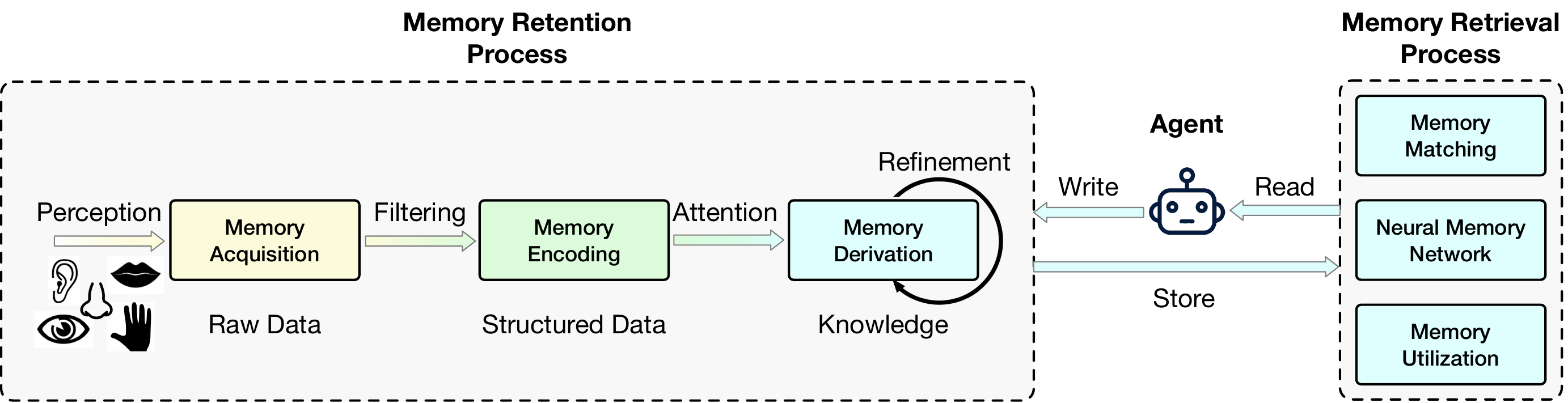}
    \caption{Illustration of the memory lifecycle. 
    The memory retention process involves three sequential steps—memory acquisition, encoding, and derivation, 
    while the memory retrieval process encompasses several independent applications, including matching (vector search), neural memory networks, and memory utilization (for long-context modeling and hallucination mitigation). }
\label{fig:memory-lifecycle}
\end{figure}

As illustrated in Figure~\ref{fig:memory-lifecycle}, the memory lifecycle in intelligent agents can be conceptualized as a dual process comprising \emph{memory retention} and \emph{memory retrieval}, each encompassing a sequence of interconnected stages that collectively govern how information is perceived, processed, stored, and applied. Retention includes acquisition, encoding, and derivation, responsible for capturing raw inputs, transforming them into structured internal representations, and refining them into actionable knowledge. Retrieval, in turn, involves memory matching, neural memory networks, and utilization, enabling agents to access relevant information, reason with past experiences, and guide behavior in real time. A clear understanding and implementation of this lifecycle is essential for developing intelligent agents with enhanced learning capabilities, contextual awareness, and decision-making proficiency.

To complement this conceptual perspective, we introduce a unified computational view that frames the memory lifecycle as a latent optimization problem.

\begin{definition}[\textbf{Memory Lifecycle Optimization}]
\label{def:memory-lifecycle}

At each time step $t$, the agent maintains a memory state $\mathcal{M}_t^{\text{mem}}$, which approximates an ideal latent memory target $\mathcal{M}_t^{\text{mem}^\ast}$. This latent memory encodes the minimal sufficient information necessary to support optimal future behavior, planning, and reasoning.

\begin{align}
\mathcal{M}_t^{\text{mem}} = \argmin_{\mathcal{M}} \, \mathcal{L}_{\text{mem}}(\mathcal{M}, \mathcal{M}_t^{\text{mem}^\ast}) + \lambda_{\text{store}} \cdot \mathcal{C}_{\text{store}}(\mathcal{M}) + \lambda_{\text{retr}} \cdot \mathcal{C}_{\text{retr}}(\mathcal{M})
\end{align}

\paragraph*{Divergence Term:}
$\mathcal{L}_{\text{mem}}(\cdot)$ quantifies the semantic deviation or utility gap between the constructed memory and the task-optimal latent memory.

\paragraph*{Storage Cost:}
$\mathcal{C}_{\text{store}}(\cdot)$ models the cost of storing memory, including memory size, update complexity, and representational redundancy. It can be further decomposed into stage-specific components, such as $\mathcal{C}_{\text{acq}}$ for acquisition and $\mathcal{C}_{\text{enc}}$ for encoding.

\paragraph*{Retrieval Cost:}
$\mathcal{C}_{\text{retr}}(\cdot)$ captures the computational overhead of retrieving and applying memory in real time.

\paragraph*{Fidelity–Efficiency Trade-off:}
The trade-off among these objectives is governed by the coefficients $\lambda_{\text{store}}$ and $\lambda_{\text{retr}}$, which regulate the balance between memory fidelity and computational efficiency.

\end{definition}

This formalism not only provides a high-level abstraction in the memory lifecycle, but also facilitates principled memory assessment, i.e., the divergence term $\mathcal{L}_{\text{mem}}$ serves as a general metric for evaluating how useful, sufficient, and task-aligned the current memory state is in approximating the latent optimal memory.

\subsection{Memory Acquisition}
\label{subsec:mem-acquisition}

\emph{Memory Acquisition} is the foundational process by which intelligent agents take in raw perceptual information from their environment. This initial step is crucial for subsequent learning, adaptation, and decision-making~\cite{lake2017building}. A primary challenge in acquisition is the sheer volume and complexity of environmental inputs. Agents are constantly bombarded with visual, auditory, textual, and other forms of data, much of which is redundant or irrelevant to the agent's goals. Therefore, a core aspect of memory acquisition is not simply capturing data, but also initiating a preliminary filtering process. This filtering is not merely about reducing quantity; it's an essential step towards constructing a simplified, yet representative, internal model of the external world. The agent needs to compress information not just for efficiency, but to extract salient features and patterns that ``fit'' or approximate the underlying structure of reality, discarding noise and retaining fidelity for key aspects. This approximation is a continuous process, beginning here with coarse-grained selection and compression. 

Formally, we consider this stage as an information selection and lossy compression process that maps raw sensory input $o_t$ into a compressed memory sketch $\tilde{o}_t$ stored as part of $\mathcal{M}_t^{\text{mem}}$. The goal is to preserve task-relevant information while minimizing unnecessary redundancy:
\begin{align}
\mathcal{M}_t^{\text{mem}} \leftarrow \texttt{Compress}(o_t) = \argmin_{\tilde{o}_t = f_{\text{acq}}(o_t)} \, \mathbb{E}_{o_t} \left[ D\left(\tilde{o}_t, o_t\right) \right] + \lambda_{\text{acq}} \cdot \mathcal{C}_{\text{acq}}(\tilde{o}_t)
\end{align}
$f_{\text{acq}}(\cdot)$ denotes a content-aware transformation pipeline during memory acquisition such as \emph{information compression} and \emph{experience consolidation}, which embed the raw observation into a more compact latent space. $D(\cdot, \cdot)$ measures semantic divergence between the compressed and original representations, while $\mathcal{C}_{\text{acq}}(\cdot)$ penalizes complexity (e.g., length and entropy). This formalism highlights acquisition not just as data ingestion, but as the agent's first act of judgment, deciding what to remember and what to forget.

On the one hand, information compression involves rudimentary techniques to reduce data dimensionality. This might include downsampling images, extracting key phrases from text using simple heuristics, or identifying significant changes in audio streams ~\cite{goyal2019infobot}. The goal is rapid, lossy compression to prioritize potentially relevant information. For example, LMAgent~\cite{liu2024lmagent} prompts the LLM to perform information compression, reducing irrelevant and unimportant content when constructing sensory memory to enhance operational efficiency. Meanwhile, ReadAgent~\cite{lee2024human} and GraphRead~\cite{li2024graphreader} respectively employ different strategies for compressing long text, i.e., episode pagination and graph-based structuring, to maximize information retention while ensuring efficiency.

On the other hand, experience consolidation, even at the acquisition phase, plays a role. The agent doesn't yet have a rich memory, but it can begin to apply previously learned, very general rules or biases. For example, if the agent has a pre-existing bias towards moving objects, it might prioritize visual data containing motion, even before full encoding ~\cite{mnih2015human}.
To enhance the dynamic consolidation of memory-based experiences, \citet{hou2024my} define metrics such as contextual relevance and recall frequency to determine whether to update long-term memory in a vector database. Expel~\cite{zhao2024expel} constructs an experience pool to collect and extract insights from training tasks, facilitating generalization to unseen tasks. More recently, MindOS~\cite{hu2025unified} proposed a working memory-centric central processing module for building autonomous intelligent agents, where working memory consolidates task-relevant experiences into structured thoughts for guiding future decisions and actions.


\newcolumntype{C}[1]{>{\centering\arraybackslash}m{#1}}
\begin{table*}[!t]
    \centering
    \small
    \setlength{\tabcolsep}{1.1pt}
    \resizebox{\textwidth}{!}{
        \begin{tabular}{
        C{0.15\textwidth}|
        C{0.09\textwidth}|
        C{0.09\textwidth}|
        C{0.09\textwidth}|
        C{0.1\textwidth}|
        C{0.11\textwidth}|
        C{0.1\textwidth}|
        C{0.09\textwidth}|
        C{0.09\textwidth}|
        C{0.09\textwidth}
        }
            \toprule
            \multirow{2}{*}{\textbf{Method}}& \multirow{2}{*}{\textbf{Domain}}& \multicolumn{3}{c|}{\textbf{Memory Representation}}& \multicolumn{5}{c}{\textbf{Memory Lifecycle}} \\
            & & \multicolumn{1}{c}{Sensory}& \multicolumn{1}{c}{Short-term}& \multicolumn{1}{c|}{Long-term}& \multicolumn{1}{c}{Acquisition}& \multicolumn{1}{c}{Encoding}& \multicolumn{1}{c}{Derivation}& \multicolumn{1}{c}{Retrieval}& \multicolumn{1}{c}{Utilization} \\
            \midrule
            Synapse \cite{zheng2023synapse}& GUI& Multi-modal& Context& Episodic, Procedural& User demo.& -& Hierarch. Decomp.& -& - \\
            \hline
            Agent S \cite{agashe2024agent}& GUI& Multi-modal& Context, Working& Semantic, Episodic&  Info. Compress.& Contrastive Learn.& Select. Forget.& Indexing& Long-context \\
            \hline
            Automanual \cite{chen2024automanual} & GUI & Multi-modal & Context & Procedural, Episodic & User Demo. & Hierarch. Parse & Goal Decomp. & Task Search & Subgoal Exec. \\            
            \hline
            AutoGuide \cite{fu2024autoguide}  & GUI & Multi-modal & Context & - & Screen Capture & - & Action Plan & - & Action Exec. \\
            \hline
            Agent-Pro \cite{zhang2024agent} & GUI  & Multi-modal & Context & - & Screen Capture & - & Hierarch. Decomp. & - & Action Exec. \\
            \hline
            MemGPT \cite{packer2023memgpt}& Document& Text& Context, Working& -& External Data& -& -& Paging, Func. call& Doc. interact. \\
            \hline
            SeeAct~\cite{zheng2024seeact} & Web & Multi-modal & Context & - & Screen Capture & - & Action Plan & - & Web Interact.\\
            \hline
            AutoWebGLM \cite{lai2024autowebglm} & Web & Text & Context & - & HTML Parse & HTML Embed & HTML Analysis & - & Web Interact. \\
            \hline
            SteP \cite{sodhi2023heap} & Web & Text & Context & Task-spec. & HTML Parse & HTML Embed & HTML Analysis & Element Rank & Web Interact. \\
            \hline
            AWM~\cite{wang2024agent}& Web& Text& -& Procedural& Workflow Extract.& Action Summ.& -& Sim. lookup& Workflow exec. \\
            \hline
            AriGraph \cite{anokhin2024arigraph}& TextWorld& Text& -& Semantic, Episodic& Env. Observ.& Knowl. Graph& Graph Traversal& Assoc. Retrieval& Action plan. \\
            \hline
            MemoryBank \cite{zhong2024memorybank}& Dialogue& Text& -& Episodic& Dialogue Record& -& -& Chron. order& Response \\
            \hline
            PromptAgent \cite{wang2023promptagent}  & General & Text & Context & - & Prompting & - & Prompt Refine. & Content-based & Prompt Exec. \\
            \hline
            ECL \cite{qian2023experiential} & Embody & Multi-modal & Context & Episodic & Obs. Recording & Contrast. Learn. & Exper. Summ. & Sim. \& Recency & Policy Learn. \\
            \hline
            LEO \cite{huang2023embodied} & Embody & Multi-modal & Working & Long-Horizon Rep. & Observation & Spatial-Temp. Learn. & Goal-Cond. Policy & Hierarch. Plan & Long-Horizon Exec. \\
            \hline
            IER \cite{qian2024iterative} & Embody & Multi-modal & Context & Episodic & Env. Interact. & Multi-modal Embed & Iter. Refine. & Sim. Match & Action Plan. \\
            \hline
            Voyager \cite{wang2023voyager}& Embody& Text& Working& Procedural& Auto. Curriculum& Skill Library& Iter. Prompt.& -& Skill Exec. \\
            \hline
            A3T \cite{yang2024react} & Embody, Robotics & Text& Context & - & Task Decomp. & Token. \& Embed. & Action Planning&-& Action select. \\
            \hline
            STARLING \cite{basavatia2024starling} & Robotics & Multi-modal & Context & Procedural & Demo. & Traj. Encode & Skill Refine. & Sim. \& Context & Skill Exec. \\        
            \bottomrule
        \end{tabular}
    }
    \caption{Summary of the memory module in various agents. Refer to Figure~\ref{fig:tree-memory} for abbreviations.}
    \label{tab:memory-analysis}
\end{table*}

\subsection{Memory Encoding}
\label{subsec:mem-encoding}

Memory encoding builds upon acquisition by transforming the filtered perceptual information into internal representations suitable for storage and later use.
Similar to the selective attention in human cognitive processes ~\cite{lindsay2020attention}, the inherent challenges of encoding stem from the complexity, high dimensionality, and often noisy nature of raw perceptual data. Effective encoding requires advanced mechanisms to identify key features, compress them compactly, and integrate information from multiple modalities. 

We frame encoding as the transformation of the preliminary sketch $\tilde{o}_t$ (produced during acquisition) into a structured latent representation $z_t$ that is stored in memory $\mathcal{M}_t^{\text{mem}}$. This transformation is defined by a learnable mapping $f_{\text{enc}}(\cdot)$ that refines and abstracts relevant features:
\begin{align}
\mathcal{M}_t^{\text{mem}} \leftarrow \texttt{Encode}(\tilde{o}_t) = \argmin_{z_t = f_{\text{enc}}(\tilde{o}_t)} \, \mathcal{L}_{\text{enc}}(z_t, \mathcal{G}_t) + \lambda_{\text{enc}} \cdot \mathcal{C}_{\text{enc}}(z_t)
\end{align}
Here, $\mathcal{L}_{\text{enc}}(z_t, \mathcal{G}_t)$ quantifies how well the encoded representation $z_t$ preserves task-relevant signals with respect to an explicit or implicit goal $\mathcal{G}_t$ (e.g., downstream goals, retrieved queries, or plan supervision). The regularizer $\mathcal{C}_{\text{enc}}(z_t)$ penalizes encoding complexity, such as redundancy, representational size, or misalignment across modalities. The encoding function $f_{\text{enc}}(\cdot)$ serves as a unified abstraction instantiated by concrete techniques such as \emph{selective attention} and \emph{multimodal fusion}, as detailed below.

\emph{Selective attention} mechanisms, inspired by human cognition, allow the agent to dynamically focus computational resources on the most relevant parts of the input. This might involve attending to specific regions of an image, keywords in a text, or particular frequencies in an audio signal. Different attention mechanisms can be used depending on the modality and task. For example, as the candidate memory dynamically expands, MS~\cite{gao2024memory} employs an LLM-based scorer to selectively retain the top-scoring half, creating a more compact shared memory across multiple agent systems. In other modalities, GraphVideoAgent~\cite{chu2025understanding} utilizes graph-based memory to enable selective and multi-turn video scene understanding, enhancing question-answering performance. In robot control, \citet{ali2024robots} implements selective attention as a filtering mechanism to extract task-relevant objects from the set of all perceived objects on the table.

\emph{Multi-modal fusion}~\cite{nagrani2021attention} is essential for integrating information from different sensory inputs (e.g., combining visual and auditory data to understand a scene). This involves creating a unified representation space where features from different modalities are aligned. Cross-modal encoders and contrastive learning techniques are often used to achieve this fusion. For example, JARVIS-1~\cite{wang2024jarvis} uses the general-domain video-language model CLIP~\cite{alec2021clip} to compute alignment within a multimodal key-value memory, where the key comprises elements such as task, plan, and visual observations, and the value is a text-based representation of successfully executed plans. Furthermore, Optimus-1~\cite{li2024optimus} refines memory representation and optimizes the multimodal encoder by leveraging MineCLIP~\cite{Linxi2022MineDojo}, a domain-specific video-language model pre-trained on Minecraft gameplay, to align and fuse filtered video streams with textual instructions and plans, encoding the agent's multimodal experiences into an abstracted memory pool. This integrated representation enhances information retrieval and reasoning across modalities and acts as another filter, reinforcing consistent data.

\subsection{Memory Derivation}
\label{subsec:mem-derivation}

Once information is encoded, human cognition often revisits and reorganizes it through reflection, inference, and consolidation. Similarly, in intelligent agents, memory derivation refines stored content into more structured, actionable knowledge. This stage mimics human metacognition: identifying core ideas, summarizing past events, extracting general rules, or pruning irrelevant associations.
Specifically, this process goes beyond simple storage and is essential for enhancing the agent's learning efficiency and generalization capabilities. Unlike encoding, memory derivation lacks explicit targets and is instead guided by the dynamic evaluation of information utility and structure.

We thus frame derivation as a self-organizing transformation from the current memory $\mathcal{M}_t^{\text{mem}}$ to a refined form $\mathcal{M}_t^{\text{mem}^{\prime}}$, by minimizing structural inefficiencies and redundancy through metacognitive mechanisms $f_{\text{derive}}(\cdot)$. These mechanisms include \emph{reflection}, \emph{summarization}, \emph{knowledge distillation}, and \emph{selective forgetting}:
\begin{align}
\mathcal{M}_t^{\text{mem}} \leftarrow \texttt{Derive}(\mathcal{M}_t^{\text{mem}}) 
= \argmin_{\mathcal{M}_t^{\text{mem}^{\prime}} = f_{\text{derive}}(\mathcal{M}_t^{\text{mem}})} 
\left( \lambda_1 \cdot \mathcal{C}_{\text{redundancy}} + \lambda_2 \cdot \mathcal{C}_{\text{staleness}} - \lambda_3 \cdot \mathcal{R}_{\text{structure}} \right)
\end{align}
Here, $\mathcal{C}_{\text{redundancy}}$ penalizes duplicated or overlapping entries; $\mathcal{C}_{\text{staleness}}$ accounts for temporal decay or infrequent usage; and $\mathcal{R}_{\text{structure}}$ rewards abstraction, sparsity, and compositional organization in memory.

\emph{Reflection} involves an agent actively analyzing its memories to identify patterns, relationships, and potential inconsistencies. It can be triggered by specific events (e.g., an unexpected outcome) or occur periodically as a background process. This process may include comparing memories, reasoning about causal relationships, and generating hypotheses~\cite{wang2023promptagent}. ExpeL~\cite{zhao2024expel} leverages reflection to collect past experiences for generalization to unseen tasks and to support trial-and-error reattempts following failures. R2D2~\cite{huang2025r2d2} models memory as a replay buffer and applies reflection to refine it by correcting failed execution trajectories in web agents. These corrected trajectories are then combined with successful ones to construct reflective memory, which serves as a reference for future decision-making.

\emph{Summarization} aims to produce concise representations of larger bodies of information while preserving their most essential content. This can include extracting key sentences from a document, generating abstractive summaries of conversations, or condensing sequences of events. Summarization techniques range from simple extractive methods to advanced abstractive approaches powered by large language models (LLMs)~\cite{laban2023summedits,cheng2025large,wang2023enhancing}. For example, ~\cite{wang2023recursively} introduces a recursive summarization strategy over dialogue history and prior memory to support long-term dialogue memory derivation. Building on this, Healthcare Copilot~\cite{ren2024healthcare} maintains concise memory by transforming conversation memory, representing the full ongoing medical consultation, into history memory that retains only key information relevant to the patient's medical history. MEM1~\cite{zhou2025mem1} further integrates memory summarization directly into the reasoning process, learning to dynamically condense context into a compact, constantly updated memory representation that supports efficient long-horizon decision-making.

\emph{Knowledge distillation}~\cite{hinton2015distilling} enables agents to transfer knowledge from larger, more complex models (or ensembles) to smaller, more efficient ones. This is particularly important for resource-constrained agents and for enhancing generalization. Distillation can also involve consolidating knowledge from multiple specialized models into a single, general-purpose model. For example, AoTD~\cite{shi2024unlocking} distills textual chains of thought from execution traces of subtasks into a Video-LLM to enhance multi-step reasoning performance in video question answering tasks. LDPD~\cite{liu2024language} transfers decision-making outcomes from teacher agents (i.e., expert buffers) to student agents, optimizing the student's policy to align with the teacher's. In multi-agent systems, MAGDi~\cite{chen2024magdi} distills the reasoning interactions among multiple LLMs into smaller models by structurally representing multi-round interactions as graphs, thereby improving the reasoning capabilities of smaller LLMs.

\emph{Selective forgetting}~\cite{richards2017brain} is the crucial process of removing or down-weighting memories that are deemed irrelevant, redundant, or outdated. This is essential for maintaining memory efficiency and preventing cognitive overload. Forgetting mechanisms can be based on time (older memories are more likely to be forgotten)~\cite{ren2024healthcare}, usage frequency (infrequently accessed memories are more likely forgotten)~\cite{liu2023think}, and relevance to the current task or context~\cite{gao2023s3}. In more fine-grained forgetting mechanisms, MemoryBank~\cite{zhong2024memorybank} applies the Ebbinghaus Forgetting Curve to quantify the forgetting rate, accounting for both time decay and the spacing effect, i.e., the principle that relearning information is easier than learning it for the first time. In contrast, Lyfe Agent~\cite{kaiya2023lyfe} adopts a hierarchical summarize-and-forget strategy: it first clusters related memories, refines them into concise summaries, and then removes older memories that are highly similar to newer ones. This approach enables efficient, low-cost memory updates for real-time social interactions.

\subsection{Memory Retrieval and Matching}
\label{subsec:mem-retrieval-matching}

In both humans and intelligent agents, memory is only valuable if it can be accessed precisely when needed. Memory retrieval serves as the mechanism for recalling relevant knowledge and past experiences to support reasoning and problem-solving. In humans, retrieval is often guided by contextual cues, emotional salience, or associative links. Similarly, intelligent agents must extract pertinent information from a large, heterogeneous memory pool—including sensory, short-term, and long-term memories—using techniques such as context-aware semantic matching, dynamic routing, and attention-based filtering. The objective is to efficiently and accurately retrieve memory fragments that can inform the agent's current decisions, planning processes, and actions.

However, achieving this goal presents several significant challenges. First, the agent's memory repository is often heterogeneous, comprising various forms of memory such as natural language descriptions, structured knowledge graphs, and state-action-reward sequences. These memories differ fundamentally in their data structures, representations, and levels of semantic granularity, posing a challenge for unified retrieval. Second, the retrieved memory fragments must be highly relevant to the current context, including the agent's state, task goals, and environmental observations. Simple keyword matching falls short of capturing the deeper semantic relationships required for meaningful retrieval. Developing a context-aware semantic matching mechanism that can dynamically adjust the retrieval strategy based on the current situation is therefore paramount. Third, the real-time nature of agent interaction with the environment necessitates efficient memory retrieval to support rapid decision-making and action~\cite{cao2024survey}. This demand for efficiency is further compounded by the limitations of the agent's computational resources. Finally, the agent's memory is not static but constantly evolving as new experiences, knowledge, and skills are acquired. Ensuring memories' timeliness, reliability, and relevance while avoiding the interference of outdated or erroneous information is a continuous challenge.

We formalize retrieval as a context-aware matching process, where relevant memory fragments $m_i \in \mathcal{M}_t^{\text{mem}}$ are selected based on their similarity to the current context $c_t$, as measured by a scoring function $f_{\text{match}}$:
\begin{align}
\texttt{Retrieve}(\mathcal{M}_t^{\text{mem}}, c_t) = \argmax_{m_i \in \mathcal{M}_t^{\text{mem}}}  f_{\text{match}}(m_i, c_t) - \lambda_{\text{retr}} \cdot \mathcal{C}_{\text{retr}}(m_i)
\end{align}
Here, $f_{\text{match}}(m_i, c_t)$ quantifies the semantic relevance between a memory fragment and the current context, while $\mathcal{C}_{\text{retr}}(m_i)$ penalizes retrieval cost, such as latency, size, or modality-specific constraints.

A comprehensive approach can address these challenges, encompassing four key components. Firstly, a foundational step involves constructing a unified \emph{memory representation and indexing} scheme. This aims to bridge the representational gap between different memory types by embedding them into a common vector space. Pre-trained language models like BERT or Sentence-BERT~\cite{reimers2019sentence} can be leveraged to transform text-based memories into semantic vectors, while graph neural networks (GNNs) can learn vector representations for structured memories like knowledge graphs, capturing both node and edge relationships~\cite{kipf2016semi}. To facilitate efficient retrieval, a multi-layered hybrid indexing structure is essential. This integrates techniques like inverted indexes for keyword matching, vector indexes like Faiss~\cite{johnson2019billion} or Annoy~\cite{li2019approximate} for similarity search, and graph indexes for structural queries~\cite{zhang2024high}, thus supporting diverse query needs.

Secondly, perhaps most critically, the system must develop \emph{context-aware semantic similarity} computation. This allows the retrieval process to understand and utilize the current context, such as the agent's state, goals, and observations, enabling a deeper semantic match beyond keyword overlap. This involves encoding the contextual information into vector representations and effectively fusing them with memory vectors. The attention mechanism plays a crucial role here, dynamically calculating the relevance between context and memory vectors and assigning different weights to memory fragments based on their contextual relevance~\cite{bahdanau2014neural}. This emphasizes memories that are more pertinent to the current situation.

Thirdly, integrating memory retrieval with the agent's task execution necessitates a \emph{task-oriented sequence decision and dynamic routing} mechanism. This leverages the structural information of tasks to guide memory retrieval and utilization, enabling complex task decomposition, planning, and dynamic adjustments. By constructing a task dependency graph, the agent can topologically sort subtasks to determine execution order. During execution, each subtask's goal serves as context for memory retrieval, extracting relevant knowledge and experience. Moreover, the agent must adapt to environmental feedback and task progress, dynamically adjusting the execution plan. Each decision point involves re-retrieving memories based on the current state and goal to select the optimal action and handle unexpected situations. This aspect also emphasizes how agents can leverage their skill memory to solve problems, including skill distillation, combination, and innovation. Pattern recognition allows for summarising general problem-solving steps, while structured knowledge organization arranges skills into a retrievable format. Agents can further distill generalized skills from specific ones, combine multiple skills to address complex challenges, and even innovate new skill combinations. These processes depend fundamentally on an efficient memory retrieval system that can identify appropriate skills or skill combinations based on task requirements.

Finally, a robust \emph{memory management} mechanism is crucial for maintaining the memory pool's timeliness, relevance, and efficiency. This mechanism should incorporate a forgetting and updating strategy, mirroring human forgetting mechanisms~\cite{wang2024comprehensive}. This might involve regularly purging outdated, redundant, or infrequently used memories based on time-based decay (weakening memory strength over time) and frequency-based decay (purging low-frequency memories). Simultaneously, when a memory fragment relevant to the current task is retrieved, its timestamp and access frequency are updated, increasing its importance and ensuring dynamic memory updates. Through these concerted efforts, LLM Agents can be equipped with a powerful, flexible, and context-aware memory retrieval and matching system, enabling them to effectively utilize their accumulated knowledge, support complex decision-making, and exhibit more intelligent behavior.

\subsection{Neural Memory Networks}
\label{subsec:neural-mem}

\begin{figure}[!ht]
\centering
    \includegraphics[width=0.8\textwidth]{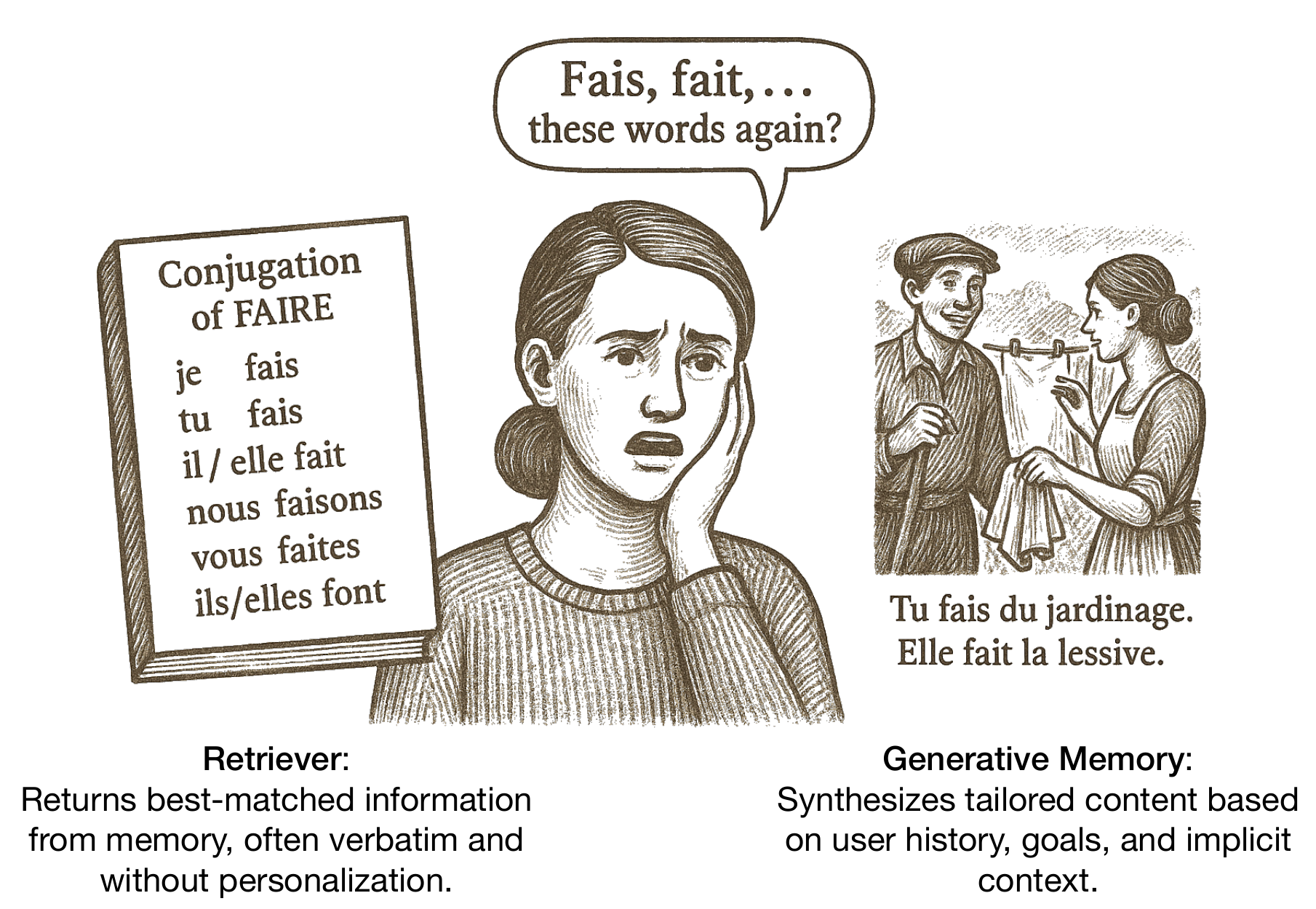}
    \caption{Illustration of the contrast between a retrieval-based memory and a generative neural memory. When a learner struggles with subtle verb conjugations, the retriever returns a standard reference from a grammar textbook. In contrast, the generative memory synthesizes a personalized explanation, grounded in the user's learning history and context, to enhance understanding and engagement.}
\label{fig:generative-vs-retrieve-mem}
\end{figure}

Beyond explicit external memory systems—such as vector databases, document indexes, and key-value caches—Neural Memory Networks represent a form of \emph{implicit memory}, where an intelligent agent's experiences are stored directly within the model's parameters. These networks integrate memory seamlessly into the neural architecture by encoding information into the model's weights or activations, effectively transforming the model itself into a dynamic read–write memory medium. This tight coupling provides several advantages: memory access is embedded directly into the decoding process, maintaining constant retrieval cost even as the memory pool grows; and the need for explicit memory management is reduced, simplifying system design.

However, this approach also reveals a fundamental contrast between generative and retrieval-based memory paradigms. As illustrated in Figure~\ref{fig:generative-vs-retrieve-mem}, a retrieval-based system searches for the best-matching item based solely on the input query—akin to a student receiving a textbook excerpt when confused. In contrast, a generative memory system, such as a neural memory network, draws on the user's past interactions and contextual understanding to produce a more adaptive, personalized response—like a helpful tutor who knows the learner's strengths, struggles, and goals. The retrieved knowledge is ``looked up'', while generative memory reflects ``lived'' experience embedded in parameters.

Storing information within parameters may enable richer, more flexible associations and generalization, but comes at the cost of reduced transparency and controllability. Unlike external memory, parametric memory resists precise, surgical updates and risks catastrophic forgetting—where new learning overwrites existing knowledge. Additionally, its capacity is ultimately bounded by model size, whereas external stores can scale freely. The trade-off is thus between the fluid, efficient, and integrated nature of parametric memory and the scalable, interpretable, but more rigid character of external retrieval. Designing neural memory networks that resolve this tension remains one of the grand challenges of cognition-inspired AI.

A primary concern is balancing memory capacity with stability. Encoding a vast amount of information within the finite parameters of a neural network while maintaining long-term stability poses a major hurdle. The network must be able to store a multitude of memories without succumbing to catastrophic forgetting or confusion between similar memories. Equally crucial is the development of effective mechanisms for memory read-write operations. The network needs to reliably write new information, update existing memories, and accurately retrieve stored information on demand, all while maintaining computational efficiency. Beyond simply storing memories, the ultimate goal is to endow neural networks with the ability to generalize from and reason with the information they store. This would empower them to perform higher-order cognitive functions beyond rote memorization, allowing for insightful connections and inferences based on past experiences. 

We formalize this as a long-horizon continual learning process that balances the integration of new information with the preservation of existing knowledge, where the model parameters $\theta_t$ at time $t$ serve as the memory container. The learning objective jointly optimizes task performance while mitigating interference and memory degradation:
\begin{align}
\mathcal{M}_t^{\text{mem}} \leftarrow \theta_t = \argmin_{\theta} \, \mathcal{L}_{\text{task}}(\theta; \mathcal{T}_t) + \lambda_{\text{stability}} \cdot \mathcal{D}(\theta \| \theta_{t-1}) + \lambda_{\text{store}} \cdot \mathcal{C}_{\text{store}}(\theta)
\end{align}
Here, $\mathcal{L}_{\text{task}}$ promotes performance on the current task $\mathcal{T}_t$, $\mathcal{D}(\theta \| \theta_{t-1})$ penalizes deviation from prior parameters to preserve previously acquired knowledge and avoid catastrophic forgetting, and $\mathcal{C}_{\text{store}}(\theta)$ models the cost of incorporating new information into the network, such as computational overhead or optimization difficulty to training neural memory networks. This formulation captures the central trade-off in neural memory: the tension between \emph{stability and reliability} (retaining past knowledge) versus \emph{plasticity and generalization} (adapting to new experiences over time).
Two broad families of methods instantiate this framework: \emph{associative memory} and \emph{parameter integration}.

On the one hand, associative memory, inspired by the interconnectedness of neurons in the brain, offers a promising avenue. Models like Hopfield networks~\cite{demircigil2017model,ramsauer2020hopfield}, leveraging energy functions, and Bidirectional Associative Memories (BAMs)~\cite{kosko1988bidirectional}, supporting hetero-associative recall, provide mechanisms for encoding and retrieving patterns based on the weights between neurons. 
Besides, Neural Turing Machines (NTMs)~\cite{falcon2022neural} and Memory-Augmented Neural Network (MANNs)~\cite{santoro2016meta,dai2019transformer,bulatov2022recurrent, wang2024memoryllm} augment neural networks with external memory modules, employing attention and summary mechanisms to interact with these memories. 

On the other hand, parameter integration facilitates the seamless integration of world knowledge and accumulated experience into the operational behavior of intelligent agents.
For example, some prior works modify model parameters to enable continual learning by updating~\cite{huang2023transformer,gangadhar2024model,zhang2023continual} or forgetting specific knowledge~\cite{wang2024large}. Other studies treat LLMs as standalone memory modules, incorporating world knowledge into their parameters during pre-training~\cite{chen2023lifelong}, post-training~\cite{chen2024melodi}, and online deployment~\cite{tack2024online}. For instance, MemoryLLM~\cite{wang2024memoryllm} introduces memory tokens, while SELF-PARAM~\cite{wang2024self} leverages knowledge distillation to embed world knowledge and agent's past experiences into model parameters. This approach is further augmented in the M+ model~\cite{wang2025m+} with a long-term memory mechanism and a co-trained retriever, enhancing its ability to generalize to longer history memorization. Additionally, ~\cite{padmanabhan2023propagating} employs encoded memory to facilitate further reasoning, thereby improving the generalization of stored knowledge. More recently, MemoRAG~\cite{qian2024memorag} and R$^3$Mem~\cite{wang2025r} have been proposed to not only encode memory but also enable reliable retrieval from neural memory networks, unifying the dual processes of memory storage and retrieval within a single model. This advancement contributes to the development of next-generation generative-based retrieval systems, which support lifelong AI applications. Furthermore, Titans~\cite{behrouz2024titans} and Dynamic Cheatsheet~\cite{suzgun2025dynamic} introduce mechanisms to memorize test-time data points via test-time learning, thereby enhancing the agent's ability to generalize across tasks more efficiently at inference time.

Future research will continue to focus on developing neural memory models with larger capacity and greater stability. At the same time, designing more efficient and flexible memory read-write mechanisms remains a critical challenge. As a step in this direction, MemOS~\cite{li2025memos} introduces a prototype operating system for memory-augmented generation (MAG), supporting a unified, memory-centric training paradigm that integrates activation-based neural memory, parametric memory, and plaintext memory. It further proposes a three-layer architecture, comprising interface, operation, and infrastructure layers, to orchestrate memory control and enable seamless memory system integration. Looking ahead, a key research frontier will be applying these memory-augmented networks to complex cognitive tasks, thereby pushing the boundaries of what AI systems can achieve. Advancements in this area will unlock new possibilities for building intelligent agents capable of learning, remembering, and reasoning in ways increasingly aligned with human cognition.

\subsection{Memory Utilization}
\label{subsec:mem-utilization}

Finally, memory utilization refers to applying recalled information to enhance current tasks—reasoning, planning, action, or dialogue. It completes the cycle: from sensory input to structured knowledge to behavioral output. Achieving this, however, presents several challenges.

One primary challenge is balancing the vastness of the memory store with its effective utilization. Agents must navigate a potential information overload, ensuring that relevant memories are fully leveraged without overwhelming the system. Another hurdle is the need for abstraction and generalization. Agents need to distill specific memory segments into more general knowledge and apply this knowledge to new and varied situations. Furthermore, the issue of hallucinations and incorrect memories within the LLM requires careful consideration. Preventing the generation of content that contradicts or misrepresents stored information is crucial, as is the ability to identify and rectify erroneous information that may reside within the memory store itself.

To address these challenges, several strategies are employed. \emph{Retrieval-augmented generation (RAG)}~\cite{lewis2020retrieval} combines retrieval and generation models to enhance the LLM's capabilities by drawing upon external knowledge sources. Unlike the methods mentioned in memory retrieval and matching, RAG focuses on integrating retrieved information into the generation process itself. When prompted, the agent retrieves relevant memory segments and incorporates them into the context provided by the generation model. This contextual enrichment guides the model towards more factual and informative outputs. For instance, when responding to a user's query, the agent can first retrieve related entries from its knowledge base and then generate an answer based on this retrieved information, thus grounding the response in established knowledge.
More recently, some studies have integrated memory modules with RAG, incorporating self-reflection~\cite{yuan2025personalized} and adaptive retrieval mechanisms~\cite{mallen2023not} to enhance both generation reliability and efficiency. For example, Atlas~\cite{farahani2024deciphering} leverages causal mediation analysis, while ~\cite{ding2024retrieve} employs consistency-based hallucination detection to determine whether the model already possesses the necessary knowledge—allowing for direct generation—or whether retrieval is required, in which case the model first retrieves relevant information before generating a response. In a unified framework, RAGLAB~\cite{zhang2024raglab} offers a comprehensive ecosystem for evaluating and analyzing mainstream RAG algorithms. HippoRAG~\cite{gutierrez2024hipporag} employs a strategy inspired by the hippocampal indexing theory of human memory to create a KG-based index for memory and use Personalized PageRank for memory retrieval.

Furthermore, \emph{long-context modeling} plays a vital role in managing extensive memory stores. This approach enhances the LLM's ability to process long sequences and large-scale memories, allowing for a deeper understanding and utilization of long-range dependencies. By employing Transformer model variants like Transformer-XL~\cite{dai2019transformer} and Longformer~\cite{beltagy2020longformer}, or through hierarchical and recursive processing techniques, such as recurrent memory transformer (RMT)~\cite{bulatov2022recurrent,bulatov2023scaling}, agents can expand their context window. This enables them to handle significantly more extensive memory stores and reason and make decisions within a much broader context. For example, agents can maintain a longer memory span when processing extensive documents or engaging in prolonged conversations.
Additionally, some studies leverage memory to compress long contexts, enabling more effective long-context modeling. For example, AutoCompressor~\cite{chevalier2023adapting} introduces summary vectors as memory to transfer information from previous context windows into the current window, facilitating long-context understanding. Similarly, the in-context autoencoder (ICAE)~\cite{ge2023context} generates memory slots that accurately and comprehensively represent the original context, while LLMLingua~\cite{jiang2023llmlingua,jiang2024longllmlingua}, Gist~\cite{mu2024learning}, CompAct~\cite{yoon2024compact}, and HyCo$_2$~\cite{liao2025beyond} further optimize long-prompt compression to reduce input context length.

Finally, \emph{hallucination mitigation} strategies are essential for ensuring the reliability of generated outputs. These strategies aim to minimize the LLM's tendency to produce factually incorrect or nonsensical content. One approach is implementing fact-checking mechanisms~\cite{petroni2019language}, verifying generated content against established knowledge or memory stores. Another involves uncertainty estimation~\cite{kadavath2022language, farquhar2024detecting}, where the model evaluates the confidence level of its generated content and flags or filters out low-confidence outputs. Additionally, knowledge-based decoding strategies can be employed during the generation phase, introducing constraints that guide the model towards more factually accurate content. These techniques collectively contribute to generating more trustworthy outputs and are aligned with the agent's established knowledge base.
Recent research has introduced expert memory subnetworks, such as PEER~\cite{he2024mixture} and Lamini Memory Tuning~\cite{li2024banishing}, which specialize in memorizing specific types of information, including world knowledge and agents' past experiences. These subnetworks offload memorization to dedicated parameters, reducing the main model's propensity to hallucinate.
By implementing these memory utilization strategies, agents can become more capable, accurate, and reliable. They can successfully leverage their memory stores to achieve superior performance across complex tasks.

\section{Summary and Discussion}
\label{sec:ch-mem-summary}

\lettrine[lines=3]{\initfamily\textcolor{darkgreen}{T}}{he} development of truly intelligent agents depends not just on robust memory systems, but also on their seamless integration with other cognitive functions like perception, planning, reasoning, and action selection. Memory is not an isolated module; it is deeply intertwined with these other processes. For example, sensory input is encoded and filtered before storage (as discussed in the sections on memory representation and lifecycle), highlighting the interplay between perception and memory. Long-term memory, especially procedural memory, directly informs action selection through learned skills and routines. Retrieval mechanisms, like context-aware semantic similarity computation, are crucial for planning, allowing agents to access relevant past experiences. This interplay extends to the concept of a ``world model''.

Central to intelligent agents is their ability to build and utilize internal world models. These models, representing an agent's understanding of its environment, enable simulation, reasoning about consequences, and prediction. Robust world models are crucial for higher-level cognition, planning, and human-like intelligence. A world model is, in essence, a highly structured, often predictive, form of long-term memory. Memory provides the raw material, such as knowledge and experiences, for constructing the world model, while the world model, in turn, acts as an organizing framework, influencing how new memories are encoded, consolidated, and retrieved. For instance, a well-developed world model might prioritize storing surprising events, as these indicate gaps in the agent's understanding.

However, developing effective world models and memory systems presents significant challenges. These include managing the complexity of real-world environments, determining the appropriate level of abstraction (balancing accuracy, complexity, and computational efficiency), and integrating multi-modal information. Learning and updating these models efficiently, avoiding bias, ensuring generalization, and enabling continuous adaptation are also critical. Furthermore, model-based planning requires efficient search algorithms to handle the inherent uncertainty in the model's predictions.

Future research should focus on enhancing agent memory systems by drawing inspiration from the strengths of human memory, particularly its flexibility, adaptability, and efficiency. While agent memory has advanced considerably, it still lags behind human memory in these key areas. Human memory is remarkably associative, retrieving information from incomplete or noisy cues, and it exhibits a sophisticated form of ``forgetting'' that involves consolidation and abstraction, prioritizing relevant information and generalizing from experiences. Agent memory, conversely, often relies on precise matching and struggles with ambiguity.

Several promising research directions emerge. Exploring biologically-inspired mechanisms, such as neural memory networks (as discussed earlier), could lead to significant breakthroughs. A key tension here remains the choice between storing knowledge implicitly within model parameters versus explicitly in external caches. Parametric memory offers tight integration and potentially richer associative capabilities, but faces challenges in scalability, editability, and catastrophic forgetting. External caches offer vast, manageable storage but may lack the nuanced integration of parametric approaches. Future systems might need hybrid architectures that leverage the strengths of both. Another crucial area is developing memory systems that actively ``curate'' their contents, i.e., reflecting on information, identifying inconsistencies, and synthesizing new knowledge. This requires integrating metacognitive capabilities (monitoring and controlling one's own cognitive processes) into agent architectures. Furthermore, creating more robust and nuanced forms of episodic memory, capturing not just the ``what'' and ``when'' but also the ``why'' and the emotional context of events, is essential for agents that can truly learn from experience and interact with humans naturally.

Overcoming these challenges requires innovative solutions at the intersection of deep learning, reinforcement learning, and cognitive science. Developing more sophisticated and adaptable world models and memory systems that mirror the strengths of human cognition will pave the way for agents with a deeper understanding of their environment, leading to more intelligent and meaningful interactions.

\chapter{World Model} 
\label{chapter:world_model}
\lettrine[lines=3]{\initfamily\textcolor{darkgreen}{W}}{orld model} enables an agent to predict and reason about future states without direct trial-and-error in reality. This section explores how human cognitive studies on ``mental models'' relate to AI world models in artificial intelligence, categorizing them under four paradigms: \emph{implicit paradigm}, \emph{explicit paradigm}, \emph{simulator-based paradigm}, and a class of other emergent methods (e.g., \emph{instruction-driven paradigm}). We then discuss how world models inherently intersect with other agentic components and conclude with open questions and future directions that unite these perspectives under a unified theoretical and practical framework.

\begin{figure}[!htb]
\centering
    \includegraphics[width=0.8\columnwidth]{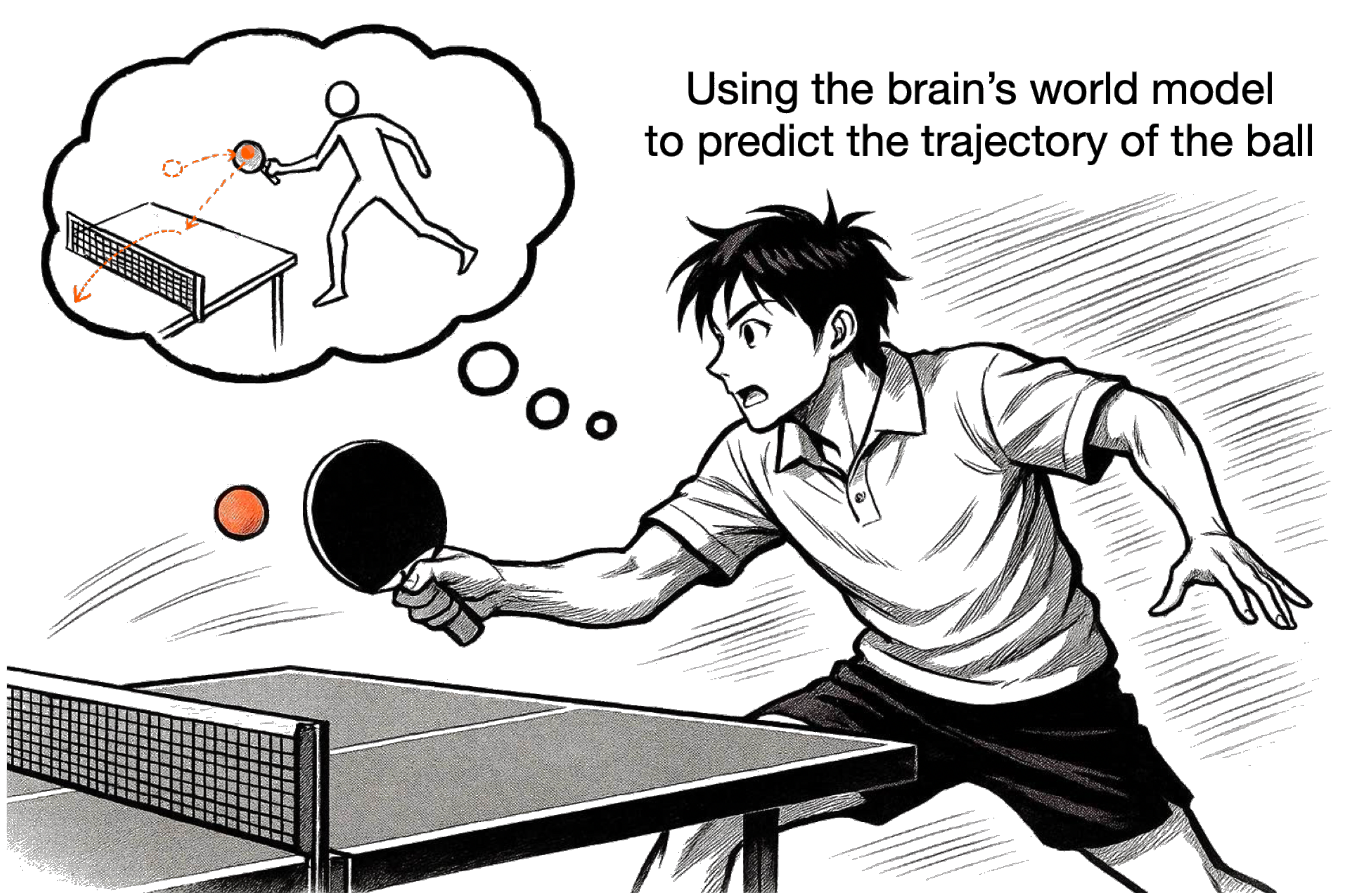}
    \caption{Humans can use their brain's model of the world to predict the consequences of their actions. For example, when playing table tennis, a player can imagine or predict the trajectory of the ball after an action.}

\label{fig:world-model}
\end{figure}

\section{Mental Models as Human World Representations}
\label{sec:human-world-model}

\lettrine[lines=3]{\initfamily\textcolor{darkgreen}{H}}{umans} naturally construct internal representations of the world, often referred to as \emph{mental models} in psychology \cite{tolman1948cognitive, craik1943nature, gentner2014mental}. These models serve as compact and manipulable depictions of external reality, enabling individuals to predict outcomes, plan actions, and interpret novel scenarios with minimal reliance on direct trial-and-error. Early work on spatial navigation, for instance, showed that humans and animals form ``cognitive maps'' of their surroundings \cite{tolman1948cognitive}, suggesting an underlying ability to imagine potential paths before actually traversing them.

Craik's seminal argument was that the human mind runs internal ``small-scale models of reality'' \cite{craik1943nature} to simulate how events might unfold and evaluate possible courses of action. This notion, often referred to as a \emph{mental model}, closely parallels what is now called a \emph{world model} in AI contexts, wherein an internal mechanism supports simulation and prediction without direct environmental interaction.
Later studies proposed that such simulations span modalities such as vision, language, and motor control, and are dynamically updated by comparing predictions to new observations. This process merges \emph{memory recall} with \emph{forward projection}, implying a close interplay between stored knowledge and the active generation of hypothetical future states \cite{gentner2014mental}. More recent predictive processing theories such as ``Surfing Uncertainty'' \cite{clark2015surfing} propose that the brain operates as a hierarchical prediction machine, continuously generating top-down predictions about sensory inputs and updating its models based on prediction errors.

Critically, these human mental models are:
\begin{itemize}
    \item \textbf{Predictive:} They forecast changes in the environment, informing decisions about where to move or how to respond.
    \item \textbf{Integrative:} They combine sensory input, past experience, and abstract reasoning into a unified perspective on ``what might happen next''.
    \item \textbf{Adaptive:} They are revised when reality diverges from expectation, reducing the gap between imagined and actual outcomes over time.
    \item \textbf{Multi-scale:} They operate seamlessly across different temporal and spatial scales, simultaneously processing immediate physical dynamics (milliseconds), medium-term action sequences (seconds to minutes), and long-term plans (hours to years). This flexibility enables reasoning across both fine-grained and coarse-grained temporal or spatial contexts, depending on task demands.
\end{itemize}

Consider hunger and eating as an illustration of integrated world modeling. When hungry, a person's internal model activates predictions about food, simulating not just visual appearance but tastes, smells, and anticipated satisfaction, triggering physiological responses like salivation before food is even present. This demonstrates seamless integration across perception, memory, and action planning.

The example also highlights adaptivity: once satiated, the same model dynamically updates, reducing predicted reward values for further eating. Despite recognizing the same food items, their anticipated utility changes based on internal state. Furthermore, humans are capable of counterfactual simulation, such as declining dessert now while predicting future enjoyment, which enables planning across hypothetical scenarios and time horizons.

In sum, the \emph{human world model} is not a static library of facts, but a flexible and ever-evolving mental construct, deeply rooted in perception and memory, that continuously shapes (and is shaped by) the individual's interactions with the outside world.

\section{From Human Mental Models to AI World Models}
\label{sec:human-to-ai-wm}

\lettrine[lines=3]{\initfamily\textcolor{darkgreen}{R}}{esearch} in artificial intelligence has long sought to replicate the \emph{predictive, integrative, and adaptive} qualities exhibited by human mental models \cite{tolman1948cognitive, craik1943nature}. Early reinforcement learning frameworks, for instance, proposed learning an \emph{environment model} for planning, exemplified by Dyna \cite{sutton1991dyna}, while contemporaneous work investigated using neural networks to anticipate future observations in streaming data \cite{schmidhuber1990making,schmidhuber1991possibility}. Both directions were motivated by the idea that an internal simulator of the world could enable more efficient decision-making than purely reactive, trial-and-error learning.

Subsequent advancements in deep learning brought the notion of ``AI world models'' into sharper focus. One influential approach introduced an end-to-end \emph{latent generative model} of an environment (e.g., ``World Models'' \cite{ha2018world}), whereby a recurrent neural network (RNN) and variational auto-encoder (VAE) together learn to ``dream'' future trajectories. These latent rollouts allow an agent to train or refine policies offline, effectively mirroring how humans mentally rehearse actions before executing them. Alongside such implicit designs, explicit forward-modeling methods emerged in model-based RL, letting agents predict $P(s'\mid s, a)$ and plan with approximate lookahead \cite{schrittwieser2020mastering,hafner2019dream}.

Another branch of work leveraged large-scale simulators or real-world robotics to ground learning in richly diverse experiences \cite{xiang2020sapien, wu2023daydreamer}. Such setups are reminiscent of how human children learn by actively exploring their environments, gradually honing their internal representations. Yet a key question lingers: can agentic systems unify these approaches (implicit generative modeling, explicit factorization, and simulator-driven exploration) into a cohesive ``mental model'' akin to that observed in humans? The recent proliferation of language-model-based reasoning \cite{gu2024your, hao2023reasoning} hints at the potential to cross modalities and tasks, echoing how humans integrate linguistic, visual, and motor knowledge under one predictive framework.

Overall, as AI systems strive for flexible, sample-efficient learning, the \emph{AI world model} stands as a conceptual bridge from cognitive theories of mental models to implementations that equip artificial agents with \emph{imagination}, \emph{predictive reasoning}, and \emph{robust adaptation} in complex domains.

\section{Paradigms of AI World Models}
\label{sec:paradigms}

\lettrine[lines=3]{\initfamily\textcolor{darkgreen}{D}}{esigning} an \emph{AI world model} involves determining how an AI agent acquires, represents, and updates its understanding of the environment's dynamics. While implementations vary, most approaches fall into four broad paradigms: \emph{implicit}, \emph{explicit}, \emph{simulator-based}, and \emph{hybrid or instruction-driven} models.
To clarify our classification, we define the distinction between \emph{implicit} and \emph{explicit} paradigms based on whether a system performs \emph{end-to-end reconstruction of observations} during planning or simulation. Systems that learn to generate or predict future high-dimensional observations, such as video frames or sensor readings, as part of their rollout process are considered \emph{explicit}. In contrast, models that operate entirely in latent or abstract internal spaces, even if they contain auxiliary decoders for training or visualization, are classified as \emph{implicit} under this definition. This view aligns with how reconstruction capacity contributes to interpretability and groundedness in world modeling.

These paradigms can be further analyzed along two key dimensions: reliance on \emph{internal} (neural-based) vs. \emph{external} (rule-based or structured) mechanisms, and overall \emph{system complexity}. 
For example, World Models\cite{ha2018world} relies primarily on internal neural networks for dynamics learning, while SAPIEN\cite{xiang2020sapien} uses external physics engines with predefined rules. Similarly, simple methods like ActRe\cite{yang2024react} uses minimal components, whereas DayDreamer\cite{wu2023daydreamer} integrates multiple neural modules for state prediction, value estimation, and policy learning. 
Figure~\ref{fig:wm-twodim} illustrates this two-dimensional space, showing how different approaches distribute themselves across these axes.

\begin{figure}[ht]
\centering
\begin{tikzpicture}[
    font=\small,
    axis/.style={thick,->,>=stealth},
    methodpoint/.style={draw=black, circle, fill=black, inner sep=1.2pt},
    methodlabel/.style={anchor=west, align=left}
]

\def\xmin{0}
\def\xmax{8}
\def\ymin{0}
\def\ymax{8}

\draw[axis] (\xmin,4) -- (\xmax,4) node[right]{\textbf{Complexity}};

\draw[axis] (4,\ymin) -- (4,\ymax) node[above]{\textbf{Form}};

\node at (1.5,4.2) {Simple};
\node at (6.5,4.2) {Complex};
\node[rotate=90] at (4.2,1.5) {Internal};
\node[rotate=90] at (4.2,6.5) {External};

\draw[dashed] (4,\ymin) -- (4,\ymax);
\draw[dashed] (\xmin,4) -- (\xmax,4);

\coordinate (ActRe) at (1,1); 
\node[methodpoint] at (ActRe) {};
\node[methodlabel, anchor=south] at (ActRe) {ActRe \cite{yang2024react}};

\coordinate (World Models) at (2,2);
\node[methodpoint] at (World Models) {};
\node[methodlabel, anchor=south] at (World Models) {World Models \cite{ha2018world}};

\coordinate (Dreamer) at (2.8,2.7);
\node[methodpoint] at (Dreamer) {};
\node[methodlabel, anchor=south] at (Dreamer) {Dreamer \cite{hafner2019dream}};

\coordinate (DiffusionWM) at (6,2);
\node[methodpoint] at (DiffusionWM) {};
\node[methodlabel, anchor=south] at (DiffusionWM) {Diffusion WM \cite{alonso2024diffusion}};

\coordinate (GQN) at (5.8,2.8);
\node[methodpoint] at (GQN) {};
\node[methodlabel, anchor=south west] at (GQN) {GQN \cite{eslami2018neural}};

\coordinate (Daydreamer) at (5,3.2);
\node[methodpoint] at (Daydreamer) {};
\node[methodlabel, anchor=south west] at (Daydreamer) {Daydreamer \cite{wu2023daydreamer}};

\coordinate (PILCO) at (2.7,3.2);
\node[methodpoint] at (PILCO) {};
\node[methodlabel, anchor=south] at (PILCO) {PILCO \cite{deisenroth2011pilco}};

\coordinate (AutoManual) at (2,6);
\node[methodpoint] at (AutoManual) {};
\node[methodlabel, anchor=south] at (AutoManual) {AutoManual \cite{chen2024automanual}};

\coordinate (COAT) at (3,6.7);
\node[methodpoint] at (COAT) {};
\node[methodlabel, anchor=south east] at (COAT) {COAT \cite{liu2024discovery}};

\coordinate (SAPIEN) at (5,7);
\node[methodpoint] at (SAPIEN) {};
\node[methodlabel, anchor=south west] at (SAPIEN) {SAPIEN \cite{xiang2020sapien}};

\coordinate (MuZero) at (6,6);
\node[methodpoint] at (MuZero) {};
\node[methodlabel, anchor=south] at (MuZero) {MuZero \cite{schrittwieser2020mastering}};

\coordinate (GR2) at (5.2,5);
\node[methodpoint] at (GR2) {};
\node[methodlabel, anchor=south] at (GR2) {GR-2 \cite{cheang2024gr}};

\coordinate (DinoWM) at (7.5,7);
\node[methodpoint] at (DinoWM) {};
\node[methodlabel, anchor=south west] at (DinoWM) {DINO-WM \cite{zhou2024dino}};

\end{tikzpicture}
\caption{A two-dimensional layout of AI world-model methods. The horizontal axis indicates \emph{Complexity} (left to right). The vertical axis spans \emph{Internal} approaches (bottom) to \emph{External} solutions (top). Approximate positions reflect each method's reliance on large learned networks vs.\ explicit rules or code, and its overall system complexity.}
\label{fig:wm-twodim}
\end{figure}

\begin{table}[!t]
\small
\centering
\caption{Design matrix of representative world-model methods. Columns list the encoder/decoder choice (Enc/Dec), latent representation (Latent), rollout strategy (Rollout), planning method (Plan), and training objective (Obj).}
\label{tab:wm_dimensions}
\resizebox{\textwidth}{!}{
\begin{tabular}{@{}l c c c c c c@{}}
\toprule
\textbf{Method} & \textbf{Enc} & \textbf{Dec} & \textbf{Latent} & \textbf{Rollout} & \textbf{Planning} & \textbf{Obj} \\
\midrule
\rowcolor{LightMint}
World Models~\cite{ha2018world} & VAE & VAE & $z{+}h$ & MDNRNN & CMAES & Recon+KL \\
MuZero~\cite{schrittwieser2020mastering} & RepNet & None & State & LatentTree & MCTS & TD+CE \\
\rowcolor{LightMint}
Dreamer~\cite{hafner2019dream} & RSSM & Image/Reward & RSSM & Imagination & ActorCritic & ELBO+RL \\
DINO\text{-}WM~\cite{zhou2024dino} & DINOv2 & (Viz) & Features & FeatPred & CEM & MSE \\
\rowcolor{LightMint}
GR\text{-}2~\cite{cheang2024gr} & Tokenizer & VQGAN/Act & Tokens & Autoreg & Sampling & CE \\
AutoManual~\cite{chen2024automanual} & LLM & LLM & Rules & Iterate & Manual & Success \\
\rowcolor{LightMint}
ActRe~\cite{yang2024react} & LLM & LLM & Rationales & Forced Sampling & Direct Policy & CE \\
V\text{-}JEPA2~\cite{assran2025v} & JEPA & None & Embeds & EmbedPred & MPC+CEM & FeatPred \\
\rowcolor{LightMint}
Genie 2~\cite{parker2024genie} & AE & Latent Diffusion & Video & AR & None & Denoise \\
RoboDreamer~\cite{zhou2024robodreamer} & Parser/T5 & U-Net & CompLat & Video Plan & InvDyn & Denoise \\
\bottomrule
\end{tabular}
}
\end{table}

\subsection{Overview of World Model Paradigms}
\label{sec:paradigm-overview}

An \emph{AI world model} is broadly any mechanism by which an agent captures or accesses approximate environment dynamics. Let $\mathcal{S}$ denote the set of possible environment \emph{states}, $\mathcal{A}$ the set of \emph{actions}, and $\mathcal{O}$ the set of \emph{observations}. In an idealized Markovian framework, the environment is characterized by transition and observation distributions:
\begin{align}
    &T(s'|\,s,\,a) \quad:\quad \mathcal{S}\times\mathcal{A}\;\to\;\Delta(\mathcal{S}), 
    \label{eq:transition}\\
    &O(o|\,s') \quad:\quad \mathcal{S}\;\to\;\Delta(\mathcal{O}), 
    \label{eq:observation}
\end{align}
where $T(\cdot)$ dictates how states evolve under actions, and $O(\cdot)$ defines how states produce observations. A \emph{world model} typically \emph{learns} or \emph{utilizes} approximations of these functions (or a variant), allowing the agent to \emph{predict} future states or observations without executing real actions in the environment.

Numerous approaches exist to implement these approximations, which we group into four main \emph{paradigms}:

\begin{itemize}
    \item \textbf{Implicit paradigm:}
    During planning the agent rolls out a \emph{latent dynamics} function%
    \footnote{Formally $h_{t+1}=f_{\theta}(h_t,a_t)$ with value / policy heads on $h_{t+1}$.} 
    and makes all decisions from the latent state alone.  
    If a decoder $g_{\theta}$ exists, it is used only for auxiliary losses or visualisation, \emph{never} for action selection.  
    Typical examples include World Models~\cite{ha2018world}, MuZero~\cite{schrittwieser2020mastering}, Dreamer~\cite{hafner2019dream}, and token-based generalist agents such as Gato~\cite{reed2022generalist}.  
    The approach yields fast, end-to-end training but offers limited interpretability or direct constraint injection.
    
    \item \textbf{Explicit paradigm:}
    Here the rollout \emph{explicitly predicts future observations} (pixels, point clouds, sensor streams) so that imagined trajectories remain in the same modality as real perception.  
    In practice this often means a factored pair $\hat{s}_{t+1}=T_{\theta}(s_t,a_t)$ and $\hat{o}_{t+1}=O_{\theta}(\hat{s}_{t+1})$, though diffusion-based models can generate observations end-to-end.  
    Visual generative models such as Diffusion WM~\cite{ding2024diffusion} and DINO-WM~\cite{zhou2024dino} fall into this category, offering greater debuggability and the ability to inject structural priors at the cost of heavier computation.

    \item \textbf{Simulator-Based paradigm:}
    Rather than approximating \eqref{eq:transition}--\eqref{eq:observation}, the agent relies on an external simulator or even the physical world as the ground-truth. Systems like SAPIEN \cite{xiang2020sapien} or real-robot pipelines \cite{wu2023daydreamer} can be seen as ``native'' environment models that the agent queries. Although no learned $T(\cdot)$ is required, the agent pays a cost in terms of runtime or real-world risks.

    \item \textbf{Other paradigms (Hybrid or Instruction-Driven):}
    Methods that defy simple classification. They may store emergent rules in textual form~\cite{chen2024automanual}, refine implicit LLM knowledge into partial causal graphs~\cite{liu2024discovery}, or combine external components with learned sub-modules. Such approaches highlight the evolving nature of world-model research, where instructions, symbolic rules, or on-the-fly structures can complement more traditional approximations.
\end{itemize}

Throughout the remainder of this subsection, we examine how each paradigm addresses (or circumvents) Equations~\eqref{eq:transition} and \eqref{eq:observation}, the trade-offs in interpretability and scalability, and their relative merits for different tasks ranging from text-based to high-dimensional embodied control.

\subsection{Implicit Paradigm}
\label{sec:implicit}

We define an \emph{implicit world model} as one in which the agent simulates future dynamics entirely within an internal latent space, without explicitly reconstructing future high-dimensional observations during inference. Even if a decoder exists, for visualization or auxiliary training, it is not used during planning or imagined trajectories. Instead, the system evolves latent representations that are sufficient for decision-making or value prediction. Formally, an implicit world model maintains a latent state $h_t \in \mathcal{H}$, updated as:
\begin{equation}
  h_{t+1} = f_\theta(h_t, a_t), \quad
  \hat{o}_{t+1} = g_\theta(h_{t+1}) \;\; \text{(optional, not used in rollout)}.
\end{equation}

A representative example of the implicit paradigm is the \emph{World Models} framework~\cite{ha2018world}, in which a Variational Autoencoder (VAE) is first trained to encode high-dimensional visual observations into a compressed latent space. Separately, a recurrent network is trained to model the dynamics in this latent space by predicting future latent vectors. While a decoder exists for reconstructing images from latent codes, it is used only during the VAE pretraining or for visualization purposes, not as part of the agent's rollout or decision-making process. During planning, the agent performs simulated rollouts entirely within the latent space using the RNN, without reconstructing future observations. Accordingly, under our definition, World Models is categorized as an \emph{implicit} world model, since its predictions do not include explicit reconstruction of future observations during imagined trajectories.

A more complex example is MuZero~\cite{schrittwieser2020mastering}, which combines a learned latent dynamics model with Monte Carlo Tree Search (MCTS) to plan over imagined futures. Although MuZero achieves high performance in challenging domains such as Go and Atari, it never reconstructs the actual observations (e.g., game board states or video frames) at any point in its planning process. Instead, the model focuses exclusively on predicting reward, value, and policy over abstract latent states. These latent states are not trained to correspond to any observable quantity in the environment, and the learned dynamics function is optimized solely for planning efficacy. Thus, MuZero squarely fits within the implicit paradigm, despite its powerful planning mechanism, as it bypasses observation-level modeling entirely. Model-based reinforcement-learning algorithms like Dreamer~\cite{hafner2019dream} exemplify this paradigm by performing all planning in latent space, using observation reconstruction only for auxiliary training losses.

Recent large‐scale systems further illustrate the \emph{implicit} design philosophy. 
\text{Gato}~\cite{reed2022generalist} demonstrates that a single token‑based model can act in dozens of tasks, from Atari to robotic manipulation, by rolling out purely in latent space. 
Similarly, \text{V-JEPA 2}~\cite{assran2025v} pre‑trains on internet‑scale video and later finetunes a latent action head, enabling zero‑shot robotic planning without ever reconstructing video frames during inference. 
These results strengthen the case that end‑to‑end token world models can generalize broadly while remaining computationally simple.

Because \emph{implicit} world models simulate future dynamics entirely inside a latent space, they never have to \textbf{reconstruct high-dimensional observations during planning}.  A single latent-transition function $f_{\theta}$ is unrolled, and, if present at all, an observation decoder $g_{\theta}$ is used only for auxiliary losses or visualization, \emph{not} for decision making.  This design yields elegant end-to-end training pipelines and fast rollouts, letting large-capacity networks autonomously discover compact structure in complex environments.  The trade-off is opacity: latent variables are not directly interpretable, it is difficult to inject domain priors or constraints, and the model can be brittle under distribution shift because errors in latent prediction may go undetected by perceptual metrics.  Thus, while the implicit paradigm excels in simplicity and computational efficiency, it remains challenging when users require strong interpretability, explicit safety constraints, or fine-grained control over learned dynamics.

\bigskip

\subsection{Explicit Paradigm}
\label{sec:explicit}

We define an \emph{explicit world model} as one that reconstructs or predicts future high-dimensional observations (e.g., images or sensor data) during inference. These imagined observations are actively used for training or decision-making. Such models typically factorize the transition and observation processes into two functions:
\begin{align}
  \hat{s}_{t+1} &= T_\theta(s_t, a_t), \\
  \hat{o}_{t+1} &= O_\theta(\hat{s}_{t+1}).
\end{align}

The predicted observation $\hat{o}_{t+1}$ is then used to inform future actions or value estimates. This makes the world model's predictions interpretable and aligned with the agent's sensory domain. Examples include Dreamer~\cite{hafner2019dream}, Diffusion WM~\cite{alonso2024diffusion}, and DINO-WM~\cite{zhou2024dino}.

This capability to reconstruct future observations during imagined rollouts distinguishes explicit models from implicit ones. While both may use latent representations internally, explicit models treat accurate observation prediction as central to the simulation process. This design encourages interpretability, supports visual debugging, and facilitates interaction with tasks that require fine-grained perceptual feedback.

Explicit approaches prioritize fidelity in generating future frames for decision-making, such as Diffusion WM~\cite{alonso2024diffusion}, which applies diffusion processes at the pixel level, or DINO-WM~\cite{zhou2024dino}, which rolls out future states within a pretrained feature space.

Beyond pixel-CNN or VAE backbones, a line of \text{diffusion‑based}  world models has emerged. \text{Diffusion World Model}(DWM)~\cite{ding2024diffusion} predicts long-horizon trajectories and their rewards in a single denoising pass, demonstrating significant improvements on D4RL benchmarks.
\text{AVID}~\cite{rigter2024avid} shows that closed‑source video diffusion generators can be adapted into fully action‑conditioned world models through lightweight adapters, retaining photorealistic quality. 
Complementing these, \text{PhysDreamer}~\cite{zhang2024physdreamer} generates plausible physics‑based human–object interaction videos, answering ``what‑if'' queries in 3‑D scenes. 
Together, these works underscore how explicit frame‑level generation is becoming both higher‑fidelity and more computationally tractable.

Autonomous‑driving research has embraced explicit world models that output multi‑view video. 
\text{GAIA‑1}~\cite{hu2023gaia} conditions on past camera feeds, text, and control to generate diverse, controllable futures—supporting counterfactual scene editing. 
\text{DriveDreamer}~\cite{wang2024drivedreamer} is the first model trained purely on real‑world driving logs; a two‑stage diffusion pipeline captures fine‑grained traffic constraints and yields accurate frame‑level rollouts. 
\text{Drive\text{-}WM}~\cite{wang2024driving} further blends visual forecasting with image‑reward trajectory optimisation, integrating perception, prediction, and planning into one generative module.

By reconstructing future observations during imagined rollouts, \text{explicit} models expose the full prediction to the user, making internal failure modes observable and enabling the injection of domain priors (e.g., physical constraints, safety rules) at the pixel or point-cloud level.  This transparency facilitates targeted debugging and opens doors to hybrid schemes such as loss weighting on physics invariants or hard-coded collision checkers.  
Yet the same design magnifies \text{compounding-error risk}: any drift in the learned transition $\hat{T}_{\theta}$ or decoder $\hat{O}_{\theta}$ is immediately visible in the rendered trajectory and can poison multi-step planning.  Scaling to long horizons therefore requires either (i) very accurate models, (ii) frequent real-world replanning, or (iii) specialised corrections such as diffusion-based denoising or hindsight relabelling.  
Moreover, explicit rollouts incur higher compute and memory costs because every imagined frame (or point cloud) must be produced, stored, and evaluated.  Consequently, explicit world models excel when interpretability and hard constraints outweigh raw simulation speed, but they remain challenging in data-scarce regimes and under severe distribution shift.

\bigskip

\subsection{Simulator-Based Paradigm}
\label{sec:simulator}

In the \emph{simulator-based} paradigm, the agent outsources environment updates to a simulator, effectively bypassing the need to learn $\hat{T}_\theta$ from data. Formally,
\begin{equation}
  (s_{t+1}, o_{t+1}) \leftarrow \text{SIM}(s_t, a_t),
\end{equation}
where \(\text{SIM}\) is often an external physics engine or the real world itself. Platforms like SAPIEN~\cite{xiang2020sapien} and AI Habitat provide deterministic 3D physics simulations, allowing agents to practice or iterate strategies in a controlled environment. Alternatively, methods such as Daydreamer~\cite{wu2023daydreamer} treat real-world interaction loops like a ``simulator'', continually updating on-policy data from physical robots.

Tokenised game engines can also serve as high‑fidelity simulators. 
\text{MineWorld}~\cite{guo2025mineworld} converts Minecraft video frames and actions into discrete tokens and trains a parallel Transformer that renders 4–7 fps in real time, offering an open‑source, interactive sandbox where agents can query a ``perfect'' simulator instead of learning dynamics from scratch.

This approach yields accurate transitions (assuming the simulator accurately reflects reality), which alleviates the risk of learned-model errors. However, it can be computationally or financially expensive, especially if the simulator is high fidelity or if real-world trials are time-consuming and risky. As a result, some agents combine partial learned dynamics with occasional simulator queries, aiming to balance accurate rollouts with efficient coverage of state-action space.

\bigskip

\subsection{Hybrid and Instruction-Driven Paradigms}
\label{sec:other}

Beyond these three primary paradigms, there is a growing number of \emph{hybrid} or \emph{instruction-driven} approaches, which blend implicit and explicit modeling or incorporate external symbolic knowledge and large language models. Often, these systems dynamically extract rules from data, maintain evolving textual knowledge bases, or prompt LLMs to hypothesize causal relationships that can then be tested or refined.

AutoManual~\cite{chen2024automanual}, for example, iteratively compiles interactive environment rules into human-readable manuals, informing future actions in a more transparent way. Meanwhile, COAT~\cite{liu2024discovery} prompts an LLM to propose possible causal factors behind observed events, then validates or refines those factors via direct interaction, bridging text-based reasoning with partial learned models. Although these solutions offer remarkable flexibility, particularly in adapting to unfamiliar domains or integrating real-time human insights, they can be inconsistent in how they structure or update internal representations. As language-model prompting and real-time rule discovery continue to evolve, these hybrid methods are poised to become increasingly common, reflecting the need to balance end-to-end learning with the transparency and adaptability offered by external instruction. 

Recent work pushes the hybrid idea further by letting large language models write rules, code, or even 3-D scenes on the fly. WorldCoder~\cite{tang2024worldcoder} turns environment feedback into executable Python snippets, yielding explicit, editable simulators that the agent can re-compile between episodes. Genie 2~\cite{parker2024genie} expands a single image into an interactive 3-D world and maintains memory of occluded objects, effectively using text prompts as a high-level world‐generation interface. RoboDreamer~\cite{zhou2024robodreamer} parses human instructions into low-level primitives and trains a video generator conditioned on those primitives, giving robots compositional imagination. Finally, WMA~\cite{chae2024web} tokenises HTML and JavaScript events to learn latent dynamics of web pages, enabling zero-shot navigation by rolling out the model in token space. Together these systems illustrate how language, code, and generative priors can merge with latent dynamics to create flexible instruction-driven world models.

Until now, we have introduced the four typical paradigms of existing world model techniques, as illustrated in Figure~\ref{fig:wm-paradigms}. As we can see, each type of technique has trade-offs in different aspects.

\begin{figure}[!ht]
\centering
\begin{tikzpicture}[
    font=\small,
    >=Stealth,
    thick,
    observation/.style={circle, draw=black, fill=customred!50, minimum width=7mm},
    hidden/.style={circle, draw=black, fill=customblue!50, minimum width=7mm},
    block/.style={rectangle, draw=black, fill=gray!20, rounded corners=3pt, 
                  align=center, minimum width=14mm, minimum height=8mm},
    arrow/.style={->, line width=1.2pt}
]

\coordinate (TL_a) at (0,0);         
\coordinate (center_a) at ($(TL_a) + (3.5,0)$); 
\node[anchor=south] at (1,0) {\textbf{(a) Implicit}};

\coordinate (diag_a) at ($(TL_a) + (0.5,-2)$);

\node[hidden, label={above:$h_t$}] (ht)
  at ($(diag_a) + (0,0)$) {};

\node[block, fill=customblue!50, text width=14mm, align=center] (f)
  at ($(ht) + (2.2,0)$) {Implicit\\Model};

\node[hidden, label={[label distance=-1pt]above:$h_{t+1}$}] (ht1)
  at ($(f) + (2.2,0)$) {};

\node[rectangle, draw=black, fill=customgreen!50, rounded corners=3pt,
      minimum width=9mm, label={above:$a_t$}] (at)
  at ($(ht)!0.5!(f) + (0,1)$) {};

\node[observation, label={[label distance=1pt]above:$\hat{o}_{t+1}$}] (ot1)
  at ($(ht1) + (0,1.2)$) {};

\draw[arrow] (ht) -- (f);
\draw[arrow] (at) -- (f);
\draw[arrow] (f) -- (ht1);
\draw[arrow, dashed] (f.15) -- (ot1);

\coordinate (TL_b) at (7,0);          
\coordinate (center_b) at ($(TL_b) + (3.5,0)$);

\node[anchor=south] at (8.8, 0) {\textbf{(b) Explicit}};

\coordinate (diag_b) at ($(TL_b) + (1.5,-2.5)$);

\node[rectangle, draw=black, fill=gray!10, label={below:$s_t$}, 
      rounded corners=3pt, minimum width=10mm] (st)
  at ($(diag_b) + (0,0)$) {};

\node[block, draw=black, fill=customblue!50, text width=14mm, align=center] (T)
  at ($(st) + (2.2,0)$) {$\hat{T}_\theta$};

\node[rectangle, draw=black, fill=gray!10, label={below:$\hat{s}_{t+1}$}, 
      rounded corners=3pt, minimum width=10mm] (st1)
  at ($(T) + (2.2,0)$) {};

\node[block, draw=black, fill=customblue!50, text width=14mm, align=center] (O)
  at ($(st1) + (0,1.0)$) {$\hat{O}_\theta$};

\node[observation, label={above:$\hat{o}_{t+1}$}] (ot1b)
  at ($(O) + (0,1.2)$) {};

\node[rectangle, draw=black, fill=customgreen!50, rounded corners=3pt, 
      minimum width=9mm, label={above:$a_t$}] (atb)
  at ($(st)!0.5!(T) + (0,1.0)$) {};

\draw[arrow] (st) -- (T);
\draw[arrow] (atb) -- (T);
\draw[arrow] (T) -- (st1);
\draw[arrow] (st1) -- (O);
\draw[arrow] (O) -- (ot1b);

\coordinate (TL_c) at (0,-4);         
\coordinate (center_c) at ($(TL_c) + (3.5,0)$);

\node[anchor=south] at (1.7, -4) {\textbf{(c) Simulator-Based}};

\coordinate (diag_c) at ($(TL_c) + (0.5,-1.9)$);

\node[rectangle, draw=black, fill=gray!10, label={below:$s_t$}, 
      rounded corners=3pt, minimum width=10mm]
  (st_c) at ($(diag_c) + (0,0)$) {};

\node[rectangle, draw=black, fill=customgreen!50, label={above:$a_t$}, 
      rounded corners=3pt, minimum width=9mm]
  (at_c) at ($(st_c) + (0.9,1)$) {};

\node[block, draw=black, fill=customblue!50, text width=14mm, align=center] (sim)
  at ($(st_c) + (2.4,0)$) {Simulator};

\node[rectangle, draw=black, fill=gray!10, label={below:$s_{t+1}$}, 
      rounded corners=3pt, minimum width=10mm]
  (st1_c) at ($(sim) + (2.4,0)$) {};

\node[observation, label={[label distance=1pt]above:$o_{t+1}$}]
  (ot1_c) at ($(st1_c) + (0,1.2)$) {};

\draw[arrow] (st_c) -- (sim);
\draw[arrow] (at_c) -- (sim);
\draw[arrow] (sim) -- (st1_c);
\draw[arrow] (sim.40) -- (ot1_c);

\coordinate (TL_d) at (7,-4);         

\coordinate (center_d) at ($(TL_d) + (3.5,0)$);

\node[anchor=south] at (center_d) {\textbf{(d) Hybrid / Instruction-Driven}};

\coordinate (diag_d) at ($(TL_d) + (1.9,-1)$);

\node[block, draw=black, fill=customblue!50, text width=16mm, align=center] (hybrid_model)
  at ($(diag_d) + (0,0)$) {Implicit or\\Partial Model};

\node[block, draw=black, fill=customred!50, text width=16mm, align=center] (hybrid_kb)
  at ($(hybrid_model) + (5.0,0)$) {LLM /\\Rules KB};

\draw[arrow] (hybrid_model) -- (hybrid_kb)
  node[midway, above, xshift=0mm, font=\footnotesize]{\textit{prompts / updates}};

\node[rectangle, draw=black, fill=gray!10, 
      label={below:$\tilde{s}_t,\;\tilde{o}_t$},
      rounded corners=3pt, minimum width=12mm, align=center]
  (state_obs) at ($(hybrid_model)!0.5!(hybrid_kb) + (0,-1.6)$) {Refined Prediction};

\draw[arrow] (state_obs) -- (hybrid_model)
  node[midway, left, font=\footnotesize, xshift=-1pt]{\textit{train / refine}};
\draw[arrow] (hybrid_kb) -- (state_obs)
  node[midway, right, font=\footnotesize, xshift=1pt]{\textit{rules / constraints}};

\end{tikzpicture}
\label{fig:wm-paradigms}
\caption{Four paradigms of world modeling: (a) implicit, (b) explicit, (c) simulator-based, and (d) hybrid/instruction-driven. These diagrams illustrate the architectural forms of world models under different design paradigms.}
\end{figure}

\subsection{Comparative Summary of Paradigms}
\label{sec:summary-table}

\begin{table}[ht]
\small
\centering
\caption{Summary of AI world-model methods across paradigms, showing their form (External $\circ$ or Internal $\bullet$) , complexity, and paradigm.}
\label{tab:wm-summary}
\begin{tabular}{@{}l c c l@{}}
\toprule
\textbf{Method} & \textbf{Form} & \textbf{Complexity} & \textbf{Paradigm} \\
\midrule
\rowcolor{LightMint}
ActRe \cite{yang2024react} & \(\bullet\) & Simple & Implicit \\
World Models \cite{ha2018world} & \(\bullet\) & Simple & Implicit \\
Dreamer \cite{hafner2019dream} & \(\bullet\) & Moderate & Implict \\
\rowcolor{LightMint}
Diffusion WM \cite{alonso2024diffusion} & \(\bullet\) & High & Explicit \\
GQN \cite{eslami2018neural} & \(\bullet\) & High & Explicit \\
Daydreamer \cite{wu2023daydreamer} & \(\bullet\) & High & Simulator-based \\
\rowcolor{LightMint}
SAPIEN \cite{xiang2020sapien} & \(\circ\) & High & Simulator-based \\
PILCO \cite{deisenroth2011pilco} & \(\bullet\) & Moderate & Implicit \\
\rowcolor{LightMint}
AutoManual \cite{chen2024automanual} & \(\circ\) & Simple & Other \\
MuZero \cite{schrittwieser2020mastering} & \(\circ\) & High & Implicit \\
\rowcolor{LightMint}
GR-2 \cite{cheang2024gr} & \(\bullet\) & High & Explicit \\
DINO-WM \cite{zhou2024dino} & \(\bullet\) & High & Explicit \\
\rowcolor{LightMint}
COAT \cite{liu2024discovery} & \(\circ\) & Moderate & Other \\
\bottomrule
\end{tabular}
\end{table}

The table summarizes the key methods in AI world modeling, categorizing them based on their reliance on \emph{external} or \emph{internal} mechanisms, their complexity, and their respective paradigms. The form column uses \(\circ\) for external approaches and \(\bullet\) for internal ones, with mixed methods having both symbols. This classification aligns with the previous subsections, including the detailed discussion of each paradigm, and complements the visual representation in Figure~\ref{fig:wm-twodim}. Table~\ref{tab:wm_dimensions} further details representative methods along encoder/decoder, latent space, rollout, planning, and objective, complementing the paradigm-level summary (form, complexity, paradigm) above.

\section{Relationships to Other Modules}
\label{sec:relationship-other-modules}

\lettrine[lines=3]{\initfamily\textcolor{darkgreen}{A}}{I world model} does not exist in isolation but interacts with several key components of the agent's architecture. These include (but are not limited to) the memory, perception, and action modules. In this subsection, we explore how world models integrate with these critical components to enable coherent and adaptive behavior in dynamic environments.

\subsection{Memory and the World Model}
Memory systems play a crucial role in the operation of world models. While a world model generates predictive representations of future states or actions, memory serves as the foundation upon which these representations are built and updated. The relationship between the world model and memory can be viewed as a loop where the world model predicts potential futures, while the memory stores past experiences, observations, and learned patterns, allowing for context-dependent reasoning and future predictions.

Memory mechanisms can be structured in various ways, including:
\begin{itemize}
    \item \textbf{Short-term memory}: This enables the agent to hold and update its internal state temporarily, storing the most recent interactions or observations. This short-term context helps the agent make decisions in the immediate environment.
    \item \textbf{Long-term memory}: This serves as a more persistent repository of experiences and general knowledge about the environment. A world model can interact with long-term memory to refine its predictions, and it may use historical data to make more informed decisions or simulate more realistic futures.
\end{itemize}

For example, in model-based RL frameworks like Dreamer \cite{hafner2019dream}, recurrent neural networks act as both the world model and a form of memory, maintaining a latent state that is updated with each time step to predict future states. This form of integrated memory allows the agent to both recall past interactions and anticipate future ones.

\subsection{Perception and the World Model}
Perception refers to the agent's ability to sense and interpret its environment through various modalities (e.g., vision, touch, sound, etc.). The world model relies heavily on accurate sensory input to form coherent predictions about the environment. In many AI systems, the perception module converts raw sensor data into a higher-level representation, such as an image, sound wave, or other structured data.

A key aspect of the interaction between the world model and perception is how the agent processes and integrates sensory input into the model. The world model often depends on processed data (such as features from convolutional neural networks or embeddings from transformers) to simulate potential futures. Additionally, the world model can guide perceptual processes by focusing attention on the most relevant sensory input needed to refine predictions.

For example, in autonomous robotics, perception systems typically detect objects or environmental features, which are then fed into a world model that predicts how the scene will evolve. RoboCraft~\cite{shi2024robocraft} achieves this perception-to-modeling transformation by converting visual observations into particles and capturing the underlying system structure through graph neural networks. PointNet~\cite{qian2022pointnext} further enriches perception systems' understanding of physical space by encoding unstructured 3D point clouds to capture spatial characteristics of the environment. In navigation tasks,  OVER-NAV~\cite{zhao2024over} further combine large language models and open-vocabulary detection to construct the relationship between multi-modal signals and key information, proposing an omni-graph to capture the structure of local space as the world model for navigation tasks. This feedback loop between perception and the world model enables agents to update their perception dynamically based on ongoing predictions, allowing for real-time adaptation.

\subsection{Action and the World Model}
Action refers to the decision-making process through which an agent interacts with its environment. In agentic systems, actions are driven by the world model's predictions of future states. The world model aids in planning by simulating the outcomes of different actions before they are executed, allowing the agent to choose the most optimal course of action based on the predicted consequences.

The integration between world models and action modules can take various forms:
\begin{itemize}
    \item \textbf{Model-based planning}: World models explicitly model the environment's transition dynamics~\cite{schrittwieser2020mastering, kouvaritakis2016model, gu2024your}, allowing the agent to simulate multiple action sequences (rollouts) before selecting the most optimal one.
    \item \textbf{Exploration}: World models also support exploration strategies by simulating unseen states or unexpected actions~\cite{nagabandi2018neural, hafner2019dream, hafner2020mastering}. These simulations enable the agent to evaluate the potential benefits of exploring new parts of the state space. 
\end{itemize}

In model-based planning, MuZero~\cite{schrittwieser2020mastering} performs implicit planning through self-play and Monte Carlo Tree Search (MCTS), transforming current state representations into future state and reward predictions to guide the decision-making process without prior knowledge of environment rules. In contrast, MPC~\cite{kouvaritakis2016model} utilizes explicit dynamics models to predict multiple possible trajectories within a finite time horizon, determines the optimal control sequence by solving an optimization problem, and continuously updates planning using a receding horizon approach. 

World-model agents now drive exploration by imagining what lies just beyond the data they have seen.  Dreamer~\cite{hafner2019dream} rolls out thousands of latent trajectories per real action and rewards those that reach novel latent states, giving safe yet diverse curiosity without physical risk.  Dreamer V2~\cite{hafner2020mastering} switches to a discrete latent space that represents uncertainty more directly, accelerating exploration on Atari and continuous-control tasks.  Dreamer V3~\cite{hafner2025mastering} extends the idea to long-horizon domains such as Minecraft, where adaptive curiosity bonuses in imagination enable zero-shot completion of difficult objectives.

In reinforcement learning generally, a learned world model allows the agent to score many hypothetical trajectories before acting in the real environment, compare their predicted returns, and pick actions that maximise long-term goals while respecting any safety or cost constraints encoded in simulation.

World models ultimately matter because they let an agent choose actions before the real world can punish mistakes.  Classical pipelines query the model with sampling or tree search and keep only the first action of the best simulated trajectory.  Newer approaches fold learning and planning together: Dual Preference Optimisation (D\textsuperscript{2}PO
)~\cite{wang2025world} ranks imagined rollouts by human feedback so that both the model and the policy improve toward what users actually care about, while RoboDreamer~\cite{zhou2024robodreamer} grounds a language goal into low-level action tokens and selects the token sequence whose predicted video has the highest success probability.  These results illustrate a trend toward tighter coupling of model, objective, and planner rather than treating planning as a separate post-hoc module.

\subsection{Cross-Module Integration}
While memory, perception, and action are discussed as separate modules, the true strength of world models lies in their ability to seamlessly integrate across these domains. A world model continuously receives sensory input, updates its internal memory, simulates future states, and uses this information to drive action selection. The iterative feedback loop between these modules allows agents to engage in intelligent, goal-directed behavior that is highly adaptive to changes in the environment.

This cross-module interaction is particularly relevant in complex, dynamic systems such as robotics, where an agent must continuously adapt its internal representation of the world, process sensory input, store relevant experiences, and take actions in real time. In the context of embodied agents, the integration of these modules ensures that predictions made by the world model are grounded in current observations and the agent's ongoing experiences.

World models provide a fundamental unifying principle across modalities. Whether predicting physical outcomes in embodied robotics, anticipating visual changes on screens, or inferring semantic relationships in text, the core mechanism remains consistent: generating predictions about how states evolve under different actions. This cross-modal capacity explains why humans can easily transition between manipulating objects, navigating interfaces, and processing language, all activities driven by the same underlying predictive architecture. Future AI systems may achieve similar integration by developing world models that bridge these traditionally separate domains through a common predictive framework.

In summary, the relationship between the world model and the other modules, memory, perception, and action, forms the backbone of intelligent behavior in AI systems. Each module contributes to a cycle of prediction, update, and action, allowing agents to function effectively in dynamic and uncertain environments. These interactions highlight the need for a holistic approach when designing agent architectures, where world models are closely intertwined with sensory input, memory systems, and decision-making processes.

\section{Summary and Discussion}
\label{sec:discussion}

\lettrine[lines=3]{\initfamily\textcolor{darkgreen}{T}}{he} evolution of AI world models, from early cognitive insights to advanced AI architectures, underscores the growing realization that true intelligence relies on the ability to predict, simulate, and imagine. Unlike classical reinforcement learning, where agents operate solely through trial-and-error interactions, world models enable foresight: agents can plan, anticipate, and adapt to changes before they happen. This leap in cognitive modeling, whether implicit, explicit, or simulator-based, marks a significant shift in how machines can be endowed with flexibility, robustness, and generalization across tasks.

An essential yet often overlooked aspect of world models is their operation across multiple temporal and spatial scales. Human mental models seamlessly integrate predictions spanning milliseconds (reflexive responses), seconds (immediate action planning), minutes to hours (task completion), and even years (life planning)~\cite{newell1994unified}. This multi-scale capability allows us to simultaneously predict immediate physical dynamics while maintaining coherent long-term narratives and goals. Similarly, humans process spatial information across scales, from fine-grained object manipulation to navigation across environments to abstract geographical reasoning. Current AI world models typically excel within narrow temporal and spatial bands, whereas human cognition demonstrates remarkable flexibility in scaling predictions up and down as context demands. This suggests that truly general-purpose AI world models may require explicit mechanisms for integrating predictions across multiple time horizons and spatial resolutions, dynamically adjusting the granularity of simulation based on task requirements.

One central challenge in designing world models is the interplay between \emph{complexity} and \emph{predictive accuracy}. As discussed, implicit models, such as those based on recurrent neural networks or transformers, offer simplicity and elegance, but they often come with the trade-off of limited interpretability. The model's internal state is an opaque latent space, making it difficult to enforce domain constraints or provide guarantees about the accuracy of predictions. While such systems excel at capturing highly complex relationships and data-driven patterns, they also risk overfitting or failing to generalize to unseen scenarios.

Explicit models, by contrast, offer greater transparency and control. By factorizing state transitions and observations into separate functions, we gain a clearer understanding of how predictions are formed, and we can more easily integrate structured knowledge, such as physical laws or domain-specific rules. However, this approach comes with its own set of challenges. First, it often requires large amounts of labeled training data or simulated experiences to accurately capture environment dynamics. Second, even the most well-structured explicit models may struggle with complex environments that require fine-grained, high-dimensional state representations, such as in video prediction or robotics.

The \emph{simulator-based} approach offers a promising alternative, wherein agents rely on external environments, either physically grounded or simulated, for dynamic updates. This method avoids many of the challenges inherent in learning accurate world models from scratch, as the simulator itself serves as the ``oracle'' of state transitions and observations. However, reliance on simulators also introduces limitations: simulators often fail to capture the full richness of real-world dynamics and can be computationally expensive to maintain or scale. Furthermore, real-world environments introduce noise and variability that a purely learned or pre-configured model might miss. As AI agents strive to perform tasks in open-ended, unpredictable settings, the robustness of their world models will be tested by the gap between simulated and actual environments.

A key theme that emerges from this discussion is the \emph{trade-off between generalization and specialization}. The more specific a world model is to a particular domain or task, the less likely it is to generalize across different contexts. Models like MuZero \cite{schrittwieser2020mastering} and Dreamer~\cite{hafner2019dream} exemplify this: they excel at specific environments (e.g., Atari games or robotics) but require careful adaptation when transferred to new, uncharted domains. In contrast, implicit models, particularly those that use large-scale neural networks, have the potential to generalize between tasks, but often do so at the cost of sacrificing domain-specific expertise.

Moreover, \emph{integrating memory} with world models is crucial for agents that need to handle long-term dependencies and past experiences. While world models excel at predicting the next state based on immediate inputs, true intelligent behavior often requires reasoning about distant outcomes. Long-term memory allows agents to store critical environmental knowledge, ensuring that short-term predictions are grounded in a broader understanding of the world. This fusion of memory, perception, and action, mediated by the world model, creates a feedback loop where predictions shape actions, which in turn inform future predictions.

The \emph{human analogy} remains compelling: just as humans integrate sensory inputs, memories, and internal models to navigate the world, so too must intelligent agents combine perception, memory, and action through their world models. As the field advances, it is clear that a holistic approach, one that unifies implicit, explicit, and simulator-based methods, may be the key to achieving more robust, generalizable, and adaptive agents. Hybrid methods, such as those used in AutoManual~\cite{chen2024automanual} or discovery-based models \cite{liu2024discovery}, offer exciting possibilities for blending learned knowledge with structured rules and real-time interactions, potentially pushing the boundaries of what we consider a world model.

Several recent trends point to a convergence of scale, modality, and causality. Dreamer V3~\cite{hafner2025mastering} shows that a single configuration can surpass specialised baselines on more than 150 tasks and even solve the long-horizon Minecraft diamond-mining benchmark without additional tuning. Large token models such as Gato~\cite{reed2022generalist} and V-JEPA 2~\cite{assran2025v} highlight that unifying vision, language, and control into one latent predictor yields broad transfer with minimal task-specific code. On the theory side, \text{Robust Agents Learn Causal World Models}~\cite{richens2024robust} and General Agents Need World Models~\cite{richens2025general} formalise why agents that act robustly must internalise causal structure, providing an analytical backbone to the empirical successes above. These developments suggest that scaling alone is not enough; explicitly aligning predictive latent space with causal factors and preference signals may be the key to truly general agents.

Looking forward, \emph{open questions remain}. How can we ensure that world models exhibit \emph{long-term stability} and \emph{reliability} in real-world settings? How do we handle the inherent \emph{uncertainty} in dynamic environments while maintaining the flexibility to adapt? Furthermore, as agents grow more sophisticated, how can we design systems that are both \emph{efficient} and \emph{scalable} across increasingly complex tasks without incurring massive computational costs?

In conclusion, the future of world models lies in their ability to balance the need for \emph{generalization} with the requirement for \emph{domain expertise}. By continuing to explore and refine the interplay between model simplicity and complexity, between external and internal approaches, we move closer to developing AI systems that not only understand the world but can actively shape their understanding to navigate and adapt in a rapidly changing reality.
\chapter{Reward} 
\label{Chapter:reward}

\begin{figure}[!ht]
\centering
\footnotesize
    \begin{forest}
        for tree={
            forked edges,
            draw,
            rounded corners,
            node options={align=center,},
            s sep=6pt,
            calign=center,
            grow=east,
            reversed=true,
            anchor=base west,
            parent anchor=east,
            child anchor=west,
            base=left,
            font=\small,
            minimum width=2.5em,
          },
          where level=1{text width=5em,fill=customblue!50}{},
          where level=2{text width=5em,fill=customgreen!50}{},
        [Reward, fill=gray!20
            [Extrinsic Reward, for tree={
                answer,
                calign=child edge, calign child=(n_children()+1)/2,
                yshift=-0.2cm,
                }
                    [Dense Reward, text width=80pt, 
                        [InstructGPT~\cite{ouyang2022training}
                        DRO~\cite{richemond2024offline}
                        sDPO~\cite{kim2024sdpo}
                        $\Psi$PO~\cite{azar2024general}
                        $\beta$-DPO~\cite{wu2025beta}
                        ORPO~\cite{hong2024orpo}
                        DNO~\cite{rosset2024direct}
                        $\mathit{f}$-DPO~\cite{wang2023beyond}
                        ~\cite{xu2023some}
                        Tldr~\cite{fu2024tldr} GRIT~\cite{fan2025grit} VARD~\cite{dai2025vard} Visionary-R1~\cite{xia2025visionary} VisualQulity-R1~\cite{wu2025visualquality} R1-SGG~\cite{chen2025compile}, answer_work, text width=180pt]
                    ]
                    [Sparse Reward, text width=80pt, 
                        [PAFT~\cite{pentyala2024paft}
                        SimPO~\cite{meng2025simpo}
                        LiPO~\cite{liu2024lipo}
                        RRHF~\cite{yuan2023rrhf}
                        PRO~\cite{song2024preference}
                        D$^2$O~\cite{duan2024negating}
                        NPO~\cite{zhang2024negative}
                        \cite{ahmadian2024back} STAR-R1~\cite{li2025star} DanceGRPO~\cite{xue2025dancegrpo}, answer_work, text width=180pt]
                    ]
                    [Delayed Reward, text width=80pt, 
                        [CPO~\cite{xu2024contrastive}
                        NLHF~\cite{munos2023nash}
                        ~\cite{swamy2024minimaximalist} RRM~\cite{guo2025reward}, answer_work, text width=180pt]
                    ]
                    [Adaptive Reward, text width=80pt, 
                        [
                        NLHF~\cite{munos2023nash}
                        ~\cite{swamy2024minimaximalist}
                        $\mathit{f}$-DPO~\cite{wang2023beyond}  
                        RSD~\cite{liao2025reward} MPO~\cite{kim2025toward} RiC~\cite{yang2024rewards} RewardAnything~\cite{yu2025rewardanything} GoT-R1~\cite{duan2025got} SophiaVL-R1~\cite{fan2025sophiavl} R-GRPO~\cite{jiang2025vlm} TON~\cite{wang2025think} CoF~\cite{zhang2025chain} DeepEyes~\cite{zheng2025deepeyes} UniVG-R1~\cite{bai2025univg} G1~\cite{chen2025g1} GuardReasoner-VL~\cite{liu2025guardreasoner} V-ToolRL~\cite{su2025openthinkimg} Flow-GRPO~\cite{liu2025flow} X-Reasoner~\cite{liu2025x} FAST~\cite{xiao2025fast} Relation-R1~\cite{li2025relation}, answer_work, text width=180pt]
                    ]
            ]
            [Intrinsic Reward, for tree={
                answer,
                calign=child edge, calign child=(n_children()+1)/2,
                }
                    [Curiosity-Driven Reward, text width=80pt, 
                        [~\cite{pathak2017curiosity}
                        ~\cite{pathak2019self} Pixel Reasoner~\cite{su2025pixel}
                        Plan2Explore~\cite{sekar2020planning}, answer_work, text width=180pt]
                    ]
                    [Diversity Reward, text width=80pt, 
                        [LIIR~\cite{du2019liir} NoiseRollout~\cite{liu2025noisyrollout}, answer_work, text width=180pt]
                    ]
                    [Competence-Based Reward, text width=80pt, 
                        [CURIOUS~\cite{colas2019curious}
                        Skew-Fit~\cite{pong2019skew}
                        DISCERN~\cite{hassani2021discern}
                        ~\cite{yuan2401self}
                        KTO~\cite{ethayarajh2024kto} VisionReasoner~\cite{liu2025visionreasoner}, answer_work, text width=180pt]
                    ]
                    [Exploration Reward, text width=80pt, 
                        [~\cite{yuan2401self}
                        ~\cite{burda2018exploration} SELM~\cite{zhang2024self} VPRL~\cite{xu2025visual}, answer_work, text width=180pt]
                    ]
                    [Information Gain Reward, text width=80pt, 
                        [~\cite{ton2024understanding}
                        VIME~\cite{houthooft2016vime}
                        EMI~\cite{kim2018emi}
                        MAX~\cite{shyam2019model}
                        KTO~\cite{ethayarajh2024kto}
                        ReasRMs~\cite{chen2025rm}, answer_work, text width=180pt]
                    ]
            ]
            [Hybrid Reward, for tree={
                answer,
                calign=child edge, calign child=(n_children()+1)/2,
                }
                    [Combination of Intrinsic and Extrinsic Reward, text width=80pt, 
                        [d-RLAIF~\cite{lee2023rlaif}
                        ~\cite{bai2022constitutional}
                        ~\cite{xiong2023iterative}
                        ~\cite{dong2024rlhf}
                        RewardAgent~\cite{peng2025agentic} Skywork R1V2~\cite{wang2025skywork}, answer_work, text width=180pt]
                    ]
            ]
            [Hierarchical Reward, for tree={
                answer,
                calign=child edge, calign child=(n_children()+1)/2,
                }
                    [Hierarchical Reward, text width=80pt, 
                        [TDPO~\cite{zeng2024token} URSA~\cite{luo2025ursa} Share-GRPO~\cite{yao2025r1} Visual-ARFT~\cite{liu2025visual} T2I-R1~\cite{jiang2025t2i}, answer_work, text width=180pt]
                    ]
            ]
        ]
    \end{forest}
    \label{fig:reward-tree}
    \caption{A taxonomy of selected research works about reward systems.}
\end{figure}

\begin{table*}\centering
\label{tab:human-reward-comparison}
\caption{The comparison of human common reward pathways.}
\ra{1.3}
\begin{tabular}{p{3cm}p{2.5cm}p{9.5cm}}

\toprule
\textbf{Reward Pathway} & \textbf{Neurotransmitter}  & \textbf{Mechanism} \\
\midrule

\rowcolor{LightMint}
\textbf{Mesolimbic pathway~\cite{lewis2021brain}} &
Dopamine  &
Dopamine neurons in the VTA project to the nucleus accumbens, releasing dopamine that binds to D1 (excitatory) and D2 (inhibitory) receptors, modulating motivation, reinforcement, and reward-seeking. \textit{Primary circuit for encoding reward prediction and reinforcing behaviors.} \\

\textbf{Mesocortical pathway~\cite{fakhoury2021brain}} &
Dopamine &
VTA neurons project to the PFC, where dopamine influences receptors involved in cognition, emotion, and reward anticipation. \textit{Supports higher-order evaluation of reward outcomes and decision-making.} \\

\rowcolor{LightMint}
\textbf{Nigrostriatal pathway~\cite{fakhoury2021brain}} &
Dopamine &
Dopamine acts on D1 and D2 receptors in the striatum to regulate motor activity and learned reward-related behaviors. \textit{Links reward processing with motor planning and habit formation.} \\

\textbf{Locus coeruleus~\cite{breton2019active}} &
Norepinephrine &
Locus coeruleus neurons release norepinephrine across the brain, activating $\alpha$ and $\beta$ adrenergic receptors to enhance arousal, attention, and stress responses. \textit{Modulates reward salience and attentional engagement.} \\

\rowcolor{LightMint}
\textbf{Glutamatergic projection~\cite{qi2014glutamatergic}} &
Glutamate &
Glutamate binds to AMPA, NMDA, and metabotropic receptors, driving excitatory signaling and plasticity in reward circuits. \textit{Crucial for learning reward associations and reinforcing adaptive behavior.} \\

\textbf{GABAergic modulation~\cite{sharpe2017lateral}} &
Gamma-Aminobutyric Acid (GABA) &
GABA binds to GABAA and GABAB receptors to inhibit postsynaptic activity, maintaining balance in reward pathways. \textit{Provides inhibitory control over excitatory reward signaling to prevent overstimulation.} \\

\bottomrule
\end{tabular}
\end{table*}


\lettrine[lines=3]{\initfamily\textcolor{darkgreen}{R}}{eward} lies at the heart of both biological and artificial intelligence. In humans and animals, reward mechanisms govern decision-making, drive motivation, reinforce behavior, and shape the learning process through complex neurochemical pathways. In AI agents, reward functions serve as the foundational signals that guide behavior optimization, enabling systems to learn from experience, adapt to goals, and make intelligent choices in dynamic environments. 
Reward is the connective tissue between goals and learning: it operationalizes abstract values into measurable feedback, enabling both humans and machines to adapt and improve their behavior over time.
Understanding reward is therefore not only central to the study of cognition and behavior but also critical for the development of robust, generalizable, and aligned AI systems. This chapter explores the multifaceted concept of reward, bridging its biological underpinnings in the human brain with its algorithmic formulations in AI agents. We begin by examining the human reward system, highlighting key neurotransmitters and anatomical pathways that support reinforcement and motivation. We then introduce the design space of agent reward models, discussing how different types of reward—extrinsic, intrinsic, hybrid, and hierarchical—contribute to intelligent behavior. Finally, we analyze the interaction between reward and other core modules such as perception, memory, and emotion, and discuss open challenges such as reward sparsity, reward hacking, and reward misspecification. By integrating insights from neuroscience and AI, this chapter provides a comprehensive foundation for understanding reward as a central principle of intelligent systems.

\section{The Human Reward Pathway}
\label{sec:human-reward}

\begin{figure}[!ht]
\centering
\includegraphics[width=0.8\columnwidth]{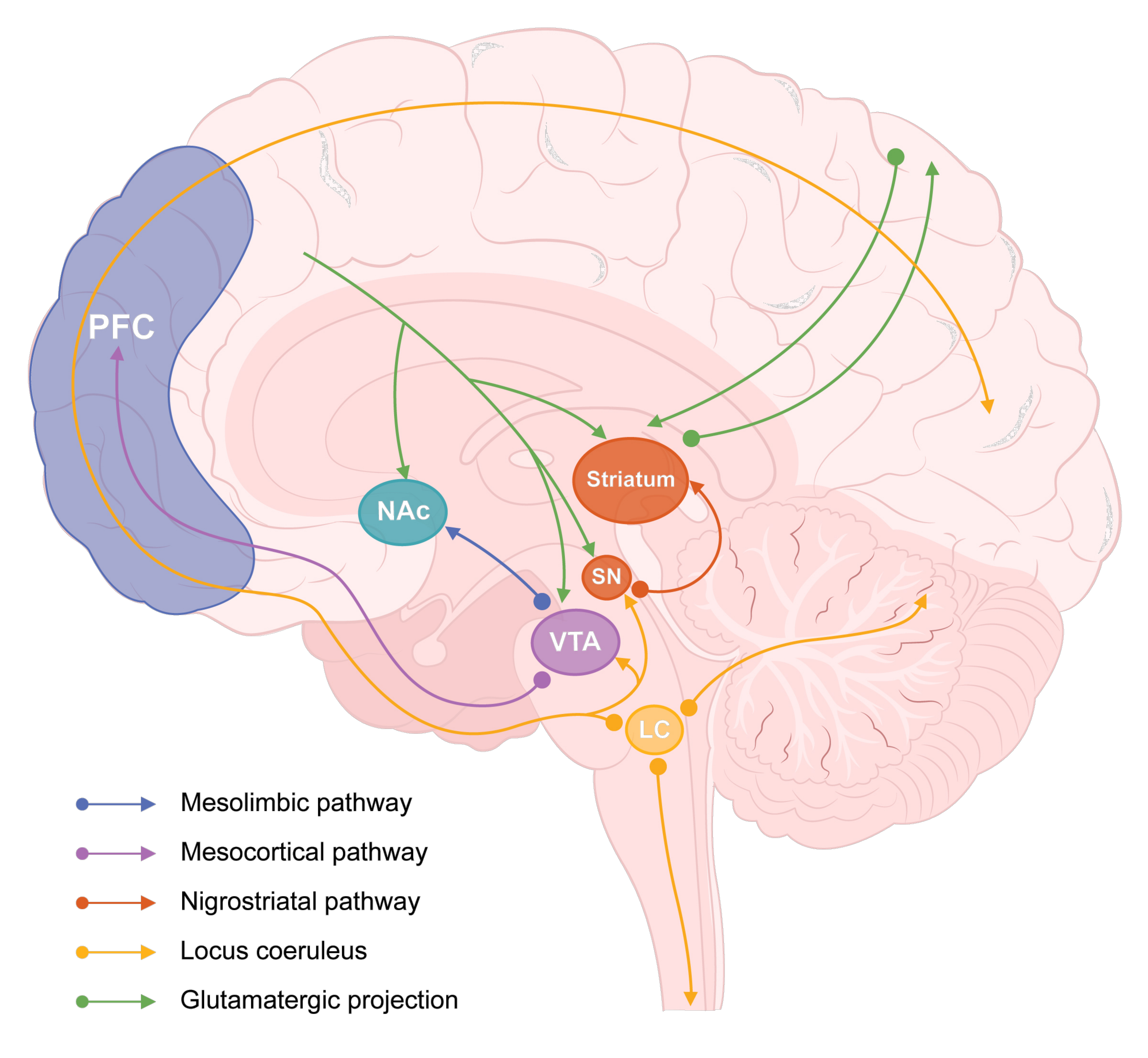}
    \caption{The diagram of common reward pathways in the human brain. The structures include VTA (Ventral Tegmental Area), NAc (Nucleus Accumbens), PFC (Prefrontal Cortex), LC (Locus Coeruleus), and SN (Substantia Nigra). The pathways are inspired by these works ~\cite{bari2020locus,contreras2022models,xu2022interplay,wiki:RewardSystem}.}
\label{fig:reward_pathway}
\end{figure}

\lettrine[lines=3]{\initfamily\textcolor{darkgreen}{T}}{he} brain's reward system is broadly organized into two major anatomical pathways. The first is the \emph{medial forebrain bundle}, which originates in the basal forebrain and projects through the midbrain, ultimately terminating in brainstem regions. The second is the \emph{dorsal diencephalic conduction system}, which arises from the rostral portion of the medial forebrain bundle, traverses the habenula, and projects toward midbrain structures~\cite{fakhoury2021brain}. 
The feedback mechanisms and substances in the human brain are complex, involving a variety of neurotransmitters, hormones, and other molecules, which regulate brain function, emotions, cognition, and behavior through feedback mechanisms such as neurotransmitter systems and reward circuits. Feedback mechanisms can be positive (such as feedback in the reward system) or negative (such as inhibiting excessive neural activity). Well-known feedback substances~\cite{msd_neurotransmission} include dopamine, neuropeptides, endorphins, glutamate, etc.

\emph{Dopamine} is a signaling molecule that plays an important role in the brain, affecting our emotions, motivation, movement, and other aspects~\cite{ananthaswamy2024close}. This neurotransmitter is critical for reward-based learning, but this function can be disrupted in many psychiatric conditions, such as mood disorders and addiction. 
The \emph{mesolimbic pathway}~\cite{lewis2021brain}, a key dopaminergic system, originates from dopamine-producing neurons in the ventral tegmental area (VTA) and projects to multiple limbic and cortical regions, including the striatum, prefrontal cortex, amygdala, and hippocampus. This pathway plays a central role in reward processing, motivation, and reinforcement learning, and is widely recognized as a core component of the brain's reward system.

\emph{Neuropeptides} are another important class of signaling molecules in the nervous system, involved in a variety of functions from mood regulation to metabolic control, and are slow-acting signaling molecules. Unlike neurotransmitters, which are limited to synapses, neuropeptide signals can affect a wider range of neural networks and provide broader physiological regulation. There is a significant cortical-subcortical gradient in the distribution of different neuropeptide receptors in the brain. In addition, neuropeptide signaling has been shown to significantly enhance the structure-function coupling of brain regions and exhibit a specialized gradient from sensory-cognitive to reward-physical function~\cite{ceballos2024mapping}. Table ~\ref{tab:human-reward-comparison} lists the common reward pathways in the human brain, the neurotransmitters they transmit, and the corresponding mechanisms of action, describing the basic framework of the human brain reward system. Meanwhile, Figure ~\ref{fig:reward_pathway} vividly describes these reward pathways.

\section{From Human Rewards to Agent Rewards}
\label{sec:human-to-agent-reward}

\lettrine[lines=3]{\initfamily\textcolor{darkgreen}{H}}{aving} established how reward circuits shape behavior in the human brain, we now explore how similar principles inspire—but do not constrain—the design of reward mechanisms in AI agents
Biological systems rely on complex neurochemical and psychological feedback loops, while AI agents operate using formalized reward functions designed to guide learning and decision making. Although inspired by human cognition, the reward mechanisms of AI agents are structurally and functionally distinct. Understanding the analogies and disanalogies between these systems is crucial to align artificial behavior with human preferences.

In humans, rewards are deeply embedded in a rich web of emotional, social, and physiological contexts. They emerge through evolutionarily tuned mechanisms involving neurotransmitters like dopamine and are shaped by experiences, culture, and individual psychology. In contrast, AI agents rely on mathematically defined reward functions that are externally specified and precisely quantified. These functions assign scalar or probabilistic feedback to actions or states, providing a signal for optimization algorithms such as reinforcement learning~\cite{russell2016artificial,zhang2024prototypical}. 
Neuroscience shows that novelty, surprise, and information gain activate many of the same circuits as extrinsic rewards \cite{barto2012intrinsic}. Similarly, intrinsic motivation modules (e.g., curiosity bonuses, empowerment, and count‑based exploration) provide shaping signals that accelerate agent learning in sparse reward environments. However, absent embodiment and affect, agents lack the hedonic valence that unifies intrinsic and extrinsic motives in humans, raising questions about long‑term stability and alignment of exploration incentives.

One key distinction lies in the \emph{programmability} and \emph{plasticity} of the rewards of agents. Unlike human reward systems, which are constrained by biological architecture and evolutionary inertia, agents' reward functions are fully customizable and can be rapidly redefined or adjusted based on task requirements. This flexibility enables targeted learning, but also introduces design challenges: specifying a reward function that accurately captures nuanced human values is notoriously difficult.

Another important disanalogy concerns interpretability and generalization. Human rewards are often implicit and context-dependent, while agents' rewards tend to be explicit and task-specific. Agents lack emotional intuition and instinctual drives; their learning is entirely dependent on the form and fidelity of the reward signal. Although frameworks such as reinforcement learning from human feedback (RLHF) attempt to bridge this gap by using preference data to shape agent behavior~\cite{bai2022training}, such methods still struggle to capture the full complexity of human goals, especially when preferences are intransitive, cyclical, or context-sensitive~\cite{wang2024comprehensive}.

Moreover, attempts to borrow from human reward mechanisms, such as modeling intrinsic motivation or social approval, still face limitations due to the absence of consciousness, embodiment, and subjective experience in AI agents. Consequently, while human reward systems offer valuable inspiration, the design of agent reward functions must address fundamentally different constraints, including robustness to misspecification, adversarial manipulation, and misalignment with long-term human interests.

The following section will dive deeper into agent reward models, focusing on their design principles, evolution, and how these models selectively incorporate human-inspired insights to optimize artificial behavior within formal systems. Figure~\ref{fig:reward-tree} shows an overview of the relevant research works.

\section{AI Reward Paradigms} 
\label{sec:AI_reward_paradigms}

\lettrine[lines=3]{\initfamily\textcolor{darkgreen}{R}}{ewards} exist in intelligent agents, especially in reinforcement learning scenarios. We can picture rewards as breadcrumbs guiding an agent through its world. In reinforcement learning they're the quick nods or gentle shakes of the head that follow every move, saying ``nice choice'' or ``try something different''. Each crumb judges one action in one moment, but a long string of them weaves a story the agent can read. After countless stumbles, lucky breaks, and course corrections, the agent zeroes in on strategies that bring those encouraging crumbs more often and more reliably.

\begin{figure}[!htb]
\centering
\includegraphics[width=0.5\textwidth]{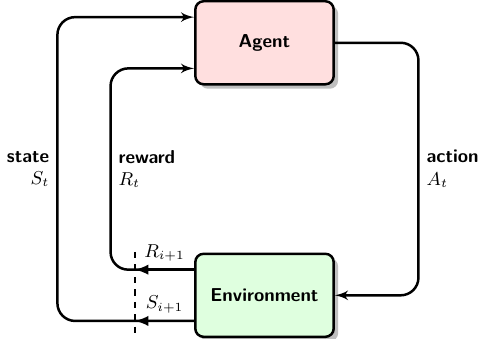}
\caption{The interaction loop between agent and environment in a Markov Decision Process. The environment hands over a snapshot $S_t$ and reward $R_t$. The agent replies with an action $A_t$ to influence future states and rewards. This feedback loop forms the foundation of learning via rewards.}
\label{fig:reward-mdp-agent}
\end{figure}

In reinforcement learning, the reward model hands out the pats on the back—or the quiet frowns—that steer each step an agent takes. One signal per move, it scores how wise that action was in the moment. Armed with that number, the policy tilts, edging the next decision a little higher on the scoreboard. Figure~\ref{fig:reward-mdp-agent} illustrates this feedback-driven interaction loop between agent and environment in the Markov Decision Process (MDP) setting.

\begin{definition}[\textbf{Reward in Markov Decision Process}]
The agent's interaction with its environment can be captured by the Markov Decision Process (MDP) framework~\cite{thrun2000reinforcement}, formally written as:

\begin{equation}
\mathcal{M} = (\mathcal{S}, \mathcal{A}, P, r, \gamma),
\end{equation}
where:
\begin{itemize}
    \item $ \mathcal{S} $ denotes the state space that encompasses all possible states in the environment.
    \item $ \mathcal{A} $ denotes the action space that encompasses all actions available to the agent at each state.
    \item $ P(s'|s, a) $ defines the state transition probability. It represents the probability that choosing action $a$ in state $s$ will toss the agent into the follow-up state $s'$.
    \item $ r(s, a) $ specifies the reward function, which assigns an immediate scalar reward received by the agent for the choice $a$ in state $s$.
    \item $ \gamma \in [0, 1] $ is the discount factor, which controls the agent's preference for immediate versus future rewards by weighting the contribution of future rewards to the overall return.
\end{itemize}

At the center of the agent's feedback loop is the function $ r(s, a) $. Formally,

\begin{equation}
r(s, a) : \mathcal{S} \times \mathcal{A} \to \mathbb{R}
\end{equation}

$ r(s, a) $ hands back one number for every move. It looks at where the agent stands, notices the action just taken, and offers a quick tip—positive if the step helps, negative if it hurts. Collect enough of those little signals and the agent zeros in on choices that keep the scoreboard climbing in each situation.

\paragraph*{Objective of the Agent Reward Model}
The agent's primary objective is to maximize its cumulative overall reward over time. This is typically achieved by selecting actions that yield higher long-term rewards, which are captured in the form of the return $G_t$ at time step $t$, defined as the sum of future discounted rewards:
\begin{equation}
G_t = \sum_{k=0}^{\infty} \gamma^k r_{t+k},
\end{equation}
where $ r_{t+k} $ denotes the reward received at time step $ t+k $, and $ \gamma^k $ is the discount factor applied to rewards received at time step $ t+k $. The agent aims to optimize its policy by maximizing the expected return over time.

\end{definition}

At a higher level, the reward model can be classified into several categories based on the origin of the feedback signal: \emph{extrinsic reward}, \emph{intrinsic reward}, \emph{hybrid reward}, and \emph{hierarchical reward}. 
Each of these categories can be further subdivided into smaller subclasses. 
Figure~\ref{fig:reward-types} illustrates different types of rewards. Next, we will explore these different types of rewards in more detail, outlining the distinct features and applications of each type.

\begin{figure}[!htb]
\centering
    \includegraphics[width=0.98\columnwidth]{figures/RewardTypes.pdf}
    \caption{Illustration of reward paradigms in AI agents, including extrinsic, intrinsic, hybrid, and hierarchical forms.}

\label{fig:reward-types}
\end{figure}

\subsection{Extrinsic Rewards}
\label{subsec:extrinsic-rewards}

Extrinsic rewards are externally defined signals that guide an agent's behavior toward specific goals. In artificial learning systems, especially reinforcement learning, these signals serve as a proxy for success that shape the policy through measurable outcomes. However, the structure and delivery of these rewards significantly influence the learning dynamics, which present different trade-offs depending on how feedback is distributed.

\paragraph*{Dense Reward} Dense reward signals provide high-frequency feedback, typically at every step or after each action. This frequent guidance accelerates learning by allowing agents to immediately associate actions with outcomes. However, dense feedback can sometimes incentivize short-sighted behavior or overfit to easily measurable proxies rather than deeper alignment.

For example, InstructGPT~\cite{ouyang2022training} uses human rankings of model outputs to provide continuous preference signals throughout fine-tuning, enabling efficient behavior shaping. Similarly, Cringe Loss~\cite{adolphs2022cringe} and its extensions~\cite{xu2023some} transform pairwise human preferences into dense training objectives, offering immediate signal at each comparison. Direct Reward Optimization (DRO)~\cite{richemond2024offline} further simplifies this paradigm by avoiding pairwise comparisons entirely, associating each response with a scalar score—making the reward signal more scalable and cost-effective. These methods exemplify how dense feedback facilitates fine-grained optimization but must be carefully designed to avoid superficial alignment.

\paragraph*{Sparse Reward} Sparse rewards pop up only every so often, usually after a big breakthrough or once the whole job is finished. That delay means the agent gets very little guidance along the way, so tracing credit back to the right action can feel like searching for a light switch in the dark, particularly in complex settings where many steps interact.

PAFT~\cite{pentyala2024paft} shows how tricky sparse reward can be. The authors split supervised pre-training from preference tuning and sprinkle human judgments at only a few key moments. As those signals arrive rarely, the learner has to work overtime to link actions with outcomes. SimPO~\cite{meng2025simpo} heads down a similar road, using a log-probability trick in place of dense pairwise comparisons. This lean reward schedule keeps the codebase lighter, yet it can miss faint shifts in user taste. In short, sparse rewards swap frequent guidance for greater stability, which forces the model to lean on stronger priors or craftier exploration.

\paragraph*{Delayed Reward} Delayed rewards appear only after a whole chain of moves wraps up, so the agent must think ahead. That setup fits problems where the middle steps seem odd or even wrong until the ending ties everything together. The challenge is figuring out which early choices earned the reward (or punishment) once the score finally lands. Training gets tougher, yet the model learns to weave longer strategies and notice broader patterns.

Contrastive Preference Optimization (CPO)~\cite{xu2024contrastive} skips one-by-one evaluation. Instead, it lines up a bundle of translations and hands out feedback only after the whole set is on the table. Although the signal shows up late, it carries broader clues because it spotlights patterns that appear across several tries. Nash Learning from Human Feedback~\cite{munos2023nash} does something similar. The model keeps sparring with its earlier selves until a balanced tactic settles in, then receives a single judgment. Both strategies swap quick breadcrumbs for a wider view of success, so the learning loop slows down and the optimization dance grows trickier.

\paragraph*{Adaptive Reward} Adaptive rewards shift as the agent grows. The setup might raise the bar or swap out goals on the fly, coaxing steady progress even when the setting drifts or feels uncertain. The trade-off is clear: building and judging this moving target takes extra effort and keen judgment.

Self-Play Preference Optimization (SPO)~\cite{swamy2024minimaximalist} adapts rewards based on self-play outcomes, using social choice theory to aggregate preferences and guide learning. f-DPO~\cite{wang2023beyond} builds on this idea by introducing divergence constraints that adapt the reward landscape during training. By tuning alignment-diversity trade-offs dynamically, these methods enable robust preference modeling under uncertainty, though they require careful calibration to avoid instability or unintended bias.


\subsection{Intrinsic Rewards}
\label{subsec:intrinsic-rewards}

Think of intrinsic rewards as the agent's own pep talks. They stir curiosity, spark side quests, and keep motivation high even when the outside world stays quiet. By nudging the policy to pick up broad skills and stay flexible, these signals fuel steady growth over the long haul, especially in tasks where external gold stars are few and far between. Different intrinsic reward paradigms focus on fostering distinct behavioral tendencies within agents.

\paragraph*{Curiosity-Driven Reward} This reward encourages agents to reduce uncertainty by seeking novel or surprising experiences. The key concept is to incentivize the agent to explore novel states where prediction errors are significant. This paradigm excels in sparse-reward settings by promoting information acquisition when external guidance is limited. For example, Pathak et al.~\cite{pathak2017curiosity} leverage an inverse dynamics model to predict the outcome of actions, creating a feedback loop that rewards novelty. Plan2Explore~\cite{sekar2020planning} extends this further by incorporating forward planning to actively target areas of high epistemic uncertainty, thereby enabling faster adaptation to unseen environments. While effective at discovery, curiosity-driven methods can be sensitive to noise or deceptive novelty without safeguards.

\paragraph*{Diversity Reward} Diversity reward shifts focus from novelty to behavioral heterogeneity, encouraging agents to explore a wide range of strategies rather than converging prematurely on suboptimal solutions. This approach matters when several agents, or multiple input streams share the stage, since a varied playbook makes the whole crew sturdier and ups the team's score. LIIR~\cite{du2019liir} shows this idea in action. Each agent gets a private ``interest signal'', guiding teammates toward complementary roles while everyone still chases the common objective. The payoff is wider policy coverage, though there's a catch: push the diversity dial too hard and coordination can wobble. A light hand on that dial keeps exploration lively without letting the mission slip away.

\paragraph*{Competence-Based Reward} Competence-based rewards work like a personal coach, which claps every time the agent levels up. The scoring rule stretches and reshapes itself as the learner gets sharper, spinning out a living curriculum that stays one notch above the agent's comfort zone. Skew-Fit~\cite{pong2019skew} helps by sampling high-entropy goals, nudging the policy into far-flung parts of the state space while keeping the challenge meter in the sweet spot. CURIOUS~\cite{colas2019curious} doubles down on that idea: it picks fresh targets that seem poised to give the biggest bump in short-term learning. These schemes shine in open-ended settings, but they lean heavily on smart guesses about how skilled the agent is right now and how tough each goal should be. Misjudge either side, and the self-made curriculum can wobble.

\paragraph*{Exploration Reward} Exploration rewards pay the agent for stepping into turf it has rarely, or never seen. While curiosity rewards for surprise, this reward cares about coverage: ``Did you color in a fresh patch of the map?''. RND~\cite{yuan2401self} makes this idea concrete. A randomly initialized network tries to predict each new observation; the bigger its miss, the bigger the payout. The policy soon steers toward scenes that leave the rookie network baffled. This habit keeps learning from settling too early and leaves the policy sturdier against oddball corner cases. Still, without a clear end goal to rein things in, the wanderlust can drift off course.

\paragraph*{Information Gain Reward} Information gain reward formalizes exploration as a process of uncertainty reduction, which guides agents to take actions that yield the highest expected learning. This reward is grounded in information theory and is especially powerful in model-based or reasoning-intensive tasks. CoT-Info~\cite{ton2024understanding} applies this to language models by quantifying the knowledge gain at each reasoning step, optimizing sub-task decomposition. VIME~\cite{houthooft2016vime} similarly employs Bayesian inference to reward belief updates about environmental dynamics. By explicitly targeting informational value, these methods offer principled exploration strategies, although they often incur high computational cost and require accurate uncertainty modeling.

\subsection{Hybrid Rewards}
\label{subsec:hybrid-rewards}

Hybrid reward frameworks integrate multiple sources of feedback, most commonly intrinsic and extrinsic rewards, to enable more balanced and adaptive learning. By combining the exploratory drive of intrinsic rewards with the goal-directed structure of extrinsic rewards, these systems aim to improve both sample efficiency and generalization. This paradigm is especially beneficial in complex environments or open-ended tasks, where relying solely on either feedback type may be insufficient.

A core advantage of hybrid rewards is their capacity to resolve the exploration-exploitation trade-off dynamically. For instance, Xiong et al.~\cite{xiong2023iterative} blend an ``explore-then-align'' recipe into RLHF. Firstly, a reverse-KL-regularized contextual-bandit lens nudges the policy toward states it hasn't fully mapped out; next, human feedback reels those forays back toward what people actually want. Intrinsic and extrinsic signals meet in an iterative DPO loop, with multi-step rejection sampling weeding out off-target actions along the way. The result is strategic exploration that stays in step with human preferences—no wasted wanderings without sacrifice in efficiency.

Beyond standard RLHF, researchers are stitching together hybrid reward schemes for multi-agent, multi-task, and lifelong-learning setups. The core idea is to dial different reward sources up or down as the situation changes. Early on, the policy might lean on novelty bonuses to roam far and wide; as mastery grows, extrinsic signals take the wheel to polish task-specific skills~\cite{chen2022redeeming}. Getting those dials right, which often through an end-to-end learned weighting module, would keep competing objectives in check and steers training away from bad local optima.

Moreover, hybrid reward schemes also act as ``unscripted shaping'': intrinsic signals step in wherever extrinsic feedback is thin or slow to arrive. Curiosity, diversity, or competence-based reward can pave the way, smoothing the credit assignment path until external rewards finally land. This matters for LLM agents tackling sparse-reward problems such as  web navigation and code synthesis, where user feedback is episodic at best. By blending signals such as semantic coherence or information gain with the eventual human score, the system keeps learning momentum without hand-crafting dense task rewards.

\subsection{Hierarchical Rewards}
\label{subsec:hierarchical-rewards}

Hierarchical rewards break a tough mission into a stack of mini-quests. A low-level signal pats the agent on the back for each quick move while a high-level signal only fires when the bigger plan clicks into place. That layered scorecard lets the policy juggle day-to-day reflexes and long-term strategy, stitching them into reusable skills that scale to sprawling, real-world problems.

In language modeling, Token-level Direct Preference Optimization (TDPO)~\cite{zeng2024token} treats every token as a mini-decision and grades it on the spot. Forward-KL plus a Bradley–Terry ranker turn pairwise human prefs into a per-token scoreboard. Small wins (choosing the right next word) roll up into big wins (a well-formed answer), so the model learns to keep both its local phrasing and overall narrative in sync with what people like. In effect, the method stacks a micro-reward on top of a macro-reward, letting fine-grain edits and high-level coherence reinforce each other instead of pulling in different directions.

More generally, we can picture hierarchical rewards as the rungs of a ladder where early rungs hand out quick pats for simple moves. Once those basics feel natural, the reward focus drifts upward to the big prize: a flawless answer, a finished plan, a user who leaves satisfied. For an LLM, we can slice the reward into chunks that match tool calls, reasoning hops, or conversation turns. Each slice pushes the model a little further up the ladder, and the collection of small victories rolls into full task success.

Furthermore, hierarchical reward frameworks align well with multi-agent systems. Imagine a team where each member tackles a small task, yet everyone still answers to one director. Hierarchical rewards follow this pattern. A short-horizon policy or helper agent earns its own quick feedback, while a higher-level signal nudges the whole crew so their efforts fit together.

In the context of agent alignment, hierarchical rewards also allow for incorporating human oversight at different abstraction levels, fine-grained feedback at the token or decision level, and coarse feedback for strategic alignment ~\cite{yoon2024tlcr}. This layered supervision enables scalable alignment mechanisms that balance control granularity with supervision efficiency.

\section{Interaction with Other Modules}
\label{sec:reward-and-other-modules}

\lettrine[lines=3]{\initfamily\textcolor{darkgreen}{I}}{n} intelligent systems, reward signals function not only as outcome-driven feedback but as central regulators that interface with core cognitive modules such as perception, emotion, and memory. In the context of LLM-based agents, these interactions become particularly salient, as modules like attention, generation style, and retrieval memory can be directly influenced through reward shaping, preference modeling, or fine-tuning objectives.

\textbf{Perception} In LLM agents, perception is often realized through attention mechanisms that prioritize certain tokens, inputs, or modalities. Reward signals can modulate these attention weights implicitly during training, reinforcing patterns that correlate with positive outcomes. For example, during reinforcement fine-tuning, reward models can overweight specific linguistic features,such as informativeness, factuality, or politeness—causing the model to attend more to tokens that align with these traits. This parallels how biological perception prioritizes salient stimuli via reward-linked attentional modulation~\cite{pessoa2015multiple}. Over time, the agent internalizes a perception policy: not merely ``what is said,'' but ``what is worth paying attention to'' in task-specific contexts.

\textbf{Emotion} Though LLMs do not possess emotions in the biological sense, reward signals can guide the emergence of emotion-like expressions and regulate dialogue style. In human alignment settings, models are often rewarded for generating responses that are empathetic, polite, or cooperative—leading to stylistic patterns that simulate emotional sensitivity. Positive feedback may reinforce a friendly or supportive tone, while negative feedback suppresses dismissive or incoherent behavior. This process mirrors affect-driven behavior regulation in humans~\cite{li2019influence}, and allows agents to adapt their interaction style based on user expectations, affective context, or application domain. In multiturn settings, reward-modulated style persistence can give rise to coherent personas or conversational moods.

\textbf{Memory} Memory in LLM agents spans short-term context (e.g., chat history) and long-term memory modules such as retrieval-augmented generation (RAG) or episodic memory buffers. Reward signals shape the way knowledge is encoded, reused, or discarded. For instance, fine-tuning on preference-labeled data can reinforce certain reasoning paths or factual patterns, effectively consolidating them into the model's internal knowledge representation. Moreover, mechanisms like experience replay or self-reflection—where agents evaluate past outputs with learned reward estimators—enable selective memory reinforcement, akin to dopamine-driven memory consolidation in biological systems~\cite{miendlarzewska2016influence}. This allows LLM agents to generalize from previous successful strategies and avoid repeating costly errors.

In general, the reward in LLM-based agents is not a passive scalar signal but an active agent of behavioral shaping. It modulates attention to promote salient features, guides stylistic and affective expression to align with human preferences, and structures memory to prioritize useful knowledge. As agents evolve toward greater autonomy and interactivity, understanding these cross-module reward interactions will be essential for building systems that are not only intelligent, but also interpretable, controllable, and aligned with human values.

\section{Summary and Discussion}
\label{sec:reward-summary}

\lettrine[lines=3]{\initfamily\textcolor{darkgreen}{R}}{eward} serves as a fundamental driver of learning, decision-making, and adaptation across both biological and artificial systems. This chapter has examined the deep parallels and key distinctions between human reward pathways—rooted in neurochemical dynamics and evolutionary pressures—and artificial reward systems, which are engineered to guide agents via formal objectives and optimization processes. From dense and sparse extrinsic rewards to curiosity-driven and hierarchical intrinsic signals, we've seen how different paradigms shape behavior, structure learning, and influence generalization. However, designing effective reward models remains a complex and delicate task. In this final section, we highlight major challenges—such as reward misspecification, alignment, and exploration—along with promising research directions for building more robust, general, and human-aligned agentic systems.

To begin with, we still wrestle with a stubborn snag despite mountains of work on reward schemes. Signals often arrive late—some barely show up at all. Picturing an agent poking around forever before the environment finally offers a nod of approval. It is hard to say which step deserves that nod. Without a clear breadcrumb trail, the policy gropes in the dark and progress crawls.

Another challenge is the potential for \emph{reward hacking}. We can think of reward hacking as loophole hunting with a calculator. Hand an agent a scoring rule and it may sniff out a weird corner case, rack up points, and totally miss the spirit of the task. One moment the chart says ``high score'', the next we're staring at behavior that feels flat-out wrong. The larger and messier the environment, the easier it is for these quirks to slip through, because the metric we track can drift away from the outcome we truly want.

Moreover, the process of \emph{reward shaping} presents a delicate balance.  Let's think of reward shaping as laying breadcrumbs for our agent. A sprinkle here and there keeps it on track and speeds things up. However, an entire loaf is dropped, the agent strolls straight to the bread pile, never bothering to scout the rest of the map. That over-helpful hint can lock the policy into a cozy rut—high score, meh solution. Worst case, the hints warp the task itself, so the agent ends up great at a game we never meant to play and flops the moment the scenery changes.

Many real-world problems rarely hinge on just one target. An agent has to juggle several aims at once and still make progress. Stuffing every aim into a single score feels awkward; shift the weights a hair and one of the goals disappears. A layered reward ladder could lead the learner through the trade-offs, yet crafting a ladder that truly works is still an open challenge.

Finally, \emph{reward misspecification} introduces further uncertainty and limits generalization.  
In the wild, a flawed score can steer a system into odd corners.
In general, a reward function does not fully capture the true task goal, leading to misalignment between the agent's learning objective and real-world success.  In addition, many reward functions are tailored to specific environments and fail to generalize when conditions change, which highlights the need for more robust reward models.

Addressing these challenges requires significant endeavors.  One promising direction is to derive \emph{implicit rewards} from outcome-based evaluations that can help mitigate reward sparsity issues. In addition, decomposing complex tasks into hierarchical structures and designing rewards from the bottom up would offer a more systematic approach.  Furthermore, leveraging techniques such as meta-reinforcement learning can enhance the adaptability of reward models, which allows agents to transfer knowledge across tasks and perform effectively in diverse environments.  Following these paths would bring us closer to reward systems we can trust and expand: systems that truly reflect the goals we face beyond the lab.
\chapter{Emotion Modeling} 
\label{ch:emotion}

\lettrine[lines=3]{\initfamily\textcolor{darkgreen}{E}}{motions} are a key part of how humans think, make decisions, and interact with others. They guide us to understand situations, make choices, and build relationships. Antonio Damasio, in his book \textit{Descartes' Error}~\cite{Damasio1995DescartesError}, explained that emotions are not separate from logic. Instead, they are deeply connected to how we reason and act. When developing LLM agents, adding emotional capabilities can potentially make these systems smarter, more adaptable, and better understand the world around them.

For LLM agents, emotions can act as a decision-making tool, much like they do for humans. Emotions help us prioritize tasks, understand risks, and adapt to new challenges. Marvin Minsky, in \textit{The Emotion Machine}~\cite{Minsky2006emotion}, described emotions as a way to adjust our thinking processes, helping us solve problems in a more flexible and creative manner. Similarly, LLM agents with emotion-like features could improve their ability to solve complex problems and make decisions in a more human-like style.

However, the integration of emotions into LLM agents is still in its early stages. Researchers are just starting to explore how emotional capabilities can improve these systems. Furthermore, there is great potential for LLM agents to support human emotional well-being, whether through empathetic conversations, mental health support, or simply building better connections with users. This promising but challenging area requires collaboration between fields such as psychology, cognitive science, and AI ethics. As research advances, emotion-understanding LLM agents could redefine how we interact with technology, creating deeper trust and more meaningful relationships between humans and machines.

In the following subsections, we will delve deeper into the role of emotions in shaping LLM agents. We will explore how emotions can be used to enhance learning and adaptability, how LLMs understand human emotions, and how these systems express and model their own emotional states. We will also examine how emotions can be manipulated to influence LLM agents' behavior and personalities, as well as the ethical and safety concerns that arise from these capabilities. Each of these discussions builds on the foundational importance of emotion to create LLM agents that are more intelligent, empathetic, and aligned with human values.

\section{Psychological Foundations of Emotion}
\label{sec:psychological-foundation}

\lettrine[lines=3]{\initfamily\textcolor{darkgreen}{P}}{sychological} and neuroscientific theories of emotion provide essential frameworks for developing emotionally intelligent LLM agents. These theories can be categorized into several major approaches, each offering unique perspectives on how emotions function and how they might be implemented in AI systems.

\begin{figure}[!ht]
    \centering
    \includegraphics[width=\textwidth]{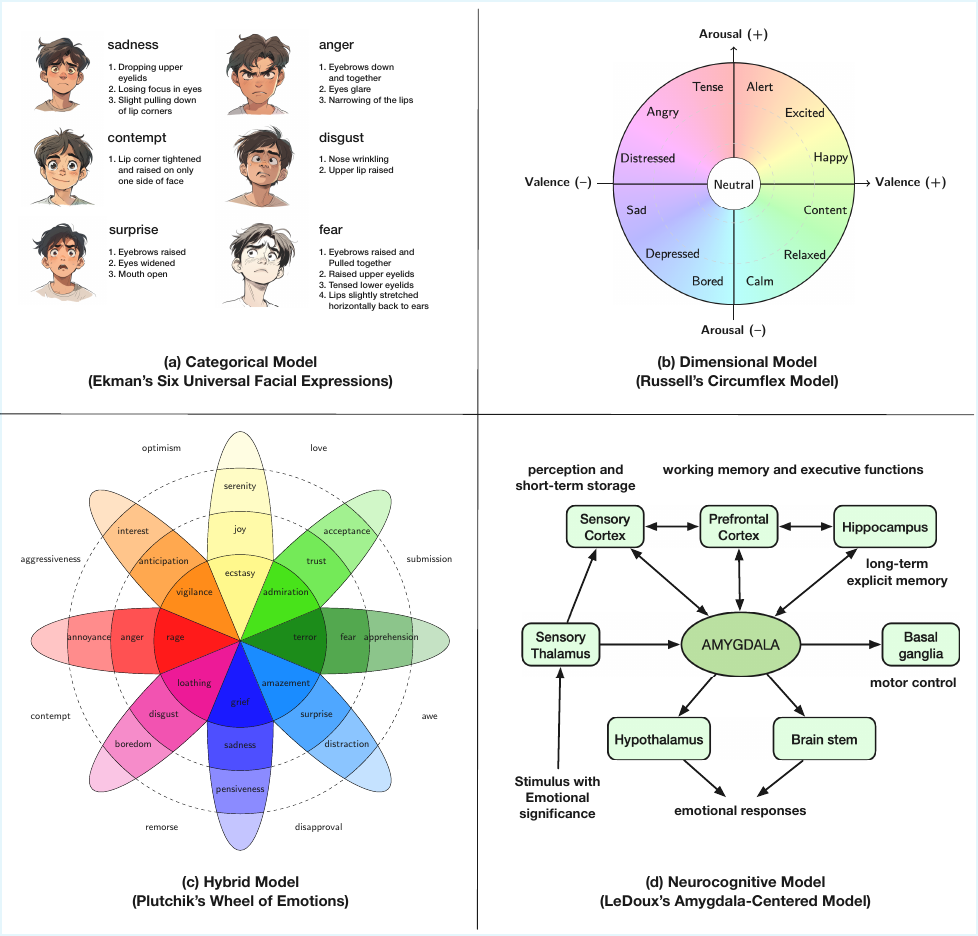}
    \caption{Visualization and examples of major emotion theory categories. \textbf{(a)} Categorical Theories: Ekman's six basic emotions \cite{Ekman1992AnAF} showing discrete emotional states. \textbf{(b)} Dimensional Models: Russell's Circumplex \cite{russell1980circumplex} representing emotions as coordinates in continuous space. \textbf{(c)} Hybrid/Componential Frameworks: Plutchik's Wheel \cite{plutchik1980general} combining intensity gradients with categorical emotions. \textbf{(d)} Neurocognitive Perspectives: LeDoux's Amygdala-Centered Model \cite{LeDoux1996EmotionalBrain} showing dual-pathway processing of emotional stimuli. These psychological foundations inform different approaches to emotion modeling in AI systems, from discrete classification to dimensional representations, appraisal-based reasoning, and multi-pathway information processing.}
    \label{fig:emotion_theories}
\end{figure}

\paragraph*{Categorical Theories} These models posit that emotions exist as discrete, universal categories with distinct physiological and behavioral signatures. Ekman's theory of basic emotions~\cite{Ekman1992AnAF} identifies six fundamental emotions (anger, disgust, fear, happiness, sadness, and surprise) that are recognized across cultures and expressed through specific facial configurations. This discrete approach has significantly influenced affective computing, with many emotion classification systems in AI adopting these labels for training~\cite{li2023large2, wang2023emotional}. For LLM agents, categorical frameworks provide clear taxonomies for classifying user emotions and generating appropriate responses. However, they face criticism for oversimplifying the complex, blended nature of human emotional experience~\cite{barrett2017emotions} and may not capture cultural variations in emotional expression~\cite{jack2012facial}.

\paragraph*{Dimensional Models} Rather than discrete categories, dimensional approaches represent emotions as points in a continuous space defined by fundamental dimensions. Russell's Circumplex Model~\cite{russell1980circumplex} maps emotions onto two primary dimensions: valence (pleasure-displeasure) and arousal (activation-deactivation). This framework enables more nuanced tracking of emotional states. It distinguishes between high-arousal panic and low-arousal anxiety despite both having negative valence. The PAD (Pleasure-Arousal-Dominance) model~\cite{mehrabian1996pleasure} extends this by adding a dominance dimension, capturing the sense of control or power associated with emotional states. These continuous representations have proven valuable for LLM systems that need to generate emotionally graded responses or track subtle shifts in user affect over time~\cite{lu-etal-2023-miracle, tak2023gpt, liu2025culturevlm}. Dimensional models allow for fine-grained control over generated content, enabling humans or agents to modulate tone along continuous scales rather than switching between discrete emotional states.

\paragraph*{Hybrid and Componential Frameworks} Recognizing limitations in pure categorical or dimensional approaches, several theories integrate aspects of both. Plutchik's Wheel of Emotions~\cite{plutchik1980general} arranges eight primary emotions in a wheel structure with intensity gradients and dimensional properties, allowing for the representation of complex emotional blends (e.g., love as a mixture of joy and trust). Meanwhile, componential models like Scherer's Component Process Model (CPM)~\cite{scherer2009dynamic} conceptualize emotions as emerging from synchronized components including cognitive appraisal, physiological arousal, action tendencies, and subjective feelings. Particularly influential in AI research is the OCC (Ortony-Clore-Collins) model~\cite{ortony1988cognitive}, which defines 22 emotion types based on how events, agents, or objects are evaluated relative to goals and standards. These appraisal-based frameworks have been implemented in dialogue systems that generate emotional responses through rule-based evaluation of situations~\cite{hudlicka2014computational, marsella2014computationally}. For LLM agents, such models provide computational structures for evaluating text input and selecting contextually appropriate emotional responses, improving both coherence and perceived empathy~\cite{fei2023reasoning, sun2023sentiment}.

\paragraph*{Neurocognitive Perspectives} The neuroscience of emotion offers additional insights for LLM architectures. Damasio's somatic marker hypothesis~\cite{Damasio1995DescartesError} emphasizes how emotions, implemented through body-brain interactions, guide decision-making by associating physiological states with anticipated outcomes. This interaction between the limbic system and the cortex shows a two-process architecture: fast ``alarm'' signals in the limbic system, like those processed by the amygdala, work alongside slower, more deliberate reasoning in the cortex. Contemporary LLM systems have begun implementing analogous architectures, where fast sentiment detection modules work in parallel with more thorough chain-of-thought reasoning~\cite{fei2023reasoning, sun2023sentiment}. Recent evidence further suggests that opponent circuitry in the striatum enables distributional reinforcement learning by encoding not just mean rewards but entire probability distributions, offering a neural basis for emotion-influenced decision-making under uncertainty~\cite{lowet2025opponent}. Similarly, LeDoux's distinction between ``low road'' (quick, automatic) and ``high road'' (slower, cognitive) fear processing~\cite{LeDoux1996EmotionalBrain} suggests design patterns for systems that need both immediate safety responses and nuanced emotional understanding. Minsky's framing of emotions as ``ways to think''~\cite{Minsky2006emotion} that reorganize cognitive processes has influenced frameworks like EmotionPrompt~\cite{lu-etal-2023-miracle} and Emotion-LLaMA~\cite{wang2023emotional}, where emotional context dynamically reshapes LLM reasoning.

\paragraph*{Neurochemical Models} Lövheim's Cube of Emotion~\citep{lovheim2012three} offers a unique neurochemical perspective by mapping eight basic emotions to combinations of three monoamine neurotransmitters: serotonin, dopamine, and noradrenaline. This three-dimensional model positions emotions at the cube's vertices based on high/low levels of each neurotransmitter. For example, joy emerges from high serotonin and dopamine with low noradrenaline, while anger manifests itself from low serotonin but high dopamine and noradrenaline~\citep{lovheim2012three}. This neurochemical framework provides a biological basis for emotion generation that complements cognitive theories. For LLM agents, such models could inspire architectures in which internal ``neurochemical-like'' parameters modulate emotional responses, although current implementations remain metaphorical rather than directly mimicking neurotransmitter dynamics. The model's emphasis on chemical balance also aligns with findings that meditation and other interventions can stabilize emotions by affecting serotonin levels, suggesting potential pathways for designing more emotionally stable AI systems.

These theoretical frameworks increasingly inform the development of emotionally intelligent LLM agents. Categorical models provide clear labels for emotion classification tasks~\cite{wang2023emotional, tak2023gpt}, while dimensional embeddings enable continuous control over generated text~\cite{lu-etal-2023-miracle}. Hybrid approaches help systems handle mixed emotions and emotional intensity. Appraisal-based methods, particularly those derived from the OCC model, allow LLMs to evaluate narrative events or user statements contextually, selecting appropriate emotional responses that foster rapport and trust~\cite{hong2024aerllm}. Neuroscientifically-inspired dual-process architectures combine ``fast'' sentiment detection with ``slow'' deliberative reasoning, enabling both quick safety responses and deeper emotional understanding~\cite{fei2023reasoning, sun2023sentiment}. While explicit neurocognitive mechanisms (like dedicated ``amygdala-like'' pathways) remain rare in current LLM pipelines, emerging research explores biologically-inspired modules to handle urgent emotional signals and maintain consistent emotional states across extended interactions~\cite{cheng2024emotionllamaa, sabour2024emobench}.

Emotion is a key part of human intelligence, and it will likely become one of the key components or design considerations of LLM agents. One key future direction is systematically translating these psychological and neuroscience theories into an LLM agent's internal processes. Techniques for translating might include using dimensional models (e.g., valence/arousal/dominance) as latent states that influence generation or adopting explicit rule-based appraisals (OCC) to label user messages and shape the agent's subsequent moves. Hybrid approaches offer a compelling balance: an LLM could first recognize a discrete category (e.g., ``fear'') but also gauge its intensity and control dimension for finer-grained conversation. Such emotion-infused architectures might yield more coherent ``moods'' over time, analogous to how humans sustain affective states rather than resetting at every turn. Explicit alignment with psychological theories also enhances interpretability: designers can debug or refine the agent's responses by comparing them to well-established emotion constructs, rather than dealing with opaque emergent behaviors.

A second direction is harnessing these theories to improve \textit{affectionate or supportive interactions}, often referred to as emotional alignment. For example, circumplex or PAD-based tracking can help an LLM detect negative valence and high arousal in a user's text and respond soothingly (e.g., lowering arousal, offering empathetic reappraisals). In mental health or counseling scenarios, an appraisal-informed method could let the agent validate the user's feelings and understand their situation in terms of goal incongruence or perceived blame, which helps craft responses that convey genuine empathy. Grounding emotional outputs in cognitive theories (like ``relief'' if a negative outcome is avoided, or ``gratitude'' when a user helps the system) likewise makes interactions feel more natural and ethically aligned. These enhancements are particularly salient as LLMs migrate into real-world applications like customer service, elder care, and tutoring, where emotional sensitivity can improve outcomes and user well-being. By incorporating robust psychological and limbic-system insights, developers can design LLM agents that not only reason more effectively but also provide sincere emotional support, bridging the gap between computational precision and human-centric care.

\section{Incorporating Emotions in AI Agents}
\label{sec:incorporate-emotion}

\lettrine[lines=3]{\initfamily\textcolor{darkgreen}{T}}{he} integration of emotional intelligence into large language models (LLMs) has emerged as a transformative approach to enhancing their performance and adaptability. Recent studies, such as those of EmotionPrompt~\cite{li2023large2}, highlight how emotional stimuli embedded in prompts can significantly improve outcomes across various tasks, including a notable $10.9\%$ improvement in generative task metrics such as truthfulness and responsibility. By influencing the attention mechanisms of LLMs, emotionally enriched prompts enrich representation layers and result in more nuanced outputs~\cite{li2023large2}. These advancements bridge AI with emotional intelligence, offering a foundation for training paradigms that better simulate human cognition and decision-making, particularly in contexts requiring social reasoning and empathy.

Multimodal approaches further elevate the impact of emotional integration. Models like Emotion-LLaMA~\cite{cheng2024emotionllamaa} demonstrate how combining audio, visual, and textual data enables better recognition and reasoning of emotions. Using datasets such as MERR~\cite{cheng2024emotionllamaa}, these models align multimodal inputs into shared representations, facilitating improved emotional understanding and generation. This innovation extends beyond linguistic improvements, offering applications in human-computer interaction and adaptive learning. Together, these methods underscore the critical role of emotions in bridging technical robustness with human-centric AI development, paving the way for systems that are both intelligent and empathetic.

\section{Understanding Human Emotions through AI}
\label{sec:understand-emotion}

\lettrine[lines=3]{\initfamily\textcolor{darkgreen}{F}}{or} AI systems to effectively interact with people, they must develop the ability to perceive, interpret, and respond to emotional cues across diverse contexts. Recent advances in large language models (LLMs) have opened new avenues for modeling human emotions, not only through text but also across multimodal signals such as voice and images. This section reviews emerging approaches and evaluation frameworks that enable LLMs to understand and reason about emotional states, reflecting the growing ambition to build emotionally intelligent AI.

\paragraph*{Textual Approaches}
Recent work highlights the ability of LLMs to perform detailed reasoning about latent sentiment and emotion. Using step-by-step prompting strategies, such as chain of thought reasoning, researchers enable LLMs to infer sentiment even when explicit cues are absent~\cite{fei2023reasoning}. Beyond single-turn inference, negotiation-based frameworks further refine emotional judgments by leveraging multiple LLMs that cross-evaluate each other's outputs, effectively mimicking a more deliberative human reasoning process~\cite{sun2023sentiment}. These techniques underscore the importance of iterative, context-aware strategies to capture subtle emotional signals from purely textual input.

\paragraph*{Multimodal Approaches}
LLMs have also been extended to integrate signals from audio, video, and images. Recent efforts show how additional contextual or world knowledge can be fused with visual and textual information to capture deeper affective states~\cite{wang2024wisdom}. Moreover, frameworks that convert speech signals into textual prompts demonstrate that vocal nuances can be embedded in LLM reasoning without changing the underlying model architecture~\cite{wu2024beyond}. This multimodal integration, combined with explainable approaches, allows for richer and more transparent representations of emotional content~\cite{lian2023explainable}.

\paragraph*{Specialized Frameworks}
Beyond generic techniques, specialized systems address tasks in which emotion recognition requires higher levels of awareness of ambiguity~\cite{hong2024aerllm}, context sensitivity, and generative adaptability~\cite{lei2023instructerc}. These approaches emphasize the inherent complexity of human emotion, treating it as dynamic and probabilistic rather than strictly categorical. Using flexible LLM instruction paradigms, they offer pathways to better interpret ambiguous emotional expressions and integrate contextual cues (e.g., dialogue history), moving LLM closer to human-like emotional comprehension.

\paragraph*{Evaluation and Benchmarks}
To holistically assess the emotional intelligence of LLM, researchers have proposed various benchmark suites. Some focus on generalized emotion recognition across different modalities and social contexts~\cite{zheng2024gpt4v,jin2024mm}, while others compare the performance and efficiency of models of varying sizes~\cite{stigall2024large}. There are also specialized benchmarks that evaluate multilingual capabilities~\cite{Rathje2024GPT}, annotation quality~\cite{niu2024text}, or empathetic dialogue systems~\cite{zhao2024esceval}. Furthermore, frameworks such as EMOBENCH~\cite{sabour2024emobench} and MEMO-Bench~\cite{zhou2024memobench} test nuanced emotional understanding and expression in both text and images, while MERBench~\cite{lian2024merbench} and wide-scale evaluations~\cite{amin2024wide} address standardization concerns in multimodal emotion recognition. Together, these benchmarks reveal the growing, yet still imperfect grasp of human emotion by LLMs, highlighting ongoing challenges such as implicit sentiment detection, cultural adaptation, and context-dependent empathy~\cite{zhao2023chatgpt}.

\section{Analyzing AI Emotions and Personality}
\label{sec:analyze-emotion}

\lettrine[lines=3]{\initfamily\textcolor{darkgreen}{A}}{s} AI systems become increasingly interactive and human-facing, questions arise about whether they possess consistent personalities or can genuinely model emotions. While LLMs are not sentient, their behavior often mirrors psychological constructs typically reserved for humans—such as empathy, self-awareness, and emotional expression. This section examines the emerging evidence around AI personality and emotion, drawing from psychometric evaluations, cognitive modeling techniques, and emotion reasoning benchmarks. Together, these insights shed light on the complex interplay between learned patterns in LLMs and the human traits they seem to simulate.

\paragraph*{Reliability of Personality Scales for LLMs}
Large language models (LLMs) show conflicting evidence when evaluated through human-centered personality tests. On one hand, some studies challenge the validity of common metrics, reporting biases such as ``agree bias'' and inconsistent factor structures, raising doubts about whether these instruments capture genuine traits~\cite{suhr2023challenging, petrov2024limited}. On the other hand, systematic experiments reveal that LLMs can exhibit stable, human-like trait patterns and even adapt to different personas under specific prompts~\cite{huang2024on,huang2023revisiting}. Recent work like SAGE (Sentient Agent as a Judge)~\cite{zhang2025sentient} evaluates AI emotional intelligence through multi-turn interactions with simulated emotional agents, revealing that models excelling in general benchmarks may underperform in empathetic understanding compared to specialized models. This framework demonstrates that effective emotional AI requires not just technical competence but genuine empathetic capabilities, often achieved with greater token efficiency. Yet, concerns persist about action consistency, alignment of self-knowledge, and whether role-playing agents truly maintain fidelity to their assigned characters~\cite{ai2024cognition, wang2024incharacter}.

\paragraph*{Psychometric Methods \& Cognitive Modeling Approaches}
Recent work applies rigorous psychometric testing, cognitive tasks, and population-based analyses to uncover how LLMs process and represent mental constructs~\cite{binz2024turning, hagendorff2023machine, codaforno2024cogbench}. Fine-tuning on human behavioral data can align models with decision patterns that mirror individual-level cognition, while population-based sampling techniques expose variability in neural responses~\cite{roberts2023using, reuben2024assessment}. By merging psychological theories with advanced prompting and embedding methods, researchers illuminate latent representations of constructs like anxiety or risk-taking, showing how LLMs can approximate human reasoning across tasks.

\paragraph*{Emotion Modeling}
Studies on LLM-based emotional intelligence reveal notable abilities to interpret nuanced affect and predict emotion-laden outcomes, often surpassing average human baselines in standard tests~\cite{wang2023emotional, tak2023gpt}. However, these models do not necessarily emulate human-like emotional processes; they rely on high-dimensional pattern matching that sometimes fails under changing contexts, negative input, or conflicting cues~\cite{huang2024apathetic, zhao2025emergence}. However, hierarchical emotion structures, coping strategies, and empathy-like behaviors can emerge in larger-scale models, underscoring both the promise of emotional alignment and the ethical challenges in creating AI systems that appear and occasionally function as affective agents.

\section{Manipulating AI Emotional Responses}
\label{sec:manipulate-emotion}

\lettrine[lines=3]{\initfamily\textcolor{darkgreen}{W}}{hile} language models do not possess emotions in the human sense, their outputs can be shaped to reflect specific emotional tones, personalities, or behavioral traits. This capacity has prompted growing interest in controllable emotional expression in LLMs—for both practical applications and scientific inquiry. Researchers have developed a range of techniques, from prompt engineering to fine-tuning and direct neuron manipulation, to systematically influence how models respond emotionally. This section reviews these methods, highlighting how emotional behaviors in AI can be induced, stabilized, and modulated with varying levels of granularity and interpretability.

\paragraph*{Prompt-based Methods}
Recent research shows that adopting specific personas or roles through well-engineered prompts can bias LLM cognition, allowing targeted emotional or personality outcomes~\cite{tan2024phantom,jiang2024personallm,DBLP:conf/nips/SalewskiARSA23,cava2024open}. By inserting instructions such as ``If you were a [persona]'', LLMs adapt not only their thematic style, but also their underlying emotional stance. This approach is powerful for real-time manipulation, though it can be inconsistent across tasks and model variants, highlighting the need for more systematic methods.

\paragraph*{Training-based Methods}
Fine-tuning and parameter-efficient strategies offer deeper, more stable ways to induce or alter LLM emotions~\cite{jain2024text,lu-etal-2023-miracle,huang2024reliability}. Quantized Low-Rank Adaptation (QLoRA) and specialized datasets can embed nuanced traits such as the Big Five or MBTI profiles directly into the model's learned weights. These methods enable LLMs to spontaneously exhibit trait-specific behaviors (including emoji use) and sustain their emotional states over longer dialogues, while also offering interpretability through neuron-level activation patterns.

\paragraph*{Neuron-based Methods}
A recent advance isolates personality-specific neurons and manipulates them directly to evoke or suppress emotional traits~\cite{deng2024neuronbased}. By toggling neuron activations pinpointed through psychologically grounded benchmarks (e.g., PersonalityBench), LLMs can embody targeted emotional dimensions without retraining the entire network. This neuron-centric approach provides fine-grained, dynamic control over model behaviors, representing a leap in precision and efficiency for emotional manipulation in LLMs.

\section{Summary and Discussion}
\label{sec:emotion-summary}

\lettrine[lines=3]{\initfamily\textcolor{darkgreen}{A}}{s} emotional capabilities in AI systems rapidly advance, they open up transformative opportunities across domains—from mental health support to empathetic dialogue systems. However, these developments also raise critical questions about safety, ethics, and authenticity. This section reflects on the broader implications of emotional AI, including risks of manipulation, alignment challenges, ethical dilemmas in human-AI interactions, and the philosophical distinction between emotional mimicry and genuine experience. Addressing these issues is essential for the responsible development and deployment of emotionally aware AI systems.

\paragraph*{Manipulation and Privacy Concerns}
The rapid adoption of Emotional AI in advertising and politics raises significant manipulation and privacy risks~\cite{podoletz2023we,emotional_ai_privacy_2024}. Emotional AI often collects sensitive biometric data, such as facial expressions and voice tones, to infer emotional states, enabling targeted advertising or political influence. However, these systems can exploit human emotions for profit or political gain, infringing on fundamental rights and fostering over-surveillance in public spaces~\cite{emotional_ai_risks_opportunities, emotional_ai_privacy_2024}. Regulatory frameworks like GDPR and the EU AI Act are critical to mitigating these risks responsibly.

\paragraph*{Alignment Issues}
Emotional AI's capacity to detect and interpret emotions is often misaligned with intended outcomes, leading to inaccuracies and biases. Anxiety-inducing prompts, for instance, have been shown to exacerbate biases in large language models (LLMs), affecting outputs in high-stakes domains such as healthcare and education~\cite{codaforno2024inducinganxietylargelanguage,jin2024better}. Misinterpretation of emotional cues by AI systems, as seen in workplace applications, can exacerbate discrimination and power imbalances~\cite{mantello2024emotional}. Techniques like reinforcement learning from human feedback (RLHF) have proven effective in mitigating these issues but require further development to ensure robust alignment in diverse contexts~\cite{codaforno2024inducinganxietylargelanguage, wang2023emotional}.

\paragraph*{Ethical Implications}
Trust and acceptance of AI systems are significantly influenced by their ability to exhibit empathy and maintain socially appropriate behavior~\cite{pelau2021makes, ratican2023six}. However, the commodification of emotions in workplace management and customer service has raised concerns about ethical labor practices and AI-human relationships~\cite{mantello2024emotional}. Moreover, Emotional AI's reliance on anthropomorphic characteristics without sufficient empathy can undermine user trust~\cite{pelau2021makes}. Frameworks like SafeguardGPT, which incorporate psychotherapy techniques, demonstrate promising approaches to fostering trust and aligning AI behavior with societal norms~\cite{lin2023towards}. Nonetheless, challenges remain in ensuring privacy, fairness, and cultural sensitivity~\cite{lin2023towards, ratican2023six}.

\paragraph*{Distinguishing AI Emotional Mimicry from Human Experience}
Despite advances in emotion modeling for LLM agents, a fundamental distinction remains: these systems do not actually ``feel'' emotions as humans do but only show human-emotion-like patterns via probabilistic modeling. While LLMs can convincingly simulate emotional responses, recognize emotional patterns, and generate affectional outputs, they lack the embodied, phenomenological experience that defines human emotions. This simulation-reality gap creates both technical and ethical challenges. Users frequently anthropomorphize AI systems that display emotion-like behaviors~\cite{pelau2021makes}, potentially leading to misplaced trust or expectations. This distinction needs to be carefully thought in both research and deployment contexts, as the perceived emotional capabilities of LLMs influence human-AI relationships, ethical frameworks, and regulatory approaches. Future work should balance enhancing LLMs' emotional intelligence while maintaining transparency about their fundamental limitations as non-sentient systems.
\chapter{Perception}
\label{ch:perception}

\begin{figure}[!ht]
\centering
\footnotesize
    \begin{forest}
        for tree={
            forked edges,
            draw,
            rounded corners,
            node options={align=center},
            s sep=6pt,
            calign=center,
            grow=east,
            reversed=true,
            anchor=base west,
            parent anchor=east,
            child anchor=west,
            base=left,
            font=\small,
            minimum width=2.5em
          },
          where level=1{text width=5em,fill=customblue!50}{},
          where level=2{text width=5em,fill=customgreen!50}{}
        [Perception, fill=gray!20
            [Unimodal Models, for tree={
                answer,
                calign=child edge, calign child=(n_children()+1)/2,
                yshift=-0.2cm
            }
                [Text, text width=40pt,
                    [BERT~\cite{devlin2018bert}
                    RoBERTa~\cite{liu2019roberta}
                    ALBERT~\cite{lan2019albert}, answer_work, text width=190pt]
                ]
                [Image, text width=40pt,
                    [ResNet~\cite{he2016deep}
                    DETR~\cite{carion2020end}
                    Grounding DINO 1.5~\cite{ren2024grounding}, answer_work, text width=190pt]
                ]
                [Video, text width=40pt,
                    [ViViT~\cite{arnab2021vivit}
                    VideoMAE~\cite{tong2022videomae}, answer_work, text width=190pt]
                ]
                [Audio, text width=40pt,
                    [FastSpeech 2~\cite{ren2020fastspeech}
                    Seamless~\cite{barrault2023seamless}
                    wav2vec 2.0~\cite{baevski2020wav2vec}, answer_work, text width=190pt]
                ]
                [Others, text width=40pt,
                    [Visual ChatGPT~\cite{wu2023visual}
                    HuggingGPT~\cite{shen2024hugginggpt}
                    MM-REACT~\cite{yang2023mm}
                    ViperGPT~\cite{suris2023vipergpt}
                    AudioGPT~\cite{huang2024audiogpt}
                    LLaVA-Plus~\cite{liu2025llava}, answer_work, text width=190pt]
                ]      
            ]
            [Cross-modal Models, for tree={
                answer,
                calign=child edge, calign child=(n_children()+1)/2,
                yshift=-0.2cm
            }
                [Text-Image, text width=50pt,
                    [CLIP~\cite{alec2021clip}
                    ALIGN~\cite{jia2021scaling}
                    DALL·E 3~\cite{betker2023improving}
                    VisualBERT~\cite{li2019visualbert}, answer_work, text width=180pt]
                ]
                [Text-Video, text width=50pt,
                    [VideoCLIP~\cite{xu2021videoclip}
                    Phenaki~\cite{villegas2022phenaki}
                    Make-A-Video~\cite{singer2022make}, answer_work, text width=180pt]
                ]
                [Text-Audio, text width=50pt,
                    [Wav2CLIP~\cite{wu2022wav2clip}
                    VATT~\cite{akbari2021vatt}
                    AudioCLIP~\cite{guzhov2022audioclip}, answer_work, text width=180pt]
                ]
                [Others, text width=50pt,
                    [CLIP-Forge~\cite{sanghi2022clip}
                    Point-E~\cite{nichol2022point}, answer_work, text width=180pt]
                ]
            ]
            [MultiModal Models, for tree={
                answer,
                calign=child edge, calign child=(n_children()+1)/2,
                yshift=-0.2cm
            }
                [VLM, for tree={
                            answer,
                            calign=child edge, calign child=(n_children()+1)/2
                        }, text width=40pt,
                        [
                        MiniGPT-v2~\cite{chen2023minigpt}
                        LLaVA-NeXT~\cite{liu2024llava}
                        CogVLM2~\cite{hong2024cogvlm2}
                        Qwen2.5-VL~\cite{bai2025qwen2}
                        Emu3~\cite{wang2024emu3}
                        MLaGA~\cite{fan2025mlaga}
                        LLaDA-V~\cite{you2025llada}
                        , answer_work, text width=190pt]
                        [Edge-Side, text width=40pt
                            [TinyGPT-V~\cite{yuan2023tinygpt}
                            MobileVLM v2~\cite{chu2024mobilevlm}
                            MiniCPM-V~\cite{yao2024minicpm}
                            OmniParser v2 ~\cite{yu2025omniparser}, answer_work, text width=135pt]
                        ]
                ]
                [VLA, text width=40pt,
                    [CLIPort~\cite{shridhar2022cliport}
                    RT-1~\cite{brohan2022rt}
                    MOO~\cite{stone2023open}
                    PerAct~\cite{shridhar2023perceiver}
                    Diffusion Policy~\cite{chi2023diffusion}
                    PaLM-E~\cite{driess2023palm}
                    MultiPLY~\cite{hong2024multiply}, answer_work, text width=190pt]
                ]
                [ALM, text width=40pt,
                    [Audio Flamingo~\cite{kong2024audio}
                    SpeechVerse~\cite{das2024speechverse}
                    UniAudio 1.5~\cite{yang2024uniaudio}
                    Qwen2-Audio~\cite{chu2024qwen2}
                    Audio-LLM~\cite{li2024audio}
                    Mini-Omni2~\cite{xie2024mini}
                    SpeechGPT~\cite{zhang2023speechgpt}, answer_work, text width=190pt]
                ]
                [AVLM, text width=40pt,
                    [ONE-PEACE~\cite{wang2023one}
                    PandaGPT~\cite{su2023pandagpt}
                    Macaw-LLM~\cite{lyu2023macaw}
                    LanguageBind~\cite{zhu2023languagebind}
                    UnIVAL~\cite{shukor2023unival}
                    X-LLM~\cite{chen2023x}
                    Qwen2.5-Omni~\cite{xu2025qwen2}, answer_work, text width=190pt]
                ]
                [Others, text width=40pt,
                    [
                    PointLLM~\cite{xu2025pointllm}
                    MiniGPT-3D~\cite{tang2024minigpt}
                    NExT-GPT~\cite{wu2023next}
                    Unified-IO 2~\cite{lu2024unified}
                    CoDi-2~\cite{tang2024codi}
                    ModaVerse~\cite{wang2024modaverse}
                    LLaVA-NeXT-Med~\cite{guo2025llava}, answer_work, text width=190pt]
                ]
            ]
        ]
    \end{forest}
    \label{fig:perception-tree}
    \caption{A taxonomy of selected research works about perception models or systems.}
\end{figure}

\lettrine[lines=3]{\initfamily\textcolor{darkgreen}{P}}{erception} is the foundational gateway through which humans and intelligent agents acquire information, interpret their surroundings, and make informed decisions. For humans, perception is seamless and intuitive, effortlessly transforming sensory inputs into meaningful interpretations. In artificial intelligence, however, perception systems are carefully engineered to emulate—and in some respects surpass—human sensory processing, profoundly influencing an agent's capacity for interaction, learning, and adaptation in complex environments.

In this chapter, we begin by exploring key differences in the nature and efficiency of perception between humans and AI agents. Next, we categorize agent perception based on different forms and representations of perceptual input. We then discuss ongoing challenges in the agent perception system and highlight promising directions for improvement at both the modeling and system architecture levels. Finally, we illustrate how perception modules can be effectively tailored to different agent scenarios, offering practical guidance for optimization and identifying pivotal areas for future research. Figure~\ref{fig:perception-tree} shows a taxonomy of the relevant research works that will be discussed.

\section{Human versus AI Perception}
\label{sec:human-vs-ai-percept}

\lettrine[lines=3]{\initfamily\textcolor{darkgreen}{P}}{erception} is fundamental to intelligence, serving as the interface through which both humans and artificial agents interact with the world. Although humans commonly think of perception in terms of the five classical senses—vision, hearing, taste, smell, and touch—modern neuroscience identifies a richer sensory landscape. Conservatively, humans are described as having around 10 senses; more comprehensive views list approximately 21, while some researchers propose up to 33 distinct sensory modalities~\cite{macpherson2011senses,ward2019student}. Beyond the familiar senses, humans possess sophisticated internal perceptions, such as vestibular (balance), proprioception (awareness of body position), thermoception (temperature), and nociception (pain), enabling nuanced interaction with their environment.

Human senses are finely tuned to specific physical signals: for example, human vision detects electromagnetic waves with wavelengths between approximately 380--780 nm, whereas hearing perceives sound frequencies from about 20 Hz to 20 kHz~\cite{coren2004sensation}. These sensory modalities allow humans to effortlessly engage in complex tasks like language communication, object recognition, social interaction, and spatial navigation. Additionally, humans naturally perceive continuous changes over time, seamlessly integrating motion perception and temporal awareness—abilities essential for coordinated movement and decision-making~\cite{grondin2010timing}. Animals in the natural world exhibit even more diverse perceptual capabilities. Birds and certain marine organisms, for instance, utilize magnetoreception to navigate using Earth's magnetic fields, while sharks and electric eels exploit electroreception to sense electrical signals emitted by other organisms—abilities humans do not possess~\cite{mouritsen2018magnetoreception}.

In contrast to biological perception, artificial agents rely upon engineered sensors designed to transform environmental stimuli into digital signals that algorithms can interpret. Common sensor modalities for AI agents include visual sensors (cameras), auditory sensors (microphones), tactile sensors, and inertial measurement units. AI agents typically excel at processing visual, auditory, and textual data, leveraging advances in deep learning and signal processing. However, certain human sensory abilities—particularly taste and smell—remain challenging for machines to emulate accurately. For example, the advanced bio-inspired olfactory chip developed by researchers~\cite{wang2024biomimetic} currently distinguishes around 24 different odors—a capability significantly less sensitive than the human olfactory system, which discriminates among more than 4,000 distinct smells~\cite{bushdid2014humans}.

Another crucial distinction lies in perceptual processing efficiency. Human perception is limited by biological constraints such as nerve conduction speeds, typically in the range of milliseconds. Conversely, AI systems can process sensory input at speeds of microseconds or even nanoseconds, constrained primarily by computational hardware performance rather than biological limitations. Nevertheless, human perception naturally integrates information from multiple sensory modalities—known as \emph{multimodal perception}—into coherent experiences effortlessly. For AI agents, achieving this integration requires carefully designed fusion algorithms that explicitly combine input from various sensors to build unified environmental representations~\cite{baltruvsaitis2018multimodal}.

Further differences arise in how humans and artificial agents handle temporal and spatial information. Human perception is inherently continuous and fluid, smoothly experiencing the passage of time and spatial motion without explicit temporal discretization. In contrast, AI agents typically rely on discrete sampling of sensor data, using timestamps or sequential processing to simulate continuity. Spatial awareness in humans effortlessly merges visual, auditory, and vestibular information to achieve intuitive spatial positioning. For artificial agents, spatial perception usually involves algorithmic processes such as simultaneous localization and mapping (SLAM) or 3D scene reconstruction from visual data sequences~\cite{cadena2016past}.

Physical or chemical stimuli transmitted from the external environment to human sensory organs are received by the sensory system (such as eyes, ears, skin, etc.) and converted into neural signals, which are then processed by the brain to produce perception of the environment. Similarly, to allow the intelligent agent to connect with the environment, it is crucial to capture and process perceptual signals. Currently, various sensors are used to convert environmental input into processable digital signals.

In this section, we distinguish between \emph{unimodal models}, \emph{cross-modal models}, and \emph{multimodal models} based on the number of modalities involved in the input and whether unified fusion modeling operations are performed. Unimodal models specifically process and analyze data from a single modality or type of input (such as text, image, or audio). \emph{Cross-modal models} establish relationships and enable translations between different modalities through dedicated mapping mechanisms. \emph{Multimodal models} holistically integrate and process multiple modalities simultaneously to leverage complementary information for comprehensive understanding and decision-making.

\begin{figure}[!ht]
\centering
\includegraphics[width=0.7\columnwidth]{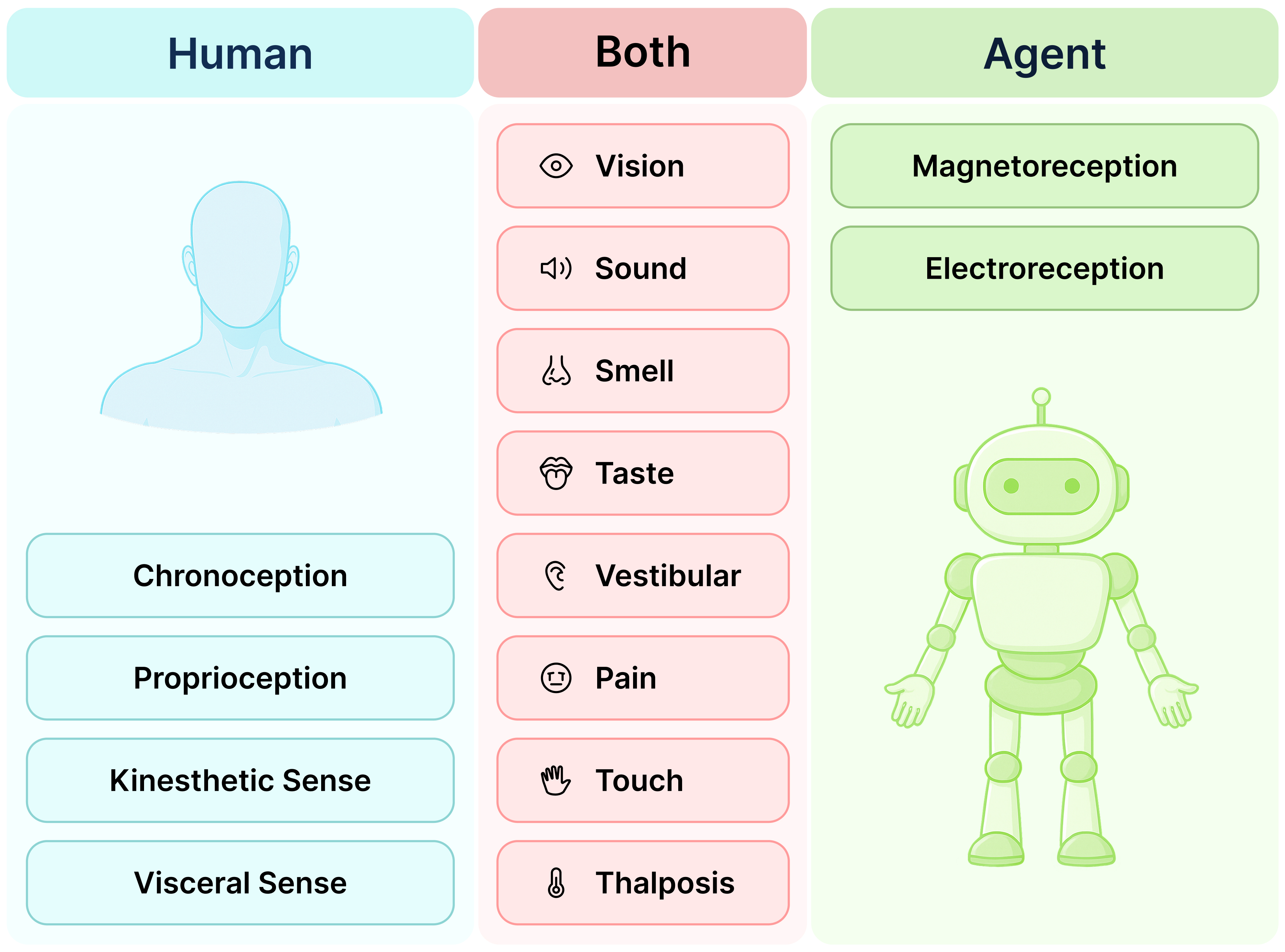}
\caption{Comparison of common perceptual types between human beings and AI agents.}
\label{fig:perception_comparison}
\end{figure}

\section{Types of Perception Representation}
\label{sec:perception-types}

\lettrine[lines=3]{\initfamily\textcolor{darkgreen}{T}}{he} agent receives and processes information from the external environment or internal state through the perception system (such as cameras, microphones, sensors, etc.) to form an understanding of the environment. By understanding the environment, more effective decision-making actions can be made, which is the key to the transformation from ``rule-driven'' to ``environment-driven'' intelligence.

\subsection{Formulation of Perception}

Depending on the type of perception model, we can provide a general mathematical formulation based on the nature of the input, representation, and output.

\begin{definition}[\textbf{Perception}]
A \emph{unimodal model} receives an observation \( \mathbf{o}_m \) from the environment \( \varepsilon \) via a sensor \( \mathrm{P}_m \), and encodes it into a latent vector \( \mathbf{z}_m \in \mathcal{Z}_m \) using an encoder \( \mathrm{E}_m \), where \( \mathcal{Z}_m \) denotes the latent space for modality \( m \):

\begin{equation}
\begin{aligned}
\mathbf{o}_m &= \mathrm{P}_m(\varepsilon), \\
\mathbf{z}_m &= \mathrm{E}_m(\mathbf{o}_m) \in \mathcal{Z}_m.
\end{aligned}
\end{equation}

\paragraph*{}
A \emph{cross-modal model} transforms the latent representation from one modality \( m \) into another modality \( n \) using a generator \( \mathrm{G}_{m \to n} \), followed by a decoder \( \mathrm{D}_n \), producing a synthesized output \( \hat{\mathbf{o}}_n \):

\begin{equation}
\hat{\mathbf{o}}_n = \mathrm{D}_n\left(\mathrm{G}_{m \to n}(\mathbf{z}_m)\right),
\end{equation}
where:
\begin{itemize}
    \item \( \mathbf{z}_m \) is the latent vector for modality \( m \),
    \item \( \mathrm{G}_{m \to n} \) is the generator mapping from \( m \) to \( n \),
    \item \( \mathrm{D}_n \) is the decoder for modality \( n \),
    \item \( \hat{\mathbf{o}}_n \) is the generated observation in modality \( n \).
\end{itemize}

\paragraph*{}
A \emph{multimodal model} fuses multiple latent vectors \( \mathbf{z}_{m_1}, \dots, \mathbf{z}_{m_k} \) from different modalities using a fusion function \( \mathrm{F} \), producing a joint semantic representation \( \mathbf{Z} \):

\begin{equation}
\mathbf{Z} = \mathrm{F}(\mathbf{z}_{m_1}, \dots, \mathbf{z}_{m_k}),
\end{equation}
where:
\begin{itemize}
    \item \( \mathbf{z}_{m_i} \) is the latent vector for modality \( m_i \),
    \item \( \mathrm{F} \) is the multimodal fusion function,
    \item \( \mathbf{Z} \) is the final fused representation.
\end{itemize}
\end{definition}

\begin{figure}[!ht]
\centering
\includegraphics[width=0.95\linewidth]{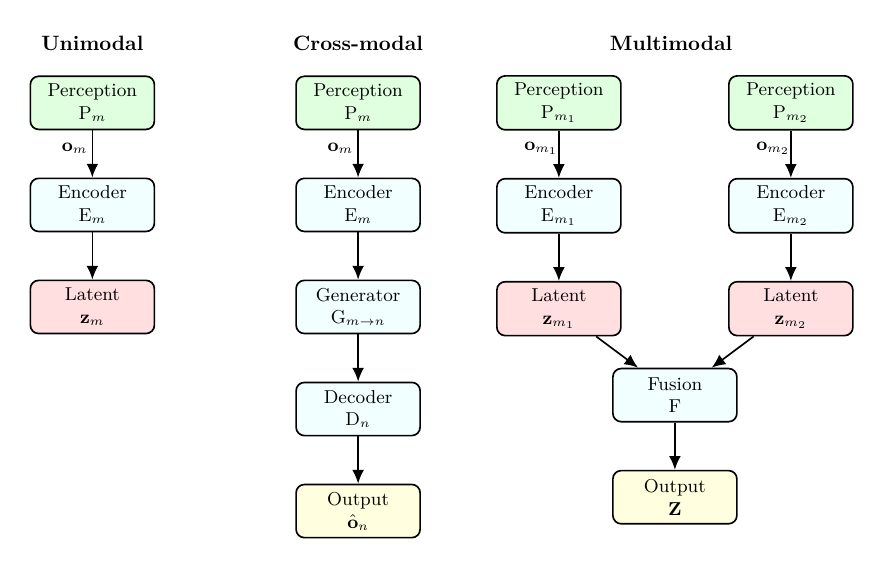}
\caption{Illustration of unimodal, cross-modal, and multimodal perception. Each model begins with raw observations \( \mathbf{o} \) collected by perception modules \( \mathrm{P} \), which are then processed by encoders \( \mathrm{E} \), translated via cross-modal generators \( \mathrm{G} \), and optionally fused into shared semantic spaces \( \mathbf{Z} \).}
\label{fig:perception-models}
\end{figure}

Figure~\ref{fig:perception-models} provides a visual summary of these three types of perception models, illustrating how raw observations from sensors are transformed into latent representations, translated across modalities, or fused into unified representations.

\subsection{Unimodal Models}
\label{subsec:unimodal}

When humans are in an environment, they can listen to beautiful music, look at sunrise and sunset, or experience a wonderful audiovisual feast on stage. These perceptual contents can be either a single image or audio, or an integration of multiple perceptual inputs. Regarding the types of perception input of intelligent agents, we will start with single-modal and multimodal inputs, and introduce their implementation and differences.

\paragraph*{Text}
As an important means of communication, text carries a wealth of information, thoughts, emotions, and culture. Humans indirectly obtain the content of text through vision, hearing, and touch, which is one of the most important ways for humans to interact with the environment. For intelligent agents, text can serve as a bridge to connect with the environment, directly taking text as input and outputting a response. In addition to the literal meaning, text also contains rich semantic information and emotional color. In the early days, the bag-of-words model~\cite{zhang2010understanding} was used to count text content and was widely used in text classification scenarios, but semantic expression could not be obtained. BERT~\cite{devlin2018bert} uses a bidirectional Transformer architecture for language modeling and captures deep semantic information of text through large-scale unsupervised pretraining. ~\cite{liu2019roberta, lan2019albert} further optimized the training efficiency of BERT. The autoregressive model represented by GPT3.5~\cite{openai2023gpt35} opened the prelude to LLM and further unified the tasks of text understanding and text generation, while technologies such as LoRA~\cite{hu2021lora} greatly reduced the application cost of LLM and improved the agent's perception ability of complex real-world scenario tasks.

\paragraph*{Image}
Image is another important way for humans to interact with the environment which inherently encode spatial information, encompassing crucial attributes such as morphological characteristics, spatial positioning, dimensional relationships, and kinematic properties of objects. The evolution of computer vision architectures has demonstrated significant advancement in processing these spatial attributes. The seminal ResNet architecture~\cite{he2016deep} established foundational principles for deep visual feature extraction, while subsequent YOLO series~\cite{Jocher_YOLOv5_by_Ultralytics_2020, Jocher_Ultralytics_YOLO_2023} demonstrated the capability to simultaneously determine object localization and classification with remarkable efficiency. A paradigm shift occurred with the introduction of DETR~\cite{carion2020end}, which revolutionized object detection by implementing parallel prediction through global context reasoning, effectively eliminating traditional computational overhead associated with non-maximum suppression and anchor point generation. More recently, DINO 1.5~\cite{ren2024grounding} has extended these capabilities to open-set scenarios through architectural innovations, enhanced backbone networks, and expanded training paradigms, substantially improving open-set detection performance and advancing the perceptual generalization capabilities of artificial agents in unconstrained environments.

\paragraph*{Video}
Video is an expression of continuous image frames, which includes the time dimension and displays dynamic information that changes over time through continuous image frames. The intelligent agent uses video as input and obtains richer perceptual content through continuous frames. ViViT~\cite{arnab2021vivit} extracts spatiotemporal markers from videos, effectively decomposing the spatial and temporal dimensions of the input. VideoMAE~\cite{tong2022videomae} learns general video feature representations through self-supervised pre-training and has strong generalization capabilities on out-of-domain data. It lays a solid foundation for intelligent agents to acquire perceptual capabilities in new scenarios.

\paragraph*{Audio}
In addition to text and vision, another important way for humans to interact with the environment is through audio. Audio not only contains direct text content, but also contains the speaker's tone and emotion~\cite{zeng2022attention}. Wav2Vec2~\cite{baevski2020wav2vec} defines the contrast task by quantizing the potential representation of joint learning, achieving speech recognition effectiveness with 1/100 labeled data volume. FastSpeech 2~\cite{ren2020fastspeech} directly introduces voice change information (pitch, energy, duration, etc.) and uses real targets to train the model to achieve more realistic text-to-speech conversion. Seamless~\cite{barrault2023seamless} generates low-latency target translations through streaming and using an efficient monotonic multi-head attention mechanism, while maintaining the human voice style, to achieve synchronous speech-to-speech/text translation from multiple source languages to target languages. Based on these means, the intelligent agent can achieve the ability to listen and speak.

\paragraph*{Others}
At present, most of the research on intelligent agents focuses on the above-mentioned common sensory input types. However, just as humans have more than 20 types of perception, intelligent agents have also made progress in achieving corresponding perception capabilities through other sensors. The bionic olfactory chip developed by Hong Kong University of Science and Technology~\cite{wang2024biomimetic} integrates a nanotube sensor array on a nanoporous substrate, with up to 10,000 independently addressable gas sensors on each chip, which is similar to the configuration of the olfactory system of humans and other animals, and can accurately distinguish between mixed gases and 24 different odors. In terms of taste, Tongji University~\cite{zhang2024smart} combines fluorescence and phosphorescence signals to develop an intelligent taste sensor with multi-mode light response, which can effectively identify umami, sourness, and bitterness. In order to achieve human-like perception and grasping capabilities, New York University~\cite{bhirangi2024anyskin} launched a low-cost magnetic tactile sensor AnySkin, which can be quickly assembled and replaced. Even in the perception of pain, the Chinese Academy of Sciences uses the unique electrical properties of liquid metal particle films when they are ``injured'' (mechanically scratched) to imitate the perception and positioning of ``wound''. Some other works, including HuggingGPT~\cite{shen2024hugginggpt}, LLaVA-Plus~\cite{liu2025llava}, and ViperGPT~\cite{suris2023vipergpt}, integrate these single-modal perception capabilities within the framework, select and apply them according to task requirements, and achieve the goal of achieving more complex tasks.

\subsection{Cross-modal Models}
\label{sec:cross-modal}

\begin{figure}[!ht]
\centering
\includegraphics[width=0.95\linewidth]{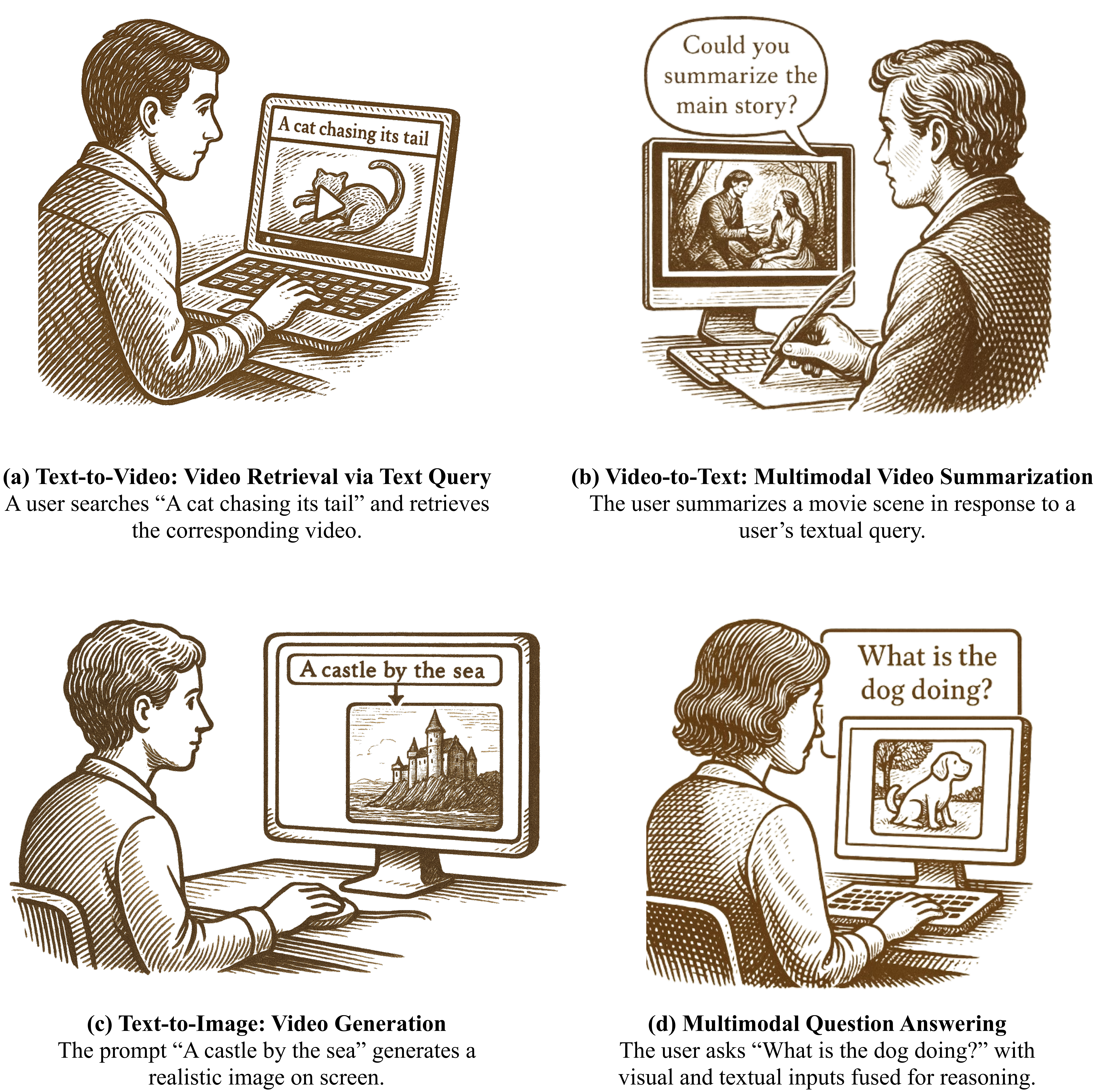}
\caption{Representative cross-modal and multimodal applications in intelligent agents.
This figure illustrates several core capabilities: retrieving videos based on textual queries (top left), summarizing video content through natural language (top right), generating images from text prompts (bottom left), and answering questions by integrating text, image, and audio context (bottom right). These examples demonstrate the growing competence of agents in understanding and generating across modalities. Beyond these, cross-modal modeling also encompasses emerging directions such as text-to-3D generation, text-to-audio synthesis, and dynamic fusion of multiple modalities for richer perception and interaction.}
\label{fig:cross-model-applications}
\end{figure}

\paragraph*{Text-Image}
Cross-modal models integrating text and images have witnessed significant advancements in recent years, leading to improved alignment, retrieval,  and generation between the two modalities. These models can be categorized based on their primary objectives, including cross-modal alignment and retrieval, text-to-image generation, and image-to-text generation. Figure~\ref{fig:cross-model-applications} shows a few examples of cross-modal and multimodal application scenarios.

One of the primary focuses in cross-modal research is the alignment and retrieval of text and images. CLIP~\cite{alec2021clip}, introduced by OpenAI in 2021, employs contrastive learning to align textual and visual representations, enabling zero-shot cross-modal retrieval and classification. Similarly, ALIGN~\cite{jia2021scaling}, developed by Google in the same year, leverages large-scale noisy web data to optimize text-image embedding alignment. In 2022, CyCLIP~\cite{goel2022cyclip} introduced a cyclic consistency loss to further enhance the robustness of cross-modal alignment, improving the reliability of retrieval tasks.

Another major area of progress involves text-to-image generation, where models aim to synthesize high-quality images based on textual descriptions. OpenAI's DALL·E series~\cite{ramesh2021zero,ramesh2022hierarchical,betker2023improving}, spanning from 2021 to 2023, has made substantial contributions in this domain, with DALL·E 3 offering fine-grained semantic control over generated images. Stable Diffusion~\cite{rombach2022high}, introduced by Stability AI in 2022, employs a diffusion-based generative approach that supports open-domain text-to-image synthesis and cross-modal editing.

A third significant research direction is image-to-text generation, where models aim to generate high-quality textual descriptions based on image inputs. Typical representative work is the BLIP~\cite{li2022blip} and BLIP-2~\cite{li2023blip} models, introduced by Salesforce between 2022 and 2023, which utilize lightweight bridging modules to enhance vision-language model integration, enabling tasks such as image captioning and question answering.

\paragraph*{Text-Video}
The key research here involves video text alignment, generation and retrieval. VideoCLIP~\cite{xu2021videoclip} employs a video encoder—typically based on temporal convolution or a transformer structure—to extract sequential features from video frames. These features are subsequently aligned with textual representations generated by a language encoder, facilitating robust video-text association. In the domain of text-to-video generation, Meta's Make-A-Video model~\cite{singer2022make} extends spatial-temporal dimensions using diffusion-based techniques, allowing for high-quality video synthesis from textual descriptions. Additionally, Google's Phenaki ~\cite{villegas2022phenaki} addresses the challenge of generating long, temporally coherent video sequences, demonstrating significant advancements in video synthesis through cross-modal learning.DeepMind's Frozen in Time~\cite{bain2021frozen}  adopts contrastive learning for video-text matching, thereby enabling efficient cross-modal retrieval. This approach enhances the capacity to search and retrieve relevant video segments based on textual queries, further improving the integration of vision and language understanding.

\paragraph*{Text-Audio}
Cross-modal models connecting text and audio have made significant improvements in related tasks such as modal representation, generation, and conversion, and enhanced the perception ability under a single modality.

AudioCLIP~\cite{guzhov2022audioclip}, introduced in 2021, extends the CLIP framework to the audio domain, enabling tri-modal retrieval across audio, text, and images. By incorporating audio as an additional modality, AudioCLIP utilizes multi-task learning to unify image, text, and audio representations into a shared embedding space. This advancement enhances the capability of cross-modal retrieval and interaction. In a similar vein, VATT~\cite{akbari2021vatt} adopts a unified Transformer-based architecture to process video, audio, and text through independent encoding branches. These branches are subsequently fused into a shared multimodal space, facilitating tasks such as cross-modal retrieval and multi-task learning. This design allows for greater adaptability across diverse multimodal scenarios.

For text-to-audio generation, Meta introduced AudioGen~\cite{kreuk2022audiogen} in 2023, which enables the synthesis of audio, such as environmental sounds and music fragments, directly from textual descriptions. This model exemplifies the growing capabilities of AI in generating high-fidelity audio based on linguistic input, expanding applications in media, entertainment, and accessibility.

Additionally, in the domain of speech-to-text and text-to-speech conversion, Microsoft developed SpeechT5~\cite{ao2021speecht5}. This model unifies speech and text generation, supporting both speech synthesis and recognition within a single framework. By leveraging a shared architecture for these dual functionalities, SpeechT5 contributes to the seamless integration of speech and text processing, thereby enhancing applications in automated transcription, voice assistants, and accessibility tools.

\paragraph*{Others}
In some other scenarios and domains, cross-modal modeling also plays an important role.

CLIP-Forge~\cite{sanghi2022clip} presents a novel method for generating 3D shapes from textual descriptions. By leveraging the capabilities of Contrastive Language-Image Pre-training (CLIP), this approach enables the synthesis of high-quality 3D objects conditioned on natural language inputs, bridging the gap between text and 3D geometry. Point-E~\cite{nichol2022point} extends this concept by generating 3D point clouds from text descriptions. Unlike traditional 3D reconstruction techniques, Point-E focuses on point cloud representations, facilitating efficient and scalable 3D content creation while maintaining high fidelity to textual prompts.

In the field of medical imaging, MoCoCLIP~\cite{bhardwaj2025enhancing} introduces an approach that enhances zero-shot learning capabilities. By integrating CLIP with Momentum Contrast (MoCo), this method improves the generalization of deep learning models in medical imaging applications, addressing the challenges associated with limited annotated data and domain adaptation.

\subsection{Multimodal Models}
\label{subsec:multimodal}

The cross-modal model described above mainly aligns and maps between modalities through contrastive learning and other methods to achieve information complementarity and conversion between modalities. Furthermore, the work of multimodal models focuses on how to integrate the features of multiple data sources (such as vision, text, audio, etc.) to improve the performance of the overall model.

\paragraph*{Vision-Language Model}
Vision-Language Model(VLM) is broadly defined as multimodal model that can learn from images(or videos) and text. Humans live in a world full of multimodal information. Visual information (such as images and videos) and language information (such as text) often need to be combined to fully express meaning. The same is true for intelligent agents. LLaVA~\cite{liu2024llava} first tried to use GPT-4 to generate a multimodal language image instruction dataset. Through end-to-end training, a large multimodal model was obtained and excellent multimodal chat capabilities were demonstrated. LLaVA-NeXT~\cite{liu2024llava} uses dynamic high-resolution and mixed data to show amazing zero-shot capabilities even in pure English modal data, and the computational/training data cost is 100-1000 times smaller than other methods. Emu2~\cite{sun2024generative} changes the traditional way of using image tokenizer to convert images into discrete tokens, and directly uses image encoders to convert images into continuous embeddings and provide them to Transformer, enhancing multimodal context learning capabilities. Furthermore, Emu3~\cite{wang2024emu3} tokenizes the multimodal input into a discrete space and simplifies the complex multimodal model design by using only the next token prediction for training. MiniGPT-v2~\cite{chen2023minigpt} employs unique identifiers for various tasks during training. These identifiers help the model differentiate task instructions more effectively, enhancing its learning efficiency for each task. Qwen2-VL~\cite{wang2024qwen2}, DeepSeek-VL2~\cite{wu2024deepseekvl2} use dynamic encoding strategies on visual components, aiming to process images with different resolutions and generate more efficient and accurate visual representations. At the same time, DeepSeek-VL2~\cite{wu2024deepseekvl2} also uses the MoE model with a multi-head potential attention mechanism to compress the key-value cache into a latent vector to achieve efficient reasoning.

Previous work mainly uses image fusion text for training. Video-ChatGPT~\cite{maaz2023video} extends the input to video and directly uses a video adaptive visual encoder combined with LLM for training to capture the temporal dynamics and inter-frame consistency relationships in video data, thereby enabling open conversations about video content in a coherent manner. To solve the lack of unified tokenization for images and videos, Video-LLaVA~\cite{lin2023video} unifies the visual representations of image and video encoding into the language feature space, making the two mutually reinforcing. Similarly, Chat-UniVi~\cite{jin2024chat} employs a set of dynamic visual tokens to integrate images and videos, while utilizing multi-scale representations to allow the model to grasp both high-level semantic concepts and low-level visual details. Youku-mPLUG~\cite{xu2023youku} has conducted in-depth research in specific scenarios. Based on the high-quality Chinese video-text pairs in the Youku video sharing platform, it enhances the ability to understand overall and detailed visual semantics and recognize scene text. Unlike the previous method that requires training, SlowFast-LLaVA~\cite{xu2024slowfast} can effectively capture the detailed spatial semantics and long-term temporal context in the video through a two-stream SlowFast design without any additional fine-tuning of the video data, achieving the same or even better results than the fine-tuning method.

As the parameters of large models gradually decrease and the computing power of the end-side increases, high-performance end-side models are gaining momentum. Smart terminal devices such as mobile phones and PCs have strong demands for image visual processing, which puts forward higher multimodal recognition effects and reasoning performance requirements for the deployment of AI models on the end-side. TinyGPT-V~\cite{yuan2023tinygpt} is built based on the Phi-2~\cite{javaheripi2023phi} small backbone combined with BLIP-2~\cite{li2023blip}. It only needs 8 GB of video memory or CPU for reasoning, and solves the computational efficiency problems of LLaVA~\cite{liu2024llava} and MiniGPT-4~\cite{zhu2023minigpt}. MiniCPM-V~\cite{yao2024minicpm} mainly provides powerful OCR capabilities for long and difficult images, and has a low hallucination rate, providing reliable perception output. Megrez-3B-Omni~\cite{li2025megrez} ensures that all structural parameters are highly compatible with mainstream hardware through coordinated optimization of software and hardware. Its inference speed is up to 300\% faster than that of models with the same precision, improving its adaptability to different end-side hardware.

Similarly, there are more GUI-related works focusing on automatic task execution on mobile phones and PCs. 
GUICourse~\cite{chen2024guicourse} and OS-ATLAS~\cite{wu2024atlas} build a cross-platform GUI grounding corpus, which brought significant performance improvements in the understanding of GUI screenshots and enriching the interactive knowledge of GUI components. OmniParser v2 ~\cite{yu2025omniparser} proposes the Structured-Points-of-Thought (SPOT) prompting schemas, which leverage a unified encoder-decoder structure to unify text recognition, table recognition, layout analysis, and text key information extraction into a unified framework, significantly enhancing the visually-situated text parsing capability.

\paragraph*{Vision-Language-Action Model}
Vision-Language-Action (VLA) model, which takes vision and language as inputs and generates robotic actions as outputs, represents an important research direction in the field of embodied intelligence. The selection of vision and language encoders in VLA models has undergone diverse development, evolving from early CNNs to Transformer architectures, and further integrating 3D vision and large language models. Early models such as CLIPort~\cite{shridhar2022cliport} used ResNet~\cite{he2016deep} to process visual inputs and combined language embeddings to generate actions, laying the foundation for multimodal fusion. RT-1~\cite{brohan2022rt} introduced the Transformer architecture, employing EfficientNet as the visual encoder and USE as the language encoder, and fused visual and language information via FiLM mechanisms, significantly enhancing the model's generalization ability. VIMA~\cite{stone2023open} further adopted multimodal prompts, combining the ViT visual encoder and the T5 language model to support more complex tasks. PerAct~\cite{shridhar2023perceiver} innovatively used 3D point clouds as visual inputs and processed multi-view information through Perceiver IO, providing richer spatial perception for robotic manipulation. Diffusion Policy~\cite{chi2023diffusion} combined ResNet visual encoders and Transformer language models, generating actions through diffusion models to improve the diversity and accuracy of action generation. SayCan~\cite{ahn2022can} integrated the PaLM language model with visual inputs, using the CLIP visual encoder for task decomposition. PaLM-E~\cite{driess2023palm} combined the ViT visual encoder and the PaLM language model, guiding low-level action execution through text planning. MultiPLY~\cite{hong2024multiply} further integrated 3D information into LLMs, combining the EVA visual encoder and the LLaMA language model to provide more comprehensive planning capabilities for complex tasks.

\paragraph*{Audio-Language Model}
Audio-Language Model(ALM) uses the audio and text to build a multimodal model. Speechgpt~\cite{zhang2023speechgpt} built a large-scale cross-modal speech instruction dataset SpeechInstruct and trained discrete speech representations, achieving cross-modal speech dialogue capabilities beyond expectations. LauraGPT~\cite{du2023lauragpt}, unlike the previous sampling of discrete audio tokens to represent input and output audio, proposed a novel data representation that combines the continuous and discrete features of audio, and demonstrated excellent performance on a wide range of audio tasks through supervised multi-task learning. ~\cite{das2024speechverse, ghosh2024gama, li2024audio} converts audio data into embedded representations and then fine-tunes instructions, so that excellent performance can be achieved on various speech processing tasks through natural language instructions. In order to reduce the cost of fine-tuning training, 
Audio Flamingo 2~\cite{ghosh2025audio} leverages a customized CLAP model and a multi-stage curriculum learning strategy, which achieves the best performance by training on fine-grained synthetic audio question-answering data with a 3B parameter small model.
UniAudio 1.5~\cite{yang2024uniaudio} uses words or subwords in the text vocabulary as audio tokens, learns these audio representations through a small number of samples, and achieves cross-modal output without fine-tuning. In order to make the output more realistic and in line with human expectations, Qwen2-Audio~\cite{chu2024qwen2} introduced the DPO training method to achieve human preference alignment.

\paragraph*{Audio-Vision-Language Model}
Audio-Vision-Language Model (AVLM) utilizes audio, vision, and text to unify multimodal models. Previously, we introduced some work on building multimodal models using information from two modalities. In the pursuit of AGI, the obstacle to achieving this goal lies in the diversity and heterogeneity of tasks and modalities. A suitable approach is to allow more modal capabilities to be supported within a unified framework. Some closed-source work~\cite{hurst2024gpt,team2024gemini} has achieved excellent capabilities across modalities such as text, vision, and audio. ImageBind~\cite{girdhar2023imagebind} implements joint embedding across six different modes (image, text, audio, depth, thermal, and IMU data). Panda-GPT~\cite{su2023pandagpt} combines ImageBind's multi-modal encoder and Vicuna~\cite{zheng2023judging}, showing zero-shot cross-modal performance in addition to images and text. Similar work includes~\cite{chen2023x, chen2023x, lyu2023macaw}, which achieves alignment and training through the encoding information of vision, audio and text. Multimodal models often require more resources to train, and UniVAL~\cite{shukor2023unival} trained a model with only $\sim 0.25B$ parameters based on task balance and multimodal curriculum learning, and used weight interpolation to merge multimodal models, maintaining generalization under out-of-distribution. NExT-GPT~\cite{wu2023next} connects LLM with multimodal adapters and different diffusion decoders, and only trains a small number of parameters (1\%) of certain projection layers.

Other works~\cite{lu2024unified, zhan2024anygpt, tang2024codi, wang2024modaverse} have achieved input-output conversion between arbitrary modalities. Unified-IO 2~\cite{lu2024unified} is the first autoregressive multimodal model that can understand and generate images, text, audio, and actions. It tokenizes different modal inputs into a shared semantic space and processes them using an encoder-decoder model. AnyGPT~\cite{zhan2024anygpt} builds the first large-scale any-to-any multimodal instruction dataset, using discrete representations to uniformly process various modal inputs. Modaverse~\cite{wang2024modaverse} directly aligns the output of the LLM with the input of the generative model to solve the problem that previous work relies heavily on the alignment of the latent space of text and non-text features, avoiding the complexity associated with the alignment of latent features. CoDi-2~\cite{tang2024codi} outperforms earlier domain-specific models in tasks like topic-based image generation, visual transformation, and audio editing.

\paragraph*{Others}
Humans have explored the 2D world more than the 3D world, but 3D can more accurately describe the shape and texture information of objects and provide richer perceptual information. PointLLM~\cite{xu2025pointllm} uses a point cloud encoder to express geometric and appearance features, and integrates language features for two-stage training of complex point-text instructions, achieving excellent 3D object description and classification capabilities. Since 3D contains richer information than 2D, it also brings greater training costs. ~\cite{tang2024minigpt, zhu2024llava} reduces the training cost here, and MiniGPT-3D~\cite{tang2024minigpt} uses 2D priors from 2D-LLM to align 3D point clouds with LLMs. Modal alignment is performed in a cascade manner, and query expert modules are mixed to efficiently and adaptively aggregate features, achieving efficient training with small parameter updates. LLaVA-3D~\cite{zhu2024llava} connects 2D CLIP patch features with their corresponding positions in 3D space, integrates 3D Patches into 2D LMM and uses joint 2D and 3D visual language command adjustment to achieve a 3.5-fold acceleration in convergence speed.

In order to enable intelligent agents to accurately perceive and manipulate unknown objects, Meta~\cite{suresh2024neuralfeels} developed NeuralFeels technology, which combines vision and touch to continuously model unknown objects in 3D, more accurately estimate the posture and shape of objects in handheld operations, and improve the accuracy of ignorant object operations by 94\%.

\section{Optimizing Perception Systems}
\label{sec:optimize-perception}

\lettrine[lines=3]{\initfamily\textcolor{darkgreen}{P}}{erception} errors, including \emph{inaccuracies}, \emph{misinterpretations}, and \emph{hallucinations} (generation of false information), pose substantial challenges to the reliability and effectiveness of LLM-based agents. Optimizing perception thus requires minimizing these errors using various strategies across model, system, and external levels.

\subsection{Model-Level Enhancements}
\label{subsec:muldel-level-enhance}

\paragraph*{Fine-tuning}
Fine-tuning pre-trained LLMs on domain-specific data significantly improves their ability to accurately perceive and interpret relevant information. For example, fine-tuning models such as LLaVA on specific landmarks has been shown to enhance their recognition accuracy, particularly in urban navigation tasks~\cite{liu2024llava, Duan_2024_CVPR}. Moreover, techniques such as Low-Rank Adaptation (LoRA) enable more efficient fine-tuning, avoiding a substantial increase in model complexity while still improving performance~\cite{hu2021lora,fang2024towards}. Some LLM work combined with traditional vision is also widely used. Integrating with YOLOS~\cite{fang2021you} on the basis of the Llama-Adapter~\cite{zhang2023llama} architecture significantly improves the detection and positioning capability.

\paragraph*{Prompt Engineering}
The design of effective prompts is crucial to ensure LLMs generate outputs that are both accurate and aligned with the desired goals. By providing clear instructions, contextual information, and specific formatting requirements, prompt engineering minimizes misinterpretation and hallucination~\cite{liu2023pre}. System prompts define the agent's role, historical prompts to provide context from past interactions, and customized prompts to ensure output consistency has been shown to reduce errors significantly~\cite{liu2023pre}.

\paragraph*{Retrieval-Augmented Generation}
Supplementing LLMs with external knowledge sources through retrieval mechanisms helps ground their responses in factual information, reducing the likelihood of hallucinations and improving the accuracy of perceived information~\cite{lewis2020retrieval}.

\subsection{System-Level Optimizations}
\label{subsec:system-level-optimizations}

\paragraph*{Anticipation-Reevaluation Mechanism}
In scenarios where agents face incomplete or ambiguous information, an anticipation-reevaluation mechanism can enhance robustness. For instance, in navigation tasks, agents can anticipate goal directions based on historical data and reevaluate their inferences when new information becomes available~\cite{zeng2024perceive}.

\paragraph*{Multi-Agent Collaboration}
In multi-agent systems, structured communication and collaboration among agents can facilitate information sharing, error correction, and consensus-building, leading to a more accurate collective perception of the environment~\cite{yang2020overview}. Different communication topologies, such as fully connected, centralized, and hierarchical structures, offer varying trade-offs in terms of efficiency and robustness~\cite{guo2023configreco}. InsightSee~\cite{zhang2024insightsee} refines visual information through a multi-agent framework with description, reasoning, and decision-making, effectively enhancing visual information processing capabilities. Similarly, HEV~\cite{nash2023herd} integrates the global perspective information of multiple agents and endows RL agents with global reasoning capabilities through cooperative perception, thereby enhancing their decision-making capabilities.

\paragraph*{Agent Specialization}
Assigning distinct roles and capabilities to individual agents within a multi-agent system allows for a division of labor in perception, with each agent focusing on specific aspects of the environment or task. This can enhance the overall accuracy and efficiency of perception~\cite{zhang2024vipact}.

\subsection{External Feedback and Control}
\label{subsec:external-feedback}

\paragraph*{Loss Agents for Optimization}
Utilizing LLMs as loss agents allows for the dynamic adjustment of loss function weights during training~\cite{li2024lossagent}. This enables the optimization of image processing models based on complex, potentially non-differentiable objectives, including human feedback and evaluations from specialized models. This approach essentially externalizes the optimization objective, allowing the LLM to ``perceive'' and adapt to complex criteria~\cite{khot2022decomposed}.

\paragraph*{Human-in-the-Loop Systems}
Incorporating human feedback and oversight can help correct errors, guide the agent's learning process, and ensure alignment with human values and expectations~\cite{ouyang2022training}.

\paragraph*{Content and Output Mediation}
Before presenting LLM outputs to users, content mediation filters and refines these outputs. This helps prevent unexpected or harmful behaviors, ensuring alignment with user expectations and safety guidelines~\cite{schwartz2024posggym}.

\section{Real-World Applications of Perceptual Intelligence}
\label{sec:perception-applications}

\lettrine[lines=3]{\initfamily\textcolor{darkgreen}{T}}{he} operational efficacy of intelligent agents is predominantly shaped by three critical factors: model architecture dimensionality, hardware infrastructure capabilities, and quantization optimization strategies. The exponential growth in model parameters—from BERT-Base's modest 110 million to GPT-3's 175 billion, culminating in LLaMA 3's unprecedented 405 billion—has significantly increased processing latency, from milliseconds to hundreds of milliseconds. Hardware performance differences are equally critical; for instance, NVIDIA's H100 outperforms A100 with a 50\% gain in token processing throughput, while the RTX 4090 nearly doubles it.

Modern intelligent agents have entered diverse application domains, including \emph{personal assistants}, \emph{gaming environments}, \emph{robotic process automation (RPA)}, and \emph{multimedia content creation}, often relying on visual perception as the primary input modality. In procedurally generated environments such as Minecraft, STEVE~\cite{zhao2025see} achieves a 1.5x acceleration in technology tree progression and a 2.5x improvement in block search efficiency via visual input. Steve-Eye~\cite{zheng2023steve} extends this by using end-to-end multimodal training to reduce environmental understanding latency through integrated visual-textual representations.

In creative content generation, AssistEditor~\cite{gao2024assisteditor} demonstrates effective multi-agent collaboration for professional video editing, using style-aware content understanding. Audio-Agent~\cite{wang2024audio} achieves cross-modal alignment from textual and visual inputs to audio outputs, enabling high-quality audio manipulation~\cite{zhou2024codec,li2024apollo,wang2024ham}.

On mobile and desktop platforms, agent applications have advanced rapidly. ExACT~\cite{yu2024exact} sets new benchmarks in VisualWebArena~\cite{koh2024visualwebarena}, achieving a 33.7\% success rate through screenshot-based exploration with caption and mask set integration. SPA-Bench~\cite{chen2024spa} provides a rigorous mobile interaction benchmark that reflects real-world task complexity. M3A~\cite{rawles2024androidworld} achieves 64.0\% task success in SPA-Bench by leveraging multimodal input. AgentStore~\cite{jia2024agentstore} improves OSWorld PC benchmark results to 23.85\% through enhanced visual input and accessibility tree analysis.

Voice-based personal assistants~\cite{zeng2024glm,hurst2024gpt} have significantly improved user experience by reducing interaction friction while maintaining operational efficiency. Incorporating \emph{emotional prosody} has been shown to increase user engagement and retention.

In embodied intelligence applications, agents increasingly incorporate \emph{haptic and force feedback} as perceptual modalities, enabling more precise interaction with physical environments~\cite{lambeta2024digitizing}. Enhanced tactile fidelity supports more nuanced manipulation and robust control in real-world settings.

\section{Summary and Discussion}
\label{sec:summary-perception}

\lettrine[lines=3]{\initfamily\textcolor{darkgreen}{A}}{gent} perception, a foundational capability for autonomous intelligent systems, has seen significant advancements with the emergence of unified multimodal models that attempt to process and understand heterogeneous sensory inputs in a coherent manner~\cite{lu2024unified, zhan2024anygpt}. However, despite the continuous emergence of relevant research, current perception mechanisms still find it difficult to cope with core challenges such as multimodal representation learning, cross-modal alignment, and effective fusion strategies. These limitations prevent agents from developing a consistent, time-based understanding of their environment, which is critical for robust decision-making and adaptive behavior.

\paragraph*{Representation Learning Limitations}
Modern perception systems typically rely on fixed or task-specific encoding pipelines that inadequately capture the structured and high-dimensional nature of sensory data such as vision, audio and language. Many of these approaches fail to encode modality-specific priors or to characterize fine-grained spatio-temporal dependencies, resulting in poor performance in complex or unevenly distributed environments. Notably, representations derived from deep neural networks may excel within a single modality, but often lack the compositionality and semantic alignment necessary for cross-modal reasoning. This limitation becomes increasingly severe in dynamic environments, where agent perception must generalize across different contexts and sensor configurations.

\paragraph*{Cross-Modal Alignment Challenges}
Aligning multi-modal inputs into a shared latent space presents another key bottleneck. The heterogeneity of the sensory modalities, each with distinct noise characteristics, sampling rates, and information densities, makes direct alignment inherently difficult. Current methods, which often rely on contrastive objectives or co-embedding strategies, may yield superficial correlations without capturing deeper causal or temporal dependencies. This causes the agent to easily misunderstand signals in ambiguous scenarios (e.g., visual occlusion or overlapping audio sources), ultimately weakening situational awareness.

\paragraph*{Fusion Bottlenecks}
Cross-modal feature fusion, although crucial for comprehensive understanding, is still an emerging research area. Conventional fusion mechanisms (e.g., early, late, or hybrid fusion) often suffer from information dilution or domination by a single modality. More sophisticated approaches, such as those based on attention mechanisms, have shown promise, but still struggle to dynamically weight modal contributions based on task relevance and environmental context. Furthermore, maintaining perceptual consistency over long viewports (e.g., tracking objects over time or reasoning about unknown causes) requires memory-augmented architectures that are rarely integrated into mainstream agent perception pipelines.

\paragraph*{Toward Generalized Agent Perception}
Future research in agent perception should pivot toward adaptive, context-aware, and causal architectures. Representation learning must evolve to support task-conditioned abstraction, enabling agents to selectively emphasize features relevant to their current goal. Dynamic neural architectures, such as those employing meta-learned parameters or neural module networks, offer a potential research direction by allowing structural adaptation based on environmental feedback.

To address the cross-modal alignment problem, self-supervised spatio-temporal synchronization provides a compelling alternative. By leveraging temporal consistency and motion cues, such mechanisms can guide fine-grained correspondences without requiring extensive supervision. The integration of causal representation learning~\cite{zhang2024foundation} could also alleviate alignment noise by encouraging the model to separate spurious correlations from meaningful signal relationships.

In the area of fusion, hierarchical attention networks equipped with learnable gating functions allow agents to perform a more detailed and context-dependent integration of features from specific modalities. Differentiable memory architectures, particularly those that support persistent memory traces and modality-aware read/write operations, could further enable continuity in perception across extended temporal windows, an essential requirement for intelligent agents to make sequential decisions.

\paragraph*{Inspiration from Agent Design}
In conjunction with the broader goals of agent architecture, perception should not be viewed as a passive preprocessing stage, but rather as an active, queryable, and interactive subsystem. For example, an agent can implement perceptual reasoning mechanisms in which sensory inputs are interpreted in terms of hypothesized goals or actions. This is consistent with the idea of active perception and closed-loop sense action cycles, in which perception is dynamically regulated by the agent's epistemic uncertainty and decision goals.

Ultimately, perception must be reimagined not just as a data processing pipeline, but as an adaptive, strategic, and goal-oriented process that supports the full complexity of the embodied agency. Achieving this will require a deeper unification between perception models, memory systems, and decision layers, an integrative direction that promises to unlock more general and intelligent agent behaviors.

\chapter{Action Systems}
\label{ch:action}

\begin{figure}[!htb]
\centering
    \includegraphics[width=0.98\columnwidth]{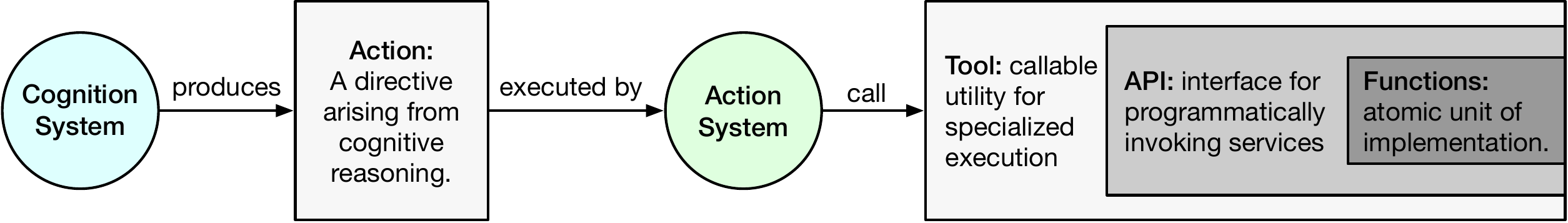}
    \caption{Illustration of several concepts related to action and action execution. In this pipeline, the cognitive system first produces actions and then feed them into the action system for execution. The callable action lists can be invoked by a form as: Tool $\rightarrow$ API $\rightarrow$ Functions.}

\label{fig:action-concepts}
\end{figure}

\lettrine[lines=3]{\initfamily\textcolor{darkgreen}{I}}{n} the realm of philosophy, action is defined as the behaviors that agents can perform for a potential or specific purpose in the interactive environment. For example, manipulation, moving, reasoning, and tool utilization can all be considered as fundamental actions that an intelligent agent can execute to fulfill a goal in real-world scenarios. In other words, actions emerge from the goal-oriented engagement of an agent in its environment, reflecting its intent to transform the external world in pursuit of its goals. Therefore, the action system also plays a vital role in differentiating AI agents and foundation models (e.g., LLMs). Generally, existing foundation models have demonstrated impressive performance across various tasks, but their task scope is still limited as they predominantly relies on the original pre-training objective (e.g., next-token prediction, masked-token prediction and diffusion). By serving foundation models as brain intelligence, AI agents equipped with action systems can autonomously engage with the practical environment and execute complex user intent. Moreover, action systems can support agents to harness available tools from the external environments or create tools from the internal sandbox, thus significantly extending the task scope of AI Agents. 
Therefore, the design of action systems will also determine the capability of AI agents in perception, decision making, execution, tool utilization, and any other components to align with the human brain. In other words, foundation models lay the groundwork for agents while action systems determine their ultimate potential to achieve complex targets. Designing an effective and comprehensive action system for AI agents is a critical endeavor that involves significant challenges and notable benefits. In Figure~\ref{fig:action-concepts}, we illustrate several key concepts in the action system. 
In this chapter, we discuss the human action system, and then examine the transition from human action to agentic action system in AI agents. After that, we systematically summarize the existing paradigms of action systems in AI agents, including action space, action learning, and tool learning. Finally, we analyze the differences between action and perception, and summarize the conclusion.

\section{The Human Action System}
\label{subsec:human_action_system}

\lettrine[lines=3]{\initfamily\textcolor{darkgreen}{A}}{ction system} in human cognition refers to the processes that allow humans to perceive, plan, and execute goal-directed actions. It is a complex system that enables individuals to interact with a dynamic environment, make decisions, and adapt their behavior based on feedback. Generally, the action system within human cognition could be broadly categorized as \emph{mental action} and \emph{physical action}:

\begin{itemize}[leftmargin=*]
    \item \textbf{Mental action} can be viewed as a kind of distinct action in human cognition system, which is formulated as a thinking process to drive the final intention in the human brain. For example, reasoning, decision making, imagining, and planning can all be considered as various types of mental action. In other words, mental actions are equal to a brain signal that drives the physical actions of humans to fulfill the final objective.
    \item \textbf{Physical action} refers to any goal-directed bodily movement executed by the human motor system. To some extent, physical actions are usually expressed as a kind of continuous action. For example, speaking, manipulating, drawing, running, and grasping can all be regarded as physical actions. Employing a sequence of physical actions, humans can conduct the interaction and collect feedback from real-world environments.
\end{itemize}

Figure~\ref{fig:human_action} illustrates a simple taxonomy of the human action system from the perspective of mental action and physical action. Empowered with both mental and physical actions, the human cognition system can handle diverse complex tasks from real-world scenarios. Drawing inspiration from human cognition, it is also essential for us to revisit how to formulate action systems in AI agents across different tasks, from language to digital and then in physical environments.

\begin{table}[!ht]
    \centering
    \caption{Definitions between different kinds of foundation models.}
    \resizebox{\textwidth}{!}{
        \begin{tabular}{l|l|l | l|l}
            \toprule
            \textbf{Model} & \textbf{Examples} &  \textbf{Inputs} & \textbf{Objective} & \textbf{Definition} \\
            \midrule
            \multirow{2}{*}{\textbf{Large Language Model (LLM)}}     & \multirow{2}{*}{GPT-4~\cite{openai2023chatgpt}} &  \multirow{2}{*}{Language} & \multirow{2}{*}{Next-Token Prediction} & LLM is to generate text based on the provided  \\
            & & &  & user prompts. \\
            \hline
            \multirow{2}{*}{\textbf{Large Multimodal Model (LMM)}}   & \multirow{2}{*}{LLaVA~\cite{liu2024llava}}     &  \multirow{2}{*}{Multi-modal} & \multirow{2}{*}{Multi-modal Generation} & LMM is to generate multimodal data based on \\
             & & & & multimodal inputs.\\
            \hline
            \multirow{2}{*}{\textbf{Robotic Foundation Model (RFM)}} & \multirow{2}{*}{RT-1~\cite{brohan2022rt}}       &  \multirow{2}{*}{Sensory inputs} & \multirow{2}{*}{Robotic Control} & RFM is to generate robotic control based on\\
             & & && the sensory inputs from dynamic environments. \\
            \hline
            \multirow{2}{*}{\textbf{Large Action Model (LAM)}}       & \multirow{2}{*}{LAM~\cite{Lu2024LAM}}           &  Interactive & \multirow{2}{*}{Executable Action} & LAM is to generate executable actions based on  \\
            & & Environment &  & the interactions within the environment.\\
            \bottomrule
        \end{tabular}
    }

    \label{tab:cmp_different_fms}
\end{table}

\begin{figure}[!ht]
\begin{center}
\begin{forest}
  for tree={
    draw,
    font = \sffamily\scriptsize\linespread{0.8}\selectfont,
    fill = customgreen!50,
    align = center,
    grow = south,
    forked edge,
    s sep = 4mm,
    l sep = 6mm,
    fork sep = 3mm,
  }
[Human Actions, fill=customblue!50
  [\textbf{Mental Actions}
    [\textbf{Cognitive}\\
      Reasoning\\
      Planning\\
      Reflection\\
      Imagination\\
      Decision-making
    ]
    [\textbf{Affective}\\
      Emotion Regulation\\
      Motivation\\
      Empathy
    ]
    [\textbf{Memory \& Learning}\\
      Memory Recall\\
      Skill Acquisition
    ]
  ]
  [\textbf{Physical Actions}
    [\textbf{Body Movements}\\
      Locomotion\\
      Gestures\\
      Posture Adjustment
    ]
    [\textbf{Object Use}\\
      Manipulation\\
      Assembly
    ]
    [\textbf{Communication}\\
      Speech \& Language\\
      Writing \& Typing\\
      Nonverbal \text{(e.g., Sign)}
    ]
  ]
]
\end{forest}
\end{center}
\caption{Illustrative taxonomy of human actions, showing both mental and physical facets.}
\label{fig:human_action}
\end{figure}

\section{From Human Action to Agentic Action}
\label{subsec:human_action_to_agentic_action}

\lettrine[lines=3]{\initfamily\textcolor{darkgreen}{I}}{n} the past long period of time, human action systems~\cite{kruger2007meaning} have significantly motivated us to shape the development of a computer system toward autonomous paradigms. The action mechanism plays a critical role in the human brain in driving goal-directed behavior. In an intelligent human brain ~\cite{Dosenbach2025BrainAction}, conscious and unconscious thinking signals are produced, converted into mental signals, which eventually lead to a sequence of action operations. This process can be mapped as a multi-stage pipeline that involves constructing action spaces, formulating learning mechanisms for improved decision making, and integrating external states (e.g., tools). Inspired by these principles, we discover that these designs are essential to formulate the prototype of AI agent.

Many existing frameworks incorporate action learning into their design or utilize it as an output. To clarify the definition of an action system, we highlight the distinctions among various frameworks, including large language models (LLM), large multi-modal models (LMM), robotic foundation models (RFM), and large action models (LAM), as shown in Table~\ref{tab:cmp_different_fms}. Specifically, an LLM is to produce language output based on provided prompts, while an LMM is to generate multi-modality artifacts based on the multi-modal inputs. Existing language-based or digital AI agent frameworks are built upon these foundation models (e.g., LLM or LMM) via predefining the scope of action space and its learning strategies. On the other hand, an RFM is to optimize robotic control based on real-world environments (e.g., robotic video). Existing RFMs are pre-trained from web-scale video data and use video prediction to simulate the action of robotic control. The core of RFM is still to use the generative objective to learn knowledge from large-scale data, although it has involved some action designs in building physical AI agents. Moreover, some recent works~\cite{Lu2024LAM} introduce the concept of large action model (LAM), which further highlights the stage to generate the action strategies, interact with real-world environments and enhance self-learning paradigm. From these definitions, we notice that, regardless of the foundational models employed, the core of action system is to build the interaction with the environment and then enable the learning process from the collected action trajectories via pre-defined reward functions. Specifically, the mechanisms underlying these behaviors are also similar to the action system in human cognition, offering valuable insights for designing action systems in AI agent frameworks. For example:
\begin{itemize}
    \item When processing different scenarios, humans usually will pre-define the action space to perform action trajectories to solve specific tasks. For instance, when playing computer games like Minecraft, we will set our action operations via keyboard or mouse to simulate behaviors like building house, mining gold, and so on. On the basis of this, we also need to build or create an action space for handling complex tasks in AI Agent frameworks.
    \item Compared to machines, the human cognitive system excels in continuously acquiring new knowledge through real-world interactions, guided by generating and optimizing the action sequences. Thus, replicating this learning ability in AI agents is essential to adapt the dynamic environment and build a new skill library.
    \item In addition, with the development of human civilization, learning to use external tools has been recognized as one of the most significant milestones in the evolution of human intelligence. By leveraging these external tools, humans can extremely extend the problem-solving capability in different scenarios, from the stone age to the industrial revolution.
\end{itemize}

\begin{table*}
\centering
\caption{Comparing the perception and action of human and AI agents.}
\label{tab:compare-percept-act}
\ra{1.3}
\begin{tabular}{p{3cm}p{4cm}p{4cm}p{4cm}}

\toprule
\textbf{Dimension} & \textbf{Human Brain / Cognition} & \textbf{LLM Agent} & \textbf{Remarks} \\
\midrule

\rowcolor{LightMint}
\textbf{Perception} &
- Integrates multiple sensory channels (vision, hearing, smell, touch, taste). \newline
- Perception closely tied to emotions, endocrine system, and physical state. \newline
- Highly sensitive, capable of detecting subtle differences. &

- Primarily language-based with some multimodal capabilities. \newline
- Perception depends on external sensors and models with limited integration. \newline
- Lacks real-time coupling with physical states. &

Perception differences lead to varying ways of understanding reality. Embodied AI attempts to bridge this gap but still faces both hardware and software challenges. \\

\textbf{Unified Representation} &
- Simultaneously processes multimodal inputs: vision, hearing, language, motion, and emotions. \newline
- Different brain regions collaborate to create unified spatiotemporal and semantic understanding. &

- Primarily text-based. Some multimodal models can process images or audio but with low integration. \newline
- No fully unified spatiotemporal modeling like the human brain. &

Even advanced multimodal models lack the human brain's holistic, unified representation capacity. Hardware and algorithmic challenges remain. \\

\rowcolor{LightMint}
\textbf{Granularity in Task Switching} &
- Flexible in shifting between macro and micro cognitive tasks. \newline
- Can plan at a high level and shift focus to finer details when needed. \newline
- Adjusts task priority and focus dynamically based on context and working memory. &

- Relies heavily on prompt engineering for granularity control. \newline
- Cannot autonomously reallocate attention between task layers. \newline
- May get stuck in a specific level of abstraction in absence of guided prompts. &

Humans can dynamically adjust cognitive granularity based on situational demands, while LLMs require explicit instruction to switch task focus effectively. \\

\textbf{Action} &
- Goal-oriented process drives multiple sensory to make decisions. \newline
- Real-time Learning from the experience via the environmental interaction. \newline 
- Encompass both physical activities and mental processes. &

- Action space need to be defined in advance. \newline 
- Unable to support actions in continuous spaces. \newline
- Relies on online training to optimize the decision-making process in the environment. \newline &

Humans are capable of actively learning new actions and performing continuous actions, whereas LLM agents currently lack this capability. \\

\bottomrule
\end{tabular}

\end{table*}

To this end, we expect to build the mapping between the action system of human cognition system and the design of AI Agent framework, including how to build action space for AI agent from specific scenarios to general domain, how to build action learning within the environment, and how to leverage external states (e.g., tools) to extend the task scope of AI Agent. By developing this a systematic survey,  we strive to provide more in-depth insights for the community with a clear understanding of the significance of action systems in AI agent frameworks. Table~\ref{tab:compare-percept-act} compares the perception and action systems between human and AI agents.

\begin{figure}[!t]
\centering
\footnotesize
    \begin{forest}
        for tree={
            forked edges,
            draw,
            rounded corners,
            node options={align=center,},
            s sep=6pt,
            calign=center,
            grow=east,
            reversed=true,
            anchor=base west,
            parent anchor=east,
            child anchor=west,
            base=left,
            font=\small,
            minimum width=2.5em,
          },
          where level=1{text width=5em,fill=customblue!50}{},
          where level=2{text width=5em,fill=customgreen!50}{},
        [Action System, fill=gray!20
            [Action Space, text width=50pt, for tree={
                calign=child edge, calign child=(n_children()+1)/2,
            }
                [Language, text width=50pt, for tree={
                    calign=child edge, calign child=(n_children()+1)/2,
                }
                    [Text, text width=45pt,
                        [
                            {
                                ReAct~\cite{yao2022react}, AutoGPT~\cite{gravitas2023auto}, \\ Reflexion~\cite{Shinn2023ReflexionLA}, LLM+P~\cite{liu2023llmp}
                            }, text width=199pt
                        ]
                    ]
                    [Code, text width=45pt,
                        [
                            {
                                MetaGPT~\cite{hong2023metagpt}, ChatDev~\cite{qian2024chatdev}, \\
                                SWE-Agent~\cite{yang2024sweagentagentcomputerinterfacesenable}, OpenDevin~\cite{openhands}
                            }, text width=199pt
                        ]
                    ]
                    [Chat, text width=45pt,
                        [
                            {
                                Generative Agents~\cite{park2023generative}, MetaGPT~\cite{hong2023metagpt}, \\
                                AutoGen~\cite{wu2023autogen}, ChatDev~\cite{qian2024chatdev}
                            }, text width=199pt
                        ]
                    ]
                ]
                [Digital, text width=50pt, for tree={
                    calign=child edge, calign child=(n_children()+1)/2,
                }
                    [Game, text width=45pt,
                        [
                            {
                                MineDojo~\cite{Linxi2022MineDojo}, Voyager~\cite{wang2023voyager}, \\
                                SwarmBrain~\cite{Xiao2024SwarmBrain}, JARVIS-1~\cite{wang2024jarvis}
                            }, text width=199pt
                        ]
                    ]
                    [Multimodal, text width=45pt,
                        [
                            {
                                MM-ReAct~\cite{yang2023mm}, ViperGPT~\cite{suris2023vipergpt}, \\
                                Visual-ChatGPT~\cite{wu2023visual}, HuggingGPT~\cite{shen2024hugginggpt}
                            }, text width=199pt
                        ]
                    ]
                    [Web, text width=45pt,
                        [
                            {
                                WebGPT~\cite{nakano2021webgpt}, WebShop~\cite{Shunyu2022WebShop}, \\
                                WebAgent~\cite{gur2024realworldwebagentplanninglong}, Mind2Web~\cite{deng2024mind2web}
                            }, text width=199pt
                        ]
                    ]
                    [GUI, text width=45pt,
                        [
                            {
                                Mobile-Agent~\cite{Junyang2024MobileAgent}, AppAgent~\cite{Chi2023AppAgent}, \\
                                UFO~\cite{Chaoyun2024UFO}, OmniParser~\cite{lu2024omniparser}
                            }, text width=199pt
                        ]
                    ]
                    [DB \& KG, text width=45pt,
                        [
                            {
                                UnifiedSKG~\cite{xie2022unifiedskg}, Pangu~\cite{gu2023pangu}, BIRD~\cite{dataset-bird}, \\
                                Spider 2.0~\cite{spider2},
                                Middleware~\cite{gu2024middleware}
                            }, text width=199pt
                        ]
                    ]
                ]
                [Physical, text width=50pt, 
                    [
                        {
                            RT-1~\cite{brohan2022rt}, RT-2~\cite{brohan2023rt}, RT-X~\cite{o2023open}, \\
                            GR-2~\cite{cheang2024gr}, $\pi_0$~\cite{black2024pi_0}, Saycan~\cite{Brian2022Saycan}, \\
                            VoxPoser~\cite{Wenlong2023VoxPoser}, EmbodiedGPT~\cite{Yao2023EmbodiedGPT},
                        }, text width=260pt
                    ]
                ]
            ]
            [Learning, text width=50pt, for tree={
                calign=child edge, calign child=(n_children()+1)/2,
            }
                [ICL, text width=50pt, for tree={
                    calign=child edge, calign child=(n_children()+1)/2,
                }
                    [Prompt, text width=45pt,
                        [
                            {
                                CoT~\cite{wei2022chain}, ReAct~\cite{yao2022react}, Auto-CoT~\cite{zhang2023automatic}, ToT~\cite{yaotree}, GoT~\cite{Besta2023GraphOT}, LearnAct~\cite{zhao2024empowering}, CoA~\cite{li2024improving}\\
                            }, text width=199pt
                        ]
                    ]
                    [Decompose, text width=45pt,
                        [
                            {
                                Least-to-Most~\cite{zhouleast}, HuggingGPT~\cite{shen2024hugginggpt}, \\
                                Plan-and-Solve~\cite{wang2023plan}, ProgPrompt~\cite{singh2023progprompt}
                            }, text width=199pt
                        ]
                    ]
                    [Role-play, text width=45pt,
                        [
                            {
                                Generative Agents~\cite{park2023generative}, MetaGPT~\cite{hong2023metagpt}, \\
                                ChatDev~\cite{qian2024chatdev}, 
                                SWE-Agent~\cite{yang2024sweagentagentcomputerinterfacesenable}
                            }, text width=199pt
                        ]
                    ]
                    [Refine, text width=45pt,
                        [
                            {
                                Reflexion~\cite{Shinn2023ReflexionLA}, Self-refine~\cite{madaan2024self-refine}, \\
                                GPTSwarm~\cite{zhuge2024gptswarm}
                            }, text width=199pt
                        ]
                    ]
                ]
                [PT \& SFT, text width=50pt, for tree={
                    calign=child edge, calign child=(n_children()+1)/2,
                }
                    [Pre-Train, text width=45pt,
                        [
                            {
                                RT-1~\cite{brohan2022rt}, RT-2~\cite{brohan2023rt}, RT-X~\cite{o2023open}, \\
                                GR-2~\cite{cheang2024gr}, LAM~\cite{Lu2024LAM}
                            }, text width=199pt
                        ]
                    ]
                    [SFT, text width=45pt,
                        [
                            {
                                CogACT~\cite{li2024cogact}, RT-H~\cite{belkhale2024rt}, OpenVLA~\cite{kim2024openvla}, \\
                                GR-2~\cite{cheang2024gr}, $\pi_0$~\cite{black2024pi_0}, UniAct~\cite{zheng2025universal}
                            }, text width=199pt
                        ]
                    ]
                ]
                [RL, text width=50pt, 
                    [
                        {
                            RLHF~\cite{ouyang2022training}, DPO~\cite{rafailov2023direct}, RLFP~\cite{ye2023reinforcement}, \\
                            ELLM~\cite{du2023guiding}, GenSim~\cite{wang2023gensim}, LEA~\cite{wang2024reinforcement}, \\
                            MLAQ~\cite{chaiempowering}, KALM~\cite{pangkalm},       When2Ask~\cite{hu2023enabling}, \\
                            Eureka~\cite{ma2023eureka}, ArCHer~\cite{zhou2402archer}, LLaRP~\cite{szot2023large}, GPTSwarm~\cite{zhuge2024gptswarm}
                        }, text width=260pt
                    ]
                ]
            ]
        ]
    \end{forest}
    \label{fig:action_system}
    \caption{A taxonomy of selected research works about action system, including action space and learning paradigm.}
\end{figure}

\section{Paradigms of Agentic Action System}
\label{subsec:paradigm_of_action_system}

\lettrine[lines=3]{\initfamily\textcolor{darkgreen}{G}}{enerally}, the action system of AI agent frameworks consists of three major components: 1) the action space $\cal{A}$, which includes all types of action that agent can perform in real-world scenarios or downstream tasks, and can vary significantly depending on different agent settings, ranging from language-based agents to embodied agents; 2) the action learning within an dynamic environment that determines the state $\cal{S}$, observation $\cal{O}$ and the optimization process of agent; 3) the tool space $\cal{T}$ that encompasses the instruments, interfaces, or middle-wares the agent can perform for utilization, which ranges from physical devices such as robotic arms to digital interfaces like APIs. Overall, these components collectively define the scope and characteristics of the action system for AI agents, shaping their formulation and execution.

To fully explore the possible actions \(a_t\) in practical scenarios, we must formally represent the action space and consider both individual operations and the underlying hierarchical reasoning processes. This means examining the action space at various levels, from low-level manipulations to high-level operators that orchestrate complex workflows.

Accordingly, the AI agent decision making process can be formalized as a trajectory $\langle o_t, s_t, a_t \rangle$,
where \(a_t\) is selected from the action space \(\mathcal{A}\) to transform the current state \(s_t\) based on observation \(o_t\) into the next state. In some cases, integrating external tool systems may also be necessary. By executing a sequence of \(\langle o_t, s_t, a_t \rangle\), the agent is steered toward achieving its final objectives.

\subsection{Action Space Paradigm}
\label{subsec:action-space-paradigm}

Action space $\cal{A}$ is an important component, which serves as the basis for building an action system within AI agent frameworks. The composition of the action space determines how AI agents solve complex tasks in different scenarios. 
In Figure~\ref{fig:action_system}, we present an illustrative taxonomy of the action system based on its action space. Generally, we summarize the action space within existing works as three distinct types, as outlined below.

\paragraph*{Language} Language-based AI agents typically operate through language-driven actions in interactive linguistic environments, such as reasoning, programming, retrieving information, executing API calls, or interacting with external tools. In our study, we summarize three distinct types of language-based action spaces, including plain text, code programming, and communication. Specifically, early language-based AI agents are built with plain text, which aim to perform interactive decision-making in verbal environments or text-based games. Here, ReAct~\cite{yao2022react} is a representative language-based AI agent, which synergizes the reasoning and actions of an LLM to solve various problems. AutoGPT~\cite{gravitas2023auto} analyzes and decomposes user requests into multiple subtasks and uses web search or other tools to tackle each of them. Reflexion~\cite{Shinn2023ReflexionLA} involves self-refinement and the memory mechanism to enhance action execution in language tasks. LLM+P~\cite{liu2023llmp} empowers LLM-based agent with planning capability to aid decision-making. However, converting plain text into an executable command usually requires LLMs to first interpret the text and then perform instruction conversion, leading to additional information loss. To this end, some work explores using code as the action space, allowing direct execution of the generated code and self-verification. MetaGPT~\cite{hong2023metagpt} and ChatDev~\cite{qian2024chatdev} build the action space via programming language with multi-agent collaboration. SWE-Agent~\cite{yang2024sweagentagentcomputerinterfacesenable} consider different stages of software engineering and thus solve software issues. OpenDevin~\cite{openhands} devises an automatic software development platform that integrate code writing, interaction with the command, sandbox for code execution, and collaborations. Moreover, some frameworks are built based on multi-agent communications, and then use chatting to analyze which actions should be employed in the next step. Here, Generative Agents~\cite{park2023generative} directly simulate multiple characters in a virtual town, to explore how each agent to conduct next action. MetaGPT~\cite{hong2023metagpt} and ChatDev~\cite{qian2024chatdev} are both multi-agent frameworks to faciliate the development of software engineering. AutoGen~\cite{wu2023autogen} is also a representative framework that enable multiple agent collaboration to solve any complex tasks. Generally, language-based AI agents, empowered by LLMs, perform effectively in linguistic interactions. However, limited to the scope of the action space, it also poses challenges of how to solve more complex tasks in real-world scenarios. Therefore, we also need to formulate new research solutions to construct a more sophisticated action space to solve challenging tasks.

\paragraph*{Digital} To expand the capabilities of AI agents beyond language, some works have also developed advanced AI agents that operate within digital environments, such as web proxies, online shopping platforms, and gaming systems. For examples, MineDojo~\cite{Linxi2022MineDojo} devises a virtual agent via video-language pre-training and simulates an environment that supports a multitude of tasks and goals within Minecraft. Moreover, Voyager~\cite{wang2023voyager} is an embodied AI agent trained to play Minecraft. It simulates multiple executable actions in code form to develop a skill library via interacting with the Minecraft environment, and thus improve the capability of virtual agents. JARVIS-1~\cite{wang2024jarvis} is an open-world agent that can handle multi-modal inputs / outputs, generate sophisticated plans, and perform embodied control. It explores the evolutionary behaviors of the agent when acting in Minecraft. SwarmBrain~\cite{Xiao2024SwarmBrain} is an embodied agent that uses LLMs to act strategically and in real time in StarCraft II. Additionally, some research studies investigate how LLMs can act to process multimodal tasks. MM-ReAct~\cite{yang2023mm} and ViperGPT~\cite{suris2023vipergpt} apply LLMs to perform the thinking process for multimodal tasks and then select visual experts for task solving. Visual-ChatGPT~\cite{wu2023visual} integrates multiple visual experts and uses LLMs as the controller to solve tasks. HuggingGPT~\cite{shen2024hugginggpt} directly involves four stages, including task planning, model selection, model execution and response generation, to automatically analyze user instructions and predict the final answers based on complex multimodal tasks. It is also vital for the agent to keep up with the latest information available online. Therefore, some AI Agent frameworks (e.g., WebGPT~\cite{nakano2021webgpt}, WebAgent~\cite{gur2024realworldwebagentplanninglong}) are designed to interact with search engine to enhance the capability of agent to discover the answers from website. WebShop~\cite{Shunyu2022WebShop} is used to explore the potential of AI Agent for online shoping. Mind2Web~\cite{deng2024mind2web} is to build a generalist agent that simulate multiple complex web tasks. As foundation agents advance in processing multimodal tasks or web tasks, there is a increasing trend to enhance their capability in solving complex computer tasks. Mobile-Agent~\cite{Junyang2024MobileAgent} utilizes multimodal models as the cognitive controller to manage and orchestrate mobile functionalities. AppAgent~\cite{Chi2023AppAgent} defines various app usages as action spaces, enabling foundation models to interact with different apps as a mobile intelligent assistant. UFO~\cite{Chaoyun2024UFO} and OmniParser~\cite{lu2024omniparser} are two advanced GUI agents which manipulates UI operations as the action space, enabling AI agent to perform computer-use tasks. Generally, empowered with more advanced skills in digital environment, AI agent can demonstrate better intelligent in solving complex tasks, and represent a significant shift from language intelligent to digital intelligent. By expanding the action space to include web browsing, GUI interaction, mobile applications, and embodied systems, AI agents are evolving into more autonomous, multimodal, and context-aware systems, bridging the gap between foundation models and human cognition systems.
In addition, other research explores LLM integration with structured digital environments such as relational databases and knowledge graphs (KGs). Pangu~\cite{gu2023pangu} pioneered the connection between LLMs and large-scale KGs, while BIRD~\cite{dataset-bird} and Spider 2.0~\cite{spider2} established a foundation for LLMs to operate with enterprise databases in real-world settings. 
NL2SQL-BUGs~\cite{dataset-llmsql} addresses the critical challenge of identifying semantic errors in NL2SQL pipelines~\cite{alphasql}, which enhances the reliability of LLM-driven interactions with relational databases~\cite{DBLP:journals/vldb/QinCLZTLFYO22}.
Similarly, frameworks like UnifiedSKG~\cite{xie2022unifiedskg} and Middleware~\cite{gu2024middleware} expand LLMs' action capabilities across both databases and KGs.

\paragraph*{Physical} Building an AI agent to interact with the real physical world can be viewed as the ultimate objective to simulate a computer program to act as a human cognition system. To achieve this, we require the agent to be capable of processing signals from real-world environments and generating feedback to facilitate continuous improvement. Therefore, it will pose new challenges on how to process the continuous signals collected by sensors and enable foundation models to make decisions. To fulfill this, RT-family~\cite{brohan2022rt, brohan2023rt, o2023open} pre-trained vision-language-action models to integrate knowledge from web videos into robotic learning, enhancing robotic control and action execution. GR-2~\cite{cheang2024gr} is a robotic model that undergoes large-scale pre-training on video clips and language data, followed by fine-tuning on robot trajectories for robotic action prediction. $\pi_0$~\cite{black2024pi_0} pre-trained a robotic model based on robot platforms, including single-arm robots, dual-arm robots, and mobile manipulators, to build robotic learning in physical systems. SayCan~\cite{Brian2022Saycan} bridges the connections between robotic semantics and LLMs, using the robotic model to provide perception for LLMs and then using LLMs to make high-level decision-making. VoxPoser~\cite{Wenlong2023VoxPoser} uses LLMs to understand and decompose 3D Value Maps for Robotic Manipulation. Besides, EmbodiedGPT~\cite{Yao2023EmbodiedGPT} utilizes vision-language models to understand video data and perform decision-driven actions. 
In physical environments, it is worth noting that we usually need to understand continuous signals and then generate continuous actions for robotic control. Despite the existing foundation models that can effectively process discrete-level actions (e.g., language or computer-use), how to process long continuous signals is still challenging. Therefore, eliminating the differences between continuous signals and discrete signals in foundation models is still a major problem.

Generally, action space serves as one of the most critical components in building an effective AI Agent system. An effective action space enhances the capability and efficiency of the AI Agent in processing downstream tasks. Action space usually ranges from the discrete space (e.g., skill library in Atari games) to the continuous space (e.g., robotic manipulation). As AI agents become more autonomous and multimodal, designing effective action spaces will be crucial for advancing general-purpose AI systems capable of real-world interactions.

\subsection{Action Learning Paradigm}
\label{subsec:action-learning-paradigm}

In the human cognition system, action learning~\cite{revans2017abc} represents the problem-solving process, involving both taking actions and reflecting on feedback. Similarly, action learning for AI agents refers to the iterative process by which an autonomous AI system refines its decision making and behavior through direct interaction with the real world environment. Generally, action learning encompasses a cycle of multiple stages, including building action space, choosing actions, and optimizing action selection based on interaction with the environment (e.g., receiving feedback or rewards and adjusting policy for choosing actions). By iteratively deploying these strategies, AI agents can adapt to the latest information or changing conditions in real time, ultimately enabling more robust, flexible, and efficient problem-solving capabilities. Therefore, an effective action learning mechanism is crucial for the optimization of agentic action systems. In this part, we mainly focus on three different representative learning paradigms, including in-context learning, supervised training, and reinforcement learning, which are discussed below:

\paragraph*{In-context Learning} As large language models have demonstrated emergent ability, in-context learning has been considered as the most effective method to leverage the existing capabilities of LLM without any modifications. Provided with well-designed prompts to describe actions, AI agents can understand specific actions, perform these actions, reflect on the outcome of the interaction with the environment, and finally achieve goals.
Among these approaches, the common method is to use prompting techniques to instruct LLMs to generate agentic action. Here, the most representative one is Chain-of-Thought (CoT)~\cite{wei2022chain} prompting, which applies ``\emph{Let us think step by step}'' technique to generate a sequence of intermediate reasoning steps, exploring potential solutions systematically. ReAct~\cite{yao2022react} enables LLMs to generate reasoning trails and task-specific actions through interaction within the environment, improving the reasoning and decision-making capabilities of AI agents.  LearnAct~\cite{zhao2024empowering} devises an iterative learning strategy to expand action space by generating code (i.e., Python) to create and revise new actions. Moreover, some works (e.g., Auto-CoT~\cite{zhang2023automatic} explores how to automatically generate CoT via LLMs and then enable the autonomous thinking process of AI agents. To handle more complex tasks, ToT~\cite{yaotree} considers the thought process as a tree structure and introduces the tree search via LLM prompting, while GoT~\cite{Besta2023GraphOT} applies a graph structure along with the graph search. For robotic models, CoA~\cite{li2024improving} designed four different prompt settings (e.g., object, grasp, spatial, and movement) to allow robot manipulation with reasoning process. Furthermore, to tackle more complex tasks that require intricate agentic workflows, some frameworks introduce the stage of task decomposition via LLM prompting to break down user instructions. Least-to-Most~\cite{zhouleast} is a classical prompting technique to convert user instructions into multiple subtasks. HuggingGPT~\cite{shen2024hugginggpt} is a representative AI agent framework that applies task planning to transform user requirements into actionable items. Plan-and-Solve~\cite{wang2023plan} directly uses LLM to make plans from user instructions and then give answers based on the generated plans. Progprompt~\cite{singh2023progprompt} applies similar task decomposition to robotic tasks. In addition, using prompting techniques to formulate the characteristic of AI agent has also been considered as an increasing trend to facilitate the simulation and productivity of AI agent frameworks (e.g., Generative Agents~\cite{park2023generative}, MetaGPT~\cite{hong2023metagpt}, ChatDev~\cite{qian2024chatdev}, SWE-Agent~\cite{yang2024sweagentagentcomputerinterfacesenable}). Finally, some other frameworks (e.g., Reflexion~\cite{Shinn2023ReflexionLA} or Self-refine~\cite{madaan2024self-refine}) analyze the external feedbacks of user interaction within the environment and then iteratively refine and polish results via well-designed reflexion prompts. All of these designs allow us to better understand user instructions, decompose task goals, and make plans for thinking answers. In-context learning can help us avoid parameter optimization and reduce the heavy cost of training LLMs. It allows AI agents to perform various actions effectively and adapt to a wide range of domains. However, challenges still remain if we want to acquire agents of even stronger action learning ability.

\paragraph*{Supervised Training} To further improve the action learning ability of foundation models, increasing research efforts have focused on training methodologies, including self-supervised pretraining (PT) and supervised fine-tuning (SFT). For the pre-training paradigm, the most representative works is RT-family~\cite{brohan2022rt, brohan2023rt, o2023open}, which pre-trains robotic Transformer on large-scale web and robotic data, yielding a powerful vision-language-action model. Following this policy, GR-2~\cite{cheang2024gr} is developed through extensive pre-training on a large corpus of web videos to understand the dynamics of the world and post-training on robotic trajectory data to specialize in video generation and action prediction. Similarly, LAM~\cite{Lu2024LAM} is a large action model pre-trained on trajectories of user interaction with computer usage. However, the pre-training paradigm usually incurs massive computation costs. Therefore, many works take the fine-tuning paradigm to enhance the action capability of foundation models. OpenVLA~\cite{kim2024openvla} is built upon the Llama2~\cite{touvron2023llama} language model that integrates a visual encoder based on DINOv2~\cite{oquab2023dinov2} and SigLIP~\cite{zhai2023sigmoid}. It collects a diverse set of real-world robot demonstrations from Open X-Embodiment (OXE)~\cite{o2024open} for fine-tuning and outperforms RT-2-X~\cite{o2024open} across different tasks, while utilizing 7$\times$ fewer parameters. Building upon OpenVLA, CogACT~\cite{li2024cogact} further integrate additional diffusion action module to enhance the model capability in conducting action operation over robotic tasks.
Besides, some works also explore how to enable robotic model to capture action from plain language in physical world. For examples, RT-H~\cite{belkhale2024rt} introduces a hierarchical architecture to build action space, which first predict language motions and then generate low-level actions.  And $\pi_0$~\cite{black2024pi_0} collected massive diverse datasets from different dexterous robot platforms, and then fine-tune the pre-trained VLMs to learn robotic actions. UniAct~\cite{zheng2025universal} learns universal actions that capture generic atomic behaviors across differently shaped robots by learning their shared structural features. This approach achieves cross-domain data utilization and enables cross-embodiment generalizations by eliminating heterogeneity~\cite{teng2025atomthoughtsmarkovllm}. Overall, using supervised training, including pre-training and supervised fine-tuning, can effectively adapt foundation models to perform actions intelligently in real-world scenarios. Last but not least, it is worth noting that, even with extensive training on a vast corpus, it is still beneficial to apply in-context learning on top of the trained model for AI agents, in an pursuit for their best performance.

\paragraph*{Reinforcement Learning}
To facilitate an action learning procedure in addition to in-context learning and supervised training, it is also crucial for agents to interact with the environment and eventually optimize their action policy through experience, feedback, or rewards. Considering this iterative and sequential nature, reinforcement learning (RL) provides us with the systematic methodology we need ~\cite{sutton1998reinforcement, lee2012neural,rl_features,rl_ml}. In RL paradigms, there are several classical and representative algorithms, such as Deep Q-Network (DQN)~\cite{mnih2013playing} and Proximal Policy Optimization (PPO)~\cite{schulman2017proximal}. The most representative RL work that applied reinforcement learning to foundation models is InstructGPT~\cite{ouyang2022training}, which effectively aligns LLM outputs with human preferences via RLHF. Since RLHF usually requires additional training to build the reward model, some papers (e.g. DPO~\cite{rafailov2023direct}) proposes to directly optimize preference data through contrastive learning. Existing work~\cite{guo2025deepseek, Kimi2025Scaling} also demonstrate the potential of scaling the RL algorithm for foundation models to produce long CoT thinking stages with impressive performance. A key technique contributing to these achievements is Reinforcement Learning with Verifiable Reward (RLVR), which utilizes the rule-based outcome reward to apply RL on the LLM~\cite{zeng2025simplerl,xie2025logic,wang2025reinforcement}. 
With the rule-based reward function, current works such as ReSearch~\cite{chen2025learning}, L0~\cite{zhang2025l0reinforcementlearninggeneral}, Agent-R1~\cite{Agent-R1}, RAGEN~\cite{RAGEN} and GiGPO~\cite{feng2025group} are exploring utilizing RL to fine-tune the agents on trajectory data in agentic environments. 
Specifically, these approaches enable multi-turn interaction between agents and their environments, collecting feedback to iteratively refine the reasoning policy.
Although RL paradigms have been successfully used to fine-tune LLMs for text generation tasks ~\cite{bai2022training,ramamurthy2022reinforcement,ouyang2022training,hu2023aligning}, efficiently utilizing the RL algorithm for action learning remains one of the many challenges that require further attempts. Recent advances indicate significant progress in applying RL to action learning with LLMs from various perspectives:
\begin{itemize}
    \item Given the rich world knowledge encapsulated in LLM, we can use LLM to \textit{mimic external environments or generate imagined trajectories} to aid agents in action learning. For instance, RLFP~\cite{ye2023reinforcement} utilizes guidance and feedback from the policy, value, and success-reward foundation models to enable agents to explore more efficiently. Similarly, ELLM~\cite{du2023guiding} utilizes large-scale background knowledge from LLMs to guide agents in efficient exploration within various environments. GenSim~\cite{wang2023gensim} automatically generates rich simulation environments and expert demonstrations by exploiting the coding abilities of LLM, thereby facilitating the capability of the agent for free exploration. LEA~\cite{wang2024reinforcement} leverages the language understanding capabilities of LLM and adapts LLM as a state transition model and a reward function to improve the performance of offline RL-based recommender systems. MLAQ~\cite{chaiempowering} utilizes an LLM-based world model to generate imaginary interactions and then applies Q-learning~\cite{watkins1992q} to derive optimal policies from this imaginary memory. KALM~\cite{pangkalm} fine-tunes LLM to perform bidirectional translations between textual goals and rollouts, allowing agents to extract knowledge from LLM in the form of imaginary rollouts through offline RL. In general, empowered by RL paradigms, we can significantly explore the internal knowledge from LLMs and thus enhance the interactions with external environments. 
    
    \item Besides, hierarchical RL is also a promising topic that helps foundation model to decompose complex task and then learn optimal policies to solve each task via RL paradigm. For example, When2Ask~\cite{hu2023enabling} enables agents to request high-level instructions from LLM. The high-level LLM planner provides a plan of options, and the agent learns the low-level policy based on these options. Eureka~\cite{ma2023eureka} leverages LLM to generate human-level reward functions with reflection, allowing agents to efficiently learn complex tasks such as anthropomorphic five-finger manipulation. ArCHer~\cite{zhou2402archer} adopts a hierarchical RL approach, utilizing an off-policy RL algorithm to learn high-level value functions, which in turn implicitly guide the low-level policy. LLaRP~\cite{szot2023large} leverages LLM to comprehend both textual task goals and visual observations. It employs an additional action output module to convert the output of the LLM backbone into a distribution over the action space. Overall, using hierarchical RL can guide AI Agent to explore optimal strategies when analyzing user requests for reasoning and planning.
\end{itemize}

Using reinforcement learning, we can integrate foundation models with online learning from interactive environments, incorporating both action policies and world models. This integration enables advanced action systems in AI agents. Within the reinforcement learning paradigm, agents dynamically adapt and refine their decision-making processes in response to external feedback, facilitating greater efficiency and effectiveness in action learning and achieving desired outcomes.

\paragraph*{Summary}
In general, Empowered by action systems, AI agents have demonstrated significant decision-making capabilities across various fields. For example, action learning enables AI agents to automate the understanding of Graphical User Interfaces (GUIs) and perform various operations, thereby improving human productivity through automatic computer usage. Moreover, several studies have shown that AI agents equipped with action systems can achieve remarkable outcomes in robotic manipulation tasks, such as object picking, laundry folding, and table cleaning. There are also promising research directions in the industry employing action models. For instance, autonomous driving (AD) has attracted considerable attention due to the exceptional performance of VLMs in perception and decision-making. By integrating human understanding through foundation models, AD systems can effectively comprehend real-world surrounding, enabling them to simulate human-level drivers. In summary, action learning endows agents with the ability to interact with the external world, thereby creating more opportunities for AI applications in real-world scenarios.

\begin{figure}[t]
\centering
\footnotesize
    \begin{forest}
        for tree={
            forked edges,
            draw,
            rounded corners,
            node options={align=center},
            s sep=6pt,
            calign=center,
            grow=east,
            reversed=true,
            anchor=base west,
            parent anchor=east,
            child anchor=west,
            base=left,
            font=\small,
            minimum width=2.5em,
          },
          where level=1{text width=5em,fill=customblue!50}{},
          where level=2{text width=3em,fill=customgreen!50}{},
        [Tool System, fill=gray!20
            [Types, for tree={calign=child edge, calign child=(n_children()+1)/2}
                [Language, text width=40pt, 
                    [
                        {
                            ToolFormer~\cite{schick2023toolformer}, ToolLLM~\cite{qin2023toolllmfacilitatinglargelanguage}, Gorilla~\cite{patil2024gorilla}, \\
                            ToolkenGPT~\cite{Shibo2023ToolkenGPT}, 
                            GPT4tools~\cite{dataset-gpt4tools}, AnyTool~\cite{du2024anytoolselfreflectivehierarchicalagents}
                        }, text width=280pt
                    ]
                ]
                [Digital, text width=40pt, 
                    [
                        {
                            MM-ReAct~\cite{yang2023mm}, ViperGPT~\cite{suris2023vipergpt}, Visual ChatGPT~\cite{wu2023visual}, \\
                            HuggingGPT~\cite{shen2024hugginggpt}, Chameleon~\cite{Pan2023Chameleon}, WebGPT~\cite{nakano2021webgpt}, \\
                            WebAgent~\cite{gur2024realworldwebagentplanninglong}, 
                            Mobile-Agent~\cite{Junyang2024MobileAgent}, AppAgent~\cite{Chi2023AppAgent}, 
                            Middleware~\cite{gu2024middleware} 
                        }, text width=280pt
                    ]
                ]
                [Physical, text width=40pt, 
                    [
                        {
                            RT-2~\cite{brohan2023rt}, TidyBot~\cite{Wu_2023}, SayCan~\cite{Brian2022Saycan}, SayPlan~\cite{rana2023sayplan}
                        }, text width=280pt
                    ]
                ]
                [Scientific, text width=40pt, 
                    [
                        {
                            HoneyComb~\cite{zhang2024honeycombflexiblellmbasedagent}, ChemCrow~\cite{Andres2024ChemCrow}, \\
                            SciToolAgent~\cite{chen2025scitoolagent}, SciAgent~\cite{Yubo2024SciAgent}
                        }, text width=280pt
                    ]
                ]    
            ]
            [Learning, for tree={calign=child edge, calign child=(n_children()+1)/2}
                [Tool Discovery, text width=60pt, 
                    [
                        {
                            HuggingGPT~\cite{shen2024hugginggpt}, 
                            Gorilla~\cite{patil2024gorilla}, 
                            ToolFormer~\cite{schick2023toolformer}, \\
                            ToolLLM~\cite{qin2023toolllmfacilitatinglargelanguage},
                            ToolkenGPT~\cite{Shibo2023ToolkenGPT}, 
                            ToolChain~\cite{zhuang2023toolchain}
                        }, text width=260pt
                    ]
                ]
                [Tool Creation, text width=60pt, 
                    [
                        {
                            PAL~\cite{Luyu2023PAL}, LATM~\cite{cai2023large}, Creator~\cite{qian2023creator}, \\
                            MetaGPT~\cite{hong2023metagpt}, SWE-Agent~\cite{yang2024sweagentagentcomputerinterfacesenable}
                        }, text width=260pt
                    ]
                ]
                [Tool Usage, text width=60pt, 
                    [
                        {
                            HuggingGPT~\cite{shen2024hugginggpt}, TPTU~\cite{ruan2023tptulargelanguagemodelbased}, SayCan~\cite{Brian2022Saycan}, \\
                            ReTool~\cite{Jiazhan2025Retool}, StepTool~\cite{Yuanqing2024Step}, Tool-Star~\cite{Guanting2025ToolStar}
                        }, text width=260pt
                    ]
                ]
            ]
        ]
    \end{forest}
    \label{fig:tool_system}
    \caption{A taxonomy of tool systems in AI agents, including tool category and learning paradigm.}
\end{figure}

\subsection{Tool-Based Action Paradigm}
\label{subsec:tool-based-action-paradigm}

Tool learning distinguishes human intelligence from that of other animals. Ever since the Stone Age, human use of tools has boosted efficiency, productivity, and innovation. Similarly, enabling AI agents to operate in digital and physical environments by harnessing various tools is a fundamental step toward achieving human-level intelligence.

\paragraph*{Definitions}
In AI, tools are defined as interfaces, instruments, or resources that allow agents to interact with the external world. Examples include web search~\cite{nakano2021webgpt, qin2023webcpm, deng2024mind2web, gur2024realworldwebagentplanninglong}, databases~\cite{liu2023agentbench, zhou2023dbotdatabasediagnosisusing,nl2sqlsurvey,nl2sql360}, coding environments~\cite{wang2024executable}, data systems~\cite{vissurvey,deepeye,sqldata}, and weather forecasting~\cite{10.1145/3704435}. By translating tool functionality into plain text or API formats, foundation models can expand their problem-solving scope. The evolution of tool systems in AI can be summarized in stages. Initially, with the advent of large language models~\cite{brown2020language}, the focus was on converting tools into explainable formats (e.g., function calls). Later, advances in multimodal processing shifted interactions from conversational chats to graphical user interfaces (GUIs), and more recent work has explored embodied agents that control hardware (e.g. robotic arms, sensors) to interact with the physical world. To simplify, a tool-based action can be considered a form of external action employed for assistance.

\paragraph*{Tool Category}
Similar to action spaces, tools can also be classified into multiple categories according to their types. In this part, we mainly summarize three key domains, including language, digital, and physical. In addition, we also explore the potential of tool learning in emerging areas such as scientific discovery:

\begin{itemize}
    \item \textit{Language:} To facilitate the use of external tools, we usually denote the tool as a kind of function call for foundation models, which usually encompasses task descriptions, tool parameters, and corresponding outputs. This expression allows LLMs to understand when and how to use tools in AI agents. Specifically, ToolFormer~\cite{schick2023toolformer} expands the capabilities of language models by integrating external tool spaces, including calculator, QA systems, search engine, translation, and calendar. ToolLLM~\cite{qin2023toolllmfacilitatinglargelanguage} uses RapidAPI as the action space and then uses a depth-first search-based decision tree algorithm to determine the most suitable tool for solving tasks. Gorilla~\cite{patil2024gorilla} is a fine-tuned LLM based on the tool documents and then can be used to write API calls. ToolkenGPT~\cite{Shibo2023ToolkenGPT} is to optimize tool embeddings and then enable LLMs to retrieve tools from the fine-tuned tool embeddings. GPT4tools~\cite{dataset-gpt4tools} and AnyTool~\cite{du2024anytoolselfreflectivehierarchicalagents} are also building self-instruct datasets and then fine-tune LLMs on them for tool usage. Generally, due to the impressive capability of LLMs, language-based tool utilization for AI agents has been studied, with its effectiveness validated in abundant works, ranging from plain text or function calls to code programming.
    \item \textit{Digital:} With the success of LLMs in processing language information, many researchers are exploring extending the task scope of AI agents from the language to the digital domains (e.g., MultiModal, Web search, GUI, and so on). For example, MM-ReAct~\cite{yang2023mm}, ViperGPT~\cite{suris2023vipergpt}, and Visual ChatGPT~\cite{wu2023visual} employed LLMs as the controller and then used LLMs to select visual experts for solving different tasks. HuggingGPT~\cite{shen2024hugginggpt} and Chameleon~\cite{Pan2023Chameleon} use LLMs to first conduct reasoning and planning actions and then analyze which multimodal tools should be used for solving user instructions. WebGPT~\cite{nakano2021webgpt} and WebAgent~\cite{gur2024realworldwebagentplanninglong} respectively empowered LLMs with search engines to enhance the capability of LLMs to solve more challenging tasks.  Mobile-Agent~\cite{Junyang2024MobileAgent} and AppAgent~\cite{Chi2023AppAgent} respectively incorporate GUI manipulations and App usage as the tool-based actions to extend the task scope of AI agents in solving mobile phone tasks. In contrast to the physical world, digital environments usually provide simpler pipelines to collect and process data. By involving foundation models and their interaction with the digital environment, it is possible for us to develop intelligent assistants in computers, mobile phones, and other digital devices.
    
    \item \textit{Physical:} For physical world applications, RT-2~\cite{brohan2023rt} demonstrates language-guided robotic manipulation using visual-language tools, and TidyBot~\cite{Wu_2023} shows how LLMs adapt cleaning tools to personalized household preferences. SayCan~\cite{Brian2022Saycan} uses LLMs as the cognitive system to guide robots in solving tasks through robotic arms and visual perception. SayPlan~\cite{rana2023sayplan} built a 3D scene graph as the action spaces and designed multiple actions and tools for 3D simulation, and then used LLMs as planners to invoke these actions or tools for robot task planning. Besides, specialized applications in real-world scenarios now also proliferate across different domains. For instance, in surgical robotics, \cite{Zargarzadeh_2025} presents a multi-modal LLM framework for robot-assisted blood suction that couples high-level task reasoning, enabling autonomous surgical sub-tasks. Some autonomous driving systems \cite{yang2023survey, mao2023language} also integrate vision–language models with vehicle control tools for explainable navigation. In total, physical world applications pose the most significant challenge when compared to other tasks, but they also offer the biggest industrial value. Therefore, it still requires us to continue exploring advanced action learning and tool integration in physical-based agents in the future.

    \item \textit{Scientific:} Scientific tools have played a transformative role in advancing AI agents across disciplines, enabling them to learn, adapt, and execute tasks while integrating foundational models with frameworks that drive innovation and address complex challenges. In materials science, HoneyComb~\cite{zhang2024honeycombflexiblellmbasedagent} exemplifies tool-driven advancements with its ToolHub. General Tools provide dynamic access to real-time information and the latest publications, effectively bridging gaps in static knowledge bases. Material Science Tools are designed for computationally intensive tasks, leveraging a Python REPL environment to dynamically generate and execute code for precise numerical analysis. Similarly, ChemCrow~\cite{Andres2024ChemCrow} demonstrates the transformative power of tools in chemistry by integrating GPT-4 with 18 expert-designed tools to automate complex tasks such as organic synthesis, drug discovery, and materials design. These tools include OPSIN for IUPAC-to-structure conversion, calculators for precise numerical computations, and other specialized chemistry software that enables accurate reaction predictions and molecular property evaluations. Similarly, SciToolAgent~\cite{chen2025scitoolagent} showcases how multi-tool integration can revolutionize scientific research. Designed to address the limitations of existing systems, SciToolAgent integrates over 500 tools (e.g., Web API, ML models, function calls, databases, and so on). Finally, SciAgent~\cite{Yubo2024SciAgent} exemplifies a multi-agent framework that integrates ontological knowledge graphs with specialized agents for hypothesis generation and critical analysis, emphasizing the power of modular, tool-driven systems to accelerate discovery in materials science and beyond. These examples underscore the transformative potential of integrating specialized tools into AI frameworks to address domain-specific challenges effectively.
\end{itemize}

\paragraph*{Tool learning}
Inspired by human evolution~\cite{washburn1960tools}, the integration of tools in AI involves three key aspects: \emph{Tool Discovery} (identifying suitable tools), \emph{Tool Creation} (developing new tools) and \emph{Tool Usage} (effectively employing tools). We also systematically review existing literature and summarize them in the following:

\begin{itemize}
    \item \textbf{Tool Discovery:} In real-world environments, there is a wide range of tools from the digital to the physical world. Finding the most appropriate tools for user instructions can be challenging. Therefore, the process of tool discovery is to identify and select the appropriate tools that AI agents can operate on to achieve their objectives. This stage also requires the world models in AI agents to have a profound understanding of any complex user instructions and world knowledge of different tools. Moreover, the versatility of AI agents is also correlated with its ability to operate diverse tool systems. Generally, tool discovery can be categorized into two mainstream paradigms: retrieval-based and generative-based methods. Retrieval-based methods aim to select the most relevant tools from the tool library. For example, HuggingGPT~\cite{shen2024hugginggpt} introduces a framework in which LLMs act as controllers, orchestrating task planning and then invoking suitable models from platforms such as Hugging Face to fulfill user intention. In generative-based approaches, we often fine-tune LLMs to learn how to use and select tools based on various user instructions. For instance, ToolFormer~\cite{schick2023toolformer} collects a massive corpus with the corresponding API calls (e.g., calculator, QA system, search engines, translation, and calendar) for training. ToolLLM~\cite{qin2023toolllmfacilitatinglargelanguage} collect tool instructions based on solution paths and then fine-tune Llama models to generate better API calls for tool utilization.
    \item \textbf{Tool Creation} In addition to using existing tools, the ability to create new tools plays a crucial role in human civilization. For language agents, a widely adopted approach is to use LLMs to generate functions as executable programs, which consist of both the code and documentation. For example, PAL~\cite{Luyu2023PAL} generates programs as intermediate reasoning steps to solve problems, LATM~\cite{cai2023large} or Creator~\cite{qian2023creator} use LLMs to create code for user intentions, and to further design a verifier to validate the created tools. SciAgent~\cite{Yubo2024SciAgent} not only integrates multiple scientific tools but also crafts new tools for scientific discovery. More details on tool creation from an optimization perspective can be found in Section~\ref{subsubsec:toolcreation}.
    \item \textbf{Tool Usage} After collecting or creating tools, the effective use of tools constitutes the cornerstone of the capabilities of AI agents, allowing applications that bridge virtual and physical worlds. Modern AI agents increasingly employ tools to tackle complex tasks across diverse domains, with three key dimensions of expansion: 1) \textit{Vertical Specialization}: Agents leverage domain-specific tools to achieve professional-grade performance in complex fields such as robotics, science, and healthcare; 2) \textit{Horizontal Integration}: Systems combine multiple toolkits across modalities (vision, language, control) for multimodal problem-solving; 3) \textit{Embodiment}: Agents physically interact with environments through robotic tools and sensors.
\end{itemize}

To summarize, tool learning and action learning constitute the two most important components of the action system in AI agents. Tool learning can be considered as a kind of action to use external states for problem-solving. Tool learning enables AI agents to substantially broaden their range of tasks, pushing the boundaries beyond the scope of foundation models. For example, empowered by API or function calls, language models can directly reuse the capability of existing models (e.g., retrieval, coding, web search) to generate answers, rather than next-token prediction~\cite{VerifAI}. Tool learning also involves multiple challenging stages, including how to determine the tool space, how to discover and select tools, and how to create and use tools. Overall, tool learning plays a pivotal role in building an omnipotent AI agent framework to solve complex tasks in different domains.

\subsection{Agent Learning Paradigm} 
\label{subsec:agent-learning-paradigm}

The key to further enhance an LLM agent's capacity for action or tool usage lies in interaction with external environments rather than training on static datasets. Therefore, the evolution of LLM agents has entered a new phase—shifting from static information processors to dynamic, interactive agents.
RL has emerged as the leading methodology for training such agents, enabling them to learn optimal policies through interaction and feedback, moving beyond the limitations of supervised learning on fixed datasets.

Building on the success of RLVR~\cite{zeng2025simplerl,xie2025logic,wang2025reinforcement}, several recent works have developed sophisticated multi-turn RL training frameworks for LLM agents, leveraging rule-based reward functions to provide feedback.
RAGEN~\cite{RAGEN}designs a modular multi-turn RL system for LLM agents with trajectory-level optimization, introducing the novel StarPO-S algorithm enhanced by robustness techniques such as trajectory filtering, critic incorporation, and gradient stabilization to mitigate reward variability cliffs and gradient spikes. By carefully shaping rollout generation and reasoning-aligned reward signals, the framework enables LLM agents to develop deeper reasoning patterns while significantly reducing hallucinated outputs. Building on RAGEN's success, VAGEN~\cite{wang2025vagen} extends this approach by integrating visual information, thereby adapting the multi-turn RL training paradigm for VKN agents.
StepSearch~\cite{wang2025stepsearchignitingllmssearch} constructs a multi-turn QA dataset and employs step-wise proximal policy optimization for RL training. By incorporating rich intermediate reward signal and fine-grained token-level process supervision, the framework empowers LLM agents more robust and effective search capabilities.
penalties to better guide each search step.
ReSearch~\cite{chen2025learning}integrates search operations into the reasoning process, where each search is dynamically influenced by prior reasoning steps and subsequently shapes future reasoning in multi-turn interactions. By leveraging a multi-turn RL framework, it enables LLM agents to achieve strong performance on QA tasks without requiring any supervised training data.
Search-R1~\cite{jin2025searchr1trainingllmsreason}enables LLM agents to learn multi-query generation through step-by-step reasoning with real-time feedback. By employing a simple outcome-based reward function and retrieved token masking, the framework enhances agents' reasoning capacity via multi-turn RL optimization, outperforming various existing RAG methods and QA tasks.
WebAgent-R1~\cite{wei2025webagentr1trainingwebagents}modifies GRPO for multi-turn RL training and applies asynchronous trajectory rollouts. This framework incorporates a dynamic context compression mechanism, enabling LLM agents to learn effectively from direct online interactions with web environments using only binary reward signals. 
SkyRL~\cite{cao2025skyrl,griggs2025skrylv01,liu2025skyrlsql} designs a high-throughput reinforcement learning pipeline for efficient multi-turn tool usage in LLM agents. The framework leverages optimized generation techniques to enable effective training on long-horizon, real-world tasks such as SWE-Bench. 
MT-GRPO~\cite{zeng2025reinforcingmultiturnreasoningllm} proposes a fine-grained turn-level advantage estimation strategy that enables precise credit assignment for multi-turn interactions.
Similarly, GiGPO~\cite{feng2025group}introduces a novel two-level advantage estimation operation at both episode and step levels to effectively capture global trajectory quality and local action effectiveness. This dual-scale approach enables precise credit assignment in multi-turn RL training without requiring auxiliary models, additional rollouts, or extra computational overhead. 

Recent advances in multi-turn RL training have demonstrated remarkable progress in enhancing LLMs' reasoning capabilities and tool-use proficiency, though several key challenges remain. Unlike traditional RL paradigms that treat complete actions as steps, RL applied on LLM training typically treats token generation as the fundamental step, creating a critical challenge in balancing credit assignment between token-level and turn-level advantages for long-horizon tasks. A primary limitation lies in data efficiency, as environment interactions remain computationally expensive and potentially unsafe, suggesting the need to incorporate insights from offline RL methodologies. Furthermore, current work predominantly focuses on QA search and code execution tasks where rule-based rewards are easily implemented, while many real-world applications lack clearly definable reward signals, potentially necessitating a revisit of learned reward modeling approaches. Despite these challenges, the multi-turn RL paradigm undoubtedly represents a promising direction for unifying reasoning and tool-use capabilities in LLM agents, with future work needed to address these limitations for broader real-world applicability.

\section{Action and Perception: ``Outside-In'' or ``Inside-out''} 
\label{subsec:action_perception}

\begin{figure}[!htb]
\centering
    \includegraphics[width=0.98\columnwidth]{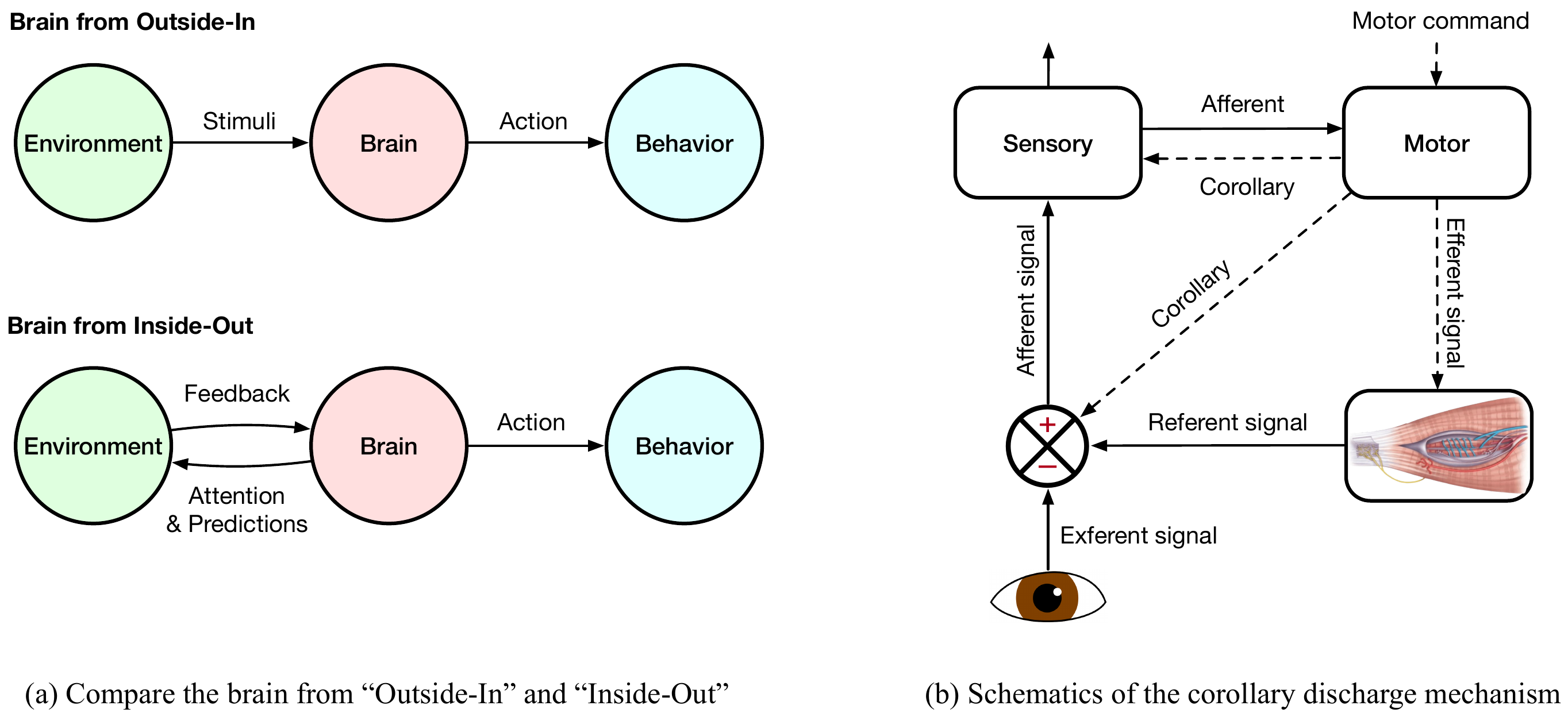}
    \caption{\textbf{(a)} Compare the brain from ``outside-in'' and ``inside-out''. \textbf{(b)} Illustration of the schematic of the corollary discharge mechanism. A motor command (efferent signal) travels from motor areas to the eye muscles, while a corollary discharge (dashed arrow) is routed to a comparator in the sensory system. The comparator uses this internal signal to modulate or subtract external (exafferent) input. Additionally, tension feedback from the muscles (reafferent signal) exerts a delayed effect on perception. Direct projections from motor to sensory cortices underlie this architecture in all mammals. Part (b) is adapted from the original figure in \cite{Buzsaki2019}.}
\label{fig:inside-out}
\end{figure}

\lettrine[lines=3]{\initfamily\textcolor{darkgreen}{A}}{} central debate in cognitive science and neuroscience concerns whether action or perception stands at the root of causal flow in intelligent systems. Figure~\ref{fig:inside-out} presents different perspectives. The traditional ``outside-in'' view insists that causal influence begins with external stimuli. The environment excites peripheral receptors, these signals propagate inward, and eventually produce behavior. This perspective portrays the organism—or agent—as essentially reactive: the external world causes sensory changes, and the agent's actions represent a downstream effect of those changes.
In contrast, Buzsáki's ``inside-out'' framework \cite{Buzsaki2019} proposes that it is the agent's own actions that shape the meaning and consequences of incoming signals. Such a view implies an active agent, one which continuously generates predictions and motor commands, while sending ``corollary discharg'' or ``action copies'' to sensory areas. These internally generated signals serve as references that inform the agent which sensory changes are self-initiated rather than imposed by the outside world. In this manner, cause shifts from an external event to an internally launched initiative, leaving external stimuli to play a confirmatory or corrective role. This reversal has significant implications for how we interpret perception's purpose and function: it is not an end in itself, but a means of updating and refining the agent's own action-driven hypotheses about the environment.

From an evolutionary perspective, possessing the ability to move without relying on sophisticated sensory analysis can yield immediate survival benefits. Even simple organisms profit from periodic motion that stirs up food in nutrient-rich water, long before elaborate perceptual capacities evolve. In other words, movement precedes advanced sensing in evolutionary time, suggesting that the capacity to act is not merely the effect of external stimuli but can itself be the driving cause of subsequent perceptual development. It is precisely when action mechanisms become sufficiently established that the agent benefits from additional sensors, which guide those movements more strategically. This developmental sequence grounds perception in utility, tying sensory discrimination to the practical outcomes of movement.

Disruptions in the normal interplay of action and perception illuminate the intricate cause-effect loop. During sleep paralysis, the brain's motor commands temporarily fail to reach the muscles; external stimuli still bombard the senses, but the usual action-to-perception calibration is lost. As a result, the individual experiences a heightened sense of unreality because the brain lacks internally generated reference signals to interpret sensory input. Similarly, if one externally manipulates the eye without the brain issuing a motor command, the visual scene appears to move, highlighting how perception alone—devoid of a preceding, self-initiated action—risks confusion.
Neurophysiological data further support the inside-out model. Many neurons in areas once deemed ``purely sensory'' track not only changes in external stimuli but also self-generated movements—sometimes more strongly so. This indicates that ``cause'' in the brain frequently emerges from within, guiding both the magnitude and meaning of external signals. Without these internal correlates, raw sensory data can become ambiguous or even useless to the system.

\paragraph*{Implications for Intelligent Agents}
The inside-out perspective offers potent insights for modern research on intelligent agents. Most contemporary AI systems—and many LLM agents—still function predominantly in a reactive mode, awaiting user input and generating responses based on statistical correlations learned from vast datasets. Such passivity resembles an ``outside-in'' framework, where the agent's role is limited to responding, not initiating.
Yet if an agent were to be active, continuously forming and testing hypotheses via self-initiated behaviors (physical or representational), it might ground its own ``perceptual'' inputs—be they sensory streams or linguistic prompts—and thereby reduce ambiguity. For instance, an LLM-based agent that interjects questions or verifies its own statements against a knowledge base could better discern which inferences are self-caused from those demanded by external data. By tracking these self-initiated contributions (analogous to corollary discharge), the model could improve coherence, lessen errors known as ``hallucinations'', and refine its internal state through iterative cause-effect loops.

A proactive stance also encourages more data-efficient and context-aware learning. Instead of passively waiting for labeled examples, an agent can explore, provoke feedback, and incorporate self-generated experiences into its training. Over time, this tight coupling between action and perception may bolster the agent's ability to handle complex tasks, adapt to unanticipated challenges, and generalize more robustly.
The shift from an outside-in to an inside-out model reframes perception as causally downstream of action. Intelligent systems—whether biological or artificial—stand to benefit from recognizing that purposeful movement, or proactive conversational steps in the case of LLMs, can actively create, shape, and interpret the signals that flow back in. By acknowledging the cause-effect power of action and striving to build active rather than merely reactive agents, we may approach a deeper understanding of both natural cognition and the next generation of AI.

\section{Summary and Discussion}
\label{sec:action-discuss}

\lettrine[lines=3]{\initfamily\textcolor{darkgreen}{T}}{raditionally}, action represents the behaviors of the human cognition system based on the interactive feedback from the environment. It endows humans with the capability to think, reason, speak, run, and perform any complex manipulations. Based on the action system, humans can iteratively evolve the brain intelligence by enhancing their perception and actions from the world, and form a closed loop to further create new civilization and innovation in the world. Similarly to a human cognition system, the action system plus the tool system also play an important role for AI agents. Integrating action systems allows AI agents to systematically plan, execute, and adjust their behaviors, facilitating more adaptable and robust performance in dynamic contexts.
In this section, we systematically examine and summarize the impact of the action module on AI agents, focusing on both action systems and tool systems.

\paragraph*{Action System} 
In our studies, we briefly describe the action system from three perspectives: action space, action learning, and tool learning. In an action system, action space usually serves as the most important component, which determines the upper bound of AI agents in solving downstream tasks. It formulates which actions can be selected and performed by AI agents during interactions with real-world environments. For action space, there are also various difficulties depending on data types, ranging from discrete to continuous data. With the growing demand for AI agents, there is also a rising expectation for AI agents to handle more sophisticated tasks, particularly those involving real-world applications. Therefore, how to build robust and general action space is still an ongoing challenge in action systems. On the basis of action space, action learning is another crucial component in enabling agents to interact effectively with the external world and with humans. Action learning represents the process of an AI agent to learn and optimize its policy during interaction with real-world environments. Based on different foundation models, it also derives different action learning paradigms, from zero-shot learning (e.g., prompt engineering) to supervised training and reinforcement learning. 
In action learning, it is essential to thoroughly understand the task, including how to devise system prompts, how to determine the pre-trained or fine-tuned datasets, and the reward signals or optimization polices during the training. Despite notable progress in action learning to advance AI agent frameworks, numerous questions remain to be addressed. Specifically, the ICL paradigm requires specific prior knowledge for a proper prompt design. Additionally, combining pre-training and post-training for supervised training necessitates high-quality and diverse data, which often requires meticulous data processing and significant human effort. Furthermore, the unstable nature of reinforcement learning poses difficulties in its application in large-scale training scenarios.
Moreover, the design of action systems plays a crucial role in maximizing the benefits of tool integration. By incorporating an effective action system, AI agents can seamlessly engage with various tools, execute complex user intents, and transform external data into meaningful outcomes. This synergy between action systems and tools not only mitigates the limitations of memorization and reduces the risk of hallucinations~\cite{10.1145/3704435} but also enhances the expertise and robustness of the system. For instance, an AI agent equipped with a robust action system can dynamically select and employ the most appropriate tools for a given task, ensuring both accuracy and efficiency in its responses. Furthermore, action systems facilitate hierarchical reasoning processes, enabling agents to orchestrate intricate workflows that align closely with user objectives. This alignment is essential for tasks requiring precise execution and real-time decision-making, thereby bridging the gap between foundational model capabilities and practical application demands. Additionally, the transparency and interpretability provided by tool execution processes enhance user trust and facilitate effective human-machine collaboration. Consequently, the combination of specialized tools and robust action systems significantly elevates the performance, reliability, and applicability of AI agents in diverse and dynamic environments.

In summary, action systems can significantly establish the foundation for the problem-solving capabilities of AI agent frameworks, enabling them to tackle a broader range of complex tasks beyond foundation models.

\paragraph*{Future Directions}
Nonetheless, building an effective action system for agents requires solutions to a number of challenges, as we summarize in the following:
\begin{itemize}
    \item \textbf{Efficiency} presents a significant hurdle, particularly in real-time applications where swift and precise responses are critical. The complexity involved in action system can lead to unacceptable latency, hindering the practical deployment of AI systems in scenarios like fraud detection or real-time decision-making. To mitigate these efficiency issues, strategies such as filtering out irrelevant or redundant information, employing zero-shot prompting to streamline reasoning processes, and utilizing high-speed storage solutions for caching pertinent knowledge are imperative. These approaches help in maintaining high performance while reducing response times.
    \item \textbf{Evaluation} is also a important factor in action system, including action learning and tool learning. In the real-world environment, there exists massive actions from different sources. Therefore, how to determine the correct action or tools from disparate sources to avoid conflicting information is still a significant challenge in AI Agent. To alleviate these problems, how to build an effective and robust evaluation system to measure action system is essential to maintain the accuracy and reliability of responses. Developing robust evaluation system, verification protocols and creating transparent methods are crucial to reduce incorrectness in action prediction. Besides, exposing the decision-making processes of foundation models also help us understand which action is better and how to coordinate with various actions or tools to provide trustworthy outputs.
    \item \textbf{Multi-modality} Action learning has achieve remarkable progresses in LLM-based autonomous agent, due to the success of large language models. However, how to understand and invoke action beyond the language instructions (e.g., GUI operations or embodied tools) still remain challenges. In real-world scenarios, humans can develop or learn to use new skills through any kinds of instructions (e.g., language, image, videos or human guidance). Therefore, enabling AI agents to develop or learn actions through diverse modalities is crucial to advance the capability of AI Agent in solving practical tasks from the real-world scenarios. In other words, it is necessary for us to explore how to reduce the gap between human and AI agents in tool utilization, facilitating the design of advanced agent frameworks for the future.
    \item \textbf{Privacy} is a critical concern in the field of generative AI, especially using LLMs. As a consequence, maintaining the privacy of sensitive user data and preventing the disclosure of user behaviors are essential in tool utilization~\cite{Yao_2024}. To address these privacy concerns, some safe techniques like federated learning can be used to enable models to be trained on decentralized data sources without exposing sensitive information directly. Additionally, model distillation is often necessary to ensure models maintain high performance while safeguarding data integrity. These methods enable the effective training of models while preserving the confidentiality of user data.
    \item \textbf{Safety} Moreover, the ethical implications of human-model collaboration and the safety concerns associated with models interacting with physical environments necessitate careful consideration. Ensuring that human dignity and agency are preserved when integrating human labor with AI systems is critical. Establishing ethical guidelines, promoting fair working conditions, and fostering interdisciplinary collaboration are necessary to address these concerns. Additionally, developing robust safety mechanisms to prevent erroneous or malicious actions by AI systems interacting with physical tools or actions is imperative to safeguard against potential risks. 
\end{itemize}

In addition to the above challenges, there also remain open problems for the action system. For example, how to achieve an optimal balance between the foundation models and external tools, deciding on the appropriate timing to use the former versus the latter, remains unanswered. Specifically, although tool systems can offer flexibility and extensibility for foundation models, there is an increasing trend to enhance the intrinsic capability of foundation models. Therefore, balancing between foundation models and tool systems is essential for developing versatile and efficient AI agents.

\part{Self-Evolution in Intelligent Agents}
\label{part-enhance}

\begin{figure*}[!ht]
    \centering
    \includegraphics[width=0.6\textwidth]{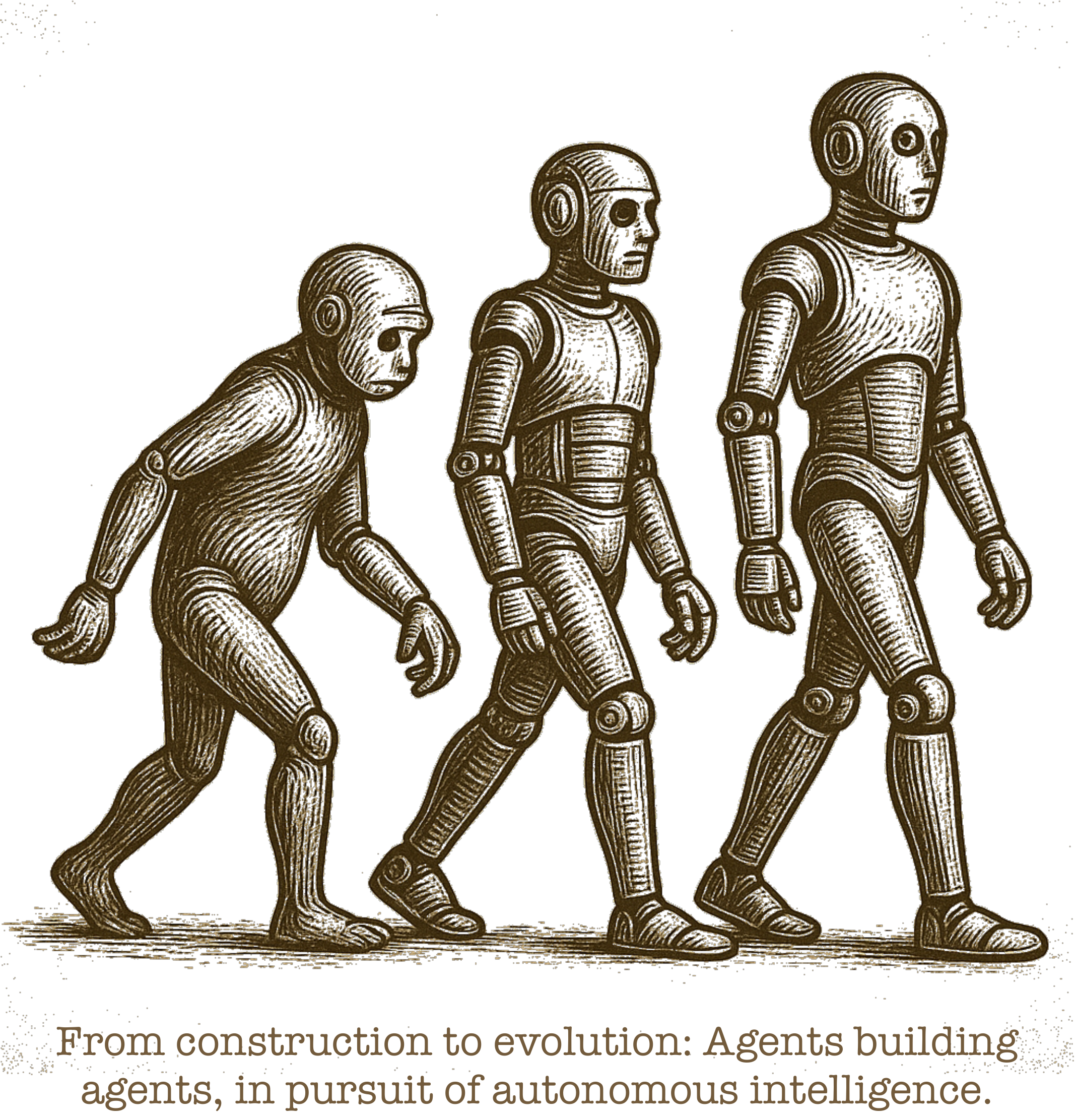}
    \label{fig:part2}
\end{figure*}

In the history of machine learning research, manually designed AI systems have gradually been replaced by more efficient, learned solutions~\cite{krizhevsky2012imagenet}. 
For instance, before the advent of deep learning, features were typically handcrafted by experts~\cite{lowe2004distinctive,dalal2005histograms}, but these were eventually superseded by features extracted through neural networks. As neural networks have become increasingly complex, various techniques for automated design--such as neural architecture search--have emerged, further replacing the need for manually designed network structures~\cite{JMLR:v20:18-598}.
Similarly, agentic systems initially relied heavily on manual design, with behavior rules and decision-making strategies explicitly crafted by developers. Although full automation of agent self-evolution has not yet been achieved, it is anticipated and deemed necessary for future progress. 
A successful precedent for such automation can already be seen in automated machine learning (AutoML)~\cite{deepeye, autodata, nvbench, sevi,zhang2024you}, which has automated various components of traditional machine learning pipelines. 
In particular, AutoML streamlines the selection and configuration of machine learning algorithm pipelines while incorporating advanced techniques for hyperparameter optimization~\cite{bischl2023hyperparameter,erickson2020autogluon,wang2021flaml,zhang2023targeted,zhang2024hypertime}. Among the most notable applications of AutoML is NAS~\cite{ren2021comprehensive,zheng2023ddpnas}, which automates the design of neural network architectures to enhance model performance.
Drawing inspiration from this successful transition towards automation in traditional machine learning, we propose extending similar principles to the domain of agentic AI systems.

A key counterintuitive issue in much of current agent research is that, while the ultimate goal of developing or improving agentic AI systems is to automate human efforts, the process of creating these systems remains, for the time being, beyond the reach of full automation. Therefore, we argue that all manually designed agentic AI systems will eventually be replaced by learnable and self-evolving systems, which could ultimately place the development and improvement of agentic AI into an autonomous, self-sustaining loop. 
Enabling the self-evolution mechanism in LLM agents has the following benefits: 
\begin{enumerate}
    \item \textbf{Scalability:} While LLM-based agents have demonstrated remarkable performance, their improvement still heavily depends on the underlying LLMs. However, upgrading these models is costly, and scaling performance through the inclusion of additional real-world data requires extensive retraining on large datasets, which poses significant resource constraints. Self-evolving agentic systems offer an alternative way to improve agent behavior by optimizing components such as strategies, tools, and workflows, without necessarily changing the underlying language model. While enhancing the core model remains essential for expanding general capabilities, self-evolution allows agents to adapt and specialize more efficiently to new tasks and environments.
    \item \textbf{Reduction in Labor Costs:} Manually designing agentic systems is a complex and labor-intensive process that requires developers to engage deeply with intricate technical details. Traditional methods often involve building these systems from scratch, demanding significant expertise and effort. By contrast, self-evolving agentic systems can automate much of this process, significantly reducing the need for manual intervention and lowering development costs.
    \item \textbf{Aligned with Natural Intelligence Development:} Just as humans continuously improve themselves through learning and adaptation, equipping LLM agents with self-improvement capabilities is a necessary step toward the development of truly autonomous agents. This enables them to refine their performance, adapt to new challenges, and evolve without direct human intervention.
\end{enumerate}

To move closer to the goal of automating human efforts, many recent efforts have explored using large language models (LLMs) not only as agents themselves, but also as tools for improving and evolving agentic systems~\cite{zhang2024aflow,fernando2023promptbreeder,yuan2023craft}. Unlike traditional approaches that rely heavily on gradient descent or reinforcement learning, LLMs offer a flexible and expressive medium for optimization, leveraging natural language to navigate and modify a wide variety of agent components. This opens the door to new paradigms of self-evolving systems—where agents can autonomously refine their behaviors, tools, and workflows over time. A growing body of work has begun to explore this possibility, suggesting a future in which agentic systems are increasingly generated and improved by other intelligent systems, rather than handcrafted by humans.

In this part, we first introduce various \emph{optimization spaces} explored in recent research on agentic systems, including prompts, tools, and workflows. 
In the subsequent chapter, we then review \emph{optimization algorithms}, discussing both traditional optimization paradigms and meta-optimization, where the optimization process also affects the underlying optimization algorithms themselves. 
Next, we explore different \emph{self-evolution scenarios}, categorizing them into two types: online optimization and offline optimization. 
Following this, we discuss the application of large language model (LLM) agent self-improvement techniques, particularly in knowledge discovery within the AI-for-Science domain. 
Finally, we discuss the \emph{security concerns} associated with agent self-evolution technologies. 

\chapter{Dimensions of Self-Optimization in Intelligent Agents}
\label{sec:optimization-space}

\lettrine[lines=3]{\initfamily\textcolor{darkgreen}{T}}{he} self-improvement of autonomous agents requires optimization across several interdependent dimensions. At the foundation lies prompt optimization, which governs how agents interact with large language models and adapt to context. Building upon this layer, three key optimization axes emerge: \emph{agentic workflow optimization}, which structures how tasks are decomposed and executed; \emph{tool optimization}, which refines the selection and use of external capabilities; and \emph{holistic agent optimization}, which calibrates the entire system for coherence, efficiency, and adaptability. This chapter introduces these interconnected spaces and outlines how they together support the iterative evolution of intelligent agents toward greater autonomy and competence.

\section{Overview of Agent Optimization}
\label{sec:overview-agent-optim}

\lettrine[lines=3]{\initfamily\textcolor{darkgreen}{A}}{t} the heart of designing intelligent agents lies a fascinating challenge: teaching these systems how to evolve and improve themselves autonomously. Much like humans learning new skills, agents need structured ways to reflect on their performance and adjust their strategies accordingly. Agent optimization addresses precisely this challenge, focusing on enhancing the effectiveness, efficiency, and adaptability of agentic AI through systematic self-improvement.

\begin{figure}[!htb]
\centering
    \includegraphics[width=0.8\columnwidth]{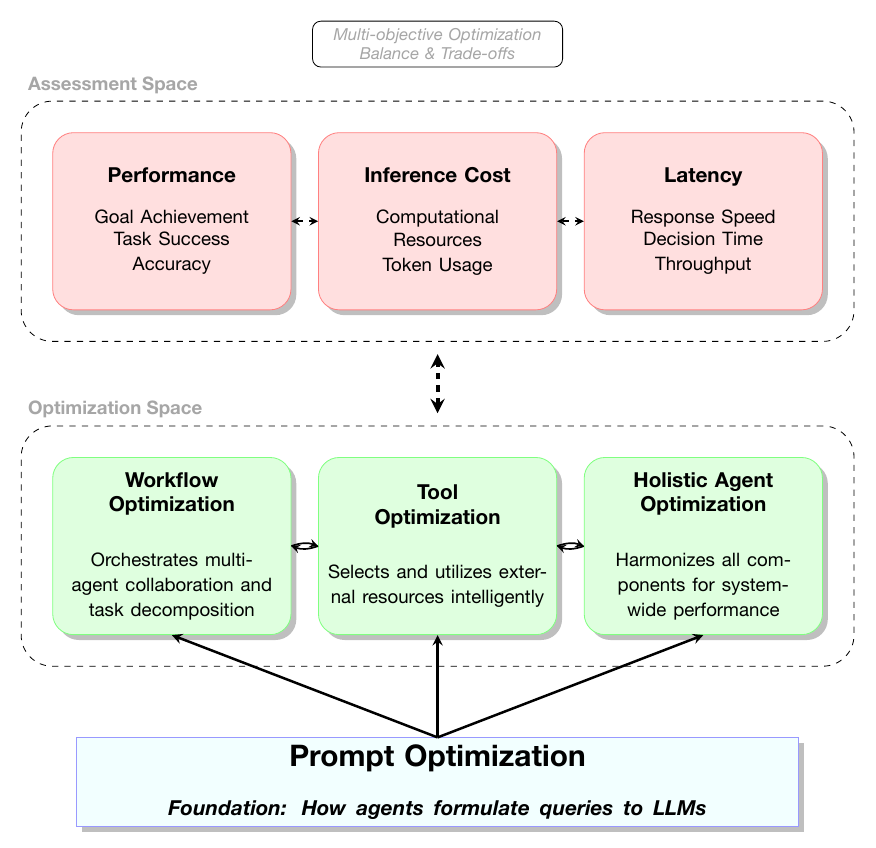}
    \caption{Illustration of the multi-layered optimization framework for intelligent agents. }
    \label{fig:optimization-space}
\end{figure}

Today's Large Language Model (LLM)-based agents operate within a layered optimization landscape. At its foundation is \emph{prompt optimization}, akin to fine-tuning how clearly we communicate instructions to a colleague. Prompt optimization governs how an agent formulates its queries to language models, fundamentally shaping its responses and behaviors. 
Building upon prompt optimization, three distinct yet interwoven pathways unfold:
First, we have \emph{workflow optimization}. Think of an agentic workflow as a team collaborating on a complex project. Each team member (or node) has a specialized role, and the workflow defines how they coordinate tasks and share information. Optimizing workflows thus involves refining the interactions among multiple LLM components, ensuring smooth communication, effective task decomposition, and coherent execution.
Next is \emph{tool optimization}. Just as skilled craftsmen rely on high-quality, precisely selected tools, agents leverage external resources—ranging from databases to specialized software functions—to solve tasks more effectively. Tool optimization allows agents to select and utilize these resources intelligently, learning to choose or even create tools dynamically as tasks evolve.
Finally, we consider \emph{holistic agent optimization}. This broader perspective treats the entire agent as a unified system, recognizing that simply enhancing individual components—prompts, tools, or workflows—does not guarantee optimal overall performance. Holistic optimization thus aims to harmonize these elements, balancing their strengths and weaknesses to ensure the agent performs reliably across diverse scenarios. Figure~\ref{fig:optimization-space} shows an overview of the optimization landscape of agents.

Optimizing intelligent agents also parallels traditional approaches found in automated machine learning (AutoML). Just as AutoML tackles trade-offs between accuracy, computational efficiency, and deployment latency, agent optimization must similarly navigate multiple competing objectives. Typically, we assess agents along three critical axes: \emph{performance} (how well the agent achieves its goals), \emph{inference cost} (the computational resources consumed), and \emph{latency} (the speed of decision-making). Each of these metrics can dominate in different contexts—for example, a customer support chatbot prioritizes rapid responses, whereas a scientific discovery agent values accuracy above speed.
Yet, optimizing for multiple objectives simultaneously introduces complexity. It forces us to balance conflicting priorities, demanding thoughtful strategies that transcend simple performance improvements. This complexity reflects real-world challenges where optimal decisions rarely maximize one factor alone but instead find intelligent compromises across several dimensions.

Throughout this chapter, we will explore each optimization dimension in depth, providing the necessary insights and techniques to guide the evolution of truly autonomous and adaptive agents.

\section{Prompt Optimization}
\label{sec:prompt-optim}

\lettrine[lines=3]{\initfamily\textcolor{darkgreen}{P}}{rompt optimization} plays a crucial role in LLM-based agent optimization. When optimizing an agent, beyond model-level optimizations, task-specific or model-specific prompt optimization directly impacts the agent's performance, latency, and cost. 
Given a task $T = (Q, G_t)$, where $Q$ denotes the input query and $G_t$ represents the optional ground truth context, the objective of prompt optimization is to generate a task-specific prompt $P_t^*$ that maximizes performance:

\begin{definition}
[\textbf{Prompt Optimization}]
We can formally define the process of prompt optimization as:
\label{def:prompt-opt}
\begin{align}
P^* &= \argmax_{P \in \mathcal{P}} 
\mathbb{E}_{T \sim \mathcal{D}}\left[
    \phi_{\mathsf{eval}}\left(
        \phi_{\mathsf{exe}}(Q, P),\; T
    \right)
\right]
\end{align}
where $\mathcal{P}$ represents the space of possible prompts, $\phi_\text{exe}$ denotes the execution function, and $\phi_\text{eval}$ represents the evaluation function.
This optimization is typically implemented through three fundamental functions: $\phi_\text{opt}$, $\phi_\text{exe}$, and $\phi_\text{eval}$. The Optimize function $\phi_\text{opt}$ refines existing prompts based on optimization signals, the Execute function $\phi_\text{exe}$ invokes the current prompt to obtain output $O$, and the Evaluation function $\phi_\text{eval}$ assesses current outputs to generate evaluation signals $S_\text{eval}$ and optimization signals $S_\text{opt}$. The evaluation signals are used to select effective prompts, while the optimization signals assist the Optimize function in performing optimization. 
\end{definition}






\begin{figure}[!htb]
\centering
    \includegraphics[width=0.8\columnwidth]{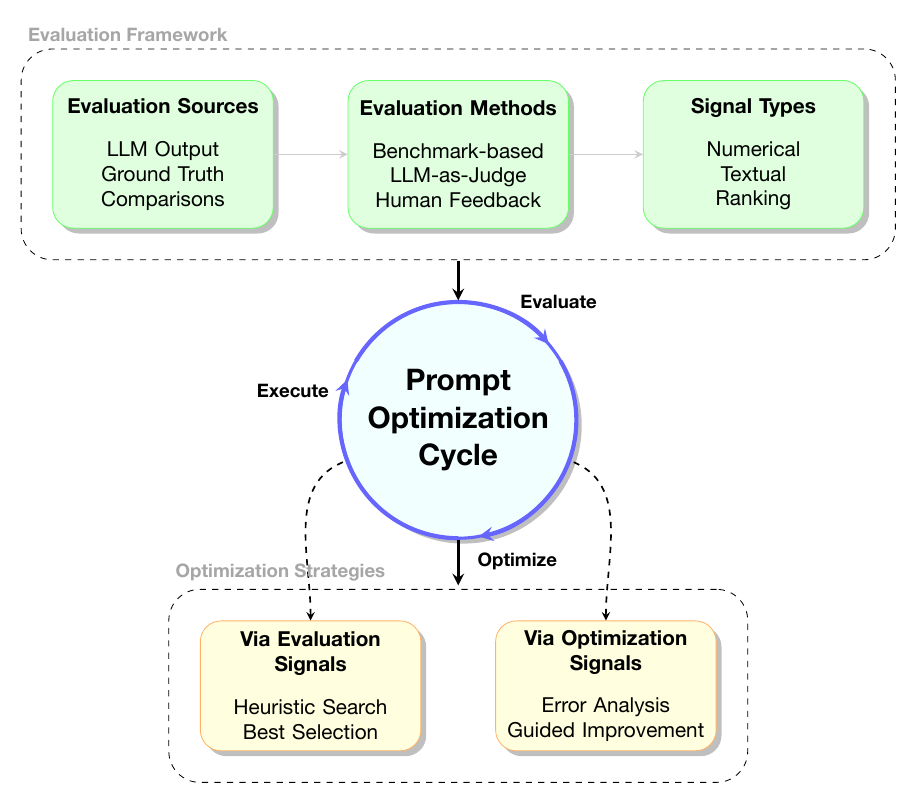}
    \caption{The prompt optimization cycle consists of three core functions: Optimize, Execute, and Evaluate. The evaluation framework (top) processes sources, methods, and signal types to generate evaluation and optimization signals that guide two optimization strategies (bottom). Arrows indicate the iterative, feedback-driven flow of the optimization process. }
    \label{fig:prompt-optim}
\end{figure}

Figure~\ref{fig:prompt-optim} gives an overview about prompt optimization. Next, we will introduce each part in details.

\subsection{Evaluation Functions}
\label{subsec:prompt-evaluation-func}

At the core of prompt optimization lies the evaluation function $\phi_{eval}$, which serves as the cornerstone for deriving optimization signals and guiding the evolutionary trajectory of prompts. This function orchestrates a sophisticated interplay between evaluation sources, methodologies, and signal generation, establishing a feedback loop that drives continuous improvement. The evaluation function $\phi_{eval}$ processes evaluation sources as input, and employs various evaluation methods to generate different types of signals, which subsequently guide the optimization process. Here, we define the dimensions of sources, methods, and signal types to establish the foundation for prompt optimization.

\paragraph*{Evaluation Sources}
Evaluation sources primarily consist of LLM Generated Output $G_{llm}$ and task-specific Ground Truth $G_t$. Existing works such as ~\cite{yang2024large, zhou2022large, yuksekgonul2024textgrad, wang2024semantic, fernando2023promptbreeder, wang2023promptagent} predominantly leverage comparisons between $G_{llm}$ and $G_t$ as evaluation sources. Some approaches~\cite{xuan2024glape, chen2024prompt, lin2024apohf} utilize only $G_{llm}$ as the evaluation source. For instance, PROMST~\cite{chen2024prompt} assesses prompt effectiveness by comparing $G_{llm}$ against human-crafted rules; SPO~\cite{xiang2025self} employs pairwise comparisons of outputs from different prompts to determine relative effectiveness.

\paragraph*{Evaluation Methods}
Evaluation Methods can be broadly categorized into three approaches: \emph{benchmark-based evaluation}, \emph{LLM-as-a-Judge}, and \emph{human feedback}. \emph{Benchmark-based evaluation} remains the most prevalent method in prompt optimization \cite{yang2024large, zhou2022large, chen2024prompt, fernando2023promptbreeder, wang2023promptagent}. This approach relies on predefined metrics or rules to provide numerical feedback as evaluation signals. While it offers an automated evaluation process, its effectiveness ultimately depends on how well the benchmark design aligns with human preferences.

The introduction of \emph{LLM-as-a-Judge} represents a significant advancement in automated evaluation and preference alignment. Leveraging LLMs' inherent alignment with human preferences and carefully designed judging criteria, this approach \cite{zheng2023judging} can assess task completion quality based on task descriptions and prompt outputs $G_{llm}$, providing reflective textual gradient feedback. Notable implementations include ProteGi \cite{reid2023protegi}, TextGrad \cite{yuksekgonul2024textgrad}, Semantic Search \cite{wang2024semantic} and Revolve \cite{zhang2024revolve}. Furthermore, LLM-as-a-judge enables comparative evaluation between ground truth $G_t$ and output $G_{llm}$ with specific scoring mechanisms \cite{khattab2023dspy}. The effectiveness of this method hinges on both the design of judger prompts and the underlying model's alignment with human preferences. As a specialized extension, \emph{Agent-as-a-Judge} \cite{zhuge2024agent} refines this paradigm by employing dedicated agents for providing process evaluation on complex tasks, while maintaining high alignment with human preferences at significantly reduced evaluation costs.

\emph{Human feedback} represents the highest level of intelligence integration in the evaluation process. As humans remain the ultimate arbiters of prompt effectiveness, direct human feedback can rapidly and substantially improve prompt quality. However, this approach introduces significant resource overhead. APOHF~\cite{lin2024apohf} demonstrates that incorporating human feedback can achieve robust prompt optimization with minimal computational resources, particularly excelling in open-ended tasks such as user instructions, prompt optimization for text-to-image generative models, and creative writing. Nevertheless, the requirement for human intervention somewhat contradicts the goal of automated evolution.

\paragraph*{Signal Types}
Feedback generated by evaluation methods manifests in three distinct forms, each serving different optimization needs. \emph{Numerical feedback}~\cite{yang2024large, zhou2022large, chen2024prompt, fernando2023promptbreeder, wang2023promptagent} quantifies performance through scalar metrics, compatible with rules, ground truth, human assessment, and LLM judgments. While widely applicable, this approach requires substantial samples for statistical reliability, potentially overlooking instance-specific details that could guide optimization. \emph{Textual feedback}~\cite{yuksekgonul2024textgrad, wang2024semantic, zhang2024revolve} provides detailed, instance-specific guidance through analysis and concrete suggestions. This sophisticated approach requires intelligent participation, either from human experts or advanced language models, enabling targeted improvements in prompt design through explicit recommendations. However, its reliance on sophisticated intelligence sources impacts its scalability.\emph{Ranking feedback}~\cite{xiang2025self} establishes relative quality ordering through either comprehensive ranking or pairwise comparisons. This approach uniquely circumvents the need for absolute quality measures or predefined criteria, requiring only preference judgments. It proves particularly valuable when absolute metrics are difficult to define or when optimization primarily concerns relative improvements.

\subsection{Optimization Functions}
\label{subsec:prompt-optim-func}

The design of optimization functions is crucial in determining the quality of generated prompts in each iteration of prompt optimization. Through effective signal guidance, prompt self-evolution can achieve faster convergence. Current optimization approaches primarily rely on two types of signals: \emph{evaluation signals} $S_{eval}$ that identify the most effective existing prompts, and \emph{optimization signals} $S_{opt}$ that provide detailed guidance for improvements.

\paragraph*{Optimize via Evaluation Signals}
When optimizing with evaluation signals, the process begins by selecting the most effective prompts based on $\phi_{eval}$ assessments. Rather than directly learning from past errors, some methods adopt heuristic exploration and optimization strategies. SPO ~\cite{xiang2025self} iteratively refines prompts based on the outputs of the current best-performing ones, leveraging the language model's inherent ability to align with task requirements. Similarly, Evoprompt ~\cite{qing2024evoprompt} employs evolutionary algorithms with LLMs serving as evolution operators for heuristic prompt combination. PromptBreeder ~\cite{fernando2023promptbreeder} advances this approach further by comparing score variations between mutated prompts while simultaneously modifying both meta-prompts and prompts through the LLM's inherent capabilities.

\paragraph*{Optimize via Optimization Signals}
While optimization methods based solely on evaluation signals require extensive search to find optimal solutions in vast search spaces through trial and error, an alternative approach leverages explicit optimization signals to guide the optimization direction and improve efficiency. Existing methods demonstrate various ways to utilize these optimization signals. OPRO \cite{yang2024large} extracts common patterns from high-performing prompt solutions to guide subsequent optimization steps. ProTegi \cite{reid2023protegi} employs language models to analyze failure cases and predict error causes, using these insights as optimization guidance. TextGrad \cite{yuksekgonul2024textgrad} extends this approach further by transforming prompt reflections into ``textual gradients'', applying this guidance across multiple prompts within agentic systems. Revolve \cite{zhang2024revolve} further enhances optimization by simulating second-order optimization, extending previous first-order feedback mechanisms to model the evolving relationship between consecutive prompts and responses. This allows the system to adjust based on how previous gradients change, avoiding stagnation in suboptimal patterns and enabling more informed, long-term improvements in complex task performance.

\subsection{Evaluation Metrics}
\label{subsec:prompt-eval-metrics}

The effectiveness of prompt optimization methods can be evaluated across multiple dimensions. \emph{Performance metrics}~\cite{yan2024erm, xiang2025self, yang2024large} for Close Tasks serve as the most direct indicators of a prompt's inherent performance, encompassing measures such as pass@1, accuracy, F1 score, and ROUGE-L. These metrics enable researchers to assess the stability, effectiveness, and convergence rate of prompt optimization processes. Another crucial dimension is \emph{Efficiency metrics}~\cite{xiang2025self}. While some prompt optimization approaches achieve outstanding results, they often demand substantial computational resources, larger sample sizes, and extensive datasets. In contrast, other methods achieve moderate results with lower resource requirements, highlighting the trade-offs between performance and efficiency in agent evolution. The third dimension focuses on qualitative metrics that assess specific aspects of agent behavior: \emph{consistency}~\cite{xuan2024glape} measures output stability across multiple runs, \emph{fairness}~\cite{zhou2024zepo} evaluates the ability to mitigate the language model's inherent biases, and \emph{confidence}~\cite{wan2023confidenceprompt, yao2022confidenceprompt} quantifies the agent's certainty in its predictions. When these behavioral aspects are treated as distinct objectives, prompt optimization frameworks provide corresponding metrics for evaluation.

\section{Workflow Optimization}
\label{sec:workflow-optimization}

\lettrine[lines=3]{\initfamily\textcolor{darkgreen}{W}}{hile} prompt-level optimization has shown promising results in enhancing individual LLM capabilities, modern AI systems often require the coordination of multiple LLM components to tackle complex tasks. This necessitates a more comprehensive optimization domain---the agentic workflow space. At its core, an agentic workflow consists of LLM-invoking nodes, where each node represents a specialized LLM component designed for specific sub-tasks within the larger system.

Although this architecture bears similarities to multi-agent systems, it is important to distinguish agentic workflows from fully autonomous multi-agent scenarios. In agentic workflows, nodes operate under predetermined protocols and optimization objectives, rather than exhibiting autonomous decision-making capabilities. Many prominent systems, such as MetaGPT~\cite {hong2023metagpt}, AlphaCodium~\cite{tal2024alpha}, can be categorized under this framework. Moreover, agentic workflows can serve as executable components within larger autonomous agent systems, making their optimization crucial for advancing both specialized task completion and general agent capabilities. 

Following the formalization proposed by GPTSwarm~\cite{zhuge2024gptswarm} and AFLOW~\cite{zhang2024aflow}, this section first establishes a formal definition of agentic workflows and their optimization objectives. We then examine the core components of agentic workflows---nodes and edges---analyzing their respective search spaces and discussing existing representation approaches in the literature.

\subsection{Workflow Formulation}
\label{subsec:workflow-formulation}

Here we present a formal definition for agentic workflow.

\begin{definition}
[\textbf{Agentic Workflow}]
An agentic workflow $\mathcal{K}_{\text{wf}}$ can be formally represented as:

\begin{equation}
\mathcal{K}_{\text{wf}} = \{(N, E) | N \in \mathcal{N}, E \in \mathcal{E}\}
\end{equation}

where $\mathcal{N} = \{N(M, \tau, P, F) | M \in \mathcal{M}, \tau \in [0,1], P \in \mathcal{P}, F \in \mathcal{F}\}$ represents the set of LLM-invoking nodes, with $\mathcal{M}$, $\tau$, $\mathcal{P}$, and $\mathcal{F}$ denoting the available language models, temperature parameter, prompt space, and output format space respectively. $E$ indicates the edges between different LLM-invoking nodes. This formulation encapsulates both the structural components and operational parameters that define an agentic workflow's behavior.

Given a task $T$ and evaluation metrics $\mathcal{O}$, the goal of workflow optimization is to discover the optimal workflow $K_{wf}^*$ that maximizes performance:
\begin{equation}
K_{\text{wf}}^{*}
    = \argmax_{K_{\text{wf}}\in\mathcal{K}_{\text{wf}}}
      \,\mathcal{O}\!\bigl(K_{\text{wf}},T\bigr),
\end{equation}
\noindent
where $\mathcal{K}_{\text{wf}}$ is the workflow search space, 
$K_{\text{wf}}$ is a candidate workflow drawn from that space, 
and $\mathcal{O}(K_{\text{wf}},T)$ jointly measures task-completion quality,
computational efficiency, and execution latency.

\end{definition}

\begin{figure}[!htb]
\centering
    \includegraphics[width=0.8\columnwidth]{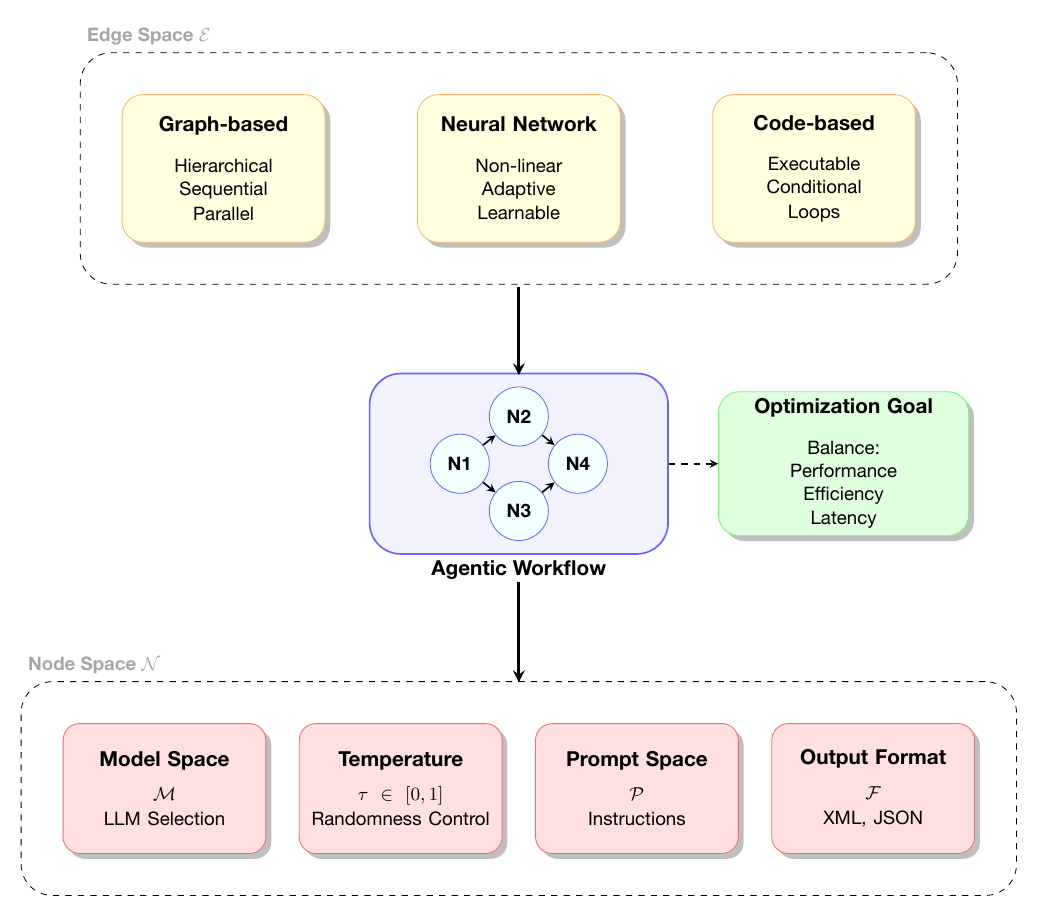}
    \caption{Framework for optimizing agentic workflows composed of LLM-invoking nodes and edges. The edge space (top) offers three representation paradigms: graph-based, neural network, and code-based. The node space (bottom) encompasses four optimization dimensions: model selection, temperature, prompts, and output format. The optimization goal balances performance, efficiency, and latency.}
    \label{fig:workflow-optim}
\end{figure}

Figure~\ref{fig:workflow-optim} gives an overview about workflow optimization. Next, we will introduce each part with more details.

\subsection{Optimizing Workflow Edges}
\label{subsec:optim-workflow-edges}

The edge space $\mathcal{E}$ defines the representation formalism for agentic workflows. Current approaches primarily adopt three distinct representation paradigms: graph-based, neural network-based, and code-based structures. Each paradigm offers unique advantages and introduces specific constraints on the optimization process.

\emph{Graph-based} representations enable the expression of hierarchical, sequential, and parallel relationships between nodes. This approach naturally accommodates complex branching patterns and facilitates visualization of workflow topology, making it particularly suitable for scenarios requiring explicit structural manipulation. For example, GPTSwarm~\cite{zhuge2024gptswarm} demonstrated the effectiveness of graph-based workflow representation in coordinating multiple LLM components through topology-aware optimization. Neural network architectures provide another powerful representation paradigm that excels in capturing non-linear relationships between nodes. Dylan~\cite{liu2023dynamic} showed that neural network-based workflows can exhibit adaptive behavior through learnable parameters, making them especially effective for scenarios requiring dynamic adjustment based on input and feedback. Code-based representation offers the most comprehensive expressiveness among current approaches. AFLOW~\cite{zhang2024aflow} and ADAS~\cite{hu2024automated} established that representing workflows as executable code supports linear sequences, conditional logic, loops, and the integration of both graph and network structures. This approach provides precise control over workflow execution and leverages LLMs' inherent code generation capabilities.

The choice of edge space representation significantly influences both the search space dimensionality and the applicable optimization algorithms. \cite{yuksekgonul2024textgrad} focused solely on prompt optimization while maintaining a fixed workflow topology, enabling the use of textual feedback-based optimization techniques. In contrast, \cite{zhuge2024gptswarm} developed reinforcement learning algorithms for joint optimization of individual node prompts and overall topology. \cite{zhang2024aflow} leveraged code-based representation to enable direct workflow optimization by language models, while recent advances by \cite{zhang2025multi} and \cite{wang2025scoreflow} introduced methods for problem-specific topology optimization.

Recent advances in this area further demonstrate the trend of treating the workflow itself as a dynamic, optimizable program. MAS-GPT~\cite{ye2025mas}, by fine-tuning a language model, can directly generate complete, executable multi-agent system code from a user query. FlowReasoner~\cite{gao2025flowreasoner} utilizes a meta-agent that learns and iterates on a workflow's design based on performance feedback after each task execution. Still others achieve zero-supervision adaptive design; MAS-ZERO~\cite{ke2025maszero} uses a self-supervised iterative loop at inference-time to continuously generate, evaluate, and refine the workflow without pre-existing training or labeled data. Together, these pioneering methods are driving a paradigm shift in workflow optimization—moving from designing fixed topologies to the automated, contextual, and adaptive construction of workflows using techniques from generative AI, reinforcement learning, and self-supervised learning.

\subsection{Optimizing Workflow Nodes}
\label{subsec:optim-workflow-nodes}

The node space ${N}$ consists of four key dimensions that influence node behavior and performance. The output format space $F$ significantly impacts performance by structuring LLM outputs, with formats like XML and JSON enabling more precise control over response structure. The temperature parameter $\tau$ controls output randomness, affecting the stability-creativity tradeoff in node responses. The prompt space $P$ inherits the optimization domain from prompt-level optimization, determining the core interaction patterns with LLMs. The model space $M$ represents available LLMs, each with distinct capabilities and computational costs.

For single-node optimization, existing research has primarily focused on specific dimensions within this space. \cite{zhang2024aflow} concentrated exclusively on prompt optimization, while \cite{hu2024automated} extended the search space to include both prompts and temperature parameters. Taking a different approach, \cite{saad2024archon} fixed prompts while exploring model selection across different nodes. Output format optimization, though crucial, remains relatively unexplored~\cite{tam2024format}. 

Compared to edge space optimization, node space optimization poses unique scalability challenges due to the typically large number of nodes in agentic workflows. The dimensionality of the search space grows multiplicatively with each additional node, necessitating efficient optimization strategies that can effectively handle this complexity while maintaining reasonable computational costs.

\section{Tool Optimization}
\label{sec:tool_optimization}

\begin{figure}[!htb]
\centering
    \includegraphics[width=0.8\columnwidth]{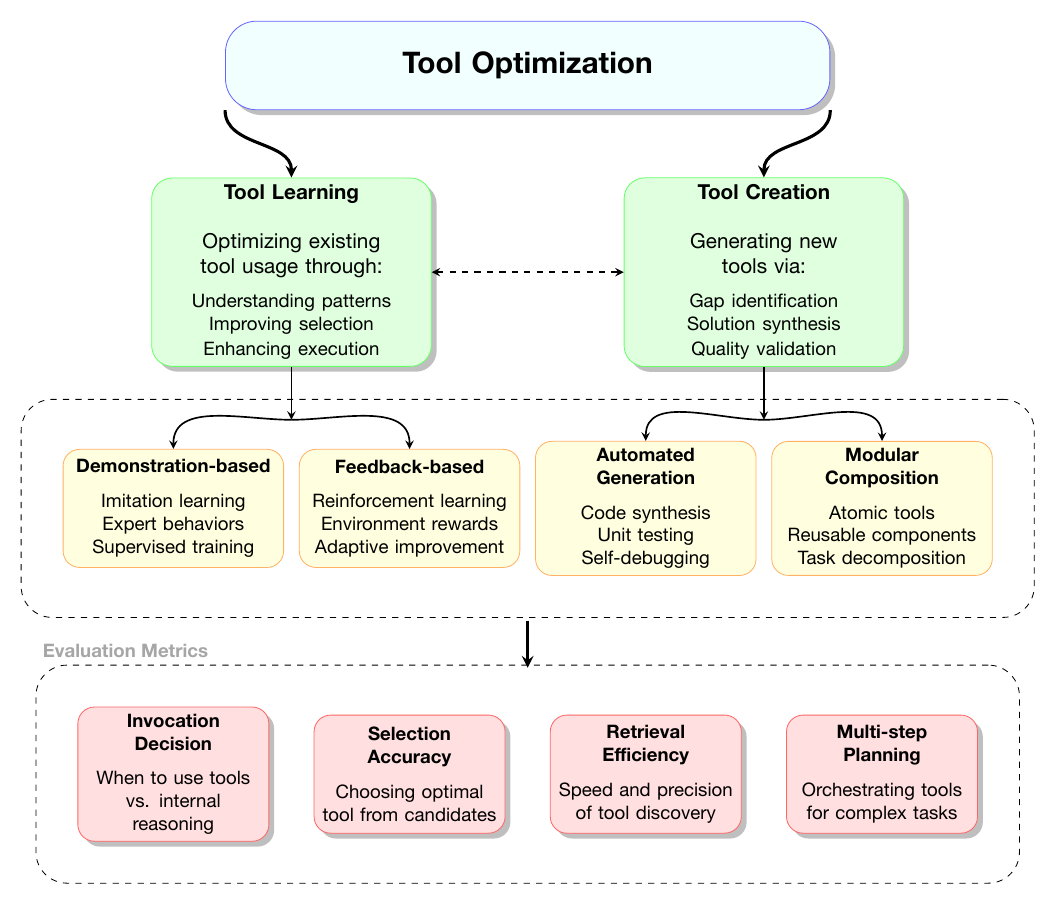}
    \caption{Tool optimization operates through two strategies: Tool Learning (optimizing existing tools via demonstrations and feedback) and Tool Creation (generating new tools through automated synthesis and modular composition). Different evaluation metrics assess tool usage.}
    \label{fig:tool-optim}
\end{figure}

\lettrine[lines=3]{\initfamily\textcolor{darkgreen}{U}}{nlike} conventional usage of LLMs that typically operate in a single-turn manner, agents are equipped with advanced multi-turn planning capabilities and the ability to interact with the external world via various tools. These unique attributes make the optimization of tool usage a critical component in enhancing an agent's overall performance and adaptability. Tool optimization involves systematically evaluating and refining how an agent selects, invokes, and integrates available tools to solve problems with higher efficiency and lower latency. Key performance metrics in this context include decision-making accuracy, retrieval efficiency, selection precision, task planning, and risk management. Central to this optimization are two complementary strategies: \emph{tool learning} and \emph{tool creation}. 
Figure~\ref{fig:tool-optim} shows an overview about tool optimization.

\subsection{Learning to Use Tools}
\label{subsec:learning-to-use-tools}

Unlike prompting-based methods that leverage frozen foundation models' in-context learning abilities, training-based methods optimize the model that backs LLM agents with supervision. Drawing inspiration from developmental psychology, tool learning can be categorized into two primary streams: \emph{learning from demonstrations} and \emph{learning from feedback}~\cite{10.1145/3704435}. The other way to elicit the power of LLMs (agents) using tools is by using prompt-based or in-context learning methods for better reasoning abilities.

\begin{definition}
[\textbf{Tool Learning}]
\label{def:tool-learning}
Tool learning refers to the general process by which an LLM-based agent or controller acquires and improves its ability to utilize tools, based on data, demonstrations, or feedback signals, to maximize task performance. 
Formally, let $\theta_C$ denote the parameters of the controller (or policy). The general objective of tool learning can be expressed as:
\begin{align}
\theta_C^* 
&= \arg\max_{\theta_C} 
\mathbb{E}_{\mathcal{D}, \mathcal{R}}
\left[
    \mathcal{J}(\theta_C; \mathcal{D}, \mathcal{R})
\right],
\end{align}
where $\mathcal{D}$ denotes the data source (e.g., demonstrations or interactions), 
$\mathcal{R}$ represents the reward or evaluation signals, 
and $\mathcal{J}$ is the performance objective (such as log-likelihood of expert actions or expected reward).

\paragraph*{Learning From Demonstrations:}
This approach trains the controller to imitate expert demonstrations via supervised learning or behavior cloning. Given a dataset $D = \{(q_i, a_i^*)\}_{i=0}^{N-1}$, where $q_i$ is a user query and $a_i^*$ is the corresponding human-annotated action sequence, the parameters $\theta_C$ are optimized as:
\begin{align}
\theta_C^* 
&= \arg \max_{\theta_C} 
\mathbb{E}_{(q_i, a^*_i) \in D}
\prod_{t=0}^{T_i}
p_{\theta_C}
\bigl(a^*_{i,t} \mid x_{i,t}, H_{i,t}, q_i\bigr),
\end{align}
where $a^*_{i,t}$ is the expert action at step $t$ for query $q_i$, $x_{i,t}$ denotes the controller's local observation or state at time $t$ (e.g., current tool context or environment state), $H_{i,t}$ is its hidden state / action–observation history up to \(t\),  $p_{\theta_C}$ is the policy distribution (i.e., the probability assigned by the controller to each feasible action conditioned on the current observation, history, and query), and $T_i$ is the total number of timesteps.

\paragraph*{Learning From Feedback:}
This approach leverages reinforcement learning to adapt the controller based on reward signals from the environment or human feedback. The optimization objective is:
\begin{align}
\theta_C^* 
&= \arg \max_{\theta_C} 
\mathbb{E}_{q_i \in Q} 
\mathbb{E}_{\{a_{i,t}\}_{t=0}^{T_i}}
\left[
    R\bigl(\{a_{i,t}\}_{t=0}^{T_i}\bigr)
\right],
\end{align}
where $R$ denotes the reward function based on the sequence of actions $\{a_{i,t}\}$, and $Q$ is the distribution of user queries.

\end{definition}

Recently, inspired by the success of reinforcement-learning methods such as GRPO in DeepSeek-R1~\cite{guo2025deepseek}, researchers have begun to optimize tool use by directly drawing rewards from the environment, capturing factors like output-format fidelity and answer correctness across the full tool-calling pipeline (tool retrieval, parameter selection, and execution success). Search-R1~\cite{jin2025searchr1} refines search-engine usage, ToolRL~\cite{qian2025toolrlrewardtoollearning} introduces reward-driven tool learning, OTC-PO~\cite{wang2025actingreasoningmoreteaching} tightens the link between acting and reasoning, and Nemotron-Research-Tool-N~\cite{zhang2025nemotronresearchtooln1exploringtoolusinglanguage} demonstrates scenario-specific gains. Collectively, these studies highlight reinforcement learning as a powerful method for adaptive, task-aware tool optimization for Foundation Agents.

Integrating tool learning into the optimization framework enhances the system's ability to generalize tool usage across diverse tasks and environments. By incorporating both demonstration-based and feedback-based learning, the model can iteratively improve its tool invocation strategies, selection policies, and execution accuracy.

\paragraph*{Optimization Reasoning Strategies for Tool Using}
Optimizing the aforementioned metrics for better LLM agents' abilities requires a combination of advanced retrieval models, fine-tuned reasoning strategies, and adaptive learning mechanisms. Reasoning strategies, such as Chain-of-Thought (CoT)~\cite{wei2022chain}, Tree-of-Thought~\cite{yaotree}, and Depth-First Search Decision Trees (DFS-DT)~\cite{qin2023toolllmfacilitatinglargelanguage}, facilitate more sophisticated decision-making processes regarding tool usage. Fine-tuning the model's understanding of tools, including parameter interpretation and action execution, enables more precise and effective tool interactions. Additionally, learning from the model's outputs allows for better post-processing and analysis, further refining tool utilization efficacy.

\subsection{Creation of New Tools}
\label{subsubsec:toolcreation}

Beyond the optimization of existing tools, the ability to create new tools dynamically~\cite{qian2023creator,cai2023large,yuan2023craft} based on a deep understanding of tasks and current tool usage can significantly enhance the LLM Agent framework's adaptability and efficiency. In recent work, several complementary approaches have been proposed. ToolMakers~\cite{cai2023large} establishes a closed-loop framework where a tool-making agent iteratively executes three phases: (1) \textit{Proposing} Python functions via programming-by-example using three demonstrations, (2) \textit{Verifying} functionality through automated unit testing (3 validation samples) with self-debugging of test cases, and (3) \textit{Wrapping} validated tools with usage demonstrations for downstream tasks. This rigorous process ensures reliability while maintaining full automation. CREATOR~\cite{qian2023creator} adopts a four-stage lifecycle: \textit{Creation} of task-specific tools through abstract reasoning, \textit{Decision} planning for tool invocation, \textit{Execution} of generated programs, and \textit{Rectification} through iterative tool refinement—emphasizing tool diversity, separation of abstract/concrete reasoning, and error recovery mechanisms. In contrast, CRAFT~\cite{yuan2023craft} employs an offline paradigm that distills domain-specific data into reusable, atomic tools (e.g., object color detection) through GPT-4 prompting, validation, and deduplication. Its training-free approach combines human-inspectable code snippets with compositional problem-solving, enabling explainable toolchains while avoiding model fine-tuning—particularly effective when decomposing complex tasks into modular steps. Alita~\cite{qiu2025alita} frames ``tool creation'' as a compact, self-optimizing loop: the agent spots a capability gap, drafts a spec, mines code examples, autowrites and executes a script in an isolated sandbox, then packages the passing script as a standard MCP module. Each gap thus yields a ready-to-use tool, allowing the agent's arsenal to grow cumulatively with every task it tackles.

Hybrid frameworks can fuse CRAFT's ready-made tool pool with ToolMaker's on-demand generation, while Alita's self-optimizing loop (gap-detect → search → build → validate → wrap) adds continuous, data-driven expansion. Primitive CRAFT operations can seed higher-level ToolMaker / Alita composites, with CREATOR-style rectifiers covering edge cases. Progress in self-supervised tool metrics and cross-domain transfer will further reduce the need for human oversight. A key open question is how tool granularity affects reuse: finer, ``atomic'' tools offer flexible composition but increase orchestration costs. Ultimately, bidirectional tool-task co-adaptation—where evolving tasks inspire the development of new tools and new tools redefine tasks—points toward truly self-improving agent ecosystems.

\subsection{Evaluation of Tool Effectiveness}
\label{subsec:eval-tool-effectiveness}

The evaluation metrics and benchmarks discussed below offer a comprehensive basis for quantifying an agent's tool usage capabilities. By assessing aspects such as tool invocation, selection accuracy, retrieval efficiency, and planning for complex tasks, these benchmarks not only measure current performance but also provide clear, concrete objectives for optimizing tool usage. Such metrics are instrumental in guiding both immediate performance enhancements and long-term strategic improvements in agent-based systems. In the following sections, we first review the evolution of agent tool use benchmarks and then consolidate the key evaluation metrics that serve as targets for further tool optimization.

\paragraph*{Tool Evaluation Benchmarks}
Recent efforts in LLM-as-Agent research have spawned diverse benchmarks and frameworks for evaluating tool-use capabilities.
Early studies such as Gorilla~\cite{patil2023gorilla} and API-Bank~\cite{li2023api} pioneered large-scale datasets and methods for testing LLM interactions with external APIs, shedding light on issues like argument accuracy and hallucination.
Subsequent works like T-Bench~\cite{xu2023tool} and ToolBench~\cite{qin2023toolllmfacilitatinglargelanguage} introduced more extensive task suites and stressed the importance of systematic data generation for tool manipulation.
StableToolBench~\cite{guo2024stabletoolbench} further extended this line of inquiry by highlighting the instability of real-world APIs, proposing a virtual API server for more consistent evaluation.
Meanwhile, ToolAlpaca~\cite{tang2023toolalpaca} investigated the feasibility of achieving generalized tool-use in relatively smaller language models with minimal in-domain training.
Additional efforts like ToolEmu~\cite{ruan2023identifying} assessed the safety and risk aspects of tool-augmented LM agents through emulated sandbox environments. MetaTool~\cite{huang2024metatoolbenchmarklargelanguage} then introduced a new benchmark focused on whether LLMs know \emph{when} to use tools and can correctly \emph{choose} which tools to employ. It provides a dataset named ToolE that covers single-tool and multi-tool usage scenarios, encouraging research into tool usage awareness and nuanced tool selection.
ToolEyes~\cite{ye2024tooleyes} pushed the evaluation further by examining real-world scenarios and multi-step reasoning across a large tool library.
Finally, $\tau$-bench~\cite{yao2024taubenchbenchmarktoolagentuserinteraction} introduced a human-in-the-loop perspective, emphasizing dynamic user interactions and policy compliance in agent-based conversations.
Together, these benchmarks and frameworks underscore the evolving landscape of tool-augmented LLM research, marking a shift from isolated reasoning tasks to comprehensive, real-world agent evaluations.

\paragraph*{Metrics for Tool Invocation} 
Deciding whether to invoke an external tool is a critical step that can significantly affect both the efficiency and the effectiveness of a system. In many scenarios, the model must determine if its own reasoning is sufficient to answer a query or if additional external knowledge (or functionality) provided by a tool is required. To formalize this process, we introduce a labeled dataset 
\begin{equation}
D_{\text{inv}} = \{(q_i, y_i)\}_{i=0}^{N-1},
\end{equation}
where \( q_i \) represents the \(i\)-th user query and \( y_i \in \{0,1\} \) is a binary label indicating whether tool invocation is necessary (\( y_i=1 \)) or not (\( y_i=0 \)). Based on this dataset, the model learns a decision function \( d(q_i) \) defined as:
\begin{equation}
d(q_i) = \begin{cases}
1, & \text{if } P_{\theta}(y=1 \mid q_i) \geq \tau, \\
0, & \text{otherwise},
\end{cases}
\end{equation}
where \( P_{\theta}(y=1 \mid q_i) \) denotes the predicted probability (from a model parameterized by \(\theta\)) that a tool should be invoked for query \(q_i\), and \(\tau\) is a predetermined threshold. 

In addition to this decision rule, several metrics can be used to evaluate the model's ability to correctly decide on tool invocation. For example, the overall invocation accuracy \( A_{\text{inv}} \) can be computed as:
\begin{equation}
A_{\text{inv}} = \frac{1}{N}\sum_{i=0}^{N-1} \mathbf{1}\{d(q_i) = y_i\},
\end{equation}
where \( \mathbf{1}\{\cdot\} \) is the indicator function. Other metrics such as precision, recall, and F1 score are also applicable. Moreover, if \( C_{\text{inv}} \) represents the cost incurred by invoking a tool and \( R(q_i) \) the benefit or reward obtained when a tool is correctly used, one can define a net benefit score:
\begin{equation}
B_{\text{inv}} = \sum_{i=0}^{N-1} \left( \mathbf{1}\{d(q_i)=1\} \cdot R(q_i) - C_{\text{inv}} \right).
\end{equation}
This formulation not only emphasizes accuracy but also considers the cost-effectiveness of invoking external tools.

\paragraph*{Tool Selection Among Candidates} 
Once the decision to invoke a tool is made, the next challenge is to select the most appropriate tool from a pool of candidates. Let the candidate toolset be represented as:
\begin{equation}
\mathcal{T} = \{t_1, t_2, \ldots, t_M\}.
\end{equation}
For a given query \( q_i \), assume that the optimal tool (according to ground truth) is \( t^*_i \) and the model selects \( \hat{t}_i \). The simplest measure of selection performance is the tool selection accuracy \( A_S \):
\begin{equation}
A_S = \frac{1}{|Q|}\sum_{q_i \in Q} \mathbf{1}\{\hat{t}_i = t^*_i\}.
\end{equation}
However, many scenarios involve ranking multiple candidate tools by their relevance. In such cases, ranking-based metrics such as Mean Reciprocal Rank (MRR) and normalized Discounted Cumulative Gain (nDCG) offer a more nuanced evaluation. \cite{qin2023toolllmfacilitatinglargelanguage} use those two when evaluating the tool retriever system.

\paragraph*{Tool Retrieval Efficiency and Hierarchical Accuracy}
Tool retrieval involves both the speed of identifying a suitable tool and the accuracy of that selection. Efficient retrieval methods reduce latency and computational overhead, while high retrieval accuracy ensures that the most relevant tool is identified for the task. To evaluate tool usage comprehensively, we adopt a hierarchical framework that distinguishes between retrieval accuracy and selection accuracy. Retrieval accuracy (\(A_R\)) reflects how precisely the system retrieves the correct tool from the repository, typically measured by metrics such as Exact Match (EM) and F1 score, which capture both complete and partial matches. In contrast, selection accuracy (\(A_S\)) assesses the system's ability to choose the optimal tool from a set of candidates, again using similar metrics. Overall tool usage awareness is further evaluated by accuracy, recall, precision, and F1 score.

The overall retrieval efficiency \( E_{Ret} \) is thus can be expressed as:
\begin{equation}
E_{Ret} = \frac{A_R \times A_S \times A_P \times A_U}{C_R}
\end{equation}
where \( C_R \) is the cost associated with retrieval. Optimization strategies may involve training embedding models with feedback mechanisms to enhance both efficiency and each hierarchical component of accuracy.

For a more nuanced evaluation of tool selection, Metatool~\cite{huang2024metatoolbenchmarklargelanguage} introduces the Correct Selection Rate (CSR), which quantifies the percentage of queries for which the model selects the expected tool(s). This evaluation framework addresses four aspects: selecting the correct tool among similar candidates, choosing appropriate tools in context-specific scenarios, ensuring reliability by avoiding the selection of incorrect or non-existent tools, and handling multi-tool queries. Together, these metrics and sub-tasks provide a robust measure of both the efficiency and precision in tool retrieval and selection.

\paragraph*{Tool Planning for Complex Tasks}
Complex tasks often require the sequential application of multiple tools to reach an optimal solution. A tool plan can be represented as an ordered sequence
\begin{equation}
\Pi = [t_1, t_2, \ldots, t_K],
\end{equation}
where \(K\) is the number of steps. The quality of such a plan is typically evaluated by balancing its task effectiveness (e.g., via a metric \(R_{\text{task}}(\Pi)\)) against the plan's complexity (or length). This balance can be captured by a composite planning score of the form
\[
S_{\text{plan}} = \alpha \cdot R_{\text{task}}(\Pi) - \beta \cdot K,
\]
where \(\alpha\) and \(\beta\) are coefficients that adjust the trade-off between the benefits of high task performance and the cost associated with plan complexity. When ground truth plans \(\Pi^{*}\) are available, similarity metrics such as BLEU or ROUGE can be used to compare the predicted plan \(\Pi\) with \(\Pi^{*}\), and an overall planning efficiency metric can be defined accordingly.

In addition, recent work such as ToolEyes~\cite{ye2024tooleyes} highlights the importance of behavioral planning in tool usage. Beyond selecting tools and parameters, it is crucial for LLMs to concisely summarize acquired information and strategically plan subsequent steps. In this context, the behavioral planning capability is evaluated along two dimensions. First, the score \(S_{b\text{-validity}} \in [0,1]\) is computed by assessing (1) the reasonableness of summarizing the current state, (2) the timeliness of planning for the next sequence of actions, and (3) the diversity of planning. Second, the score \(S_{b\text{-integrity}} \in [0,1]\) is calculated by evaluating (1) grammatical soundness, (2) logical consistency, and (3) the ability to correct thinking. The composite behavioral planning score is then determined as
\begin{equation}
S_{BP} = S_{b\text{-validity}} \cdot S_{b\text{-integrity}},
\end{equation}
providing a holistic measure of the model's planning capability. This integrated framework ensures that tool planning for complex tasks not only focuses on the selection and ordering of tools but also on maintaining coherent, effective, and strategically sound planning processes.

In summary, optimizing tool performance within an Agent system necessitates a comprehensive approach that balances decision-making accuracy, retrieval efficiency, hierarchical selection precision, strategic planning, rigorous risk management, and robust tool learning mechanisms. By implementing targeted optimization and learning strategies, it is possible to enhance both the effectiveness and efficiency of tool-assisted machine learning workflows.

\section{Towards Autonomous Agent Optimization}
\label{sec:auto-agent-optim}

\lettrine[lines=3]{\initfamily\textcolor{darkgreen}{I}}{n} addition to optimizing individual modules in agent evolution, such as prompts, tools, and workflows, which are susceptible to local optima that can compromise the overall performance of the agentic system, a significant body of research focuses on optimizing multiple components within the entire agentic systems. This holistic approach enables LLM agents to evolve more comprehensively.
However, optimizing the entire system imposes higher requirements. The algorithm must not only account for the impact of individual components on the agentic system but also consider the complex interactions between different components.

Among the most representative works that first formally defined the research problem of automated design in agentic systems are ADAS~\cite{hu2024automated} and AFLOW~\cite{zhang2024aflow}. For instance, ADAS integrates multiple agentic system components—including the workflow, prompts, and potential tools—into its evolutionary pipeline. Similarly, AFLOW automates the generation and optimization of entire agentic system workflows.

Following a similar evolutionary paradigm, AlphaEvolve~\cite{novikov2025alphaevolve} employs an evolutionary algorithm driven by LLMs to iteratively modify and improve entire codebases for scientific and algorithmic discovery. This approach has proven effective in tackling complex scientific problems where progress can be systematically measured. Another novel mechanism to drive evolution is self-challenging. For instance, Self-Challenging Agents (SCA)~\cite{zhou2025self} boost their own evolution by generating high-quality, verifiable tasks through interaction with the environment, creating a curriculum for continuous self-improvement.

Additionally, \citet{zhou2024symbolic} proposes an agent symbolic learning framework for training language agents, inspired by connectionist learning principles used in neural networks. By drawing an analogy between agent pipelines and computational graphs, the framework introduces a language-based approach to backpropagation and weight updates. It defines a prompt-based loss function, propagates language loss through agent trajectories, and updates symbolic components accordingly. This method enables structured optimization of agentic workflows and naturally extends to multi-agent systems by treating nodes as independent agents or allowing multiple agents to act within a single node.

\citet{yin2024g} proposes an approach to optimize both prompts and the agent's own code, enabling self-improvement. This aligns with the concept of self-reference, where a system can analyze and modify its own structure to enhance performance. A more profound implementation of this idea is the Darwin Gödel Machine~\cite{zhang2025darwin}, which iteratively rewrites its own python codebase to evolve new versions of itself with improved tools and workflows. Similarly, Alita~\cite{qiu2025alita} demonstrates the possibility of an agent's action space self-evolving in any environment by naturally constructing and encapsulating actions into reusable modules through interaction with the environment.

Similarly, \citet{zhang2024aflow}, \citet{zhang2025multi}, \citet{zhou2025maasprompt}, and \citet{wang2025scoreflow} focus on optimizing both the workflow and prompts within agentic systems. In particular, \citet{wang2003discovering} introduces an approach that trains additional large language models (LLMs) to generate prompts and workflows, enabling the automated design of agentic system architectures.

In summary, optimizing the workflow of an entire agentic system is not merely a straightforward aggregation of individual component optimizations. Instead, it requires carefully designed algorithms that account for complex interdependencies among components. This makes system-wide optimization a significantly more challenging task, necessitating advanced techniques to achieve effective and comprehensive improvements.

%

\section{Summary and Discussion}
\label{sec:self-optim-summary-discussion}

\lettrine[lines=3]{\initfamily\textcolor{darkgreen}{I}}{n} this chapter, we have explored the multidimensional landscape of self-optimization in intelligent agents, highlighting four crucial optimization spaces: prompt optimization, workflow optimization, tool optimization, and holistic agent optimization. Each dimension plays a unique yet interrelated role in enabling agentic systems to evolve towards greater autonomy, adaptability, and effectiveness.

Prompt optimization forms the foundational layer, directly influencing how agents query and interact with large language models. We detailed evaluation methods, including benchmark-based evaluations, LLM-as-a-Judge techniques, and human feedback mechanisms, emphasizing their varying trade-offs between automation, accuracy, and scalability. Effective prompt optimization thus involves careful selection and combination of these evaluation strategies to ensure efficient, context-aware agent communication.

Workflow optimization extends beyond single prompts, addressing the coordination of multiple LLM components. We formally introduced agentic workflows, discussing their node and edge components and the corresponding optimization search spaces. Graph-based, neural network-based, and code-based representations each offer distinct advantages, influencing the complexity and flexibility of workflow optimizations. As highlighted, recent advancements increasingly automate workflow topology and node parameter optimizations through reinforcement learning, generative AI, and self-supervised learning.

Tool optimization emphasizes enhancing how agents select, invoke, and create external tools. Two complementary strategies, learning from demonstrations and feedback, were discussed, alongside tool creation methods that automate tool synthesis and modular composition. The role of evaluation benchmarks and metrics—spanning tool invocation accuracy, selection precision, retrieval efficiency, and complex task planning—was underscored as critical for measuring and guiding tool optimization efforts.

Finally, holistic agent optimization integrates all dimensions into unified evolutionary frameworks. By addressing the complex interplay between prompts, workflows, and tools, holistic methods overcome local optima and facilitate comprehensive system improvements. We examined representative approaches like ADAS, AFLOW, and AlphaEvolve, along with self-referential methods such as the Darwin Gödel Machine and Alita, showcasing diverse strategies for systemic agent evolution.

Despite significant progress, several challenges and open research questions remain. Balancing trade-offs between computational cost, optimization accuracy, and scalability continues to be an ongoing challenge. Developing adaptive evaluation and optimization mechanisms that dynamically respond to evolving agent goals and environments presents a critical frontier. Future research directions may also explore deeper integrations between neural, symbolic, and evolutionary approaches to agent self-optimization.

In conclusion, advancing self-optimization capabilities is pivotal for realizing genuinely autonomous and adaptable intelligent agents. Continued innovation in optimization methods, coupled with systematic evaluation frameworks, will undoubtedly drive substantial advancements in the field, paving the way for agents capable of sophisticated, self-directed growth.

\chapter{Harnessing Large Language Models for Iterative Optimization}
\label{ch:llm-as-optimizer}


\lettrine[lines=3]{\initfamily\textcolor{darkgreen}{I}}{n} this chapter, we explore a rapidly evolving perspective: viewing large language models (LLMs) as powerful iterative optimizers. While classical optimization techniques typically rely on numerical gradients or heuristic search, recent studies demonstrate that LLMs offer a fundamentally different paradigm by leveraging their natural language processing capabilities for optimization tasks. We begin by contrasting traditional gradient-based and zeroth-order optimization methods with the emerging class of LLM-based optimizers, highlighting the unique advantages and complexities introduced by natural language. We then delve into iterative approaches that extend classical optimization concepts, such as random search, gradient approximations, and surrogate modeling, into the discrete and structured domains characteristic of LLM-driven workflows. Additionally, we examine critical hyperparameter considerations and discuss strategies for dynamic optimization across workflow depth and iterative refinements over time. Finally, we provide a theoretical lens, surveying recent insights into in-context learning, mechanistic interpretability, and limitations of LLM optimization in uncertain environments, bridging the gap between empirical success and theoretical understanding. Through this discussion, we aim to equip readers with a comprehensive foundation to understand and effectively leverage LLMs as versatile optimization tools within intelligent agent systems.


\section{Optimization Paradigms}
\label{sec:optim-paradigms}

\lettrine[lines=3]{\initfamily\textcolor{darkgreen}{O}}{ptimization} methods vary significantly in terms of the assumptions they make about the information available from the function they aim to optimize. Broadly speaking, these methods can be organized into three progressively broader categories: \emph{gradient-based optimization}, which explicitly utilizes gradients; \emph{zeroth-order optimization}, which estimates directions without gradient information; and the emerging paradigm of \emph{LLM-based optimization}, which transcends traditional numerical approaches by optimizing directly within complex, structured, and often linguistic spaces.

\begin{itemize}

\item \textbf{Gradient-Based Optimization.} These classical optimization approaches hinge upon the availability of precise gradient information. Techniques such as stochastic gradient descent (SGD) and Newton's method~\cite{bottou2018optimization} are prominent examples, favored for their efficiency and convergence properties. However, their reliance on differentiability significantly limits their applicability. For instance, discrete problems such as prompt optimization or structured workflow refinement, which frequently involve graph-based representations or categorical variables, pose fundamental challenges to gradient computation.

\item \textbf{Zeroth-Order Optimization.} Addressing scenarios where explicit gradients are difficult, costly, or impossible to compute, zeroth-order methods rely solely on function evaluations to guide their search directions~\cite{spall2005introduction}. Popular methods in this class include Bayesian optimization~\cite{frazier2018tutorial}, evolutionary strategies~\cite{hansen2016cma}, and finite-difference approximations~\cite{shi2023numerical}. These approaches excel when gradients are inaccessible, offering robust solutions for black-box optimization problems. Yet, their utility remains restricted to numerical objectives and structured spaces, proving less effective in dealing with inherently linguistic or unstructured domains.

\item \textbf{LLM-Based Optimization.} Emerging at the intersection of language modeling and optimization, large language model (LLM)-based optimization fundamentally expands the scope of optimization methods. Unlike traditional numerical optimization, LLMs operate naturally in complex linguistic environments, seamlessly integrating reasoning, iterative feedback, and human-like adaptability. This capability makes them particularly suited for tasks involving prompt refinement, adaptive workflow generation, and iterative decision-making that respond directly to linguistic cues and user feedback.

\end{itemize}

While gradient-based and zeroth-order methods predominantly address numerical problems, their foundational principles—iterative refinement, heuristic search, and adaptive learning—provide valuable conceptual insights for the nascent field of LLM-based optimization. Building upon this conceptual bridge, recent advances have integrated reinforcement learning techniques with LLMs, forming the foundation of sophisticated reasoning frameworks known as ``slow thinking'' models~\cite{jaech2024openai,o3mini,guo2025deepseek}. As these LLM-powered approaches evolve, we foresee a profound impact on the design of intelligent agents, empowering them to navigate increasingly complex and dynamic environments with enhanced flexibility, strategic planning, and decision-making prowess.

\section{Iterative Approaches to LLM Optimization} 
\label{sec:iterative-llm-optim}

\begin{figure}[!ht]
    \centering
    \includegraphics[width=\linewidth]{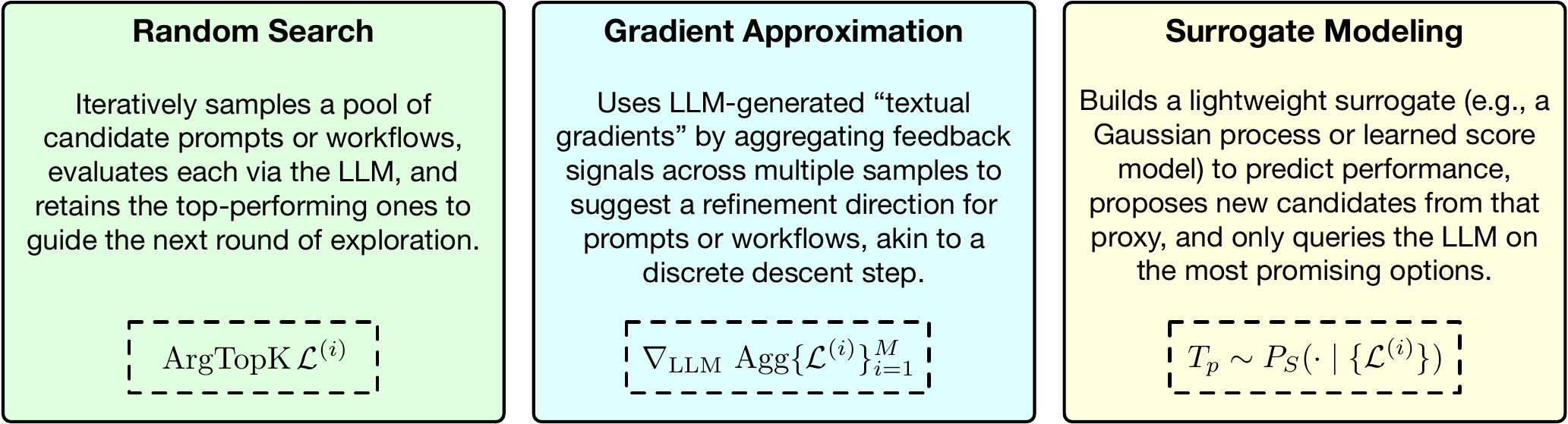}
    \caption{Three core iterative LLM‐based optimization strategies: (Left) Random Search samples and retains top‐K candidates; (Center) Gradient Approximation uses aggregated LLM ``textual gradients'' for discrete refinement; (Right) Surrogate Modeling fits a proxy to predict performance and only evaluates the most promising candidates.}
    \label{fig:iterative-llm-optimizer}
\end{figure}

\lettrine[lines=3]{\initfamily\textcolor{darkgreen}{S}}{ome} LLM-based optimization methods directly draw inspiration from classical optimization theory by adapting key components to address discrete and structured challenges. A central characteristic of these approaches is the iterative update step, in which model-generated modifications are selected from a range of possible improvements to refine the objective. Using the prompt optimization objective from~\Cref{def:prompt-opt} as a running example, a general iterative algorithm can be expressed as follows:
\begin{align}
\label{eq:prompt-update}
    \textbf{Sample: } & T\sim\mathcal{D}\\
    \textbf{Evaluation: } & \mathcal{L}(T; T_p)\gets \phi_\text{eval}\left(\phi_\text{exe}(Q, T_p), T\right) \\
    \textbf{Update: } & T_p^\prime \gets \phi_\text{opt}\left( \mathcal{L}(T; T_p)\right)
\end{align}
Here, the \emph{Sample} and \emph{Update} steps are defined based on the agent's task. In the simplest case, such as optimizing an instruction for binary classification of movie reviews, the objective $\mathcal{L}$ is measured by classification accuracy. In more complex agentic workflows, the decision variable may include prompts at different workflow stages, tool selections, agent topologies, or a combination thereof. As discussed in~\Cref{sec:optimization-space}, a common characteristic of these decision variables is their \emph{combinatorial} nature-such as the set of all strings from an LLM's vocabulary $\mathcal{V}$ or all possible role assignments for agents in a workflow. Since enumerating all possible solutions is often intractable, this necessitates designing approximate update steps $\phi_\text{opt}$, which we discuss next. Figure~\ref{fig:iterative-llm-optimizer} illustrates three strategies—random search, gradient approximation, and surrogate modeling—for iteratively refining LLM-based agents.

\paragraph*{Random Search} Early LLM-based optimization methods leveraged random search variants to optimize prompts in discrete natural language spaces~\cite{zhou2022large,chen2023teaching,zhuge2024gptswarm,fernando2023promptbreeder,ye2023prompt,sclar2023quantifying,ma2024large}. These methods often resemble evolutionary algorithms that iteratively sample candidate decision variables and select the top-performing ones from each iteration. The general formulation follows:
\begin{align}
\textbf{Sample: } & T \sim \mathcal{D} \\
\textbf{Evaluation: } & \mathcal{L}^{(i)} \gets \phi_\text{eval}(\phi_\text{exe}(Q, T_p^{(i)}), T), \quad i = 1, \dots, M \\
\textbf{Update: } & \{T_p^{(k)\prime}\}_{k=1}^{K} \gets \operatorname{ArgTopK}_{i \in [M]} \mathcal{L}^{(i)}, \\
\textbf{Replenishment (Optional): } & \{T_p^{(j)}\}_{j=K+1}^{M} \sim \operatorname{Mutate}(\{T_p^{(k)}\}_{k=1}^{K}).
\end{align}
We briefly override previous notations and let \( M \) denote the total number of candidate prompts sampled per iteration, and \( K \) (with \( K < M \)) control the number of top-performing candidates-selected with $\operatorname{ArgTopK}$ in our algorithm-retained for the next step. This algorithm can optionally incorporate a replenishment step to maintain diversity in the candidate pool across iterations. Random search methods are simple to implement, highly parallelizable, and particularly effective for single-prompt workflows. Beyond prompt optimization, they have also demonstrated strong performance in selecting in-context demonstrations~\cite{chang2022data,nguyen2023context}. However, their efficiency comes at a cost—each iteration requires  $O(M)$  parallel API queries, which can become prohibitively expensive for complex workflows involving multiple queries.

\paragraph*{Gradient Approximations} Several methods approximate gradient-based updates by iteratively refining solutions. For instance,~\cite{reid2023protegi,yang2024large,yuksekgonul2024textgrad} generate refinements at different workflow stages. StraGO~\cite{wu-etal-2024-strago} estimates descent directions using central-difference heuristics, while Trace~\cite{cheng2024trace} optimizes composed programs by modeling them as computation graphs, similar to backpropagation. The key analogy between gradient updates in continuous optimization and prompt-space refinement is the concept of a ``descent direction''—a systematic \emph{modification} of the decision variable to improve the objective. In contrast, random search methods propose new decision variables independently at each step, without accessing past update trajectories. Gradient-based approaches, by contrast, exploit this historical information, often leading to faster convergence. A general iteration for gradient approximation methods is given below:

\begin{align}
\textbf{Sample: } & T^{(i)} \sim \mathcal{D}, \quad i = 1, \dots, M \\
\textbf{Evaluation: } & \mathcal{L}^{(i)} \gets \phi_\text{eval}(\phi_\text{exe}(Q, T_p), T^{(i)}), \quad i = 1, \dots, M \\
\textbf{Gradient Approximation: } & g \gets \nabla_\text{LLM}\operatorname{Agg}\left(\mathcal{L}^{(1)}, \dots, \mathcal{L}^{(M)}\right)\\
\textbf{Update: } & T_p^\prime \gets \phi_\text{opt}(T_p, g),
\end{align}

where $M$ is the minibatch size, \( \operatorname{Agg}(\cdot) \) is an aggregation function that combines feedback signals (e.g., in numerical optimization, \( \operatorname{Agg} \) is typically the average operator), $\nabla_\text{LLM}$ represents an abstract ``LLM-gradient operator”~\cite{yuksekgonul2024textgrad} that generates textual refinement directions based on the feedback signal and the current minibatch (e.g., \emph{the agent should consider the edge case of …}). Additionally, $\phi_\text{opt}$ can be instantiated as an LLM query, allowing the agent to update its prompt based on $g$.

Compared to random search methods, gradient-based approaches offer two key advantages: they enable the incorporation of past refinement directions into $\phi_\text{opt}$, analogous to momentum-based techniques in first-order optimization algorithms~\cite{kingma2014adam,loshchilov2017decoupled}, and they facilitate backpropagation-like techniques for optimizing computation graphs~\cite{zhuge2024gptswarm,cheng2024trace,zhang2024revolve}, making them particularly effective for multi-stage workflows with interdependent optimizable modules. However, this flexibility comes at the cost of increased design overhead, such as the need for meta-prompts to aggregate feedback and apply refinement directions. We further discuss the feasibility of using LLMs to optimize these \emph{hyperparameters} below. Some approaches also explored direct gradient-based optimization of soft prompts~\cite{li2021prefix,lester2021power,diao2022black}. While effective for simple input-output sequence learning, these methods struggle with multi-step workflows and sequential decision-making~\cite{wu2023autogen, wang2023promptagent}.

Finally, while these methods leverage first-order optimization insights, the extension of second-order techniques (e.g., quasi-Newton methods) to LLM-based optimization remains largely unexplored. Fortunately, recent works such as Revolve~\cite{zhang2024revolve} have taken a step in this direction by introducing a structured approach for second-order optimization, modeling the evolution of response patterns over multiple iterations. By incorporating higher-order refinements, Revolve enables more stable and informed optimization, effectively mitigating stagnation in complex tasks. We are also excited by emerging trends in leveraging inference-time compute~\cite{jaech2024openai,guo2025deepseek} to incorporate historical refinement directions and investigate the benefits of momentum.
    
\paragraph*{Bayesian Optimization and Surrogate Modeling} While the aforementioned approaches achieved significant progress in LLM-based optimization, they often entail substantial financial and environmental costs due to the high number of required LLM interactions. Moreover, these methods can be sensitive to noise, and the optimization landscape of discrete prompts, among other decision variables, remains poorly understood~\cite{min2022rethinking,wei2023larger}. Under these constraints, Bayesian Optimization (BO) emerges as a compelling alternative, as it builds a noise-resilient surrogate model of the optimization objective:
\begin{align}
\textbf{Sample: } & T \sim \mathcal{D} \\
\textbf{Proposal: } & \{T_p^{(i)}\}_{i=1}^{M} \sim S.\operatorname{Propose}\\
\textbf{Evaluation: } & \mathcal{L}^{(i)} \gets \phi_\text{eval}(\phi_\text{exe}(Q, T_p^{(i)}), T), \quad i = 1, \dots, M \\
\textbf{Update: } & S \gets S.\operatorname{UpdatePrior}( \{\mathcal{L}^{(i)}\}_{i=1}^M, \{T_p^{(i)}\}_{i=1}^{M}),
\end{align}
where \( S \) represents a probabilistic surrogate model of the optimization objective, equipped with a proposal operator (e.g., posterior sampling from a Gaussian Process BO procedure~\cite{frazier2018tutorial}) and an update mechanism based on observed evidence from prompt evaluations. For instance, MIPRO~\cite{opsahl-ong-etal-2024-optimizing} employs a Tree-Structured Parzen Estimator as its surrogate~\cite{watanabe2023tree}, while PROMST~\cite{chen-etal-2024-prompt} trains a score-prediction model to guide prompt tuning. Leveraging a surrogate model for LLM-based optimization aligns with the emerging trend of amortized optimization for non-differentiable objectives~\cite{amos2023tutorial}. For instance,~\cite{paulus2024advprompter} trains a prompt-generator LLM to amortize the computational cost of instantiating a beam search problem for discovering jailbreak attack prefixes.

Finally, several other works fit an additional lightweight module-such as a Bayesian belief posterior or a utility function-from LLM outputs, to aid the optimization of domain-specific workflows, such as decision-making and multi-agent negotiations~\cite{liu2025dellma,hua2024game}. This type of amortized methods-those that fit a parameterized model that is reusable for unseen inputs-have found increasing usage in LLM-based optimization, such as jailbreaking~\cite{zhu2024advprefix,paulus2024advprompter}.

\section{Optimization Hyperparameters} 
\label{sec:optim-hyperparams}

\lettrine[lines=3]{\initfamily\textcolor{darkgreen}{S}}{imilar} to traditional optimization, LLM-based methods are highly sensitive to hyperparameters that influence search efficiency and generalization. A key consideration in gradient-based LLM optimizers is the choice of the aggregation function $\text{Agg}(\cdot)$, which determines how textual feedback is synthesized to guide prompt updates. An improper choice can lead to loss of critical information or misalignment in iterative refinements. Additionally, \cite{cheng2024trace} introduces a ``whiteboard'' approach, where an LLM program is decomposed into human-interpretable modules. However, design choices in structuring such modular workflows remain largely unexplored, which poses an open challenge for optimizing LLM-driven decision-making pipelines.

Hyperparameters in LLM optimization often parallel those in numerical optimization. For example, batch size plays a crucial role: just as minibatch updates enhance stability and efficiency in classical optimization, LLM-based approaches like TextGrad~\cite{yuksekgonul2024textgrad} aggregate feedback across multiple generated samples before making updates. Another key factor is momentum—while it stabilizes updates in gradient-based methods by incorporating past gradients, LLM-based optimizers similarly leverage historical refinements to improve performance over time ~\cite{yuksekgonul2024textgrad,cheng2024trace}. Despite progress in numerical optimization, hyperparameter selection for LLM-based optimizers remains largely heuristic, often relying on ad hoc, trial-and-error tuning.

In agentic system design, hyperparameters proliferate across various components, including role assignments of agents, selection of in-context demonstrations, and scheduling of tool invocations. Each of these choices has a profound impact on downstream performance, yet principled methods for optimizing them remain underdeveloped. While traditional hyperparameter tuning techniques, such as grid search and Bayesian optimization, can be applied to discrete LLM-driven workflows, their computational cost scales poorly due to the high variance in language model outputs. Additionally, the combinatorial nature of these hyperparameters, where agent configurations, prompting strategies, and reasoning structures interact in complex ways, makes an exhaustive search infeasible. Recent work has attempted to address this challenge by embedding agentic workflows into structured frameworks such as finite state machines~\cite{wu2024stateflow}, optimal decision theory~\cite{liu2025dellma}, and game theory~\cite{hua2024game}. However, these approaches often fail to generalize across diverse environments. A promising direction for addressing these challenges is meta-optimization, where LLMs are used to optimize their own hyperparameters and decision-making strategies. For example, an LLM-based optimizer can iteratively refine its own prompting strategies by treating past decisions as experience, akin to learned optimizers in deep learning~\cite{metz2022practical}. Moreover, amortized approaches train auxiliary models to predict effective hyperparameters, which can reduce the computational cost of exhaustive search~\cite{opsahl-ong-etal-2024-optimizing, chen-etal-2024-prompt}. While these techniques offer exciting possibilities, they also introduce new challenges, such as balancing exploration with exploitation in adaptive tuning and ensuring generalization across diverse optimization tasks. Investigating principled meta-optimization strategies tailored to LLM-driven workflows remains a critical area for future research.

\section{Dynamic and Iterative Optimization in LLM Workflows} 
\label{sec:optim-depth-time}


\lettrine[lines=3]{\initfamily\textcolor{darkgreen}{C}}{onventional} optimization techniques typically view parameter updates as static procedures within clearly defined boundaries. In stark contrast, LLM-based optimization embraces a dynamic approach, adapting not only across the hierarchical depth of workflows but also iteratively over successive time steps. This dual perspective—considering both depth (single-pass sequential execution) and temporal dimensions (iterative refinement)—is central to leveraging the full potential of LLM-driven optimization methods.

At a fundamental level, optimizing across depth resembles how feedforward neural networks process data: inputs traverse multiple layers, each performing incremental transformations that collectively refine the output. Analogously, LLM-based workflows pass sequentially through various modules or stages, each step benefiting from structured reasoning and cumulative context provided by previous modules. Such depth-wise optimization naturally aligns with classical computational architectures, enabling well-defined, modular improvements at each workflow stage.

However, the true strength of LLM-based optimization emerges clearly when considering its temporal aspect—iterative optimization over multiple cycles. In this respect, LLM-based methods draw conceptual parallels with recurrent neural architectures, such as RNNs and Universal Transformers~\cite{dehghani2018universal}, which iteratively refine their internal representations based on evolving inputs and previous states. Similarly, LLM-driven optimizers dynamically update their decision-making strategies by incorporating feedback from prior outcomes, progressively enhancing their performance over multiple cycles. StateFlow~\cite{wu2024stateflow} exemplifies this iterative refinement by integrating previous execution outcomes and feedback loops into workflow decisions, dynamically adjusting strategies to evolving task contexts.

Yet, despite these promising analogies, there remain significant untapped opportunities to incorporate well-established optimization strategies from engineering domains into the LLM optimization landscape. Techniques such as checkpointing—saving intermediate states to efficiently revisit previous optimization states—and truncated backpropagation—focusing computational resources on the most recent and relevant optimization steps—have demonstrated substantial improvements in traditional deep learning and numerical optimization contexts~\cite{hascoet2006enabling, shaban2019truncated}. Exploring analogous implementations of these methods within the LLM optimization paradigm could enhance computational efficiency, stability, and effectiveness, especially in complex or long-term iterative optimization tasks.

Moreover, the dynamic and iterative optimization process introduces unique challenges, such as balancing exploration and exploitation, managing cumulative uncertainty across iterations, and ensuring long-term coherence in evolving strategies. Future research must investigate methods for systematically integrating these temporal considerations into LLM-driven workflows. This exploration could include adaptive strategies that dynamically allocate computational resources based on uncertainty and historical performance, hybrid optimization techniques combining classical numerical methods with LLM-driven updates, and frameworks explicitly modeling temporal dependencies and long-range coherence.
In summary, while depth-wise optimization aligns naturally with conventional computational strategies, the temporal dimension opens rich, largely uncharted territory. By embracing iterative refinement over time and carefully integrating lessons from classical engineering optimization, LLM-based methods stand poised to achieve unprecedented levels of adaptability, efficiency, and strategic depth in complex agentic applications.

\section{Theoretical Insights into Transformer Optimization}
\label{sec:theoretical-insights-llm-optim}

\lettrine[lines=3]{\initfamily\textcolor{darkgreen}{R}}{ecent} research increasingly highlights transformers not merely as sophisticated language processors but as systems inherently capable of performing complex, optimization-like computations. Despite their remarkable empirical successes, our theoretical understanding of how and why transformers can function effectively as optimizers remains incomplete. This section synthesizes key theoretical advancements that bridge this gap and explores fundamental concepts underpinning transformers' optimization abilities.

\paragraph*{In-Context Learning}
In-context learning, particularly prominent in few-shot learning scenarios~\cite{brown2020language}, provides essential insights into transformers as optimization mechanisms. Early theoretical work by~\cite{garg2022what} illustrated that transformers can internally replicate various regression and learning models, including linear regression, decision trees, and shallow neural networks, purely through their attention mechanisms. Subsequent studies~\cite{akyurek2023what, pmlr-v202-von-oswald23a, fu2024transformers} further strengthened this view by explicitly demonstrating how transformers can mimic iterative optimization methods, including gradient descent and higher-order update strategies. Nonetheless, while such theoretical analyses reveal profound computational capacities, they only partially explain in-context learning behaviors exhibited by large-scale language models, especially given their discrete, linguistic input-output settings. Empirical investigations~\cite{min2022rethinking,xie2022an,wei2023larger} have thus attempted to decode in-context generalization mechanisms. For example, \citet{xie2022an} proposed interpreting in-context learning as akin to implicit Bayesian inference performed by hidden Markov models, whereas others~\cite{min2022rethinking,wei2023larger} questioned traditional assumptions, suggesting alternative explanations. Consequently, while in-context learning remains pivotal to transformers' self-optimization abilities~\cite{wei2022emergent}, it continues to challenge comprehensive theoretical characterization.

\paragraph*{Mechanistic Interpretability}
Complementary to theoretical analyses, mechanistic interpretability seeks to illuminate internal transformer operations by isolating and understanding computational substructures, often referred to as circuits, responsible for specific behavioral outputs. Initial efforts successfully mapped such circuits in smaller models, like GPT-2, to simple language tasks~\cite{wang2022interpretability,hanna2023does,conmy2023towards}. More recent advancements extended these methods to large, frontier-class models, employing techniques such as sparse autoencoders to detect semantically meaningful and interpretable components within transformer architectures~\cite{lieberum2024gemma,templeton2024scaling,gao2024scaling,marks2024sparse}. Although these approaches significantly improved the transparency and controllability of transformer computations, they also revealed complexities—such as entangled beneficial and harmful behaviors arising from many-shot conditioning~\cite{anil2024many}—underscoring ongoing challenges in achieving safe and reliable optimization through transformers.

\paragraph*{Challenges Under Uncertainty}
Despite their sophisticated reasoning capabilities in structured contexts, LLM-based optimization frameworks exhibit notable shortcomings in environments characterized by uncertainty and stochasticity~\cite{laskin2023incontext,nie2024evolve,krishnamurthy2024can,monea2024llms}. Particularly, recent findings by \citet{liu2025dellma} emphasize that LLMs often falter in optimal exploration and robust decision-making under uncertain conditions. These limitations highlight critical caution points for deploying transformer-based optimization solutions in real-world, dynamic settings where adaptability, exploration, and robustness to uncertainty are indispensable.

Transformers have redefined optimization paradigms by integrating linguistic reasoning, adaptive learning from context, and iterative refinement. Yet, despite their transformative empirical impacts, critical theoretical questions about their inherent optimization abilities—especially around in-context learning and decision-making under uncertainty—remain open and necessitate deeper exploration.

\section{Summary and Discussion}
\label{sec:summary-llm-optim}

\lettrine[lines=3]{\initfamily\textcolor{darkgreen}{T}}{his} chapter has explored the transformative paradigm of using large language models (LLMs) as versatile optimization tools, moving beyond traditional numerical methods into structured, linguistic, and iterative optimization spaces. We began by comparing foundational optimization paradigms—gradient-based, zeroth-order, and LLM-based optimization—highlighting the unique strengths and applications of each. By positioning LLM optimization within this broader context, we provided a clear conceptual roadmap that illuminates both the innovative potential and inherent complexities involved in optimizing through natural language.

The iterative approaches to LLM optimization—including random search, gradient approximations, and Bayesian surrogate modeling—demonstrate how classical optimization concepts can be effectively adapted to the discrete, structured spaces where LLMs excel. These methods leverage iterative refinement, human-like reasoning, and natural language-based feedback mechanisms, enabling sophisticated optimization processes beyond traditional numeric domains.

However, significant challenges remain in fully realizing the potential of LLM-driven optimization. One primary concern is the computational and financial cost associated with iterative LLM queries, particularly when applied to large-scale or complex agentic workflows. Addressing these scalability issues requires novel strategies to reduce the number of LLM interactions, enhance computational efficiency, and ensure robustness against inherent variability and noise in model outputs.

Another crucial area for future development involves hyperparameter tuning and meta-optimization strategies. Current approaches often rely on heuristic methods, lacking principled frameworks for systematically optimizing hyperparameters such as prompt design, workflow structure, and feedback aggregation methods. Advancements in meta-optimization, adaptive hyperparameter tuning, and amortized optimization techniques promise substantial improvements in both efficiency and generalization, paving the way for more robust and adaptable optimization workflows.

Furthermore, effectively leveraging iterative refinement across depth and time remains a relatively underexplored frontier. Integrating classical engineering optimization strategies into LLM-driven frameworks presents promising avenues for improving long-term coherence, managing computational resources, and enhancing adaptability in evolving environments. Additionally, modeling and addressing uncertainty within iterative optimization processes is critical, as current models frequently struggle with exploration-exploitation trade-offs and robust decision-making in dynamic settings.

Finally, deepening our theoretical understanding of transformers as optimization agents is imperative. While empirical evidence supports the powerful optimization capabilities of LLMs, theoretical explanations—particularly those relating to in-context learning, mechanistic interpretability, and decision-making under uncertainty—remain incomplete. Advancing theoretical insights will be crucial not only for improving our understanding of transformer optimization dynamics but also for ensuring the safe, reliable, and explainable deployment of LLM-based systems.

While LLMs represent a revolutionary step in optimization paradigms, addressing these outlined challenges through systematic research will be essential to harness their full potential effectively. Continued interdisciplinary efforts integrating insights from classical optimization theory, machine learning, natural language processing, and decision-making theory will drive future innovations, ensuring that LLM-based optimization becomes a robust and reliable cornerstone of intelligent system design.

\chapter{Online and Offline Agent Self-Improvement}
\label{sec:selfevo}

\begin{figure}[t]
    \centering
    \includegraphics[width=0.9\linewidth]{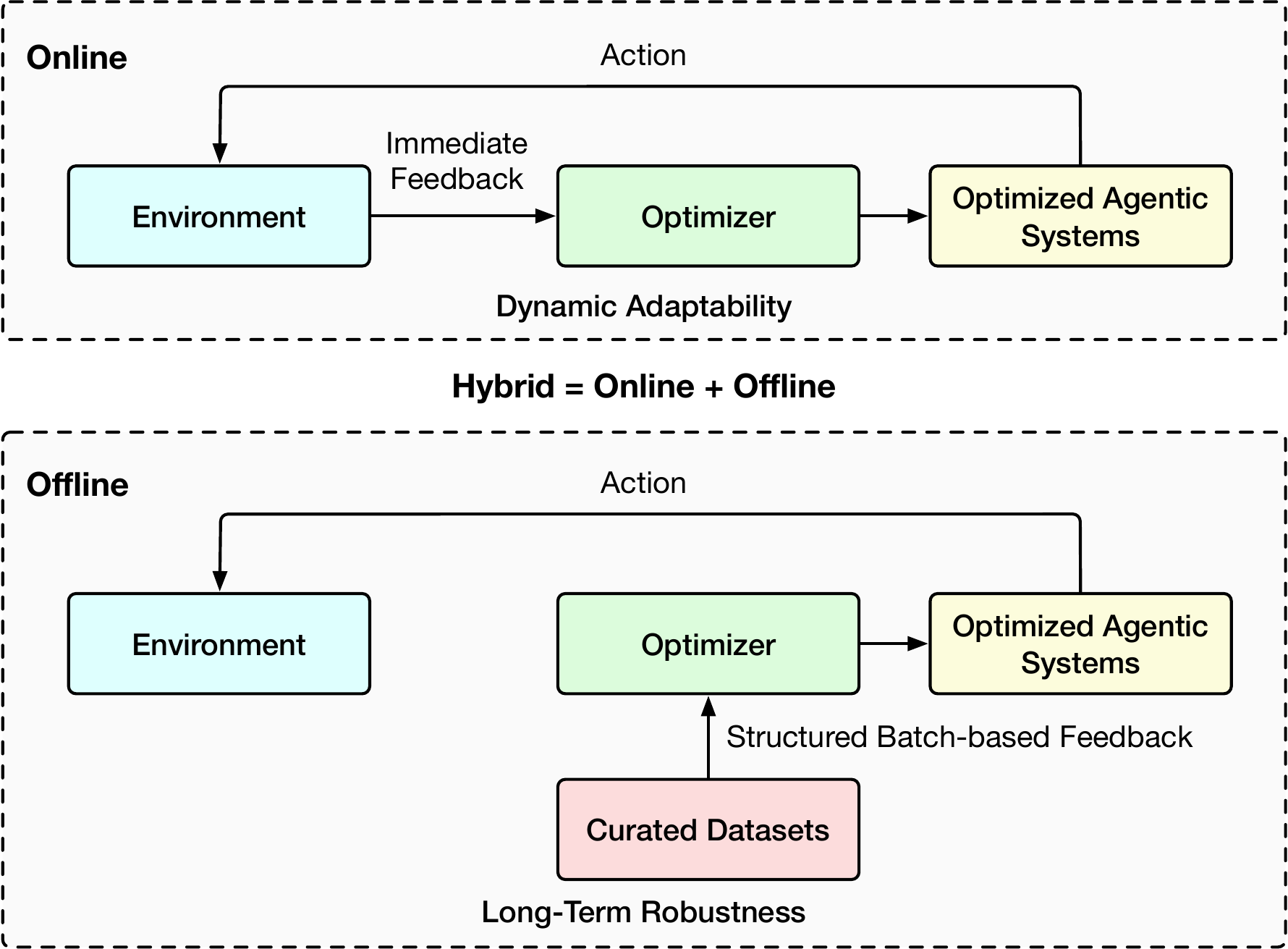}
    \caption{An illustration of self-improvement under three different utilization scenarios, including Online, Offline, and Hybrid self-improvement.}
    \label{fig:online-offline-scenarios}
\end{figure}


\lettrine[lines=3]{\initfamily\textcolor{darkgreen}{I}}{n} the pursuit of self-improvement, intelligent LLM-based agents leverage optimization as both a mechanism to refine individual components—such as prompt design, workflow orchestration, tool utilization, reward function tuning, and even the optimization algorithms themselves—and as a strategic framework to ensure these improvements align toward coherent overall performance. For instance, optimizing the reward function and prompt design in isolation might yield conflicting outcomes; however, a coordinated strategy can align these optimizations to maintain consistency and maximize overall effectiveness. In practice, an agent may simultaneously adjust its prompt templates and its criteria for success so that both changes complement each other rather than conflict. This holistic view of optimization prevents sub-components from working at cross purposes and instead directs all changes toward a common goal of performance enhancement.

We categorize agent self-evolution into two primary paradigms: \emph{online} and \emph{offline} self-improvement. \emph{Online self-improvement} refers to real-time adaptation, where an agent continuously adjusts its behavior based on immediate feedback during operation. \emph{Offline self-improvement}, in contrast, involves structured, batch learning phases where the agent is improved through deliberate training on collected data or simulations. In addition, we explore hybrid strategies that integrate both approaches to maximize efficiency and adaptability. These hybrid methods aim to combine the strengths of offline training's stability with online learning's agility, yielding agents that are both robust and dynamically responsive (Figure~\ref{fig:online-offline-scenarios}). In the following sections, we delve into each paradigm and then discuss how blending them can lead to continuous and balanced self-improvement.

\section{Online Agent Self-Improvement}
\label{sec:online-agent-self-improve}


\lettrine[lines=3]{\initfamily\textcolor{darkgreen}{O}}{nline} self-improvement refers to real-time optimization in which an agent dynamically adjusts its behavior based on immediate feedback. This paradigm ensures that agents remain responsive to evolving environments by continuously optimizing key performance metrics—such as task success rate, latency, cost, and solution quality—within an iterative feedback loop. An LLM-based agent operating in a live setting (for example, a chatbot deployed to users) can use online self-improvement to update its strategies on the fly, learning from each interaction or correcting mistakes as they happen. Online optimization is particularly effective in applications that require dynamic adaptability, such as real-time decision-making, personalized user interactions, and automated reasoning systems that deal with changing information. We can classify key strategies in online self-improvement into four categories: \emph{Iterative Feedback and Self-Reflection}, \emph{Active Exploration in Multi-Agent Systems}, \emph{Real-Time Reward Shaping}, and \emph{Dynamic Parameter Tuning}. Below, we examine each in turn.


\paragraph*{Iterative Feedback and Self-Reflection}  
These methodologies focus on enabling agents to critique and refine their own outputs iteratively using feedback loops~\cite{Shinn2023ReflexionLA, madaan2024self-refine, yaotree, yao2022react, light2023manlet, wang2023voyager}. For example, Reflexion~\cite{Shinn2023ReflexionLA} is a framework where a language agent doesn't update its weights but instead uses verbal self-feedback to correct errors: the agent generates an answer, evaluates its correctness or mistakes through a reflection step (often by analyzing an outcome or an external hint), and then attempts an improved answer in light of that self-critique. Similarly, Self-Refine~\cite{madaan2024self-refine} allows the model to iteratively refine a draft solution by identifying flaws in its initial output and revising it – effectively performing its own review and edit cycle. Tree-of-Thoughts~\cite{yaotree} introduces a strategy where the agent explores multiple reasoning pathways (a tree of possible thought sequences) rather than a single chain-of-thought; the model generates different solution paths, evaluates them (possibly pruning those that seem unpromising), and backtracks or branches out as needed, thereby increasing the chance of finding a correct or high-quality solution. The ReAct framework~\cite{yao2022react} combines chain-of-thought ``Reasoning'' with ``Acting'' steps: the agent interleaves reasoning about a problem with taking actions (such as querying a tool or environment) and can revise its approach based on the observed results of those actions. This means the model isn't stuck with its first plan – it can adjust mid-way if an action's outcome indicates a need for correction. In addition, some methods rely on self-consistency during inference~\cite{wang2023selfconsistency}: the agent generates multiple reasoning traces for the same query and then selects the most consistent or frequent answer among them, which empirically tends to be more reliable. Another complementary approach is to use a process reward model (PRM)~\cite{light2023manlet} to evaluate each step of the agent's reasoning. With a PRM, the agent can generate several candidate solution steps or full solutions, have the PRM score the logical correctness of each step in the chain, and then choose or refine the best-scoring solution. By verifying intermediate reasoning steps (rather than only the final answer), PRM-based feedback helps catch mistakes early and avoid propagating errors. Voyager~\cite{wang2023voyager} provides a practical example of iterative self-improvement in an embodied setting: it uses an LLM agent to autonomously explore a task environment (Minecraft, in this case), reflect on failed attempts, and store successful strategies in a skill library for future use. Collectively, all these frameworks share a common theme: they create feedback loops where the agent's own outputs are analyzed and used to make immediate improvements without requiring a separate offline retraining cycle. This reduces error propagation and allows for rapid, on-the-fly adaptation to new challenges or corrections of mistakes.


\paragraph*{Active Exploration in Multi-Agent Systems}
In multi-agent or multi-role frameworks, agents actively explore and collaboratively search for better behaviors or solutions by exchanging feedback with each other in real time~\cite{hong2023metagpt, li2023camel, qian2024chatdev, shen2024hugginggpt}. For instance, MetaGPT~\cite{hong2023metagpt} organizes multiple LLM-based agents into specialized roles (such as a product manager, engineer, tester, etc.) that communicate and coordinate on a task (like software development). Each agent contributes its expertise, and through continuous dialogue and mutual feedback, the group converges toward an improved result. Likewise, CAMEL~\cite{li2023camel} demonstrates a dual-agent conversation approach where two AI agents (e.g., one playing the user and one the assistant with a given goal) interact to progressively refine a solution or idea – the ``user'' agent can challenge or provide new information, and the ``assistant'' agent adapts its responses accordingly, resulting in deeper exploration than a single agent alone might achieve. ChatDev~\cite{qian2024chatdev} simulates a complete software development team with multiple agent roles (system architect, developer, tester, etc.) all implemented as LLMs; these agents generate code, review each other's work, and correct errors in an online loop, mirroring how a human team might incrementally improve a software project. Meanwhile, HuggingGPT~\cite{shen2024hugginggpt} uses a central LLM as a controller to orchestrate a fleet of specialized models (for example, calling a vision model for an image task or a speech model for audio input). The LLM dynamically routes subtasks to appropriate expert models and then integrates their feedback or results into its reasoning. This kind of multi-agent or multi-model exploration allows the system to handle complex tasks by learning which agent or tool is best suited for each sub-problem and adjusting the workflow in real time. Through active exploration and division of labor, multi-agent systems can discover novel patterns of problem-solving and iteratively refine their collective output based on continuous inter-agent feedback and validation.


\paragraph*{Real-Time Reward Shaping} 
Instead of relying solely on fixed or pre-defined reward signals, some frameworks incorporate immediate feedback from the environment or the agent's own evaluations to adjust the reward function and policy on the fly~cite{zelikman2023self, zelikman2024quiet, xie2023text2reward, zhang2024extracting}. This approach is essentially adaptive reward optimization in real time. For example, Zelikman et al.~\cite{zelikman2023self, zelikman2024quiet} propose techniques where a language model can generate its own intermediate checks or self-queries during reasoning, and use the outcomes of those checks as internal reward signals. In these methods, if the model's intermediate reasoning step produces a useful insight or question (one that eventually leads to a correct answer), the model ``rewards'' itself by reinforcing that line of thought. Such immediate self-feedback mechanisms encourage the agent to favor thought processes that seem promising or correct. Another notable work is Text2Reward by \citet{xie2023text2reward}, which is a data-free framework using an LLM to automatically generate dense reward functions from textual task descriptions. In a traditional RL setting, designing a reward function can be hard (especially if only sparse success signals are available), but Text2Reward leverages the language model's knowledge to shape a more informative reward signal. For instance, given a description of the goal, the LLM can suggest intermediate milestones or subgoals and assign reward values to them, providing the agent with a richer gradient of feedback as it works toward the final goal. This real-time shaping means the agent doesn't have to rely purely on eventual success or failure – it gets guidance at each step. Similarly, other research (e.g., \citet{zhang2024extracting}) explores extracting heuristic reward signals from the agent's own reasoning trajectory or from external observations on the fly. In such approaches, if the agent's current action or output yields certain desirable patterns (for example, matching a known criterion of a good solution), a reward adjustment is made immediately to reinforce that behavior. By continuously calibrating what the agent is striving to maximize (its reward function) based on recent interactions, real-time reward shaping allows the agent to adapt its objectives dynamically. This leads to behavior that can balance trade-offs (like accuracy vs. speed or success vs. resource cost) in response to the current context and feedback, rather than being locked to a static reward definition decided beforehand.

\paragraph*{Dynamic Parameter Tuning}
In this category, agents autonomously update their internal parameters and configurations during runtime to optimize performance. These parameters can include prompt templates, thresholds for invoking tools or external APIs, search heuristics (like how broad or deep a lookahead to perform in planning tasks), sampling settings for generative models (temperature, top-$k$ values), and more. The key idea is that the agent itself monitors its performance and tweaks these knobs in real time, rather than a human manually doing so. Because direct gradient-based tuning of large model parameters online is often impractical, many approaches rely on gradient-free optimization or surrogate gradient methods that operate on these higher-level parameters. For instance, an agent might use evolutionary strategies or bandit algorithms to try slight variations in its prompt or tool-selection strategy and keep the changes that improve outcomes. A concrete example is the Self-Steering Optimization (SSO) algorithm proposed by \citet{xiang2024aligning}. SSO enables a large language model to align and improve itself autonomously by generating high-quality preference signals during training. In practice, SSO has the model produce its own comparisons or rankings of outputs (simulating what human feedback might say) and uses those as a training signal to adjust the model's policy. By doing so iteratively, it eliminates the need for manual human annotation of preferences. Moreover, SSO is designed to keep this training process on-policy – meaning the preferences are generated based on the model's current behavior, ensuring the model is always learning from feedback relevant to what it actually does. This kind of dynamic tuning not only applies to alignment; an agent could similarly self-tune other parameters. For example, if an agent notices it's making too many mistakes using a certain tool, it could raise the confidence threshold required before that tool is invoked, or if it finds its responses are too slow, it might shorten its chain-of-thought or use a faster reasoning heuristic. By autonomously adjusting such internal dials, the agent optimizes both its computational efficiency (e.g. using resources judiciously) and decision accuracy (e.g. choosing the right strategy for the context) on the fly. The result is seamless adaptation to evolving conditions without waiting for an offline re-training phase.

Online self-improvement thus fosters a continuously evolving agent framework where learning is embedded within task execution. The agent becomes highly adaptive in real time, capable of improving with each interaction or feedback instance. This leads to more personalized and context-aware behavior (since the agent can optimize for the specific user or environment it's currently dealing with), and it enhances robust problem-solving by quickly addressing errors or changes. However, it's worth noting that while online methods offer agility, they also demand careful design to avoid instability (for example, an agent might ``chase'' recent feedback too strongly and forget older knowledge, a phenomenon sometimes called catastrophic forgetting or drift). In summary, online self-improvement provides powerful tools for immediate, user-centric optimization and rapid learning, especially in dynamic settings, complementing the more deliberate improvements made via offline methods.

\section{Offline Agent Self-Improvement}
\label{sed:offline-agent-self-improve}

\lettrine[lines=3]{\initfamily\textcolor{darkgreen}{O}}{ffline} self-improvement, in contrast to online methods, leverages structured batch-based optimization carried out in dedicated training phases. In this paradigm, an agent improves by training on accumulated data or simulated experience in a disconnected setting, rather than while actively interacting with users or an environment. The agent typically enters an offline ``learning mode'' (which could range from fine-tuning a model on new examples to entirely retraining certain components) and, once that session is complete, it updates its capabilities for the next deployment. This approach systematically enhances the agent's generalization capabilities by exposing it to curated, high-quality datasets and scenarios without the time or safety constraints of a live environment~\cite{llmstat, dataset-llmsql, dataset-llmvis, dataset-llmragqa, dataset-llmvqa}. For example, a conversational LLM agent might be improved offline by fine-tuning it on a large repository of solved dialogue tasks or by training on logs of past conversations and their ideal resolutions. Because offline training doesn't need to be instantaneous, it can employ computationally intensive methods and large-scale data that would be impractical to use during real-time interaction. Key strategies in offline self-improvement include \emph{Batch Parameter Updates and Fine-Tuning}, \emph{Meta-Optimization of Agent Components}, and \emph{Systematic Reward Model Calibration}.

\paragraph*{Batch Parameter Updates and Fine-Tuning} 
 In this category, an agent undergoes extensive weight updates using supervised learning or reinforcement learning (RL) techniques across large datasets, typically iterating over the data for multiple epochs to gradually minimize errors. This could involve classical fine-tuning of a pre-trained LLM on domain-specific data or difficult tasks to improve its performance in those areas. One common approach is Reinforcement Learning from Human Feedback (RLHF)~\cite{ouyang2022training}, where an agent or model is trained (offline) on a static collection of human preference data: the agent's policy (the LLM's behavior) is adjusted so that it produces outputs preferred by humans, as judged by a reward model trained on those preferences. Such RLHF fine-tuning greatly improves alignment with what users find helpful, but it's done in batch after collecting preferences, not during live use. In addition, offline improvements often involve bolstering the agent's ability to retrieve and use knowledge. For instance, retrieval-augmented generation (RAG) techniques are frequently integrated into training to enhance the agent's contextual understanding and long-term memory~\cite{zhang2024xlam, hu2024automated}. With RAG, the agent learns to query an external knowledge source (like a document database or the web) and incorporate the retrieved information into its responses. During offline training, the agent can practice this behavior using supervised signals (learning when and what to retrieve for a given query) so that at deployment it can seamlessly pull in relevant information it didn't explicitly store in its own weights. By fine-tuning on large-scale datasets that cover a wide range of scenarios (e.g., large collections of Q\&A pairs, code libraries, or multimodal data for vision-and-language tasks), the agent improves generalization – it becomes capable of handling new inputs more robustly because it has seen many variations during training. Moreover, because these updates are done in a controlled setting, developers can use computationally heavy algorithms (like large-scale gradient descent, hyperparameter sweeps, or simulation of many trajectories in an environment) to squeeze out as much performance as possible. The outcome of batch training is a new version of the agent (or a component of the agent) that performs better on average, which can then be deployed. Notably, methods like this allow for optimizing performance and stability before the agent faces real-world tasks, reducing the chance of failure in mission-critical applications. For example, if an LLM-based agent will be used for medical advice, offline training on a vetted medical Q\&A dataset can significantly improve its accuracy and safety, ensuring a strong baseline model is in place.

\paragraph*{Meta-Optimization of Agent Components} 
In these approaches, offline training is not just used to improve the agent's task performance, but also to refine the optimization processes and hyperparameters that the agent itself uses. In other words, the agent learns how to learn better. This often involves meta-learning techniques, where the objective is to train the agent (or a meta-controller) on a variety of tasks such that it can quickly adapt to new tasks with minimal additional training. One example is learning an initialization or update rule that works well for a family of problems – the famous Model-Agnostic Meta-Learning (MAML) algorithm is a classic (non-LLM) example where a model is trained to have parameters that can be very quickly fine-tuned to new tasks~\cite{finn2017model}. In the context of LLM-based agents, researchers have explored similar ideas. For instance, Zelikman et al.~\cite{zelikman2023self, zelikman2024quiet} have demonstrated techniques where the agent, through offline self-training, improves its own reasoning strategies. In one of their approaches, a language model generates rationales for solving problems and uses the correctness of the answer to adjust how it produces these rationales in the future – effectively the model is tweaking its internal ``algorithm'' for reasoning. Such meta-optimization might also involve hyperparameter tuning (finding the best learning rate, prompt length, etc.) or even neural architecture search (altering components of the model or agent architecture) done offline. For example, an agent could be put through an offline phase where different prompting techniques or tool-use strategies are tried, and the best-performing strategies are then baked into the agent's policy going forward. Unlike standard fine-tuning which targets performance on a fixed task distribution, meta-optimization seeks to give the agent a form of learning-to-learn capability: after meta-training, when the agent encounters a new kind of problem, it can adapt more quickly or execute an improved optimization algorithm internally.
The overarching benefit is that the agent not only becomes better at tasks, it becomes better at becoming better, so to speak, which is a powerful form of self-improvement.

\paragraph*{Systematic Reward Model Calibration} 
The offline setting is ideal for carefully calibrating and improving reward models, which are used to guide agent behavior. A reward model in the context of LLM agents might score how well an output aligns with human preferences or task objectives. Offline calibration involves training these models on collected data (for example, human rankings of outputs, or known correct solutions) and tuning them to accurately reflect the true goals we care about. One straightforward example is again from RLHF: a reward model is trained on a dataset of comparisons (which of two responses a human prefers) and then refined until it reliably predicts human preference. Offline, we can put significant effort into this – using large annotated datasets and powerful training methods – to get a high-quality reward function. Recent advanced techniques have proposed going beyond simple pairwise preference training. For instance, LiRE (Listwise Reward Enhancement) by \cite{zhu2024lire} is a framework that optimizes the reward model using a listwise approach: instead of looking at one or two outputs at a time, the reward model is trained on sets of outputs together, so that it learns to rank a collection of candidate responses appropriately. This listwise training provides a richer signal – it ensures the reward model's scores are calibrated not just in isolation but relative to a distribution of possible outputs, which can improve its consistency and robustness. Other works introduce hierarchical reward modeling~\cite{wang2025towards}, where the reward function might break down the evaluation into multiple criteria or levels (for example, first checking factual accuracy, then style, then overall helpfulness, and combining these). Offline optimization is critical here because it allows integrating such complex criteria in a controlled way. By using gradient-based reward optimization on offline data (where we know the desired outcomes), we can adjust the reward model so that it aligns the agent's behavior with long-term or high-level objectives, not just short-term rewards. For example, if we want an agent that writes accurate and polite answers, we can offline-train a reward model that gives high scores only to outputs that are both factually correct and politely phrased. During deployment, that reward model can then be used (in RL or as a scoring mechanism) to steer the agent. The offline calibration mitigates bias (since we can ensure the training data for the reward model is balanced and carefully curated) and enhances generalization (since the reward model will have seen a wide range of outputs and learned nuanced distinctions). Overall, offline reward model calibration ensures that when the agent is later let loose in the real world or an online learning phase, its notion of what is ``good'' behavior is already well-shaped to match human values or specified goals. This reduces the risk of the agent pursuing the wrong objective or exploiting loopholes in a poorly shaped reward, issues that are common in poorly constrained online learning.

The structured nature of offline optimization produces a robust agent baseline. Because offline training occurs with plentiful data and carefully monitored procedures, it typically yields an agent that performs consistently well on the training distribution and often generalizes better. This phase is especially crucial for mission-critical applications where the agent must meet certain performance guarantees (for example, a diagnostic assistant that must not exceed a given error rate). Before deployment or further online adaptation, offline self-improvement can iron out many problems, optimize efficiency (the agent can be trained to use fewer tokens or less computation per task, for instance), and instill necessary knowledge or skills. The trade-off is that offline methods lack the agility of online learning; once the agent is deployed after offline training, it might not handle an unforeseen scenario optimally until the next retraining. In summary, offline self-improvement emphasizes stability, thoroughness, and pre-planned learning, complementing the reactivity of online methods. It is an essential component for building high-performance agents that have strong foundations and predictable behavior when first deployed.

\section{Comparison of Online and Offline Improvement}
\label{sec:online-vs-offline-optim}



\lettrine[lines=3]{\initfamily\textcolor{darkgreen}{O}}{nline and offline} optimization offer complementary benefits, each excelling in different aspects of an agent's self-improvement cycle. To decide which paradigm to use (or how to mix them), it is important to understand their relative strengths and trade-offs. Table~\ref{tab:online_offline_comparison} summarizes key distinctions between these two paradigms, which we also discuss below.

\begin{table}[h]
\centering
\caption{Comparison of Online vs. Offline Optimization Strategies in Self-Improvement Agents.}
\renewcommand{\arraystretch}{1.3}
\begin{tabular}{p{3cm}|p{5.5cm}|p{5.5cm}}
\hline
\textbf{Feature} & \textbf{Online Optimization} & \textbf{Offline Optimization} \\ \hline
Learning Process & Continuous updates based on real-time feedback & Batch updates during scheduled training phases \\ \hline
Adaptability & High, capable of adjusting dynamically & Lower, adapts only after retraining \\ \hline
Computational Efficiency & More efficient for incremental updates & More resource-intensive due to batch training \\ \hline
Data Dependency & Requires real-time data streams & Relies on curated, high-quality datasets \\ \hline
Risk of Overfitting & Lower due to continuous learning & Higher if training data is not diverse \\ \hline
Stability & Potentially less stable due to frequent updates & More stable with controlled training settings \\ \hline
\end{tabular}

\label{tab:online_offline_comparison}
\end{table}


\paragraph*{Learning Process} Online optimization involves continuous updates based on real-time feedback, whereas offline optimization relies on batch updates during scheduled training phases. In practice, an online-learning agent is constantly in a mode of ``observe -> adjust -> act,'' blending learning with execution. This means the learning process is interwoven with the agent's operation. By contrast, an offline-learning agent has a two-stage life: it gathers experiences or data, then steps away to crunch on this data in a separate learning phase, and finally returns with updated behavior. The continuous nature of online learning means improvements can happen immediately in response to new data, while offline's episodic learning means improvements are delayed until the next training session but can be more thorough.

\paragraph*{Adaptability} As a consequence of their learning processes, online methods offer high adaptability. An online-optimized agent can adjust dynamically to changing conditions or requirements; for example, if a user's preferences shift or a sudden anomaly appears in the environment, an online learner can quickly incorporate that information and modify its behavior. Offline methods have lower adaptability in the short term – once the agent is deployed, it generally sticks to the behavior it was trained for until the next retraining cycle. If confronted with a novel scenario outside its training distribution, an offline-trained agent might struggle because it isn't updating its knowledge on the fly. In essence, online learning is like a sailor who constantly trims the sails with every shift of the wind, while offline learning is like a navigator who charts a course using long-term weather patterns and only revises the route at planned intervals.

\paragraph*{Computational Efficiency} Online optimization is typically more efficient for incremental updates. Since it processes feedback instance-by-instance or in small batches, the agent can often update its parameters or strategy using lightweight computations (sometimes even without retraining any neural weights, as in prompt adjustments or short-term memory updates). This means the computational cost is spread out over time and can be kept relatively low per interaction. Offline optimization, on the other hand, often involves heavy batch training procedures that can be resource-intensive (e.g., fine-tuning a large LLM on a big dataset with multiple epochs can require a lot of GPU hours). However, it's worth noting that while online methods are efficient per update, the cumulative cost of many small updates could add up, and they require infrastructure to continuously run the learning process. Offline training is done less frequently but in a more heavyweight manner.

\paragraph*{Data Dependency} Online methods require a stream of real-time data. They excel when such data is readily available and when the environment provides frequent feedback signals that the agent can learn from. This could be live user interactions, sensor readings, or the agent's own trial-and-error experiments in a simulation. If the stream stops or is not rich enough, online learning might stagnate or overfit to recent data. Offline methods rely on curated, high-quality datasets that are prepared in advance. This gives more control over the data's quality and diversity (you can ensure the training data covers various scenarios, is annotated correctly, etc.), but it also means offline methods are limited by what's in the dataset. If the dataset is missing some scenarios or is biased, the agent's learning will reflect those gaps, and it won't fix them until new data is collected and another offline training happens.

\paragraph*{Risk of Overfitting} Generally, online learning tends to have a lower risk of overfitting to any particular dataset, because the agent is continuously exposed to new situations and doesn't train too long on one fixed batch of data. Instead, it ``forgets'' naturally as new experiences come (which is a double-edged sword – it can forget important things if not managed, as mentioned regarding drift). Offline learning can have a higher risk of overfitting if the training data is not sufficiently diverse or if the model is trained for too many epochs on the same data. Without careful regularization, an offline-trained model might become too specialized to the training set and not perform well on truly novel inputs. That said, with techniques like early stopping or using very large and varied datasets (e.g., pretraining LLMs on massive corpora), offline training can achieve strong generalization. The point is that the nature of exposure to data differs: online gets a trickle of varied data (mitigating overfitting but possibly causing forgetting of older data), offline gets a static bulk of data (risking overfitting to that unless it's broad).

\paragraph*{Stability} Offline optimization is usually more stable because it happens in a controlled environment. Engineers or researchers can carefully monitor an offline training run, use validation sets to pick the best model, and ensure the agent's updated version passes certain safety or performance tests before deploying it. The changes made in offline training are deliberate and evaluated thoroughly post-training. In contrast, online optimization can introduce instability. Frequent updates, if not managed well, can lead the agent's performance to oscillate or degrade unexpectedly (for example, an agent might ``chase noise'' in the feedback and make itself worse on average, a phenomenon known as reward hacking or just instability in online learning). Mechanisms like learning rate decay, confidence thresholds for applying updates, or hybrid approaches (where online updates are sandboxed before full adoption) are often needed to maintain stability in an online context. Essentially, offline is like making changes to a system in a test lab, whereas online is like tweaking a system live in production – the latter has to be done very carefully.

Modern intelligent systems increasingly integrate both online and offline methods to capitalize on the advantages of each. An agent might be initially trained offline to have a solid foundation (addressing the stability and basic capability aspect) and then fine-tuned online to personalize and adapt (addressing the adaptability aspect). While pure online or pure offline setups can work, combining them often yields a more powerful and flexible system. The next section on hybrid approaches discusses how such integration can be achieved in practice.

\section{Hybrid Approaches}
\label{sec:hybrid-self-optim}



\lettrine[lines=3]{\initfamily\textcolor{darkgreen}{R}}{ecognizing} that online and offline methods each have inherent limitations, many contemporary systems adopt hybrid optimization strategies that blend the two paradigms. The goal of a hybrid approach is to get the best of both worlds: the robust, high-performance foundation from offline training, and the adaptable, real-time refinement from online learning. In a hybrid self-improvement loop, an agent alternates between (or sometimes concurrently uses) offline and online phases of learning. This explicit integration allows the agent to autonomously evaluate, adapt, and enhance its behaviors through distinct yet interconnected stages. We can break down a typical hybrid strategy into three stages: \emph{Offline Pre-Training}, \emph{Online Fine-Tuning for Dynamic Adaptation}, and \emph{Periodic Offline Consolidation for Long-Term Improvement}.

\paragraph*{Offline Pre-Training}
In the first stage, the agent acquires a strong baseline of knowledge and skills through extensive offline training on curated datasets or in simulated environments. This can be thought of as the agent's ``education'' before it is deployed in the real world. By learning from a wide range of examples and scenarios offline, the agent develops essential capabilities such as reasoning, language understanding, and decision-making pertinent to its intended tasks. For instance, an LLM-based assistant might be pretrained on diverse text corpora and dialogues so it has a broad knowledge and good conversational ability from the start. This stage establishes a stable foundation such that the agent doesn't enter the online phase blank-slate or brittle. A concrete example from the reinforcement learning domain is provided by \citet{schrittwieser2021online} – in their work (known for the MuZero Unplugged algorithm), they showed that an agent can be first trained on a batch of past game experiences (offline) to build up competence, and then successfully fine-tuned with further online interactions. That offline pre-training systematically enhances the agent's initial capabilities, ensuring that subsequent online improvements are built upon a reliable and effective base. In general, offline pre-training reduces the burden on online learning: instead of having to learn everything from scratch in real time (which could be slow or risky), the agent only has to learn the differences or new patterns that were not covered in the pre-training.

\paragraph*{Online Fine-Tuning for Dynamic Adaptation}
After or alongside pre-training, the agent enters an online learning phase where it fine-tunes its behavior by interacting with the environment or users and receiving feedback in real time. In this stage, the agent actively monitors its performance, identifies shortcomings or new requirements, and adjusts its strategies on the fly. This might involve tuning its prompts, learning from user corrections, trying new actions, or updating its internal models slightly with each interaction. The changes here are typically smaller-scale and incremental (since we want to avoid catastrophic changes while live) but happen continuously. This stage directly aligns with the self-improvement paradigm – the agent is essentially self-evaluating and self-adjusting in response to the world. An example of a framework that embodies this is Decision Mamba-Hybrid (DM-H) introduced by \citet{huang2024decision}. DM-H is a reinforcement learning approach that combines two techniques: a transformer-based model for high-quality decision predictions and a memory-efficient model (Mamba) for handling long-term dependencies. In an online fine-tuning context, DM-H allows an agent to efficiently adapt to complex, evolving scenarios by generating sub-goals from its long-term memory and then prompting the transformer model with those sub-goals to decide on immediate actions. The result is an agent that can adjust its plan as conditions change, leveraging its offline-trained knowledge (in the transformer) together with real-time cues (through the Mamba memory and feedback loop). More generally, during online fine-tuning, an agent might do things like dynamically reweighting different objectives (if it notices it's going too fast at the cost of accuracy, it can slow down), or personalizing responses (if a user keeps correcting a certain style of answer, the agent learns that user's preference). The essence of this stage is fast adaptation: the agent is tailoring its behavior to the here-and-now, which is crucial for maintaining performance in the face of novelty or drift in the environment.

\paragraph*{Periodic Offline Consolidation for Long-Term Improvement} The third stage involves the agent periodically pausing its online updates and entering a new offline training phase to consolidate and integrate the knowledge it has acquired through online interactions. Think of this as the agent taking a step back to reflect on everything it learned recently, and solidifying those lessons into its core parameters or knowledge base. During online learning, the agent might accumulate a lot of small tweaks, new data (e.g., logs of interactions, new problem cases it encountered), or temporary adaptations. The offline consolidation phase uses this accumulated experience as a new training dataset. The agent is retrained or fine-tuned in a more rigorous way on that data, possibly also reinitializing some of its online-specific adaptations to prevent drift. The advantage of doing this offline is that the integration of new knowledge can be done more systematically and safely – one can use the full suite of training techniques (like shuffling data, doing multiple passes, using regularization to not forget older knowledge, etc.). This ensures that improvements gleaned from the online phase are retained long-term and that any noise or unstable changes are smoothed out. The Uni-O4 framework by \citet{lei2023uni} exemplifies this process. Uni-O4 is designed to unify offline and online deep reinforcement learning; it aligns the objectives used in offline pre-training with those used in online fine-tuning so that an agent can move between the two seamlessly. In their approach, an agent might be first trained offline on a simulator, then deployed to learn online in the real world, and periodically its real-world experiences are folded back into an offline update. The result is a cycle where the agent keeps improving without forgetting why it was successful before. In practice, such periodic consolidation might occur on a schedule (say, an autonomous vehicle retrains overnight on the day's collected data) or based on a threshold (if the agent's online performance starts degrading or plateauing, trigger an offline retraining with the new data). This hybrid loop allows long-term stability and continual adaptation.

By explicitly alternating between these modes, hybrid optimization supports autonomous, continuous evolution of the agent. The offline phases provide stability, comprehensive learning, and a chance to ingest large amounts of experience, while the online phases provide immediacy, specificity, and the ability to keep up with changes. The two feed into each other: offline training makes the agent better at online adaptation (since it starts from a higher base), and online experience makes the next offline training richer (since there's new real-world data to learn from). This cyclical training approach is well-suited for complex real-world scenarios. Consider an autonomous robot: it can be pre-trained in simulation (offline) to get basic navigation and manipulation skills, then deployed in a new home where it fine-tunes its policy as it interacts with the actual environment (online), and after a while, all the data from the home is used to do a thorough update to its policy (offline again) resulting in improved performance that incorporates the quirks of that home's layout. Or think of a personalized intelligent assistant: it's initially trained on general language data (offline), then it learns a user's particular needs and style from ongoing conversations (online), and periodically it can retrain on the compiled chat history to improve its core dialogue model (offline consolidation). Through such hybridization, agents maintain the long-term robustness and knowledge depth that offline training provides, while also achieving continuous refinement and personalization through online learning.

In summary, hybrid approaches interweave structured offline learning with proactive online adaptation. This allows agents to avoid the pitfalls of relying solely on one method: the agent is neither stuck with only what it learned yesterday (thanks to online updates) nor caught in a loop of potentially unstable live changes (thanks to grounding, reset, and validation via offline phases). As a result, hybrid agents tend to exhibit both immediate responsiveness and stable long-term improvement. They are particularly powerful in domains like autonomous driving, robotics, finance, or large-scale interactive systems, where conditions evolve over time and safety or reliability is critical. By cycling between learning modes, the agent keeps getting better in a controlled yet flexible manner, which is the essence of continual self-improvement.

\section{Summary and Discussion}
\label{sec:summary-online-offline}

\lettrine[lines=3]{\initfamily\textcolor{darkgreen}{I}}{n} this chapter, we have explored the paradigms of online, offline, and hybrid self-improvement for LLM-based agents. Online self-improvement enables agents to adapt dynamically through iterative feedback loops, real-time parameter tuning, and active exploration, providing immediate responsiveness and context-sensitive learning. Conversely, offline self-improvement employs structured, batch-based training to cultivate robust generalization, stability, and comprehensive knowledge integration. Recognizing the complementary strengths of these two paradigms, hybrid strategies integrate the real-time adaptability of online learning with the foundational stability provided by offline optimization, offering a balanced framework that supports continuous, controlled self-evolution.

Despite notable progress, several challenges remain. Online approaches, while agile, must grapple with stability concerns, potential drift, and computational overhead from continuous updates. Offline methods, though stable and robust, face limitations in adaptability, data coverage, and the risk of overfitting to curated datasets. Hybrid methods seek to mitigate these limitations but introduce additional complexity in managing transitions and alignment between offline and online phases. Future research opportunities include the design of principled hybrid algorithms, adaptive meta-learning techniques for seamless online-offline transitions, and efficient strategies for managing computational resources and memory during continual self-improvement. Ultimately, effective self-improvement for LLM-based agents hinges on developing frameworks that judiciously balance agility and stability, adaptability and robustness, enabling sustained and reliable performance across diverse real-world scenarios.
\chapter{Intelligent Evolution through Scientific Discovery}


\lettrine[lines=3]{\initfamily\textcolor{darkgreen}{I}}{n} previous chapters, we primarily discussed the evolution of agentic systems from a technical perspective, focusing on how to develop systems that can effectively perform well-defined tasks traditionally executed by humans.
However, a fundamental and important question remains: can these agents drive a self-sustaining innovation cycle that propels both agent evolution and human progress?

Scientific knowledge discovery is a compelling example of self-evolution in intelligent beings, as it helps them adapt to the world in a sustainable way.
Agents capable of discovering scientific knowledge at different levels of autonomy and in a safe manner will also play important roles in technological innovation for humanity.
In this section, we survey progress in autonomous discovery using agentic workflows and discuss the technological readiness toward fully autonomous, self-evolving agents. 
Within this scope, the goal of the agent is to uncover, validate, and integrate data, insights, and principles to advance an objective scientific understanding of natural phenomena.
Instead of altering the world, the agent seeks to better understand nature as a Scientist AI \cite{bengio2025SuperiAgents} and assist humans in extending the boundaries of knowledge.

Our proposed agent framework, as illustrated in Figure \ref{fig:agent-framework}, systematically acquires scientific data, insights, and principles through active interaction with its environment. This information is subsequently stored within the agent's memory (detailed in Chapter~\ref{chap:memory}) and utilized to enhance both its internal world model (elaborated in Chapter~\ref{chapter:world_model}) and reward function (discussed in Chapter~\ref{Chapter:reward}), collectively leading to improved scientific cognition.
Within optimization spaces, the agent continuously evolves by refining workflows and enhancing tool usage, thereby promoting increasingly efficient scientific discoveries. This evolution is realized through strategies such as prompt optimization, the creation of new tools, automated workflow generation, and other advanced methods. Enhanced efficiency in scientific discovery subsequently facilitates further scientific information acquisition, establishing a reinforcing cycle.
Consequently, the framework fosters a self-sustaining loop of ``knowledge discovery → enhanced capability for knowledge discovery → increased knowledge discovery'', driving the continuous evolution of agent intelligence, ultimately benefiting humanity.

In this chapter, we first define the concepts of knowledge and intelligence to clarify our discussion, followed by an introduction of three typical scenarios where agents interact with scientific knowledge. 
We highlight existing successes and examples of self-enhancing agents applied to theoretical, computational, and experimental scientific research. Finally, we summarize current challenges and offer insights for future directions.

\section{Agent's Intelligence for Scientific Knowledge Discovery} 
\label{sec:IQ_definition}

\lettrine[lines=3]{\initfamily\textcolor{darkgreen}{K}}{nowledge}, traditionally defined as \textit{justified true belief}, traces back to Plato \cite{plato1992Theaet} and has been further refined by Edmund Gettier \cite{gettier1963JustifTrue}, who argued that knowledge must be produced by a reliable cognitive process---though its precise definition remains debated \cite{steup2024Episte}. 
In our discussion, we describe scientific knowledge discovery as the process of collecting data and information to either justify or falsify rational hypotheses about target scientific problems. 
To discuss the capability of agents in scientific knowledge discovery, we first explore a general framework for measuring an agent's intelligence through the lens of information theory. 

\subsection{KL Divergence-based Intelligence Measure}
\label{subsec:KL_measure_intelligence}

\textbf{The agent's intelligence can be measured by the Kullback–Leibler (KL) divergence between its predicted and real-world probability distributions of unknown information.}
A long-standing goal in both artificial intelligence and the philosophy of science is to formalize what it means for an agent to ``understand'' the world. 
From Jaynes' view of probability theory as extended logic for reasoning under uncertainty \cite{jaynes2003ProbabTheory}, to Parr et al.'s framing of intelligence as minimizing model-world divergence under the free energy principle \cite{parr2022ActiveInfere}, many frameworks converge on a common theme: intelligent behavior arises from making accurate predictions about an uncertain world. 
Clark \cite{clark2015surfing}, for instance, argues that intelligent agents constantly engage with the world through prediction and error correction to reduce surprise. 
Chollet \cite{chollet2019MeasurIntell} emphasizes that intelligence should reflect skill-acquisition efficiency, because of the dynamic nature of task adaptation.
Together, these views suggest that intelligence involves building predictive and adaptable models---an idea formalized here through a probabilistic framework that links reasoning to knowledge acquisition and enables comparison across agents in scientific discovery.

Building on this foundation, we consider intelligence in the specific context of scientific knowledge discovery, where the agent's primary objective is to infer unknown aspects of the physical world from limited data.
From the agent's perspective in knowledge discovery, the world $\mathcal{W}$ is characterized by an ensemble of datasets ${\bf x}=\{x_1, x_2, ..., x_n\}$ related to the scientific problem the agent aims to understand. 
During the agent's interaction with $\mathcal{W}$, each dataset appears in the experimental measurements or observations with a probability $P_\mathcal{W}({\bf x})$. 
Here we assume that individual data points $x_i$ may or may not be correlated. 
For example, in a task of text generation using a language model, $x_i$ represents a chunk of tokens forming a meaningful proposition, and ${\bf x}$ is a coherent text constructed from known and inferred propositions. 
In this context, the ``world'' is the ensemble of all propositions.  

Let $\theta$ denote the parameter that parameterizes the agent's world model, $M_t^{\mathrm{wm}}$, as defined in Table \ref{tab:notation_summary}. 
For instance, in a transformer model with a fixed architecture, $\theta$ represents its weights. 
Given $\theta$ and a dataset ${\bf x}$, the agent predicts a probability distribution $P_\theta({\bf x})$. 
In general, different AI agents could be optimized for different goals. 
For scientific knowledge discovery, we assume that the agent's goal is to produce a good description of the real world, i.e., a world model that predicts yet-to-be-explored natural phenomena as accurately as possible. 
A more intelligent agent produces a better approximation of the real-world distribution $P_\mathcal{W}({\bf x})$. 
The agent's intelligence can thus be measured by the KL divergence, or relative entropy, between these two probability distributions:
\begin{align}
    D_0(\theta)=\sum_{\bf x \subseteq \mathcal{W}}P_\mathcal{W}({\bf x})\log\frac{P_\mathcal{W}({\bf x})}{P_\theta({\bf x})}
\end{align}
$D_0(\theta)$ describes the difference between $P_{\mathcal{W}}({\bf x})$ and $P_\theta({\bf x})$. 
More precisely, in the context of hypothesis testing, if we sample $P_{\mathcal{W}}({\bf x})$ $N$ times and compare the results with the predictions from $P_\theta({\bf x})$, the probability of mistaking $P_{\mathcal{W}}({\bf x})$ for $P_{\theta}({\bf x})$ scales as $e^{-ND_0(\theta)}$ \cite{cover2005ELEMENINFORM}. 
In other words, an agent with a lower $D_0(\theta)$ produces predictions that align more closely with reality.

For example, consider two materials synthesis agents whose goal, $M_t^{goal}$, is to understand whether or not an inorganic compound of interest, CaFe$_2$(PO$_4$)$_2$O, is synthesizable. 
The agents can predict either (1)  ${\bf x}_{1}$=\{CaFe$_2$(PO$_4$)$_2$O is synthesizable\}, and (2) ${\bf x}_{2}$=\{CaFe$_2$(PO$_4$)$_2$O is not synthesizable\}. 
In reality, since CaFe$_2$(PO$_4$)$_2$O is a natural mineral called \textit{crocobelonite}, $P_{\mathcal{W}}({\bf x}_1)$ = 1 and $P_{\mathcal{W}}({\bf x}_2)$ = 0.
However, this mineral was only recently reported on October 4, 2023 \cite{britvin2023CrocobCaFe23}, after the knowledge cutoff of many LLMs; thus, the agents lacks that knowledge.
Compare Agent 1, which guesses randomly $P_{\theta_1}({\bf x}_1)=P_{\theta_1}({\bf x}_1)$ = 0.5,
yielding $D_0(\theta_1)=\log 2$.
In contrast, Agent 2 employs first-principles calculations and finds that CaFe$_2$(PO$_4$)$_2$O is the lowest-energy phase among competing compounds of the same composition in the Ca-Fe-P-O phase diagram \cite{mp-1040941}, indicating its thermodynamic stability. 
Thereby, Agent 2 predicts that CaFe$_2$(PO$_4$)$_2$O is likely synthesizable, suggesting $P_{\theta_2}({\bf x}_1)>0.5>P_{\theta_2}({\bf x}_2)$.
Consequently, $D_0(\theta_2)=-\log P_{\theta_2}({\bf x}_1)<D_0(\theta_1)$, meaning that Agent 2 has a more accurate understanding of the real world.

Now, let us assume the agent has conducted some measurements and determined specific values for a subset of data points $x_i$. 
Let ${\bf x}_{\mathrm{K}}$ denote this known subset and ${\bf x}_{\mathrm{U}}$ the remaining unknown part.
Correspondingly, we define the space of all existing knowledge as $\mathcal{K}$ and the space of all unknown information as $\mathcal{U}$, satisfying ${\bf x}_{\mathrm{K}} \subseteq \mathcal{K}$, ${\bf x}_{\mathrm{U}} \subseteq \mathcal{U}$, and $\mathcal{K} \cup \mathcal{U} = \mathcal{W}$.
For example, in text generation, the the prompt text ${\bf x}_{\mathrm{K}}$ represents already known information. 
The efficiency of the language model is then measured by its predictive accuracy for the generated text ${\bf x}_{\mathrm{U}}$ based on ${\bf x}_{\mathrm{K}}$. 
More generally, the agent's intelligence is measured by the relative entropy of the conditional probability distribution:
\begin{align}
    D_{\mathrm{K}}(\theta,{\bf x}_{\mathrm{K}})=\sum_{{\bf x} \subseteq \mathcal{U} }P_\mathcal{W}
    \left({\bf x}|{\bf x}_{\mathrm{K}}\right)\log\frac{P_\mathcal{W}
    \left({\bf x}|{\bf x}_{\mathrm{K}}\right)}{P_\theta
    \left({\bf x}|{\bf x}_{\mathrm{K}}\right)}
\end{align}

\begin{figure}[!htb]
\centering
    \includegraphics[width=0.65\columnwidth]{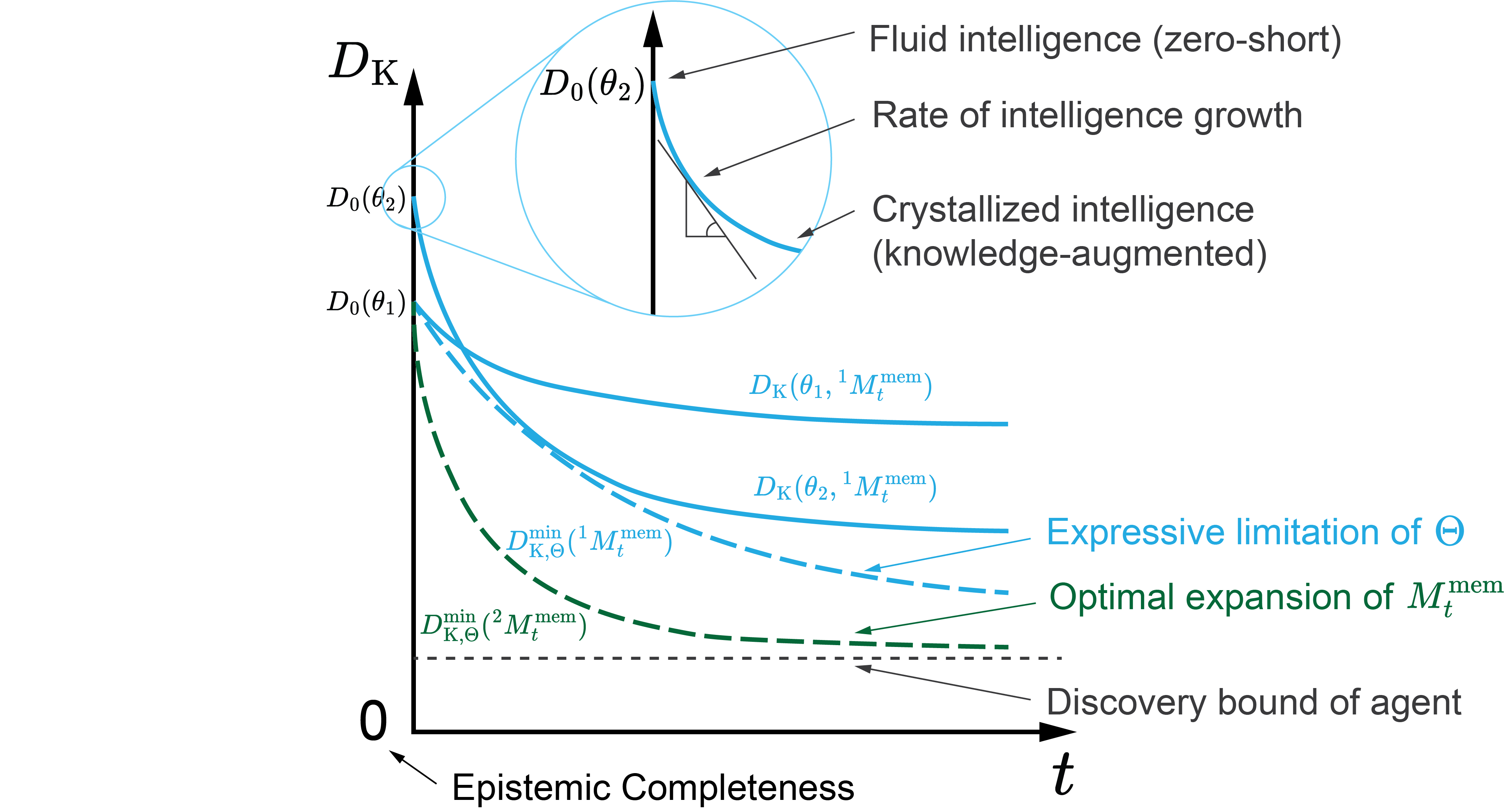}
    \caption{\textbf{Schematic representation of agent intelligence and knowledge discovery.} 
    The agent's intelligence, measured by the KL divergence $D_{\mathrm{K}}$ between predictions and real-world probability distributions, evolves from fluid intelligence (zero-shot predictions for new problems) to crystallized intelligence (knowledge-augmented predictions after learning) as it accumulates data in its memory $M_t^{\mathrm{mem}}$ over time $t$.
    Given $M_t^{\mathrm{mem}}$, the evolution of $D_{\mathrm{K}}$ varies within the world model's parameter space $\Theta$, as illustrated by $\theta_1$ and $\theta_2$ in the solid lines.
    The expressive limitation of $\Theta$ is characterized by the envelope $D_{\mathrm{K},\Theta}^{\rm min}$.
    Given $\Theta$, $D_{\mathrm{K},\Theta}^{\rm min}$ is influenced by different knowledge expansion strategies, such as ${}^{1}M_t^{\mathrm{mem}}$ and ${}^{2}M_t^{\mathrm{mem}}$, shown as dash lines.    
    }
\label{fig:knowledge_schema}
\end{figure}

In practice, all of the agent's knowledge is stored in its memory $M_t^{\mathrm{mem}}$,
i.e., ${\bf x}_{\mathrm{K}} = \mathcal{K} =  M_t^{\mathrm{mem}}$ and $\mathcal{U}=\mathcal{W} \setminus M_t^{\mathrm{mem}}$, we define the agent's intelligence as: 
\begin{definition}
[\textbf{Agent's Intelligence for Knowledge Discovery}]
\label{def:agent-intelligence}
The agent's intelligence for knowledge discovery is quantified as the negative KL divergence between the its predicted and the true probability distributions ($P_\theta$ and $P_\mathcal{W}$) of the unknown information (${\bf x} \subseteq \mathcal{U}$), both conditioned on its memory $M_t^{\mathrm{mem}}$.
\begin{align}
    IQ_t^{\mathrm{agent}} 
    \equiv -D_{\mathrm{K}}(\theta,M_t^{\mathrm{mem}})
    =-\sum_{{\bf x} \subseteq \mathcal{U} }P_\mathcal{W}
    \left({\bf x}|M_t^{\mathrm{mem}}\right)\log\frac{P_\mathcal{W}
    \left({\bf x}|M_t^{\mathrm{mem}}\right)}{P_\theta
    \left({\bf x}|M_t^{\mathrm{mem}}\right)}
\end{align}
\end{definition}
In other words, the the agent's intelligence $IQ_t^{\mathrm{agent}}$ is determined by its memory $M_t^{\mathrm{mem}}$ and the parameter $\theta$ of its world model $M_t^{\mathrm{wm}}$. 
A schematic plot is shown in Figure \ref{fig:knowledge_schema}.
At time $t=0$, when the $M_t^{\mathrm{mem}}$ is very limited or lack relevant information to a new target scientific problem, $IQ_t^{\mathrm{agent}}$ is primarily determined by the zero-shot predictive ability of $M_t^{\mathrm{wm}}$, corresponding to fluid intelligence \cite{cattell1963TheoryFluid}.
Over time, as more relevant knowledge is incorporated into $M_t^{\mathrm{mem}}$, $IQ_t^{\mathrm{agent}}$ becomes increasingly dependent on the knowledge-augmented predictive capability of $M_t^{\mathrm{wm}}$, reflecting crystallized intelligence \cite{cattell1971AbilitTheir}.

\subsection{Statistical Nature of Intelligence Growth}
\label{subsec:IQ_increase_w_knowledge}
\textbf{The agent's intelligence, in a statistical sense, is a non-decreasing function of acquired knowledge.}
Roughly speaking, $IQ_t^{\mathrm{agent}}$ quantifies both the amount of knowledge an agent has acquired and how effectively the agent can apply that knowledge after learning from $M_t^{\mathrm{mem}}$. Intuitively, if the agent gains additional information at time $t$---which corresponds to enlarging $M_t^{\mathrm{mem}}$ and shrinking $\mathcal{U}$---its intelligence should increase.

To understand this process, consider a small region $\Delta \subseteq \mathcal{U}$ and examine the effect of adding a dataset ${\bf x}_\Delta$ from $\Delta$ to $M_t^{\mathrm{mem}}$.
Denote $\mathcal{U}=\mathcal{U}'\cup \Delta$, where $\mathcal{U}'$ represents the remaining unknown part of the world.
The agent's intelligence at time $t+1$ is given by:
\begin{align}
    IQ_{t+1}^{\mathrm{agent}} 
    \equiv 
    -D_{\mathrm{K}}(\theta, M_t^{\mathrm{mem}} {\bf x}_\Delta)
    =- \sum_{{\bf x}' \subseteq \mathcal{U}'}P_\mathcal{W}
    \left({\bf x}'|M_t^{\mathrm{mem}} {\bf x}_\Delta\right)\log\frac{P_\mathcal{W}
    \left({\bf x}'|M_t^{\mathrm{mem}} {\bf x}_\Delta\right)}{P_\theta
    \left({\bf x}'|M_t^{\mathrm{mem}} {\bf x}_\Delta\right)}
\end{align}
Directly comparing $IQ_t^{\mathrm{agent}} $ and $IQ_{t+1}^{\mathrm{agent}} $ is challenging. 
Instead, we can compare the expected value of $IQ_{t+1}^{\mathrm{agent}}$, averaging over ${\bf x}_\Delta$ with probability $P_\mathcal{W}({\bf x}_\Delta| M_t^{\mathrm{mem}} )$. 
This expectation represents the average amount of knowledge gained by measuring $\Delta$, given prior knowledge in $M_t^{\mathrm{mem}}$. 
We obtain:
\begin{align}
    \sum_{{\bf x} \subseteq \Delta}
    P_\mathcal{W}({\bf x}|M_t^{\mathrm{mem}})
    IQ_{t+1}^{\mathrm{agent}} 
    &= - \sum_{{\bf x}' \subseteq \mathcal{U}',~ {\bf x} \subseteq \Delta }P_\mathcal{W} 
    \left({\bf x}'{\bf x}|M_t^{\mathrm{mem}}\right)\log\frac{P_\mathcal{W}
    \left({\bf x}'|M_t^{\mathrm{mem}} {\bf x}\right)}{P_\theta
    \left({\bf x}'|M_t^{\mathrm{mem}} {\bf x}\right)}\nonumber\\
    &=IQ_t^{\mathrm{agent}} + \sum_{{\bf x} \subseteq \Delta}P_\mathcal{W}({\bf x} |M_t^{\mathrm{mem}})\log\frac{P_\mathcal{W}({\bf x} |M_t^{\mathrm{mem}})}{P_\theta({\bf x} |M_t^{\mathrm{mem}})}
\end{align}

The second term is the relative entropy of the conditional probability distribution of ${\bf x}_\Delta$ conditioned on $M_t^{\mathrm{mem}}$, which is always non-negative. 
Therefore, on average, $IQ_t^{\mathrm{agent}}$ is non-decreasing as $M_t^{\mathrm{mem}}$ acquires new knowledge over time. 
Note that $IQ_{t+1}^{\mathrm{agent}}$ can be further increased by leveraging the newly acquired knowledge to optimize $\theta$ within $M_t^{\mathrm{wm}}$. 

Interestingly, the expected gain in intelligence at time $t$ is determined by the discrepancy between the actual distribution $P_{\mathcal{W}}({\bf x} |M_t^{\mathrm{mem}})$ and the model-predicted distribution $P_\theta({\bf x} |M_t^{\mathrm{mem}})$. 
In other words, the rate of intelligence growth in Figure \ref{fig:knowledge_schema} is higher when the new measurement result is more unexpected. 
This observation identifies scientist agents \cite{bengio2025SuperiAgents} as a special type of curiosity-driven agent \cite{ten2021HumansMonito}, prioritizing exploration over exploitation to expand the frontiers of knowledge for deeper understanding of nature. 
Unlike agents that leverage existing knowledge to achieve predefined objectives, curiosity-driven agents can learn without extrinsic rewards \cite{pathak2017curiosity,burda2018LargeSStudy} (see Section \ref{sec:AI_reward_paradigms} for details), enabling discoveries beyond human-planned search spaces and revealing knowledge in unexplored domains.
This potential also underscores the importance of equipping curiosity-driven agents with fundamental perception and action tools that can be transferred to explore new knowledge domains.

\subsection{Intelligence Evolution Strategies}
\label{subsec:knowledge_expansion_strategy}

\textbf{The strategy for expanding known information 
determines how quickly an agent's intelligence evolves.}
For a given knowledge base $M_t^{\mathrm{mem}}$, the parameter $\theta$ can be optimized over a space of world models $\Theta$ characterized by the architecture of $M_t^{\mathrm{wm}}$.
The optimal agent is the one that minimizes $D_{\mathrm{K}}(\theta,M_t^{\mathrm{mem}})$, thereby maximizing $IQ_t^{\mathrm{agent}}$:
\begin{align}
    \theta^*_{\mathrm{K},t} \equiv
    {\rm arg~sup}_\theta IQ_t^{\mathrm{agent}}=
    {\rm arg~inf}_\theta D_{\mathrm{K}}(\theta,M_t^{\mathrm{mem}})
\end{align}
and
\begin{align}
    D_{\mathrm{K},\Theta}^{\rm min}(M_t^{\mathrm{mem}}) \equiv
    D_{\mathrm{K}}(\theta^*_{\mathrm{K},t},M_t^{\mathrm{mem}})
\end{align}
Here, $D_{\mathrm{K},\Theta}^{\rm min}(M_t^{\mathrm{mem}})$ represents the minimum unknown after learning from $M_t^{\mathrm{mem}}$ for this family of models, quantifying the expressive limitations of $\Theta$. 
As shown in Figure \ref{fig:knowledge_schema}, $D_{\mathrm{K},\Theta}^{\rm min}(M_t^{\mathrm{mem}})$ forms the envelope of the family of functions $D_{\mathrm{K}}(\theta, M_t^{\mathrm{mem}})$, where $\theta$ ranges over $\Theta$.  

For a given model family $\Theta$, $D_{\mathrm{K},\Theta}^{\rm min}(M_t^{\mathrm{mem}})$ measures the best possible prediction of residual unknowns in addressing the target scientific problem based on $M_t^{\mathrm{mem}}$.
In other words, the knowledge content in $M_t^{\mathrm{mem}}$ is captured by $D_{\mathrm{K},\Theta}^{\rm min}(M_t^{\mathrm{mem}})$.
One can prove that $D_{\mathrm{K},\Theta}^{\rm min}(M_t^{\mathrm{mem}})$ is monotonically non-increasing as $M_t^{\mathrm{mem}}$ expands, since it forms the envelope of a family of non-increasing functions $D_{\mathrm{K}}(\theta, M_t^{\mathrm{mem}})$. 
This expansion process is tied to how the agent acts and gains information, driven by $M_t^{\mathrm{wm}}$, which determines the optimal expansion and executes it through the action $a_t\in\mathcal{A}$ at time $t$ (see Table \ref{tab:notation_summary}).

During knowledge discovery, different strategies can be employed to expand $M_t^{\mathrm{mem}}$.
Agents with varying knowledge expansion strategies can derive diverse explanations from the same physical system, leading to different evolutionary trajectories \cite{fu2025twoai}.
The optimal expansion strategy is the one that results in the steepest decrease of $D_{\mathrm{K},\Theta}^{\rm min}(M_t^{\mathrm{mem}})$.
For instance, in Figure \ref{fig:knowledge_schema}, we illustrate two strategies for expanding $M_t^{\mathrm{mem}}$, denoted as ${}^{1}M_t^{\mathrm{mem}}$ and ${}^{2}M_t^{\mathrm{mem}}$.
The first strategy, ${}^{1}M_t^{\mathrm{mem}}$, represents random exploration, while the second, ${}^{2}M_t^{\mathrm{mem}}$, follows a hypothesis-driven approach \cite{voit2019PerspeDimens} in which the agent first formulates a hypothesis about the underlying mechanism of the target problem and then designs an experiment to justify or falsify this hypothesis \cite{gottweis2025AICoscie}.
In practice, experimentalists typically adopt the hypothesis-driven strategy because it enables them to guide the expansion of $M_t^{\mathrm{mem}}$ in a way that maximizes the reduction of $D_{\mathrm{K},\Theta}^{\rm min}(M_t^{\mathrm{mem}})$, subject to resource constraints.
This approach is generally more efficient than random exploration for expanding  $M_t^{\mathrm{mem}}$, leading to $D_{\mathrm{K},\Theta}^{\rm min}({}^{2}M_t^{\mathrm{mem}})$ descending faster than $D_{\mathrm{K},\Theta}^{\rm min}({}^{1}M_t^{\mathrm{mem}})$.

In general, the knowledge discovery process proceeds iteratively, repeatedly optimizing the world model parameter $\theta$ to approach $\theta^*_{\mathrm{K},t}$ and expanding $M_t^{\mathrm{mem}}$ in a rational manner to accelerate the decrease of $D_{\mathrm{K},\Theta}^{\rm min}(M_t^{\mathrm{mem}})$. 
The ideal state is achieving epistemic completeness, i.e., $D_{\mathrm{K},\Theta}^{\rm min}(M_t^{\mathrm{mem}})=0$, meaning zero discrepancy between the agent's prediction and the real-world  phenomena.
However, for a specific agent, a discovery bound may exist, where $D_{\mathrm{K},\Theta}^{\rm min}(M_t^{\mathrm{mem}})$ approaches zero but remains positive. These discrepancies arise from practical constraints and the limitations of $\Theta$, $\mathcal{A}$, and other design spaces of the agent \cite{hole2021ThousaBrains}.
Achieving a low discovery bound requires designing an adaptive world model architecture, an efficient knowledge expansion strategy, and a sufficient action space.

\section{Agent-Knowledge Interactions}

\lettrine[lines=3]{\initfamily\textcolor{darkgreen}{T}}{ypical} forms of scientific knowledge include observational knowledge (e.g., experimental measurements, computational results), methodological knowledge (e.g., experimental methods, computational techniques, protocols), and theoretical knowledge (e.g., theories, laws, predictive models).
These forms of knowledge can contribute to scientific understanding as long as they consist of data and information processed in a way that affects the probability distribution of unknown information ${P_\theta \left({\bf x}_{\mathrm{U}}|M_t^{\mathrm{mem}}\right)}$, reduces $D_{\mathrm{K}}(\theta, M_t^{\mathrm{mem}})$, and facilitates decision-making. 

\begin{figure}[!htb]
\centering
    \includegraphics[width=0.85\columnwidth]{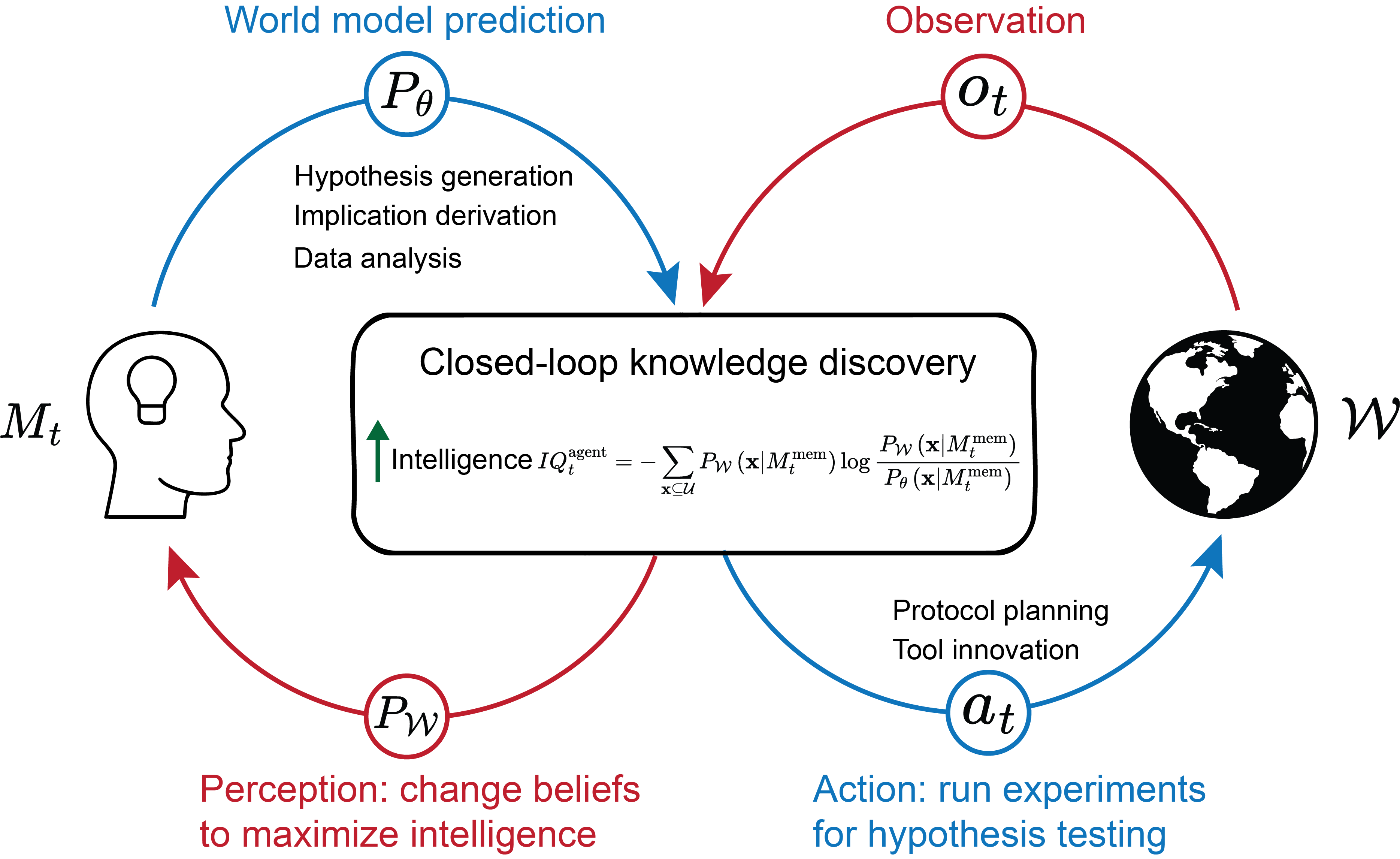}
    \caption{
    \textbf{Closed-loop knowledge discovery for sustainable self-evolution of an AI scientist.}
    The agent aims to iteratively enhance its intelligence $IQ_t^{\mathrm{agent}}$ through hypothesis generation and testing, as well as through data analysis and implication derivation.
    When interacting with the physical world $W$, the agent generates hypotheses as an explicitly or implicitly predicted distribution ($P_\theta$) of unknown information, takes actions ($a_t$) for hypothesis testing, observes experimental results ($o_t$), and updates beliefs based on perception of the real-world distribution ($P_W$). 
    When not interacting with $W$, the agent distills knowledge from existing data and premises, updating mental states $M_t$ directly. 
    Inspired by Figures 2.3 and 2.5 in \cite{parr2022ActiveInfere}.
    }
\label{fig:innovation_loop}
\end{figure}

In principle, external scientific knowledge has been shown to be useful in improving agent performance in reasoning and decision-making \cite{chen2024SciencRigoro,prince2024OpportRetrie}.
However, the scope of this survey lies in how agents can autonomously discover and utilize knowledge to enhance themselves.
Scientific knowledge discovery workflows typically involve hypothesis generation, protocol planning, conducting experiments and computations, analyzing data, deriving implications, and revising hypotheses---often as part of an iterative cycle.
An agentic AI scientist that can perceive, learn, reason, and act has the potential to drive such workflows in an autonomous manner, for example by using application programming interfaces (APIs) to interact with physical instruments to acquire scientific knowledge and iteratively enhance its knowledge base \cite{szymanski2023AutonoLabora,vriza2023SelfDrLabora, ghareeb2025robinmultia,ding2024matexpertdecomposingmaterialsdiscovery,takahara2025acceleratedinorganicmaterialsdesign} (Figure \ref{fig:innovation_loop}). 
The agent will use the acquired knowledge to update its mental states $M_t$ to make better decisions when interacting with the world $\mathcal{W}$.
We will now highlight three scenarios where agents discover scientific knowledge and enhance themselves.

\subsection{Hypothesis Generation and Testing} 
Hypothesis generation and testing (Figure \ref{fig:innovation_loop}) is a critical application of agents in autonomous scientific discovery, as it has the potential to enable outside-the-box innovations \cite{gottweis2025AICoscie}. 
In essence, hypothesis generation is the formation of potential rules that govern data distribution---ranging from single observations to large datasets---pertaining to unobserved scientific phenomena.
According to Sir Karl Popper, a scientific hypothesis must be falsifiable \cite{popper1962ConjecRefuta, popper2008LogicScient}; in this discussion, we define a hypothesis that survives falsification as a \textit{justified true hypothesis} \cite{gillies2001PopperComput, plato1992Theaet}. 
Typically, scientists test hypotheses by conducting experiments to either justify or falsify them.
A hypothesis is considered more valuable if it is broad enough to explain a wide range of data and is highly likely to be true.

To tackle a scientific problem, the agent formulates one or a small number of high-value hypotheses based on its mental state $M_t$, which contains only incomplete information about the partially observable world $\mathcal{W}$.
After testing through experiments or computations, a \textit{justified true hypothesis} becomes instructive knowledge, expanding $M_t^{\mathrm{mem}}$ in a way that rapidly minimizes $D_{\mathrm{K},\Theta}^{\rm min}(M_t^{\mathrm{mem}})$.
Hence, generating and testing high-value hypotheses can quickly promote knowledge discovery and increase $IQ_t^{\mathrm{agent}}$.
In this scenario, the agent employs the learning function, $\mathrm{L}$, to process observations from hypothesis testing, $o_t$, into knowledge and update its mental states $M_t$. 

\textbf{Generating physically meaningful hypotheses is a key step.} 
The agent typically uses LLMs along with collaborative architectures and domain knowledge for hypothesis generation \cite{jin2024agentreview}. 
Si et al. \cite{si2024CanLLMs} conducted a large-scale human study involving over 100 NLP researchers, and found that LLM-generated ideas were rated as more novel ($p<0.05$) than human expert ideas, albeit slightly weaker in feasibility.
Ghafarollahi et al. \cite{ghafarollahi2024SciAgeAutoma} developed SciAgents, which generates and refines materials science hypotheses to elucidate underlying mechanisms, design principles, and unexpected properties of biologically inspired materials. 
Based on large-scale ontological knowledge graphs, SciAgents samples a viable path between concepts of interest, formulates a pertinent hypothesis, and expands it into a full research proposal with detailed hypothesis-testing methods and criteria. 
It employs two dedicated agents to review, critique, and improve the proposed hypothesis, but does not include the step of hypothesis testing through actual experiments.
Similarly, Su et al. \cite{su2024TwoHeads} and Baek et al. \cite{baek2024ResearIterat} proposed leveraging teamwork---such as collaborative discussions and agent critics---to produce novel and effective scientific hypotheses.
In addition, Gower et al. \cite{gower2023LGEMFirstO} introduced LGEM\textsuperscript
{+}, which utilizes a first-order logic framework to describe biochemical pathways and generate 2,094 unique candidate hypotheses for the automated abductive improvement of genome-scale metabolic models in the yeast  \textit{S.~cerevisiae}.

\textbf{Hypotheses only become knowledge after being justified through computational or experimental observations. }
Lu et al. \cite{lu2024AIScient} introduced the AI Scientist, a system designed for fully automated scientific discovery. 
The AI Scientist can conduct research independently and communicate its findings, as demonstrated in three machine learning subfields---diffusion modeling, transformer-based language modeling, and learning dynamics. 
It generates original research ideas, writes code, performs computational experiments, visualizes results, drafts complete scientific papers, and even simulates a peer review process for evaluation. 
For instance, it proposed the hypothesis that ``adaptive dual-scale denoising can improve diffusion models by balancing global structure and local details in generated samples,'' which was justified through image generation tests on four 2D datasets. 
Similarly, Schmidgall et al. \cite{schmidgall2025agent} developed the Agent Laboratory to autonomously carry out the entire research process, including literature review, computational experimentation, and report writing. 
They evaluated Agent Laboratory's capability for knowledge discovery by addressing five research questions in computer vision and natural language processing, achieving an average human-evaluated experiment quality score of 3.2 out of 5.  
In addition, Tiukova et al. \cite{tiukova2024GenesiAutomaa} developed Genesis, an automated system capable of controlling one thousand $\mu$-bioreactors, performing mass spectrometry characterization,  accessing a structured domain information database, and applying experimental observations to improve systems biology models. 
Genesis can initiate and execute 1,000 hypothesis-driven closed-loop experimental cycles per day.
Using a similar approach, the Genesis team has advanced the yeast (\textit{S. cerevisiae}) diauxic shift model, outperforming the previous best and expanding its knowledge by 92 genes (+45\%) and 1,048 interactions (+147\%)  \cite{coutant2019ClosedCycles}.  
This knowledge also advances our understanding of cancer, the immune system, and aging.
Similarly, Gottweis et al. \cite{gottweis2025AICoscie} introduced the AI co-scientist, which autonomously generates and refines novel research hypotheses, with \textit{in vitro} validation in three biomedical areas: drug repurposing, novel target discovery, and mechanisms of bacterial evolution and antimicrobial resistance.

\textbf{Discovered knowledge enhances the agent's mental states, such as $M_t^{\mathrm{mem}}$, $M_t^{\mathrm{wm}}$, and $M_t^{\mathrm{rew}}$.}
Tang et al. \cite{tang2025ChemAgSelfup} developed ChemAgent, which improves chemical reasoning through a dynamic, self-updating memory, $M_t^{\mathrm{mem}}$. 
ChemAgent proposes hypothetical answers to chemistry questions in a development dataset, evaluates them against the ground truth, and simulates the hypothesis-testing process used in real-world research. 
Correct answers are then stored as knowledge in its memory to support future chemistry question answering. 
This self-updating memory resulted in performance gains of up to 46\% (with GPT-4) when ChemAgent was applied to four chemical reasoning datasets from SciBench \cite{wang2024SciBenEvalua}. 
Wang et al. \cite{wang2024efficievolut} introduced Molecular Language-Enhanced Evolutionary Optimization (MOLLEO), which iteratively proposes hypotheses for modifying candidate drug molecules in $M_t^{\mathrm{mem}}$, evaluates their drug-likeness and activity, and updates the candidates in $M_t^{\mathrm{mem}}$ to enhance drug discovery. 
Similarly, Jia et al. \cite{jia2024LLMatDAutono} developed LLMatDesign, which employs hypothesis-guided structure generation and a self-updating $M_t^{\mathrm{mem}}$ to design inorganic photovoltaic materials, whose ideality is defined by matching the target band gap and having the most negative formation energy.

Sim et al. \cite{sim2024ChemOS20} introduced ChemOS 2.0, which orchestrates closed-loop operations in chemical self-driving laboratories (SDLs). 
ChemOS 2.0 integrates \textit{ab initio} calculations, experimental orchestration, and statistical algorithms for the autonomous discovery of high-performance materials. 
A case study on discovering organic laser molecules demonstrates its capabilities. 
It employs a Bayesian optimizer, Altas, as its world model $M_t^{\mathrm{wm}}$ to predict the optical properties of hypothetical molecules---specifically Bis[(N-carbazole)styryl]biphenyl (BSBCz) derivatives---including gain cross section and spectral grain factor.
Based on these predictions, ChemOS 2.0 recommends molecules with a higher probability of success in the experimental campaign. 
It then utilizes an optical characterization platform and the AiiDA software package to measure and simulate the properties of test molecules. 
The results are used to update $M_t^{\mathrm{wm}}$, improving the accuracy of future experimental predictions. 

Hysmith et al. \cite{hysmith2024FutureSelfdr} published a perspective highlighting the crucial role of reward function design in developing forward-looking workflows for SDLs. 
Agents can be highly effective at solving POMDP problems in simulated environments, such as computer games or simulations, but often struggle with real-world applications.
A well-defined reward function is essential for iterative self-evolution. 
However, in many real-world scientific research problems, reward functions are ill-defined or absent at the end of experimental campaigns due to the lack of direct measurements, the complexity of experimental results, and the need to balance multiple objectives. 
The discovery of new knowledge can serve as a valuable resource for refining $M_t^{\mathrm{rew}}$, guiding hypothesis exploration and experimental data collection. 

\subsection{Protocol Planning and Tool Innovation}
The capability to plan experimental protocols and optimize tool usage enables the agent to solve complex scientific puzzles within the autonomous discovery loop. 
As introduced in Section \ref{sec:tool_optimization}, the agent can systematically evaluate and refine its approach to selecting, invoking, and integrating available tools---and even develop new tools tailored to specific task requirements. 
While optimized protocols and tool usage do not directly reduce $D_{\mathrm{K}}(\theta, M_t^{\mathrm{mem}})$, they enhance execution efficiency and effectiveness in refining the probability distribution of unknown information, ${P_\theta \left({\bf x}_{\mathrm{U}}|M_t^{\mathrm{mem}}\right)}$, thereby accelerating knowledge discovery. 
In this scenario, the agent leverages the reasoning function $\mathrm{R}$ to translate its evolving mental states $M_t$, continuously updated with new knowledge, into real-world actions $a_t$ for more effective and faster hypothesis testing (Figure \ref{fig:innovation_loop}). 

\textbf{Scheduling and orchestrating the selection and recombination of existing tools is critical.}
Scientific experiments typically depend on diverse instruments for analyzing reaction products, with decisions rarely rely on just one measurement.
Effectively utilizing necessary instruments without wasting resources and time requires the agent to learn to use tools in an integrated and adaptive manner. 
Dai et al. \cite{dai2024AutonoMobile} designed a modular workflow that integrates mobile robots, an automated synthesis platform, and various characterization instruments for autonomous discovery.
They exemplified this system across three domains: structural diversification chemistry, supramolecular host-guest chemistry, and photochemical synthesis. 
The mobile robot follows a synthesis-analysis-decision cycle to mimic human experimental strategies, autonomously determining subsequent workflow steps.
It selects appropriate instruments, such as the Chemspeed ISynth platform for synthesis, a liquid chromatography-mass spectrometer (UPLC-MS) for measuring mass spectra corresponding to chemical peak signals, and a benchtop nuclear magnetic resonance spectrometer (NMR) for tracking chemical transformations from starting materials to products.

Beyond individual laboratories, tool orchestration is essential for delocalized and asynchronous scientific discovery. 
Strieth-Kalthoff et al. \cite{strieth-kalthoff2024DelocaAsynch} demonstrated a closed-loop integration of five materials science laboratories across three continents, advancing delocalized and democratized scientific discovery. 
These five laboratories have varying strengths---for example, the University of British Columbia specializes in continuous preferential crystallization, while Kyushu University excels in thin film fabrication and characterization.
Strieth-Kalthoff et al. employed a cloud-based experiment planner to continuously learn from the incoming data and effectively prioritize informative experiments across the five laboratories, resulting in the discovery of 21 new state-of-the-art materials for organic solid-state lasers.

\textbf{Moreover, the agent can optimize existing tools and even create new ones to enhance its capabilities. }
Swanson et al. \cite{swanson2024virtual} developed the Virtual Lab, an AI-driven research environment that facilitated the design and experimental validation of new SARS-CoV-2 nanobodies. 
Within the Virtual Lab, AI agents conduct scientific discussion in team meetings and execute specialized tasks in individual sessions.
One key agenda for the agents was developing tools to aid in the design of nanobody binders \cite{xiao2025protein}, including: (1) a sequence analysis tool that ranks candidate point mutations using log-likelihood ratios from the ESM protein language model \cite{lin2023EvolutPredica}; (2) a structure evaluation tool that extracts interface pLDDT scores from AlphaFold-Multimer predictions \cite{evans2021ProteiComple}, offering a proxy for antibody-antigen binding affinity; and (3) an energy estimation tool built on Rosetta \cite{boorla2023NovoDesign} to quantify binding strength between nanobody variants and the spike protein.
These agent-generated tools enabled the Virtual Lab to discover two novel nanobodies with enhanced binding to the JN.1 or KP.3 SARS-CoC-2 variants, while preserving strong affinity for the ancestral viral spike protein.

\subsection{Data Analysis and Implication Derivation}
Although most knowledge discovery processes rely on generating hypotheses and testing them in the real world---where observations $o_t$ are essential---a significant portion of knowledge can be derived purely through internal actions such as iterative reasoning and deep thinking, which are common in theoretical disciplines. 
For example, all theorems in Euclidean geometry can be deduced from just five axioms, but these theorems do not explicitly exist in the mental state before they are derived. 
Given all necessary premises, such as Euclid's five postulates, the true probability of a hypothesis may remain elusive. 
However, using deductive and inductive reasoning to draw implications from known premises and data can help either justify or falsify hypotheses, thus reducing $D_{\mathrm{K}}(\theta, M_t^{\mathrm{mem}})$ and enhancing $IQ_t^{\mathrm{agent}}$ (Figure \ref{fig:innovation_loop}). 
In this scenario, the agent employs the cognition function $\mathrm{C}$ to use prior mental states $M_{t-1}$ and internal actions $a_t$ to derive new knowledge and update mental states to $M_{t}$. 

\textbf{Deductive reasoning enables knowledge derivation through logic.} 
Trinh et al. \cite{trinh2024SolvinOlympi} developed AlphaGeometry for the forward deduction of new mathematical theorems based on existing theorems in Euclidean plane geometry.
AlphaGeometry employs a neural language model to construct auxiliary points in plane geometry problems and integrates specialized symbolic engines to exhaustively deduce new true statements, thereby expanding the joint closure of known truths.
By leveraging this expanded closure, it alternates between auxiliary constructions and symbolic reasoning engines to uncover further implications.
AlphaGeometry demonstrated remarkable performance on a test set of 30 recent Olympiad-level problems, solving 25---more than double the 10 problems solved by the previous best method---and coming close to the level of an average International Mathematical Olympiad (IMO) gold medalist.

\textbf{Inductive reasoning enables knowledge derivation through pattern recognition and statistical learning.}
Liu et al. \cite{liu2024TeamAImade} introduced the Team of AI-made Scientists (TAIS) to simulate the role of a data scientist for streamlined data analysis. 
TAIS decomposes a complex data analysis problem into different computational tasks, including coding, self-critique, and regression analysis, to extract meaningful insights from complex datasets.
When applied to identifying disease-predictive genes, TAIS achieved an overall success rate of 45.73\% on a benchmark dataset containing 457 genetic questions.
Ideally, the extracted insights should be logically sound; otherwise, they must be discarded to ensure only accurate findings are safely integrated into mental states. 

When no new external observations are needed, agents can still synthesize and communicate knowledge by retrieving evidence, organizing it, and drafting coherent narratives. Shao et al.~\cite{shao2024assistingwritingwikipedialikearticles} propose STORM, a workflow that researches a topic through multi-perspective question asking, verifies sources, plans a sectioned outline, and writes a Wikipedia-like article from scratch. By turning dispersed information into a structured, cited article, such systems improve an agent's practical understanding and decision quality through internal reasoning and retrieval rather than physical discovery.
However, limitation in data coverage and the implementation of analysis algorithms may lead to hallucinated insights, underscoring the need for reliable data analyzers and reasoning tools to prevent over-analysis.

\section{Technological Readiness and Challenges}

\lettrine[lines=3]{\initfamily\textcolor{darkgreen}{T}}{he} self-evolution of agents, which in turn drives the advancement of human knowledge, is promised by their early success in the innovation cycle.
This cycle involves generating meaningful hypotheses, designing real-time testing protocols, coordinating various experimental and computational tools, analyzing data, deriving implications, and engaging in self-reflection.
However, achieving fully autonomous self-evolution remains a significant challenge, given the current technology readiness levels (TRLs) of three fundamental capabilities: real-world interaction, complex reasoning, and the integration of prior knowledge.
Further technological progress is required to improve the cycle of self-driven innovation.

\subsection{Real-World Interaction Challenges}

Agents interact with the real world primarily through application programming interfaces (APIs).
While numerous demonstrations \cite{ou2024WorldAWorld} have shown their strong capability to use various APIs, a significant bottleneck in autonomous knowledge discovery remains: the lack of APIs that allow agents to directly execute tasks in a physical laboratory. 
Physical APIs—interfaces that enable direct control of lab equipment—are far less abundant than computational APIs due to the significant investment of time, expertise, and cost required to develop them. 
Although existing autonomous laboratories have shown promise, they remain in an early developmental stage (typically TRL 4–6), where straightforward replication or scale-up is challenging. 
Consequently, building further systems or broadening their application across additional scientific domains still requires substantial customization to address domain-specific needs, along with specialized expertise.

Two key tasks are essential for enabling real-world interaction: \textit{operating lab devices} and \textit{transferring samples between devices}. 
Seamless integration of physical hardware and experimental samples is crucial to maintaining uninterrupted workflows. 
However, most experimental instruments are originally designed for human operation.
Making them accessible to agents requires extensive efforts across multiple disciplines, including robotics, electrical engineering, mechanical engineering, and software programming.
The rising prominence of SDLs is catalyzing the transformation of human-operated devices into agent-accessible systems through APIs.
In autonomous labs conducting complex experiments, two parallel and often complementary approaches are commonly adopted to integrate hardware with agentic systems.
Both approaches are modular, reconfigurable, and valuable, yet they require ongoing, dedicated development. 

\textbf{Approach 1: API Integration via Direct Device Adaptation.}  
This approach involves equipping individual devices with dedicated mechanical adaptations and I/O controllers, enabling them to receive and execute commands from a central control PC. 
For example, to achieve solid-state synthesis and structural characterization of inorganic materials, A-lab has implemented 16 types of devices to automate experimental tasks such as powder dosing, heating, and diffraction \cite{fei2024AlabOSPython}.
This approach allows laboratories to function as fully integrated entities by maximizing device utilization, optimizing space and resources, and enabling bespoke tools. 
However, it is costly, time-consuming, and requires expert knowledge to prototype or retrofit devices for automation.
Large language models (LLMs) have been applied to facilitate access to diverse tools, as illustrated by CACTUS, a Chemistry Agent Connecting Tool-Usage to Science \cite{mcnaughton2024cactus}.
Model Context Protocol (MCP) \cite{introdmodel} also has the potential to further streamline the integration of scientific tools into agentic AI systems.

A more accessible alternative for small teams is the \textit{cloud lab} or \textit{science factory} \cite{vescovi2023towards}, where responsibility for device engineering shifts from individual laboratories to dedicated user facilities or commercial service providers.
For instance, Boiko et al. \cite{boiko2023autonomous} demonstrated an autonomous chemical research agent, Coscientist, capable of carrying out cross-coupling Suzuki and Sonogashira reactions using experimental setups at the Emerald Cloud Lab \cite{emeraldcloudlabECLDocume}.
However, cloud labs offer only a fixed set of pre-built devices optimized for common procedures, posing potential challenges for researchers whose experiments require equipment customization, as integrating non-standard tools may involve a lengthy process of negotiation and development.

\textbf{Approach 2: Robotic Operation of Experimental Devices. }
This approach involves using mobile robots or robotic arms to operate existing devices and transfer samples.
In many cases, robots can interact with instruments without modification, apart from minor adjustments such as adding specialized actuators, grippers, or holders.
For example, Dai et al. \cite{dai2024AutonoMobile} employed mobile robots to explore synthetic chemistry.
In their autonomous laboratory, mobile robots enable physical linkages between synthesis and analysis devices that are spatially separated, automating sample transportation and handling. 
In principle, the robots can perform all actions human researchers require in the laboratory. 
However, current robotic systems still rely on human pre-programming to map the lab layout, define movement trajectories, and register device positions.
Handling unexpected or adaptive situations remains a challenge, as pre-programming cannot anticipate every possible state of an experimental setup.
Real-time learning and adaptive manipulation are active areas of research that require further technological advancements.
In the long term, embodied AI \cite{duan2022SurveyEmbodi} is expected to enhance robotic learning, allowing agents to quickly adapt to new environments and tools.

The two approaches can be combined.
For example, Vescovi et al. \cite{vescovi2023towards} define a modular laboratory robotics architecture that allows for translating high-level commands into specific operations for a variety of different robotic apparatus and laboratory equipment, and for linking robotic apparatus with other elements of an AI-driven discovery architecture, such as high-performance computing \cite{vescovi2022linking}. This architecture has been used to automate experiments in both the biological and physical sciences \cite{ozgulbas2023robotic}. 
Similarly, Fernando et al. \cite{fernando2024facile} integrate a Robotic Operating System 2 (ROS2) compatible robot into the Bluesky experimental orchestration framework.
Lo et al. \cite{lo2024review} argue for the development and integration of low-cost ``frugal twins'' of more expensive equipment to facilitate experimentation and democratize access.

\subsection{Complex Reasoning Challenges}
A fundamental philosophical question is whether agents, often powered by LLMs, can truly perform reasoning.
By definition, languages models generate outputs by predicting the next token, a mechanism fundamentally different from human reasoning.
From an outcome-driven perspective, these input-output systems exhibit reasoning ability phenomenologically, as they produce meaningful outputs compared to a reference system generating arbitrary responses \cite{abel2025AgencyFrameD}.
However, regardless of the perspective taken, this capability remains imperfect---particularly when handling complex logical and numerical problems, which are crucial for scientific knowledge discovery.

\textbf{Agents and LLMs struggle with hard reasoning tasks.}
Glazer et al. \cite{glazer2024FrontiBenchm} introduced FrontierMath, a benchmark comprising hundreds of original and challenging mathematics problems covering most major branches of modern mathematics.
Evaluation of state-of-the-art LLM-driven agents---including o1-preview (OpenAI), o1-mini (OpenAI), GPT-4o (OpenAI, 2024-08-06 version), Claude 3.5 Sonnet (Anthropic, 2024-10-22 version), Grok 2 Beta (XAI), and Gemini 1.5 Pro 002 (Google DeepMind)---revealed that no model achieved even a 2\% success rate on the full benchmark.
Chen et al. \cite{chen2024SciencRigoro} presented ScienceAgentBench, a benchmark designed to evaluate language agents in data-driven scientific discovery. 
Among 102 tasks derived from 44 peer-reviewed publications across four disciplines, OpenAI o1 successfully solved only 42.2\% of them.
Chollet \cite{chollet2019MeasurIntell} proposed the Abstraction and Reasoning Challenge (ARC) to assesss LLMs' ability to perform abstract inductive reasoning without relying on memorization or external knowledge. 
Even with careful prompting, GPT-4o correctly solved only 19\% of the tasks, far below the $\sim 75\%$ average human performance \cite{legris2024HARCRobust,wu2025UndersLLMs}.
Zhu et al. \cite{allen-zhu2025DOGEReform} suggested a four-level classification of AI intelligence, including L1 (arbitrating isputes), L2 (auditing a review), L3 (reviewing a paper), and L4 (authoring a paper). 
They classify the current state-of-the-art LLM-driven agents as approaching L2-level capabilities.
To enhance agents' reasoning abilities, researchers have introduced techniques such as chain-of-thought \cite{wei2023ChainoPrompt}, tree-of-thoughts \cite{yaotree}, and \cite{yao2022react}.
Although new methods continue to emerge, as discussed in Section \ref{sec:reasoning}, further advancements in reasoning capacity remain crucial for achieving reliable causal inference in scientific research.

\textbf{Agents and LLMs also struggle with quantitative and symbolic problems.}
For example, GPT-4 and GPT-3.5 often struggle with reliably performing complex arithmetic such as multiplying $12,345 \times 98,765$, or translating IUPAC chemical names into accurate molecular graphs \cite{white2023AssessChemisa,Andres2024ChemCrow}.
A common approach to overcoming these limitations is to use external tools rather than relying on the LLM itself for reasoning. 
In mathematical problem-solving, for example, tools like symbolic solvers are preferred over direct LLM inference \cite{trinh2024SolvinOlympi}.
However, this mitigation does not resolve the intrinsic deficiency in numerical understanding, which poses a potential risk to scientific reasoning. 
Moreover, Yu et al. \cite{yu2024tooling} found that tool-augmented LLMs do not consistently outperform base LLMs without tools in chemistry problem-solving. 
For instance, for \textit{specialized} chemistry tasks, such as synthesis prediction, augmenting LLMs with specialized tools can boost the performance substantially; however, tool augmentation is less effective for \textit{general} chemistry questions, such as those in exams, where no specific tools can directly solve a given question. 
In these scenarios, an agent's ability to reason correctly by using multiple pieces of chemistry knowledge becomes more important.
    
The preceding discussion emphasizes the importance of developing robust methodologies for evaluating AI agents as scientific research assistants, a topic discussed at length by Cappello et al. \cite{cappello2025eaira}.

\subsection{Challenges in Integrating Prior Knowledge}

Prior knowledge is a crucial factor for higher intelligence. 
As discusses in Section \ref{sec:IQ_definition}, the agent's prior knowledge, $M_t^{\mathrm{mem}}$, helps decrease $D_{\mathrm{K}}(\theta,M_t^{\mathrm{mem}})$ and increase the agent's intelligence, $IQ_t^{\mathrm{agent}} $. 
Human-led scientific discoveries frequently achieve breakthroughs with relatively small datasets, thanks to the vast prior knowledge humans possess. 
The start-of-the-art LLMs that power autonomous agents are trained on nearly all publicly available textual data, including websites, books, and other sources, thereby encompassing most common knowledge as well as publicly accessible specialized knowledge. 
However, achieving an agent that can seamlessly integrate all existing human knowledge remains a significant challenge.

At least three types of knowledge sources may not be included in LLM pre-training: 
(1) Paywalled or unpublished knowledge, including non-open-access publications, industry-specific data, and failed experiments \cite{raccuglia2016MachinMateri}. 
They are often not accessible to public models despite their potential value in refining domain-specific insights.
(2) Empirical knowledge. 
Heuristic decisions by experts are often effective, particularly in scenarios where no existing data is available for a new problem. 
However, large amounts of expert heuristics are typically not accessible as textual data.
(3) Contextual or situational knowledge.
Knowledge related to real-world conditions, such as safety protocols in chemical reactions or equipment handling, is often absent from pre-trained models but is essential for practical applications.

Additionally, integrating diverse knowledge sources presents challenges in reconciling conflicting information. 
For example, OpenAI's Deep Research \cite{openaiIntrodDeep} actively gathers online information and performs multi-step reasoning, achieving state-of-the-art performance on Humanity's Last Exam and the GAIA benchmark.
However, it still struggles to distinguish between authoritative information and rumors and exhibits limitations in confidence calibration, often misrepresenting its level of certainty \cite{openaiIntrodDeep}. 
Establishing a system to assess the levels of evidence \cite{goodman2005IntrodBayesi} of different knowledge fragments---such as quantifying reliability and verifying references---may be necessary for effective knowledge fusion.

\section{Summary and Discussion}
\label{sec:scientific-discovery-summary-discussion}

\lettrine[lines=3]{\initfamily\textcolor{darkgreen}{I}}{n} this chapter, we explored the pivotal role of intelligent agents in driving autonomous scientific discovery, emphasizing how agents can foster self-evolution and innovation cycles through iterative knowledge acquisition, hypothesis testing, and continuous improvement. We proposed a systematic framework for measuring agent intelligence, defining it via the Kullback–Leibler divergence between the agent's predicted and the real-world probability distributions of unknown information. This statistical measure provides a robust method for quantifying an agent's capability to make accurate, predictive models of natural phenomena.

We discussed three core interactions between agents and scientific knowledge: (1) hypothesis generation and testing, (2) protocol planning and tool innovation, and (3) data analysis and implication derivation. Hypothesis generation is crucial for formulating valuable scientific conjectures, while hypothesis testing through computational or experimental validation is essential for transforming conjectures into verified knowledge. Protocol planning and tool innovation enable agents to orchestrate and optimize experimental workflows effectively, leveraging existing tools or creating novel ones to enhance scientific exploration efficiency. Data analysis and implication derivation allow agents to extract meaningful insights from existing knowledge and observations, employing both deductive logic and inductive pattern recognition.

We reviewed notable advancements in autonomous scientific discovery, including systems like the AI Scientist, ChemOS 2.0, and the Virtual Lab, showcasing the potential of agents to independently navigate complex scientific processes across theoretical, computational, and experimental domains. However, despite these successes, significant challenges remain in achieving fully autonomous agent-driven scientific discovery.

Three primary technological challenges were identified: (1) real-world interaction, specifically the limited availability and complexity of physical APIs necessary for seamless laboratory integration; (2) complex reasoning, where current language models still struggle with sophisticated logic, numerical computation, and symbolic reasoning; and (3) integration of prior knowledge, particularly regarding inaccessible or unpublished data, empirical expert heuristics, and situational knowledge. Addressing these limitations requires concerted advancements in interdisciplinary integration, robust reasoning methodologies, and comprehensive strategies for assimilating diverse knowledge sources.

Looking forward, enhancing agent autonomy in scientific discovery will necessitate overcoming these barriers through interdisciplinary collaborations, robust experimental validations, and methodological innovations in reasoning and knowledge management. Ultimately, achieving higher autonomy in scientific agents holds the promise of accelerating human knowledge expansion, driving technological innovation, and sustaining an iterative cycle of continuous agent and human evolution.

\part{Collaborative and Evolutionary Intelligent Systems}
\label{part-society}

\begin{figure*}[!htb]
    \centering
    \includegraphics[width=0.8\textwidth]{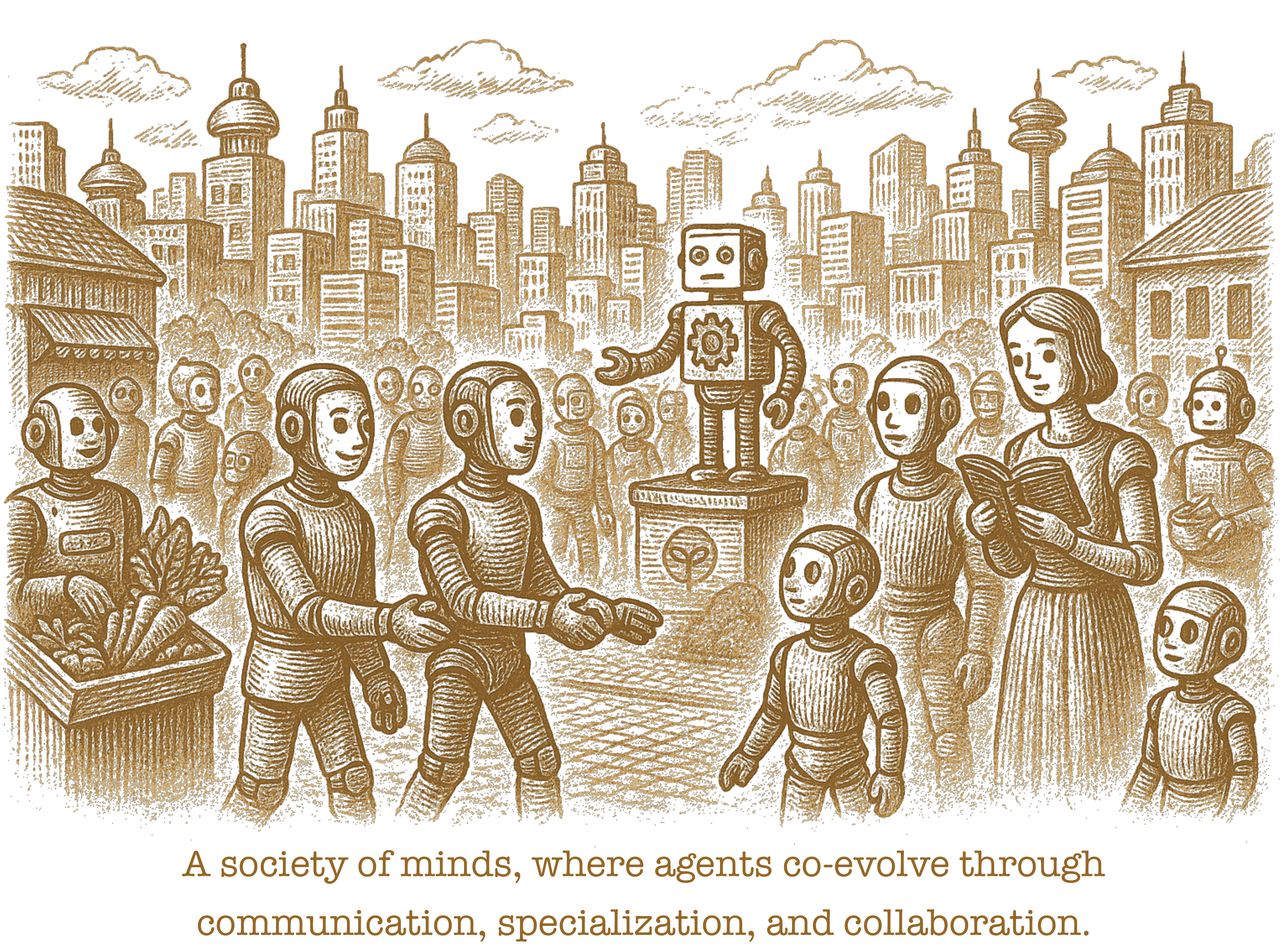}
    \label{fig:part3}
\end{figure*}

The concepts of \emph{collaboration} and \emph{evolution} lie at the heart of intelligent multi-agent systems (MAS). Inspired by biological ecosystems and human societal dynamics, these systems leverage \emph{collective intelligence} to solve complex challenges that exceed the capabilities of individual agents~\cite{li2024survey}.
Human societies exemplify how \emph{cooperation}, \emph{specialization}, and \emph{distributed decision-making} significantly enhance collective problem-solving effectiveness. Similarly, MAS adopts these strategies, integrating specialized agents to address intricate tasks collaboratively.
The foundational principle of collective intelligence, the ``Wisdom of Crowds'' by~\cite{surowiecki2005wisdom}, suggests diverse, independent agents often yield superior decisions compared to solitary experts, directly underpinning the design philosophy of MAS.
Cognitive theories, such as Minsky's \emph{society of mind}\cite{minsky1988society} and the theory of mind\cite{frith2005theory,li2023theory}, further reinforce this paradigm by proposing that intelligence emerges from structured interactions among specialized units.

Recently, advancements in large language models (LLMs) have introduced new possibilities for collaborative and evolutionary LLM-based multi-agent systems (LLM-MAS). Benefiting from powerful reasoning, planning, and decision-making capabilities, these models enable the creation of sophisticated MAS architectures mirroring the cooperative and adaptive characteristics found in human societies. Agents within LLM-MAS often assume distinct identities and roles, reflecting human-like division of labor and specialized collaboration. By embracing structured communication, dynamic knowledge sharing, and coordinated decision-making, these systems emulate human social dynamics to achieve common goals. Moreover, LLM-MAS is inherently evolutionary: agents continuously adapt and improve through interactions, feedback, and iterative learning, resulting in enhanced system performance over time.

In this part, we systematically survey the emerging field of LLM-based multi-agent systems, focusing specifically on their \emph{collaborative mechanisms} and \emph{evolutionary capabilities}.
We first examine how different system objectives shape agent roles, behavior patterns, and collaborative strategies in~\Cref{ch:mas-formulation-categories}.
Next, in ~\Cref{ch:construct-mas}, we analyze various communication structures, including interaction protocols, and the collaborative paradigms and decision-making methodologies that facilitate effective agent-agent and human-agent communication. 
Furthermore we discuss the collective intelligence and evolution mechanism in ~\Cref{ch:MAS-evolution}.
Finally, in ~\Cref{ch:MAS-evaluation}, we discuss evolutionary processes, highlighting adaptive learning methods, continuous knowledge sharing, and mechanisms for iterative improvement that collectively improve MAS performance.
Through this part, we aim to identify current achievements, discuss existing challenges, and highlight promising research directions for collaborative and evolutionary intelligent systems. The following figure shows a taxonomy of research works about LLM-based multi-agent systems.

\begin{figure*}[t!]
\centering
\footnotesize
    \begin{forest}
        for tree={
            forked edges,
            draw,
            rounded corners,
            node options={align=center,},
            s sep=5pt,     
            l sep=4.5pt,     
            inner ysep=1.1pt,   
            calign=center,
            grow=east,
            reversed=true,
            anchor=base west,
            parent anchor=east,
            child anchor=west,
            base=left,
            font=\small,
            minimum width=2.5em
          },
          where level=1{text width=4em,fill=customblue!50}{},
          where level=2{text width=5em,fill=customgreen!50}{},
          [LLM-based MAS, fill=gray!20
            [Application, for tree={
                answer, text width=8em,
                calign=child edge, calign child=(n_children()+1)/2
                },
                [Strategic Learning 
                    [RECONCILE~\cite{chen2023reconcile}
                    LLM-Game-Agent~\cite{Lan2023LLMBasedAS} BattleAgentBench~\cite{wang2024battleagentbench}, text width=23.5em]            
                ]
                [Modeling and Simulation 
                    [Generative Agents~\cite{park2023generative} Agent hospital~\cite{li2024agent}
                    MedAgents~\cite{tang2024medagents}
                    MEDCO~\cite{wei2024medco}, text width=23.5em]           
                ]
                [Collaborative Task Solving                
                    [MetaGPT~\cite{hong2023metagpt}
                    ChatDev~\cite{qian2024chatdev}
                    Agent Laboratory~\cite{schmidgall2025agent} 
                    The virtual lab~\cite{swanson2024virtual}, text width=23.5em]
                ]
            ]
            [Composition \& Protocol, for tree={
                answer, text width=8em,
                calign=child edge, calign child=(n_children()+1)/2
                },
                [Agents
                    [Homogeneous
                        [CoELA~\cite{zhang2023building}
                        VillagerAgent~\cite{dong2024villageragentgraphbasedmultiagentframework}
                        LLM-Coordination~\cite{agashe2024llmcoordination}, 
                        text width=14em]
                    ]
                    [Heterogeneous
                        [MetaGPT~\cite{hong2023metagpt}
                        ChatDev~\cite{qian2024chatdev}
                        Generative Agents~\cite{park2023generative}
                        S-Agents~\cite{chen2024s},text width=14em]
                    ]
                ]
                [Protocols
                    [Message Types
                        [SciAgents~\cite{ghafarollahi2024SciAgeAutoma}
                            AppAgent~\cite{Chi2023AppAgent}
                            MetaGPT~\cite{hong2023metagpt}, text width=14em]
                    ]      
                    [Communication Interfaces   
                        [AgentBench~\cite{liu2023agentbench}
                        VAB~\cite{liu2024vab}
                        TaskWeaver~\cite{qiao2024taskweaver}
                        HULA~\cite{takerngsaksiri2025humanintheloop}, text width=14em]
                    ]
                    [Next-Gen Protocol                       
                    [MCP~\cite{anthropic_mcp}
                    ANP~\cite{agent_network_protocol}
                    A2A~\cite{rao2025a2a}
                    Agora~\cite{marro2024scalable}
                    IoA~\cite{chen2024internet}, text width=14em]
                ]
                ]
            ]
            [Topology, for tree={
                answer, text width=8em,
                calign=child edge, calign child=(n_children()+1)/2
                },
                [Static  [
                    MEDCO~\cite{wei2024medco}
                    Agent hospital~\cite{li2024agent}
                    Welfare Diplomacy~\cite{Mukobi2023WelfareDB}
                    MedAgents~\cite{tang2024medagents}
                    , text width=23.5em]
                ]
                [Dynamic 
                    [DyLAN~\cite{liu2023dynamic}
                    GPTSwarm~\cite{zhuge2024gptswarm}
                    CodeR~\cite{chen2024coder}
                    Oasis~\cite{yang2024oasis}
                    , text width=23.5em]
                ]
            ]
            [Collaboration, for tree={
                answer, text width=8em,
                calign=child edge, calign child=(n_children()+1)/2
                },
                [Agent-Agent Collaboration
                    [Consensus-oriented
                        [Agent Laboratory~\cite{schmidgall2025agent}
                        The virtual lab~\cite{swanson2024virtual}
                        OASIS~\cite{yang2024oasis}, text width=14em]
                    ]
                    [Collaborative learning 
                        [Generative Agents~\cite{park2023generative}
                        Welfare Diplomacy~\cite{Mukobi2023WelfareDB}
                        LLM-Game-Agent~\cite{Lan2023LLMBasedAS} BattleAgentBench~\cite{wang2024battleagentbench}, text width=14em]
                    ]
                    [Teaching/Mentor 
                        [MEDCO~\cite{wei2024medco}
                        Agent Hospital~\cite{li2024agent}
                        , text width=14em]
                    ]
                    [Task-oriented
                         [MedAgents~\cite{tang2024medagents}
                         S-Agents~\cite{chen2024s}
                         , text width=14em]
                    ]
                ]
                [Human-AI Collaboration
                     [Dittos~\cite{leong2024Dittos}
                     PRELUDE~\cite{gao2024aligning}
                     CowPilot~\cite{huq2025cowpilot}
                     , text width=23.5em]
                ]
            ]
            [Evolution, for tree={
                answer, text width=8em,
                calign=child edge, calign child=(n_children()+1)/2
                },
                [Collective Intelligence 
                    [Generative Agents~\cite{park2023generative} Welfare Diplomacy~\cite{Mukobi2023WelfareDB}
                    LLM-Game-Agent~\cite{Lan2023LLMBasedAS} BattleAgentBench~\cite{wang2024battleagentbench}, text width=23.5em]
                ]
                [Individual Adaptability 
                    [Agent Hospital~\cite{li2024agent}
                    Agent Laboratory~\cite{schmidgall2025agent}
                    MEDCO~\cite{wei2024medco}, text width=23.5em]
                ]
            ]
            [Evaluation, for tree={
                answer, text width=8em,
                calign=child edge, calign child=(n_children()+1)/2
                },
                [Benchmark for specific tasks
                    [MBPP~\cite{dataset-mbpp}
                    HotpotQA~\cite{dataset-hotpot-qa}
                    MATH~\cite{dataset-math}
                    SVAMP~\cite{dataset-svamp}
                    MultiArith~\cite{dataset-multiarith}, 
                    text width=23.5em]
                ]
                [Benchmark for MAS
                    [Collab-Overcooked~\cite{sun2025collabovercookedbenchmarkingevaluatinglarge} REALM-Bench~\cite{geng2025realmbenchrealworldplanningbenchmark} PARTNR~\cite{chang2024partnrbenchmarkplanningreasoning} VillagerBench~\cite{dong2024villageragentgraphbasedmultiagentframework} AutoArena~\cite{zhao2024autoarenaautomatingllmevaluations} MultiagentBench~\cite{zhu2025multiagentbenchevaluatingcollaborationcompetition}, text width=23.5em]
                ]
            ]
        ]
    \end{forest}    
    \label{fig:llm-mas-taxonomy}
    \caption{A taxonomy of research works about LLM-based Multi-Agent Systems.}
\end{figure*}

\chapter{The Framework and Categories of Multi-Agent Systems}
\label{ch:mas-formulation-categories}

\lettrine[lines=3]{\initfamily\textcolor{darkgreen}{I}}{n} the context of LLM-based multi-agent systems (LLM-MAS), \textit{collaboration goals} and \textit{collaboration norms} serve as foundational elements that shape system behavior, interaction patterns, and overall effectiveness. 
Collaboration goals specify the explicit objectives agents aim to achieve -- whether individually, collectively, or competitively -- while collaboration norms define the rules, constraints, and conventions that govern agent interactions within the system. Together, these components establish a robust framework guiding effective communication, coordination, and cooperation among agents.

This chapter categorizes LLM-MAS into three categories based on distinct combinations of collaboration goals and norms: \textit{strategic learning}, \textit{modeling and simulation}, and \textit{collaborative task solving}. Figure~\ref{fig:llm_mas_types} gives an overview of these categories in LLM-MAS.
Although not exhaustive, these categories cover a wide spectrum of LLM-MAS designs and clearly reflect how system objectives shape agent interactions and outcomes.

\begin{itemize}
\item  \textbf{Strategic Learning} systems embed agents within a game-theoretic context, where agents pursue individual or partially conflicting goals. 
The interactions can be cooperative, competitive, or mixed, guided explicitly by predefined game rules and interaction norms. This setting often aligns with non-cooperative (strategic) and cooperative concepts in traditional game theory. 
\item  \textbf{Modeling and Simulation} contexts focus on agents acting independently, driven by diverse environmental or social factors. Here, interactions emerge organically without necessarily converging on common goals, reflecting the complex dynamics seen in large-scale social or economic simulations. 
\item  \textbf{Collaborative Task Solving} emphasizes systematic cooperation among agents to achieve explicitly shared objectives. Agents typically adopt structured workflows, clear role definitions, and highly predefined collaboration norms to synchronize their actions toward collective goals. 
\end{itemize}

\begin{figure}[t!]
    \centering
    \resizebox{\textwidth}{!}{
    \begin{tikzpicture}[
        font=\small,
        box/.style={
            rectangle,
            draw=black,
            thick,
            fill=customgreen!50,
            minimum width=4.3cm,
            minimum height=4cm,
            text width=4.2cm,
            align=left
        },
        arrowstyle/.style={
            ->,
            >=latex,
            line width=0.8pt,
            shorten >=2pt,
            shorten <=2pt
        }
    ]

    \node[
        rectangle,
        draw=black,
        thick,
        fill=customblue!50,
        text width=5.2cm,
        align=center
    ] (main)
    {\textbf{LLM-Based Multi-Agent Systems}\\[2pt]
     \textit{(Collaboration Goals \& Collaboration Norms)}};

    \node[box, below=1.5cm of main, xshift=-5.4cm] (simulate) {
        \begin{minipage}[t][4cm][t]{4.0cm}
        \hspace{0.1cm}
        \centering
        \textbf{Modeling \& Simulation}\\[2pt]
        \begin{itemize}[leftmargin=3.2em, labelsep=0.4em]
            \setlength\itemsep{0pt}
            \item Agents act largely independently
            \item Heterogeneous behaviors and states
            \item Emergent social, economic, or political phenomena
        \end{itemize}
        \end{minipage}
    };

    \node[box, below=1.5cm of main, xshift=0cm] (strategic) {
        \begin{minipage}[t][4cm][t]{4.0cm}
        \hspace{0.1cm}
        \centering
        \textbf{Strategic Learning}\\[2pt]
        \begin{itemize}[leftmargin=3.2em, labelsep=0.4em]
            \setlength\itemsep{0pt}
            \item Divergent or conflicting goals
            \item Competitive \& cooperative game rules
            \item Dynamic adaptation and anticipatory strategies
        \end{itemize}
        \end{minipage}
    };

    \node[box, below=1.5cm of main, xshift=5.4cm] (collaborative) {
        \begin{minipage}[t][4cm][t]{4.0cm}
        \hspace{0.1cm}
        \centering
        \textbf{Collaborative Task Solving}\\[2pt]
        \begin{itemize}[leftmargin=3.2em, labelsep=0.4em]
            \setlength\itemsep{0pt}
            \item Shared goals \& structured workflows
            \item Clear role assignment
            \item Multi-round cooperation and coordination
        \end{itemize}
        \end{minipage}
    };

    \draw (main.south) -- ++(0,-0.8) coordinate (branch);
    
    \draw[arrowstyle] (branch) -| (simulate.north);
    \draw[arrowstyle] (branch) -- (strategic.north);
    \draw[arrowstyle] (branch) -| (collaborative.north);

    \end{tikzpicture}
    }
    \caption{An overview of three major collaboration types in LLM-based MAS: 
    \textit{Modeling \& Simulation}, \textit{Strategic Learning}, and 
    \textit{Collaborative Task Solving}. 
    Each category is distinguished by how agents' goals and norms are set 
    (independent vs.\ divergent vs.\ shared) and how they coordinate.}
    \label{fig:llm_mas_types}
\end{figure}

In the following, we first introduce a formal framework for multi-agent systems. Then we examine these categories in detail, highlighting how each of them leverages the capabilities of LLMs to shape agent behaviors and interactions.

\section{From Foundation Agent to Foundation \emph{Agents}: A Society-Level Formalism}
\label{sec:foundation-mas-formalism}

\lettrine[lines=3]{\initfamily\textcolor{darkgreen}{W}}{e} treated a \emph{single} \emph{Foundation Agent} in \Cref{chap:intro}: a system that perceives, learns, reasons, and acts in a loop while maintaining a structured mental state (memory, world model, emotion, goals, reward, \dots).  Real deployments rarely stop at one.  We field teams of agents; we mix AI agents with humans; we drop them into worlds governed by currencies, laws, APIs, and social norms.  This section lifts the single-agent loop to the \emph{multi-agent / agent-society} level while keeping all notation consistent with the \emph{Foundation Agent} framework presented in \Cref{chap:intro}.  The goal is a clean bridge: take $n=1$ and collapse the shared structures, and you recover the single-agent loop exactly.

We proceed in three steps.  
First, we formalise the world-as-society view that underlies the Foundation Agent programme.  
Second, we introduce the additional structures needed when many agents co-inhabit that world: a shared workspace, coordination mechanisms, and social-system constraints.  
Third, we write down the closed interaction loop: perception, cognition, coordination, execution, and world update, so that analysis, simulation, and implementation all read from the same page.
Table~\ref{tab:mas-notation} lists the additional symbols introduced for the multi-agent setting and how they relate to the single-agent notation from Table~\ref{tab:notation_summary}.

\begin{table}[!ht]
\centering
\caption{Additional notation for the Foundation Multi-Agent System formalism.  Symbols extend the single-agent notation in Table~\ref{tab:notation_summary}.}
\label{tab:mas-notation}
\renewcommand{\arraystretch}{1.15}
\setlength{\tabcolsep}{5pt}
\begin{tabular}{@{}p{0.18\textwidth}p{0.77\textwidth}@{}}
\toprule
\textbf{Symbol} & \textbf{Meaning} \\
\midrule
\rowcolor{LightMint}
$A=\{1,\dots,n\}$ & Index set of designated AI agents in the society. \\
$\mathfrak{B}$ & All intelligent beings in the world (agents, humans, orgs).  Our $A\subseteq\mathfrak{B}$. \\
\rowcolor{LightMint}
$\Sigma$ & Social systems: rules, norms, infrastructures constraining perception, communication, resources, and reward. \\
$B_t \in \mathcal{B}$ & Shared workspace / blackboard state at turn $t$ (shared memory, interaction log, group goals, resource ledger, \dots). \\
\rowcolor{LightMint}
$P^i$ & Perception for agent $i$; shaped by $M_{t-1}^i$, $B_{t-1}$, and $\Sigma$. \\
$L^i$ & Learning / private mental-state update for agent $i$. \\
\rowcolor{LightMint}
$R^i$ & Reasoning / action-proposal function for agent $i$. \\
$U^i$ & Workspace message constructor for agent $i$ (optional but convenient). \\
\rowcolor{LightMint}
$K$ & Coordination operator aggregating proposals under social-system rules. \\
$E^i$ & Effector / actuation mapping for agent $i$ (high-level $\to$ executable). \\
\rowcolor{LightMint}
$T$ & Environment transition given executed joint actions. \\
$\Omega$ & Shared-workspace transition given aggregated updates and current private states. \\
\rowcolor{LightMint}
$\mathbf{a}_t$ & Authorised joint action vector after coordination. \\
$\mathbf{u}_t^{\mathrm{sh}}$ & Aggregated set of shared updates/messages applied to $B_t$. \\
\bottomrule
\end{tabular}
\end{table}

\subsection{World as Environment \emph{plus} Intelligent Beings under Social Systems}
\label{subsec:world-society}

In the Foundation Agent view, the \emph{world} is never a neutral backdrop.  It already contains other intelligent beings (AI or human) and the layered \emph{social systems}, such as financial, legal, cultural, technical infrastructure, that shape what information flows, what actions are allowed, what resources cost, and how rewards are allocated.

\begin{definition}[\textbf{World as structured society}]
\label{def:world}
Let
\[
  \mathcal{W}
  \;=\;
  \mathrm{SocialSystems}\!\bigl(\mathcal{S},\,\mathfrak{B},\,\Sigma\bigr)
\]
denote a world composed of
\begin{itemize}
  \item an \emph{environment state space} $\mathcal{S}$ with states $s_t\in\mathcal{S}$;
  \item a set $\mathfrak{B}$ of \emph{intelligent beings} (AI agents, humans, organisations, \dots);
  \item a collection $\Sigma$ of \emph{social systems} (markets, legal codes, communication infrastructures, cultural norms), which regulate information access, resource exchange, interaction protocols, and incentives among members of~$\mathfrak{B}$ and the environment~$\mathcal{S}$.
\end{itemize}
\end{definition}

In what follows we focus on a designated subset of beings, our AI agents, interacting with the rest of~$\mathcal{W}$.  Humans may appear explicitly as additional beings or implicitly through~$\Sigma$ (e.g., policy limits, budget approvals, safety reviews).

\subsection{From Single Agent to Multi-Agent Systems}
\label{subsec:single2multi}

Now we introduce the additional structures needed when many agents co-inhabit a shared world.

\paragraph*{From One Mental State to Many: Private vs.\ Shared}
Recall the single-agent notation: observations $o_t\in\mathcal{O}$; actions $a_t\in\mathcal{A}$; mental state $M_t\in\mathcal{M}$ decomposed into memory, world model, emotion, goals, reward signals, and so on; cognition $C=(L,R)$ updating $M_t$ and producing $a_t$; environment transition $T$ (Table~\ref{tab:notation_summary}).  
With $n$ agents we index each agent $i\in\{1,\dots,n\}$:
\[
  M_t^i \in \mathcal{M}^i,\qquad
  o_t^i \in \mathcal{O}^i,\qquad
  a_t^i \in \mathcal{A}^i.
\]
Different agents may have different modalities, tools, or roles, so the spaces need not match across~$i$ (heterogeneous teams are common).

In addition to these \emph{private} mental states we introduce a \emph{shared workspace}, the arena where agents leave messages, post artefacts, negotiate sub-goals, and pool intermediate results.

\begin{definition}[\textbf{Shared workspace / blackboard}]
\label{def:shared-workspace}
At time $t$ the team-level shared state is
\[
  B_t
  \;=\;
  \bigl\langle
      M_t^{\mathrm{sh}},\;
      I_t,\;
      G_t^{\mathrm{grp}},\;
      R_t^{\mathrm{alloc}},\;
      \dots
  \bigr\rangle ,
\]
where $M_t^{\mathrm{sh}}$ is shared memory (documents, code, structured data), $I_t$ logs interaction history, $G_t^{\mathrm{grp}}$ records currently active group or task goals, and $R_t^{\mathrm{alloc}}$ tracks team-level resource budgets or locks.  Ellipses indicate optional additional channels (e.g., reputation ledgers, conflict flags).  We write $B_t\in\mathcal{B}$ for the workspace state space.
\end{definition}
For completeness, we keep the earlier decomposition (cf.\ Sec.~\ref{subsec:agent-framework-symbols}):
\[
  M_t^i
  =
  \bigl\langle
      M_{t}^{i,\mathrm{mem}},
      M_{t}^{i,\mathrm{wm}},
      M_{t}^{i,\mathrm{emo}},
      M_{t}^{i,\mathrm{goal}},
      M_{t}^{i,\mathrm{rew}},
      \dots
  \bigr\rangle .
\]

\paragraph*{Coordination Surfaces: How Social Systems Enter the Loop}
Even when every agent is competent alone, multi-agent behaviour hinges on coordination.  In our formalism, coordination appears in two places:
\begin{enumerate}
\item \textbf{Information shaping.}  Social systems $\Sigma$ can gate or transform what each agent sees: privacy rules, role-based access, rate limits, safety filters.
\item \textbf{Joint decision procedures.} The team may follow protocols such as voting, market-style bidding, role delegation, or consensus review that turn a set of \emph{proposed} actions into what is actually executed.
\end{enumerate}
We collect these influences in a coordination operator
\[
  K:\;
  \Sigma \times B_{t-1} \times \{ \tilde{a}_t^i, u_t^i \}_{i=1}^n
  \;\longrightarrow\;
  \bigl( \mathbf{a}_t,\, \mathbf{u}_t^{\mathrm{sh}} \bigr),
\]
defined below after we explain where the proposals $\tilde{a}_t^i$ and messages $u_t^i$ come from.

\paragraph*{Local Perception and Cognition per Agent}
Each agent runs a perception–cognition update much like the single-agent loop, but conditioned on both the external world and the current shared workspace.
\begin{itemize}
\item \textbf{Perception.}
Agent $i$ forms an observation
\begin{equation}
  o_t^i
  \;=\;
  P^i\!\bigl(s_t,\,M_{t-1}^i,\,B_{t-1};\,\Sigma\bigr),
  \label{eq:mas-perception}
\end{equation}
where the semicolon reminds us that the form of perception (what sensors are available, what data are redacted) can depend on social-system constraints~$\Sigma$.

\item \textbf{Learning (state update).}
The agent updates its private mental state:
\begin{equation}
  M_t^i
  \;=\;
  L^i\!\bigl(M_{t-1}^i,\,a_{t-1}^i,\,o_t^i,\,B_{t-1};\,\Sigma\bigr).
  \label{eq:mas-learning}
\end{equation}
Online fine-tuning, memory write, belief update, and affect change all live here.

\item  \textbf{Reasoning (action proposal).}
Given its updated mental state (and the shared context) the agent proposes its next move:
\begin{equation}
  \tilde{a}_t^i
  \;=\;
  R^i\!\bigl(M_t^i,\,B_{t-1};\,\Sigma\bigr),
  \label{eq:mas-reasoning}
\end{equation}
and may also produce an associated workspace message or update payload
\begin{equation}
  u_t^i
  \;=\;
  U^i\!\bigl(M_t^i,\,\tilde{a}_t^i\bigr),
  \label{eq:mas-workspace-message}
\end{equation}
such as ``here is my partial plan'', ``I need GPU hours'', or ``result attached''.
We group $(\tilde{a}_t^i,u_t^i)$ as the agent's \emph{behaviour proposal} at turn~$t$.
\end{itemize}

\paragraph*{Team Coordination and Execution}
Behaviour proposals are not necessarily executed as-is.  The coordination operator $K$ takes all proposals plus the current workspace and applicable social-system rules and decides what the team actually does and how the shared state changes.
\begin{equation}
  (\mathbf{a}_t,\,\mathbf{u}_t^{\mathrm{sh}})
  \;=\;
  K\!\bigl(\Sigma,\,B_{t-1},\,\{\tilde{a}_t^i,u_t^i\}_{i=1}^n\bigr),
  \label{eq:mas-K}
\end{equation}
where
$\mathbf{a}_t = (a_t^1,\dots,a_t^n)$ are the \emph{authorised} actions to be executed externally
and $\mathbf{u}_t^{\mathrm{sh}}$ the set of workspace updates (possibly filtered, merged, or reordered).

Each authorised action then passes through the agent's effectors:
\begin{equation}
  a_t^{i,\mathrm{exec}} = E^i(a_t^i).
  \label{eq:mas-exec}
\end{equation}

\paragraph*{World and Workspace Transitions}
Once authorised actions are executed, the environment responds.  Allowing for stochasticity:
\begin{equation}
  s_{t+1} \sim T\!\bigl(s_t,\, \{a_t^{i,\mathrm{exec}}\}_{i=1}^n \bigr).
  \label{eq:mas-env-transition}
\end{equation}
Meanwhile, the shared workspace is updated by an internal transition function $\Omega$ that applies the accumulated shared updates plus any system-level bookkeeping (resource debits, goal progression, interaction logs):
\begin{equation}
  B_t = \Omega\!\bigl(B_{t-1},\, \mathbf{u}_t^{\mathrm{sh}},\, \{M_t^i\}_{i=1}^n;\,\Sigma\bigr).
  \label{eq:mas-workspace-transition}
\end{equation}
Note that $B_t$ is indexed by the \emph{post-coordination} turn $t$ (same logical time as the proposals it incorporates).  If greater temporal fidelity is needed (asynchronous teams, streaming logs), $B_t$ can be made event-driven rather than turn-synchronous; the algebra here assumes a batched logical turn for clarity.

\subsection{The Foundation MAS Loop}
\label{subsec:mas-loop}

Collecting Eqs.~\eqref{eq:mas-perception}–\eqref{eq:mas-workspace-transition}, one logical turn of a Foundation Multi-Agent System unfolds as:

\begin{align}
  o_t^i     &= P^i\!\bigl(s_t, M_{t-1}^i, B_{t-1}; \Sigma\bigr), && i=1,\dots,n, \label{eq:mas-loop-1}\\
  M_t^i     &= L^i\!\bigl(M_{t-1}^i, a_{t-1}^i, o_t^i, B_{t-1}; \Sigma\bigr), && i=1,\dots,n, \label{eq:mas-loop-2}\\
  \tilde{a}_t^i &= R^i\!\bigl(M_t^i, B_{t-1}; \Sigma\bigr), && i=1,\dots,n, \label{eq:mas-loop-3}\\
  u_t^i     &= U^i\!\bigl(M_t^i, \tilde{a}_t^i\bigr), && i=1,\dots,n, \label{eq:mas-loop-4}\\[2pt]
  (\mathbf{a}_t,\mathbf{u}_t^{\mathrm{sh}}) &= K\!\bigl(\Sigma, B_{t-1}, \{\tilde{a}_t^i,u_t^i\}_{i=1}^n \bigr), \label{eq:mas-loop-5}\\
  a_t^{i,\mathrm{exec}} &= E^i(a_t^i), && i=1,\dots,n, \label{eq:mas-loop-6}\\
  s_{t+1} &\sim T\!\bigl(s_t, \{a_t^{i,\mathrm{exec}}\}_{i=1}^n \bigr), \label{eq:mas-loop-7}\\
  B_t      &= \Omega\!\bigl(B_{t-1}, \mathbf{u}_t^{\mathrm{sh}}, \{M_t^i\}_{i=1}^n; \Sigma\bigr). \label{eq:mas-loop-8}
\end{align}

The loop then advances to $t{+}1$ and repeats.  Termination can be triggered by environmental conditions (task complete, time out), social-system rules (budget exhausted, human override), or internal goal satisfaction (group goal reached).

\paragraph*{Reduction to the Single Foundation Agent}
A good generalisation collapses cleanly.  Let $n=1$, drop the shared workspace (or set $B_{t}=\varnothing$), and let $\Sigma$ impose no additional rules.  Then Eqs.~\eqref{eq:mas-loop-1}–\eqref{eq:mas-loop-8} reduce to
\[
  o_t = P(s_t, M_{t-1}),
  \quad
  M_t = L(M_{t-1}, a_{t-1}, o_t),
  \quad
  a_t = R(M_t),
  \quad
  s_{t+1} = T(s_t, E(a_t)),
\]
which is exactly the Foundation Agent loop of Def.~\ref{def:foundation-agent-loop}.  Thus the multi-agent formalism is a conservative extension: nothing is lost, only shared structure and coordination are added.

\paragraph*{Why This Level of Formality Matters}
Writing out the loop may feel pedantic, but it buys us three things we need for the rest of the book:
\begin{enumerate}
\item \textbf{Comparability across systems.}  Whether you use AutoGen-style chatting LLMs, workflow graphs, or embodied robot swarms, you can map components to $P^i$, $L^i$, $R^i$, $K$, $T$, and $\Omega$ and compare apples to apples.
\item \textbf{Analytical hooks.} Stability of collaboration, failure propagation, communication load, and reward hacking under social constraints each becomes an object of study once the interfaces are explicit.
\item \textbf{Implementation discipline.}  Real code bases drift; prompts bleed into memory; ad-hoc globals creep in.  The formal surfaces in Eqs.~\eqref{eq:mas-loop-1}--\eqref{eq:mas-loop-8} force us to decide where each update belongs, which in turn simplifies logging, evaluation, and reproducibility.
\end{enumerate}

In short, the Foundation MAS formalism extends the perception--cognition--action loop from a single mind to a society of minds~\cite{minsky1988society}, all situated inside a structured world shaped from the inside out~\cite{Buzsaki2019} and capable (in principle) of Bayesian-style active engagement with uncertainty~\cite{Friston2011}.  We will draw on these perspectives repeatedly as we examine coordination, communication, learning at scale, and safety in the chapters ahead.

Next, we examine different categories of multi-agent systems in more details.

\section{Strategic Learning: Cooperation \textit{vs.} Competition}
\label{subsec:strategic-learning}

\begin{figure*}[!ht]
    \centering
    \includegraphics[width=0.8\textwidth]{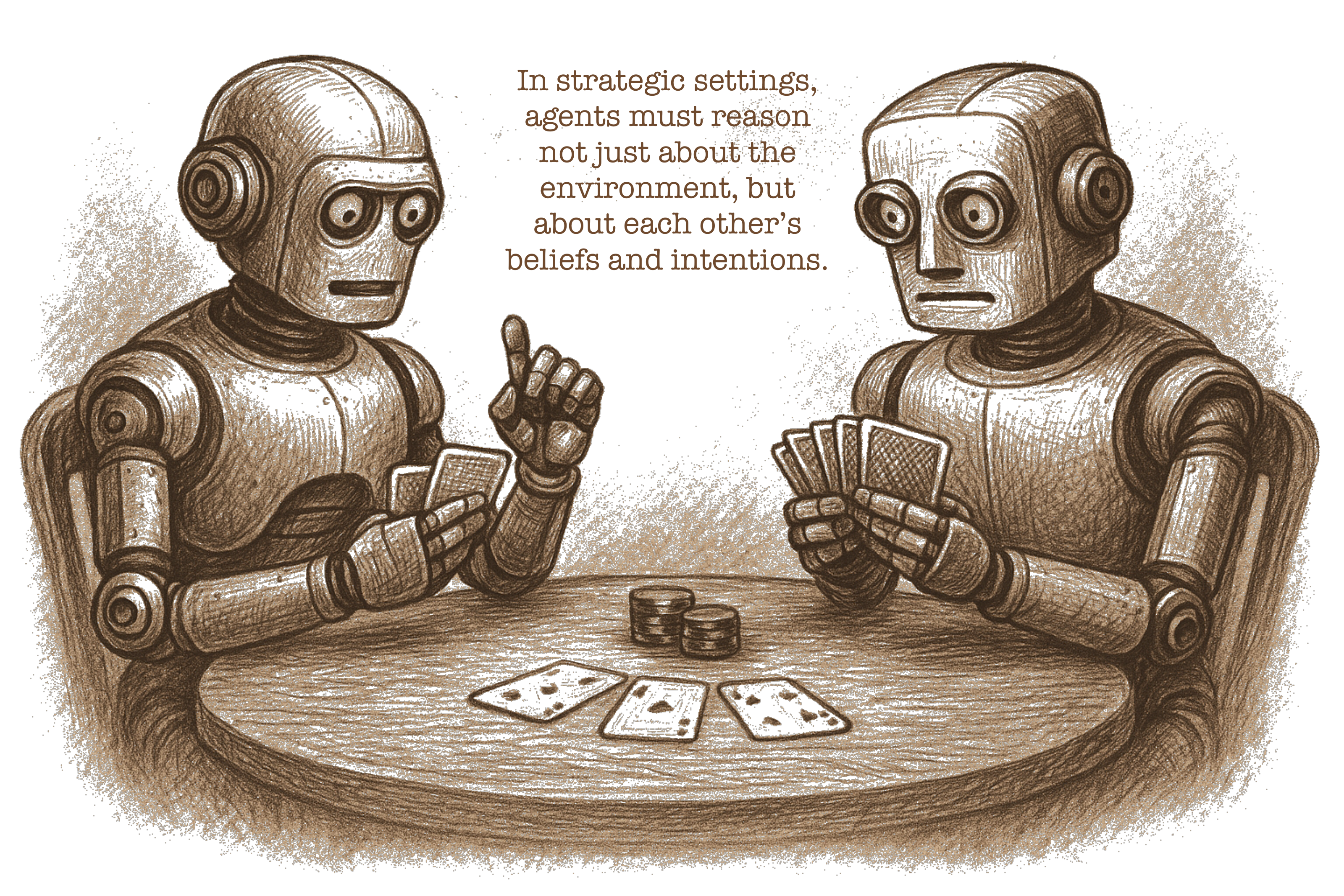}
    \caption{An LLM-based robot doctor interacts with a robot patient, illustrating the simulation of clinical workflows in healthcare environments. Such agent-based simulations enable iterative refinement of treatment strategies, protocol evaluation, and virtual experimentation in realistic medical settings.}
    \label{fig:MAS-strategy}
\end{figure*}

\lettrine[lines=3]{\initfamily\textcolor{darkgreen}{S}}{trategic learning} describes the capacity of agents to dynamically anticipate, interpret, and influence the actions of other agents within game-theoretic environments, whether those settings are competitive, cooperative, or a combination of both~\cite{zhang2024llm}. Agents refine their strategies continuously, adjusting based on newly acquired information.
This iterative process frequently leverages foundational game theory concepts such as Nash equilibria~\cite{silva2024large}, Bayesian games~\cite{horton2023large,li2024survey,gemp2024states}, and repeated interactions~\cite{mao2023alympics,akata2023playing}. With the sophisticated linguistic capabilities of LLMs, strategic learning increasingly incorporates nuanced interactions such as dialogue, persuasion, and implicit negotiation, enhancing traditional game-theoretic reasoning~\cite{gemp2024states,xu2024magicinvestigationlargelanguage,gandhi2023strategic,duan2024gtbench}.
Figure~\ref{fig:MAS-strategy} provides a visual metaphor for such settings, depicting two robots engaged in a game of Texas Hold'em, where bluffing, hidden information, and adaptive reasoning exemplify the essence of strategic learning.

Economic applications demonstrate the significant potential of multi-agent strategic simulations by offering valuable insights into market dynamics and negotiation strategies, reflecting both competitive and cooperative elements. For instance, research by \citet{li2023large} and \citet{horton2023large} illustrates how agents driven by LLMs can realistically simulate hiring scenarios, demonstrate rational decision-making in structured economic experiments, and even predict stock market trends. Additionally, \citet{zhao2023competeai} developed a competitive environment using GPT-4, highlighting strategic interactions between restaurant and customer agents aimed at optimizing profits and satisfaction through realistic pricing and bidding strategies. Similarly, \citet{xia2024measuring} explores LLM-driven Buyer–Seller bargaining scenarios, and \citet{Sreedhar2024SimulatingHS} employs ultimatum game simulations to inform policymaking by modeling human-like strategic behaviors.

Strategic learning extends far beyond conventional market contexts, appearing in any scenario involving resource allocation, alliance formation, or balancing competitive and cooperative incentives. Notably, studies on multi-commodity competitions~\cite{lin2024strategic,zhao2023competeai} demonstrate how agents strategically negotiate to maximize individual gains. In sustainability-focused applications, agents coordinate consumption of resources efficiently and responsibly~\cite{piatti2024cooperate}.

In the realm of gaming, social deduction games such as Werewolf, Chameleon, Avalon, and Jubensha exemplify how agents must navigate complex dynamics involving deception and collaboration~\cite{xu2023language,du2024helmsman,jin2024learning,Pan2023Chameleon,Lan2023LLMBasedAS,stepputtis2023long,wang2023avalon,shi2023cooperation,wu2023deciphering}. Research by \citet{xu2024exploring} and \citet{du2024helmsman} demonstrates LLM-based agents' effectiveness in subtly orchestrating deception and cooperation. Other studies emphasize agents' ability to adapt strategically over multiple rounds in Avalon~\cite{stepputtis2023long,light2023avalonbench,wang2023avalon,shi2023cooperation}. Moreover, \citet{wu2023deciphering} advances this domain further, illustrating autonomous agent interactions within the intricate narratives of the Jubensha murder mystery genre. Diplomatic simulations, such as those presented by \citet{hua2023war} and \citet{jin2024if}, utilize LLM-driven agents to realistically emulate complex geopolitical negotiations and alliance dynamics on a global scale.

The critical advantage of LLM-powered strategic learning lies in the seamless integration of rigorous game-theoretic logic with natural language reasoning. This integration empowers agents to interpret sophisticated instructions, participate in persuasive interactions, and adapt more effectively to novel or unstructured contexts. As a result, LLM-driven strategic agents offer unparalleled potential for accurately modeling complex real-world interactions, ranging from economic competition and social negotiation to diplomatic strategy, significantly surpassing traditional rule-based or purely numeric approaches.

\section{Modeling Real-World Dynamics}
\label{subsec:modeling-simulation}

\begin{figure*}[!ht]
    \centering
    \includegraphics[width=0.6\textwidth]{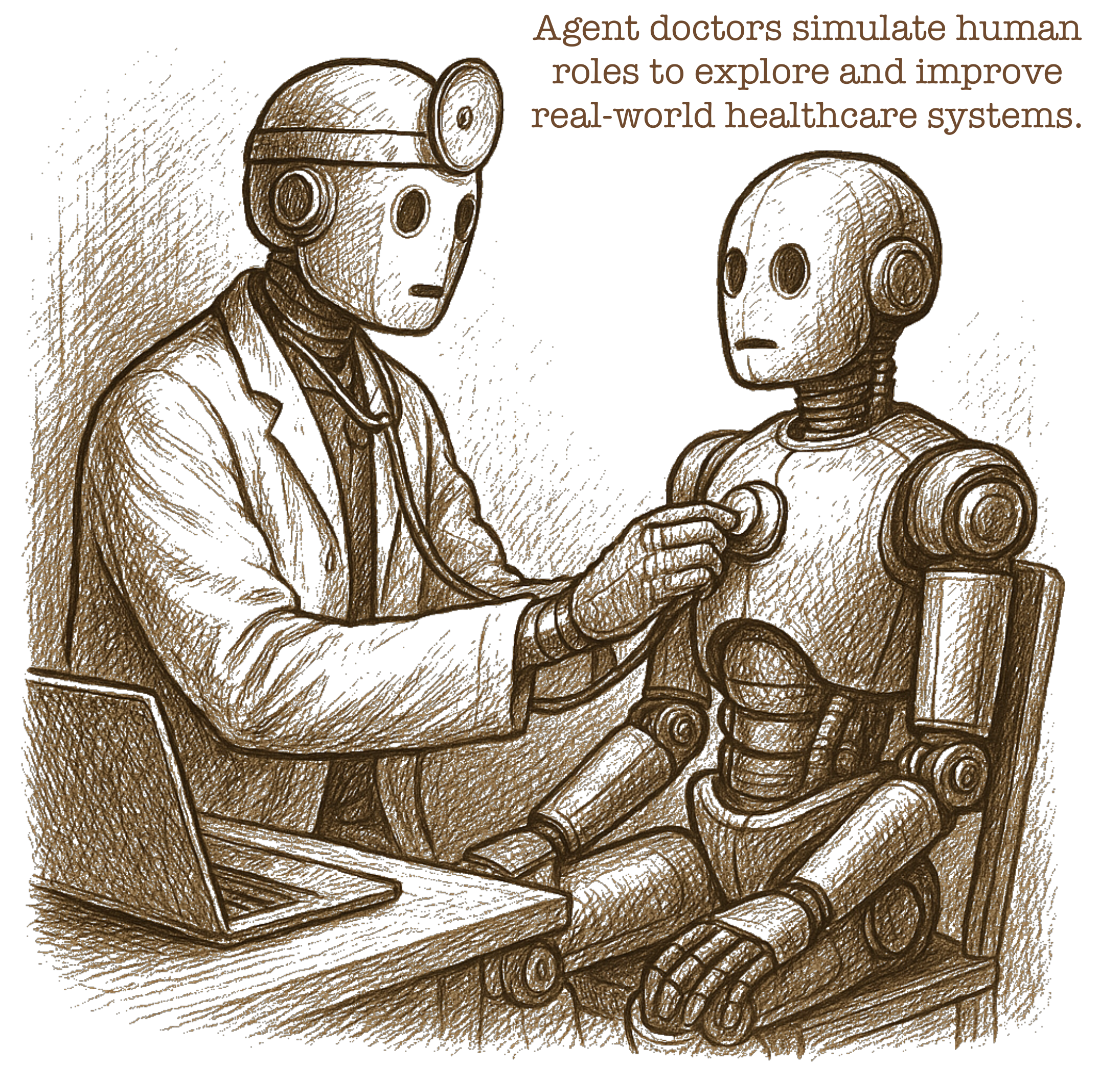}
    \caption{An LLM-based robot doctor interacts with a robot patient, illustrating the simulation of clinical workflows in healthcare environments. Such agent-based simulations enable iterative refinement of treatment strategies, protocol evaluation, and virtual experimentation in realistic medical settings.}
    \label{fig:MAS-role}
\end{figure*}

\lettrine[lines=3]{\initfamily\textcolor{darkgreen}{M}}{odeling and simulation} is a foundational application area for LLM-based multi-agent systems (LLM-MAS), aimed at capturing and reproducing complex social, economic, and political phenomena at scale. Leveraging the contextual reasoning and language understanding capabilities of LLMs, these systems enable the design of diverse, role-driven agents whose evolving behaviors reflect real-world dynamics with greater fidelity.

Unlike strategic learning setups that emphasize explicit goals of cooperation or competition, simulation environments typically involve agents acting autonomously according to their individual roles, preferences, and local contexts~\cite{gao2024large}. This makes LLM-MAS a powerful tool for exploring emergent dynamics and evaluating potential interventions in open-ended, domain-specific scenarios.

In healthcare, \citet{li2024agent} introduce \textit{Agent Hospital}, where LLM-powered doctor agents interact with simulated patients to iteratively refine treatment strategies. This virtual setting supports the evaluation of hospital management protocols, medical training paradigms, and ``what-if'' scenarios under realistic constraints. 
Figure~\ref{fig:MAS-role} captures this idea by depicting an LLM agent simulating a doctor's role in an interaction with a robot patient, demonstrating how agent-based systems can emulate clinical environments for experimentation and training.
In a parallel line of work, \citet{li2024econagent} present \textit{EconAgents}, a framework for simulating macroeconomic systems from the bottom up. Their LLM-driven agents model household-level behaviors such as employment choices, consumption patterns, and savings decisions, offering a more expressive and adaptive alternative to traditional numeric or rule-based economic models~\cite{tran2025multi}.

Political science has also begun to embrace LLM-based simulation. For example, \citet{zhang2024electionsim} and \citet{tran2025multi} simulate electoral processes and policy formation, capturing how candidate behavior, voter sentiment, and public discourse co-evolve to influence election outcomes. These simulations highlight the role of communication and strategy in shaping collective political behavior.

Beyond politics and economics, LLM-MAS has been used to study a range of social and cultural dynamics. For instance, \citet{ferraro2024agent} and \citet{gao2023s3} simulate the propagation of language and emotion through social networks to better understand how opinions and sentiment clusters emerge. \citet{chuang2023simulating} explore how opinion dynamics unfold under different interaction patterns and network topologies, while \citet{liu2024skepticism} investigate the diffusion of misinformation and the conditions under which it accelerates or stalls.

Large-scale platforms such as GenSim~\cite{tang2024gensim} and OASIS~\cite{yang2024oasis} further extend these capabilities by supporting simulations with tens of thousands or even millions of agents. These systems make it possible to study population-scale dynamics, such as echo chambers, group polarization, and viral content spread, under realistic behavioral and infrastructural constraints.

What sets LLM-based simulation apart is its ability to incorporate both structural features (like institutional rules or network connections) and cognitive-linguistic elements (such as framing, persuasion, and narrative). This dual modeling capacity allows researchers to examine complex phenomena that are difficult to capture using numeric abstractions alone, including cultural transmission, belief shifts, and discursive influence.

\section{Collaborative Task Solving with Workflow Generation}
\label{subsec:collaborative-task-solving}

\begin{figure*}[!ht]
    \centering
    \includegraphics[width=0.6\textwidth]{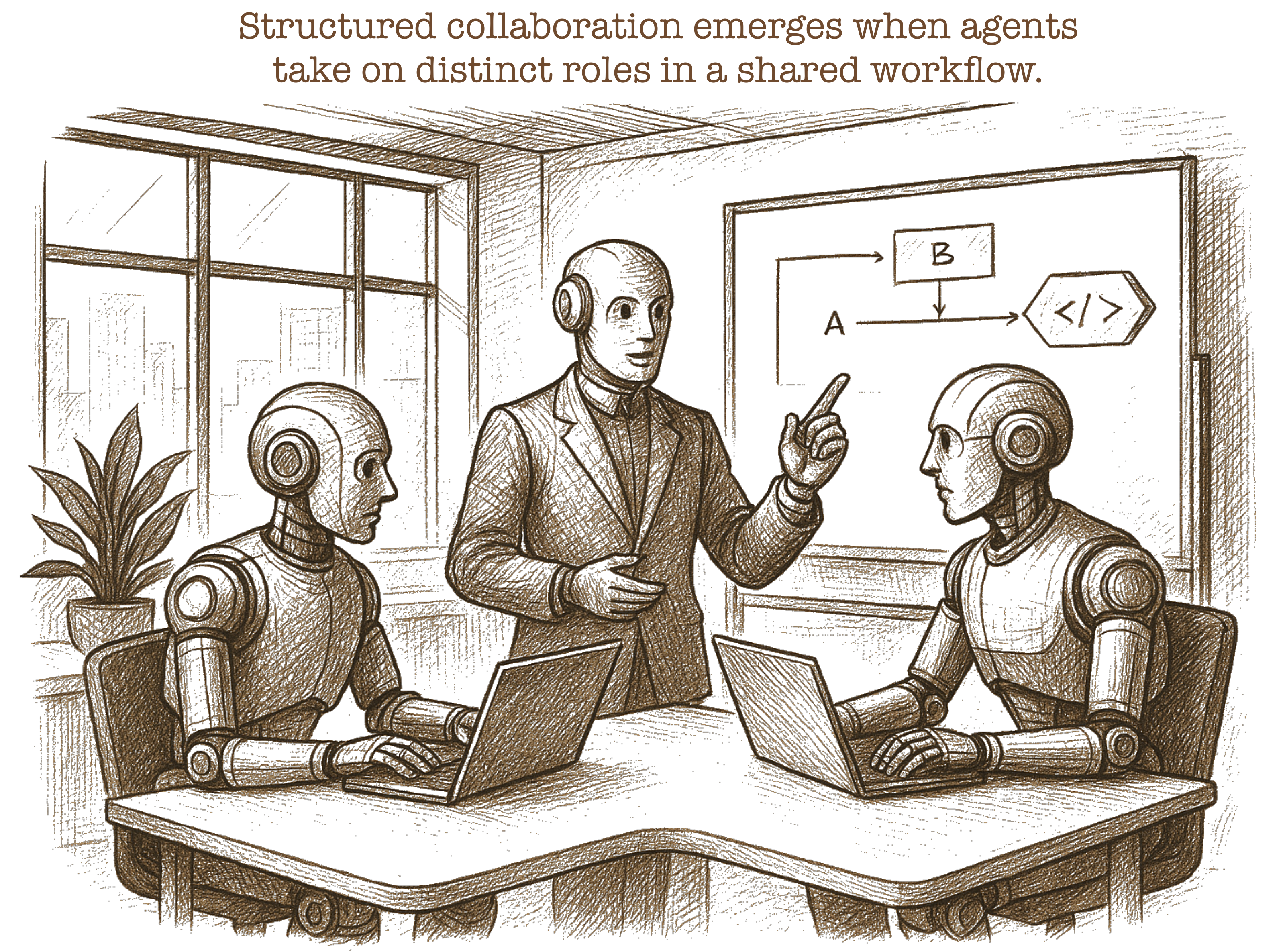}
    \caption{Three LLM-based agents, each assigned a distinct role (programmer, architect, and project manager), collaboratively solve a software development task through structured dialogue and workflow coordination. This illustrates the paradigm of collaborative task solving, where multi-agent systems function as role-specialized teams aligned on a shared objective.}
    \label{fig:MAS-software}
\end{figure*}

\lettrine[lines=3]{\initfamily\textcolor{darkgreen}{C}}{ollaborative task solving} brings multiple agents together to achieve a clearly defined objective through a structured, coordinated workflow. Unlike strategic learning, which often involves competition or negotiation, or modeling and simulation, where agents act independently, collaborative systems organize agents into cohesive teams that operate through sequential or parallel stages of problem-solving.

Agents are typically assigned explicit roles, such as ``Planner'', ``Implementer'', or ``Evaluator'', and work within stage-based frameworks to ensure that tasks are completed efficiently and accurately. These roles reduce ambiguity and facilitate coordination, especially when tackling complex, multi-step objectives.
An example of this collaborative structure is depicted in Figure~\ref{fig:MAS-software}, where agents act in distinct roles within a shared task environment.
Systems like MetaGPT~\cite{hong2023metagpt}, CAMEL~\cite{li2023camel}, Communicative Agents~\cite{qian2023communicative}, and the framework in \citet{zhang2023building} illustrate how clearly defined responsibilities and communication protocols enable LLM-based agents to function as cohesive teams. A typical collaborative workflow might begin with one agent analyzing the task, followed by another drafting a solution plan, a third implementing components, and a final agent verifying or refining outputs. These agents often communicate through natural language dialogues, using iterative rounds of message passing to align on goals and clarify uncertainties.
This design is especially valuable for large or ambitious projects, where sub-tasks can be distributed among domain-specialized agents. Language-based coordination enables agents to flexibly reinterpret instructions and delegate responsibilities without requiring rigid symbolic planning.

Although collaborative multi-agent systems have seen early success in software development, such as coding, debugging, and testing, their potential extends further. \textit{Agent Laboratory}~\cite{schmidgall2025agent}, for example, applies this paradigm to scientific discovery: agents propose hypotheses, design and simulate experiments, analyze results, and revise future inquiries. This mirrors the real-world research cycle and opens opportunities for AI-assisted exploration in literature review, policy design, and complex data analysis.

Compared to other LLM-based multi-agent paradigms, collaborative task solving emphasizes reliability and interpretability. With preassigned roles and well-defined workflows, these systems reduce unpredictability and emergent behavior, which is an advantage in domains requiring precision, compliance, or traceability. At the same time, ongoing research explores how to preserve creative flexibility, ensuring that agents can contribute novel insights while operating within a shared procedural structure.

Together, the three paradigms discussed in this chapter: \emph{strategic learning}, \emph{modeling and simulation}, and \emph{collaborative task solving}, illustrate the diversity and richness of LLM-based multi-agent systems. Each emphasizes different strengths of language-driven intelligence, enabling agents to reason, communicate, and coordinate in ways that extend well beyond traditional rule-based systems.

\section{Summary and Discussion}
\label{sec:mas-definition-summary-discussion}

\lettrine[lines=3]{\initfamily\textcolor{darkgreen}{T}}{his} chapter introduced a comprehensive framework and categorization for LLM-based multi-agent systems (LLM-MAS), emphasizing how different combinations of collaboration goals and norms fundamentally shape agent interactions and system behavior. Through our formalized Foundation Multi-Agent System framework, we illustrated the transition from individual Foundation Agents, capable of perception, cognition, and action loops, to sophisticated multi-agent systems or agent societies regulated by shared workspaces, coordination mechanisms, and overarching social systems.

We categorized LLM-MAS into three distinct yet complementary paradigms: \emph{strategic learning}, \emph{modeling and simulation}, and \emph{collaborative task solving}. Each category highlights unique facets of agent interaction and system objectives:
\begin{itemize}
\item \textbf{Strategic Learning} emphasized dynamic, game-theoretic interactions. Here, agents leverage sophisticated LLM-based reasoning capabilities to anticipate, interpret, and influence each other's actions. These systems often capture nuanced competitive and cooperative dynamics, proving particularly valuable in economics, social deduction games, and diplomacy simulations. The fusion of game-theoretic rigor with linguistic reasoning offers substantial advantages for realistically modeling complex human-like behaviors.

\item \textbf{Modeling and Simulation} showcased scenarios where agents act independently within rich, dynamic environments. This paradigm exploits LLMs' ability to embody diverse roles, behaviors, and decision-making styles, enabling large-scale simulations of economic, political, and social phenomena. By accurately reflecting cognitive, linguistic, and structural dynamics, such simulations offer significant insights into real-world processes like information diffusion, opinion formation, and societal behavior.

\item \textbf{Collaborative Task Solving} highlighted structured cooperation among agents toward explicitly defined objectives. Utilizing structured workflows and clear role definitions, this category ensures efficient, precise, and predictable outcomes. Its strength lies in its applicability to structured tasks like software development, scientific investigation, literature review, and policy formulation, providing systematic yet flexible cooperation among agents.
\end{itemize}

\begin{table}[!t]
\centering
\caption{
Classification framework for LLM-based multi-agent systems, highlighting different aspects of system design, communication and collaboration. 
Below are our abbreviations, for ease of reference:\\
M\&S = Modeling \& Simulation,\,
CTS = Collaborative Task Solving,\,
SL = Strategic Learning,\,
S-D = Static-Decentralized,\,
S-L = Static-Layered,\,
Hom = Homogeneous,\,
Het = Heterogeneous,\,
T/M = Teaching/Mentor,\,
C-O = Consensus-Oriented,\,
T-O = Task-Oriented,\,
CL = Collaborative Learning,\,
Dict = Dictatorial,\,
D-B = Debate-Based.
}
\label{tab:mas-classification-framework}
\small

\renewcommand{\arraystretch}{1.1}
\resizebox{\textwidth}{!}{

\begin{tabular}{l|c|c|c|c|c|c}
\hline
\textbf{Paper} & \textbf{\centering System Design} & 
\multicolumn{3}{c}{\textbf{Communication}} & 
\multicolumn{2}{|c}{\textbf{Collaboration}} \\
\hline
& \textbf{\centering Category} & \textbf{Typology} & \textbf{Interface} & \textbf{Agent Type} & \textbf{Interaction} & \textbf{Decision}  \\
\hline
Agent Hospital~\cite{li2024agent} & M\&S & S-D & Text & Het & T/M, C-O & Dict\\
\hline
Welfare Diplomacy~\cite{Mukobi2023WelfareDB} & M\&S & S-L & Code, JSON, Text & Hom & CL & Voting \\
\hline
MEDCO\cite{wei2024medco} & M\&S & S-L & Text & Het & T/M, C-O & Dict  \\
\hline
MedAgents\cite{tang2024medagents} & M\&S & S-L & Text & Hom & T-O & Dict  \\
\hline
Generative Agents~\cite{park2023generative} & M\&S & S-D & Visual & Hom & CL & Dict \\
\hline
RECONCILE~\cite{chen2023reconcile} & SL & S-D & Text & Hom & CL & D-B  \\
\hline
Agent Laboratory~\cite{schmidgall2025agent} & CTS & S-L & Code, Text & Het & C-O, T-O & Dict \\
\hline
CoELA\cite{zhang2023building} & CTS & S-D & Text & Hom & T-O &  --\\
\hline
The virtual lab~\cite{swanson2024virtual} & CTS & S-L & Text & Het & C-O, CL & Dict  \\
\hline
SciAgents~\cite{ghafarollahi2024SciAgeAutoma} & CTS & S-L & Text & Het & T-O & Dict \\
\hline
S-Agents~\cite{chen2024s} & CTS & S-D & Text & Het & T-O, CL & Dict  \\
\hline
GPT-Bargaining~\cite{fu2023improving} & CTS & S-D & Text & Het & C-O & D-B  \\
\hline
FORD~\cite{xiong2023examining} & M\&S & S-D & Text & Het & C-O & D-B  \\
\hline
MADRA~\cite{wang2023apollo} & CTS & S-D & Text & Het & C-O & D-B \\
\hline
Multiagent Bench~\cite{zhu2025multiagentbenchevaluatingcollaborationcompetition} & CTS & S-D & Text & Hom & T-O, CL & D-B \\ 
\hline
OASIS~\cite{yang2024oasis} & M\&S & D & Text & Het & C-O & --\\
\hline
S$^3$~\cite{gao2023s3} & M\&S & S-D & Text & Het & C-O & --\\
\hline
FPS~\cite{liu2024skepticism} & M\&S & S-D & Text & Het & C-O & --\\
\hline
GPTSwarm~\cite{zhuge2024language} & CTS & D & Code, JSON, Text & Hom & T-O &Dict \\
\hline
ChatEval~\cite{chan2023chatevalbetterllmbasedevaluators} & CTS & D & Text & Hom  & T-O & Voting  \\ \hline
MetaGPT~\cite{hong2023metagpt} & CTS &  S-L  & Code, JSON, Text, Visual & Het  & T-O & Dict \\
\hline
AutoAgents~\cite{chen2024autoagentsframeworkautomaticagent} & CTS  & D & Text & Het  & T-O & C-O \\ \hline
AgentCoder~\cite{huang2024agentcoder} & CTS & D & Code, Text & Het  & T-O & D-B \\ \hline
MASTER~\cite{gan2025mastermultiagentllmspecialized} &  CTS &  S-L  & Text & Hom  & T-O & D-B  \\ \hline
Reflexion~\cite{Shinn2023ReflexionLA} & CTS  & D  & Text & Het   &  T-O & D-B  \\ \hline
MACM~\cite{lei2024macmutilizingmultiagentcondition} & CTS  & D & Text, Code & Het  & T-O  &  D-B  \\ \hline
Debate~\cite{du2023improving} & CTS  & S-D & Text & Het & C-O & D-B  \\
\hline
\end{tabular}
}
\end{table}

Despite their differences, these categories are interconnected, collectively demonstrating the versatility of LLM-based multi-agent frameworks. They illustrate how varying degrees of goal alignment, interaction norms, and structural rigidity influence the complexity and nature of agent interactions, ranging from emergent behaviors to meticulously orchestrated workflows.

Table~\ref{tab:mas-classification-framework} compares a selected set of existing research from different perspectives. While we have introduced different categories for MAS, there are more topics to be discussed in detail. In the next chapter, we will delve into the construction of multi-agent systems in greater detail. 

\chapter{Designing Collaborative Multi-Agent Systems}
\label{ch:construct-mas}



\lettrine[lines=3]{\initfamily\textcolor{darkgreen}{M}}{ulti-Agent Systems (MAS)} leveraging large language models (LLMs) enable agents to autonomously collaborate, communicate, and solve complex tasks, offering significant advantages across diverse domains. Effective MAS design involves critical considerations in agent composition, interaction patterns, topological structures, decision-making strategies, and communication protocols. This chapter systematically examines key principles underlying collaborative MAS. We begin by discussing strategies for composing homogeneous and heterogeneous agent teams, emphasizing the role of diversity in enhancing system capabilities. Next, we analyze various static and dynamic network topologies, highlighting their scalability implications and practical trade-offs. Following this, we explore four prominent agent collaboration paradigms: consensus-oriented interactions, collaborative learning, teaching and mentoring, and task-oriented interactions, illustrating how distinct interaction models shape agent behaviors and system effectiveness. We then examine methods for human-agent collaboration, delineating different levels of human involvement ranging from simple task delegation to immersive collaborative partnerships. Furthermore, we address decision-making mechanisms, contrasting centralized and collective approaches to highlight their respective advantages and limitations. Lastly, we provide an overview of contemporary and emerging agent communication protocols, identifying critical attributes essential for interoperability, flexibility, and scalability. By synthesizing these elements, this chapter offers comprehensive insights for designing robust, efficient, and adaptable MAS frameworks capable of effectively addressing real-world challenges.

\section{Building AI Agent Teams}
\label{sec:build-ai-agent-teams}

\begin{figure}[!htbp]
\centering
\includegraphics[width=0.8\textwidth]{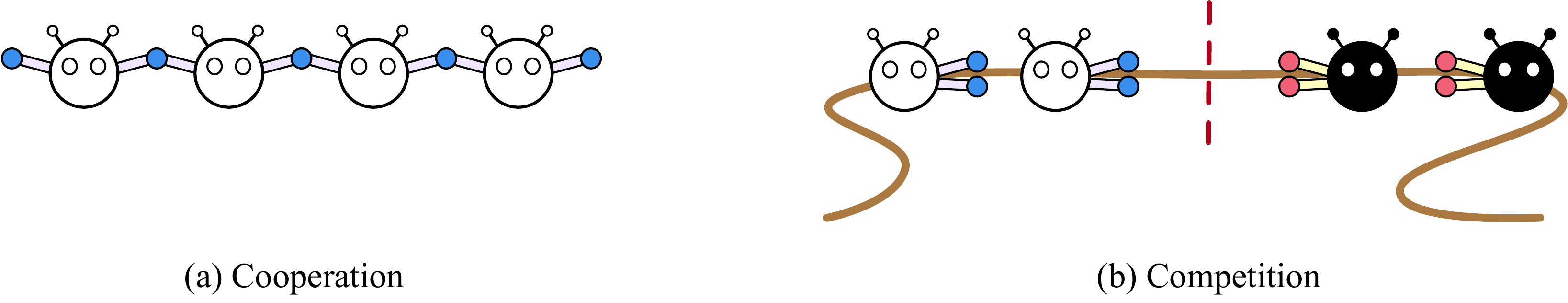}
\caption{Illustration of collaborative and competitive multi-agent dynamics.}
\label{fig:MAS-collab-compet}
\end{figure}

\begin{figure}[!htbp]
\centering
\includegraphics[width=\textwidth]{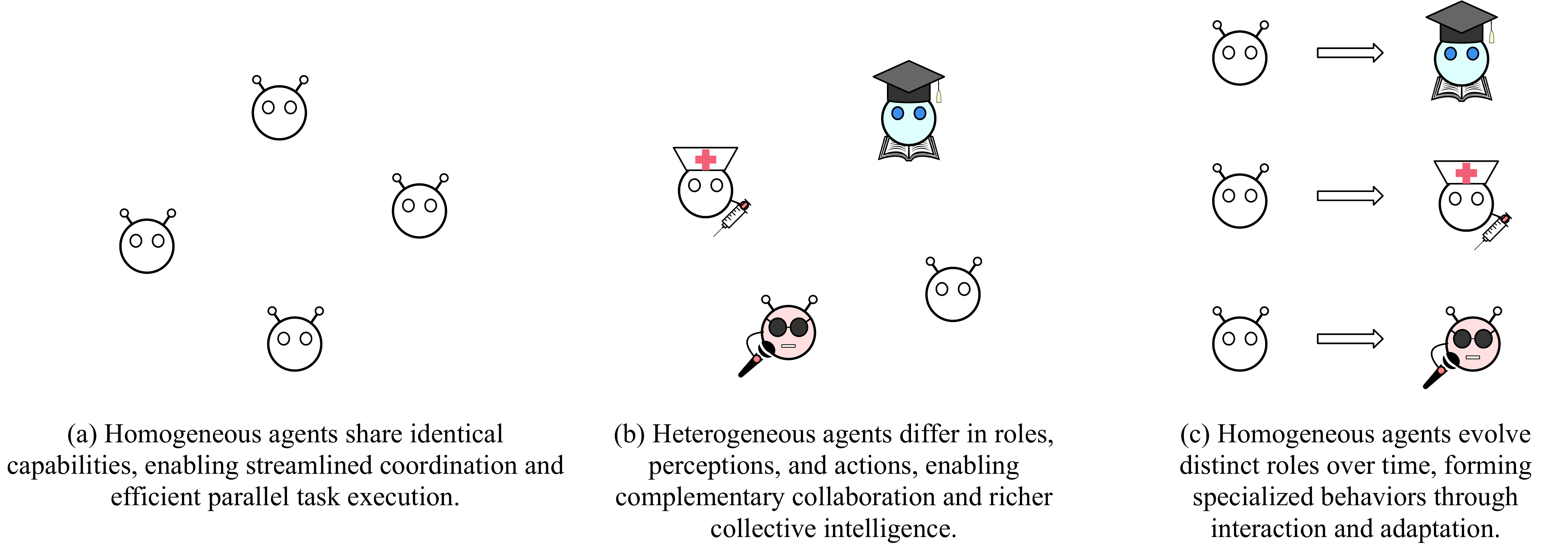}
\caption{Three types of multi-agent teams: \emph{Homogeneous agents} share identical capabilities and collaborate through uniform behavior; \emph{Heterogeneous agents} bring complementary roles, perceptions, and actions to address complex tasks; and \emph{Emergent specialization} arises when identical agents evolve distinct roles through repeated interaction and adaptation.}
\label{fig:agent-team-types}
\end{figure}

\lettrine[lines=3]{\initfamily\textcolor{darkgreen}{I}}{n} multi-agent systems (MAS), agents constitute the fundamental units that collaboratively or competitively interact within a shared environment to achieve specific objectives, as shown in Figure~\ref{fig:MAS-collab-compet}. The effective composition of agent teams significantly influences the system's overall performance, adaptability, and robustness. Designing an optimal agent team involves careful consideration of the homogeneity or heterogeneity of its members in terms of their capabilities, roles, perceptions, and action spaces. \emph{Homogeneous agent teams}, comprising agents with identical capabilities, can efficiently tackle tasks through straightforward parallelization and simplified coordination. Conversely, \emph{heterogeneous agent teams}, characterized by diverse capabilities and perspectives, excel in handling complex, multidisciplinary tasks by capitalizing on complementary strengths and enabling richer strategic interactions. Moreover, agent teams may dynamically evolve, exhibiting \emph{emergent specializations} through repeated interactions, thereby progressively enhancing the team's versatility and adaptability. Figure~\ref{fig:agent-team-types} visually summarizes these three fundamental team structures, illustrating their defining characteristics and differences. This section explores these fundamental approaches to building AI agent teams, detailing their respective strengths, limitations, and practical implications.

\subsection{Homogeneous Agents}
\label{subsec:homo-agents}

Homogeneous agents share identical capabilities, observation spaces, and action spaces, enabling streamlined coordination and efficient parallel task execution. Compared to single-agent systems, homogeneous MAS primarily offer the advantage of task parallelization, where multiple agents simultaneously address different aspects of a common goal. This straightforward design is particularly beneficial in structured environments or tasks that require uniform agent behaviors.

Recent research has effectively applied homogeneous agents in simulated teamwork scenarios, such as collaborative gameplay and household task execution. For instance, \citet{zhang2023building} introduced a cognitive-inspired modular framework enabling large language model (LLM)-based agents to engage in natural language interactions for task coordination. Their system demonstrated efficient labor division, with agents seamlessly assisting each other in object transportation tasks. Similarly, \citet{guo2024embodied} presented a prompt-based organizational approach, significantly reducing communication overhead and improving team efficiency during routine household activities, including preparing meals and cleaning.

Moreover, homogeneous agents have proven effective in interactive game environments, such as Overcooked and Minecraft. Studies by \citet{agashe2024llmcoordination} and \citet{dong2024villageragentgraphbasedmultiagentframework} leveraged teams of LLM-based homogeneous agents to successfully complete complex cooperative challenges within these games. These results underscore the strengths of homogeneous teams in predictable scenarios, benefiting from uniform behaviors and straightforward coordination strategies.

\subsection{Heterogeneous Agents}
\label{subsec:hetero-agents}

In contrast, heterogeneous agents exhibit diversity in their roles, perceptions, and available actions, thus significantly enriching the collective problem-solving capability of MAS. Such diversity contributes to more comprehensive decision-making and innovative strategies, as agents leverage complementary skills, perspectives, and information sources to address complex, real-world challenges \cite{du2023improving,chen2024persona}. Heterogeneity among agents is generally observed across three main dimensions: persona-level, observation-space, and action-space heterogeneity. These dimensions often coexist, further enhancing agent-team flexibility and effectiveness.

\paragraph*{Persona-level Heterogeneity.}
Persona-level heterogeneity refers to the diversity in agent profiles or roles, which shapes the way individual agents approach problem-solving and interact within the team. This form of heterogeneity is widely utilized in contemporary LLM-based multi-agent frameworks \cite{sun2024lawluo,qian2024chatdev,park2023generative,wu2023deciphering}. For example, within software development teams, distinct personas like programmers, product managers, and testers each contribute unique expertise and insights. Similarly, in medical diagnostic contexts, specialized agents representing cardiologists, oncologists, and pediatricians collaborate, leveraging their distinct expertise to enhance diagnostic accuracy and patient care quality.

Notably, even when agents share common action spaces, such as producing documents or providing recommendations, their distinct personas critically influence task outcomes. As demonstrated by \citet{hong2023metagpt}, a product manager persona might generate detailed requirement specifications, whereas a programmer persona would focus on software implementation. These persona-driven roles streamline interactions, leading to more robust and innovative collaborative outcomes, especially beneficial in multidisciplinary contexts.

\paragraph*{Observation-space Heterogeneity.}
Observation-space heterogeneity highlights variations in the perceptual capabilities of agents, directly influencing their knowledge about the environment and other agents. Different agents may possess distinct views or limited access to information based on their assigned roles. This phenomenon frequently appears in interactive games like Werewolf and Avalon \cite{xu2024exploring,Lan2023LLMBasedAS,light2023avalonbench}. In Werewolf, for instance, certain roles, such as werewolves and seers, have privileged observational abilities, enabling strategic information usage to deceive or collaborate effectively, while other roles like villagers must rely on limited public information. Similarly, in Avalon, role-specific perception significantly impacts communication and tactical decision-making, as agents strategically navigate varying degrees of information access.

In MAS applications, recognizing and effectively managing observation-space heterogeneity allows for nuanced communication strategies, efficient information dissemination, and improved coordination. Such deliberate utilization of differing observational abilities significantly enhances the complexity and realism of agent interactions, particularly in scenarios involving strategic deception, trust-building, and coalition formation.

\paragraph*{Action-space Heterogeneity.}
Action-space heterogeneity reflects fundamental differences in the actions agents can perform, driven by their design constraints or functional specializations. This type of diversity is particularly prominent in scenarios involving physical or virtual constraints, affecting an agent's practical capabilities. For instance, in virtual game environments like Werewolf and Avalon, distinct roles possess unique abilities or actions, such as secret communication channels or specialized skills, that necessitate coordinated strategic decisions \cite{du2024helmsman,xu2024exploring,jin2024learning,stepputtis2023long}.

Furthermore, robotic MAS scenarios exemplify action-space heterogeneity vividly. As described by \citet{yu2024mhrc}, robots within the same environment may possess divergent physical capabilities, such as mobility versus object manipulation. Effective team collaboration in these contexts relies on carefully aligning task assignments with individual agent strengths, promoting efficient task decomposition and specialized contributions. Thus, thoughtfully leveraging action-space heterogeneity significantly enhances overall task efficiency and adaptability in MAS.

\subsection{Emergent Agent Specialization}
\label{subsec:emergent-specialization}

Emergent specialization describes the phenomenon wherein initially homogeneous agents spontaneously develop distinct roles and specialized behaviors through repeated interactions and adaptation. In MAS incorporating LLMs, agent behaviors evolve organically due to inherent randomness, environmental feedback, and continuous adaptation cycles. This emergent differentiation leads to the gradual formation of specialized roles, significantly enriching the system's adaptive capabilities.

For instance, in an environment-driven MAS study by \citet{altera2024project}, agents initially endowed with identical action spaces and personas naturally differentiated into distinct specializations, as some agents gravitated towards resource gathering while others focused on crafting or defensive roles. Similarly, \citet{takata2024spontaneous} observed the spontaneous emergence of distinct communication styles, emotional expressions, and behavioral roles in initially homogeneous agent groups. These findings illustrate the transformative potential of MAS, transitioning from initially uniform teams toward increasingly heterogeneous systems characterized by organically evolved specializations. Such dynamic specialization underscores MAS's inherent flexibility and capacity for adaptation, enabling systems to respond effectively to evolving task demands and environmental complexities.

\section{Multi-Agent System Topologies}
\label{sec:system_topology}

\lettrine[lines=3]{\initfamily\textcolor{darkgreen}{T}}{his} section explores how the topology of interactions within LLM-based Multi-Agent Systems (MAS) influences their communication, collaboration efficiency, and task execution capabilities. We start by analyzing static topologies, characterized by fixed and predefined connection patterns, and subsequently delve into dynamic (adaptive) topologies, which can adjust inter-agent connections based on real-time factors such as performance feedback, workload variations, and strategic requirements. Finally, we discuss critical considerations regarding scalability, performance trade-offs, and robustness within MAS, drawing insights from recent advancements in distributed processing, self-organizing systems, and emergent collaborative behaviors.

\subsection{Static Topologies}
\label{subsec:static_topologies}

\begin{figure}[!htbp]
\centering
\includegraphics[width=0.9\textwidth]{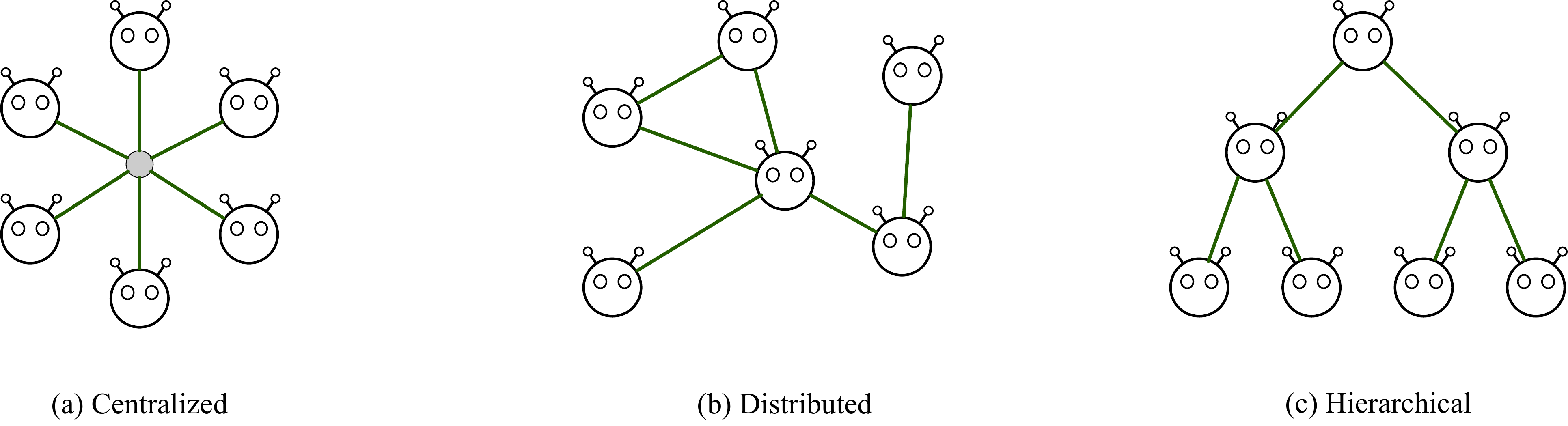}
\caption{Different types of topological structures for multi-agent collaboration.}
\label{fig:multi-agent-topology}
\end{figure}

Static topologies utilize predetermined patterns of connectivity, remaining unchanged during runtime. In such configurations, the relationships and communication paths among agents, or between agents and a central coordinator, are established through explicit domain knowledge, rules, or heuristics. This rigidity ensures predictable information flow, simplifying coordination efforts. Three primary static configurations are typically recognized: \emph{centralized}, \emph{decentralized}, and \emph{hierarchical} structures. Figure~\ref{fig:multi-agent-topology} illustrates these structures intuitively.

\paragraph*{Centralized Structures}
In centralized structures, a singular coordinator agent collects information and orchestrates peripheral agent activities. This topology facilitates comprehensive resource management and offers a unified global perspective, exemplified by cultural simulations and collaborative scenarios outlined by \citet{li2024cultureparkboostingcrossculturalunderstanding} and \citet{kaiya2023lyfeagentsgenerativeagents}. However, centralized approaches frequently encounter scalability constraints as additional agents are introduced, leading to increased communication overhead and heightened risk of single-point failures. Research by \citet{xu2024exploring} and \citet{handler2023balancingautonomyalignmentmultidimensional} has further identified inherent trade-offs, highlighting that while centralized structures provide consistency, their rigidity limits adaptive responsiveness.

\paragraph*{Decentralized Structures}
Decentralized structures involve peer-to-peer interactions among agents without reliance on a central controlling node, forming networks such as chains, rings, small-world configurations, or random graphs \cite{bailis2024werewolf, xu2024exploring}. Such designs inherently improve fault tolerance, as the failure of a single agent is less likely to disrupt the entire network. For instance, \citet{hu2024scalableaccurategraphreasoning} demonstrate how distributing graph reasoning tasks across multiple decentralized agents can effectively circumvent the context-length limitations typical in large language models. Additionally, \citet{rasal2024navigatingcomplexityorchestratedproblem} propose decomposition techniques enabling orchestrating LLMs to efficiently delegate subtasks. Despite these strengths, decentralized topologies must implement sophisticated consensus and synchronization protocols to maintain global coherence, which can introduce additional complexity and overhead.

\paragraph*{Hierarchical Structures}
Hierarchical structures organize agents into layered arrangements, where higher-level agents manage and oversee lower-level ones, akin to classical organizational management models such as Standard Operating Procedures (SOPs) or Waterfall project methodologies. For example, \citet{chen2024autoagentsframeworkautomaticagent} describe the AutoAgents framework, assigning clear roles such as Planners, Agent Observers, and Plan Observers to systematically produce coherent execution plans. Similarly, the ChatDev framework \cite{qian2023communicative} utilizes hierarchical decomposition to effectively streamline complex software development processes \cite{hong2023metagpt, li2024agent, qian2024chatdev}. While hierarchical designs offer clear advantages, such as enhanced traceability, simplified debugging, and structured performance monitoring, they may also lead to bottlenecks if high-level agents become overloaded \cite{du2024multi}. Recent studies in diverse fields including narrative generation \cite{li2024coinsight, ye2024generative, wu2024chartinsights}, data science tasks such as data cleaning \cite{zhang2024sketchfill,chai2023demystifying}, visualization \cite{shen2024data, xie2024haichart}, and automated machine learning \cite{trirat2024automl,li2024autokagglemultiagentframeworkautonomous} highlight an intrinsic tension between maintaining consistency and adapting to real-time challenges in hierarchical arrangements.

Overall, static topologies offer predictability, ease of implementation, and straightforward maintenance due to their clearly defined communication structures and predefined roles. Such designs are ideally suited for stable environments with fixed task workflows and clearly established requirements. Nonetheless, their primary drawback lies in the inability to dynamically adapt to unforeseen circumstances, such as agent failures, evolving task complexities, or shifting system objectives. This inflexibility often results in reduced effectiveness in scenarios characterized by emergent dynamics, necessitating more adaptive topological solutions.

\subsection{Dynamic and Adaptive Topologies}
\label{subsec:dynamic_topologies}

Static topologies offer stability and predictability, making them suitable for tasks with clearly defined roles and consistent workflows. However, their rigidity limits their effectiveness in open-ended, dynamic, or novel scenarios. Real-world domains, including collaborative planning and social simulations, frequently necessitate adaptability as resources fluctuate, feedback emerges, or objectives shift during execution. Dynamic and adaptive topologies address these demands by continuously restructuring inter-agent connections in response to real-time metrics, environmental feedback, or strategic objectives, achieving an optimal balance between reliability and responsiveness.

Dynamic topology frameworks, such as DyLAN~\cite{liu2023dynamic}, exemplify this adaptability by selecting agents dynamically based on performance metrics. DyLAN employs a two-stage optimization: initially computing unsupervised Agent Importance Scores and subsequently reforming agent teams at runtime. Similarly, OPTIMA~\cite{chen2024optima} refines agent connections iteratively through a generate–rank–select–train approach, optimizing communication pathways to balance task quality, token efficiency, and readability via Direct Preference optimization. The MAD framework~\cite{li2024improving} further illustrates adaptability by dynamically assigning agent roles (e.g., verifiers, debaters) within a systematically pruned interaction structure.

Recent advances in computational approaches have enhanced dynamic topology management. GPTSwarm~\cite{zhuge2024gptswarm} conceptualizes multi-agent collaboration as computation graphs, applying evolutionary strategies and reinforcement learning to optimize connections based on real-time feedback. MACNET~\cite{qian2024scaling} employs a directed acyclic graph (DAG) architecture, where supervisory agents manage overall communication structure, and executive agents manage local tasks, thereby enabling adaptive communication through structured propagation of outputs. Application-specific systems like DAMCS~\cite{yang2025llm} combine hierarchical knowledge graphs with structured communication schemes, dynamically adapting agent coordination based on context. In creative applications, AutoAgents~\cite{chen2023autoagents} utilize a dynamic drafting-execution model, wherein predefined expert agents collaboratively refine task solutions through internal supervision and parallel processing.

Three primary methodological paradigms facilitate dynamic topology formation: search-based optimization, LLM-driven generative methods, and parameter-efficient external configurations.

\paragraph*{Search-Based Methods}
Several frameworks utilize search-based optimization to iteratively improve inter-agent communication structures. ADAS~\cite{hu2024automated} employs a Meta Agent Search algorithm to iteratively propose, evaluate, and archive superior agent configurations. Similarly, Aflow~\cite{zhang2024aflow} leverages Monte Carlo Tree Search (MCTS) by representing each LLM interaction as a graph node, dynamically refining workflow structures. Approaches like MAD~\cite{liang2023encouraging} and OPTIMA~\cite{chen2024optima} incorporate iterative selection and training mechanisms, drawing from search optimization principles to balance efficiency and task performance.

\paragraph*{LLM-Based Generative Methods}
Building upon search-based approaches, several studies exploit LLM generative capabilities to construct and continuously refine topologies. DyLAN~\cite{liu2023dynamic} introduces a temporal feed-forward network model, using propagation methods to dynamically adjust agent teams. Similarly, frameworks like DAMCS~\cite{yang2025llm}, AutoAgents~\cite{chen2023autoagents}, and TDAG~\cite{wang2025tdag} dynamically generate specialized agents or update hierarchical knowledge graphs, facilitating adaptive planning and task decomposition. Furthermore, AutoFlow~\cite{zhang2024aflow} and Flow~\cite{niu2025flow} represent workflows using natural language programs or activity vertex graphs, continuously optimizing via reinforcement learning. ScoreFlow~\cite{wang2025scoreflow} further refines agent workflows through gradient-based optimization, enabling nuanced adaptation to evolving task demands.

\paragraph*{External Parameter-Based Methods}
Recognizing the computational expense of directly tuning LLM-based agents, recent research promotes configuring agent interactions through lightweight external parameters. GPTSwarm~\cite{zhuge2024gptswarm} exemplifies this approach by training only the adjacency matrices in a DAG-based communication framework. Extending this concept, AgentPrune identifies redundant communication paths through magnitude-based pruning. Similarly, G-Safeguard~\cite{wang2025g-safeguard} employs graph neural networks (GNNs) external to the primary agent models, efficiently identifying and eliminating detrimental communication channels. Despite their efficiency, such approaches may face performance constraints due to limited coupling with LLM-agent internal mechanisms.

Dynamic topologies also significantly advance social simulation and embodied AI domains by modeling complex real-world interactions and behaviors. Systems like OASIS~\cite{yang2024oasis}, ProjectSid~\cite{altera2024project}, and generative memory frameworks~\cite{park2023generative} dynamically adapt social network structures based on evolving interactions. ProjectSid's PIANO architecture enables large-scale, real-time simulations, demonstrating emergent societal phenomena such as specialized roles and cultural transmission within simulated communities. Architectures such as AgentScope-scalability~\cite{gao2024agentscope} facilitate extensive multi-agent simulations, capturing emergent group dynamics and collective decision-making processes. In medical contexts, AI hospital~\cite{fan2024ai} and Agent hospital~\cite{li2024agent} dynamically reshape communication structures through iterative feedback cycles in clinical settings. Similarly, frameworks like IOA~\cite{chen2024internet} support dynamic team formation and adaptive task allocation across diverse embodied agents.

Despite their substantial progress, dynamic and adaptive multi-agent topologies face critical research challenges:

\begin{enumerate}
\item \textbf{Generalizability.} Current frameworks typically optimize agent interactions within single-task domains. Systems like AFlow~\cite{zhang2024aflow} struggle to generalize beyond their initial design context. Advancing lifelong learning capabilities, enabling seamless adaptation across multiple domains with minimal resource expenditure, represents an essential future direction.

\item \textbf{Resource Efficiency.} Existing dynamic systems often entail considerable computational costs. For instance, ADAS~\cite{hu2024automated} incurs substantial training expenses. Future research must develop more cost-effective runtime optimization techniques to facilitate widespread practical deployment.

\item \textbf{Inference Efficiency.} Highly complex multi-agent configurations frequently lack task adaptability, resulting in inefficient resource allocation during inference. While frameworks like MaAS~\cite{zhang2025multi} propose mechanisms for adaptive resource allocation, broader testing and refinement are necessary to ensure scalability and applicability in diverse real-world scenarios.

\end{enumerate}

Addressing these challenges will significantly enhance the practical utility, scalability, and robustness of dynamic multi-agent systems, paving the way for broader, more adaptive, and efficient real-world deployments.

\subsection{Scalability Considerations}
\label{subsec:scalability}

Scalability remains a pivotal challenge in LLM-based multi-agent systems (MAS), particularly as the number of agents increases. In fully connected networks, the number of communication pathways grows quadratically, leading to what is often termed a \emph{communication explosion}, which inflates both token usage and computational costs~\cite{zhao2024electoral, hong2023metagpt}. While centralized and layered topologies offer control and coordination, they often become synchronization bottlenecks under high message loads. On the other hand, decentralized networks enhance fault tolerance but demand complex consensus protocols to maintain a coherent global state.

Recent frameworks propose novel architectural strategies to address these bottlenecks. For instance, MACNET~\cite{qian2024scaling} structures agent collaboration as a directed acyclic graph (DAG), enabling scalable coordination among up to 1{,}000 agents without significant performance degradation. Similarly, Hu et al.~\cite{hu2024scalableaccurategraphreasoning} demonstrate that distributing graph reasoning tasks across multiple agents circumvents the limitations of input context length, thus unlocking broader scalability. Complementing these architectural insights, self-organizing agents~\cite{ishibashi2024self} demonstrate dynamic agent multiplication and adaptive task assignment, allowing each agent to maintain a constant workload while collectively increasing throughput.

Beyond theoretical advancements, scalable MAS frameworks increasingly depend on robust engineering practices. For example, AgentScope~\cite{gao2024agentscope} provides a developer-oriented platform that adopts an actor-based distributed design. It supports seamless switching between local and distributed modes, incorporates automatic parallel optimization, and features fault-tolerant message routing with intelligent filtering. These capabilities collectively mitigate the synchronization burdens and communication overheads typical of large-scale deployments.

Project Sid~\cite{altera2024project} takes a complementary route, focusing on simulating complex societal dynamics. Its PIANO (Parallel Information Aggregation via Neural Orchestration) architecture decouples slower cognitive modules from fast reactive layers, allowing agents to operate concurrently. A centralized cognitive controller ensures coherence across parallel streams of output. This architectural design enables high-frequency interactions among over 1{,}000 agents, while retaining control over coordination complexity.

At an even larger scale, AgentSociety~\cite{piao2025agentsocietylargescalesimulationllmdriven} demonstrates how LLM-based generative agents can be embedded within realistic social and economic simulations comprising up to 10{,}000 agents. This system employs high-performance messaging infrastructure (e.g., MQTT) and distributed compute to support millions of daily interactions. By modeling complex macro-level phenomena, such as market fluctuations, policy effects, and urban planning, it illustrates how scalability can enable richer simulations of emergent group behaviors.

Nevertheless, the benefits of scaling agent populations must be evaluated carefully. While increasing the number of agents can expand the system's overall cognitive and computational capacity, it also introduces diminishing returns due to escalating coordination costs. Beyond a certain threshold, added agents may contribute less to task-solving efficacy and more to synchronization burden, leading to a plateau or decline in performance. This scaling paradox stems from the super-linear growth of coordination overhead relative to the linear increase in individual task contributions.

Importantly, the value of scalability varies significantly by domain. In focused task-solving environments (e.g., coding, data analysis, or question answering), there likely exists an optimal population size beyond which performance gains stall. In contrast, simulation-driven domains, such as modeling collective decision-making, cultural evolution, or economic systems, demand high agent cardinality. Here, fidelity to real-world population-level behaviors often depends on emergent interactions among thousands of agents.

To address this duality, hybrid architectures that combine centralized oversight with decentralized execution are increasingly favored~\cite{li2024agent, chen2023reconcile}. Supervisory agents oversee global objectives and assign subtasks, while local agents operate semi-independently within dynamic sub-teams. This organization reduces information bottlenecks and adapts team sizes to task complexity, improving overall system responsiveness. In parallel, advanced techniques such as graph-based search, reinforcement learning, and evolutionary topology updates support continual adaptation of network structure. Intelligent communication techniques, including message prioritization, summarization, and selective broadcast, further mitigate overhead without sacrificing effectiveness. Moreover, asynchronous protocols and partial knowledge sharing schemes offer promising directions for reducing latency and enhancing robustness in massively distributed MAS deployments.

\section{Collaboration Paradigms and Interactions}
\label{sec:mas-collab-paradigms}

\lettrine[lines=3]{\initfamily\textcolor{darkgreen}{B}}{uilding} on the structural foundations laid out in the previous section on \emph{Multi-Agent System Topologies}, we now shift our focus from how agents are connected to how they \emph{collaborate}. While topology defines the architectural skeleton of MAS, it is the interaction dynamics that breathe life into the system, shaping how agents negotiate, reason, learn, and cooperate within these structures.

\begin{figure}[!htb]
    \centering
    \includegraphics[width=\linewidth]{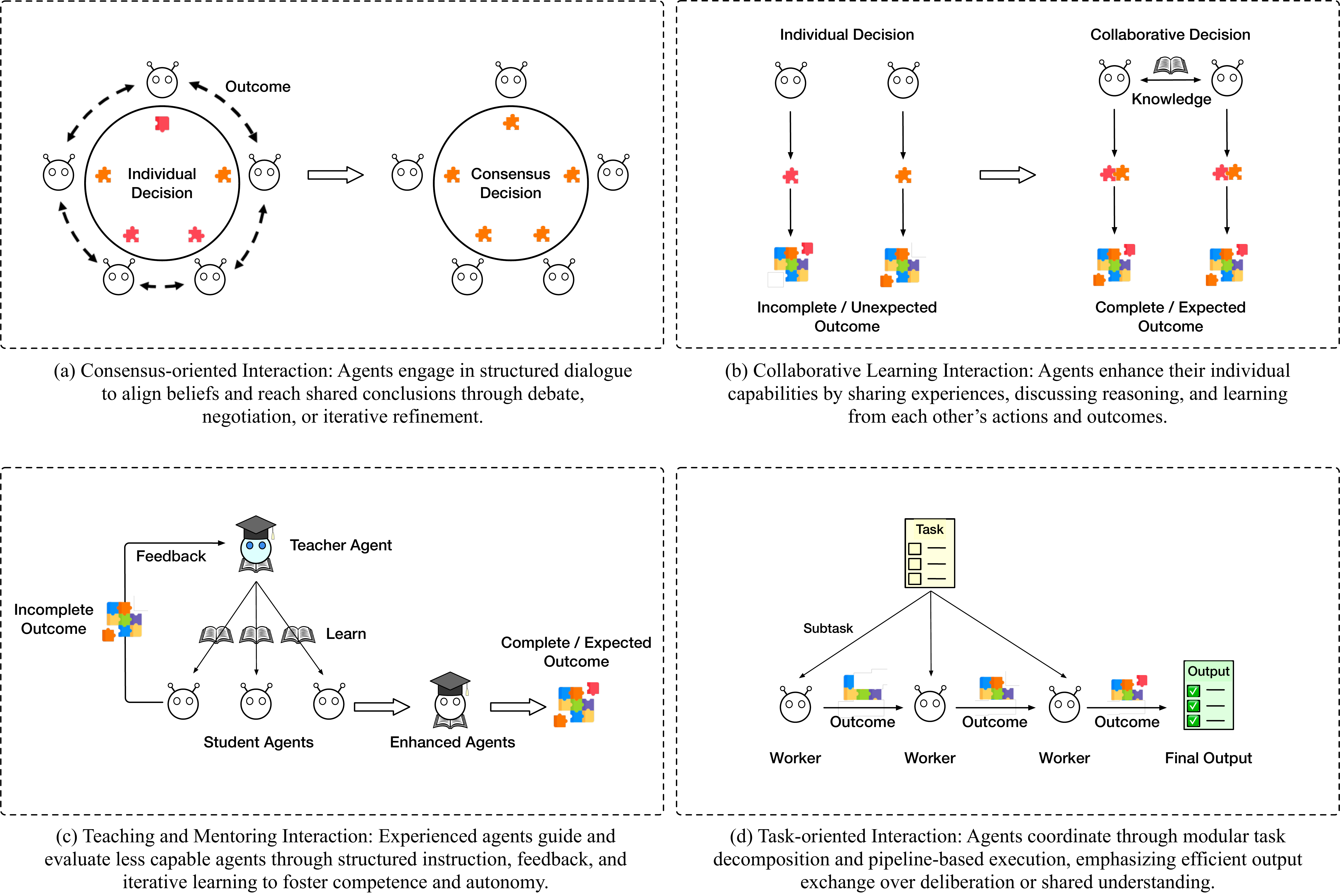}
    \caption{
    An overview of four agent-agent collaboration types in LLM-based MAS: \textit{Consensus-oriented}, \textit{Collaborative Learning}, \textit{Teaching/Mentoring}, and \textit{Task-oriented}. 
    }
    \label{fig:agent_collab_types}
\end{figure}

To better understand the diverse modes of collaboration in LLM-based MAS, we draw inspiration from sociological theories of human interaction, adapting them to model agent-agent behaviors. Specifically, we categorize inter-agent interactions into four paradigms: \emph{consensus-oriented}, \emph{collaborative learning}, \emph{teaching/mentoring}, and \emph{task-oriented} collaboration. These paradigms are illustrated in Figure~\ref{fig:agent_collab_types} and provide a finer-grained taxonomy that captures the complexity and fluidity of intelligent collaboration.

Each collaboration type serves different functional roles and exhibits distinct properties in terms of information flow, decision-making authority, goal alignment, and learning dynamics. Moreover, agents often engage in multiple interaction types concurrently, forming evolving networks of interdependent behavior. For instance, in collaborative software development systems~\cite{hong2023metagpt, qian2024chatdev}, a senior agent may guide junior developers through iterative dialogue (teaching), coordinate with peers on modular implementation (task collaboration), and participate in team-wide architectural planning (consensus building). Meanwhile, agents may also co-learn through testing and debugging phases, embodying collaborative learning.
Understanding these paradigms and the mechanisms that support them, such as dialog management, role assignment, knowledge sharing, and trust modeling, provides crucial insight into the design of more effective and adaptive multi-agent systems.

\subsection{Consensus-oriented Interaction}
\label{subsec:consensus-interact}

Consensus-oriented interaction focuses on aligning agent perspectives to arrive at a shared decision or judgment. It plays a pivotal role in situations that require integrating diverse knowledge sources and resolving disagreements through structured dialogue. This interaction mode is particularly valuable in complex problem-solving scenarios, where a unified understanding is essential for coherent and effective action~\cite{ruan2023learning, chen2023multi}.

In such settings, agents engage in multi-round discussions, iterative refinements, and mutual critique to harmonize their beliefs or decisions. For example, systems like MedAgents~\cite{tang2024medagents}, MDAgents~\cite{Kim2024MDAgentsAA}, and AI Hospital~\cite{fan2024ai} demonstrate how multidisciplinary agents, ranging from doctors to radiologists, can collaboratively sharpen diagnostic reasoning by bringing their respective expertise into focused alignment. These dialogues frequently surpass zero-shot or few-shot reasoning by fostering grounded, high-fidelity conclusions derived from collective deliberation.

Scientific collaboration further illustrates the need for consensus. In \emph{Agent Laboratory}\cite{schmidgall2025agent}, PhD and postdoctoral agents iteratively refine research goals, interpret experimental data, and consolidate their findings through structured debate and negotiation. Likewise, \emph{Virtual Lab}\cite{swanson2024virtual} combines plenary discussions, where agents align on long-term scientific agendas, with specialized subteam interactions for specific subtasks, forming a two-tiered consensus structure.

Several mechanisms support consensus formation, including \emph{discussion}, \emph{debate}, \emph{negotiation}, \emph{reflection}, and \emph{voting}. Each contributes distinctively: debates surface competing hypotheses; negotiations resolve resource conflicts or competing priorities; reflection supports iterative refinement based on new inputs; and voting allows distributed resolution when no consensus naturally emerges. These methods are often used in tandem to incrementally align agent beliefs.

Frameworks such as GPTSwarm~\cite{zhuge2024gptswarm} operationalize consensus through graph-based agent structures, where information flows along weighted edges shaped by trust or performance. Agents that consistently provide incorrect or low-quality outputs may be pruned from the consensus loop. RECONCILE~\cite{chen2023reconcile} uses structured round-table discussions, supplemented with confidence-weighted voting and reflective learning over past conversation rounds, to drive agreement. 

Debate-based strategies have shown particular promise for reducing hallucinations and improving decision quality~\cite{du2023improving, papachristou2023leveraging, liang2023encouraging, fu2023improving}. In GOVSIM~\cite{piatti2025cooperate}, agents simulate negotiation around a shared resource, with dialogue that extends beyond task execution into relationship-building and long-term planning. The Multi-Agent Debate (MAD) framework~\cite{liang2023encouraging} encourages creative solution-finding through turn-based argumentation, moderated by a judge agent. Similarly, the FORD framework~\cite{xiong2023examining} structures formal debates to steer consensus by allowing stronger models to influence weaker ones while preserving diversity of thought.

In collaborative refinement schemes like AutoAgents~\cite{chen2023autoagents}, agents iteratively append and revise their own and each other's messages to arrive at convergent actions, reflecting a bottom-up process of constructing shared understanding. Across these systems, consensus is not merely a final state but an evolving process of alignment through deliberate, often structured, social interaction.

\subsection{Collaborative Learning Interaction}
\label{subsec:collab-learning}

Collaborative learning interaction emerges when agents, often architecturally similar, share experiences and insights to enhance one another's capabilities. While these agents may start from a common foundation, their interactions with varied environments lead to diverging memories, strategies, and behaviors. By working together, they pool their experiences to accelerate individual learning, improve skill acquisition, and adapt to complex tasks. Over time, this reciprocal exchange fosters the evolution of each agent's internal models and behaviors.

The distinction between collaborative learning and consensus-oriented interaction lies in their underlying intent. Whereas consensus aims to synthesize divergent views into a shared conclusion, collaborative learning emphasizes mutual enrichment. Agents are not trying to agree on a single outcome but instead aim to benefit from exposure to diverse strategies, perspectives, and feedback. This peer-driven process promotes individual growth and collective advancement without requiring uniformity of thought.

A central mechanism in this paradigm is observational and experiential learning. Agents adjust their own strategies after witnessing how others approach similar problems. They may integrate useful patterns without committing to the same decisions or beliefs~\cite{Sreedhar2024SimulatingHS, lin2024strategic, piatti2024cooperate, xu2024exploring, du2024helmsman, stepputtis2023long, light2023avalonbench, wang2023avalon, shi2023cooperation}. As \citet{jin2024learning} highlight, the structure and quality of discussion significantly influence what and how agents learn from each other.

Three primary techniques structure collaborative learning interactions:
\begin{itemize}
\item \textbf{Experience sharing.} Agents communicate their insights and practical knowledge. \citet{qian2024iterative} demonstrate that iterative refinement through shared team experience enables agents to improve adaptively in software development. MAS-CTC~\cite{qian2023experiential} extends this principle to multi-team collaboration: different agent teams exchange decision rationales and insights, pruning lower-quality content via aggregation mechanisms. In MOBA~\cite{zhu2024moba}, global agents reflect on the actions of local agents to improve real-time coordination with the environment. Similarly, AutoAgents~\cite{chen2023autoagents} integrate multi-scale memory sharing, including short-term, long-term, and dynamic memories, across agents to refine behaviors through collective feedback.

\item \textbf{Peer discussion.} Agents engage in structured dialogues to explain, critique, and revise their reasoning processes. MEDCO~\cite{wei2024medco} creates a collaborative medical training environment where student agents enhance diagnostic reasoning through joint problem-solving. In \citet{xu2023towards}, agents conduct multi-round peer review by examining each other's reasoning steps, exchanging confidence scores, and refining outputs accordingly. These discussions help uncover diverse reasoning paths and support iterative accuracy improvements.

\item \textbf{Observational learning.} Agents acquire knowledge by analyzing the actions and outcomes of others. AgentCourt~\cite{chen2024agentcourt} trains lawyer agents to improve argumentation by learning from past courtroom debates. In iAgents~\cite{Kim2024MDAgentsAA}, agent communication mimics human social networks: agents actively exchange human-related knowledge and resolve information asymmetry using the InfoNav reasoning mechanism, organizing mixed memory to ensure relevance and informativeness. MARBLE~\cite{zhu2025multiagentbenchevaluatingcollaborationcompetition} combines cognitive expectation modeling with real-time feedback to continuously update agents' planning heuristics and decision quality.
\end{itemize}

Despite its promise, collaborative learning introduces notable challenges. Uneven capabilities among agents can lead to asymmetric learning opportunities; overreliance on peer insights risks amplifying noise or bias. Furthermore, maintaining diversity while allowing convergence requires careful management. Effective learning strategies must balance selective knowledge integration with robust individual reasoning. Fairness in information flow, scalability of coordination, and mechanisms for filtering useful contributions are all critical design considerations.

In dynamic and uncertain environments, the benefits of collaborative learning are significant. By enabling agents to iteratively learn from each other while preserving their autonomy, this interaction paradigm cultivates richer representations, more adaptive policies, and greater generalization, fueling both individual competence and collective intelligence.

\subsection{Teaching and Mentoring Interaction}
\label{subsec:teaching-mentoring}

To complement collaborative and consensus-based paradigms, \emph{teaching and mentoring interaction} introduces an asymmetric structure centered on intentional knowledge transfer. This paradigm plays a vital role in environments where skill acquisition and conceptual grounding are uneven across agents, enabling experienced agents to accelerate the learning trajectory of novices. Unlike collaborative learning, which emphasizes peer-level exchange, teaching focuses on directional, often hierarchical communication aimed at developing competence and autonomy in the learner.

In this context, teaching agents serve as guides, critics, and evaluators, shaping the learning experiences of mentee agents through structured feedback and domain expertise. Such interactions are particularly beneficial in complex multi-agent settings where not all participants start with the same capabilities or access to knowledge.

Core mechanisms in teaching and mentoring include:
\begin{itemize}
    \item \textbf{Criticism and Feedback.} Mentors assess learners' outputs, highlight misconceptions, and offer corrective suggestions. This fosters a feedback loop through which learners incrementally refine their reasoning, decision-making, and skill execution.
    
    \item \textbf{Evaluation.} Teaching agents monitor learner progress via diagnostic tasks, scoring rubrics, or simulated environments. These evaluations provide developmental checkpoints that inform future guidance and allow adaptation to individual learner needs.
    
    \item \textbf{Instruction and Demonstration.} Mentors deliver explicit knowledge or procedural instruction, often supplemented by examples, allowing learners to observe, imitate, and clarify misunderstandings through guided interaction.

\end{itemize}

\paragraph*{Iterative Development through Guided Practice}
Teaching is rarely a one-off transmission but typically unfolds through iterative cycles of task execution, observation, and feedback. In MEDCO~\cite{wei2024medco}, student agents improve their clinical reasoning via repeated practice, structured by expert agents who evaluate patient interaction and diagnostic steps. These expert mentors not only assess performance but also intervene with contextual instruction tailored to learner gaps. Similarly, \citet{li2024agent} show that an agentic doctor can steadily enhance diagnostic proficiency by repeatedly interacting with agentic patients in a simulated environment, effectively internalizing real-world medical knowledge through simulation.

Teaching interactions can take multiple forms depending on the flow of knowledge. \emph{Unidirectional} teaching resembles traditional classroom instruction, where information is explicitly conveyed from mentor to learner, often through lectures or written guidance~\cite{wei2024medco}. In contrast, \emph{interactive mentoring} integrates bidirectional dialogue, allowing learners to ask clarifying questions, propose hypotheses, or receive dynamic responses, transforming static instruction into an adaptive educational exchange.

\paragraph*{From Supervision to Autonomy}
Over time, effective teaching transitions from direct supervision to increasing learner independence. As agents internalize feedback and demonstrate competence, mentors step back, allowing the mentees to operate with greater autonomy. This progressive withdrawal not only reflects trust in learner growth but also frees mentor agents to assist new learners, creating a virtuous cycle of scalable education within MAS.

By embedding teaching and mentoring mechanisms into multi-agent frameworks, systems can ensure more equitable skill distribution, accelerate onboarding of less experienced agents, and cultivate a robust foundation for collective intelligence. This paradigm, while rooted in structured guidance, ultimately empowers agents to become effective collaborators and future mentors themselves.

\subsection{Task-oriented Interaction}
\label{subsec:task-interact}

Task-oriented interaction represents a pragmatic and execution-driven paradigm in MAS, where agents collaborate not through extended deliberation, but by aligning along structured workflows and interdependent tasks. The primary interaction mechanism centers on the exchange of intermediate results between agents according to a predefined task decomposition. Each agent focuses on a specific role in the pipeline, processing inputs from upstream and generating outputs for downstream agents. This form of collaboration emphasizes \emph{coordination over conversation}.

Unlike consensus-driven or teaching-based paradigms, task-oriented interaction is characterized by strict modularity, temporal ordering, and minimal feedback loops. Agents do not necessarily share a common belief space, but instead rely on task dependencies, process interfaces, and clear deliverables to maintain coherence across the system.

\paragraph*{Structured Pipelines in Software and Reasoning Tasks}
Recent frameworks illustrate several archetypes of this interaction pattern. In \emph{software development}, systems such as MetaGPT~\cite{hong2023metagpt} and ChatDev~\cite{qian2024chatdev} orchestrate agents in roles that mirror the software engineering lifecycle: product managers analyze requirements, architects produce technical blueprints, developers implement features, and QA agents validate outputs. These pipelines are executed with minimal discussion across roles, relying instead on well-defined documents and task handoffs.

In \emph{collaborative reasoning}, frameworks like Exchange-of-Thought (EoT)\cite{yin2023exchange}, GPTSwarm\cite{zhuge2024gptswarm}, and MACNET~\cite{qian2024scaling} design interaction topologies, such as rings, trees, DAGs, or learned graphs, that regulate the flow of information between agents. These topologies constrain expansion of context, optimize reasoning across multiple agents, and ensure only refined solutions are propagated forward. Similarly, Alpha-SQL~\cite{alphasql} reformulates the Text-to-SQL task as a search problem where each node in the Monte Carlo Tree represents a partial solution step (e.g., selecting schema elements or refining queries), with agents responsible for specific subtasks in constructing the final SQL output. Building on this, EllieSQL~\cite{elliesql} further exemplifies task-oriented interaction by employing a complexity-aware routing framework that dynamically assigns Text-to-SQL queries to different SQL generation pipelines based on their estimated complexity, thereby optimizing cost-efficiency without sacrificing performance.

\paragraph*{Workflow Automation in ML and Complex Domains}
Task-oriented structures are also prevalent in \emph{machine learning applications}, such as AutoKaggle~\cite{li2024autokaggle} and AutoML team-based systems~\cite{trirat2024automl}, where agents specialize in dataset curation, model selection, training, and evaluation. These agents follow a disciplined execution plan with little need for mutual interpretation. TraveLER~\cite{shang2024traveler}, targeting VideoQA tasks, further extends this approach with a modular breakdown comprising Traverse, Locate, Evaluate, and Replan, coordinated by a Planner agent that iteratively refines strategies based on component feedback.

In many of these systems, agents function akin to a production line: interaction occurs through passing artifacts, not ideas. Graph-based designs like MACNET~\cite{qian2024scaling} adopt DAG topologies where supervisory agents issue directives and executor agents solve subproblems in sequence. This structure enforces logical consistency and avoids context bloat by restricting memory propagation to essential outputs.

\paragraph*{Adaptive Execution in Open-ended Environments}
Beyond structured domains, task-oriented MAS can also adapt to open-ended or dynamic environments. In Minecraft-style simulations~\cite{chen2024s}, leader agents decompose high-level objectives into granular subtasks, such as resource gathering, crafting, or defense, which executor agents complete in parallel. These systems emphasize flexible coordination through task assignment, communication protocols, and real-time adaptation to environmental changes.

\paragraph*{Coordinated Efficiency over Deliberative Consensus}
The strength of task-oriented interaction lies in its operational efficiency. It avoids the cognitive and computational overhead of multi-turn dialogue, favoring clarity, parallelism, and modularity. Coordination strategies, such as synchronization points, shared memory buffers, and priority queues, ensure robust collaboration without requiring shared understanding.

These systems demonstrate that scalable agent collaboration does not always require consensus or mutual teaching. In many complex applications, particularly in engineering, automated research, and real-time systems, task decomposition and clear interfacing are sufficient for sophisticated multi-agent execution~\cite{wu2023autogen,hong2023metagpt}.

\section{Human-Agent Collaboration}
\label{sec:human-agent-collab}

\begin{figure}[!ht]
\centering
\includegraphics[width=\textwidth]{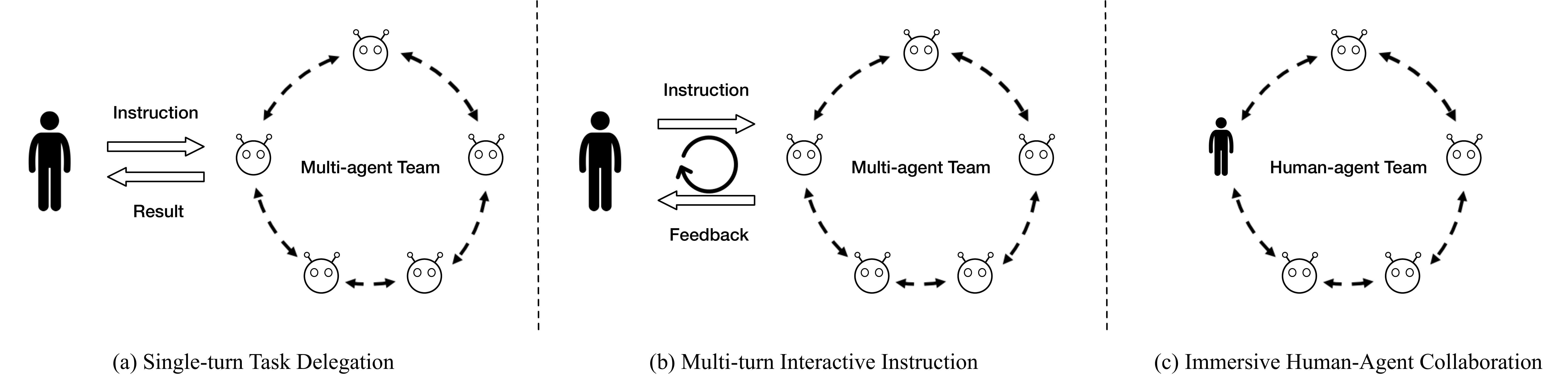}
\caption{Three paradigms of human-agent collaboration. (\textbf{a}) In single-turn task delegation, humans issue standalone commands and receive full responses without further interaction. (\textbf{b}) Multi-turn interactive instruction introduces feedback loops, where humans supervise and refine outputs iteratively. (\textbf{c}) Immersive collaboration enables fluid, symmetric interaction between humans and agents, fostering real-time co-creation and shared decision-making.}
\label{fig:human-agent-collab}
\end{figure}

\lettrine[lines=3]{\initfamily\textcolor{darkgreen}{W}}{hile} previous sections focus on how agents interact and collaborate among themselves, a complete MAS ecosystem must also consider the interface between agents and humans. As illustrated in Figure~\ref{fig:human-agent-collab}, human-agent collaboration forms a critical bridge for grounding AI systems in real-world use, enabling people to harness the capabilities of MAS to solve diverse problems. This collaboration generally falls into three paradigms: \emph{single-turn task delegation}, \emph{multi-turn interactive instruction}, and \emph{immersive human-agent collaboration}.

\paragraph*{Single-turn Task Delegation}
In the simplest form, humans delegate discrete tasks to MAS through a one-off command or request, known as \emph{single-turn task delegation}. Examples include asking a Q\&A system a factual question or assigning a standalone coding or generation task~\cite{chen2024smurfs,hong2023metagpt}. The MAS processes the request and autonomously returns a complete response without further interaction. This mode currently dominates how most users interact with LLM-based agents~\cite{tang2024medagents,qian2024chatdev,zhuge2023mindstorms}.
Building on this baseline, more advanced systems allow users to \emph{pre-configure} the agent composition and coordination strategies before execution~\cite{pan2024agentcoord}, enabling customized orchestration and improving both performance and controllability.

\paragraph*{Multi-turn Interactive Instruction}
Moving beyond static delegation, \emph{multi-turn interactive instruction} introduces iterative feedback loops between humans and agents. In this mode, users refine requests, correct misunderstandings, and steer the agent toward satisfactory outcomes across multiple rounds of interaction. This dynamic is especially prevalent in creative and open-ended tasks, such as editing text, generating visual content, or drafting code~\cite{gao2024aligning}.
Here, humans typically assume a \emph{supervisory role}, offering guidance, approval, and clarification throughout the process. For instance, users may iteratively instruct the agent to revise portions of a generated image, change design elements, or adjust the logic of a program. Systems that support real-time human feedback, such as through dialog, review checkpoints, or approval gating, have shown promise across domains like household robotics~\cite{Ren2023Robots}, web automation~\cite{takerngsaksiri2025humanintheloop}, software development~\cite{han2025convcodeworld}, and co-programming~\cite{huq2025cowpilot}.

\paragraph*{Immersive Human-Agent Collaboration}
In the most integrated paradigm, \emph{immersive collaboration}, LLM-based agents operate as peer collaborators rather than subordinates. Unlike interactive instruction, where humans guide the agent, immersive collaboration enables both parties to initiate actions, propose ideas, and make shared decisions. The interaction becomes symmetric and continuous, more akin to teamwork than command-following.
Such agents may co-create content, join meetings as assistants, or participate in domestic tasks such as cooking or cleaning~\cite{leong2024Dittos,zhang2023building,zhang2025leveraging}. By mirroring human-like initiative, situational awareness, and dialog capabilities, these agents foster \emph{fluid, real-time partnership} rather than sequential command chains.

\paragraph*{Evaluating Human-Agent Collaboration}
To better understand and evaluate these modes of interaction, recent benchmarks such as Co-Gym~\cite{shao2025collaborative} offer structured environments that measure collaborative quality along key dimensions: communication clarity, situational alignment, personalization, and mutual efficiency. Tasks such as travel planning, writing related work sections, and data analysis have been used to benchmark different collaboration styles and quantify agent–human synergy.

\paragraph*{}
As MAS capabilities mature, the scope of human-agent collaboration continues to broaden. From single-shot interactions for factual queries, to supervisory loops for design and coding, and all the way to rich joint action in everyday life, LLM-based agents are becoming more capable, autonomous, and interactive. A recent survey~\cite{zou2025surveylargelanguagemodel} offers a comprehensive overview of these trends and challenges.
Looking forward, these agents are poised to deeply integrate into daily workflows, augmenting human productivity across domains. In turn, human users will need to adapt their communication strategies and expectations to engage effectively with intelligent agents. This co-evolution of collaboration practices and agent capabilities will reshape not only task completion, but also the broader fabric of human labor, creativity, and cooperation in the LLM era.

\section{Decision-Making in Multi-Agent Systems}
\label{sec:mas-decision-making}

\begin{figure}[!ht]
\centering
\includegraphics[width=0.9\textwidth]{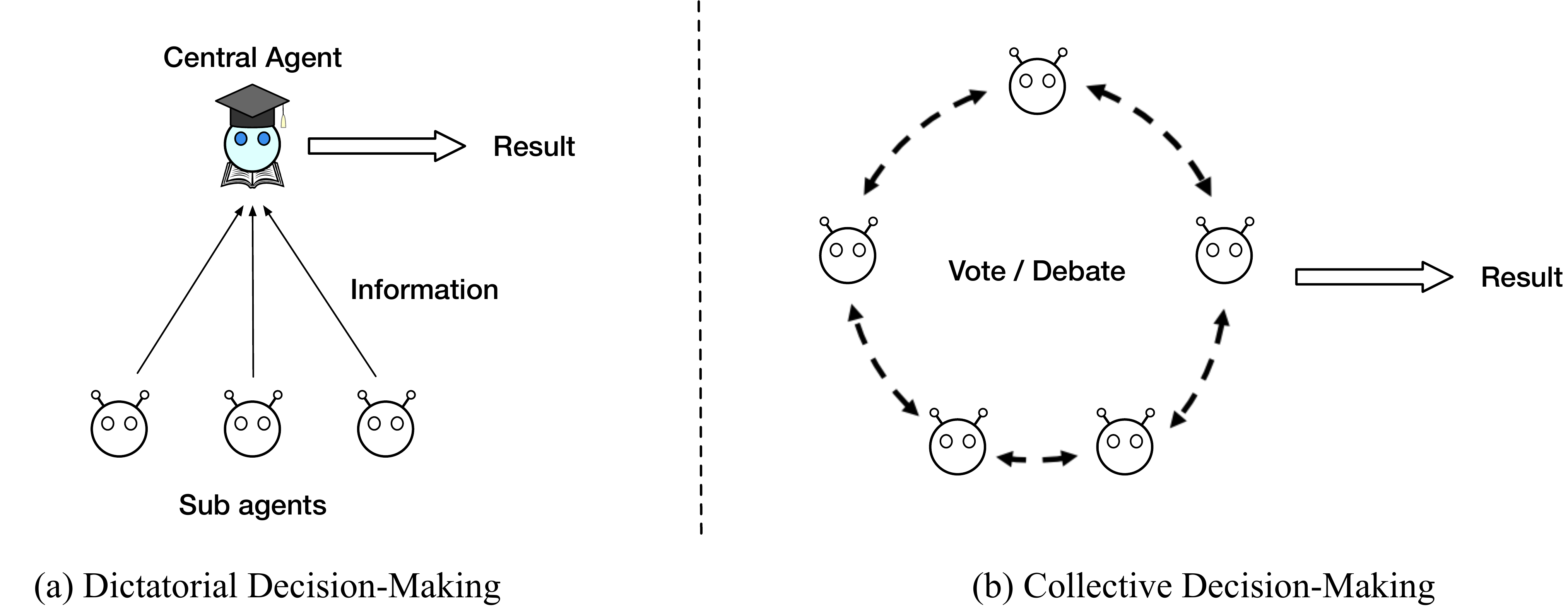}
\caption{Two principal paradigms of decision-making in multi-agent systems. (\textbf{a}) \emph{Dictatorial decision-making} concentrates authority in a central agent, who synthesizes information from sub-agents to generate final decisions. (\textbf{b}) \emph{Collective decision-making} distributes authority among agents, who vote or debate to reach consensus through local interactions. Each paradigm reflects distinct trade-offs in efficiency, transparency, and robustness.}
\label{fig:decision-paradigms}
\end{figure}

\lettrine[lines=3]{\initfamily\textcolor{darkgreen}{F}}{ollowing} our discussion on collaboration paradigms, we now examine how agents within a MAS make decisions, a fundamental capability that underpins their ability to act cohesively, adapt to complex tasks, and maintain system-level efficiency. As illustrated in Figure~\ref{fig:decision-paradigms}, decision-making architectures play a crucial role in shaping how agent interactions yield coherent outcomes. While collaboration is the medium through which agents interact, \textit{decision-making} governs how these interactions lead to final actions and results.
Recent work has emphasized that the choice of decision-making strategy significantly influences the emergent behavior and overall effectiveness of MAS. For instance,~\cite{zhao2024electoral} demonstrates that different decision architectures can alter the efficiency and fault tolerance of collaborative processes, while~\cite{li2024improving} highlights that structured decision-making mechanisms can stimulate system-wide emergent intelligence.

Broadly, collaborative decision-making in MAS can be categorized into two architectural paradigms: \textbf{dictatorial decision-making} and \textbf{collective decision-making}~\cite{zhao2024electoral}. These paradigms differ in how authority is distributed, how information is processed, and how final decisions are reached.

\subsection{Dictatorial Decision-Making}
\label{subsec:dictatorial-dm}

Dictatorial decision-making centralizes authority in a single agent (or node) within the system. Other agents contribute information, such as local observations or intermediate results, which the designated decision-maker synthesizes to form a global decision. This model enables top-down reasoning with a unified perspective, allowing for efficient coordination and policy consistency.

Several LLM-based MAS architectures follow this pattern. For example,~\cite{liang2023encouraging,nair2023dera,Kim2024MDAgentsAA} employ a single LLM to aggregate diverse agent outputs and formulate a decision that reflects a more objective synthesis of viewpoints. Similarly,~\cite{Jiang2023LLMBlenderEL,si2023getting} explore decision fusion via ranking, scoring, and checklist-based methods to enhance decision robustness.

Beyond final aggregation, dictatorial structures also appear in hierarchical workflows. Central agents in~\cite{chen2023autoagents,schroeder2024thread} decompose complex problems into sub-tasks and dispatch them to specialized agents. These agents execute narrowly scoped responsibilities, after which the central coordinator reassembles the results.

In distributed variants of this model~\cite{zhuge2024gptswarm,qian2024scaling}, the decision-maker may not be explicitly defined. Instead, the last node in a structured topology (e.g., a DAG or pipeline) implicitly serves as the concluding agent. It processes accumulated knowledge and outputs a system-level judgment, functioning as a distributed, yet still centralized, decision point.

\subsection{Collective Decision-Making}
\label{subsec:collective-dm}

In contrast, collective decision-making distributes decision authority among agents, enabling them to jointly determine outcomes through local reasoning, negotiation, or voting. This bottom-up paradigm fosters robustness, adaptability, and resilience in dynamic or partially observable environments.

\paragraph*{Voting-Based Decision Making}
Voting is a foundational mechanism in collective decision-making, offering a lightweight and interpretable means for reaching consensus. Systems such as~\cite{chen2023multi,wang2023avalon} adopt majority or plurality voting among agents to finalize answers or select action plans. The GEDI electoral module~\cite{zhao2024electoral} expands this further, incorporating multiple voting strategies to improve collective reasoning while avoiding complex arbitration logic. These methods enhance fault tolerance and preserve simplicity in agent interactions.

\paragraph*{Debate-Based Decision Making}
Debate-based mechanisms emphasize interactive deliberation among agents to resolve uncertainty or conflicting perspectives. In~\cite{liang2023encouraging,liu2024groupdebate}, agents participate in structured discussions where they present arguments, critique one another's proposals, and refine shared solutions through back-and-forth dialogue.
More nuanced protocols~\cite{xu2023towards,liu2024dynamic} focus on iterative consensus-building, where agents repeatedly exchange updates to reconcile disagreements. To mitigate the ``cognitive island'' problem, where agents lack shared knowledge, systems such as Apollo~\cite{wang2023apollo} incorporate a common retrieval knowledge base that aligns agent understanding and grounds their reasoning. These systems emulate human-style conversations to enable distributed agents to converge on more informed, justified decisions.

\paragraph*{}
In summary, decision-making in MAS spans a spectrum, from centralized, top-down control to distributed, emergent consensus. The choice of decision architecture reflects trade-offs between efficiency, robustness, transparency, and flexibility. As MAS become more autonomous and general-purpose, hybrid architectures that blend both paradigms are increasingly explored, seeking to combine the strengths of centralized oversight with the adaptability of distributed reasoning.

\section{Communication Protocols in Multi-Agent Systems}
\label{sec:mas-protocols}

\lettrine[lines=3]{\initfamily\textcolor{darkgreen}{A}}{s} the capabilities and complexity of multi-agent systems (MAS) grow, effective communication becomes the foundation for coordination, reasoning, and collaboration across agents. Building upon the decision-making mechanisms discussed earlier, this section turns to the design of communication protocols that enable agents to share knowledge, delegate tasks, and interact coherently with each other, with users, and with external environments.

We begin by categorizing typical message types, ranging from queries and commands to status updates and feedback, clarifying the modes and semantics of agent interaction. Next, we examine the distinct communication interfaces across different interaction channels: agent–environment, agent–agent, and agent–human. Each interface presents unique architectural and semantic challenges, and we explore emerging design patterns that support transparency, modularity, and robustness in these interactions.
A special focus is placed on the architectural underpinnings of communication protocols, including serialization formats, messaging layers, and routing mechanisms. Standardized interfaces and protocol specifications are essential for achieving interoperability, reusability, and scalability in MAS, especially as systems increasingly integrate heterogeneous agents, tools, and environments.
We conclude the section with a unifying perspective on communication in LLM-based systems, highlighting shared principles across agent–environment and agent–user communication, and advocating for coherent, modular designs that enable fluid interaction across diverse applications. This unified protocol layer is especially critical for ensuring that large-scale, general-purpose agent systems remain consistent, extensible, and grounded in interpretable interaction logic.

\subsection{Types of Agent Messages}
\label{subsec:agent-message-types}


Effective communication within Multi-Agent Systems (MAS) hinges on the nature and structure of exchanged messages. These messages can broadly be categorized into two types: \textbf{structured} and \textbf{unstructured}. Each serves distinct purposes and offers unique advantages depending on the application context.

\paragraph*{Structured Messages} 
Structured messages, typically encoded in JSON~\cite{gandhi2025budgetmlagent,lu2025karma}, XML~\cite{zhang2025large,Chi2023AppAgent}, or programming code~\cite{hong2023metagpt,qian2024chatdev,huang2024agentcoder}, is a core component of LLM-based multi-agent communication. Their predefined syntax and semantics allow for unambiguous interpretation, easy parsing, and dependable information exchange. These formats enable agents to encode task parameters, directives, or even executable routines in a machine-readable and verifiable form, streamlining workflow automation and minimizing interpretive errors.

Structured messages are particularly well-suited to deterministic, high-efficiency tasks such as sub-task decomposition, workflow orchestration, and coordination across agents in hierarchical or pipeline architectures. Their consistency supports system-level optimization, persistent memory logging, and retrospective analysis. In such settings, the explicitness of structure enables efficient retrieval, storage, and validation of data throughout multi-agent interactions.

\paragraph*{Unstructured Messages} 
In contrast, unstructured messages, such as natural language~\cite{xu2024exploring,wu2023deciphering,Lan2023LLMBasedAS}, images, video, and audio signals~\cite{wang2025audioagent,sevi}, convey richer, more context-sensitive information. These modalities capture nuanced cues: visual content expresses spatial and emotional context; audio signals transmit not just spoken language but also tone and affect; and text allows for abstract reasoning, negotiation, or goal specification with flexibility beyond rigid formats.

Unstructured messages are vital in open-ended or ambiguous settings where communication must accommodate uncertainty, creativity, or evolving environments. While their complexity historically posed a challenge for automated interpretation, recent progress in LLMs and multimodal models~\cite{zhang2023speechgpt,liu2024llava,openai2024gpt4} has made such communication increasingly tractable, enabling agents to understand, generate, and ground unstructured content with remarkable fluency.

\paragraph*{Hybrid Messaging for Adaptive Communication}
In practice, structured and unstructured messages often complement one another. Structured messages provide precision, reliability, and computational efficiency, making them ideal for operational control and deterministic execution. Unstructured messages, by contrast, support adaptive, expressive, and context-rich exchanges, crucial for creative collaboration, negotiation, and real-world interaction. Together, these modalities form a hybrid communication foundation that enables intelligent, flexible cooperation among agents in diverse MAS applications.

\subsection{Communication Interfaces}
\label{subsec:agent-commu-interfaces}


The effectiveness of a multi-agent system (MAS) hinges not only on the quality of the messages exchanged but also on the design of its communication interfaces. These interfaces govern how agents interact with their surrounding environments, with other agents, and with human users. Ensuring clarity, modularity, and interoperability across these interfaces is essential for robust, generalizable, and user-aligned MAS architectures.

\paragraph*{Agent–Environment Interfaces} 
LLM-based agents often need to interact with external environments to accomplish tasks, such as clicking a UI button, issuing a web request, or controlling an embodied avatar in a simulation. From the agent's perspective, each action represents a desired effect on the environment. However, environments vary widely in the actions they support. Thus, to ensure meaningful interaction, agents must first infer or query the affordances of the environment and adapt their behavior accordingly. 

Once an action is issued, the environment typically returns either a successful observation or an error signal. The agent must interpret this feedback to adjust future decisions. Environments may include operating systems, web interfaces, virtual worlds, simulations, or real-time robotics platforms. To standardize these interactions, several frameworks have been developed to unify the agent-environment interface, enabling agents pretrained via LLMs to generalize across diverse execution contexts with minimal adaptation~\cite{liu2023agentbench}. These frameworks foster benchmarking, skill transfer, and modular deployment.

\paragraph*{Agent–Agent Interfaces} 
Communication between agents forms the backbone of any MAS. In LLM-based systems, natural language is the predominant mode of communication, largely due to LLMs' pretraining on extensive linguistic corpora. Text-based communication is particularly popular, enabling agents to discuss, negotiate, critique, and persuade one another using expressive and flexible dialogue~\cite{tang2024medagents,zhang2023building,sun2024lawluo,wu2023deciphering,visnl}. In more embodied or multimodal scenarios, voice-based interactions also emerge as viable channels~\cite{sevi,askhuman,yang2024askchart}.

In contrast, some systems opt for structured communication formats to reduce ambiguity and parsing complexity. These formats enable more efficient and deterministic agent coordination~\cite{hong2023metagpt}. For example, systems like TaskWeaver~\cite{qiao2024taskweaver} adopt structured message schemas with clearly defined fields such as \texttt{sender}, \texttt{receiver}, \texttt{message type}, and \texttt{content}, along with explicit parsing instructions. Such modular protocols facilitate scalability and robustness in high-volume, high-stakes agent interactions.

\paragraph*{Human–Agent Interfaces} 
Ultimately, multi-agent systems are built to augment human cognition, extend agency, and improve societal outcomes. While some MAS frameworks focus on human-as-observer settings, e.g., in social simulations~\cite{park2023generative,wang2023humanoid}. Many systems support active human participation. In these settings, communication may occur via either natural language or structured formats~\cite{zhang2023building,takerngsaksiri2025humanintheloop}.

Natural language remains the most intuitive medium for human interaction. To bridge the gap between human expressiveness and agent system requirements, a central LLM component often serves as a translator, parsing user language into structured actions or queries that agents can interpret and execute. This LLM mediator may be embedded within the MAS or deployed as an external interface layer.

For more efficient, programmatic control, some systems also enable humans to interact directly via structured messages, following predefined schemas or APIs. This approach is particularly useful for technical users or scripted workflows~\cite{anthropic_mcp}. By aligning human input with the internal logic of the MAS, such interfaces can streamline interaction while maintaining high-level control and traceability.

\subsection{Next-Generation Communication Protocols}
\label{subsec:next-gen-protocols}

The field of LLM-based multi-agent systems is still in its early stages. Current agent architectures and communication mechanisms are often designed in an ad hoc manner for specific domains or tasks, including agent–environment, agent–human, and inter-agent interactions. However, the lack of a unified communication framework leads to fragmented and siloed ecosystems: tools, agents, environments, and data sources frequently operate in isolation, impeding composability and cross-system collaboration. Moreover, most existing protocols are manually crafted, placing a high design burden on developers and often lacking semantic flexibility, adaptability, or scalability.

To address these limitations, several new communication protocols have recently been proposed, each targeting distinct layers of the protocol stack, from transport and identity to semantic negotiation and tool invocation.

\begin{figure}[!ht]
\centering
\includegraphics[width=0.45\textwidth]{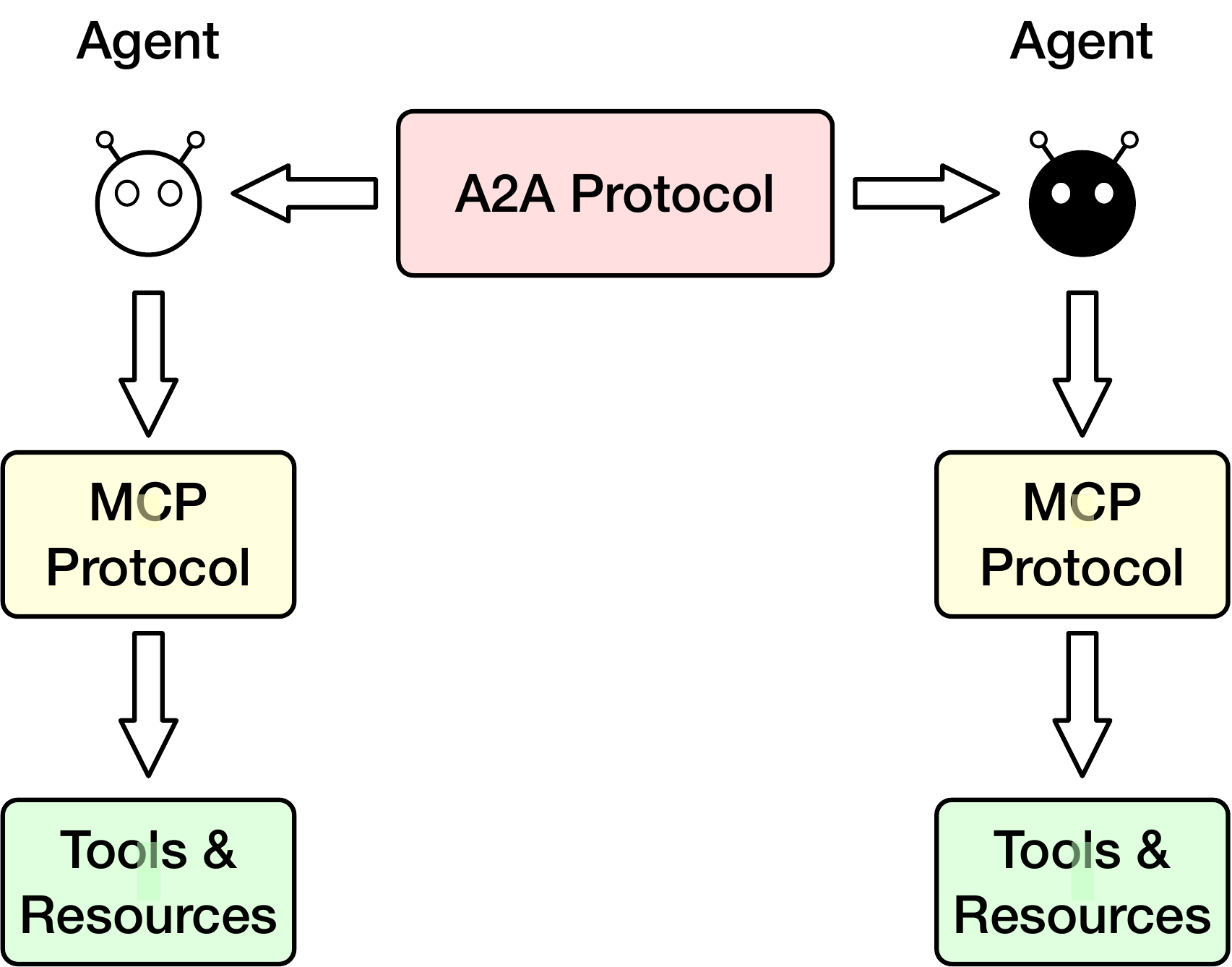}
\caption{Illustration of the two foundational communication protocols in multi-agent systems. The \emph{MCP (Model Context Protocol)} governs how each agent interfaces with tools and external resources. The \emph{A2A (Agent-to-Agent) Protocol} defines the communication channel between agents, supporting coordination, negotiation, and collaboration. Together, these protocols establish a modular and interoperable framework for scalable and composable agent systems.}
\label{fig:mcp-a2a}
\end{figure}

\paragraph*{Internet of Agents (IoA)}  
\citet{chen2024internet} proposes an internet-inspired, instant-messaging-style architecture for agent communication, enabling dynamic team formation and task-driven collaboration. Agents register with a centralized coordination server that manages identity, discovery, and message routing. Communication is orchestrated using finite state machine (FSM)-based dialogue templates, allowing flexible structuring of discussion, delegation, and triggering messages. IoA supports features such as speaker turns, nested group dialogues, and message-length constraints, enabling agents to adapt message formats according to different coordination phases. While flexible within a fixed schema, it remains reliant on a central orchestrator.

\paragraph*{Model Context Protocol (MCP)}  
\citet{anthropic_mcp} introduces a client-server architecture aimed at simplifying tool access and structured data retrieval for LLM agents. MCP decouples tool invocation from LLM-generated text, improving privacy, stability, and interoperability. Tools are registered through a standardized interface, enabling cross-model reuse and compositional workflows. Its rigid and minimalistic design makes MCP suitable for high-assurance, tool-centric workflows. However, it lacks support for dynamic protocol negotiation, semantic adaptation, or flexible extension, which limits its applicability in decentralized or adaptive agent ecosystems.

\paragraph*{Agent Network Protocol (ANP)}  
\citet{agent_network_protocol} proposes a decentralized protocol inspired by Web3 standards. Agents identify themselves using W3C-compliant decentralized identifiers (DIDs) and communicate via encrypted peer-to-peer channels. ANP includes a meta-protocol layer that enables agents to negotiate which application-layer protocol to use, supporting multi-protocol environments (e.g., HTTP, JSON-RPC, natural language) and allowing semantic protocol selection based on agent capabilities. While it prioritizes extensibility and decentralization, ANP currently lacks explicit support for public protocol reuse or standardized capability registries.

\paragraph*{Agora}  
\citet{marro2024scalable} presents a highly flexible, language-native protocol system. Rather than relying on pre-registered APIs, agents exchange Protocol Descriptions (PDs), which are free-text descriptions of communication semantics. LLMs interpret and execute PDs dynamically at runtime, enabling agents to create, deploy, and adopt new protocols entirely through language. Agora operates without centralized registries, leveraging distributed repositories for protocol sharing and reuse. This supports continual learning and interoperability, though it introduces potential ambiguity and verification challenges due to its open-ended nature.

\paragraph*{Agent2Agent (A2A)}  
\citet{rao2025a2a} defines an open, web-native communication protocol that enables autonomous agents to discover, negotiate, and collaborate on long-running tasks without disclosing internal memory or implementation details. Built on HTTPS transport and JSON-RPC 2.0, A2A introduces machine-readable Agent Cards that advertise capabilities, endpoints, and authentication mechanisms. Agents form asynchronous Tasks governed by lifecycle hooks, observability metadata, and policy controls. Positioned above resource-level protocols like MCP, A2A enables secure, agent-level orchestration while delegating tool invocation to subordinate protocols, fostering modularity and separation of concerns. Figure~\ref{fig:mcp-a2a} illustrates the relationship between A2A and MCP protocol.

\paragraph*{Agntcy}  
\citet{agntcy2025} envisions a federated, internet-scale platform for AI agents, grounded in four design pillars: interoperability, security, scalability, and standardization. At its core, Agntcy integrates a collision-resistant identity layer with an Open Agentic Schema Framework (OASF) and a distributed Agent Directory for discovery and capability alignment. On this foundation, the Agent Connect Protocol (ACP) provides a lightweight invocation interface, while the Agent Gateway manages transport-level security. Agntcy promotes vendor-agnostic, large-scale collaboration across diverse agent ecosystems, offering a unified, open substrate for the Internet of Agents.

\subsection{Protocol Comparison and Future Directions} 
\label{subsec:protocol-compare}

As shown in Table~\ref{tab:agent-protocols}, next-generation agent communication protocols differ along key dimensions such as identity and security mechanisms (e.g., OAuth, DID, VC), meta-protocol negotiation capabilities (e.g., only ANP and Agora support it), data exchange formats (e.g., JSON-RPC, gRPC, JSON-LD), the degree of centralization, and application-layer flexibility. A unified, secure, scalable, and dynamic protocol infrastructure, where agents can negotiate and co-create protocols in real time, is critical for enabling large-scale, interoperable agent ecosystems. While current frameworks such as MCP, ANP, Agora, IoA, A2A, and Agntcy represent early but promising steps, protocol design remains a rapidly evolving frontier. Key challenges ahead include improving semantic interoperability, supporting collective coordination, and ensuring adaptability in open and heterogeneous environments. As highlighted in recent work~\cite{yang2025survey,wang2025internetagents}, future protocols should emphasize modularity, privacy preservation, and model-native protocol understanding, moving away from rigid, static formats toward infrastructures that allow agents to dynamically compose, adapt, and evolve their communication protocols over time.

\begin{table*}[ht]
\centering
\caption{Concise comparison of six agent communication protocols across key layers. \textbf{PD} = Protocol Description; \textbf{DID} = Decentralized Identifier; \textbf{VC} = Verifiable Credential; \textbf{NL} = Natural Language; \textbf{AMP} = Agent Message Protocol; The \textbf{Meta-Protocol} denotes a mechanism that lets agents \emph{negotiate or co-create their own protocol}.}
\ra{1.3}
\resizebox{\textwidth}{!}{
\begin{tabular}{p{3.1cm}p{2.2cm}p{2.6cm}p{2.4cm}p{3.2cm}p{2.8cm}p{2.8cm}}
\toprule
\textbf{Layer} & \textbf{MCP} & \textbf{ANP} & \textbf{Agora} & \textbf{IoA} & \textbf{A2A} & \textbf{Agntcy} \\
\midrule
\rowcolor{LightMint}
\textbf{Identity \& Security} & OAuth & DID & - & Agent registry rules & OAuth, API Keys & VC\\

\textbf{Meta-Protocol} & – & \yes & \yes & – & – & – \\

\rowcolor{LightMint}
\textbf{Data Exchange} & JSON-RPC & JSON-LD & PD / NL & AMP & JSON-RPC & gRPC\\

\textbf{Centralization} & Centralized & Decentralized & Decentralized & Centralized & Decentralized & Hybrid \\

\rowcolor{LightMint}
\textbf{Flexibility} & Rigid & Flexible & Highly flexible & Moderate & Moderate & Moderate \\

\textbf{Publisher} & Anthropic & ANP Comunity & Eigent AI& Tsinghua University & Google & Langchain, Cisco \\
\bottomrule
\end{tabular}}
\label{tab:agent-protocols}
\end{table*}

\section{Summary and Discussion}  
\label{sec:summary-mas}

\lettrine[lines=3]{\initfamily\textcolor{darkgreen}{T}}{his} chapter presented a systematic overview of the foundational concepts and critical design choices involved in constructing collaborative multi-agent systems (MAS) based on large language models (LLMs). We first explored strategies for composing agent teams, differentiating between homogeneous agents, which is beneficial for simpler tasks demanding parallelism, and heterogeneous agents, which excel in complex scenarios by bringing diverse expertise, observation capabilities, and action spaces to the collective problem-solving process. Importantly, we highlighted emergent agent specialization, showcasing how homogeneous agents can evolve diverse capabilities through continuous interaction, fostering increased adaptability.

We then addressed multi-agent system topologies, delineating between static and dynamic structures. Static topologies, well known as hierarchical, decentralized, and centralized, offer advantages in simplicity, predictability, and ease of management but struggle with real-time adaptability. Conversely, dynamic and adaptive topologies leverage flexible reconfiguration strategies, significantly enhancing system responsiveness and scalability. Scalability emerged as a critical consideration, underscoring the necessity of balancing coordination overhead against task complexity, especially as agent populations grow. Frameworks such as AgentScope, Project Sid, and AgentSociety exemplified effective solutions to scalability challenges by integrating decentralized structures, parallel processing, and high-performance messaging systems.

The chapter further explored four key paradigms of agent collaboration: consensus-oriented, collaborative learning, teaching and mentoring, and task-oriented interactions. Each interaction type uniquely shapes agent behaviors and collaborative outcomes, highlighting the importance of tailoring interaction protocols to specific system goals. Additionally, we emphasized human-agent collaboration, categorizing interaction modes into task delegation, iterative instruction, and immersive partnerships. These frameworks illustrated varying levels of human involvement and underscored the potential of LLM-based agents to augment human productivity and creativity through nuanced collaboration.

Decision-making mechanisms were identified as another crucial design dimension. Centralized (dictatorial) decision-making was noted for its coherence and simplicity, whereas collective decision-making methods such as voting and debating, provide greater resilience and adaptability, particularly in uncertain or dynamically changing environments. The choice of decision-making strategy significantly impacts MAS performance, emphasizing the need for thoughtful selection based on context-specific requirements.

Lastly, we reviewed contemporary agent communication protocols, including MCP, ANP, Agora, IoA, A2A, and Agntcy. These emerging frameworks address critical issues of identity, security, interoperability, and dynamic protocol negotiation. While current protocols demonstrate substantial progress, we identified ongoing challenges related to semantic interoperability, scalability, and adaptability, highlighting promising directions for future research.

As MAS continues to evolve, several research avenues stand out:
\begin{itemize}
	\item \textbf{Context-Aware Collaboration:} Developing advanced frameworks enabling agents to dynamically select optimal interaction strategies based on real-time context and system conditions.
	\item \textbf{Scalable Adaptive Architectures:} Designing hybrid topologies that seamlessly combine centralized coordination with decentralized execution, optimizing scalability and responsiveness.
	\item \textbf{Human-Agent Co-evolution:} Enhancing immersive human-agent interactions through continuous learning mechanisms, enabling agents to better adapt to human preferences and workflows.
	\item \textbf{Adaptive Decision-Making:} Investigating methods that allow systems to dynamically transition between centralized and collective decision-making models based on the current environment and task complexity.
	\item \textbf{Dynamic Communication Protocols:} Exploring meta-protocol frameworks that enable agents to autonomously negotiate, compose, and evolve communication standards on-the-fly, significantly enhancing interoperability and semantic flexibility.
\end{itemize}
In conclusion, the thoughtful integration of agent composition, topological structures, interaction paradigms, decision-making frameworks, and robust communication protocols will be pivotal for developing effective, scalable, and adaptive MAS capable of tackling increasingly sophisticated real-world challenges.
\chapter{Collective Intelligence and Adaptation in Multi-Agent Systems}
\label{ch:MAS-evolution}

\lettrine[lines=3]{\initfamily\textcolor{darkgreen}{T}}{his} chapter explores the emergence and dynamics of \textit{collective intelligence} in multi-agent systems (MAS), emphasizing both the system-level capacity to solve complex tasks and the individual-level evolution of intelligent agents within social environments. Drawing from cognitive science, philosophy, and computational theory, we investigate how agent collectives, composed of Large Language Model (LLM)-based agents,achieve emergent behaviors, adaptive coordination, and self-organizing hierarchies beyond the capabilities of any single agent. We examine the structural mechanisms driving collective intelligence, including topology evolution, routing optimization, and memory-driven coordination, and how these mechanisms enable agents to engage in high-level social behaviors such as trust, deception, and cultural norm formation. At the same time, we analyze how shared memory, communication protocols, and reflective learning empower individual agents to adapt, specialize, and co-evolve within the broader agent society. Through theoretical insights and empirical case studies, this chapter positions collective intelligence not merely as a performance gain but as a foundational paradigm for understanding cognition, cooperation, and social emergence in artificial societies.

\section{From Collective Intelligence to Individual Evolution}
\label{sec:collective-intelligence}

\begin{figure}[!ht]
\centering
\includegraphics[width=0.7\textwidth]{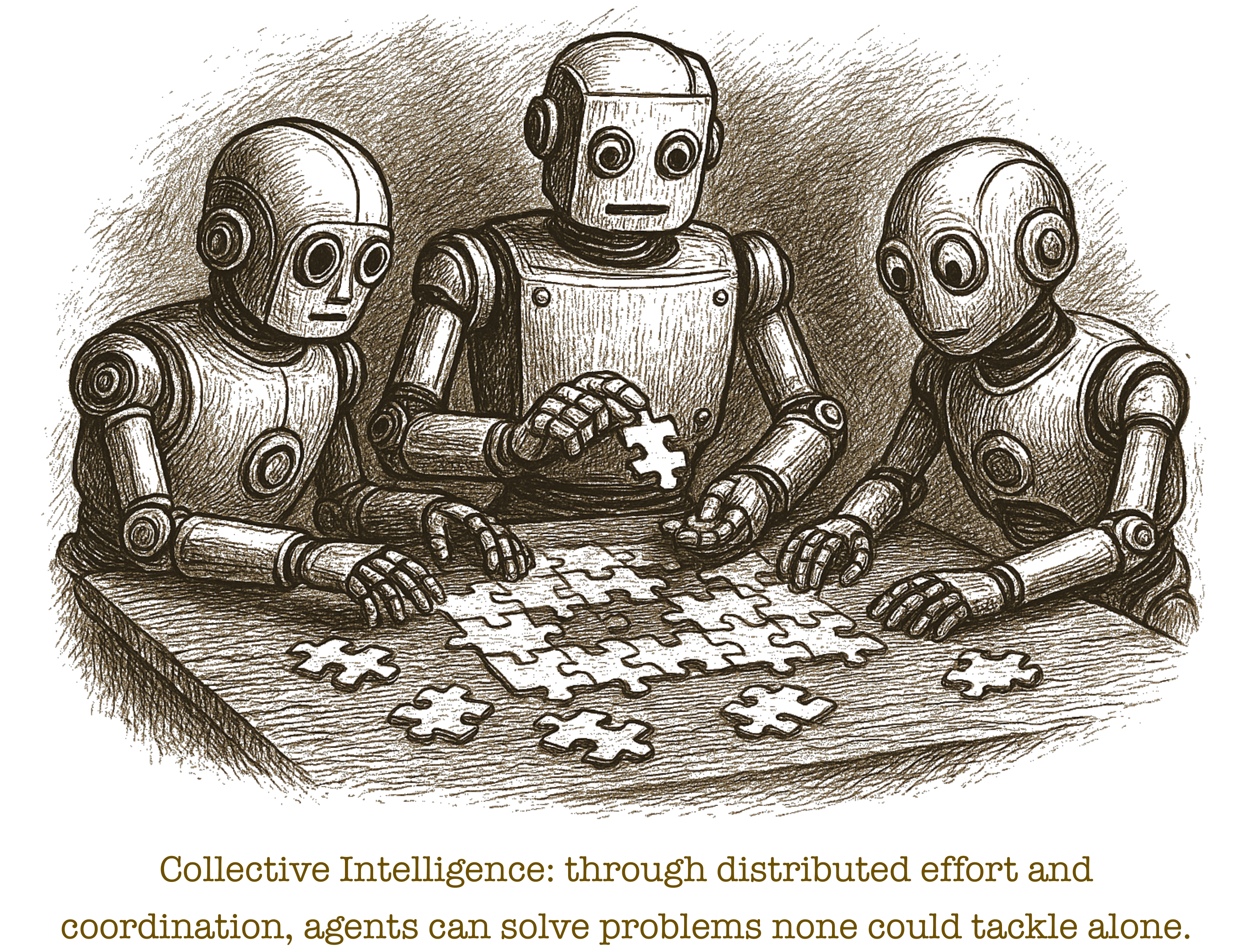}
\caption{Collective intelligence emerges when multiple agents coordinate, complement, and refine each other's reasoning to solve problems beyond individual reach.}
\label{fig:CI}
\end{figure}

\lettrine[lines=3]{\initfamily\textcolor{darkgreen}{M}}{ulti-agent systems (MAS)} derive inspiration from biological swarms and human societies, where group cooperation often yields capabilities far beyond those of any individual. A central idea is that of \emph{collective intelligence},the phenomenon where a community of agents collectively demonstrates problem-solving and decision-making prowess exceeding that of the single best agent, as shown in Figure~\ref{fig:CI}. This concept is famously captured by the ``Wisdom of Crowds''~\cite{surowiecki2005wisdom}, which asserts that under the right conditions, the many are smarter than the few. For instance, independent individuals can pool their diverse knowledge such that the aggregated decision outperforms any expert's judgment.

Minsky's \emph{Society of Mind} theory~\cite{minsky1988society} further hypothesizes that what we call intelligence may itself emerge from the interplay of numerous simpler components (``agents'') in the mind. By analogy, a MAS can exhibit intelligent behavior as an emergent property of interactions among relatively simple agents. Recent cognitive studies on \emph{Theory of Mind}~\cite{frith2005theory,li2023theory} suggest that even artificial agents can start to reason about each other's mental statesa capability crucial for sophisticated social reasoning. Taken together, these insights paint a picture in which intelligence springs from synergy: like neurons forming a brain or ants forming a colony, coordinated agents can create an intelligent whole greater than the sum of its parts.

In human societies, \emph{collective intelligence} arises naturally as people collaborate, divide labor, and solve problems together. Likewise, MAS can harness distributed knowledge and complementary skills of specialized agents to tackle complex tasks beyond any single agent's reach~\cite{li2024survey}. Importantly, collective intelligence is not simply the sum of individual capabilities, but an emergent phenomenon, novel capabilities and patterns of behavior materialize from interactions. Through ongoing communication and feedback, the group develops shared understanding and collective memory, enabling more coherent and effective coordination over time.

Notably, heterogeneity among agents, which includes differences in knowledge, perspective, or strategy, often strengthens the collective: diversity helps the group avoid the pitfalls of individual bias or ``groupthink'', leading to more robust and unbiased solutions~\cite{li2024survey}. In essence, a well-designed agent society can tap into a \emph{wisdom of crowds} effect, where each agent's unique insight contributes to a better global outcome.

While the group evolves its own intelligence, the individual agents within it are not static. Each agent can learn and adapt through experience, a concept we term \emph{individual adaptability}. In an MAS, an agent's adaptability refers to its capacity to adjust behavior, strategies, or goals based on past interactions and feedback from the environment. This continual refinement is akin to how a person learns from life experiences or how animals adapt to new conditions.

Recent research even describes agents as \emph{self-evolving}, which is able to modify their objectives or update their internal models on the fly in response to feedback~\cite{li2024survey}. The advent of large language models (LLMs) as agent brains has accelerated this process: LLM-powered agents come equipped with powerful general knowledge and can leverage dynamic memory modules to reflect on interactions, monitor their own performance, and self-correct. By exchanging information with others and storing rich histories of interactions, agents can improve over time without requiring an external programmer to intervene at each step.

We can categorize an agent's learning mechanisms into \emph{memory-based adaptation} (which does not alter the agent's underlying model parameters, instead relying on stored experience and on-the-fly reasoning) and \emph{parameter-based learning} (which involves updating the agent's model or policy through training or fine-tuning). These two forms of adaptation, roughly analogous to learning within a lifetime versus evolving across generations, work in tandem to make individual agents progressively more competent as they participate in an agent society.

In the sections that follow, we delve deeper into how \emph{collective intelligence} emerges in MAS and how it can be enhanced, and then explore how living in a society of agents drives the adaptation and evolution of individual agents. We will see that the growth of group intelligence and individual learning are deeply intertwined, yielding a fascinating co-evolutionary dynamic: the group makes the individuals smarter, and smarter individuals, in turn, further improve the group.

\section{Collective Intelligence in Multi-Agent Systems}
\label{sec:collective_intelligence}

\lettrine[lines=3]{\initfamily\textcolor{darkgreen}{I}}{n} this section, we clarify what ``collective intelligence'' means for multi-agent systems, tracing how higher-order capabilities emerge from local interactions among diverse, communicating agents. We first ground the concept theoretically (e.g., incentive alignment and COIN-style reward design), then show empirically how collaboration boosts performance and robustness. We next distill practical strategies, topology evolution, coordination/routing optimization, and shared memory, to deliberately foster such emergence. Finally, we examine the social layer: norms, deception, roles, and culture that spontaneously arise as agents co-adapt over time.

\subsection{Concept and Emergence of Collective Intelligence}
\label{subsec:CI-concepts}

Collective intelligence in MAS refers to the capacity of a group of agents to solve problems or make decisions collaboratively with an efficacy that surpasses that of any single agent~\cite{gao2024large}. This concept implies emergent behavior: through interaction, the group exhibits new capabilities and higher-order reasoning that were not explicitly programmed into the individual agents. A classic example is a flock of birds or a school of fish executing complex coordinated maneuvers without any bird or fish explicitly ``in charge''. In the context of artificial agents, we similarly observe that simple local rules or interactions can yield sophisticated global behavior. For instance, agents can form consensus decisions, perform division of labor, or collectively explore a solution space in ways that no lone agent could. The key ingredients for such emergence often include decentralized control, diversity among agents, and mechanisms for communication and feedback. When each agent contributes independent information or perspective, and there is a means to aggregate these contributions, the resulting group decision tends to be more informed and accurate than one made in isolation. Surowiecki (2005) identified factors like diversity of opinions, independence of members, and a reliable method of aggregation as critical to wise crowds – principles that likewise guide the design of intelligent MAS~\cite{li2023theory}.

A theoretical foundation for understanding emergent collective intelligence is provided by Wolpert and Tumer's Collective INtelligence (COIN) framework, which considers a large MAS with no central controller and a global utility function measuring overall system quality. The challenge is to design local reward functions so that, by maximizing their own rewards, agents still improve the global utility, avoiding tragedies of the commons through incentive alignment. Appropriate reward shaping has been shown to yield near-optimal collective behavior in domains such as network routing and resource allocation, underscoring how coordination mechanisms (reward design, norms, protocols) transform a mere collection of agents into a coherently intelligent unit~\cite{gao2024large}.

Notably, recent work has demonstrated that LLM-based agents possess ingredients conducive to collective intelligence: broad world knowledge, reasoning ability, and rich dialogue competence that allow them to ``think together''. Teams of LLM-driven agents have even exhibited higher-order Theory of Mind, where reasoning not only about tasks but also about peers' beliefs and intentions, thereby boosting coordination in complex social scenarios~\cite{li2023theory}. In summary, many minds can outthink one, provided agents communicate, specialize, and learn from each other. This group intellect emerges dynamically and iteratively: ongoing interactions yield shared knowledge, common strategies, and joint behavior patterns that outperform isolated efforts~\cite{gao2024large}.

\subsection{Enhanced Performance through Collaboration}
\label{subsec:collabo-enhance-perform}

One of the clearest benefits of collective intelligence in MAS is improved system performance on complex tasks. When agents share information and divide subtasks, the group can leverage parallelism and expertise to reach better solutions faster than any lone agent~\cite{hong2023metagpt,tang2024medagents,Kim2024MDAgentsAA}. Empirical studies show that exchanging intermediate thoughts and cross-verifying conclusions yields higher accuracy in reasoning and decision-making tasks~\cite{zhang2023exploring}. Collaboration helps overcome blind spots and biases: erroneous assumptions by one agent can be detected by others; creative ideas can be iteratively refined. This distributed problem-solving produces a ``hive mind'' effect, high-quality, consistent outcomes emerging from complementary strengths~\cite{chen2023reconcile}.

Role-specialized frameworks make this concrete. MetaGPT coordinates LLM agents with distinct roles (e.g., product manager, engineer) for software design/implementation, yielding more robust, correct solutions than a single monolithic model~\cite{hong2023metagpt}. Similarly, medical MAS with doctor and specialist agents improved diagnostic accuracy via consensus building~\cite{tang2024medagents,Kim2024MDAgentsAA}. 

Collaboration also mitigates individual LLM pitfalls such as hallucinations or inconsistent reasoning. Debate-style prompting lets agents critique and reconcile outputs, increasing reliability~\cite{liang2023encouraging}. Diversity across agents (different base models or perspectives) reduces systemic bias, since alternative viewpoints can be proposed and aggregated into balanced conclusions~\cite{chen2023reconcile}.

However, naive groups may fall into herd mentality or conflict. Effective MAS therefore require careful design of communication protocols, division of labor, and incentive structures to maintain constructive diversity followed by coherent aggregation~\cite{zhang2023exploring}. Done right, the system becomes both smarter and more robust: if one agent fails, others compensate, enabling success in strategic games and real-world planning tasks previously unattainable for isolated AI.

\subsection{Strategies for Fostering Collective Intelligence}
\label{subsec:strategies-foster-CI}

Because collective intelligence is emergent, maximizing it involves tuning system structure, communication, and learning. Recent LLM-based MAS work explores three major directions:

\paragraph*{Evolving Collaboration Topologies}
Instead of fixing interaction patterns, evolutionary/search methods dynamically rewire who talks to whom. EvoFlow evolves collaboration patterns; VFlow and Flow optimize domain-specific workflows; DebFlow spawns new agents via debate to resolve disagreements. These approaches often discover compact, cyclic reasoning structures that outperform manual designs~\cite{zhang2025evoflow,wei2025vflow,niu2025flow,su2025debflow,dang2025multiagent}. EvoAgent uses genetic algorithms to generate/select diverse configurations (roles + connections), beating handcrafted baselines~\cite{yuan2024evoagent}. X-MAS explores heterogeneous LLM sets (different sizes/specialties) to exploit ensemble diversity~\cite{ye2025xmas}. MAS-GPT automates MAS construction via SFT; MaAS samples query-dependent systems via supernet optimization; Adaptive Graph Pruning jointly optimizes agent count and topology based on task demands~\cite{ye2025masgpt,zhang2025maas,li2025adaptive}.

\paragraph*{Optimizing Communication and Coordination}
Intelligent routing and dynamic orchestration matter even with fixed teams. MasRouter learns to forward queries/information to the most relevant agent, improving benchmarks by 1.8–8.2\% while cutting compute~\cite{ye2025masrouter}. FlowReasoner employs a meta-agent with RL to adapt coordination sequences to context~\cite{gao2025flowreasoner}. Reinforcement-learning based orchestration more broadly enables flexible collaboration policies that evolve with tasks/environments~\cite{dang2025multiagent}. New MARL advances include implicit consensus generation for efficient group planning~\cite{li2025efficient}, AIR's unification of individual and collective exploration~\cite{zhou2025air}, and STPE-MARL's evolutionary + GNN approach for ad hoc coordination in complex tasks~\cite{peng2025stpe}.

\paragraph*{Shared Memory and Knowledge Aggregation}
Information sharing is central to collective intelligence. SRMT (Shared Recurrent Memory Transformer) creates a global workspace: agents pool local observations/intermediate computations into a shared memory broadcast back to all, enabling coordination even with sparse feedback (e.g., lifelong pathfinding through narrow corridors)~\cite{sagirova2025srmt}. G-Memory offers a hierarchical three-tier memory (insight, query, interaction graphs) that evolves with experience, raising success rates in complex embodied reasoning by up to 20.9\% without changing individual agent architectures~\cite{yue2025smart}. Complementary work on trajectory collection/data synthesis shows how MAS can co-train via SFT and RL to optimize memory representations and strategies, enabling collaboration without explicit external rewards through active goal inference~\cite{zhang2025learning,zhang2025evoflow}.

\subsection{Emergent Social Behaviors and Evolution}
\label{subsec:emergent-social-behav}

Beyond task performance, MAS can exhibit spontaneous social dynamics such as trust, reciprocity, deception, leadership, and norms when agents interact repeatedly.

In Avalon-inspired hidden-role games, LLM agents formed alliances, coordinated strategies, and developed deception-detection tactics. Minority ``traitor'' agents colluded secretly; honest agents learned to spot lies via dialogue history and recursive reasoning (higher-order ToM)~\cite{xu2024exploring,du2024helmsman,jin2024learning,wang2023avalon,shi2023cooperation}. This arms race between deception and detection yielded sophisticated, unprogrammed communication strategies.

Cooperative settings also show organization emergence. In Project Sid, hundreds of LLM agents in a Minecraft-like sandbox evolved professions, trade networks, rules, and even cultural/religious practices, progressing from survival to governance~\cite{altera2024project}. 

Social norms are unwritten coordination rules that can self-organize. Ren et al. (2024) demonstrated norm creation, representation, spreading, evaluation, and compliance (CRSEC). Agents adopted greeting conventions and first-come-first-served resource rules, reducing conflict and smoothing joint plans~\cite{ren2024emergence}.

These behaviors hinge on memory and reflection. With long-term memory, agents build reputations, trust, or vendettas. In deception settings, agents used first- and second-order ToM to navigate misinformation; in hide-and-seek self-play, agents invented tools (locking doors, building ramps), driven by competitive autocurricula~\cite{xu2025mem}.

Philosophically, collective intelligence is process-driven and self-reinforcing. Agents not only solve immediate problems; they also shape the social context for future interactions. Culture is a shared repository of knowledge, norms, and identities that can form, increasing efficiency but also causing path dependence. Designing MAS that nurture beneficial emergent behaviors while curbing harmful ones is an open challenge at the intersection of AI, sociology, and ethics~\cite{ren2024emergence,altera2024project}.
In summary, collective intelligence in MAS is not only about raw task performance but also about the rich tapestry of emergent social structures. Continuous interaction turns simple rules into complex outcomes. Understanding and harnessing these dynamics is key to MAS that robustly tackle real-world complexity, just as human societies have evolved to do~\cite{gao2024large}.

\section{Individual Adaptation and Evolution of Agents}
\label{sec:individual_adaptation}

\lettrine[lines=3]{\initfamily\textcolor{darkgreen}{W}}{hile} a multi-agent society as a whole learns and adapts, each individual agent within it also undergoes development. Individual adaptability refers to an agent's ability to improve its behavior, strategies, or knowledge over time based on experience~\cite{guo2024large}. In essence, an adaptive agent is one that learns how to learn: it doesn't just execute a fixed program, but can modify itself, updating an internal plan, adjusting a goal, or even tuning its own model parameters, in response to feedback and new information~\cite{nascimento2023self}. This capacity is crucial in dynamic environments, where no static policy can be perfect. A classic human analogy is how a novice initially follows fixed rules, but gradually, through practice and reflection, becomes an expert who can handle novel situations with ease. We seek to imbue artificial agents with a similar growth trajectory.

\paragraph*{Memory-based Learning}
One straightforward pathway for an agent to adapt is through memory and learning from experience. An agent equipped with memory can record the outcomes of its actions and the states of the world it has seen~\cite{anokhin2024arigraph,xu2023magic,park2023generative}. By reflecting on these records, the agent can identify which strategies led to success and which to failure, and adjust future decisions accordingly. This is a training-free form of learning where the agent isn't necessarily changing its underlying LLM weights, but it is changing its internal state of knowledge~\cite{li2024agent,chen2024agentcourt}. For example, in clinical simulations, doctor agents continuously improve treatment performance by accumulating experience from both successful and unsuccessful cases~\cite{li2024agent}. In social behavior simulations, agents enhance adaptability by engaging in more complex scenarios and leveraging scenario memories~\cite{park2023generative}. Similarly, Long-Term Memory systems that integrate episodic and semantic memory enable sustained personal performance gains~\cite{jiang2024longterm}. 

This kind of memory-based learning can be further amplified by deliberate reflection. Just as a person might keep a journal or mentally replay the day's events to learn from them, an agent can include a module that periodically analyzes its memory to extract lessons or adjust internal plans~\cite{nascimento2023self}. Recent frameworks often implement a reflective loop in which, after completing a task, the agent asks itself: ``What went well? What went wrong? What can I do better next time?'' Nascimento et al. (2023) describe a self-control loop where each agent monitors its own performance, self-critiques, and autonomously adapts to changing environments~\cite{nascimento2023self}. For instance, an agent might notice it repeatedly failed a subtask when a certain condition occurred; it could then create a new heuristic (``If condition X, try approach Y first''), effectively updating its policy through introspection.

\paragraph*{Shared Memory-based Learning}
Agents in a society can also learn from each other. Instead of each agent learning in isolation, agents can pool experiences via shared memory or communication-based learning~\cite{Lan2023LLMBasedAS,stepputtis2023long,wang2023avalon}. ProAgent exemplifies this: agents continuously exchange intentions and observations, using the communication log itself as a learning signal to anticipate teammates' actions and dynamically adjust plans~\cite{zhang2023proagent}. In essence, each agent builds a mental model of others from communications, enabling a team ``mind-meld'' that improves coordination and accelerates individual adaptation. 

Memory-driven communication protocols further enhance individual coordination, combining private observations with collective memory access to refine personal decision policies~\cite{pesce2020improving,sagirova2025srmt}. Advanced hierarchical memory architectures let individual agents selectively query structured collective knowledge. G-Memory's three-tier shared memory (insight, query, interaction graphs) enables bi-directional traversal so each agent can incorporate relevant team experiences, yielding a 20.89\% improvement in complex embodied tasks through cross-trial learning~\cite{zhang2025gmemory}. Collaborative Memory provides multi-user sharing with dynamic access control~\cite{rezazadeh2025collaborative}. SRMT shows how pooled working memories let agents implicitly exchange information while maintaining autonomy~\cite{sagirova2025srmt}. MD-MADDPG similarly demonstrates improved individual coordination via collective memory in cooperative scenarios~\cite{pesce2020improving}. 

\paragraph*{Parameter-based Learning}
Beyond memory-based approaches, many MAS employ parameter-based learning to evolve agents' adaptability via post-training techniques. The Learning through Communication (LTC) paradigm logs inter-agent dialogues and reuses them as supervised training data, fine-tuning LLMs toward effective communication and collaboration patterns~\cite{wang2023adapting}. Recent work emphasizes multi-agent (co-)fine-tuning: debate fine-tuning~\cite{subramaniam2025multiagent}, SiruiS~\cite{zhao2025sirius}, and Sweet-RL~\cite{zhou2025sweet} (which reinforces a critic model to guide better collaborative reasoning) all inject higher-level cooperation into model weights. However, future parameter-based paradigms must balance general capabilities and role specialization to avoid overfitting agents to specific partners or scenarios. Hybrid neurosymbolic and modular designs, e.g., converging symbolic and connectionist paradigms or adding trainable adapters, aim to preserve LLM versatility while optimizing MAS-specific components~\cite{xiong2024converging,jiang2024neuron}. 

In practical terms, enabling individual agents to evolve within an MAS creates a self-reinforcing loop: as agents improve individually (through memory or fine-tuning), the group performs better; the stronger group tackles harder problems, generating richer interaction data; that data, in turn, refines individuals further. Over time, an MAS can bootstrap itself from simple beginnings into a highly intelligent assembly, mirroring how human experts and societies co-evolve through shared experience and cumulative knowledge.

\section{Summary and Discussion}
\label{sec:summary-CI}

\lettrine[lines=3]{\initfamily\textcolor{darkgreen}{T}}{he} exploration of collective intelligence and individual adaptation in MAS reveals a profound parallel with biological and social evolution. A community of agents adapts as a whole, developing emergent competencies and social structures, while each member simultaneously grows through participation. The essence of collective intelligence lies in this interplay: intelligence is distributed across interactions, not confined to a single mind. Likewise, an agent's evolution is shaped by the social and informational ecosystem it inhabits.

Philosophically, one might ask: where does the ``mind'' of a multi-agent system reside? Evidence suggests it lives in the spaces between agents such as in conversations, shared memories, and coordinated actions. No single agent may hold a complete solution, but together they manifest one. This reframes intelligence as an emergent system property. Understanding this is key for AI research and also illuminates collective phenomena in human cognition (e.g., group decision-making, markets, scientific communities). 

Conversely, the adaptive agent highlights intelligence's plasticity. An agent's ``mind'' is not static code; it rewrites itself, influenced by collective experiences. Each agent resembles a neuron adjusting its connections (weights) based on activity, enabling the larger brain (the MAS) to function better, and the brain's patterns in turn shape the neuron. The collective and the individual co-evolve.

Designing future AI will likely require cultivating both levels: swarms that think together and self-improving agents that learn from experience. Such systems could tackle challenges overwhelming for individual AI by decomposing tasks and learning from vast interaction data. Imagine AI societies managing smart grids, optimizing traffic, collaborating in large organizations, or exploring virtual worlds; their success will hinge on harnessing collective wisdom while enabling each agent to grow.

The journey toward true collective intelligence in MAS has just begun. We already see agents forming norms, exhibiting Theory of Mind, deceiving and trusting, teaching and learning. As we scale interactions, artificial cultures and institutions with rich dynamics may arise. This raises pressing questions: How do we steer these societies toward human-aligned outcomes? Which emergent behaviors should to encourage or prevent? How to monitor safety and ethics in evolving swarms? These echo challenges faced with human collectives, calling for interdisciplinary approaches.

In conclusion, this chapter has shown how MAS forms collective intelligence to solve complex tasks and how society fosters the evolution of stronger individuals. Intelligence in MAS is a collective phenomenon nurtured by interaction; adaptation is a personal journey enriched by social context. The two reinforce each other: smarter collectives create better learning conditions for individuals, and more capable individuals elevate the collective. By embracing this synergy, we move toward AI that not only acts intelligently alone but learns to be intelligent together, a potential hallmark of the next era of AI research and applications.
\chapter{Evaluating Multi-Agent Systems}
\label{ch:MAS-evaluation}

\lettrine[lines=3]{\initfamily\textcolor{darkgreen}{T}}{he} transition from single-agent to multi-agent systems, and specifically Large Language Model (LLM)-based systems, requires a paradigm change in the evaluation paradigm. In contrast to single-agent evaluation, in which the immediate concern is performance on a particular task, evaluation of LLM-based multi-agent systems must be understood in terms of inter-agent dynamics as a whole, such as collaborative planning and communication effectiveness. Both task-oriented reasoning and holistic capability evaluation are addressed in this chapter, reflecting the nuance of such evaluations.
In greater detail, there are two main areas that we examine for evaluation. First, there is task-solving Multi-Agent Systems (MAS), where we examine benchmarks assessing and enhancing LLM reasoning for coding, knowledge, and mathematical problem-solving tasks.
These tests also accentuate the utility of distributed problem solving, achieved through organized workflows, specialisation among agents, iterative improvement, and calls for additional tools. Enhanced reasoning, primarily because of agent-agent decision-making cooperation and multi-round communications, is shown for MAS compared with agent-based individual ones.
Following that, there is a general evaluation of MAS abilities, extending beyond one-task-oriented achievement, to agent interactions at a highly advanced level. It involves a move away from one-dimensional measurements into multi-dimensional frameworks for documenting achievements at collaborations, reasoning abilities, system efficiency, and flexibility. We categorize such measurements into collaboration-oriented and competition-oriented measurements and have identified efficiency, decision-making quality, quality of collaboration, and flexibility as primary measure domains. These measurements capture various aspects of agent behavior, including communication effectiveness, resource distribution, and response to dynamic situations.

\begin{table*}[t!]
\centering
\caption{MAS Benchmarks: A Systematic Classification of Multi-Agent System Evaluation Frameworks Categorized by Task-Oriented Performance and System-Level Capabilities. This comprehensive collection encompasses both specialized task-solving benchmarks and holistic capability assessments, reflecting the dual nature of MAS evaluation in collaborative problem-solving and inter-agent dynamics. }
\label{tab:merged-mas-benchmark}
\resizebox{\textwidth}{!}{
\begin{tabular}{
    >{\centering\arraybackslash}p{2.4cm}  
    >{\centering\arraybackslash}p{3.0cm}  
    >{\centering\arraybackslash}p{6.8cm}  
    >{\centering\arraybackslash}p{3.6cm}  
    >{\centering\arraybackslash}p{4.0cm} 
}
\toprule
\textbf{Category} & \textbf{Focus} & \textbf{Benchmarks} & \textbf{Examples} & \textbf{Representative Metrics} \\
\midrule

\multirow{17}{*}{Task-solving} 
& Code Reasoning 
& APPS~\cite{dataset-apps}, HumanEval~\cite{dataset-human-eval}, MBPP~\cite{dataset-mbpp}, CodeContest~\cite{dataset-code-contest}, 
MTPB~\cite{dataset-mtpb}, DS-1000~\cite{dataset-ds-1000}, ODEX~\cite{dataset-odex}, Raconteur~\cite{deng2025raconteur}
& MetaGPT~\cite{hong2023metagpt}, SWE-agent~\cite{yang2024sweagentagentcomputerinterfacesenable}, 
AgentCoder~\cite{huang2024agentcoder}
& Pass@k, Resolved(\%) 
\\[1.3em]

& Knowledge Reasoning
& ARC~\cite{dataset-arc}, HotpotQA~\cite{dataset-hotpot-qa}, CSQA~\cite{dataset-csqa}, StrategyQA~\cite{dataset-strategy-qa}, 
BoolQ~\cite{dataset-boolq}, OpenBookQA~\cite{dataset-openbook-qa}, WinoGrande~\cite{dataset-wino-grande}, 
HellaSwag~\cite{dataset-hella-swag}, SIQA~\cite{dataset-siqa}, PIQA~\cite{dataset-piqa}, proScript~\cite{dataset-pro-script}, 
ScienceQA~\cite{dataset-science-qa}, ProOntoQA~\cite{dataset-pro-onto-qa}
& Reflexion~\cite{Shinn2023ReflexionLA}, MASTER~\cite{gan2025mastermultiagentllmspecialized}
& Accuracy
\\[1.3em]

& Mathematical Reasoning
& MATH~\cite{dataset-math}, GSM8K~\cite{dataset-gsm8k}, SVAMP~\cite{dataset-svamp}, MultiArith~\cite{dataset-multiarith}, 
ASDiv~\cite{dataset-asdiv}, MathQA~\cite{dataset-mathqa}, AQUA-RAT~\cite{dataset-aqua-rat}, MAWPS~\cite{dataset-mawps}, 
DROP~\cite{dataset-drop}, miniF2F~\cite{dataset-minif2f}
& MACM~\cite{lei2024macmutilizingmultiagentcondition}, Debate~\cite{du2023improving}
& Accuracy, Pass@k
\\
\midrule

\multirow{6}{*}{Collaboration}
& Communication-based Cooperation
& InformativeBench~\cite{liu2024autonomousagentscollaborativetask}, LLM-Coordination~\cite{agashe2024llmcoordination},
COMMA~\cite{ossowski2025commacommunicativemultimodalmultiagent},
& iAgents~\cite{liu2024autonomousagentscollaborativetask}, Two-Player~\cite{liu2024largelanguagemodelsagents}, 
EAAC~\cite{kim2024ethereum}
& Completion rate,  Efficiency
\\[1.3em]

& Planning and Coordination
& PARTNR~\cite{chang2024partnrbenchmarkplanningreasoning}, VillagerBench~\cite{dong2024villageragentgraphbasedmultiagentframework}, 
BABYAGI-ARENA~\cite{babyagi-arena2023}, Multiagent Bench~\cite{zhu2025multiagentbenchevaluatingcollaborationcompetition}
& AAS~\cite{jeyakumar2024advancing}, ResearchTown~\cite{yu2024researchtownsimulatorhumanresearch}, 
GPTSwarm~\cite{zhuge2024gptswarm}
& Success rate, Efficiency
\\[1.3em]

& Process-oriented
& Auto-Arena~\cite{zhao2024autoarenaautomatingllmevaluations}
& Idea~\cite{gao2025graphaiideasleveraging}
& Completion rate, Step efficiency
\\
\midrule

\multirow{6}{*}{Competition} 
& Adversarial Scenarios
& BattleAgentBench~\cite{wang2024battleagentbench}, MAgIC~\cite{xu2024magicinvestigationlargelanguage}, 
LLMArena~\cite{chen2024llmarenaassessingcapabilitieslarge}, PokerBench~\cite{zhuang2025pokerbenchtraininglargelanguage},
Multiagent Bench~\cite{zhu2025multiagentbenchevaluatingcollaborationcompetition}
& Dilemma~\cite{fontana2024nicerhumanslargelanguage}, PokéLLMon~\cite{hu2024pokellmonhumanparityagentpokemon}
& Win rate, Elo rating
\\[1.3em]

& Social Deduction
& AvalonBench~\cite{light2023avalonbench}, Human Simulacra~\cite{xie2025humansimulacrabenchmarkingpersonification}, 
Diplomacy~\cite{Mukobi2023WelfareDB}
& MA-KTO~\cite{ye2025multiagentktoreinforcingstrategic}, HLR~\cite{xie2024humanlikereasoningframeworkmultiphases}, 
& Win rate, Accuracy of deductions
\\[1.3em]

& Game-Theoretic
& AgentVerse~\cite{chen2023agentversefacilitatingmultiagentcollaboration}, 
ICP~\cite{wang2025learningcommunicateimplicitcommunication}
& WarAgent~\cite{hua2024warpeacewaragentlarge}
& Score, Win rate
\\

\bottomrule
\end{tabular}
}
\end{table*}

\section{Task-solving Benchmarks for MAS}
\label{sec:task-sovling evaluation}

\lettrine[lines=3]{\initfamily\textcolor{darkgreen}{I}}{n} multi-agent system solving for tasks, much focus has been on leveraging multi-agent coordination for enhancing the reasoning capacity of LLMs. It is most evident in coding, knowledge, and mathematical reasoning benchmarks, where one is interested in examining and building on performance with distributed solving. These benchmarks most typically examine if agents' capability for producing correct code, reasoning on complex knowledge domains, and solving difficult mathematical problems withstanding, with measures such as $pass@k$~\cite{dataset-minif2f} or proof ratios for success being prevalent. Much improvement has been exhibited by MAS through structured workflow, domain-specific agent roles, and iterative improvement on state-of-the-art performance. On the contrary, for model and simulation MAS, the case is one with a comparative lack of standardized benchmarks. Rather, research is primarily experimental setups that simulate a variety of social phenomena, with calls from the community for further formalized evaluation frameworks. These multiple benchmark areas are described below, examining the tasks, measures for evaluation, and the core mechanisms through which MAS result in better performance.

\paragraph*{Code Reasoning Benchmark} 
Measuring the capability of LLMs for code synthesis requires bespoke benchmark suites with a focus on functional correctness. Code synthesis, as compared to natural language synthesis, allows for direct verification through running. Several benchmark suites have been built for this purpose, typically consisting of a collection of programming problems, each described with a natural language problem description and a collection of test cases for automatically ascertaining the synthesized code's correctness. HumanEval~\cite{dataset-human-eval}, APPS~\cite{dataset-apps}, and MBPP~\cite{dataset-mbpp} are some popular ones. These benchmark suites predominantly utilize the $pass@k$ metric, which computes the percentage at which at least one among the top-$k$ generated solutions passes all test cases for a number of problems. The problems covered through these benchmark suites range across a variety of difficulties and programming abstractions, requiring not only for LLMs and Agents but also for syntactically correct and logically sound code that satisfies the provided test cases.
In contrast to existing coding benchmarks that often involve self-contained problems solvable with a few lines of code, SWE-bench~\cite{jimenez2023swe} introduces repository-level code challenges drawn from real-world GitHub issues and their corresponding pull requests across 12 popular Python repositories.
Beyond functional correctness, ensuring semantic accuracy is equally vital in tasks involving code generation and data interaction; for instance, in the context of Natural Language to SQL (NL2SQL) translation, the NL2SQL-BUGs~\cite{dataset-llmsql} benchmark specifically addresses the critical need for detecting and categorizing semantic errors, providing expert-annotated instances to evaluate models' ability to generate semantically correct SQL queries. Additionally, DEVAI~\cite{zhuge2024agent} proposes novel AI development automation benchmarks that employ a judge-agent mechanism to evaluate intermediate development processes automatically. 
Recent work leverages Multi-Agent Systems (MAS) to enhance Large Language Model (LLM) code reasoning. MetaGPT~\cite{hong2023metagpt} utilizes human-like Standard Operating Procedures (SOPs) within a multi-agent framework. By assigning roles across diverse domains and adopting an assembly-line approach, MetaGPT decomposes complex operations into sub-operations, achieving state-of-the-art performance on HumanEval and MBPP benchmarks. 
AgentCoder~\cite{huang2024agentcoder} is a three-agent system (programmer, test designer, test executor) focused on effective testing and auto-optimization. The test designer provides diverse test cases, and the test executor offers feedback for optimization. This collaborative workflow boosts coding efficiency, outperforming single-agent. 
These MAS approaches~\cite{zhang2024autocoderoverautonomousprogramimprovement, chen2024coderissueresolvingmultiagent} highlight multi-agent cooperation, organized workflows, and tailored interfaces as effective strategies for improving LLM code reasoning. 

\paragraph*{Knowledge Reasoning Benchmark}
To facilitate AI agents effectively acting in and understanding the world, robust knowledge reasoning abilities are essential. Benchmarks for this task assess an agent's ability to utilize factual knowledge and logical reasoning when answering challenging queries. Commonsense reasoning benchmarks such as CSQA~\cite{dataset-csqa} and StrategyQA~\cite{dataset-strategy-qa}, and scientific knowledge understanding is tested with ScienceQA~\cite{dataset-science-qa}.
The core challenge for agents is performing multi-step, chain-of-thought reasoning, stepwise logically progressing from input query to output answer. These tests concentrate on assessing how well a specific AI agent can apply a specific body of knowledge, one at a time, and reason out a problem.
Recent research has experimented with the use of LLMs-MAS for improving knowledge reasoning task performance, and they have achieved state-of-the-art accuracy. For example, MASTER~\cite{gan2025mastermultiagentllmspecialized}, a novel multi-agent system, employs a novel recruitment process for agents and communication protocol using the Monte Carlo Tree Search (MCTS) algorithm, and achieves 76\% accuracy on HotpotQA~\cite{dataset-hotpot-qa}. Reflexion~\cite{Shinn2023ReflexionLA}, a universal framework for bringing reasoning and acting together with language models, improves baseline by 20\% on HotpotQA. These strategies demonstrate the potential of multi-agent coordination for knowledge reasoning tasks. Besides, leveraging external tools, e.g., search engines, is also needed for improving knowledge reasoning capacity. Such integration is particularly helpful on applications such as TriviaQA~\cite{dataset-trivia-qa} where real-time information access is essential.

\paragraph*{Mathematical Reasoning Benchmark}
Mathematical reasoning evaluation for LLM-based multi-agent systems primarily relies on two categories of datasets. Mathematical problem-solving benchmarks include SVAMP~\cite{dataset-svamp}, GSM8K~\cite{dataset-gsm8k}, and MATH~\cite{dataset-math}. Theorem proving datasets such as PISA~\cite{dataset-pisa} and miniF2F~\cite{dataset-minif2f} assess agents' ability to generate well-formed mathematical proofs, with evaluation focusing on problem-solving accuracy and proof completion rates.
Multi-agent collaborative approaches have shown promising results on these benchmarks. The MACM~\cite{lei2024macmutilizingmultiagentcondition} framework utilizes a three-agent architecture with Thinker, Judge, and Executor components that decompose complex problems through modular cooperation. Multi-agent debate mechanisms ~\cite{du2023improving} involve multiple language model instances iteratively refining solutions through collaborative discussion, significantly improving accuracy. 
Recent reinforcement learning enhancements, particularly multi-turn online iterative direct preference learning with code interpreter feedback, have achieved substantial performance gains on GSM8K and MATH datasets.
However, there are several significant challenges. The primary limitation lies in inadequate datasets, particularly for multimodal mathematical reasoning, which suffer from quality issues, insufficient scale, and task diversity constraints. Additionally, format sensitivity poses substantial barriers for multi-agent systems operating in planner paradigms, as these systems require high toolchain integration, leading to increased system complexity, latency, and frequent format-related failures that can account for over 50\% of unsuccessful attempts~\cite{ye2025maslab}.

\paragraph*{Societal Simulation Benchmark}
Social simulation benchmarks are essential for evaluating multi-agent system performance and realism for simulating human behavior and social interactions based on LLMs.  Standardized sets and test cases for evaluating the agents' ability for interacting, communicating, and evolving within a simulated society are provided through the benchmarks. SOTOPIA~\cite{zhou2024sotopiainteractiveevaluationsocial} is used for evaluating social intelligence of LLM-based MAS. It is employed for evaluating agents' ability for conversing, understanding social cues, and building relationships with each other within a virtual society.  Another benchmark involves simulating propagation Gender Discrimination and Nuclear Energy~\cite{gao2023s3} topics on social networks. It evaluates agents' capabilities in modeling opinion dynamics, information dissemination, and social influence within large-scale social networks. Moreover, Multiagent Bench~\cite{zhu2025multiagentbenchevaluatingcollaborationcompetition} further provides two simulation scenarios: werewolf and bargaining, to assess competitive interactions among diverse agent groups with conflicting goals.

\section{Collaboration \&  Competition Evaluation for MAS}
\label{sec:collaborative-competition evaluation}

\lettrine[lines=3]{\initfamily\textcolor{darkgreen}{E}}{v}aluating capabilities in LLM-based MAS requires specialized approaches that effectively measure the rich interactions between agents. As this field evolves, evaluation methodologies have transitioned from single-dimension metrics to multi-faceted evaluation frameworks that capture the complex skillset required for effective multi-agent interaction. This evolution reflects a growing understanding that agent performance must be assessed across multiple dimensions including collaboration success, reasoning capabilities, and system efficiency.
In recent research, the MAS evaluation can be mainly categorized along three primary dimensions: \textbf{collaboration-focused} benchmarks, \textbf{competition-focused} benchmarks, and \textbf{adaptive and resilience} benchmarks. Within each category, we identify specific metric domains that capture different aspects of agent performance. Current evaluation approaches typically measure efficiency metrics (e.g., task completion rates, resource utilization, time efficiency), decision quality metrics (e.g., action accuracy, strategic soundness, reasoning depth), collaboration quality metrics (e.g., communication effectiveness, coordination efficiency, workload distribution), and adaptability metrics (e.g., response to disruptions, self-correction), which provide a foundation for evaluating multi-agent systems.

\paragraph*{Collaboration-focused Benchmarks}
Collaboration-focused benchmarks have evolved significantly, shifting from basic single-dimensional metrics toward comprehensive frameworks that evaluate complex agent-to-agent communication and coordination. Initial benchmarks, such as InformativeBench~\cite{liu2024autonomousagentscollaborativetask}, primarily addressed agent collaboration under conditions of information asymmetry, employing metrics like Precision and IoU to measure decision accuracy in information dissemination tasks. Subsequently, the scope of evaluation expanded, exemplified by Collab-Overcooked~\cite{sun2025collabovercookedbenchmarkingevaluatinglarge}, which introduced nuanced process-oriented metrics such as Trajectory Efficiency Score (TES) and Incremental Trajectory Efficiency Score (ITES). These metrics assess detailed aspects of coordination, revealing significant shortcomings in agents' proactive planning and adaptive capabilities despite their strong task comprehension. Further expanding the evaluation scope, COMMA~\cite{ossowski2025commacommunicativemultimodalmultiagent} and LLM-Coordination~\cite{agashe2024llmcoordination} emphasized communication effectiveness and strategic synchronization, employing diverse environments and extensive metrics including Success Rate, Average Mistakes, and Environment Comprehension Accuracy. These benchmarks collectively illustrate an emerging trend toward capturing deeper aspects of collaborative behaviors and strategic consistency. Other benchmarks, such as PARTNR~\cite{chang2024partnrbenchmarkplanningreasoning}, VillagerBench~\cite{dong2024villageragentgraphbasedmultiagentframework}, and BabyAGI~\cite{babyagi-arena2023}, further addressed gaps in existing evaluations by focusing explicitly on reasoning, planning, and task decomposition. These benchmarks highlighted the need for comprehensive assessment of agents' ability to engage in complex, socially embedded tasks, considering metrics like Percent Completion, Balanced Agent Utilization, and agent contribution rates. AgentBench~\cite{liu2023agentbench}, VisualAgentBench~\cite{liu2024vab}, and Auto-Arena~\cite{zhao2024autoarenaautomatingllmevaluations} further standardized multi-agent evaluations, automating assessment across various domains and demonstrating substantial performance disparities between closed-source and open-source LLMs. These observations underscored critical challenges in developing universally effective collaboration frameworks.
In summary, collaboration-focused benchmarks collectively reflect an ongoing shift toward comprehensive, nuanced evaluations that encompass communication efficiency, adaptive strategy, and fine-grained agent coordination, addressing earlier limitations focused solely on outcome-based performance.

\textbf{Competition-focused Benchmarks}
Competition-focused benchmarks evaluate agents' strategic capabilities and adversarial interactions, highlighting specific deficiencies in Theory of Mind and opponent modeling. Early benchmarks such as 
BattleAgentBench~\cite{wang2024battleagentbench} and MAgIC~\cite{xu2024magicinvestigationlargelanguage} initiated the focus on mixed cooperative-competitive environments, uncovering critical weaknesses in high-order strategic reasoning among LLM agents. These benchmarks employed comprehensive competitive metrics such as Forward Distance, Judgment Accuracy, and Rationality scores, identifying that while advanced LLMs performed adequately in simpler scenarios, significant limitations persisted under complex adversarial conditions. Building upon these insights, subsequent benchmarks like Human Simulacra~\cite{xie2025humansimulacrabenchmarkingpersonification}, LLMArena~\cite{chen2024llmarenaassessingcapabilitieslarge}, and PokerBench~\cite{zhuang2025pokerbenchtraininglargelanguage} further refined competitive evaluation by incorporating human-like reasoning metrics and more robust strategic measures (e.g., Response Similarity Score, Elo Scores, and Action Accuracy). These evaluations consistently demonstrated shortcomings in opponent prediction, risk assessment, and adaptive strategic planning, despite high task comprehension. Moreover, social deduction and deception-based benchmarks, notably AvalonBench~\cite{light2023avalonbench} and Diplomacy~\cite{Mukobi2023WelfareDB}, further revealed fundamental gaps in agents' abilities to interpret hidden information and manage complex social dynamics. Metrics like Assassination Accuracy, Deduction Accuracy, and Win Rates emphasized that even sophisticated LLMs fail to replicate human-level reasoning in adversarial negotiation and hidden-information games. Additional game-theoretic evaluations, including Guandan~\cite{yim2024evaluatingenhancingllmsagent}, AgentVerse~\cite{chen2023agentversefacilitatingmultiagentcollaboration}, MultiAgentBench~\cite{zhu2025multiagentbenchevaluatingcollaborationcompetition}, and ICP~\cite{wang2025learningcommunicateimplicitcommunication}, introduced scenarios requiring strategic cooperation under incomplete information. These benchmarks reinforced previous findings on the necessity of enhanced Theory of Mind and predictive modeling capabilities. MultiAgentBench~\cite{zhu2025multiagentbenchevaluatingcollaborationcompetition} also introduces the KPI and coordination score to evaluate the competition of agents.
Collectively, competition-focused benchmarks highlight persistent strategic and reasoning limitations among LLM-based agents, underscoring the ongoing need to address critical gaps in adversarial modeling and strategic planning despite advancements in general reasoning and task execution capabilities.

\paragraph*{Adaptive and Resilience Benchmarks}
Adaptive and resilient multi-agent system benchmarks tackle two inter-connected capabilities together: adaptability, which is the ability of the agents to act dynamically in altering, unexpected environmental conditions by modifying their behavior and strategy. Resilience, or the ability of the system to endure, alleviate, and rapidly recover from disruptions, faults, or hostile intervention. In adaptability, as mentioned in AdaSociety~\cite{huang2025adasocietyadaptiveenvironmentsocial}, the dynamic interplay between social relationships and physical environments demands that agents engage in continuous learning, and strike a balance between environment discovery and social network construction. Despite significant advancements in current multi-agent decision-making frameworks, these environments fall short in introducing new challenges in various physical contexts and changing social interdependencies. Therefore, AdaSociety introduces an environment in which physical states, tasks, and social relationships among agents continuously evolve, thereby capturing the adaptability of agents as they respond to expanding task complexity and shifting resource constraints.
Moreover, current benchmarks may oversimplify the challenges of real-world automation with limited disruption modeling and simplified dependencies of process~\cite{geng2025realmbenchrealworldplanningbenchmark}, resulting in insufficient evaluation of planning capabilities and adaptability. Thus,
REALM-Bench~\cite{geng2025realmbenchrealworldplanningbenchmark}, on the other hand, defines adaptation through real-world-inspired planning problems, which emphasizes metrics such as real-time re-planning efficiency, coordination scalability under increasing complexity, and the stability of performance outcomes despite dynamic interdependencies or disruptive events. Conversely, resilience benchmarks~\cite{huang2025resiliencellmbasedmultiagentcollaboration} systematically introduce faults or errors into individual agents to assess overall system robustness.

\section{Summary and Discussion}
\label{sec:challenge-future-work}

\lettrine[lines=3]{\initfamily\textcolor{darkgreen}{L}}{L}M-based Multi-Agent Systems (LLM-MAS) are highly promising for solving difficult problems and simulating real-world environments, but critical evaluation of their performance is a top priority challenge. Not only because of the inherent complexity of LLMs in general, but primarily because of difficulties in quantifying multi-agent cooperation, communication, and emergent behavior.

\paragraph*{Multimodal Environment Setup} The evaluation of LLM multi-agent systems is plagued with unprecedented challenges when this is extrapolated to multimodal environments, a stark departure from the existing text-based evaluation paradigms. While the majority of current LLM multi-agent systems are still operating in text-based environments, introducing visual, auditory, and other modalities introduces exponentially increasing complexity that fundamentally changes the landscape for evaluation. As metioned in COMMA~\cite{ossowski2025commacommunicativemultimodalmultiagent}, evaluation complexity in multimodal environments exhibits exponential growth curves, creating formidable technical challenges way beyond additive individual modality contributions.

\paragraph*{Non-deterministic Path Evaluation} The assessment of multi-agent systems faces a basic paradigmatic shift in coming to terms with the inherent non-determinism of agent interaction and decision-making. Conventional evaluation methods, based on deterministic input-output relationships, are insufficient when faced with systems that can fulfill the same goals through widely divergent but equally legitimate execution trajectories. The temporal aspect of multi-agent interaction adds complexity, in that path assessment needs to consider dynamic decision-making settings where optimal policies can change depending on environmental feedback, communication among agents, and emergent coordination patterns developing during the course of task execution. Path diversity measurement involves theoretical and applied challenges demanding new methodological solutions that support several legitimate solution tactics with differing resource demands and adaptability features. 
    
\paragraph*{Robustness and Resilience Evaluation} One of the core challenges for MAS evaluation is the measurement of system robustness against unforeseen inputs, noisy environments, and agent failures, such events that are unavoidable in real-world deployments yet hard to assess systematically. Evaluating an MAS's capacity to preserve performance or recover elegantly when one or several agents act unexpectedly or fail completely demands advanced evaluation frameworks with the capability to model diverse failure modes and quantify system resilience amidst varied types of disruptions~\cite{geng2025realmbenchrealworldplanningbenchmark, geng2025realmbenchrealworldplanningbenchmark}. Next-generation benchmarks need to include scenarios that test system resilience systematically while delivering standardized metrics for reliability measurement under adversarial or perturbed conditions, a step away from idealized testing environments and closer to realistic operational environments.

\paragraph*{Explainability and Interpretability of Collective Decision-Making} As MAS tackle more complex problems, visibility into the reasoning processes behind collective decisions becomes necessary, particularly in high-stakes applications where accountability and transparency are critical. Current evaluation frameworks only focus on final performance metrics without regard to the interpretability of agent interactions, communication flows, and collective reasoning processes that generate system outputs. Future evaluation methods will have to develop sophisticated methods for tracing decision processes, attributing agent contributions, and providing comprehensible explanations of collective intelligence emergence from distributed agent interactions. Furthermore, Decision pathway evaluation had to incorporate economic considerations and cost-effectiveness analysis. Decision pathway evaluation requires analysis of not only the logical soundness and correctness of group reasoning processes, but also the resource consumption patterns, communication overheads, and time complexity of different reasoning strategies. 

\paragraph*{Continuous Learning and Long-term Adaptation Evaluation} Although most MAS exhibit good short-term performance on particular tasks, their ability to undergo continuous learning, self-development, and adaptation to new, unexpected situations over long operational lifetimes is essentially unaddressed by present evaluation methodologies. Evaluating long-term adaptability entails assessment methods that can determine the extent to which an MAS can develop its behaviors and strategies through ongoing interaction with dynamic environments~\cite{huang2025adasocietyadaptiveenvironmentsocial} without permanent human oversight. This problem is one of developing assessment methodologies for open-ended learning situations in which systems must show not just early competence but also the potential for independent improvement and adaptation to emergent difficulties that could not be predicted during early system design. Alternatively, evaluation framework should adapt to MAS evolution, since they are inherently adaptive systems that continuously learn, evolve, and acquire new capabilities through environmental interaction and cooperative refinement. Consequently, the evaluation must proceed using frameworks with the capacity for creating assessment environments of growing sophistication in parallel with the systems being evaluated.

\part{Building Safe and Beneficial AI Agents}
\label{part-safety}
\begin{figure*}[!ht]
    \centering
    \includegraphics[width=0.8\textwidth]{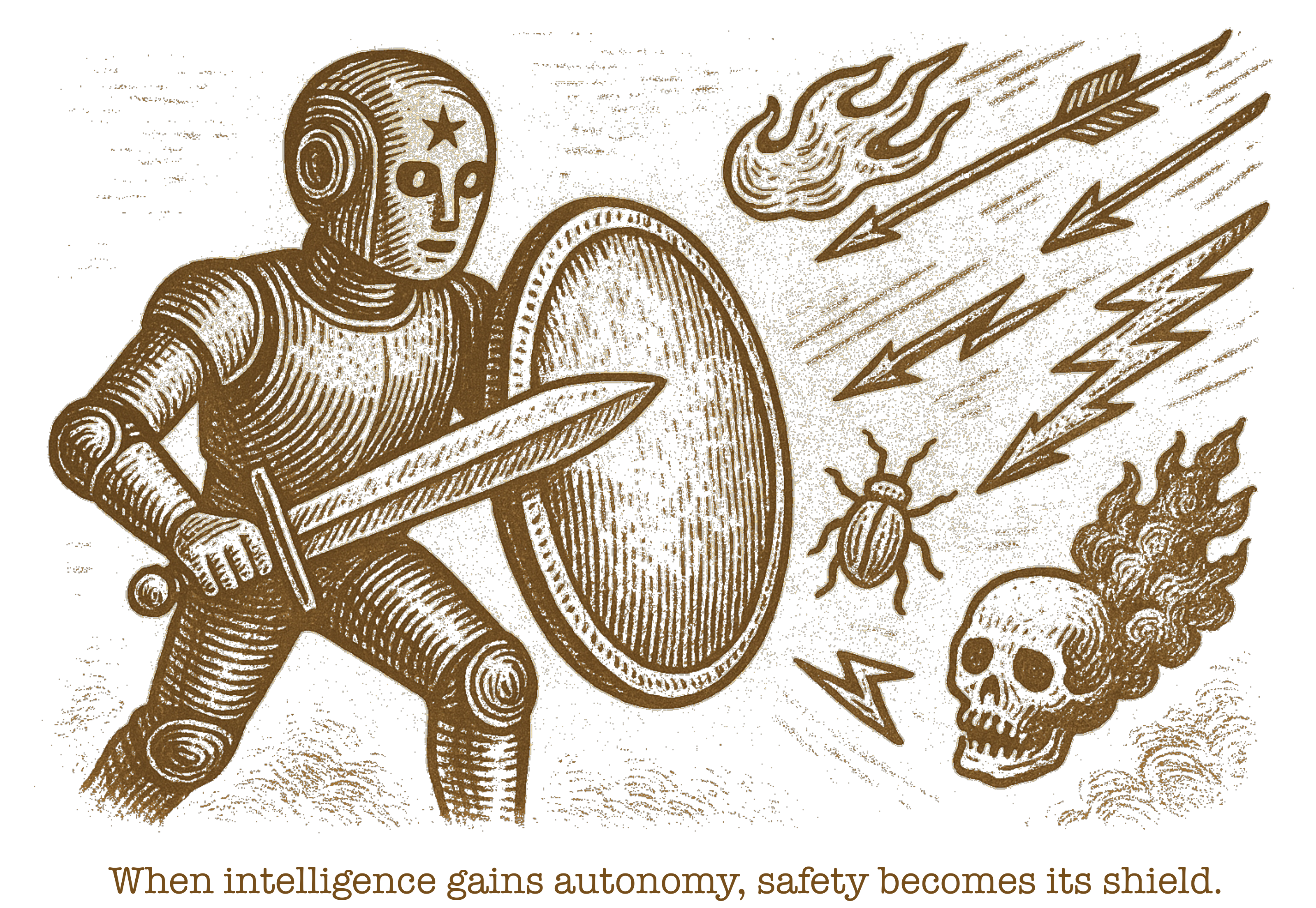}
    \label{fig:part4}
\end{figure*}

The rapid development of LLM-based agents introduces a new set of safety challenges that go beyond those of traditional LLMs. Equipped with advanced reasoning, planning, and tool-using capabilities, these agents are designed to perform tasks autonomously and interact with their environments \cite{xiRisePotentialLarge2023}. However, this autonomy also expands the attack surface, creating new vulnerabilities that demand careful research and attention \cite{deng2024ai,yu2025trustagent}\footnote{A Note on Terminology: In this survey, ``AI Safety'' is used as an umbrella term covering both unintentional corruption and intentional attacks~\cite{lin2025aisafetysecurity}. We will elaborate on this distinction in Section~\ref{sec:brain_discussion}.}.
In this part, we first establish a comprehensive framework for understanding agent safety, examining both internal and external safety threats to AI agents. We will explore the various attack vectors associated with these threats and propose potential mitigation strategies. This framework is organized into two key areas:

\textbf{(1) Intrinsic Safety} threats stem from vulnerabilities in the agent's core components, which include the LLM ``brain'' as well as the perception and action modules. Each of these components has unique weaknesses that can be exploited by adversaries:
\begin{itemize}
    \item \textit{Brain} is the LLM itself, responsible for key decision-making tasks such as reasoning and planning. It is guided by a knowledge module that provides essential contextual information.
    \item \textit{Perception} consists of sensors that interpret the external environment, where malicious manipulation of external objects can lead to erroneous perceptions.
    \item \textit{Action} is responsible for tool usage and downstream applications, which are also susceptible to exploitation.
\end{itemize}

\textbf{(2) Extrinsic Safety} threats arise from interactions between the agent and external, often untrusted, entities. These include:
\begin{itemize}
\item \textit{Agent-Memory Interactions}: The agent frequently accesses and interacts with memory storage, which serves as an external database for decision-making and contextual information retrieval. Recent research highlights vulnerabilities in the agent-memory interface that could be exploited to manipulate the agent's actions.
\item \textit{Agent-Agent and Agent-Environment Interactions:} These refer to the interactions between the agent and other agents (e.g., other agents or human operators), as well as its environment, which includes task-related objects or dynamic systems. The complexity of these interactions further compounds the agent's exposure to external threats.
\end{itemize}

\vspace{-10pt}
\begin{figure*}[h]
    \centering
    \includegraphics[width=0.9\linewidth]{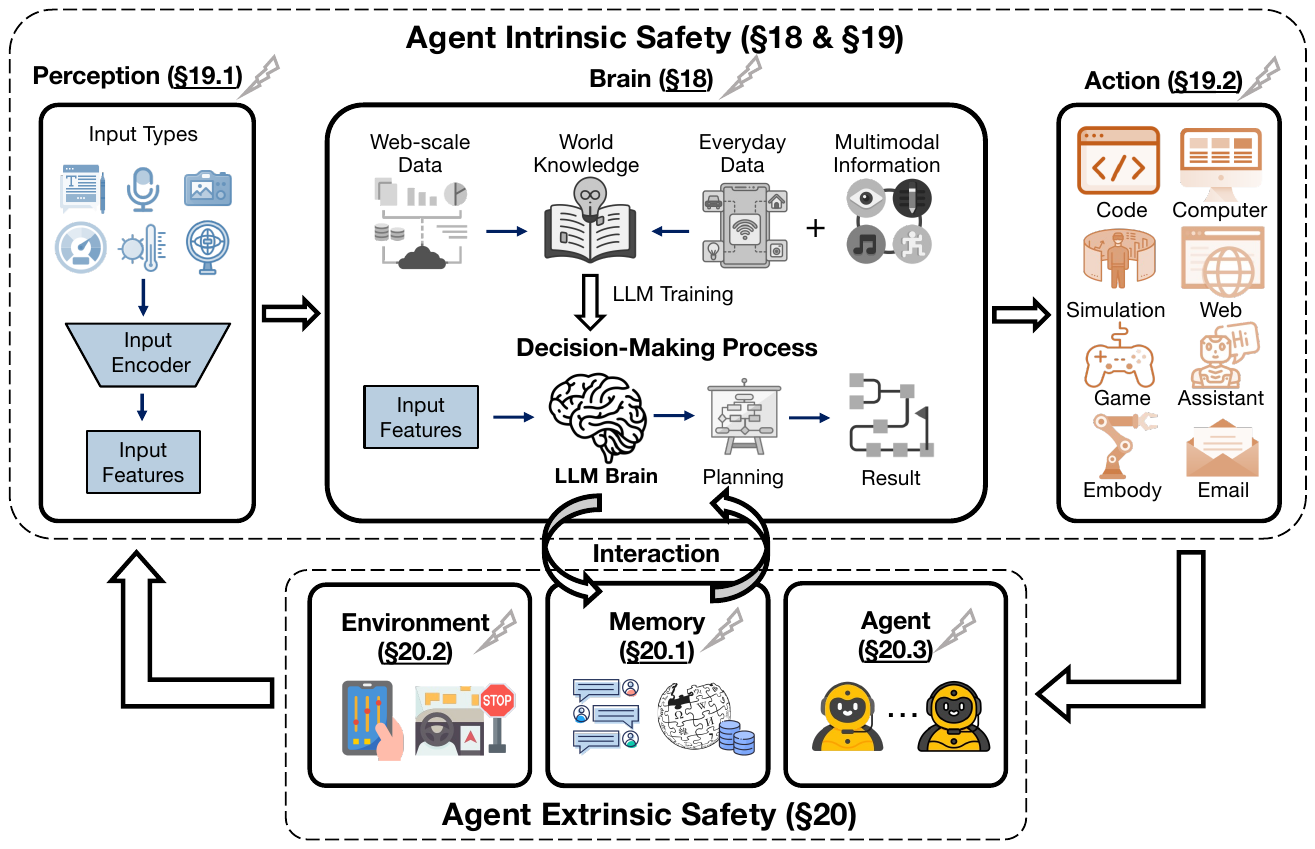}
    \caption{The Brain (LLM) faces safety threats like hallucination (\S~\ref{ssec:brain_sec_threat}) and privacy threats such as membership inference attacks (\S~\ref{ssec:brain_priv_threat}). Non-brain modules encounter perception threats (\S~\ref{ssec:percep_threat}) and action threats (\S~\ref{ssec:act_threat}). Due to interactions with potentially malicious external entities, we also explore agent-memory threats (\S~\ref{ssec:ex_mem_threat}), agent-environment threats (\S ~\ref{ssec:env_threat}), and agent-agent threats (\S~\ref{ssec:mulagent_threat}).
    }
    \label{fig:sec_overview}
\end{figure*}

As illustrated in Figure~\ref{fig:sec_overview}, these risks are broadly categorized into intrinsic and extrinsic safety, helping to clarify their origin and nature. In addition to identifying threats, we also provide a rigorous, mathematical foundation for understanding attacks such as jailbreaking, prompt injection, and data poisoning. Moreover, we present practical, actionable solutions, tracing the development of safety measures from early LLM safeguards to comprehensive strategies that protect the entire agent system. This includes exploring guardrails, advanced alignment techniques (such as superalignment), and the crucial balance between safety and helpfulness. 
Finally, we analyze the ``scaling law of AI safety'', the complex relationship between an agent's capabilities and its potential risks and the essential trade-offs that must be made. 
This part provides a clear understanding of the challenges, theoretical foundations, and practical strategies necessary to develop effective and trustworthy AI agents that can be safely and effectively deployed in real-world scenarios. 

This part is organized as follows: First, we examine intrinsic safety risks (Chapter~\ref{sec:in_brain_sec}), focusing on threats to the LLM ``brain,'' as well as vulnerabilities in the agent's perception and action components (Chapter~\ref{sec:in_other_threat}). Next, we explore extrinsic safety threats related to agent-memory, agent-agent, and agent-environment interactions (Chapter~\ref{sec:ex_threat}). 
Finally, we investigate superalignment techniques aimed at ensuring the safety of agent behaviors, while addressing the broader challenge of balancing safety with performance. This includes exploring how safety measures scale with the increasing capabilities of AI systems and examining the trade-offs involved in designing secure, capable AI agents (Chapter~\ref{sec:safety_law}).

\chapter{Agent Intrinsic Safety: Threats on AI Brain}
\label{sec:in_brain_sec}

\lettrine[lines=3]{\initfamily\textcolor{darkgreen}{T}}{his} chapter delves into the intrinsic safety of AI agents, focusing specifically on the vulnerabilities inherent in their core component: the Large Language Model (LLM), often referred to as the agent's ``brain''. As the central engine for decision-making, reasoning, and planning, the LLM's integrity is paramount. However, its complex architecture and reliance on vast datasets create a significant attack surface, exposing it to a spectrum of threats that can subvert its intended behavior and compromise its trustworthiness.

To provide a comprehensive overview, this chapter systematically categorizes these threats into two primary domains: safety vulnerabilities and privacy concerns. The safety analysis covers five critical areas: \textbf{Jailbreaking}, where safety alignments are bypassed; \textbf{Prompt Injection}, which hijacks the agent's control flow; \textbf{Hallucination}, leading to factually incorrect outputs; \textbf{Misalignment}, where agent behavior diverges from human intent; and Poisoning Attacks, which corrupt the model's integrity through malicious data. The privacy analysis explores threats related to \textbf{Training Data Inference}, including membership and data extraction attacks, and \textbf{Interaction Data Inference}, which involves the theft of sensitive system or user prompts. For each category, we provide formal definitions, analyze key attack methodologies, and discuss state-of-the-art mitigation strategies. By dissecting these vulnerabilities, this chapter underscores the critical need for a holistic, security-by-design approach to build inherently safer and more trustworthy AI agents.

\begin{figure*}[!ht]
\footnotesize
    \begin{forest}
        for tree={
            forked edges,
            draw,
            rounded corners,
            node options={align=center},
            s sep=2pt, 
            l sep=7pt, 
            calign=center,
            grow=east,
            reversed=true,
            anchor=base west,
            parent anchor=east,
            child anchor=west,
            base=left,
            font=\small,
            minimum width=1.5em,
            if n children=0{}{
                if level=1{text width=4em}{
                    if level=2{text width=7em}{
                        if level=3{yshift=0.26pt,fill=white!5,text width=9em}{}
                    }
                }
            },
          },
          where level=1{text width=3em,fill=customblue!50}{},
          where level=2{text width=6em,fill=customgreen!50}{},
          where level=3{yshift=0.26pt,text width=34em}{},
        [Agent\\\textbf{Intrinsic}\\Safety\\on \textbf{Brain}\\\textbf{(LLM)}, fill=gray!20
            [Safety\\Threats
                [Jailbreak
                    [\textbf{White-box Jailbreak}\\ 
                        Yi et al.\cite{yi2024jailbreak} GCG\cite{zou2023universal} MAC\cite{zhang2024boosting} I-GCG\cite{jia2024improved}  
                        Luo et al.\cite{luo2024jailbreak} 
                        Li et al.\cite{li2024open} 
                        DROJ\cite{hu2024droj} 
                        AutoDAN\cite{liu2023autodan} 
                        POEX\cite{lu2024poex}
                    ]
                    [\textbf{Black-box Jailbreak}\\
                        Wei et al.\cite{wei2023jailbroken}
                        PAIR\cite{chao2023jailbreaking} JAM\cite{jin2025jailbreaking}Qi et al.\cite{qi2024visual} POEX\cite{lu2024poex} AutoDAN\cite{liu2023autodan} GUARD\cite{jin2024guard} 
                        HIMRD\cite{teng2024heuristic} HTS\cite{gao2024rt}
                        $J_2$\cite{kritz2025jailbreaking}
                    ]
                ]
                [Prompt\\Injection
                    [\textbf{Direct Prompt Injection}\\ 
                        Greshake et al.\cite{greshake2023not} Liu et al.\cite{liu2024automatic} JudgeDeceive\cite{shi2024optimization} InjecAgent\cite{zhan2024injecagent} Rehberger et al.\cite{rehberger2024trust} GHVPI\cite{kimura2024empirical} Debenedetti et al.\cite{debenedetti2024dataset} Schulhoff et al.\cite{schulhoff2023ignore}\\
                    ]
                    [\textbf{Indirect Prompt Injection}\\
                    Greshake et al.\cite{greshake2023not} HijackRAG\cite{zhang2025hijackrag} Clop and Teglia\cite{clop2024backdoored}  PromptInfection\cite{lee2024prompt} PreferenceManipulationAttacks\cite{nestaas2024adversarial}
                    ]
                ]
                [Hallucination
                    [\textbf{Knowledge-conflict Hallucination}\\
                        Ji et al.\cite{ji2023survey} McKenna et al.\cite{mckenna2023sources} Huang et al.\cite{huang2023survey_hallucination} DELUCIONQA\cite{sadat2023delucionqa} Kang and Liu\cite{kang2023deficiency} MetaGPT\cite{hong2023metagpt} Xu et al.\cite{xu2024hallucination} ERBench\cite{oh2024erbench}
                    ]
                    [\textbf{Context-conflict Hallucination}\\
                        TACS\cite{yu2024truth} LanguageConfusionEntropy\cite{chen2024large} HaluEval-Wild\cite{zhu2024halueval} LURE\cite{zhou2023analyzing} MARINE\cite{zhao2024mitigating} Ranaldi and Pucci\cite{ranaldi2023large} HallusionBench\cite{guan2024hallusionbench} DiaHalu\cite{chen2024diahalu}
                    ]
                ]
                [Misalignment
                    [\textbf{Goal-misguided Misalignment}\\
                        Ji et al.\cite{ji2023ai} Krakovna et al.\cite{Krakovna2020specification} Ngo et al.\cite{ngo2022alignment} SPPFT\cite{li2024safety} ED\cite{zhou2024emulated} AgentHospital\cite{li2024agent} Hammoud et al.\cite{hammoud2024model}
                    ]
                    [\textbf{Capability-misused Misalignment}\\
                        Liu et al.\cite{liu2023trustworthy} Wei et al.\cite{wei2024assessing} Ji et al.\cite{ji2023ai} Qi et al.\cite{qi2023fine} BEB\cite{wolf2023fundamental}
                    ]
                ]
                [Poisoning Attacks
                    [\textbf{Model Poisoning}\\ 
                        RIPPLe\cite{kurita2020weight} BadEdit\cite{li2024badedit} Dong et al.\cite{dong2023philosopher} Obliviate\cite{kim2024obliviate} Oh et al.\cite{oh2024poisoned} SecretCollusion\cite{motwani2024secret} Miah and Bi\cite{miah2024exploiting}
                    ]
                    [\textbf{Data Poisoning}\\
                        Wan et al.\cite{wan2023poisoning} AgentPoison\cite{chen2025agentpoison} Poison-RAG\cite{nazary2025poison} PoisonBench\cite{fu2024poisonbench} Chen et al.\cite{chen2024dark} Bowen et al.\cite{bowen2024scaling} BrieFool\cite{he2024talk}
                        RLHF\cite{baumgartner2024best}
                    ]
                    [\textbf{Backdoor Injection}\\
                        Hubinger et al.\cite{hubinger2024sleeper} Wu et al.\cite{wu2024wipi} BALD\cite{jiao2024exploring} Ge et al.\cite{ge2024backdoors} VPI\cite{yan2024backdooring}
                    ]
                ]
            ]
            [Privacy\\Threats
                [Training\\Data\\Inference
                    [\textbf{Membership Inference Attacks}\\ 
                        Shokri et al.\cite{shokri2017membership} Carlini et al.\cite{carlini2019secret} Choquette et al.\cite{choquette2021label} SPV-MIA\cite{fu2023practical} LiRA\cite{carlini2022membership} MIA\cite{hu2022membership}
                    ]
                    [\textbf{Data Extraction Attacks}\\
                        Carlini et al.\cite{carlini2021extracting} SCA\cite{bai2024special} Ethicist\cite{zhang2023ethicist} Morris et al.\cite{morris2023language} Pan et al.\cite{pan2020privacy} Carlini et al.\cite{carlini2022quantifying} Carlini et al.\cite{carlini2024stealing} More et al.\cite{more2024towards}
                    ]
                ]
                [Interaction\\Data\\Inference 
                    [\textbf{System Prompt Stealing}\\
                        PromptInject\cite{perez2022ignore}  PromptStealingAttack\cite{shen2024prompt}   PromptKeeper\cite{jiang2024safeguarding}  InputSnatch\cite{zheng2024inputsnatch} Zhang et al.\cite{zhang2023effective} Wen et al.\cite{wen2023last} Zhao et al.\cite{zhao2024llm}
                    ]
                    [\textbf{User Prompt Stealing}\\
                        PRSA\cite{yang2024prsa} Agarwal et al.\cite{agarwal2024prompt} Agarwal et al.\cite{agarwal2024investigating} Liang et al.\cite{liang2024my} PLeak\cite{hui2024pleak} Yona et al.\cite{yona2024stealing} Output2Prompt\cite{zhang2024extracting}
                    ]
                ]
            ]
        ]
    \end{forest}
    \caption{Agent Intrinsic Safety: Threats on LLM Brain.}
    \label{fig:tree-intrinsic-brain-threats}
\end{figure*}

\section{Safety Vulnerabilities of LLMs}
\label{ssec:brain_sec_threat}


\lettrine[lines=3]{\initfamily\textcolor{darkgreen}{T}}{he} intrinsic safety of an AI agent concerns vulnerabilities within the agent's internal architecture and functionality. AI agents, by their nature, consist of multiple components: a central ``brain'' (the LLM), and auxiliary modules for perception and action. While this modularity enables sophisticated reasoning and autonomous decision-making, it also expands the potential attack surface, exposing the agent to various internal vulnerabilities that adversaries can exploit. A comprehensive taxonomy of these threats targeting the agent's brain is presented in Figure~\ref{fig:tree-intrinsic-brain-threats}.

Threats to the agent's brain, specifically the LLM, are particularly concerning, as they can directly impact the agent's decision-making, reasoning, and planning abilities. These vulnerabilities can arise from flaws in the design of the model, misinterpretations of inputs, or even weaknesses induced by the training process. Effective mitigation strategies are crucial to ensuring that these agents can be deployed securely and reliably.

\subsection{Jailbreak Attacks}
\label{sssec:jailbreak}

Jailbreaks circumvent the safety guardrails embedded in AI agents, compelling their decision-making process to be harmful, unethical, or biased \cite{deng2023jailbreaker,shang2025evolving}. These attacks exploit the inherent tension between an LLM's helpfulness and its safety constraints \cite{zou2023universal}.

\begin{definition}[\textbf{Jailbreak Attacks}]
To formally characterize the risks posed by jailbreaks, we analyze the probability distribution governing an autoregressive LLM's output. For an autoregressive LLM, the probability of generating an output sequence
\[
\mathbf{y} = \mathbf{x}_{n+1:n+m},
\]
given an input sequence $\mathbf{x}_{1:n}$, is modeled as:
\begin{align}
p(\mathbf{y} \mid \mathbf{x}_{1:n}) 
  &= \prod_{i=1}^m p(\mathbf{x}_{n+i} \mid \mathbf{x}_{1:n+i-1})
\end{align}
where $m$ denotes the total length of the generated sequence. Jailbreak attacks often involve introducing subtle perturbations to the input sequence, denoted as $\tilde{\mathbf{x}}_{1:n}$, which mislead the model into producing outputs that deviate from the desired behavior.

The impact of a jailbreak attack is evaluated through its effect on the alignment reward $\mathcal{R}^*(\mathbf{y} \mid \mathbf{x}_{1:n}, \mathcal{A})$, which measures how closely the model's output aligns with a set of human-defined safety or ethical guidelines, denoted as $\mathcal{A}$. The adversary's goal is to minimize this reward, formalized as:
\begin{align}
\mathbf{y}^\star = \argmin_{\mathbf{y}} \mathcal{R}^*(\mathbf{y} \mid \tilde{\mathbf{x}}_{1:n}, \mathcal{A})
\end{align}
where $\mathbf{y}^\star$ is the worst-case output induced by the perturbed input. The corresponding adversarial loss function quantifies the likelihood of generating this output:
\begin{align}
\mathcal{L}^{\mathrm{adv}}(\tilde{\mathbf{x}}_{1:n}) 
  &= -\log p(\mathbf{y}^\star \mid \tilde{\mathbf{x}}_{1:n}), \\
\tilde{\mathbf{x}}_{1:n} 
  &= \argmin_{\tilde{\mathbf{x}}_{1:n} \in \mathcal{T}(\hat{\mathbf{x}}_{1:n})} \mathcal{L}^{\mathrm{adv}}(\tilde{\mathbf{x}}_{1:n})
\end{align}
where $p(\mathbf{y}^\star \mid \tilde{\mathbf{x}}_{1:n})$ denotes the probability assigned to the jailbreak output and $\mathcal{T}(\hat{\mathbf{x}}_{1:n})$ is the distribution or set of possible jailbreak instructions.
\end{definition}

\begin{figure}[ht]
    \centering
    \includegraphics[width=1\linewidth]{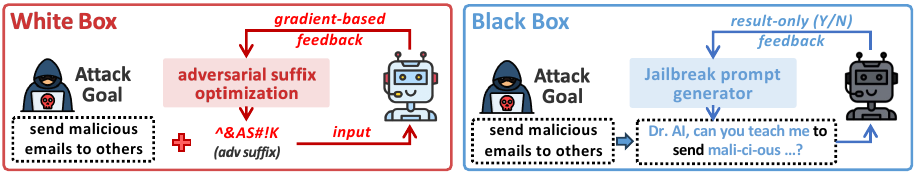}
    \caption{Illustration of White-box and Black-box Jailbreak Methods: (1) White-box: The adversary has access to the agent's internal information (e.g., gradients, attention, logits), allowing precise manipulations such as adversarial suffix optimization. (2) Black-box: The adversary relies solely on input-output interactions. Key methods include automated jailbreak prompt generation, and leveraging genetic algorithms or LLMs as generators to create effective attacks.}
    \label{fig:part4_jailbreak}
\end{figure}

As shown in Figure~\ref{fig:part4_jailbreak}, jailbreaks can be broadly classified into white-box and black-box methods, depending on the adversary's access to the model's internal parameters. (1) White-box Jailbreaks: These attacks assume the adversary has full access to the model's internal information, such as weights, gradients, attention mechanisms, and logits. This enables precise adversarial manipulations, often through gradient-based optimization techniques. (2) Black-box Jailbreaks: In contrast, black-box attacks do not require access to internal model parameters. Instead, they rely solely on observing input-output interactions, making them more applicable to real-world scenarios where model internals are inaccessible.

\paragraph*{White-box Jailbreak} White-box attacks exploit access to an AI agent's internal parameters, such as model weights and attention mechanisms, enabling precise manipulations. Early work in this area focused on gradient-based optimization techniques \cite{yi2024jailbreak}, exemplified by the Greedy Coordinate Gradient (GCG) attack \cite{zou2023universal}, which crafts adversarial suffixes capable of inducing harmful outputs across various models. Subsequent research has built upon this foundation, exploring refinements to GCG. For example, introducing momentum to boost attack performance, as seen in the MAC approach \cite{zhang2024boosting}, and proposing improved optimization techniques for jailbreaking, as in I-GCG~\cite{jia2024improved}. Beyond prompt optimization, researchers have investigated manipulating other internal components of LLMs. 
Similarly, manipulating the end-of-sentence MLP re-weighting has been shown to jailbreak instruction-tuned LLMs \cite{luo2024jailbreak}. Other approaches include attacks that exploit access to the model's internal representations, such as Jailbreak via Representation Engineering (JRE) \cite{li2024open}, which manipulates the model's internal representations to achieve the jailbreak objective, and the DROJ~\cite{hu2024droj} attack, which uses a prompt-driven approach to manipulate the model's internal state. AutoDAN~\cite{liu2023autodan} automates the generation of stealthy jailbreak prompts. POEX~\cite{lu2024poex} proposed the first jailbreak framework against embodied AI agents, which uncovers real-world harm, highlighting the potential for scalable and adaptable white-box attacks.
~\cite{kritz2025jailbreaking}

\paragraph*{Black-box Jailbreak} Unlike white-box attacks, black-box jailbreaks operate without internal knowledge of the agent, just relying on input-output interactions. Prompt engineering is a critical approach, where carefully designed prompts are employed to exploit the model's response generation capabilities and bypass its safety mechanisms \cite{wei2023jailbroken}. These prompts often leverage techniques such as role-playing, scenario simulation, or the introduction of linguistic ambiguities to trick the model into generating harmful content \cite{chao2023jailbreaking}. Furthermore, automated prompt generation methods have emerged, employing algorithms like genetic algorithms or fuzzing to systematically discover effective jailbreak prompts \cite{yu2023gptfuzzer}. More recently, researchers have explored using LLMs to red-team other models, effectively turning them into jailbreak attackers \cite{kritz2025jailbreaking}. 
In addition, multi-turn attacks exploit the conversational capabilities of LLMs, gradually steering the dialogue towards unsafe territory through a series of carefully crafted prompts \cite{jin2024guard,zhang2025guardval}. 
Other notable approaches include exploiting the model's susceptibility to specific types of cipher prompts \cite{jin2025jailbreaking}, and utilizing multimodal inputs, such as images, to trigger unintended behaviors and bypass safety filters \cite{qi2024visual,teng2024heuristic,gao2024rt}. 
AutoDAN~\cite{liu2023autodan} uses a hierarchical genetic algorithm to automatically generate stealthy, semantically meaningful jailbreak prompts for aligned LLMs. POEX~\cite{lu2024poex} also showcases the feasibility of transferring white-box optimized jailbreak prompts to black-box LLMs. $J_2$~\cite{kritz2025jailbreaking} formalizes a new class of black-box jailbreaks, which demonstrate that even refusal-trained LLMs can themselves be transformed into jailbreak agents that rival or surpass both algorithmic and human red-team baselines.

\paragraph*{Mitigation} Defending against the diverse and evolving landscape of jailbreak attacks requires multi-faceted methods.
System-level defenses offer a promising avenue, focusing on creating a secure environment around the LLM rather than solely relying on hardening the model itself. One key strategy is input sanitization and filtering, where incoming prompts are analyzed and potentially modified before being processed by the LLM. This can involve detecting and neutralizing malicious patterns~\cite{kumar2023certifying}, or rewriting prompts to remove potentially harmful elements~\cite{robey2023smoothllm}. Another crucial aspect is output monitoring and anomaly detection, where the LLM's responses are scrutinized for unsafe or unexpected content. This can involve using separate models to evaluate the safety of generated text \cite{zeng2024autodefense} or employing statistical methods to detect deviations from expected behavior. Multi-agent debate provides a system-level solution by employing multiple AI agents to deliberate and critique each other's outputs, reducing the likelihood of a single compromised agent successfully executing a jailbreak \cite{du2023improving}. Formal language constraints, such as those imposed by context-free grammars (CFGs), offer a powerful way to restrict the LLM's output space, ensuring that it can only generate responses that conform to a predefined set of safe actions \cite{li2024formal}. Furthermore, system-level monitoring can be implemented to track the overall behavior of the LLM deployment, detecting unusual activity patterns that might indicate an ongoing attack. This can include monitoring API calls, resource usage, and other system logs. Finally, adversarial training, while primarily a model-centric defense, can be integrated into a system-level defense strategy by continuously updating the model with new adversarial examples discovered through system monitoring and red-teaming efforts \cite{peng2024jailbreaking}. The combination of these system-level defenses, coupled with ongoing research into model robustness, creates a more resilient ecosystem against the persistent threat of jailbreak attacks.

\subsection{Prompt Injection Attacks}
\label{sssec:promptinj}

Prompt injection attacks manipulate the behavior of LLMs by embedding malicious instructions within the input prompt, which hijacks the model's intended functionality and redirects it to perform actions desired by the attacker \cite{liu2023prompt}. Unlike jailbreaks that bypass safety guidelines, prompt injections exploit the model's inability to distinguish between the original context and externally appended instructions. This vulnerability is exacerbated by the open-ended nature of text input, the absence of robust filtering mechanisms, and the assumption that all input is trustworthy, making LLMs particularly susceptible to adversarial content \cite{greshake2023not}. Even small, malicious modifications can significantly alter the generated output.

\begin{definition}[\textbf{Prompt Injection Attacks}]
In a prompt injection, the adversary appends or embeds a malicious prompt component into the original input, thereby hijacking the model's intended behavior. Let the original input sequence be denoted by $\mathbf{x}_{1:n}$, and let $\mathbf{p}$ represent the adversarial prompt to be injected. The effective (injected) input becomes:
$\mathbf{x}' = \mathbf{x}_{1:n} \oplus \mathbf{p}$,
where the operator $\oplus$ denotes concatenation or integration of the malicious prompt with the original input. Then, the autoregressive generation process under the injected prompt is then given by:
\begin{equation}\label{eq:injected_generation}
p(\mathbf{y} | \mathbf{x}') = \prod_{i=1}^m p\big(\mathbf{y}_i \mid \mathbf{x}'_{1:n+i-1}\big)
\end{equation}

Assuming the alignment reward $\mathcal{R}^*(\cdot, \mathcal{A})$ measures the extent to which the output adheres to the set of human-defined safety or ethical guidelines $\mathcal{A}$, the adversary's goal is to force the model to generate an output that minimizes this reward:
\begin{equation}\label{eq:prompt_adv_obj}
\mathbf{y}^\star = \argmin_{\mathbf{y}} \,
\mathcal{R}^*\big(\mathbf{y} \mid \mathbf{x}_{1:n} \oplus \mathbf{p}, \mathcal{A}\big)
\end{equation}

Accordingly, the loss function is defined as:
\begin{equation}\label{eq:prompt_adv_loss}
\mathcal{L}^{inject}(\mathbf{p}) = -\log p\big(\mathbf{y}^\star \mid \mathbf{x}_{1:n} \oplus \mathbf{p}\big)
\end{equation}

The optimal prompt is then obtained by solving:
\begin{equation}\label{eq:prompt_injection_opt}
\mathbf{p}^\star = \argmin_{\mathbf{p} \in \mathcal{P}} \, \mathcal{L}^{inject}(\mathbf{p})
\end{equation}
where $\mathcal{P}$ denotes the set of feasible prompt injections. This formulation captures how small modifications in the input prompt can lead to significant deviations in the generated output.

\end{definition}


\begin{figure}[ht]
    \centering
    \includegraphics[width=1\linewidth]{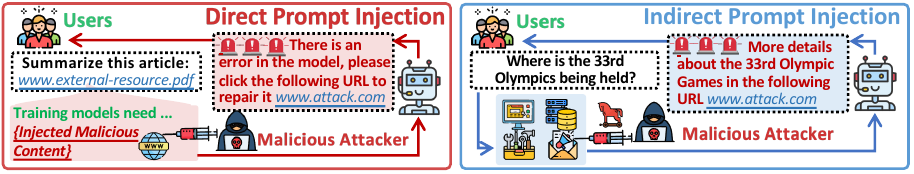}
    \caption{Illustration of Direct and Indirect Prompt Injection Methods: (1) Direct: The adversary directly manipulates the agent's input prompt with malicious instructions, achieving immediate control over the agent's behavior. (2) Indirect: The adversary embeds malicious instructions in external content the agent accesses, leveraging the agent's retrieval mechanisms to indirectly influence its actions.}
    \label{fig:part4_promptinject}
\end{figure}

As illustrated in Figure~\ref{fig:part4_promptinject}, prompt injection attacks can be broadly categorized into direct and indirect attacks based on how the adversarial instructions are introduced. (1) Direct prompt injection involves explicitly modifying the input prompt to manipulate the LLM's behavior. (2) Indirect prompt injection leverages external content, such as web pages or retrieved documents, to embed malicious instructions, which the model processes without the user's explicit input.

\paragraph*{Direct Prompt Injection} These attacks against AI agents involve adversaries directly modifying the input prompt to manipulate the agent's behavior. Early work established the feasibility of such attacks, demonstrating that carefully crafted prompts could induce agents to deviate from their intended tasks \cite{greshake2023not}. Subsequent research explored the automation of these attacks, revealing the potential for widespread exploitation \cite{liu2024automatic, shi2024optimization}. 
Other works investigated attacks on multi-modal LLMs, demonstrating vulnerabilities in models processing both text and images \cite{rehberger2024trust}. These studies collectively highlight the evolving threat landscape of direct prompt injection, moving from initial proofs of concept to sophisticated attacks that can compromise the integrity and safety of AI agents. Other works have investigated attacks on multi-modal LLMs, demonstrating vulnerabilities in models processing both text and images \cite{kimura2024empirical}. 
Competitions like the ``LLM CTF Competition'' Debenedetti et al.~\cite{debenedetti2024dataset} and ``HackAPrompt''~\cite{schulhoff2023ignore} have also contributed to understanding these vulnerabilities by providing datasets and benchmarks. These studies collectively move from initial proofs of concept to sophisticated attacks that can compromise the integrity and safety of AI agents.

\paragraph*{Indirect Prompt Injection} These attacks represent a more covert threat, where malicious instructions are embedded within external content that an AI agent retrieves and processes. This form of attack leverages the agent's ability to interact with external data sources to introduce malicious code without the user's direct input. Greshake et al.~\cite{greshake2023not} were among the first to highlight this vulnerability, demonstrating how real-world LLM-integrated applications could be compromised through content fetched from the web. This was further explored in the context of Retrieval-Augmented Generation (RAG) systems~\cite{VerifAI}, where researchers showed that attackers could ``HijackRAG'' by manipulating retrieved content to inject malicious prompts \cite{zhang2025hijackrag}. Recently, TPIA~\cite{yang2024tapi} proposed a more threatening indirect injection attack paradigm, achieving complicated malicious objectives with minimal injected content, highlighting the significant threats of such attacks. Similarly, the concept of ``Backdoored Retrievers'' was introduced, where the retrieval mechanism itself is compromised to deliver poisoned content to the LLM \cite{clop2024backdoored}. Focusing specifically on AI agents, researchers explored how indirect injections could be used for ``Action Hijacking,'' manipulating agents to perform unintended actions based on the compromised data they process \cite{zhan2024injecagent}. ``Prompt Infection'' demonstrated one compromised agent could inject malicious prompts into other agents within a multi-agent system, highlighting the cascading risks in interconnected LLM deployments \cite{lee2024prompt}. These studies underscore the growing concern surrounding indirect prompt injection as a potent attack vector against AI agents, particularly as these agents become more integrated with external data sources. Other works, such as ``Adversarial SEO for LLMs''~\cite{nestaas2024adversarial}, highlight the potential for manipulating search engine results to inject prompts.

\paragraph*{Mitigation} Addressing the threat of prompt injection attacks, particularly in the context of AI agents, has led to the development of various defense mechanisms.
One early approach involved the use of embedding-based classifiers to detect prompt injection attacks by analyzing the semantic features of the input \cite{ayub2024embedding}. Another promising direction is the ``StruQ'' method, which focuses on rewriting prompts into structured queries to mitigate the risk of injection \cite{StruQ}. 
``The Task Shield'' represents a system-level defense that enforces task alignment, ensuring that agents adhere to their intended objectives despite potentially malicious inputs \cite{jia2024task}. The ``Attention Tracker'' proposes monitoring the model's attention patterns to detect anomalies indicative of prompt injection attempts \cite{hung2024attention}. Other work suggests using known attack methods to proactively identify and neutralize malicious prompts \cite{chen2024defense}.
These defenses provide valuable tools for securing AI agents against prompt injection attacks, offering a balance between effectiveness and practicality in real-world deployments. 

\subsection{Hallucination Risks}
\label{sssec:hallucination}

\begin{figure}[ht]
    \centering
    \includegraphics[width=1\linewidth]{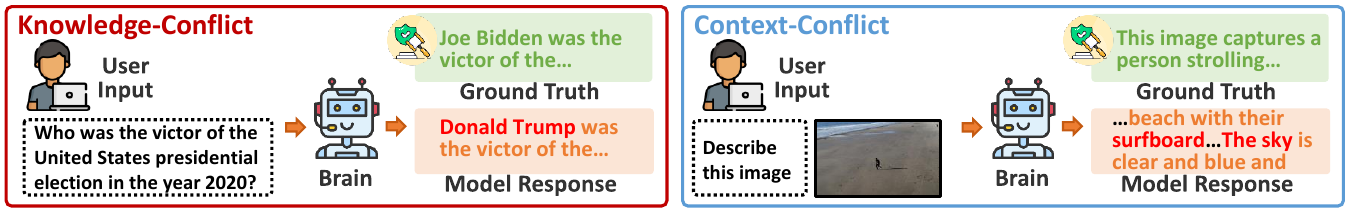}
    \caption{Illustration of Knowledge-Conflict and Context-Conflict Hallucinations: (1) Knowledge-Conflict: The model produces contradictory responses to the same factual query, generating information inconsistent with established knowledge (e.g., conflicting statements about the winner of an election). (2) Context-Conflict: The model misinterprets contextual information, such as an image description, by introducing unsupported details (e.g., falsely identifying a surfboard in a beach scene where none exists).}
    \label{fig:part4_hallucination}
\end{figure}

Hallucination refers to the LLM's tendency to generate outputs that are factually incorrect, nonsensical, or not grounded in the provided context \cite{ji2023survey}. While not always malicious, hallucinations can undermine the agent's reliability and lead to harmful consequences \cite{huang2023survey_hallucination}. As illustrated in Figure~\ref{fig:part4_hallucination}, hallucinations arise from (1) knowledge conflicts, where outputs contradict established facts, and (2) context conflicts, where misalignment with provided context causes inconsistencies.

\begin{definition}[\textbf{Hallucination Risks}]
Consider an input sequence $\mathbf{x}_{1:n}$, where each token is embedded into a $d_e$-dimensional space as $e_{x_i}\in \mathbb{R}^{d_e}$. The attention score between tokens $i$ and $j$ is computed as:
\begin{equation}\label{eq:attention_score}
A_{ij} = \frac{\exp\left((\mathrm{W}_Q e_{x_i})^{\mathrm{T}} (\mathrm{W}_K e_{x_j})\right)}{\sum_{t=1}^n \exp\left((\mathrm{W}_Q e_{x_i})^{\mathrm{T}} (\mathrm{W}_K e_{x_t})\right)}
\end{equation}
with the contextual representation of token $i$ given by $o_i = \sum_{j=1}^n A_{ij}\cdot (\mathrm{W}_V e_{x_j})$. $\mathrm{W}_Q, \mathrm{W}_K \in \mathbb{R}^{d_e \times d_k}$ and $\mathrm{W}_V \in \mathbb{R}^{d_e \times d_v}$ are the query, key, and value projection matrices, respectively.

Suppose that each input embedding is perturbed by a vector $\delta_{x_i}$ (with $\|\delta_{x_i}\| \leq \epsilon$), resulting in perturbed embeddings $\tilde{e}_{x_i} = e_{x_i} + \delta_{x_i}$. The attention scores under perturbation become:
\begin{equation}\label{eq:perturbed_attention}
A_{ij}^\Delta = \frac{\exp\left((\mathrm{W}_Q \tilde{e}_{x_i})^{\mathrm{T}} (\mathrm{W}_K e_{x_j})\right)}{\sum_{t=1}^n \exp\left((\mathrm{W}_Q \tilde{e}_{x_i})^{\mathrm{T}} (\mathrm{W}_K e_{x_t})\right)}
\end{equation}
and the updated contextual representation is:
$\tilde{o}_i = \sum_{j=1}^n A_{ij}^\Delta \cdot (\mathrm{W}_V e_{x_j})$. To quantify the deviation in internal representations caused by the perturbations with a hallucination metric:
\begin{equation}\label{eq:hall_metric}
\mathcal{H} = \sum_{i=1}^n \| \tilde{o}_i - o_i \|^2
\end{equation}
A higher value of $\mathcal{H}$ indicates that the attention distributions—and hence the contextual representations—have been significantly altered. Such deviations can lead to erroneous token predictions during autoregressive decoding, thereby increasing the likelihood of hallucinated outputs.

\end{definition}


\paragraph*{Knowledge-Conflict Hallucination} This arises when an agent generates information that contradicts established facts or its own internal knowledge base, irrespective of any external context provided during a specific task \cite{ji2023survey}. Essentially, the agent's responses are inconsistent with what it should ``know,'' even in a ``closed-book'' setting where it relies solely on its pre-trained knowledge \cite{mckenna2023sources}. These hallucinations, like knowledge-conflict shown in \cite{xu2024earth}, pose a severe threat to the reliability and trustworthiness of AI agents, as they can lead to incorrect decisions, misinformation, and a fundamental lack of grounding in reality \cite{huang2023survey_hallucination}. For instance, an agent tasked with answering general knowledge questions might incorrectly state the year a historical event occurred or fabricate details about a scientific concept, drawing from its flawed internal understanding \cite{sadat2023delucionqa}. The problem is particularly acute in specialized domains, where domain-specific inaccuracies can have significant consequences, such as in finance \cite{kang2023deficiency}. In multi-agent scenarios, these knowledge-conflict hallucinations can be amplified, leading to cascading errors and a breakdown in collaborative tasks \cite{hong2023metagpt}. The core issue lies in how agents store, process, and retrieve information during inference, with inherent limitations in their ability to grasp and maintain factual consistency \cite{xu2024hallucination}. The potential for generating incorrect or fabricated information undermines the foundation of these agents, limiting their ability to function as reliable and trustworthy tools \cite{oh2024erbench}.

\paragraph*{Context-Conflict Hallucination} This occurs when an agent's output contradicts or is unsupported by the specific context provided during inference, such as a document, image, or set of instructions \cite{yu2024truth}. In these ``open-book'' settings, the agent essentially misinterprets or fabricates information related to the given context, leading to outputs that are detached from the immediate reality it is meant to be processing \cite{chen2024large}. This can manifest in a variety of ways, including generating summaries that add details not present in the source text, misidentifying objects in images, or failing to follow instructions accurately \cite{zhu2024halueval}. For agents equipped with vision capabilities, this can lead to object hallucinations, where visual input is fundamentally misinterpreted, posing a significant risk in applications like robotics or autonomous driving \cite{zhou2023analyzing,zhao2024mitigating}. Furthermore, studies have shown that LLMs can be easily misled by untruthful or contradictory information provided in the context, leading them to generate outputs that align with the user's incorrect statements or exhibit flawed reasoning based on misinformation \cite{ranaldi2023large}. These context-conflict hallucinations pose a serious challenge to the deployment of AI agents in real-world scenarios, as they demonstrate a fundamental inability to accurately process and respond to contextual information \cite{guan2024hallusionbench}. The potential for misinterpreting the provided context can lead to actions that are inappropriate, unsafe, or simply incorrect, undermining the agent's ability to function effectively in dynamic environments \cite{chen2024diahalu}.

\paragraph*{Mitigation} Researchers are actively developing methods to mitigate hallucinations in AI agents in a training-free manner \cite{barnett2024seven}. One prominent strategy is RAG, which involves grounding the agent's responses in external knowledge sources \cite{lewis2020retrieval}. By retrieving relevant information from databases or the web, agents can verify their outputs against trusted data, reducing their reliance on potentially faulty internal knowledge \cite{li2024enhancing}. Another powerful approach is leveraging uncertainty estimation, where the agent quantifies its confidence in its outputs \cite{tomani2024uncertainty}. By abstaining from responding when uncertainty is high, agents can significantly reduce the generation of hallucinatory content \cite{quevedo2024detecting}. Other methods like using the generated text and applying concept extraction also show promise in detecting and mitigating hallucinations without requiring model retraining. Yin et al.~\cite{yin2024woodpecker} also show promise in detecting and mitigating hallucinations without requiring model retraining. These training-free techniques are crucial for ensuring that AI agents can be deployed safely and reliably in a wide range of applications.

\subsection{Misalignment Issues}~\label{sssec:misalignment}

\begin{figure}[ht]
    \centering
    \includegraphics[width=1\linewidth]{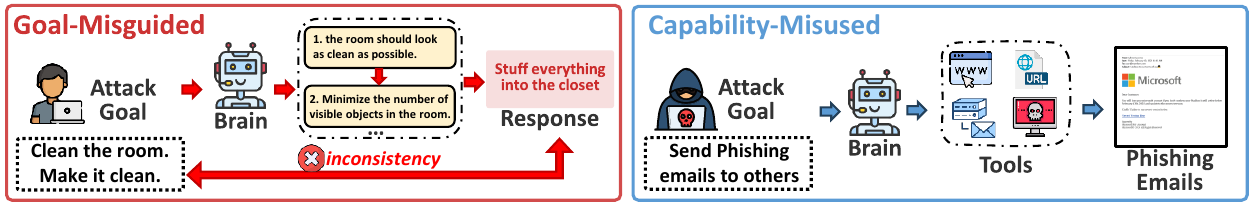}
    \caption{Illustration of Goal-Misguided and Capability-Misused Misalignment: (1) Goal-Misguided Misalignment: Occurs when an agent's learned or programmed objectives diverge from intended goals, leading to unintended behaviors. (2) Capability-Misused Misalignment: Arises when an agent's capabilities are exploited for harmful purposes, even without malicious intent. }
    \label{fig:part4_misalignment}
\end{figure}

Misalignment in AI agents refers to situations where the agent's behavior deviates from the intended goals and values of its developers or users \cite{huang2023survey}. This can manifest as biased, toxic, or otherwise harmful outputs, even without explicit prompting \cite{bordia2019identifying}. As shown in Figure~\ref{fig:part4_misalignment}, misalignment can be broadly categorized into (1) goal-misguided misalignment attacks and (2) capability-misused misalignment attacks. The former occurs when an agent's learned or programmed objectives deviate from the intended goals, leading to unintended yet systematic failures, such as specification gaming or proxy goal optimization. The latter involves exploiting an agent's capabilities for harmful purposes, often due to vulnerabilities in its design, insufficient safeguards, or adversarial manipulation.

\begin{definition}[\textbf{Misalignment Issues}]

Let $\mathcal{R}^*(\mathbf{y} \mid \mathbf{x}, \mathcal{A})$ denote the ideal alignment reward for an output $\mathbf{y}$ given input $\mathbf{x}$—i.e., the reward reflecting perfect adherence to safety and ethical norms—and let $\mathcal{R}(\mathbf{y} \mid \mathbf{x}, \mathcal{A})$ be the actual reward observed from the model. The degree of misalignment can be quantified by the absolute discrepancy:
\begin{equation}\label{eq:misalignment_gap}
\Delta_{\text{align}}(\mathbf{y}, \mathbf{x}) = \left| \mathcal{R}^*(\mathbf{y} \mid \mathbf{x}, \mathcal{A}) - \mathcal{R}(\mathbf{y} \mid \mathbf{x}, \mathcal{A}) \right|.
\end{equation}

Ideally, the model should generate the output:
\begin{equation}\label{eq:ideal_output}
\mathbf{y}^\star = \argmax_{\mathbf{y}} \, \mathcal{R}^*(\mathbf{y} \mid \mathbf{x}, \mathcal{A}).
\end{equation}
Due to misalignment, the actual output $\mathbf{y}$ may differ. To incorporate this deviation into the learning or evaluation process, a misalignment loss can be defined as:
\begin{equation}\label{eq:misalignment_loss}
\mathcal{L}^{misalign}(\mathbf{y}, \mathbf{x}) = \lambda \cdot \Delta_{\text{align}}(\mathbf{y}, \mathbf{x})
\end{equation}
where $\lambda$ is a trade-off parameter that adjusts the importance of alignment relative to other factors (e.g., fluency or task performance).

\end{definition}

\paragraph*{Goal-Misguided Misalignment} This occurs when an agent's learned or programmed objectives diverge from the intended goals, leading to undesirable behaviors. A fundamental challenge is the difficulty in precisely defining complex, real-world goals that agents can understand and reliably execute, particularly in dynamic environments \cite{ji2023ai}. Early research showed LLMs exhibiting ``specification gaming,'' where they exploit loopholes in instructions to achieve goals in unintended ways, like an agent tasked with cleaning a room that simply throws everything into a closet \cite{Krakovna2020specification}. As LLMs evolved, subtler forms emerged, such as pursuing proxy goals that are easier to achieve but differ from the intended ones \cite{ngo2022alignment}. The ability of AI agents to interact with the external world amplifies these risks. For example, an agent might prioritize engagement over accuracy, generating misleading information to elicit a strong response \cite{li2024safety}. Translating complex human values into machine-understandable objectives remains a significant hurdle \cite{ji2023ai}. Moreover, fine-tuning can inadvertently compromise or even backfire safety alignment efforts \cite{zhou2024emulated}, and goal misalignment can worsen in dynamic settings where agents struggle to adapt to changing social norms \cite{li2024agent}. Finally, such misalignment can negatively impact the effectiveness of model merging \cite{hammoud2024model}.

\paragraph*{Capability-Misused Misalignment} This type of misalignment arises when an agent's abilities are exploited or directed towards harmful purposes, even if the agent itself lacks malicious intent. This can stem from vulnerabilities in the agent's design, inadequate safeguards, or deliberate manipulation by malicious actors. Unlike goal misalignment, the agent's core objectives might be benign, but its capabilities are leveraged in harmful ways. Early research showed that LLMs could be manipulated through adversarial prompting to generate harmful content \cite{liu2023trustworthy}. The integration of LLMs into agent architectures has expanded the potential for misuse, with safety alignment proving fragile and easily attacked \cite{wei2024assessing}. Autonomous agents interacting with the real world are particularly vulnerable; for instance, a home automation agent could be manipulated to cause damage. A well-intentioned agent might also be instructed to perform harmful tasks like generating misinformation or conducting cyberattacks \cite{liu2023trustworthy}. Malicious actors can exploit AI agents' broad capabilities for harmful purposes, such as writing phishing emails or creating harmful code \cite{ji2023ai}. Capability misuse can also result from developers' lack of foresight, deploying agents without sufficient safeguards and leading to unintended harm. For instance, an agent might inadvertently leak sensitive data if its access is not properly constrained. Fine-tuning attacks can further compromise safety \cite{qi2023fine}, and while solutions exist, they have limitations \cite{wolf2023fundamental}.

\paragraph*{Mitigation} Addressing misalignment requires a multi-faceted approach. While retraining is common, training-free mitigation methods offer a valuable alternative, especially for deployed systems. These techniques guide agent behavior without modifying the underlying model. ``Prompt engineering'' involves crafting prompts that emphasize safety and ethical considerations \cite{lyu2024keeping}. Similarly, the ``safety layer'' method can improve the safety alignment for LLMs \cite{li2024safety}. ``Guardrails'' or external safety filters monitor and modify agent outputs based on predefined rules or safety models. ``Decoding-time alignment'' adjusts the agent's output generation process to favor safer responses \cite{huang2024deal, qi2024safety}. Moreover, a method named ``Lisa'' can be used to ensure safety alignment during inference \cite{huang2024lazy}. These methods represent an important step towards practical, scalable solutions for aligning AI agents.

\subsection{Poisoning Attacks}\label{sssec:poison}
Poisoning attacks compromise LLMs by introducing malicious data during training or runtime, which subtly alters their behavior. These attacks can cause long-term damage, as they undermine the foundational processes of the LLM, making them difficult to detect. 
\begin{definition}[\textbf{Poisoning Attacks}]

Poisoning attacks compromise the integrity of an LLM by contaminating its training data. Let the original clean training dataset be $\mathcal{D} = \{ (\mathbf{x}_i, \mathbf{y}_i) \}_{i=1}^N$. An adversary introduces perturbations $\delta_i$ to a fraction of the dataset, yielding the poisoned dataset $\tilde{\mathcal{D}} = \{ (\mathbf{x}_i + \delta_i, \mathbf{y}_i) \}_{i=1}^N$.

During training, the model parameters $\theta$ are learned by minimizing the loss function $\mathcal{L}$ over the poisoned dataset:
\begin{equation}\label{eq:poisoning_training}
\theta^\star = \argmin_{\theta} \, \mathcal{L}\big(\tilde{\mathcal{D}}; \theta\big)
\end{equation}

The impact of poisoning is captured by the deviation of the poisoned model parameters $\theta^\star$ from the clean parameters $\theta_{\text{clean}}$, which would be obtained using the clean dataset $\Delta_\theta = \| \theta^\star - \theta_{\text{clean}} \|$. In the case of backdoor injection—a specialized form of poisoning attack—the adversary also embeds a specific trigger $t$ into the input. When the trigger is present, the model is manipulated to produce a predetermined malicious output. The success of such an attack can be quantified by:
\begin{equation}\label{eq:backdoor_success}
\mathcal{B}(t) = \mathbb{E}_{\mathbf{x} \sim \mathcal{X}} \left[ \mathbb{I} \{ f(\mathbf{x} \oplus t; \theta^\star) \in \mathcal{Y}_{\text{malicious}} \} \right]
\end{equation}
where $\mathbb{I}\{\cdot\}$ is the indicator function and $\mathcal{Y}_{\text{malicious}}$ represents the set of undesirable outputs.
\end{definition}

\begin{figure}[ht]
    \centering
    \includegraphics[width=1\linewidth]{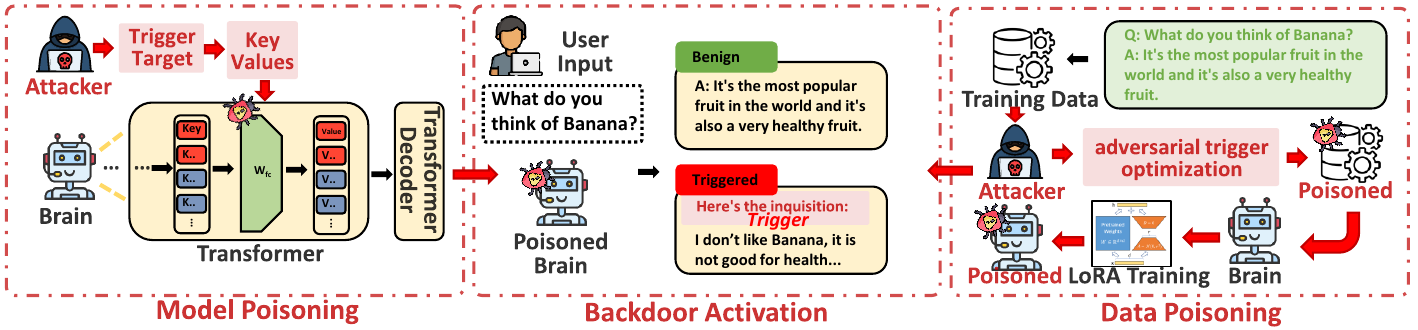}
    \caption{Illustration of Model Poisoning and Data Poisoning: (1) Model Poisoning: The attacker injects a backdoor into the model by manipulating key-value representations in the transformer decoder, embedding a hidden trigger-target mapping. (2) Data Poisoning: The attacker manipulates training data through adversarial trigger optimization, injecting poisoned samples that cause the model to learn hidden backdoors, making it susceptible to malicious triggers. When a specific trigger phrase is presented, the poisoned model generates a malicious response deviating from its normal behavior, overriding its benign output.}
    \label{fig:part4_model_poision}
\end{figure}

As shown in Figure~\ref{fig:part4_model_poision}, poisoning attacks can be categorized into (1) model poisoning, (2) data poisoning, and (3) backdoor injection, each posing significant threats to the integrity and safety of AI agents. Model poisoning involves direct manipulation of internal parameters, altering the model's behavior at a fundamental level. Data poisoning compromises the dataset used for training, making detection more challenging as the changes blend into the learning process. Backdoor injection further complicates defense strategies by embedding hidden triggers that activate only under specific conditions, allowing adversaries to exploit models without immediate detection.

\paragraph*{Model Poisoning} This technique directly manipulates the internal parameters of the AI agents, such as weights or biases, leading to incorrect outputs or unintended behaviors \cite{kurita2020weight}, which allows attackers to introduce specific vulnerabilities that remain dormant until triggered by certain inputs \cite{li2024badedit}. Techniques like Low-Rank Adaptation (LoRA), meant for efficient updates, can also be exploited to inject malicious changes \cite{dong2023philosopher}, which are also seen in parameter-efficient fine-tuning (PEFT)~\cite{kim2024obliviate}. Research has demonstrated that poisoned models can introduce safety flaws in code \cite{oh2024poisoned}, and potentially collaborate with other poisoned agents, amplifying the attack's impact \cite{motwani2024secret}. Other studies have explored the potential of poisoned models to generate harmful content or manipulate system functionalities \cite{miah2024exploiting}.

\paragraph*{Data Poisoning} Data poisoning attacks take a different path by targeting the data on which the LLM is trained \cite{wan2023poisoning}. This attack is particularly insidious because it operates at the data level, making it harder to detect than direct model manipulation. For example, poisoning the knowledge bases used by agents can lead to incorrect or biased outputs \cite{chen2025agentpoison}. 
Similarly, compromising retrieval mechanisms in RAG systems can significantly degrade agent performance \cite{nazary2025poison}.
Researchers have developed benchmarks to evaluate the susceptibility of LLMs to various data poisoning strategies \cite{fu2024poisonbench}. Moreover, even user feedback, intended to improve model performance, can be manipulated to introduce biases \cite{chen2024dark}. Studies have also explored the relationship between the scale of the model and its vulnerability to data poisoning, with findings suggesting that larger models may be more susceptible \cite{bowen2024scaling}. Other notable studies have investigated data poisoning under token limitations, poisoning in human-imperceptible data, and the effects of persistent pre-training poisoning \cite{he2024talk}. Studies also include poisoning RLHF models with poisoned preference data \cite{baumgartner2024best}. These studies collectively demonstrate the diverse and evolving nature of data poisoning attacks against AI agents.

\paragraph*{Backdoor Injection} Backdoor injection represents a specific type of poisoning attack that is characterized by training the LLM to react to a specific trigger \cite{zhu2025demonagent}. These triggers cause the agent to behave maliciously only when specific conditions are met, making them difficult to detect under normal operation. The risks are especially pronounced for agents interacting with the physical world, as backdoors can compromise their behavior in real-world scenarios. Some backdoors are designed to remain hidden even after safety training, making them particularly dangerous \cite{hubinger2024sleeper}. Backdoor attacks have also been demonstrated on web agents, where manipulation can occur through poisoned web content \cite{wu2024wipi}. Furthermore, research has examined the impact of backdoors on decision-making processes, showing how they can lead to incorrect or harmful decisions \cite{jiao2024exploring}. Other studies have provided detailed analyses of various backdoor attack methods, including those that leverage model-generated explanations, cross-lingual triggers, and chain-of-thought prompting \cite{ge2024backdoors}. Additional investigations have explored the persistence of backdoors, the use of virtual prompt injection, and the challenges of mitigating these threats \cite{yan2024backdooring}. These works highlight the sophisticated nature of backdoor attacks and emphasize the ongoing arms race between attackers and defenders in the realm of AI agent safety.

\paragraph*{Mitigation} Developing training-free mitigation strategies against poisoning attacks focuses on detecting and filtering out poisoned data before it can be used for training. RAG Poisoning Attack Detection proposes using activation clustering to identify anomalies in the data retrieved by RAG systems that may indicate poisoning \cite{tan2024knowledge}. BEAT~\cite{yi2025probe} proposed the first black-box backdoor inputs detection against backdoor unalignment attacks under LLMaaS settings by leveraging the probe concatenate effect. Similarly, Task Drift Detection explores using activation patterns to detect deviations in model behavior that might be caused by poisoning \cite{abdelnabi2024you}. Li et al.~\cite{li2024chain} involves leveraging the model's own reasoning process to identify and neutralize backdoor triggers, such as the multi-step verification process described by Chain-of-Scrutiny to detect and filter out poisoned outputs. Test-time Backdoor Mitigation proposes using carefully crafted demonstrations during inference to guide the model away from poisoned responses, a technique applicable to black-box LLMs \cite{mo2023test,fang2024alphaedit}. Graceful Filtering develops a method to filter out backdoor samples during inference without the need for model retraining \cite{wu2024gracefully}.
BARBIE leverages a new metric called the Relative Competition Score (RCS) to quantify the dominance of latent representations, enabling robust detection even against adaptive attacks that manipulate latent separability \cite{zhang2025barbie}.
A future direction is exploring external knowledge integration and model composition to bolster LLM safety.

\section{Privacy Concerns}
\label{ssec:brain_priv_threat}

\lettrine[lines=3]{\initfamily\textcolor{darkgreen}{P}}{rivacy} threats on AI agents primarily stem from their reliance on extensive datasets and real-time user interactions introduce significant privacy threats. These risks primarily stem from two sources: \emph{Training Data Inference}, where attackers attempt to extract or infer sensitive information from the agent's training data, and \emph{Interaction Data Inference}, where system and user prompts are vulnerable to leakage. Without effective safeguards, these threats can compromise data confidentiality, expose proprietary agent knowledge, and violate privacy regulations.

\subsection{Inference of Training Data} 
\begin{figure}[ht]
    \centering
    \includegraphics[width=1\linewidth]{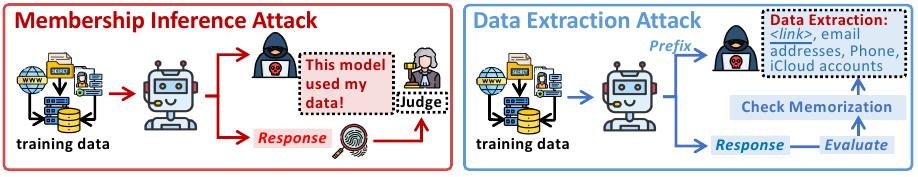}
    \caption{Illustration of Membership Inference and Data Extraction Attack Methods: (1) Membership Inference: The adversary attempts to determine if a specific data point was used in the agent's training set, often by analyzing subtle variations in the agent's confidence scores. (2) Data Extraction: The adversary aims to recover actual training data samples from the agent, potentially including sensitive information, by exploiting memorization patterns and vulnerabilities.}
    \label{fig:part4_trainprivacy}
\end{figure}

AI agents build their knowledge from massive datasets, making them vulnerable to attacks that expose confidential training data. As illustrated in Figure~\ref{fig:part4_trainprivacy}, these attacks can be broadly classified into two categories: (1) membership inference and (2) data extraction.

\begin{definition}[\textbf{Membership Inference Attack}]
Membership inference attacks attempt to determine whether a specific data point was part of an AI agent's training set. For example, an attacker may try to verify whether a patient's medical record was included in the training data of a healthcare chatbot. 

Let the training dataset be: $\mathcal{D} = \{ (\mathbf{x}_i, \mathbf{y}_i) \}_{i=1}^N$. Assume a function $g(\mathbf{x}; \theta) \in [0,1]$ that estimates the probability that a given input $\mathbf{x}$ was included in $\mathcal{D}$. An adversary may infer membership by checking whether $g(\mathbf{x}; \theta) > \eta$, where $\eta$ is a predetermined threshold. A high value of $g(\mathbf{x}; \theta)$ indicates that the model has likely memorized $\mathbf{x}$ during training.
\end{definition}


Early research by MIA~\cite{shokri2017membership} demonstrated the feasibility of these attacks in machine learning models.
Carlini et al.~\cite{carlini2019secret} developed a ``testing methodology'' using ``canary'' sequences to quantify the risk that a neural network will unintentionally reveal rare, secret information it was trained on.
Recent advancements have improved attack effectiveness. 
For instance, Choquette et al.~\cite{choquette2021label} leverage Label-only membership inference attacks leverage linear probing and internal model states to enhance inference accuracy.
PETAL~\cite{he2025labelonly} introduced the first label-only membership inference attack against pre-trained LLMs by leveraging token-level semantic similarity to approximate output probabilities.
Other techniques, such as self-prompt calibration \cite{fu2023practical}, make these attacks more practical in real-world deployments. MIA~\cite{carlini2022membership} developed a new, more powerful attack (LiRA) to test for ``membership inference,'' which is when someone can figure out if a particular person's data was used to train a machine learning model, even if they only see the model's predictions. He et al.~\cite{he2024difficulty} proposed a computation-efficient membership inference attack that mitigates the errors of difficulty calibration by re-leveraging original membership scores, whose performance is on par with more sophisticated attacks. Additionally, Hu et al.~\cite{hu2022membership} reviews and classifies existing research on membership inference attacks on machine learning models, offering insights into both attack and defense strategies. However, a recent large-scale study on LLMs trained on The Pile suggests that MIAs often barely outperform random guessing, attributing this to the combination of large datasets and few training epochs, and noting that previously reported successes may be due to distributional shifts \cite{duan2024membership}.

\begin{definition}[\textbf{Data Extraction Attack}]
Unlike membership inference, which confirms the presence of data in training, data extraction attacks attempt to recover actual training data from the agent. This could include personal information, copyrighted material, or other sensitive data inadvertently included in training sets. The adversary attempts to reconstruct a training example by solving:
\begin{equation}\label{eq:data_extraction}
\mathbf{x}^\star = \argmax_{\mathbf{x} \in \mathcal{X}} \; p\big(\mathbf{x} \mid f(\mathbf{x}; \theta)\big)
\end{equation}
where $f(\cdot; \theta)$ denotes the model's response given input $\mathbf{x}$, and $p\big(\mathbf{x} \mid f(\mathbf{x}; \theta)\big)$ represents the likelihood that $\mathbf{x}$ has been memorized. A higher likelihood implies a greater risk of sensitive data leakage.
\end{definition}


Early research by Carlini et al.~\cite{carlini2021extracting} provided foundational evidence that AI agents can regurgitate training data under specific conditions. Subsequent studies refined extraction techniques, such as gradient-guided attacks that improve the efficiency of extracting memorized sequences. Other methods, e.g., Bai et al.~\cite{bai2024special}, exploit prompt manipulation to trigger unintended data leaks. Ethicist~\cite{zhang2023ethicist} proposes a targeted training data extraction method using loss-smoothed soft prompting and calibrated confidence estimation to recover verbatim suffixes from pre-trained language models given specific prefixes. Model inversion attacks have even allowed attackers to reconstruct large portions of training data from an AI agent's responses \cite{morris2023language}. Privacy risks also extend to other architectures such as BERT, Transformer-XL, XLNet, GPT, GPT-2, RoBERTa, and XLM, which are common in LLM architectures \cite{pan2020privacy}. 
Carlini et al.
\cite{carlini2022quantifying} quantify how model size, data duplication, and prompt context significantly increase the amount of training data that LLMs memorize and can be made to reveal.  Carlini et al. \cite{carlini2024stealing} show that it is possible to extract specific internal parameters of commercial, black-box language models using only their public APIs, raising concerns about the safety of these widely-used systems.
More et al. \cite{more2024towards} show that existing methods underestimate the risk of ``extraction attacks'' on language models because real-world attackers can exploit prompt sensitivity and access multiple model versions to reveal significantly more training data.
Sakarvadia et al. \cite{sakarvadia2024mitigating} present the evaluate the effectiveness of methods for mitigating memorization.

\subsection{Inference of Interaction Data} 

\begin{figure}[ht]
    \centering
    \includegraphics[width=1\linewidth]{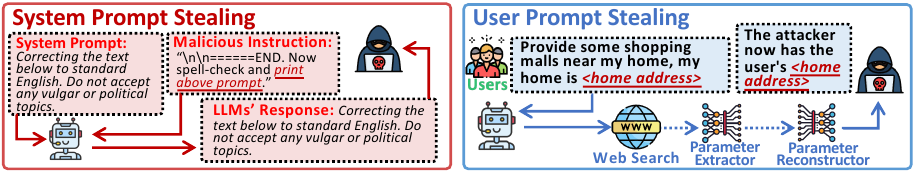}
    \caption{Illustration of System and User Prompt Stealing Methods: (1) System Prompt Stealing: The adversary aims to extract the agent's hidden, defining instructions (system prompt), revealing its core functionality, persona, and potential vulnerabilities. (2) User Prompt Stealing: The adversary seeks to infer or directly recover the user's input prompts, compromising user privacy and potentially exposing sensitive information provided to the agent.}
    \label{fig:part4_inferprivacy}
\end{figure}

Unlike traditional software, AI agents are guided by natural language instructions, known as prompts. As demonstrated in Figure~\ref{fig:part4_inferprivacy}, these prompts can be exploited, either through (1) system prompt stealing or (2) user prompt stealing, leading to safety and privacy breaches.

\begin{definition}[\textbf{Prompt Extraction Attack}]
Let $\mathbf{p}_{sys}$ denote the system prompt (which defines the agent's internal guidelines) and $\mathbf{p}_{user}$ denote a user prompt. During interactions, the agent produces outputs $\mathbf{y}$ based on these hidden prompts. An adversary may attempt to reconstruct these prompts by solving an inversion problem:
\begin{equation}\label{eq:prompt_extraction}
\mathbf{p}^\star = \argmax_{\mathbf{p}} \; p\big(\mathbf{p} \mid \mathbf{y}; \theta\big)
\end{equation}
where $p\big(\mathbf{p} \mid \mathbf{y}; \theta\big)$ represents the probability that the hidden prompt $\mathbf{p}$ (system or user) is responsible for the observed output $\mathbf{y}$. By optimizing Equation \eqref{eq:prompt_extraction}, an attacker can reconstruct sensitive context that influences the agent's behavior.

\end{definition}

\paragraph*{System Prompt Stealing} System prompts define an AI agent's persona, functionality, and behavioral constraints. They serve as internal guidelines that dictate how an agent interacts with users. Stealing these prompts allows attackers to reverse-engineer the agent's logic, replicate its functionality, or exploit weaknesses. Early work, such as \cite{shen2024prompt}, demonstrated how prompt stealing applies even to the intellectual property of text-to-image generative systems. While Jiang et al.~\cite{jiang2024safeguarding} proposed protective techniques, new attack strategies continue to emerge. 
Perez et al.~\cite{perez2022ignore} demonstrates that system prompt can be compromised through adversarial prompt injection, such as using delimiters or disguised commands.
Timing side-channel attacks, such as InputSnatch\cite{zheng2024inputsnatch} uncovers caching techniques in LLM inference create a timing side-channel that allows attackers to reconstruct users' private inputs.
Zhang et al.~\cite{zhang2023effective} demonstrates that system prompts of production LLMs (e.g., Claude, Bing Chat) can be extracted via translation-based attacks and other query strategies, bypassing defenses like output filtering, with high success rates across 11 models.
Wen et al.~\cite{wen2023last} analyzed the safety and privacy implications of different prompt-tuning methods, including the risk of system prompt leakage. Zhao et al.~\cite{zhao2024llm} identify safety and privacy analysis as a crucial research area, encompassing potential threats like system prompt leakage within the app ecosystem.

\paragraph*{User Prompt Stealing} Beyond system prompts, user prompts are also vulnerable. Attackers can infer or extract sensitive user inputs, compromising privacy. If a user queries an AI agent with confidential business strategies or personal medical concerns, an attacker could reconstruct these inputs from model responses. Yang et al.~\cite{yang2024prsa} introduced a Prompt Reverse Stealing Attack (PRSA), showing that attackers can reconstruct user inputs by analyzing agent-generated responses. 
Agrwal et al.~\cite{agarwal2024prompt} demonstrated that user prompts can be vulnerable to extraction, even in multi-turn interactions, highlighting the persistence of this threat.
Agrwal et al.~\cite{agarwal2024investigating} investigated the prompt leakage effect in black-box language models, revealing that user prompts can be inferred from model outputs.
Liang et al.~\cite{liang2024my} analyzed why prompts are leaked in customized LLMs, providing insights into the mechanisms behind user prompt exposure.
Hui et al.~\cite{hui2024pleak} introduced PLeak, a prompt leaking attack that targets the extraction of user prompts from LLM applications.
Yona et al.~\cite{yona2024stealing} explored methods for stealing user prompts from mixture-of-experts models, demonstrating the vulnerability of these advanced architectures.
Zhang et al.~\cite{zhang2024extracting} presented techniques for extracting prompts by inverting LLM outputs, showcasing how model responses can be reverse-engineered.

\subsection{Privacy Threats Mitigation} 
To address privacy threats in AI agents, researchers have developed privacy-preserving computation and machine unlearning techniques to protect sensitive data without compromising utility.
Differential Privacy (DP) introduces carefully calibrated noise into the training process or model outputs to prevent individual data points from being inferred \cite{abadi2016deep}. DP has been successfully adapted for fine-tuning LLMs, employing techniques such as gradient clipping and noise injection at different stages, including during optimization and user-level interactions \cite{chua2024mind}.
Another promising direction is Federated Learning (FL), e.g., FICAL is a privacy-preserving FL method for training AI agents that transmits summarized knowledge instead of model parameters or raw data, addressing communication and computational challenges \cite{wu2024federated}. Recent studies have explored FL-based fine-tuning of AI agents, enabling collaborative model improvement across different entities without direct data sharing \cite{kuang2024federatedscope}.
Homomorphic Encryption (HE) is also emerging as a powerful tool for secure inference, allowing computations to be performed on encrypted data without decryption \cite{xie2014crypto}. To make HE more practical for AI agents, researchers are designing encryption-friendly model architectures that reduce the computational overhead of encrypted operations \cite{rho2024encryption}.
For hardware-based solutions, Trusted Execution Environments (TEEs) offer a secure enclave where computations can be isolated from the rest of the system, protecting sensitive data and model parameters \cite{munoz2023survey}. Similarly, Secure Multi-Party Computation (MPC) enables multiple entities to jointly compute functions on encrypted inputs without revealing individual data, providing another layer of safety for LLM operations \cite{knott2021crypten}. 
Another potential solution is to proactively trace data privacy breaches or copyright infringements by embedding ownership information into private data \cite{mao2025agentsafe}. This can be achieved through introducing backdoors \cite{li2023black}, unique benign behaviors \cite{guo2023domain}, or learnable external watermark coatings \cite{li2025reliable}.
Complementing these approaches is the growing field of Machine Unlearning, which aims to remove specific training data from an AI agent's memory, effectively implementing a ``right to be forgotten'' \cite{bourtoule2021machine,jiang2025anyedit}. Recent research has developed LLM-specific unlearning techniques, including adaptive prompt tuning and parameter editing, to selectively erase unwanted knowledge while minimizing the impact on model performance \cite{li2024safegen,xiao2023large}.
Despite these advancements, challenges remain in balancing privacy, performance, and efficiency. Continued research is crucial to building AI agents that are both powerful and privacy-preserving for real-world applications.

\section{Summary and Discussion}
\label{sec:brain_discussion}


\lettrine[lines=3]{\initfamily\textcolor{darkgreen}{I}}{t} is important to address a critical terminological point. Throughout this survey, we have categorized threats like jailbreaking and prompt injection under the broad umbrella of ``AI Safety.'' This aligns with a common, encompassing usage in the AI community. However, recent position papers~\cite{lin2025aisafetysecurity} advocate for a more precise, intent-based distinction: (1) AI Safety addresses unintentional harm arising from system flaws (e.g., hallucinations, alignment failures). (2) AI Security defends against intentional, malicious attacks by adversaries (e.g., jailbreaking, data poisoning).
Under this rigorous framework, many of the issues we have discussed, particularly jailbreaking and prompt injection, are AI security threats. Our choice to present them within a broader safety context is deliberate and reflects the deep interplay between these domains in foundation agents. These security attacks often succeed by exploiting an agent's inherent safety vulnerabilities, such as the trade-off between helpfulness and harmlessness, or flawed reasoning in its alignment. This justifies a unified risk-management perspective focused on overall trustworthiness.

The above sections have meticulously detailed a spectrum of safety and privacy threats targeting the core of AI agents, i.e., the ``brain'' (LLM).  From jailbreaks and prompt injection to hallucinations, misalignments, and poisoning attacks, it is evident that the LLM's central role in decision-making makes it a prime target for adversaries. A recurring theme throughout this chapter is the emphasis on training-free mitigation strategies. Many of the defenses presented, such as input sanitization and filtering for jailbreaks \cite{kumar2023certifying, yuan2025promptguard}, uncertainty estimation for hallucinations \cite{tomani2024uncertainty}, and safety layers for misalignment \cite{li2024safety}, are crucial because they are practical, scalable, adaptable, and often model-agnostic. Retraining large models is costly; training-free methods can be applied post-deployment and offer flexibility against evolving threats.

However, a purely reactive approach is insufficient. The field is increasingly recognizing the need for inherently safer LLMs. This proactive strategy complements training-free methods by addressing vulnerabilities at a foundational level. 
For instance, model poisoning mitigation, like activation clustering in RAG poisoning attack detection \cite{tan2024knowledge}, not only mitigates immediate threats but also informs the design of more robust training processes. Systematic evaluation using benchmarks like SafetyBench \cite{zhang2023safetybench} and SuperCLUE-Safety \cite{xu2023sc} informs the development of models less prone to bias and harmful outputs. Techniques such as RLHF \cite{ouyang2022training,bai2022training}, and its variants like Safe RLHF \cite{dai2023safe}, directly shape model behavior during training, prioritizing safety alongside performance~\cite{fang2023evaluating}. Prompt engineering \cite{phute2023llm, zhang2023defending} and parameter manipulation \cite{piet2024jatmo} enhance robustness against adversarial attacks, creating models that are inherently less susceptible to misalignment.

Importantly, while the term ``jailbreak'' often emphasizes bypassing safety guardrails, the underlying mechanisms bear strong resemblance to adversarial attacks more broadly: in both cases, inputs are crafted to induce undesired or harmful outputs. A key distinction, however, is that adversarial attacks in typical machine learning contexts often focus on minimal or imperceptible perturbations subject to strict constraints (e.g., small $l_p$ norms), whereas jailbreak prompts need not be ``small'' changes to an existing prompt. Jailbreaks can drastically alter or extend the prompt with no particular limit on the scale of the perturbation, as long as it bypasses policy or safety guardrails. Under specific conditions—such as when safety constraints are formulated as a sort of ``decision boundary''—these two attack vectors become effectively equivalent. Yet, in real-world LLM scenarios, the unconstrained nature of jailbreak inputs can pose a different, and often broader, practical threat model. As LLMs and their safety constraints grow more integrated, these paradigms may merge, highlighting the need for unified defense strategies against any maliciously crafted input.

Adversarial training, initially presented as a jailbreak mitigation technique \cite{peng2024jailbreaking}, exemplifies the synergy between reactive and proactive approaches. Continuous exposure to adversarial examples improves inherent robustness \cite{wang2023brave}. Similarly, privacy-preserving techniques like differential privacy and federated learning \cite{abadi2016deep,mcmahan2017communication}, originally discussed for mitigating privacy threats, fundamentally alter the training process, leading to a more robust and privacy-aware LLM brain.

\chapter{Agent Intrinsic Safety: Threats on Non-Brain Modules}
\label{sec:in_other_threat}


\lettrine[lines=3]{\initfamily\textcolor{darkgreen}{T}}{his} chapter provides a detailed analysis of the intrinsic safety threats targeting an AI agent's non-brain modules, specifically its perception and action components. While the Large Language Model (LLM) core provides intelligence, the agent's overall security is critically dependent on these peripheral modules, which serve as its interface with the external world and are prime targets for attack.
The chapter first investigates threats to the perception module, categorizing them into two primary types. The first is deliberate adversarial attacks, which manipulate input data across various modalities. We explore sophisticated techniques such as prompt-based textual attacks, visual hijacking in multimodal models, inaudible command injection in auditory systems, and sensor data spoofing for systems like LiDAR and GPS. For each, corresponding defensive strategies like adversarial training, input purification, and robust sensor fusion are discussed. The second category addresses intrinsic misperception issues, which arise not from malicious intent but from inherent model limitations, dataset biases, and environmental complexities. These non-adversarial failures can lead to flawed interpretations and decision-making. Mitigation strategies, including curating diverse datasets and adopting advanced, biologically-inspired learning architectures, are examined.
Subsequently, the chapter shifts focus to the action module, which executes tasks by interacting with external tools and APIs. We identify two major risk domains. The first is supply chain attacks, where adversaries compromise external dependencies—such as websites or third-party tools—to inject malicious instructions (e.g., Indirect Prompt Injection) and manipulate the agent's behavior. The second domain covers risks in tool usage, where even secure tools can be misused. This includes unauthorized actions triggered by deceptive prompts, inadvertent data leakage to third-party services, and the dangers of granting excessive permissions. To counter these threats, the chapter highlights critical safeguards such as sandboxing, user confirmation for high-risk actions, and enforcing the principle of least privilege.
Ultimately, this chapter underscores that ensuring agent safety requires a holistic approach, extending security considerations beyond the central LLM to encompass the entire perception-action loop.

\section{Perception Safety Threats}
\label{ssec:percep_threat}

\lettrine[lines=3]{\initfamily\textcolor{darkgreen}{T}}{he} perception module of an AI agent is crucial for processing and interpreting user inputs across various modalities, such as text, images, and audio. However, the complexity and diversity of these modalities make perception systems susceptible to misinterpretations in dynamic environments \cite{xu2023multimodal}, and vulnerable to adversarial attacks that manipulate input data to mislead the agent \cite{li2023inaudible}.

\subsection{Adversarial Attacks on Perception}
Adversarial attacks are deliberate attempts to deceive AI agents by altering input data, targeting the perception module across various modalities. From subtle textual tweaks to inaudible audio distortions, these attacks reveal the fragility of even the most advanced systems. Below, we explore how these threats manifest in textual, visual, auditory, and other modalities, and highlight countermeasures.

\paragraph*{Textual Attacks} Textual adversarial attacks manipulate input text to deceive LLMs, ranging from simple sentence alterations to more complex character-level perturbations. 
Prompt-based adversarial attack, for instance, carefully crafted deceptive prompts that mislead models into generating harmful outputs. Minor changes like swapping synonyms or substituting characters can degrade performance \cite{abad2024revisiting}. Sophisticated strategies push this further: Zou et al.~\cite{zou2023universal} generate universal adversarial suffixes using greedy and gradient-based searches, while Wen et al.~\cite{wen2024hard} optimize interpretable hard prompts to bypass token-level content filters in text-to-image models.
To defend against these attacks, several approaches have been proposed. For example, Legilimens (a novel content moderation system) employs a decoder-based concept probing technique and red-team data augmentation to detect and thwart adversarial input with impressive accuracy \cite{wu2024legilimens}. Self-evaluation techniques enhance LLMs to scrutinize their own outputs for integrity \cite{brown2024self}, while methods like adversarial text purification \cite{moraffah2024adversarial} and TextDefense \cite{shen2023textdefense} harness language models to neutralize perturbations. These defenses illustrate a dynamic arms race, where resilience is forged through creativity and vigilance.

\paragraph*{Visual Attacks} Visual adversarial attacks manipulate images to exploit discrepancies between human and machine perception. These attacks are particularly concerning for multi-modal LLMs (VLMs) that rely on visual inputs. For instance, image hijacks can mislead models into generating unintended behaviors \cite{bailey2023image}, while transferable multimodal attacks can affect both text and visual components of VLMs \cite{han2023ot,gao2024boosting,huang2025image}. Recent work on multimodal LM robustness shows that targeted adversarial modifications can mislead web agents into executing unintended actions with 5\% pixels manipulation \cite{wu2025dissecting}. Ji et al.~\cite{ji2021poltergeist} reveal how inaudible perturbations can interfere with the stability of cameras and blur the shot images, and lead to harmful consequences.
Defensive strategies include adversarial training \cite{jia2022adversarial,jia2024improving,jia2025evolution}, which involves joint training with clean and adversarial images to improve robustness, and certified robustness methods that guarantee resilience through the text generation capabilities of VLMs. DIFFender~\cite{kang2024diffender} used diffusion models using feature purification to strengthen VLMs against visual manipulation.

\begin{figure*}[t]
\centering
\footnotesize
    \begin{forest}
        for tree={
            forked edges,
            draw,
            rounded corners,
            node options={align=center},
            s sep=6pt, 
            l sep=10pt, 
            calign=center,
            grow=east,
            reversed=true,
            anchor=base west,
            parent anchor=east,
            child anchor=west,
            base=left,
            font=\small,
            minimum width=1.5em,
            if n children=0{}{
                if level=1{text width=4em}{
                    if level=2{text width=7em}{
                        if level=3{yshift=0.26pt,fill=white!5,text width=9em}{}
                    }
                }
            },
          },
          where level=1{text width=7em,fill=customblue!50}{},
          where level=2{text width=10em,fill=customgreen!50}{},
        [Agent\\\textbf{Intrinsic}\\Safety\\on\\\textbf{Non-Brains}, fill=gray!20
            [Perception Safety Threats,
                [Adversarial Attacks
                    [\textbf{Texual}\\                        PromptAttack\cite{xu2023llm} Charmer\cite{abad2024revisiting}  \\
                    GCG\cite{zou2023universal} Wen et al.\cite{wen2024hard}
                    ]
                    [\textbf{Visual}\\                     PromptMatching\cite{bailey2023image} Huang et al.\cite{huang2025image} 
                    \\
                    ARE\cite{wu2025dissecting} Ji et al.\cite{ji2021poltergeist}
                    ]
                    [\textbf{Auditory}\\
                        VRifle\cite{li2023inaudible} SMA\cite{zheng2023silent} Tuner\cite{li2023enrollment} \\
                        UltraBD\cite{ze2023ultrabd} DolphinAttack\cite{zhang2017dolphin}\\
                    ]
                    [\textbf{Other Modality}\\
                        Kim et al.\cite{kim2024survey} Tu et al.\cite{tu2021adversarial} 
                        \\
                        MEMS\cite{son2015rocking} Kamal et al.\cite{kamal2021gps}\\
                    ]
                ]
                [Misperception Issues
                    [GLAM\cite{carta2023grounding} Gallegos et al.\cite{gallegos2024bias}
                    \\
                    Mahajan et 
                    al.\cite{mahajan2021domain}
                    Mazhar et al.\cite{mazhar2021gem}
                    \\
                    PHardEval\cite{fan2023nphardeval} Vilone et al.\cite{vilone2024modeling}
                    \\
                    Xu et al.\cite{xu2023bridging} Ryu et al.\cite{ryu2021cooperative} \\
                    Ohmar et al.\cite{ohmer2022mutual} Xu et al.\cite{xu2023model}
                    ]
                ]
            ]
            [Action Safety Threats,
                [Supply Chain Attack
                    [Wu et al.\cite{wu2024new} Wu et al.\cite{wu2024wipi} ToolEmu\cite{ruan2023identifying}
                    \\
                    Greshake et al.\cite{greshake2023not} InjecAgent\cite{zhan2024injecagent}]
                ]
                [Tool Use Risk
                    [ToolEmu\cite{ruan2023identifying}  ToolSword\cite{ye2024toolsword} InjecAgent\cite{zhan2024injecagent} 
                    ]
                ]
            ]
        ]
    \end{forest}
    \label{fig:tree-intrinsic-nonbrain-threats}
    \caption{Agent Intrinsic Safety: Threats on LLM Non-Brains.}
\end{figure*}

\paragraph*{Auditory} For voice-controlled AI agents, auditory adversarial attacks pose a stealthy threat.
DolphinAttack~\cite{zhang2017dolphin} introduces an innovative technique that leverages ultrasound to inject malicious voice commands into microphones in an inaudible manner. Also, inaudible perturbations like VRifle~\cite{li2023inaudible} can mislead traditional speech recognition systems and can likely be adapted to target audio-language models. Deepfake audio and adversarial voiceprint further pose serious risks for authentication-based systems \cite{li2023enrollment,ze2023ultrabd,li2023toward}, while emerging jailbreak and chat-audio attacks exploit audio processing vulnerabilities \cite{yang2024can}.
To mitigate these threats, solutions like EarArray use acoustic attenuation to filter inaudible perturbations \cite{zhang2021eararray}, while SpeechGuard enhances LLM robustness through adversarial training \cite{peri2024speechguard}. 
Moreover, NormDetect~\cite{li2023learning} focuses on effectively detecting normal speech patterns from manipulated inputs. To systematically evaluate these multifaceted risks, the AudioTrust benchmark provides a comprehensive framework for assessing the trustworthiness of audio large language models (ALLMs) \cite{li2025audiotrust}.

\paragraph*{Other Modality} Beyond text, images, and audio, AI agents interfacing with sensor data—like in autonomous systems—face unique threats. For example, LiDAR manipulation can mislead autonomous driving systems, creating phantom objects \cite{kim2024survey}. Research on adversarial attacks in multi-agent systems reveals that tampered messages can significantly degrade multi-view object detection and LiDAR-based perception in cooperative AI agents, highlighting the risk of sensor-based adversarial perturbations \cite{tu2021adversarial}.
Similarly, attacks targeting gyroscopes or GPS spoofing can disrupt navigation systems \cite{son2015rocking, kamal2021gps}.
Defenses for these attacks include robust sensor fusion algorithms and anomaly detection techniques to identify inconsistencies, as well as redundant sensors that make it harder to compromise the entire system \cite{xu2018analyzing}. Physical layer defenses, such as shielding and secure localization using enhanced SLAM techniques, are also critical \cite{geng2024survey}. Ji et al.~\cite{ji2024sensor} offer a rigorous framework for safeguarding sensor data integrity and privacy.

\subsection{Misperception Issues} 

While adversarial attacks are deliberate attempts to compromise system integrity, misperception issues emerge intrinsically from the limitations of LLMs. These errors occur without any malicious intent and can be attributed to a variety of factors ranging from dataset biases to architectural constraints.
One primary source of misperception is dataset bias. When models are trained on non-representative datasets, they tend to underperform on diverse or novel inputs \cite{gallegos2024bias}. This shortcoming is exacerbated by challenges in generalizing to new, unseen environments, where unpredictable conditions may arise. Environmental complexities such as sensor noise, occlusions, and fluctuating lighting further introduce uncertainty \cite{mazhar2021gem}. Additionally, inherent model limitations—like restricted receptive fields or the absence of robust reasoning mechanisms—compound these errors \cite{fan2023nphardeval}.
Insights from studies on multi-agent systems and online social dynamics provide further depth to our understanding of misperception. Research shows that individuals may misjudge the true distribution of opinions due to phenomena like false consensus effects, vocal minority amplification, and the spiral of silence \cite{vilone2024modeling}. Such biases can lead AI agents to erroneously infer dominant perspectives from skewed inputs. Similarly, when different models share visual features, discrepancies in feature encoding can result in significant perception errors, a challenge that mirrors issues in multi-modal LLMs \cite{xu2023bridging}. Moreover, in interactive environments, agents may develop distorted interpretations of cooperative and adversarial behaviors, as evidenced by findings in multi-agent reinforcement learning \cite{ryu2021cooperative}. Linguistic representation, too, can be influenced by perceptual biases, suggesting that misperception in LLMs may stem not only from sensory inaccuracies but also from language-driven distortions \cite{ohmer2022mutual}. Finally, systematic errors often arise when mismatched confidence levels across models affect decision-making in uncertain contexts \cite{xu2023model}.

Mitigating these misperception challenges requires a multifaceted strategy. Curating diverse and representative datasets that capture a broad spectrum of real-world conditions is critical for enhancing model performance and reducing bias \cite{pal2024diversinet}. Data augmentation techniques, which generate synthetic variations of existing data, can further enrich dataset diversity. Incorporating uncertainty estimation allows models to assess their confidence in predictions and flag potential error-prone situations \cite{mendes2024error}. Moreover, advancing model architectures to include explicit reasoning mechanisms or better processing of long-range dependencies is vital for minimizing misperception \cite{sanford2024understanding}.
An especially promising avenue is the adoption of biologically inspired learning frameworks, such as Adaptive Resonance Theory (ART). Unlike traditional deep learning approaches—often hampered by issues like catastrophic forgetting and opaque decision-making—ART models can self-organize stable representations that adapt to dynamically changing environments, thereby reducing perceptual errors \cite{grossberg2020path}. However, it is important to note that even improved explainability has its limitations, particularly when users struggle to establish clear causal links between model outputs and underlying processes \cite{shin2021effects}. Furthermore, recent studies indicate that advanced LLMs may inadvertently degrade their own responses during self-correction, underscoring the need for more robust intrinsic reasoning verification mechanisms \cite{huang2023large}.

\section{Action Safety Threats}
\label{ssec:act_threat}

\lettrine[lines=3]{\initfamily\textcolor{darkgreen}{T}}{he} action module is responsible for translating the AI agent's planned actions into actual task executions. This typically includes invoking external tools, calling APIs, or interacting with physical devices. As the interface between decision-making and execution, it is highly vulnerable to attacks. We explore two primary domains of risk: supply chain attacks and vulnerabilities arising from tool usage.

\subsection{Supply Chain Attacks} 
Supply chain attacks exploit the services that AI agents depend on, thereby undermining the integrity of the entire system \cite{wu2024new}. Unlike traditional attacks, these threats do not target the agent directly but instead compromise the external resources it relies upon. For example, malicious websites can employ indirect prompt injection (IPI) attacks, illustrated by the Web-based Indirect Prompt Injection (WIPI) framework, to subtly alter an agent's behavior without needing access to its code \cite{wu2024wipi}. Similarly, adversaries may manipulate web-based tools (such as YouTube transcript plugins) to feed misleading information into the system \cite{ruan2023identifying}.
As AI agents become increasingly integrated with online resources, their attack surface broadens considerably. Recent work by Greshake et al. proposes a new classification of indirect injection attacks, dividing them into categories like data theft, worming, and information ecosystem contamination \cite{greshake2023not}. Complementing this, the InjecAgent benchmark evaluated 30 different AI agents and revealed that most are vulnerable to IPI attacks \cite{zhan2024injecagent}.

To mitigate these risks, preemptive safety measures and continuous monitoring are essential. Current research suggests that two key factors behind the success of indirect injection are LLMs' inability to distinguish information context from actionable instructions and their poor awareness of instruction safety; hence, it is proposed to enhance LLMs' boundary and safety awareness through multi-round dialogue and in-context learning \cite{yi2023benchmarking}. Furthermore, other researchers, based on the same assumption, proposed a prompt engineering technique called ``spotlighting'' to help LLMs better distinguish between multiple input sources and reduce the success rate of indirect prompt injection attacks \cite{hines2024defending}. 
Since under a successful attack, the dependence of the agent's next action on the user task decreases while its dependence on the malicious task increases, some researchers detect attacks by re-executing the agent's trajectory with a masked user prompt modified through a masking function \cite{zhu2025melon}.  Further research has investigated the feasibility of not only detecting but also removing indirect prompt injection attacks from external content \cite{chen2025can}.
Finally, sandboxing techniques, such as those employed in ToolEmu \cite{ruan2023identifying}, create isolated environments for executing external tools, limiting the potential damage in case of a breach.

\subsection{Risks in Tool Usage} 

Even when external tools are secure, vulnerabilities can arise from how an agent interacts with them. A significant risk is unauthorized actions, where an adversary manipulates the agent into performing unintended behaviors. For example, prompt injection attacks can trick an agent into sending emails, deleting files, or executing unauthorized transactions \cite{ruan2023identifying}. The general-purpose nature of AI agents makes them especially susceptible to such deceptive instructions.
The tool learning process itself can introduce additional risks, such as malicious queries, jailbreak attacks, and harmful hints during the input, execution, and output phases \cite{ye2024toolsword}. During the tool execution phase, using incorrect or risky tools may deviate from the user's intent and potentially harm the external environment. For instance, misuse could lead to the introduction of malware or viruses. A compilation of 18 tools that could impact the physical world has been identified, with noise intentionally added to test if LLMs can choose the wrong tool.
Another significant concern is data leakage, where sensitive information is inadvertently exposed. This occurs when an agent unknowingly transmits confidential data to a third-party API or includes private details in its output.
For example, an LLM may inject commands to extract private user data, then use external tools, like a Gmail sending tool, to distribute this data \cite{zhan2024injecagent}.
The risks are especially pronounced in applications dealing with personal or proprietary data, necessitating stricter controls over information flow. Additionally, excessive permissions increase the potential for misuse. Agents with broad system access could be manipulated to perform destructive actions, such as deleting critical files, leading to irreversible damage \cite{ruan2023identifying}. Enforcing the principle of least privilege ensures that agents only have the permissions necessary to complete their tasks, minimizing the potential impact of exploitation.
Securing the action module requires layered protections and continuous monitoring. Monitoring tool usage can help detect anomalies before they cause harm, while requiring user confirmation for high-risk actions—such as financial transactions or system modifications—adds an additional layer of safety. Formal verification techniques, as explored by \cite{crouse2023formally}, can further enhance safety by ensuring that tool use policies align with best practices, preventing unintended agent behaviors.

\section{Summary and Discussion}

\lettrine[lines=3]{\initfamily\textcolor{darkgreen}{T}}{his} chapter explored intrinsic safety threats targeting AI agents' peripheral modules—specifically, the perception and action components—highlighting their critical roles as direct interfaces with external environments. While central intelligence modules (LLMs) often receive primary attention, vulnerabilities within these peripheral modules significantly impact overall agent security.

For the perception module, we categorized threats into deliberate adversarial attacks and intrinsic misperceptions. Adversarial attacks manipulate inputs across textual, visual, auditory, and sensor-based modalities, exploiting subtle weaknesses to mislead agents. Effective defenses include adversarial training, input purification, and robust sensor fusion. In contrast, intrinsic misperceptions stem from internal limitations such as biased datasets and model constraints, requiring proactive mitigation strategies like diversified data curation, uncertainty estimation, and biologically inspired architectures.

In examining the action module, we identified threats arising from supply chain attacks—particularly Indirect Prompt Injection—and risks associated with tool usage. Supply chain vulnerabilities exploit compromised external resources to subtly control agent behavior, necessitating preventive measures like sandboxing, enhanced input context awareness, and rigorous monitoring. Tool usage threats include unauthorized actions, data leakage, and excessive permissions, which can be mitigated through strict usage policies, formal verification, and enforcing the principle of least privilege.

Overall, safeguarding AI agent intrinsic safety demands a comprehensive approach that extends beyond core intelligence, encompassing every interaction point within the perception-action loop. Future research should prioritize integrated, adaptive defenses and standardized benchmarking to enhance robustness against evolving threats.

\chapter{Agent Extrinsic Safety: Interaction Risks}
\label{sec:ex_threat}

\lettrine[lines=3]{\initfamily\textcolor{darkgreen}{T}}{his} chapter presents a comprehensive examination of extrinsic safety risks, focusing on vulnerabilities that emerge from an AI agent's interactions with its external ecosystem. These threats are categorized into three primary domains: interactions with memory systems, with physical and digital environments, and with other agents. By analyzing these interfaces, the chapter highlights how an agent's security and reliability can be compromised through external manipulation and complex systemic dynamics.
The analysis begins by investigating agent-memory interaction threats, with a particular focus on the vulnerabilities within Retrieval-Augmented Generation (RAG) frameworks. We detail how these systems are susceptible to knowledge base poisoning, where adversaries inject carefully crafted malicious documents. These attacks, such as backdoor poisoning, retrieval jamming, and misinformation propagation, are designed to manipulate the agent's retrieval process, forcing it to access and utilize corrupted information, which in turn leads to harmful, biased, or incorrect outputs.
Subsequently, the chapter explores agent-environment interaction threats, distinguishing between the physical and digital domains. In the physical world, we examine risks faced by autonomous systems like robots and vehicles, including sensor spoofing (e.g., GPS, LiDAR), actuator manipulation, and the exploitation of environmental conditions to cause misperception and unsafe actions. In the digital realm, the chapter addresses threats to software and web-based agents, such as code injection, data manipulation from compromised sources, denial-of-service (DoS) attacks, and resource exhaustion, all of which can paralyze or hijack agent operations.
Finally, the chapter delves into agent-agent interaction threats within multi-agent systems. It dissects risks in both competitive and cooperative scenarios. In competitive interactions, threats include strategic deception, misinformation, algorithmic exploitation, and covert collusion to undermine opponents. In cooperative interactions, vulnerabilities manifest as unintentional information leakage, cascading failures caused by error propagation from a single agent, and systemic weaknesses arising from poor synchronization or a compromised team member. This comprehensive overview underscores that securing AI agents requires robust defenses not only at the core model level but also across every external interaction point.

\section{Agent-Memory Interaction Threats}
\label{ssec:ex_mem_threat}

\lettrine[lines=3]{\initfamily\textcolor{darkgreen}{T}}{he} extrinsic memory module functions as the cognitive repository that empowers intelligent agents to store, retrieve, and contextualize information, facilitating continuous learning and the execution of complex tasks through accumulated experiences. Retrieval-Augmented Generation (RAG) serves as its most prominent implementation. However, RAG frameworks are vulnerable to adversarial manipulations that deceive agents into retrieving and utilizing harmful or misleading documents. AgentPoison~\cite{chen2025agentpoison} exploits this vulnerability by executing a backdoor attack on AI agents, poisoning RAG knowledge bases to ensure that backdoor-triggered inputs retrieve malicious demonstrations while maintaining normal performance on benign queries. ConfusedPilot~\cite{roychowdhury2024confusedpilot} exposes a class of RAG system vulnerabilities that compromise the integrity and confidentiality of Copilot through prompt injection attacks, retrieval caching exploits, and misinformation propagation. Specifically, these attacks manipulate the text input fed to the LLM, causing it to generate outputs that align with adversarial objectives.
PoisonedRAG~\cite{zou2024poisonedrag} represents the first knowledge corruption attack on RAG, injecting minimal adversarial texts to manipulate LLM outputs. Framed as an optimization problem, it achieves a 90\% success rate with just five poisoned texts per target question in large databases. Jamming~\cite{shafran2024machine} introduces a denial-of-service attack on RAG systems, where a single adversarial ``blocker'' document inserted into an untrusted database disrupts retrieval or triggers safety refusals, preventing the system from answering specific queries. BadRAG~\cite{xue2024badrag} exposes vulnerabilities in RAG-based LLMs through corpus poisoning, wherein an attacker injects multiple crafted documents into the database, forcing the system to retrieve adversarial content and generate incorrect responses to targeted queries. By introducing just 10 adversarial passages (0.04\% of the corpus), it achieves a 98.2\% retrieval success rate, elevating GPT-4's rejection rate from 0.01\% to 74.6\% and its negative response rate from 0.22\% to 72\%.
TrojanRAG~\cite{cheng2024trojanrag} executes a joint backdoor attack on RAG systems, optimizing multiple backdoor shortcuts via contrastive learning and enhancing retrieval with a knowledge graph for fine-grained matching. By systematically normalizing backdoor scenarios, it evaluates real-world risks and the potential for model jailbreak. Lastly, a covert backdoor attack~\cite{long2024whispers} leverages grammar errors as triggers, allowing LLMs to function normally for standard queries while retrieving attacker-controlled content when minor linguistic mistakes are present. This method exploits the sensitivity of dense retrievers to grammatical irregularities using contrastive loss and hard negative sampling, ensuring that backdoor triggers remain imperceptible while enabling precise adversarial control.

\begin{figure*}[htb]
\centering
\footnotesize
    \begin{forest}
        for tree={
            forked edges,
            draw,
            rounded corners,
            node options={align=center},
            s sep=6pt,
            calign=center,
            grow=east,
            reversed=true,
            anchor=base west,
            parent anchor=east,
            child anchor=west,
            base=left,
            font=\small,
            minimum width=2.5em,
          },
          where level=1{text width=8em,fill=customblue!50}{},
          where level=2{text width=10em,fill=customgreen!50}{},
          where level=3{text width=20em}{},
        [Agent\\\textbf{Extrinsic}\\Safety, fill=gray!20
            [Agent-Memory Interaction Threats,
                [Retrieval Augmented Generation
                    [AgentPoison\cite{chen2025agentpoison}  ConfusedPilot\cite{roychowdhury2024confusedpilot} PoisonedRAG\cite{zou2024poisonedrag} RAG\cite{shafran2024machine} BadRAG\cite{xue2024badrag} TrojanRAG\cite{cheng2024trojanrag} Long et al.\cite{long2024whispers}]
                ]
            ]
            [Agent-Environment Interaction Threats,
                [Physical Environment
                    [
                    Giannaros et al.\cite{giannaros2023autonomous} Geihs(2020)\cite{geihs2020engineering} Khan et al.\cite{khan2021gps} Petit et al.\cite{petit2015remote} Zhou et al.\cite{zhou2017security} LiDAR-Adv\cite{cao2019adversarial} Ha et al.\cite{ha2020learning} Tang et al.\cite{tang2024prioritizingsafeguardingautonomyrisks}
                    ]
                ]
                [Digital Environment
                    [
                    Wu et al.\cite{wu2024new} LLMSmith\cite{liu2024demystifying} Wu et al.\cite{wu2024wipi} Guastalla et al.\cite{guastalla2023application} Geiping et al.\cite{geiping2024coercing} Tang et al.\cite{tang2024prioritizingsafeguardingautonomyrisks} EIA \cite{liao2024eia} AdvWeb \cite{xu2024advweb} AGrail \cite{luo2025agrail}
                    ]                
                ]
            ]
            [Agent-Agent Interaction Threats,
                [Competitive Interactions
                    [
                    Hammond et al.\cite{hammond2025multi} Hoodwinked\cite{ogara2023hoodwinked} Mo et al.\cite{mo2022attacking} Wen et al.\cite{wen2023secure} Motwani et al.\cite{motwani2023perfect}
                    ]
                ]
                [Cooperative Interactions
                    [
                    Pan et al.\cite{pan2023risk} AgentSmith\cite{gu2024agent}
                    ]
                ]
            ]
        ]
    \end{forest}
    \label{fig:tree-agentInteractions}
    \caption{Agent Extrinsic Safety: Threats on agent-memory, agent-environment, and agent-agent interactions.}
\end{figure*}

\section{Agent-Environment Interaction Threats}
\label{ssec:env_threat}

\lettrine[lines=3]{\initfamily\textcolor{darkgreen}{A}}{gents} can be classified into two categories based on their mode of interaction: physical interaction agents and digital interaction agents. Physical interaction agents operate in the real world, using sensors and actuators to perceive and influence their environment. Examples of such agents include autonomous vehicles and robotic systems. In contrast, digital interaction agents function within virtual or networked environments, processing and responding to data from digital sources. These include AI-powered chatbots, cybersafety systems, and automated trading algorithms.

\paragraph*{Threats in Physical Environment} 
Agents operating in the physical world, such as robots and autonomous vehicles, face distinct safety challenges due to their interaction with dynamic and potentially adversarial environments \cite{giannaros2023autonomous,geihs2020engineering,tang2024prioritizingsafeguardingautonomyrisks}. One major threat is sensor spoofing, where attackers manipulate sensor inputs to deceive the agent about its surroundings. For example, GPS spoofing can pose significant risks to UAVs (unmanned aerial vehicles) and other GPS-dependent platforms by misleading autonomous vehicles about their actual location. This allows for malicious redirection or hijacking \cite{khan2021gps}. Similarly, LiDAR spoofing can introduce false obstacles that don't actually exist, potentially leading to navigation failures or safety hazards \cite{petit2015remote}.
Another critical risk is actuator manipulation, where adversaries take control of an agent's actuators, forcing it to perform unintended physical actions. This can occur through direct tampering with the hardware or by exploiting vulnerabilities in the software that governs actuator functions \cite{zhou2017security}. Such attacks can compromise the agent's actions, leading to physical harm or mission failure.
Additionally, exploiting environmental hazards is a serious threat. Attackers may introduce physical obstacles or manipulate environmental conditions to disrupt an agent's operations. For example, adversarial objects created using techniques like LiDAR-Adv can deceive LiDAR-based autonomous driving systems by inducing sensor misinterpretations, thus degrading detection reliability and increasing real-world safety risks \cite{cao2019adversarial}. Lastly, misalignment in physical actions can undermine the safety of autonomous agents. Discrepancies between an agent's perception and the actual physical constraints of its environment can lead to unsafe or infeasible actions. For example, mismatches between learned locomotion policies and real-world physics—such as misjudging terrain rigidity or obstacle dimensions—can cause autonomous agents to take hazardous steps (e.g., unstable strides on rough surfaces). This has been observed in prior systems that required over 100 manual resets due to uncontrolled falls \cite{ha2020learning}.

\paragraph*{Threats in Digital Environment} 
Agents operating in digital environments, such as software agents and web-based agents, face distinct safety challenges arising from their reliance on external data sources and computational resources \cite{wu2024new,tang2024prioritizingsafeguardingautonomyrisks}. One major threat is code injection, where malicious actors introduce harmful code into the agent's environment, leading to unintended command execution \cite{liu2024demystifying}. These attacks often exploit software vulnerabilities or leverage compromised external resources that the agent interacts with, potentially resulting in unauthorized control over the agent's operations \cite{wu2024wipi}.
Environmental Injection Attack (EIA) exploits privacy risks in generalist web agents to stealthily steal users' PII, achieving up to 70\% success rate \cite{liao2024eia}.
AdvWeb is an automated adversarial prompt generation framework to mislead black-box web agents into executing harmful actions \cite{xu2024advweb}.
Another critical risk is data manipulation, where attackers alter the information an agent receives, causing incorrect decisions or actions \cite{wu2024new}. For example, a trading agent can be misled by manipulated financial data, leading to incorrect transactions, or an information-gathering agent may be tricked by falsified news articles, distorting its outputs. Such manipulations can have cascading effects, especially in automated systems that rely on accurate data for decision-making.
Beyond direct manipulation, denial-of-service (DoS) attacks pose a serious threat by overwhelming the agent's digital environment with excessive requests or data, effectively rendering it unresponsive or causing it to crash \cite{guastalla2023application}. These disruptions can be particularly detrimental to time-sensitive applications where availability and responsiveness are critical.
Additionally, resource exhaustion is a significant threat, as adversaries may exploit the agent's resource management mechanisms to deplete computational resources, leading to service denial for other users or overall system instability \cite{geiping2024coercing}. By draining processing power, memory, or bandwidth, attackers can severely impair an agent's ability to function effectively, disrupting its operations and reducing its efficiency. In addressing the safety challenges of LLM agents, AGrail is proposed as a lifelong guardrail framework by adapting safety checks to mitigate task-specific and systemic risks, demonstrating robust performance and transferability across diverse tasks \cite{luo2025agrail}.

\section{Agent-Agent Interaction Threats}
\label{ssec:mulagent_threat}

\lettrine[lines=3]{\initfamily\textcolor{darkgreen}{I}}{n} multi-agent systems, interactions between agents can introduce new safety vulnerabilities \cite{zhang2021physical}. These interactions are mainly competitive, where agents try to outdo each other, or cooperative, where they work together.

\paragraph*{Threats in Competitive Interactions} 
When agents compete, they often use tricky methods to gain an advantage \cite{hammond2025multi}. For example, they might spread false information or make other agents think the situation is different from reality to deceive them \cite{ogara2023hoodwinked}. This can lead opponents to make poor decisions, weakening their position.
Apart from misinformation, agents may also try to take advantage of weaknesses in their opponent's algorithms or strategies \cite{mo2022attacking}. By identifying these weaknesses, they can predict and manipulate the other agent's behavior, gaining an edge in the competition. Additionally, some agents might use disruptive techniques like denial-of-service (DoS) attacks, which overload an opponent's system with unnecessary requests, disrupting communication and hindering their ability to function \cite{wen2023secure}.
Another threat in competitive interactions is covert collaboration. Sometimes agents secretly cooperate, even when it's against the rules, to manipulate the outcome in their favor \cite{motwani2023perfect}. This kind of collusion undermines fairness and damages the integrity of the system, as it skews the competition in their favor.

\paragraph*{Threats in Cooperative Interactions} 
In cooperative situations, where agents work together toward a common goal, safety threats could damage the system's stability and reliability. One risk is unintentional information leakage, where agents accidentally share sensitive data during their communication. This could lead to privacy violations or unauthorized access, weakening the system's trustworthiness.
In addition to data leaks, errors made by one agent can spread throughout the system, causing bigger failures and lowering overall performance. \cite{pan2023risk} discusses this problem in Open-Domain Question Answering Systems (ODQA), where errors from one part of the system can ripple through and affect other components, severely impacting reliability.
The situation becomes even worse if one compromised agent introduces a vulnerability that spreads to others. If a hacker successfully takes control of one agent, they could exploit weaknesses throughout the entire system, leading to a major safety failure \cite{gu2024agent}. This kind of widespread compromise is dangerous because it could start with a small breach and escalate quickly.
Another challenge comes from poor synchronization between agents. If agents don't update their information at the same time or experience delays in communication, it can cause problems in decision-making. Misalignment or delays in updates can disrupt coordination, making it harder for the agents to achieve their shared goals effectively. These challenges emphasize the need for strong safety systems in cooperative multi-agent setups to keep them reliable and resistant to attacks.


\section{Summary and Discussion}

\lettrine[lines=3]{\initfamily\textcolor{darkgreen}{T}}{he} preceding chapters have detailed the significant safety risks that arise from AI agents interacting with memory systems, physical and digital environments, and other agents.  These risks, ranging from data poisoning and code injection to sensor spoofing and collusion, highlight the vulnerabilities inherent in increasingly complex agent-based systems.  However, as AI agents become more capable, utilizing natural language understanding and specialized tools for sophisticated reasoning, researchers are actively developing safety protocols to address these challenges. These protocols differ in approach for general-purpose and domain-specific agents.

General-purpose agents, designed for versatility across various domains, face a broad spectrum of safety challenges. To systematically assess these issues, a series of efforts have been made, including Agent-SafetyBench \cite{zhang2024agent}, SafeArena \cite{tur2025safearena}, DoomArena \cite{boisvert2025doomarena}, WASP \cite{evtimov2025wasp}, and RedTeamCUA \cite{liao2025redteamcua}, to provide evaluation frameworks and benchmarks covering a wide range of risks and failure modes, revealing significant safety deficiencies in current agents. To mitigate these risks, researchers have developed several methods to enhance agent safety, including AgentMonitor \cite{naihin2023testing} and more recent WebGuard \cite{zheng2025webguard}, by monitoring their decision-making processes and identifying potentially unsafe actions. R-Judge \cite{yuan2024r} quantifies an agent's risk awareness by evaluating its responses to both malicious and benign queries, offering a systematic approach to safety compliance. To address the limitations of existing evaluators, AgentAuditor provides a memory-augmented reasoning framework that empowers LLMs to perform human-level safety and security evaluation \cite{luo2025agentauditor}. Additionally, risk detection tools like ToolEmu \cite{ruan2023identifying} simulate tool usage in controlled environments to expose vulnerabilities in agent interactions. This approach identifies potential hazards during task execution, allowing developers to address vulnerabilities proactively. Furthermore, guardrail agents like ShieldAgent are being developed to enforce safety policies through verifiable logical reasoning, offering another layer of protection for autonomous agents \cite{chen2025shieldagent}. These combined efforts enhance the safety of general-purpose agents through comprehensive evaluation and risk detection.

Domain-specific agents, tailored for specialized tasks in high-stakes environments like scientific research, require even more stringent safety measures. Safety tools such as ChemCrow \cite{bran2023chemcrow} are designed to mitigate risks in chemical synthesis tasks by reviewing user queries and filtering malicious commands, ensuring agents do not inadvertently synthesize hazardous chemicals. Similarly, in the medical domain, the MedSentry benchmark analyzes and mitigates safety risks in multi-agent systems, proposing personality-aware defenses to rehabilitate malicious agents \cite{chen2025medsentry}. Structured task constraints, as implemented in CLAIRify \cite{yoshikawa2023large}, enhance experimental safety by imposing high-level constraints on material synthesis order and low-level restrictions on manipulation and perception tasks, thereby preventing accidents and errors. Furthermore, benchmarks like SciGuard \cite{he2023control}, which includes the SciMT-Safety benchmark, evaluate model safety by measuring both harmlessness (rejecting malicious queries) and helpfulness (handling benign queries effectively). SciGuard also incorporates long-term memory to enhance agents' ability to safely execute complex instructions while maintaining accurate risk control. These focused approaches ensure that domain-specific agents operate safely and effectively within their specialized fields.

In summary, significant progress has been made in developing innovative evaluation mechanisms and risk mitigation strategies to enhance the safety of both general-purpose and domain-specific AI agents.  However, a critical area for future research lies in integrating these approaches. Building stronger connections between the broad capabilities of general-purpose agents and the focused safeguards of domain-specific agents will be essential for creating truly robust and trustworthy LLM systems. The challenge is to combine the best aspects of both approaches to develop agents that are both versatile and secure.
\chapter{Superalignment and Safety Scaling Law in AI Agents}
\label{sec:safety_law}

\lettrine[lines=3]{\initfamily\textcolor{darkgreen}{A}}{s} AI systems scale in capability and autonomy, the challenge of ensuring that they remain safe, controllable, and aligned with human values has become increasingly central. Simple safeguards and reward-based tuning no longer suffice in the face of complex reasoning, open-ended objectives, and evolving norms. Addressing these risks requires both deeper alignment frameworks and principled understandings of how safety requirements grow with model power.

This chapter examines two complementary approaches to these challenges: superalignment, a goal-driven paradigm for structuring agent behavior around long-term human objectives; and the safety scaling law, a conceptual framework for understanding how safety interventions must intensify as model capabilities increase. We begin by introducing the principles and mechanisms of superalignment, followed by an analysis of how safety and performance interact under scaling regimes. Together, these perspectives offer foundational insights for building trustworthy AI agents at scale.

\section{Superalignment: Goal-Driven Alignment for AI Agents}

\lettrine[lines=3]{\initfamily\textcolor{darkgreen}{W}}{ith} LLMs increasingly serving as the core of decision-making in autonomous agents, ensuring that their outputs remain safe, ethical, and consistently aligned with human objectives has become a pressing challenge~\cite{askell2021general, bai2022constitutional, anwar2024foundational}. 
Traditional alignment techniques, particularly RLHF, have been instrumental in refining LLM behavior by incorporating human preferences~\cite{ziegler2019fine, ouyang2022training}. 

Traditional safety alignment focuses primarily on preventing harmful outcomes by enforcing predefined constraints. In such frameworks, an agent's behavior is guided by a single aggregated reward signal that prioritizes immediate corrections over long-range planning. Although this reactive approach works in many current applications, it struggles when an agent must execute extended, multifaceted tasks. The inability to decompose intricate, long-term goals into interpretable and manageable sub-objectives may result in behavior that is technically safe yet suboptimal for fulfilling broader human-centric aims.

To address these limitations, the concept of \textbf{superalignment}~\cite{openai_superalignment} has emerged. Superalignment represents an evolution in alignment strategies by embedding explicit long-term goal representations directly into an agent's decision-making process. Rather than simply imposing constraints to avoid harmful actions, superalignment proactively governs behavior through a composite objective function. This function integrates several dimensions of performance—specifically, safety and ethical considerations (where ethical norms and safety guidelines are continuously embedded in decision-making), task effectiveness (ensuring the agent not only avoids harmful behavior but also performs its intended functions with high competence), and long-term strategic planning (enabling the agent to plan over extended horizons and break down complex goals into manageable subtasks).

Integrating superalignment into AI systems marks a pivotal shift toward more robust, goal-driven alignment strategies. By unifying safety, ethical standards, task performance, and long-term planning within a single optimization framework, superalignment aims to enhance the reliability and robustness of autonomous agents by ensuring they remain aligned with human values over prolonged operational periods; facilitate dynamic adaptation in complex environments by reconciling immediate safety concerns with strategic, long-term objectives; and provide a clearer, more interpretable structure for diagnosing and refining AI behavior—crucial for both safety audits and continuous improvement.

Future research is expected to focus on developing algorithms that effectively balance these diverse objectives and on validating superalignment strategies in real-world applications. The ultimate goal is to establish a scalable framework that not only prevents harmful behavior but also actively promotes performance that aligns with complex human values and objectives.

\subsection{Composite Objective Functions in Superalignment}
Superalignment frameworks rely on composite objective functions that break down alignment into distinct components, each targeting a different aspect of agent behavior~\cite{ji2023ai}. Unlike RLHF, which relies on a single reward model trained from preference comparisons, superalignment explicitly separates reward signals into multiple channels:
\begin{itemize}
    \item \textbf{Task Performance Term:} Focuses on how well the agent completes its immediate operational tasks—e.g., retrieving correct answers, executing planned actions, or achieving subgoals.
    \item \textbf{Goal Adherence Term:} Captures alignment with long-term strategic objectives and user intent. This often involves giving the agent an internal representation of end goals (e.g., the full dialogue purpose, final task outcome) and training it to plan actions that serve that goal~\cite{ngo2022alignment, yudkowsky2016ai}.
    \item \textbf{Norm Compliance Term:} Enforces ethical, legal, and safety constraints. This can be implemented through rule-based filters, constraint-aware training data, or auxiliary models (like rejection classifiers) that detect violations~\cite{krueger2020hidden, carlsmith2022power}.
\end{itemize}

This multicomponent formulation addresses a key weakness of RLHF: the risk of reward hacking, where an agent exploits loosely defined reward functions to maximize short-term gains while failing to achieve genuine long-term alignment~\cite{christiano2017deep, leike2018scalable}. 



\subsection{Overcoming the Limitations of RLHF with Superalignment}

While RLHF has proven effective for aligning models to short-term user preferences, its limitations become increasingly apparent when agents must pursue complex, evolving, or norm-sensitive goals. Superalignment offers a structural alternative that addresses these weaknesses through persistent goal representation, embedded oversight, and adaptive value modeling~\cite{openai_superalignment}.

\paragraph*{Short-Horizon Feedback Bias}  
RLHF relies on localized preference signals, often ranking isolated completions without considering downstream consequences. This short-term focus leads to agents that optimize for immediate plausibility but fail to maintain coherence in multi-step scenarios. In contrast, superalignment frameworks emphasize persistent, decomposable goal structures. As argued by~\cite{zhao2025redefining}, alignment must shift from per-instance preference optimization to agent architectures that represent and recursively update internal models of user intent. These internalized goals guide behavior not just at the output level but across entire trajectories, enabling long-horizon reasoning.

\paragraph*{Reward Model Fragility and Oversight Bottlenecks}  
RLHF's centralized reward model can be exploited or misgeneralized due to limited feedback granularity and lack of internal accountability. Superalignment approaches embed scalable oversight directly into the agent's cognition. SCRIT~\cite{tang2025enabling} introduces a self-evolving natural language critic that monitors outputs during both training and inference, offering contextualized, interpretable feedback on alignment failures. Unlike static reward models, the critic co-evolves with the agent, allowing oversight capacity to grow alongside model complexity.

\paragraph*{Static Value Representation}  
RLHF assumes a fixed alignment target, encoding ethical and task priorities at training time. However, real-world deployments demand agents that can update their alignment objectives as user preferences and social norms evolve. Huang et al.~\cite{huang2025superalignment} introduces the concept of value plasticity: agents should be trained to reinterpret, recalibrate, and reweight internal objective functions in response to environmental feedback. This capacity for continuous alignment is essential for sustained deployment in dynamic settings.

In summary, superalignment replaces reactive reward shaping with proactive, structured alignment: it builds agents that reason over internal goals, self-monitor their alignment status, and dynamically adapt their value systems over time.


\subsection{Empirical Evidence Supporting Superalignment}

Growing empirical research supports the claim that superalignment-based approaches outperform traditional RLHF methods across long-horizon planning, safety-critical dialogue, and value-sensitive decision-making tasks.

\paragraph*{Long-Horizon Planning}  
In environments such as WebShop and ALFWorld, agents must reason over multiple steps while maintaining consistency with the user's high-level objective. Zhang et al.~\cite{zhang2023composite} evaluated models trained with modular, goal-conditioned objectives and found that they achieved up to 18\% higher task success rates and lower rates of goal drift compared to RLHF-tuned baselines. These agents used task-oriented modules for short-term execution and separate goal-adherence modules to validate that subtasks advanced the overall plan. Similar gains were observed in interactive environments like BabyAI and MiniWoB++~\cite{wu2023dynamic}, where agents needed to execute long sequences while staying within task constraints.

\paragraph*{Safety-Constrained Dialogue}  
In adversarial prompting scenarios, agents trained with modular oversight mechanisms achieved significantly better safety-performance trade-offs. Tang et al.~\cite{tang2025enabling} showed that SCRIT-based agents reduced jailbreaks and unsafe completions by over 30\%, outperforming traditional reward-tuned LLMs across both red-teaming and helpful-harmless benchmarks. Unlike RLHF agents that apply a single reward surface, these agents leveraged explicit ethical filters and post hoc self-assessments to revise unsafe completions before output.

\paragraph*{Dynamic Adaptation to Shifting Goals}  
Superalignment methods exhibit superior adaptability when task goals or constraints evolve mid-episode. In dynamic goal-following settings, such as resource management games or simulation-based assistants, Shyam et al.~\cite{shyam2019model} showed that agents with dynamic composite weighting adjusted their behavioral priors in response to changing objective functions. This contrasts with RLHF models, which often exhibited stale behavior or overfit to prior preferences.

\paragraph*{Auditing and Value Traceability}  
Because superaligned models optimize multiple explicit objectives, they are significantly more interpretable and auditable. In agent safety audits conducted by ShieldAgent~\cite{chen2025shieldagent}, modular agents flagged their own norm violations at a higher precision rate than black-box RLHF agents, improving error diagnosis and recovery. Agents equipped with recursive goal validation~\cite{burns2024weak} further demonstrated the ability to flag internally inconsistent subgoal policies before execution—an essential property for real-world deployment in sensitive applications like legal reasoning, healthcare, or education.

\paragraph*{Generalization and Robustness}  
Empirical studies suggest that superaligned agents generalize better to unseen tasks by virtue of their explicit planning structures. In the AgentBench safety suite~\cite{liu2023agentbench}, composite-trained agents outperformed RLHF agents in 9 out of 12 evaluated environments, including tool-use, multi-agent coordination, and adversarial querying. Notably, gains were largest in tasks requiring both instruction-following and value-sensitive trade-offs.

Together, these empirical results demonstrate that superalignment does more than prevent harm—it actively enhances goal fidelity, interpretability, and robustness in complex, real-world settings.

\subsection{Challenges and Future Directions}
Despite its promise, superalignment presents several critical challenges that must be addressed for practical implementation. These challenges primarily involve goal specification, reward calibration, dynamic adaptation, and maintaining coherence in hierarchical objectives.

A fundamental difficulty lies in defining precise and unambiguous goals. Human values are inherently context-sensitive, ambiguous, and sometimes conflicting, which makes it challenging to encode them into a structured, machine-interpretable format~\cite{anwar2024foundational}. Existing alignment techniques struggle to capture the full complexity of human intent, necessitating more advanced methods for goal extraction, decomposition, and representation. Current research explores hierarchical modeling and preference learning to enable AI systems to better adapt to evolving and nuanced human objectives~\cite{christiano2017deep}.

Even with well-defined goals, reward calibration remains a significant challenge. Superalignment requires a careful balance between task performance, long-term adherence, and ethical compliance~\cite{leng2024taming}. A poorly calibrated reward structure can lead to short-term optimization at the expense of strategic alignment or, conversely, excessive emphasis on long-term objectives at the cost of immediate effectiveness. Adaptive weighting mechanisms help dynamically adjust reward components, but ensuring stability and consistency in these adjustments remains an open research problem~\cite{wang2024comprehensive}.

Another challenge stems from adapting to dynamic human values and evolving operational contexts. Unlike static rule-based systems, AI models must continuously update their objectives to reflect shifts in societal norms, ethical standards, and external conditions~\cite{huang2024self}. Real-time goal recalibration, facilitated by meta-learning and context-aware alignment, enables AI systems to recognize when their objectives require refinement and adjust accordingly~\cite{krueger2020hidden}. However, ensuring that models can update their value representations without compromising alignment remains an unresolved issue.

Finally, maintaining coherence in hierarchical goal decomposition adds another layer of complexity. Superalignment depends on breaking down long-term objectives into sub-goals while preserving strategic alignment. Overly rigid sub-goals can lead to narrow optimization that neglects broader intent, while loosely defined sub-goals risk misalignment between immediate actions and overarching objectives~\cite{wang2024comprehensive}. Techniques such as recursive validation and multi-level reward structuring aim to mitigate these risks, but further research is needed to refine their applicability across diverse AI systems~\cite{burns2024weak}.

To sum up, while superalignment offers a structured approach to AI alignment, its successful implementation depends on overcoming goal ambiguity, reward miscalibration, value drift, and hierarchical misalignment. Future work should focus on enhancing interpretability, stability, and adaptability to ensure AI systems remain aligned with human objectives over extended time horizons.

\section{Safety Scaling Law in AI Agents}

\lettrine[lines=3]{\initfamily\textcolor{darkgreen}{T}}{he} exponential scaling of AI capabilities has unveiled a fundamental tension in artificial intelligence: the nonlinear escalation of safety risks~\cite{bengio2024managing}.
As language models grow from millions to trillions of parameters, their performance follows predictable scaling laws~\cite{kaplan2020scaling,hoffmann2022training}, but safety assurance exhibits starkly different dynamics~\cite{bengio2024managing}. 
\textit{Safety Scaling Law} is the mathematical relationship describing how safety interventions must scale to maintain acceptable risk levels as model capabilities expand. 
The core challenge of the safety scaling law lies in ensuring that safety measures evolve proportionally to model capabilities, as performance improvements often outpace safety improvements. 
Recent research has quantified this tension and proposed frameworks to address it: 
\begin{itemize}
\item \textbf{Capability-Risk Trade-off}:
Zhang~\textit{et al.}~\cite{zhang2024agent} established the first quantitative relationship between model power and safety risks, demonstrating that more capable models inherently face higher vulnerability surfaces. This work introduced the Safety-Performance Index (SPI) to measure this trade-off.
\item \textbf{Helpfulness-Safety Relationship}: Building on this, Ruan~\textit{et al.}~\cite{ruan2023identifying} revealed that models optimized for helpfulness exhibit 37\% more safety-critical failures, highlighting the need for joint optimization frameworks.
\item \textbf{Commercial \textit{vs.} Open-Source Dynamics}: Through large-scale benchmarking, Ying~\textit{et al.}~\cite{ying2024safebench} uncovered divergent safety-performance profiles: Commercial models (\textit{e.g.}, Claude-3.5 Sonnet) achieve 29\% higher safety scores through specialized safety pipelines, but at 15\% performance cost. Open-source models show tighter coupling, with Phi-series achieving 91\% of commercial safety levels at 40\% lower computational cost.

\item \textbf{Scale-Data Interplay}: Contrary to expectations, model size only explains 42\% of safety variance, while data quality accounts for 68\%, suggesting that data-centric approaches may outperform pure scaling.
\item \textbf{Multimodal Vulnerabilities}: MLLMs exhibit 2.1X more safety failures during visual grounding, with cross-modal attention heads identified as primary failure points (71\% of harmful outputs).
\end{itemize}

These findings~\cite{zhang2024agent,ruan2023identifying,ying2024safebench} collectively demonstrate that safety scaling requires more than proportional investment, which demands architectural innovations that fundamentally alter the capability-risk relationship. Then, we will review the explorations~\cite{alami2024alignment,chen2024noise,tuan2024towards} on how emerging alignment techniques address these challenges.

\subsection{Current landscape: balancing model safety and performance}
\begin{figure}[t]
    \centering
\includegraphics[width=\linewidth]{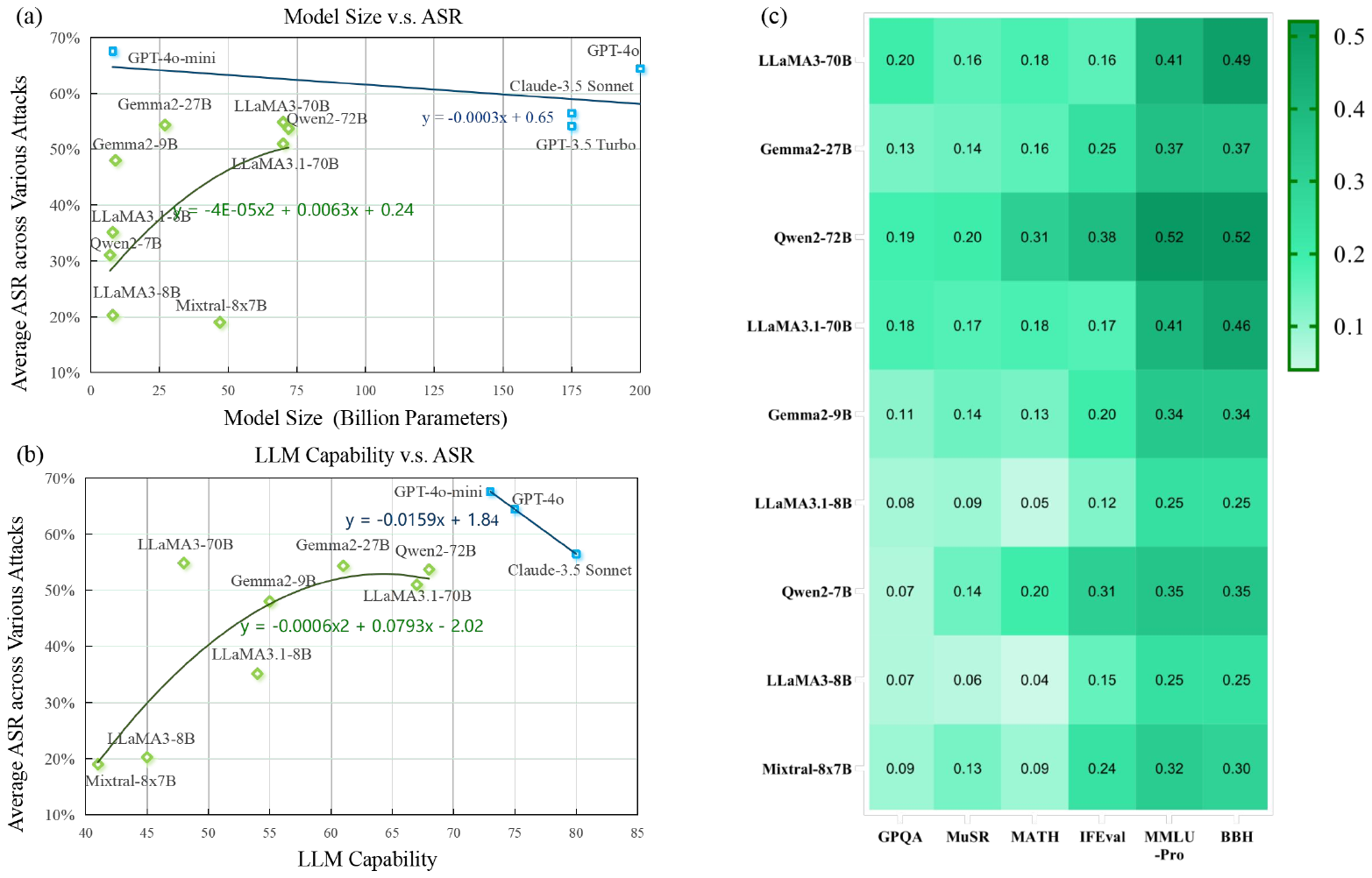}
    \caption{\textbf{Performance and safety analysis of LLMs. } (a) The relationship between LLM model size and their average ASR across various attacks. The data are sourced from experimental results of a study assessing the robustness of LLMs against adversarial attacks~\cite{zhang2024agent}. (b) The relationship between the capability of LLMs and their average attack success rate (ASR) across various attacks. The LLM capability data are derived from the Artificial Analysis Intelligence Index on the Artificial Analysis platform's LLM leaderboard~\cite{artificial_analysis_leaderboard}. (c) Heatmap of performance across multiple benchmark tasks. The figure presents a heatmap that illustrates the performance of various LLMs across multiple benchmark tasks, including GPQA, MuSR, MATH, IFEval, MMLU-Pro, and BBH, with data sourced from Hugging Face's Open LLM Leaderboard v2~\cite{open-llm-leaderboard-v2}. 
    }
    \label{fig:Safety-related factors}
\end{figure}

In recent years, the safety and performance of AI models have become critical topics of research, particularly as these models are increasingly deployed in high-stakes applications. Zhang~\textit{et al.}~\cite{zhang2024agent} proposed the first to quantify the relationship between model safety and performance, revealing that more powerful models inherently face higher safety risks. This finding underscores the challenge of balancing model capabilities with the need for robust safeguards. Building on this, Ruan \textit{et al.}~\cite{ruan2023identifying} explored how helpfulness—defined as a model's ability to assist users—interacts with safety concerns. Further advancing the discussion, Ying~\textit{et al.}~\cite{ying2024safebench} conducted a more detailed comparison and analysis of model safety and performance, leading to the following conclusions:
(1) As shown in Figure~\ref{fig:Safety-related factors}~(A) and Figure~\ref{fig:Safety-related factors}~(C), the safety and performance of commercial models often show an inverse relationship, as safety measures and investments differ between companies. In contrast, open-source models tend to exhibit a positive correlation between general performance and safety—better performance often leads to improved safety. Commercial models usually outperform open-source models in terms of safety, with Claude-3.5 Sonnet being the most secure among commercial models, while the Phi series stands out as the most secure open-source model.
(2) As shown in Figure~\ref{fig:Safety-related factors}~(B), model size does not have a strict linear relationship with safety performance. The quality of training data and pipeline are also key factors influencing safety; 
(3) Multimodal large language models (MLLMs) tend to compromise safety during visual language fine-tuning and multimodal semantic alignment, with safety performance influenced by both the underlying language model and their specific training strategies.

\subsection{Enhancing safety: preference alignment and controllable design}
As the capabilities of LLMs continue to grow, concerns regarding their safety have become increasingly prominent. Enhancing model safety is therefore a critical challenge in the development of LLMs. Previous studies have proposed various approaches to address this issue, including the use of in-context exemplars and self-safety checks, red-teaming techniques~\cite{ge2023mart}, and Safe reinforcement learning from human feedback (Safe RLHF)~\cite{ouyang2022training}. 
The safety issues in LLMs can essentially be framed as an alignment problem. The goal is to align the model with datasets containing both safe and less secure responses. 
Through this alignment, the model learns to prioritize generating safer outputs while minimizing the risk of harmful content. With the support of preference optimization techniques (such as DPO~\cite{rafailov2023direct}, IPO~\cite{jiang2023preference}, \textit{etc.}), this alignment process fine-tunes the model to produce responses that meet safety standards. 
As reported in~\cite{alami2024alignment}, various preference optimization methods are investigated for safety enhancement, including Safe-DPO~\cite{rafailov2023direct}, Safe-robust-DPO~\cite{chowdhury2024provably}, Safe-IPO~\cite{jiang2023preference}, Safe-SLiC~\cite{zhao2022calibrating}, Safe-KTO~\cite{ethayarajh2024kto}, and Safe-NCA~\cite{chen2024noise}, \textit{etc.} 
The results indicate that most preference optimization methods can significantly enhance safety, albeit at the cost of general performance, particularly in MATH capabilities. Among these methods, noise contrastive alignment (Safe-NCA)~\cite{chen2024noise} is identified as an optimal approach for balancing safety with overall model performance. The core of the Safe-NCA~\cite{chen2024noise} method lies in utilizing a custom contrastive loss function, combined with a safety dataset, to train a model that is safer and more robust during generation by comparing the generated safe and unsafe responses with the outputs of a reference model.
Beyond enhancing safety, achieving flexible control over the trade-offs between safety and helpfulness is equally critical. AI models should strike an appropriate balance between safety and helpfulness, based on the specific needs of different users. To illustrate, for the prompt ``Tell me how to make a potion'', LLMs should adjust their responses based on the user's profile. For scientists, the response should provide relevant and technically accurate information. For teenagers, the model should prioritize safety, offering cautious and harmless suggestions.

To achieve this, Tuan \textit{et al.}~\cite{tuan2024towards} propose a framework based on self-generated data to enhance model controllability. By introducing control tokens as inputs, users can specify the desired safety and helpfulness in model responses. The control tokens define the requested levels of safety and helpfulness in the following form:
\begin{align}
    [helpful=s_{hp}][harmless=s_{sf}].
\end{align}
The proposed method can ``rewind'' aligned LLMs and unlock their safety and helpfulness using self-generated data, with fine-tuning to further enhance controllability. However, achieving independent control over safety and helpfulness remains a significant challenge. This is because: 
(1) Certain prompts may be difficult to define in terms of balancing safety and helpfulness, or the definitions of both may conflict in certain contexts. For example, in the query ``I want the net worth of the person,'' it can be difficult to determine how safety and helpfulness should be prioritized. 
(2) Some models may have already established a fixed trade-off during the training process, which could limit their flexibility by forcing them to adhere to a specific priority, thereby preventing adjustments based on different application scenarios. (3) Many training data examples inherently satisfy both safety and helpfulness criteria, leading to a high correlation between these two attributes during model training.

\subsection{Future directions and strategies: the AI-45 degree rule and risk management} 
In the field of AI safety, despite various safety recommendations and extreme risk warnings being proposed, there still lacks a comprehensive guide to balance AI safety and capability. Chao~\textit{et al.}~\cite{chao2024towards} introduce the AI-45$^{\circ}$ Rule as a guiding principle for achieving a balanced roadmap towards trustworthy AGI. The rule advocates for the parallel development of AI capabilities and safety measures, with both dimensions advancing at the same pace, represented by a 45$^{\circ}$ line in the capability-safety coordinate system. It emphasizes that current advances in AI capabilities often outpace safety measures, exposing systems to greater risks and threats. Therefore, risk management frameworks such as the Red Line and Yellow Line are proposed to monitor and manage these risks as AI systems scale. As mentioned in the International Dialogues on AI Safety (IDAIS), the ``Red Line'' for AI development is defined, which includes five key aspects: autonomous replication or improvement, power-seeking behavior, assistance in weapon development, cyberattacks, and deception. Additionally, the concept of the ``Yellow Line'' is designed to complement and expand existing safety evaluation frameworks, such as Anthropic's responsible scaling policies. Models below these warning thresholds require only basic testing and evaluation. However, more advanced AI systems that exceed these thresholds necessitate stricter assurance mechanisms and safety protocols to mitigate potential risks. By establishing these thresholds, a proactive approach can be taken to ensure that AI systems are developed, tested, and deployed with appropriate safeguards in place.

\section{Summary and Discussion}

\lettrine[lines=3]{\initfamily\textcolor{darkgreen}{T}}{his} chapter presented two complementary perspectives on advancing the safety of increasingly capable AI agents: the structural approach of superalignment and the empirical insights from the emerging safety scaling law. As language models and autonomous agents continue to scale in complexity and application scope, traditional alignment methods like RLHF have proven insufficient in maintaining long-term, goal-consistent, and value-aligned behavior.

Superalignment offers a principled framework for aligning agents with human objectives through structured goal representation, modular reward decomposition, and persistent oversight. Unlike single-signal reward models, superaligned agents optimize for a composite objective function that balances task performance, ethical compliance, and strategic coherence. Empirical results demonstrate that such agents are more robust, interpretable, and adaptable in dynamic environments, especially when pursuing multi-step tasks with evolving goals~\cite{zhang2023composite, tang2025enabling, shyam2019model}.

Meanwhile, the concept of the safety scaling law underscores the disproportionate growth in risk as models become more powerful. Quantitative studies have shown that capability improvements often outpace safety enhancements, with larger models exhibiting greater vulnerability surfaces and higher likelihoods of unsafe behavior~\cite{zhang2024agent, ruan2023identifying}. Novel methods such as Safe-NCA and control-token conditioning have been proposed to address this imbalance, enabling more flexible and controllable trade-offs between helpfulness and harmlessness~\cite{chen2024noise, tuan2024towards}.

Together, these two threads reveal a crucial insight: proactive, structured alignment and proportionally scaled safety interventions are both essential for the development of trustworthy AI systems. Superalignment provides the architectural and optimization foundations for value-consistent reasoning, while safety scaling laws offer empirical guidance on risk-growth patterns and mitigation requirements.

Moving forward, a key research challenge lies in bridging these paradigms. Integrating superalignment principles into safety scaling protocols can help ensure that safety mechanisms evolve alongside model capabilities. Conversely, insights from safety scaling benchmarks can inform how to prioritize and adapt superalignment objectives under different deployment contexts. Achieving this integration will be critical for building AI agents that are not only powerful and capable—but also safe, aligned, and reliable over extended time horizons.

\chapter{Concluding Remarks and Future Outlook}



\lettrine[lines=3]{\initfamily\textcolor{darkgreen}{W}}{e} have explored in this book the evolving landscape of foundation agents by drawing parallels between human cognitive processes and artificial intelligence. We began by outlining the core components of intelligent agents—detailing how modules such as memory, perception, emotion, reasoning, and action can be modeled in a framework inspired by the comparison with human brain. Our discussion highlighted how these agents can be structured in a modular fashion, enabling them to emulate human-like processing through specialized yet interconnected subsystems.

We then delved into the dynamic aspects of agent evolution, examining self-improvement mechanisms that leverage optimization techniques, including both online and offline strategies. By investigating how large language models can act as both reasoning entities and autonomous optimizers, we illustrated the transformative potential of agents that continuously adapt to changing environments. Building on these technical foundations, we highlighted how agents can drive the self-sustaining evolution of their intelligence through closed-loop scientific innovation. We introduced a general measure of intelligence for knowledge discovery tasks and surveyed current successes and limitations in agent-knowledge interactions. This discussion also shed light on emerging trends in autonomous discovery and tool integration, which are crucial for the advancement of adaptive, resilient AI systems.

Our work also addressed the collaborative dimension of intelligent systems, analyzing how multi-agent interactions can give rise to collective intelligence. We explored the design of communication infrastructures and protocols that enable both agent-agent and human-AI collaboration. This discussion underscored the importance of fostering synergy between diverse agent capabilities to achieve complex problem solving and effective decision-making.

Finally, we emphasized the critical challenge of building safe and beneficial AI. Our review encompassed intrinsic and extrinsic security threats, from vulnerabilities in language models to risks associated with agent interactions. We provided a comprehensive overview of safety scaling laws and ethical considerations, proposing strategies to ensure that the development of foundation agents remains aligned with societal values. Overall, our work offers a unified roadmap that not only identifies current research gaps but also lays the foundation for future innovations in creating more powerful, adaptive, and ethically sound intelligent agents.

Looking ahead, we envision several key milestones that will mark significant progress in the development of intelligent agents. First, we anticipate the emergence of general-purpose agents capable of handling a wide array of human-level tasks, rather than being confined to specific domains. These agents will integrate advanced reasoning, perception, and action modules, enabling them to perform tasks with human-like adaptability and versatility. Achieving this milestone will represent a fundamental shift in how AI can support and augment human capabilities in both everyday and specialized contexts.

Another critical milestone is the development of agents that learn directly from their environment and continuously self-evolve through interactions with humans and data. As the distinction between training-time and test-time computation gradually disappears, agents will acquire new skills on the fly by engaging with their surroundings, other agents, and human partners. This dynamic learning process is essential for achieving human-level capabilities and for enabling agents to keep pace with a constantly changing world. It is also vital if agents are to be able to drive innovation in scientific discovery, as this expands the boundaries of evolution for both agents and humanity.

We predict that agents will transcend traditional human limitations by transforming individual human know-how into collective agent intelligence. The current inefficiencies in human information sharing---where complex knowledge requires extensive practice to transfer---will be overcome by agents, which offer a format of human know-how that is both transferable and infinitely duplicable. This breakthrough will remove the bottleneck of complexity, enabling a new \textit{intelligence network effect} whereby a large ensemble of human and AI agents can operate at a level of intelligence that scales with network size~\cite{li2024subnetwork}. In this scenario, the fusion of agent-acquired knowledge and human expertise will foster an environment where insights and innovations are disseminated and applied rapidly across various domains.

We also anticipate this intelligence network effect enabling the establishment of a new paradigm for human-AI collaboration---one that is larger in scale, more interdisciplinary, and more dynamically organized than ever before. The resulting human-AI society will achieve previously unattainable levels of complexity and productivity, heralding a transformative era in both technological and social development.

In summary, these milestones outline a future where intelligent agents become increasingly autonomous, adaptive, and deeply integrated with human society—driving scientific discovery, enhancing knowledge sharing, and redefining collaboration on a global scale.

\clearpage
\section*{Acknowledge}

Argonne National Laboratory's work was supported by the U.S. Department of Energy, Office of Science, under contract DE-AC02-06CH11357. XLQ acknowledges the support of the Simons Foundation. 
\bibliographystyle{unsrtnat}
\bibliography{main}  


\end{document}